\documentclass[mscthesis, 11pt, oneside, openany]{usiinfthesis}

\usepackage[utf8]{inputenc} 
\usepackage[T1]{fontenc}
\usepackage{textcomp}

\usepackage[font=small,labelfont=bf]{caption}
\usepackage{float}
\usepackage{blindtext}

\usepackage{tocloft}
\usepackage{enumitem}

\usepackage{graphicx}
\usepackage{caption}
\usepackage{subcaption}

\usepackage{url}
\makeatletter
\g@addto@macro{\UrlBreaks}{\UrlOrds}
\makeatother
\PassOptionsToPackage{hyphens}{url}
\setcitestyle{authoryear,open={[},close={]}}

\newfloat{Equation}{htbp}{equ}[chapter]
\newcommand{\listequationsname}{Equations}

\usepackage{xcolor}
\usepackage{listings,lstautogobble}
\definecolor{gray}{gray}{0.5}
\colorlet{commentcolour}{green!50!black}
\colorlet{stringcolour}{red!60!black}
\colorlet{keywordcolour}{blue}
\colorlet{exceptioncolour}{yellow!50!red}
\colorlet{commandcolour}{magenta!90!black}
\colorlet{numpycolour}{blue!60!green}
\colorlet{literatecolour}{magenta!90!black}
\colorlet{promptcolour}{green!50!black}
\colorlet{specmethodcolour}{violet}
\newcommand*{\literatecolour}{\textcolor{literatecolour}}
\newcommand*{\pythonprompt}{\textcolor{promptcolour}{{>}{>}{>}}}
\lstdefinestyle{python}{
	language=python,
	showtabs=true,
	tab=,
	tabsize=4,
	basicstyle=\ttfamily\footnotesize,
	stringstyle=\color{stringcolour},
	showstringspaces=false,
	keywordstyle=\color{keywordcolour}\bfseries,
	emph={as,and,break,class,continue,def,yield,del,elif ,else,%
		except,exec,finally,for,from,global,if,in,%
		lambda,not,or,pass,print,raise,return,try,while,assert,with},
	emphstyle=\color{blue}\bfseries,
	emph={[2]True, False, None},
	emphstyle=[3]\color{commandcolour},
	morecomment=[s]{"""}{"""},
	commentstyle=\color{commentcolour}\slshape,
	emph={array, matmul, ones, transpose, float32},
	emphstyle=[4]\color{numpycolour},
	emph={[5]assert,yield},
	emphstyle=[5]\color{keywordcolour}\bfseries,
	emph={[6]range},
	emphstyle={[6]\color{keywordcolour}\bfseries},
	literate=*%
	{:}{{\literatecolour:}}{1}%
	{=}{{\literatecolour=}}{1}%
	{-}{{\literatecolour-}}{1}%
	{+}{{\literatecolour+}}{1}%
	{*}{{\literatecolour*}}{1}%
	{**}{{\literatecolour{**}}}2%
	{/}{{\literatecolour/}}{1}%
	{//}{{\literatecolour{//}}}2%
	{!}{{\literatecolour!}}{1}%
	{<}{{\literatecolour<}}{1}%
	{>}{{\literatecolour>}}{1}%
	{>>>}{\pythonprompt}{3},
	frame=trbl,
	rulecolor=\color{black!40},
	backgroundcolor=\color{gray!5},
	breakindent=.5\textwidth,
	frame=single,
	breaklines=true,
	basicstyle=\ttfamily\footnotesize,%
	keywordstyle=\color{keywordcolour},%
	emphstyle={[7]\color{keywordcolour}},%
	emphstyle=\color{exceptioncolour},%
	literate=*%
	{:}{{\literatecolour:}}{2}%
	{=}{{\literatecolour=}}{2}%
	{-}{{\literatecolour-}}{2}%
	{+}{{\literatecolour+}}{2}%
	{*}{{\literatecolour*}}2%
	{**}{{\literatecolour{**}}}3%
	{/}{{\literatecolour/}}{2}%
	{//}{{\literatecolour{//}}}{2}%
	{!}{{\literatecolour!}}{2}%
	{<}{{\literatecolour<}}{2}%
	{<=}{{\literatecolour{<=}}}3%
	{>}{{\literatecolour>}}{2}%
	{>=}{{\literatecolour{>=}}}3%
	{==}{{\literatecolour{==}}}3%
	{!=}{{\literatecolour{!=}}}3%
	{+=}{{\literatecolour{+=}}}3%
	{-=}{{\literatecolour{-=}}}3%
	{*=}{{\literatecolour{*=}}}3%
	{/=}{{\literatecolour{/=}}}3%
}
\lstnewenvironment{python}
{\lstset{style=python}}
{}

\usepackage[titletoc,page]{appendix}
\makeatletter\@openrightfalse\makeatother

\usepackage{acro}
\DeclareAcronym{ai}{short=AI,long=Artificial Intelligence}
\DeclareAcronym{ml}{short=ML, long=Machine Learning}
\DeclareAcronym{dl}{short=DL, long=Deep Learning}
\DeclareAcronym{dnn}{short=DNN, long=Deep Neural Network}
\DeclareAcronym{rnn}{short=RNN, long=Recurrent Neural Network}
\DeclareAcronym{rl}{short=RL, long=Reinforcement Learning}
\DeclareAcronym{irl}{short=IRL, long=Inverse Reinforcement Learning}
\DeclareAcronym{nn}{short=NN, long=Neural Network}
\DeclareAcronym{cnn}{short=CNN, long=Convolutional Neural Network}
\DeclareAcronym{ann}{short=ANN, long=Artificial Neural Network}
\DeclareAcronym{il}{short=IL, long=Imitation Learning}
\DeclareAcronym{gpu}{short=GPU, long=Graphics Processing Unit}

\DeclareAcronym{idsia}{short=IDSIA, long=Istituto Dalle Molle di Studi sull’Intelligenza Artificiale}
\DeclareAcronym{epfl}{short=EPFL, long=Swiss Federal Institute of Technology in Lausanne}
\DeclareAcronym{ecal}{short=ECAL, long=Lausanne Arts School}
\DeclareAcronym{ros}{short=ROS, long=Robot Operating System}

\DeclareAcronym{ik}{short=IK, long=Inverse Kinematics}

\DeclareAcronym{3d}{short=3D, long=three-dimensional}
\DeclareAcronym{2d}{short=2D, long=two-dimensional}
\DeclareAcronym{1d}{short=1D, long=one-dimensional}

\DeclareAcronym{ev}{short=EV, long=Expected value}

\DeclareAcronym{led}{short=LED, long=Light Emitting Diode}
\DeclareAcronym{rgb}{short=RGB, long=Red Green Blue}

\DeclareAcronym{ir}{short=IR, long=Infrared}
\DeclareAcronym{gui}{short=GUI, long=Graphical User Interface}
\DeclareAcronym{mas}{short=MAS, long=Multi-Agent System}
\DeclareAcronym{dai}{short=DAI, long=Distributed Artificial Intelligence}
\DeclareAcronym{mal}{short=MAL, long=Multi-Agent Learning}
\DeclareAcronym{ma}{short=MA, long=Multi-Agent}
\DeclareAcronym{pid}{short=PID, long=Proportional Integral Derivative}
\DeclareAcronym{decpomdp}{short=Dec-POMDP, long=Decentralised partially observable Markov decision process}

\DeclareAcronym{mse}{short=MSE, long=Mean Squared Error}
\DeclareAcronym{bce}{short=BCE, long=Binary Cross Entropy}
\DeclareAcronym{ce}{short=CE, long=Cross Entropy}
\DeclareAcronym{auc}{short=AUC, long=Area Under the ROC Curve}
\DeclareAcronym{roc}{short=ROC, long=Receiver Operating Characteristic}
\DeclareAcronym{r2}{short=R$^2$, long=R Squared}

\DeclareAcronym{tpr}{short=TPR, long=True Positive Rate}
\DeclareAcronym{fpr}{short=FPR, long=False Positive Rate}

\DeclareAcronym{icc}{short=ICC, long=Instantaneous Center of Curvature}

\lstdefinelanguage{algebra}
{morekeywords={import,sort,constructors,observers,transformers,axioms,if,
else,end},
sensitive=false,
morecomment=[l]{//s},
}

\title{Simulation of robot swarms for learning\\ communication-aware coordination} %compulsory
\author{Giorgia Adorni} %compulsory
\begin{committee}
\advisor{Dr.}{Alessandro}{Giusti} %compulsory
\coadvisor{Dr.}{Jèrôme}{Guzzi}{} %optional
\end{committee}
\Day{16} %compulsory
\Month{October} %compulsory
\Year{2020} %compulsory, put only the year
\place{Lugano} %compulsory

\dedication{To everyone who taught me something} %optional
\openepigraph{``It has become appallingly obvious that our technology has 
exceeded our humanity.''}{Albert Einstein} %optional

\begin{document}
\maketitle %generates the titlepage, this is FIXED
\let\cleardoublepage\clearpage
\frontmatter %generates the frontmatter, this is FIXED
\begingroup
\begin{abstract}
%\addcontentsline{toc}{chapter}{Abstract}  % added 
In recent years, robotics research has dedicated extensive attention to cooperative multi-agent problems, in which agents have a common goal that requires them to collaborate and possibly communicate, to achieve it.
We investigate imitation learning algorithms to address this issue, providing new insights and promising solutions. 
These methods learn a controller by observing demonstrations of an expert, such as the behaviour of a centralised omniscient controller, which can perceive the entire environment, including the state and observations of all agents. 

Performing tasks with complete knowledge of the state of a system is relatively easy, but centralised solutions might not be feasible in real scenarios since agents do not have direct access to the state but only to their own observations.
%FIXME add input e output * rifrasa
%To overcome this issue, we train end-to-end \glspl{nn}, usually classifiers or 
%regressors, that only exploit local observations and communications, learning 
%decentralised solution via imitation of a centralised control, which indicates the 
%action to be performed.
To overcome this issue, we train end-to-end \acp{nn}, usually classifiers or regressors, that take as input local observations obtained from an omniscient centralised controller, in other words, the agents' sensor readings and the communications received, producing as output the action to be performed and the communication to be transmitted.

In this study, we focus on two different scenarios: distributing the robots in space such that they stand at an equal distance from each other and colouring the robots in space depending on their position with respect to the others in the group.
Both are examples of cooperative tasks based on the use of a distributed controller. While the second cannot be solved without allowing an explicit exchange of messages between the agents, in the first one, a communication protocol is unnecessary, although it may increase performance.

The experiments are run in Enki, a high-performance open-source simulator for planar robots, which provides collision detection and limited physics support for robots evolving on a flat surface. Moreover, it can simulate groups of robots hundreds of times faster than real-time.

%In addition to analysing typical supervised learning approaches, which directly 
%learn a mapping from observations to actions, we concentrate on more 
%challenging situations where the communication is not provided to the network, 
%instead it is a latent variable which has to be inferred.

The results show how applying a communication strategy improves the performance of the distributed model, letting it decide which actions to take almost as precisely and quickly as the expert controller.
\end{abstract}

\begin{acknowledgements}
%\addcontentsline{toc}{chapter}{Acknowledgements}  % added 
I would like to thank my supervisors Alessandro Giusti and Jérôme Guzzi for their 
precious advice and help, my family who have never stopped believing in me and 
to Elia who shared with me this extraordinary journey, always encouraging and 
supporting me, despite all the difficulties.

A special thank goes also to Università della Svizzera Italiana and the University of 
Milano–Bicocca for allowing me to take part in this double degree program and 
to all those people who have accompanied me and made this experience special. 

\end{acknowledgements}
\endgroup

\tableofcontents 
\newpage

\begingroup
\let\cleardoublepage\clearpage
\listoffigures %optional
\addcontentsline{toc}{chapter}{List of Figures}
\clearpage%\cleardoublepage
\listoftables %optional
\addcontentsline{toc}{chapter}{List of Tables}
%\cleardoublepage
\listof{Equation}{\listequationsname}
\addcontentsline{toc}{chapter}{List of \listequationsname}
\lstlistoflistings
\addcontentsline{toc}{chapter}{List of Listings}
\endgroup

\makeatletter
\renewcommand\mainmatter{\clearpage\@mainmattertrue\pagenumbering{arabic}}
\makeatother

\mainmatter

\begingroup
\let\cleardoublepage\clearpage
\acresetall
\chapter*{Introduction}
\addcontentsline{toc}{chapter}{Introduction}
\markboth{}{}
\label{chap:intro}
In this work, we consider cooperative multi-agent scenarios, in which multiple robots collaborate and possibly communicate to achieve a common goal \cite[][]{ismail2018survey}.
Homogeneous Multi-Agent Systems (MAS) are composed of $N$ interacting agents, which have the same physical structure and observation capabilities, so they can be considered to be interchangeable and cooperate to solve a given task \cite[][]{stone2000multiagent, vsovsic2016inverse}.
This system is characterised by a state $S$ — which can be decomposed into sets of local states for each agent and the set of possible observations $O$ for each agent — obtained through sensors and the set of possible actions  $A$ for each agent.

The objective of this study is to use \ac{il} approaches to learn decentralised solutions via imitation of an omniscient centralised control. To do so, we train end-to-end \acp{nn}, either classifiers or regressors, that only exploit local observations and communications to decide the action to be performed. This controller is the same for each agent, which means that given an identical set of observations as input, likewise, for each of these, the outputs will be equivalent \cite[][]{ross2011reduction, tolstaya2020learning}.

This research project focuses on two different multi-agent scenarios, in which $N$ robots, all oriented in the same direction, are initially randomly placed along the $x$-axis. We consider the Thymio as holonomic since their movements are limited to only one dimension. This premise simplifies our system, in which consequently we have to keep into account only geometric constraints and not kinematic ones.

In the first scenario, visualised in Figure \ref{fig:task1}, the objective is to distribute the robots in space such that they stand at equal distances from each other. To do so, each agent updates its state — its absolute position — by performing actions — moving forward and backwards along the $x$-axis — based on the observations received from the environment — the distances from neighbours.
\begin{figure}[!htb]
	\begin{center}
		\begin{subfigure}[h]{0.49\textwidth}
		\centering
		\includegraphics[width=\textwidth]{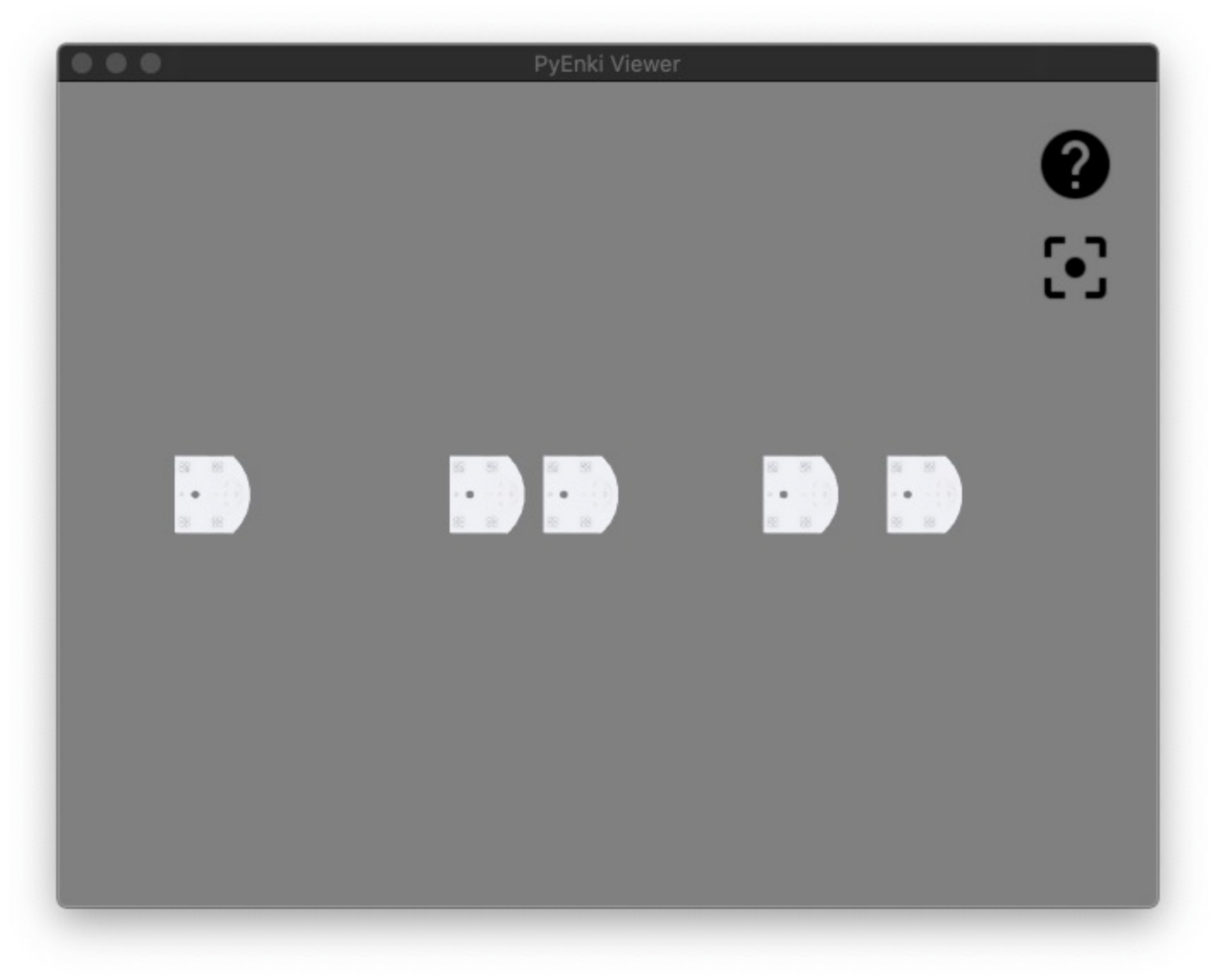}
		\end{subfigure}
		\hfill
		\begin{subfigure}[h]{0.49\textwidth}
			\centering
			\includegraphics[width=\textwidth]{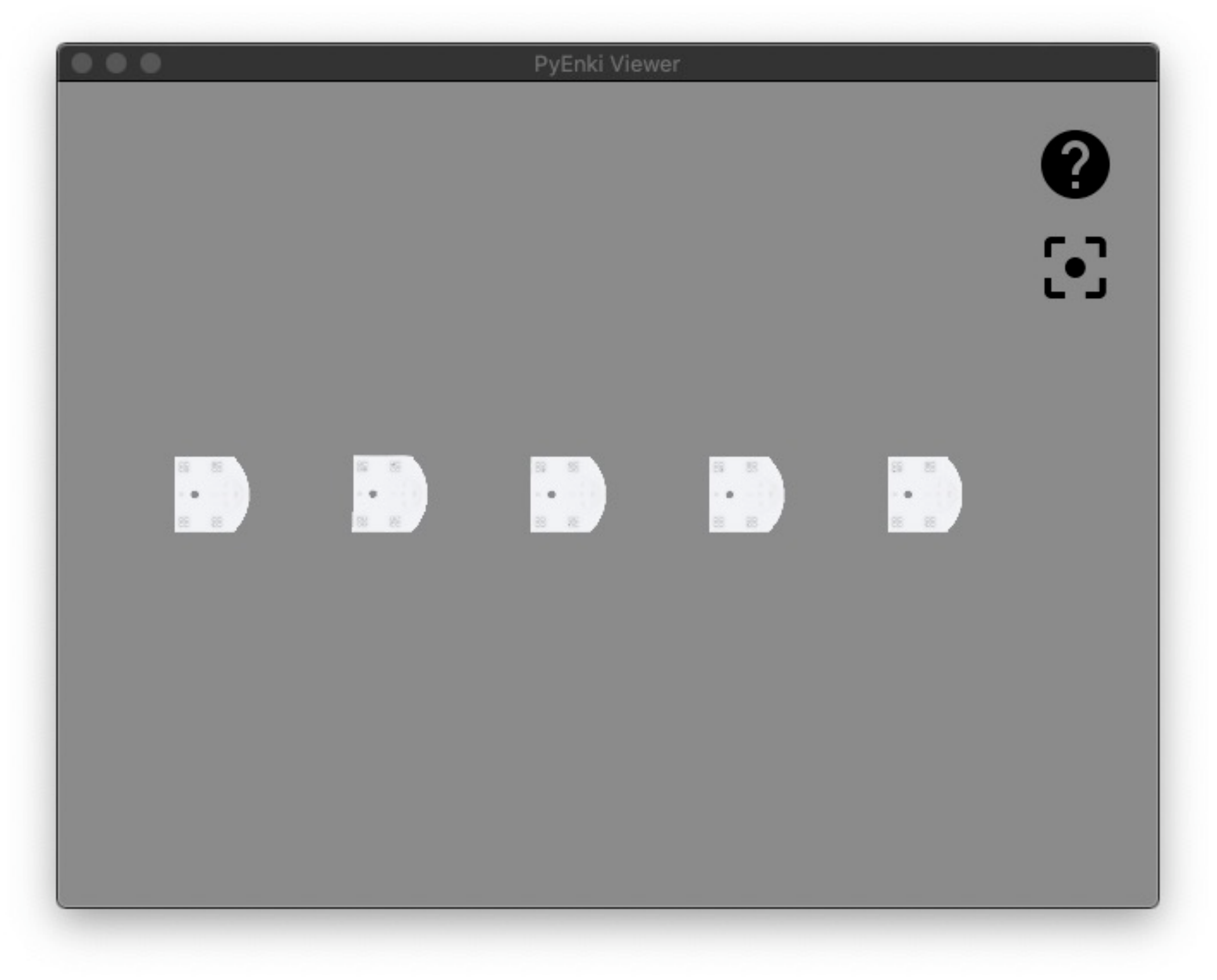}
		\end{subfigure}
	\end{center}
	\vspace{-0.5cm}
	\caption[Visualisation of the simulation of the first task.]{Visualisation of the 
	initial and final configurations obtained simulating the first task.}
	\label{fig:task1}
%	\vspace{-0.5cm}
\end{figure}
Perceiving the environment using its own sensors, in particular sensing the 
distances from the neighbours, each robot achieves the goal moving towards the 
target by minimising the difference between the values recorded by the front and 
rear sensors, trying to maintain the maximum achievable speed. This means that 
each robot is at the same distance from the one in front and the one behind, that 
should be the same distance among all the agents.

\bigskip
In the second scenario, shown in Figure \ref{fig:task2}, assuming that the agents 
in space are divided into two sets, the objective is to colour them depending on 
their group membership. 
For the sake of simplicity, we decide the group each robot belongs to based on 
the total number of robots: in case of an even number of agents, those in the first 
half of the row belong to the first group and the remaining to the second, while in 
the case of an odd number of robots, the same reasoning is applied and the 
central agent is assigned to the first set.
To perform this task, each agent performs actions — colouring the top \ac{led} 
in red or blue — based on the observations received from the environment — the 
messages received from neighbours. 
\begin{figure}[!htb]
	\begin{center}
		\begin{subfigure}[h]{0.49\textwidth}
			\centering
			\includegraphics[width=\textwidth]{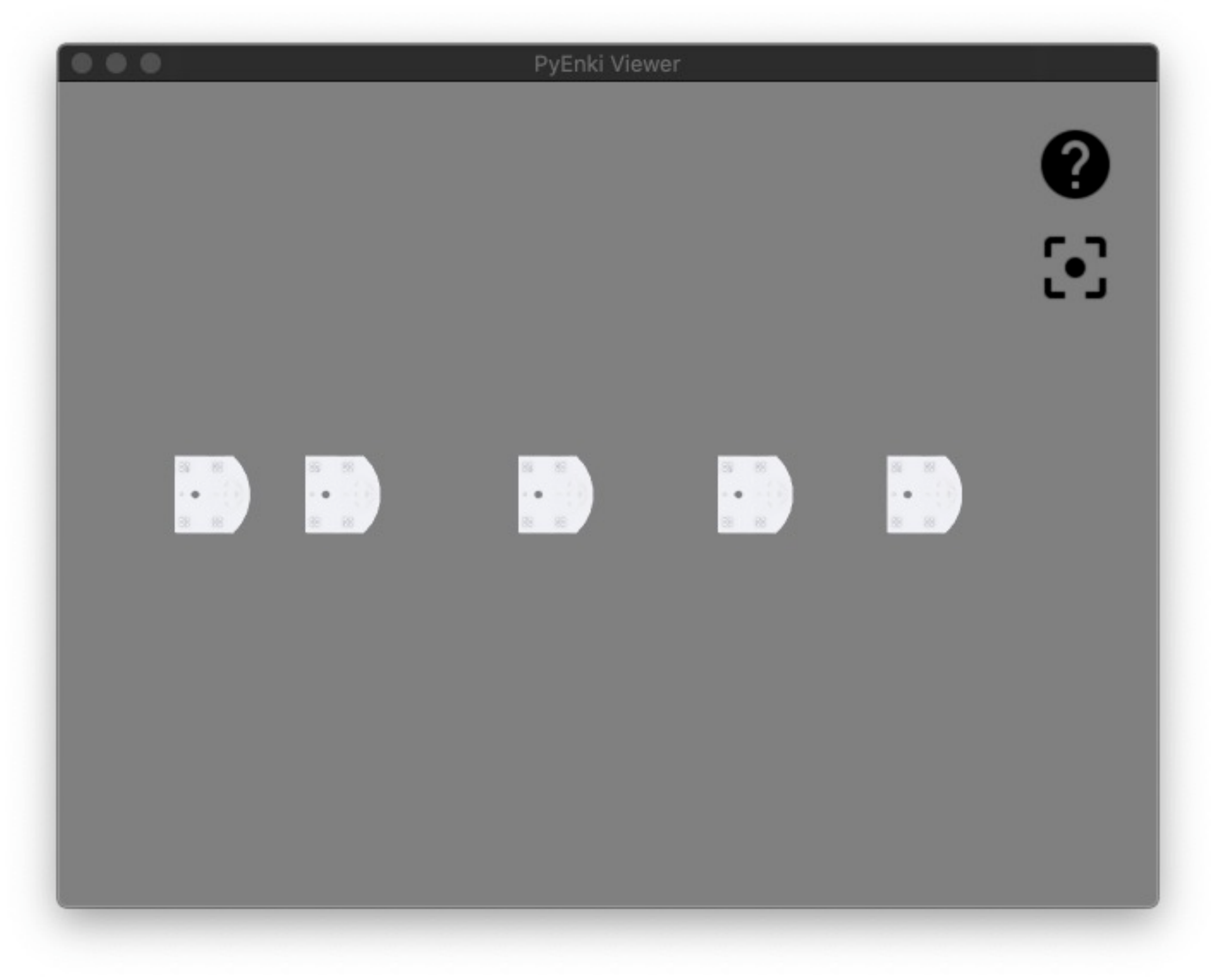}
		\end{subfigure}
		\hfill
		\begin{subfigure}[h]{0.49\textwidth}
			\centering
			\includegraphics[width=\textwidth]{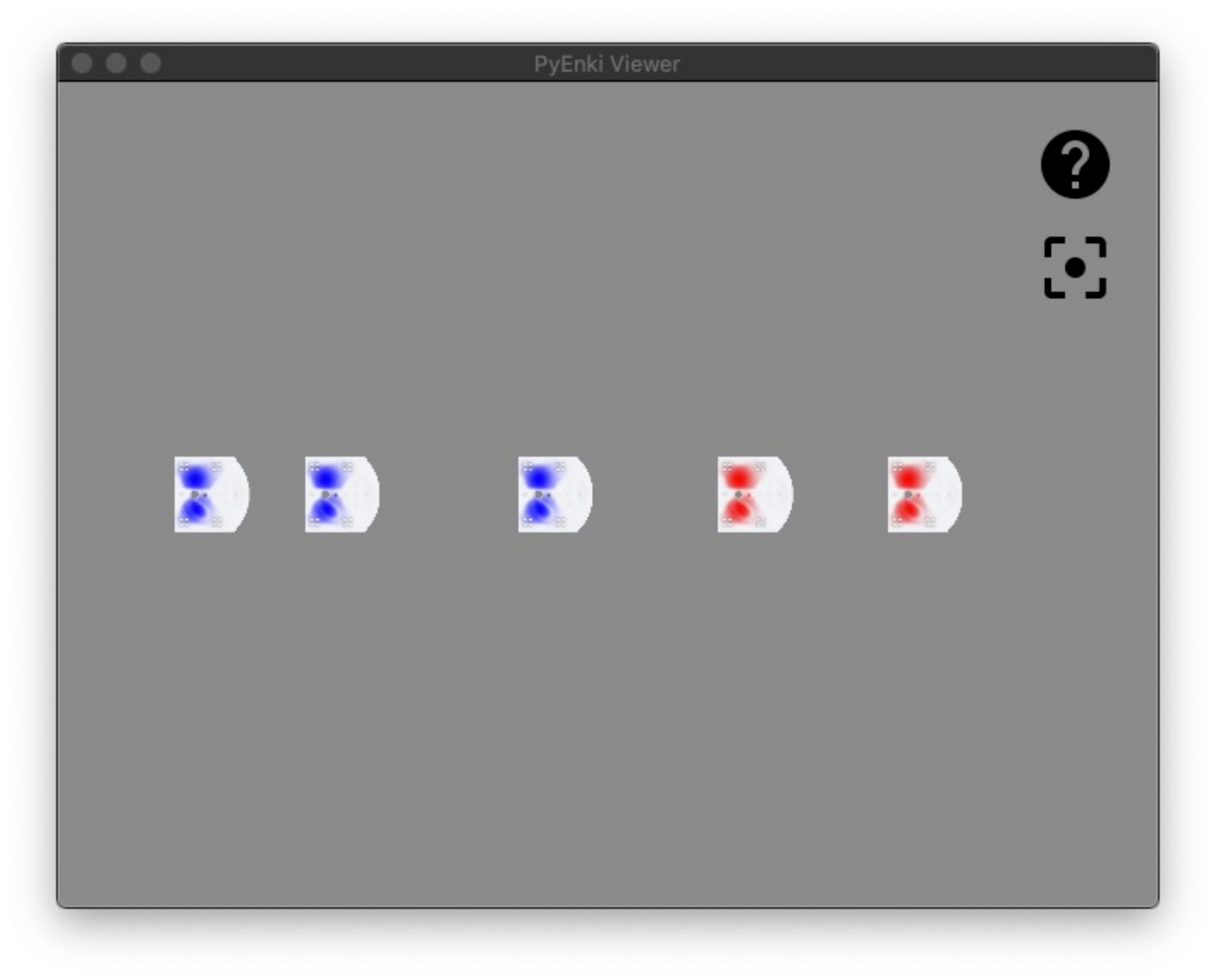}
		\end{subfigure}
	\end{center}
	\vspace{-0.5cm}
	\caption[Visualisation of the simulation of the second task.]{Visualisation of the 
		initial and final configurations obtained simulating the second task.}
	\label{fig:task2}
\end{figure}
This time, sensing the distance from neighbours does not provide useful 
information about the order of the agents, therefore they are not considered to 
accomplish this task. Instead, the only way for the robots to understand their 
ordering is necessarily using a communication protocol. For example, starting 
from the extreme agents, the first and the last in the row, or those that cannot 
receive any communication respectively from back and front, all the robots 
retransmit the received value, increased by one. In this way, the agents will in a 
sense, learn to count in order to understand which is the correct value to transmit.

Both the scenarios are examples of cooperative tasks based on the use of a 
distributed controller. While the second cannot be solved without allowing an 
explicit exchange of messages between the agents, in the first one a 
communication protocol is not necessary, but it may nonetheless increase 
performance.
To solve these tasks we have adopted two different methodologies, depending on 
whether the communication is used or not.
First of all, we analyse typical supervised learning approaches, which directly learn 
a mapping from observations to actions. This method can be applied only to the 
first task, in which a very simple ``distributed network'' that takes as input an 
array containing the response values of the sensors — either 
\texttt{prox\_values},  \texttt{prox\_comm} or  \texttt{all\_sensors} — and 
produces as output an array containing one float that represents the speed of the 
wheels, which is assumed to be the same both right and left, so that the robot 
moves straight.
After that, we concentrate on more challenging situations where the 
communication is not provided to the network, instead, it is a latent variable 
which has to be inferred \cite[][]{le2017coordinated}.
Throughout the experiments, we show the effectiveness of our methods, 
comparing approaches with and without communication. We also analyse the 
effects of varying the inputs of the networks, the initial distance between the 
robot and the number of agents chosen.

\subsubsection*{Outline}
\label{subsec:outline}

The thesis is composed of 6 chapters, whose main points are presented as follows:
\begin{itemize}
	\item Chapter \ref{chap:stateoftheart} summarises the previous research 
	on the topic, evaluating the approaches adopted by the authors;
	
	\item Chapter \ref{chap:background} provides the background knowledge 
	needed to properly understand the research contents;
	
	\item Chapter \ref{chap:impl} presents the tools used for the data collection 
	and all the additional frameworks we relied on;
		
	\item Chapter \ref{chap:methods} thoroughly illustrates the methodology used, 
	their benefits and limitations, also including descriptions of the kind of data 
	used and how they are collected;

	\item Chapter \ref{chap:experiments} explores the analysis conducted and 
	shows evaluation results;	
	
	\item The \hyperref[chap:concl]{Conclusion} addresses the results of the 
	experiments, concludes the thesis by discussing the implications of our findings, 
	possible improvements and outlines future works.
	
\end{itemize}

\acresetall
\setcounter{chapter}{0}
\chapter{Literature review}
\label{chap:stateoftheart}
This chapter discusses previous research about the topic, providing a brief 
introduction to the approach we adopted.
%, evaluating, in particular, their shortcomings. 

\bigskip
In recent years, the application of \ac{ai} and \ac{dl} techniques to multi-agent 
cooperative problems has become increasingly successful.
Gasser and Huhns, in \emph{Distributed Artificial Intelligence} 
\cite[][]{gasser2014distributed}, have addressed the issue of coordination and 
cooperation among agents with the combination of distributed systems and 
\ac{ai}, a discipline known as \ac{dai}. 

Similarly, Panait and Luke, in their work \emph{Cooperative multi-agent learning: 
The state of the art} \cite[][]{panait2005cooperative}, report that \ac{mal}, the 
application of machine learning to problems involving multiple agents, has 
become a popular approach which deals with unusually large search spaces. 
In addition, they mention three of the most classic techniques used to solve 
cooperative \ac{ma} problems, that are Supervised and Unsupervised Learning, 
and \ac{rl}. These methods are distinguished according to the type of feedback 
provided to the agent: the target output in the first case, no feedback is provided 
in the second, and reward based on the learned output in the last one.

Given these three options, the vast majority of articles in this field used 
reward-based methods to approach \ac{mal} scenarios, in particular \ac{rl}, that 
make they possible to achieve sophisticated goals in complex and uncertain 
environments \cite[][]{oliehoek2012decentralised}. 
However, this technique is notoriously known for being hard, in particular it is 
difficult to design a suitable reward function for the agents to optimise, which 
precisely leads to the desired behaviour in all possible scenarios. This problem is 
further exacerbated in multi-agent settings \cite[][]{hadfield2017inverse, 
oliehoek2012decentralised}.

\ac{irl} addresses this problem by using a given set of expert trajectories to 
derive a reward function under which these are optimal. 
Nevertheless, this technique imposes often unrealistic requirements 
\cite[][]{vsovsic2016inverse}.

\ac{il} is a class of methods that has been successfully applied to a wide range of 
domains in robotics, for example, autonomous driving 
\cite[][]{schaal1999imitation, stepputtis2019imitation}. They aim to overcome the 
issues aforementioned and, unlike reward-based methods, the model 
acquires skills and provides actions to the agents by observing the desired 
behaviour, performed by an expert \cite[][]{song2018multi, zhang2018deep, 
billard2008survey}.
Using this approach the models learn how to extract relevant information from 
the data provided to them, directly learning a mapping from observations to 
actions. 
A more challenging situation occurs when the model also has to infer the 
coordination among agents that is implicit in the demonstrations, using 
unsupervised approaches to imitation.

In this direction, literature suggests that cooperative tasks sometimes cannot be 
solved using only a simple distributed approach, instead it may be necessary to 
allow an explicit exchange of messages between the agents.
In \emph{Multi-agent reinforcement learning: Independent vs. cooperative 
agents} \cite[][]{tan1993multi}, Tan affirms that cooperating learners should use 
communication in a variety of ways in order to improve team performance: they 
can share instantaneous informations as well as episodic experience and learned 
knowledge.
Also Pesce and Montana propose the use of inter-agent communication for 
situations in which the agents can only acquire partial observations and are faced 
with a task requiring coordination and synchronisation skills 
\cite[][]{pesce2019improving}. Their solution consists in an explicit 
communication system that allows agents to exchange messages that are used 
together with local observations to decide which actions to take.

Our work is based on \emph{Learning distributed controllers by backpropagation}
\cite[][]{marcoverna2020}, which proposes an approach in which a distributed 
policy for the agents and a coordination model are learned at the same time. 
This method is based on an important concept introduced in \emph{Coordinated 
multi-agent imitation learning} \cite[][]{le2017coordinated}: the network has the 
ability to autonomously determine the communication protocol.
Likewise, we use imitation learning approaches to solve the problem of 
coordinating multiple agents, introducing a communication protocol, which 
consists in an explicit exchange of messages between the robots. The 
communication is not provided to the network, instead, it is a latent variable 
which has to be inferred.   
The results show the effectiveness of this communication strategy developed, as 
well as an illustration of the different patterns emerging from the tasks.

\acresetall
\chapter{Background}
\label{chap:background}

This chapter provides some background concepts to understand the material 
presented in this thesis. 
We start Section \ref{sec:ddr} with a review about differential drive robots and 
their kinematics, providing some useful notions about the types of sensors used 
and finally summarising the theory of control.
We conclude with a refresher on \acp{nn}, in Section \ref{sec:nn}.

\section{Differential drive robots}
\label{sec:ddr}
The robot we use for this study is Thymio II, a non-holonomic differential drive 
robot.

A differential wheeled vehicle is mobile robot, that typically consists of a rigid 
body 
\begin{figure}[!htb]
	\centering
	\includegraphics[width=.55\textwidth]{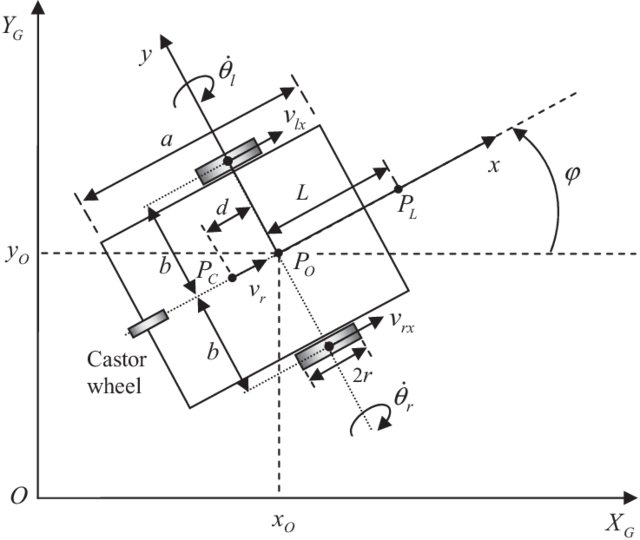}
	\caption[Non-holonomic differential drive mobile robot.]{Configuration of a 
		non-holonomic differential drive mobile robot 
		\cite[][]{shojaei2011adaptive}.}
	\label{fig:differentialdrive}
\end{figure}

\noindent
and a mobile base with a system of wheels that allows movement in the 
environment. 
Its movements are based on two, independently powered and 
controlled, fixed wheels, placed on either side of the robot body, with a common 
axis of rotation, and one passive caster wheel, whose job is to keep the robot 
statically balanced. 
Thus, both drive and steering functions are provided. In fact, it is possible to 
impose different values of velocity for the two fixed wheels, allowing the robot to 
rotate and move back and forth \cite[][]{siciliano2010robotics}. It is important to 
note that the robot is non-holonomic since it cannot translate laterally.

The \ac{icc}, is the point which lies along the horizontal axis, common to the left 
and right wheels, perpendicular to the plane of each wheel. Its position changes 
over time as a function of the individual wheel velocities, in particular of their 
relative difference. Since the rate of rotation $\omega$ is the same for both 
wheels, the following expressions hold:
\begin{Equation}[!h]
	\centering
	\begin{equation}
	\omega \, (R + b) = V_{rx}
	\end{equation}
	\begin{equation}
	\omega \, (R - b) = V_{lx}
	\end{equation}
	\caption[Right and left linear velocities of a differential drive robot.]{Right and 
	left linear velocities of a differential drive robot, computed as function of the 
	rotation rate, the distance between the wheels, and the distance from the 
	\ac{icc} and the robot reference point.}
	\label{eq:velocities}
\end{Equation}

\noindent
where $b$ is the distance between the centres of the two wheels, $V_{rx}$ and 
$V_{lx}$ are the right and left wheel velocities, and $R$ is the signed distance 
from the \ac{icc} to the midpoint between the wheels $P_O$ 
\cite[][]{dudek2010computational}. 
Moreover, at any specific time instant $t$ we can compute $R$ and  $\omega$ as 
follows:
\begin{Equation}[!h]
	\centering
	\begin{equation}
	R = b \, \frac{V_{rx} + V_{lx}}{V_{rx} - V_{lx}}
	\end{equation}
	\caption[Function to compute the distance from the ICC to the robot 
	reference point.]{Function to compute $R$, the signed distance from the 
	\ac{icc} to the midpoint between the wheels $P_O$, or robot reference point, 
	using the right and left linear velocities and the distance between the centres of 
	the two wheels.}
	\label{eq:r}
\end{Equation}

\begin{Equation}[!h]
	\centering
	\begin{equation}
	\omega = \frac{V_{rx} - V_{lx}}{2b}
	\end{equation}
	\caption[Function to compute the angular velocity of the robot.]{Function to 
	compute $\omega$, the angular velocity of the robot, using the right and left 
	linear velocities and the distance between the centres of the two wheels.}
	\label{eq:omega}
\end{Equation}

\bigskip
As a consequence, by varying the speed of the two wheels, $V_{rx}$ and 
$V_{lx}$, the trajectories that the robot takes changes as well:
\begin{itemize}
	\item $V_{rx} = V_{lx}$: if the same speed is set to both wheels, we have a 
	forward linear motion in a straight line. $R$ becomes infinite, and there
	is effectively no rotation $\omega = 0$.
	 
	\item $V_{rx} = - V_{lx}$: if both wheels are driven with equal speed but in the 
	opposite direction, we have a rotation about the midpoint of the wheel axis, or 
	in-place rotation. $R=0$, the \ac{icc} coincide with $P_O$ and $\omega = - 	
	\frac{V}{b}$.
	
	\item $V_{lx} = 0$: if the left wheel is not powered, we have a  
	counter-clockwise rotation about the left wheel. $R=b$ and 
	$\omega=\frac{V_{rx}}{2b}$.
	
	\item $V_{rx} = 0$: if the right wheel is not powered, we have a clockwise 
	rotation about the right wheel. $R=-b$ and $\omega=-\frac{V_{lx}}{2b}$.
\end{itemize}

Differential drive robots, however, are sensitive to slight changes in velocity, and 
even small errors can affect the robot trajectory. Moreover, they are also 
susceptible to variations in the ground.

\subsection{Sensors}
\label{subsec:sensors}

A significant feature of the Thymio II robot, is the a large number of sensors with 
which it is equipped.

The adoption of external sensing mechanism is of crucial importance to allow a 
robot to interact with its environment and achieve high-performance 
\cite[][]{fu1987robotics, siciliano2010robotics}. 

It is possible to classify the sensors in two principal categories, according to their 
function: \emph{proprioceptive} sensors, that measure the internal state and deal 
with the the detection of variables used for the control, such as the robot position, 
and \emph{exteroceptive} sensors that measure the external state of the 
environment, dealing with the detection of variables such as range and proximity, 
often used for robot guidance as well as object detection. 

For the purposes of this thesis we are interested in the study of proximity sensors, 
in particular the \ac{ir} sensors, that used by Thymio II.

Proximity sensors are an example of exteroceptive sensors: they gather 
information from the environment around the robot, such as distance to objects. 
They are 
also \emph{active} sensors: they emit their own energy, usually light, and measure 
the reflection. 
Among the advantages of this type of sensors is the fact that the infrared 
beams generated by the sources can be used unobtrusively, since they are 
invisible to the human eye.

\ac{ir} sensors are composed of an emitter, which radiates invisible infrared light, 
and a receiver, that measures the intensity of the reflection and the quantity of 
light that comes back.
If the reflection is strong enough — a large part of the light is reflected from 
the object and returns to the robot — it can be inferred that the obstacle is 
relatively close and lies within a certain range of the sensor, depending on 
the received intensity. If the object is farther away, only a small 
part of the light comes back \cite[][]{fu1987robotics}.
However, this estimate can be significantly affected by some property of the 
obstacle, such as its colour and reflectivity, but also by the presence of external 
light sources and the temperature of the environment — e.g. a black object 
reflects less light than a white one, placed at the same distance 
\cite[][]{mordechai2018elements}.

\subsection{Control theory}
\label{subsec:control}
The problem of controlling a robot is of a primary importance: to achieve a given 
task, an agent must acts based on the perception received from the environment. 
To do so, robots use a controller that takes decisions according to algorithms 
developed to accomplish the goal.

We can distinguish two main techniques of control: \emph{open-loop control} 
sets in advance the parameters of the algorithm and never observes the result of 
its actions to adjust them to the actual state, \emph{closed-loop control} or 
feedback-based control, measures the error between the desired state of the 
system and the actual one, and uses this feedback to to adjust the control and 
decide the next action to take \cite[][]{mordechai2018elements}.

These methods can be used to compute trajectories that lead the robot from an 
initial configuration to a final one. To achieve a more appropriate behaviour we 
are interested in the use of closed-loop control systems 
\cite[][]{siegwart2011introduction}.

\begin{figure}[!htb]
	\centering
	\includegraphics[width=.8\textwidth]{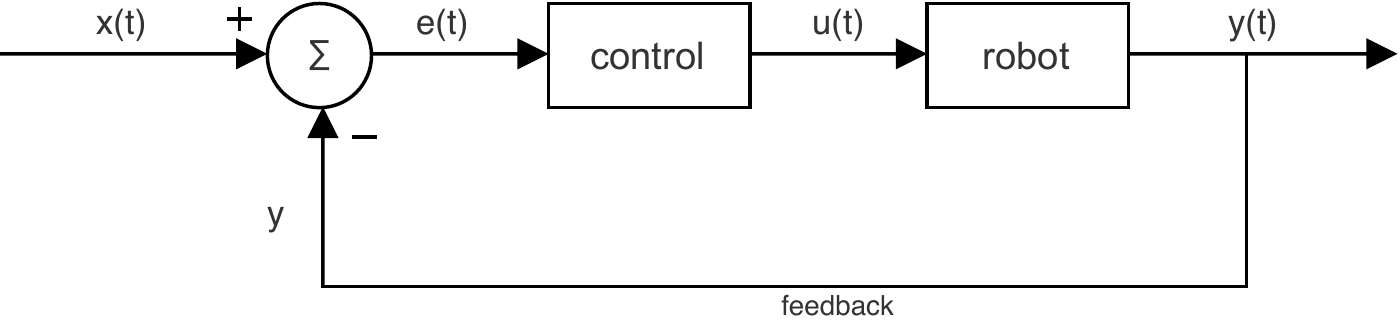}
	\caption{Closed-loop control system.}
	\label{fig:closedloopcontrol}
\end{figure}
This model can be formally defined using the variables introduced below: 
\begin{itemize}
	\item $x(t)$, called command or set point (SP), represents the reference 
	value, or a desired output. It cannot be imposed directly on the robot, instead it 
	is transformed into a control value $u(t)$.
	
	\item $u(t)$, called control variable, is the output of the control and the 
	input of the robot.
	
	\item $y(t)$, called process variable (PV), represents an observable part of the 
	the actual state of the robot. It is the feedback signal combined with the set 
	point to compute the error value.

	\item $e(t)$ is the error. It is computed as the difference between the value of 
	the set point and the process variable, as shown in Equation 
	\ref{eq:systemerror}. The error is used to generate the new control signal $u(t)$.
	\begin{Equation}[!h]
		\centering
		\begin{equation}
		e(t) = x(t) - y(t)
		\end{equation}
		\caption{Calculation of the error value $e(t)$ of the system.}
		\label{eq:systemerror}
	\end{Equation}	
\end{itemize} 

The closed-loop control algorithms sometimes can be very sophisticated. For the 
purpose of this study, we design two algorithms, the first one is a \ac{pid} 
controller, while the second a Bang Bang controller.

The implementation of any feedback controller requires the availability of the 
robot configuration at each time instant. 

\subsubsection{Proportional (P) controller}
\label{subsubsec:pid}

Considering real situations in which agents do not have access to their states but 
only to their local observations, the controller we design for achieve the goal of 
the first task is a Proportional (P) Controller,  a particular variant of \ac{pid} 
Controller, with only the $K_p$ term, more details are provided in the Section 
\ref{subsubsec:manualtask1}.

The closed-loop control uses a feedback to adjust the control while the action 
takes place in proportion to the existing error. This function, given a desired 
output $x(t)$, or set point, produces an output $y(t)$, or process variable, such 
that the error $e(t)$ is obtained as the difference between the value of the set 
point and the process variable. Finally, the control variable $u(t)$ is the output of 
the \ac{pid} controller and is computed as follows:
\begin{Equation}[!h]
	\centering
	\begin{equation}
	u(t) = K_p * e(t)
	\end{equation}
	\caption[Proportioal PID controller.]{Proportional \ac{pid} controller.}
	\label{eq:pid}
\end{Equation}

The value of the proportional gain can be tuned to yield satisfactory 
performance so that the system is stable.

\subsubsection{Bang-bang controller}
\label{subsubsec:bangbang}
The controller we design to achieve the goal of the first task, of which more 
details are provided in the Section \ref{subsubsec:experttask1}, is a variant of the 
Bang-bang algorithm.

In this case we considered instead a situation in which agents have access to their 
states: given the set point and the variable measured, the goal and the actual 
position of the robot, the error is computed as the difference between the two 
quantities.
What we want to achieve is that the error is 0, to do so, when the error is negative, 
the robot should move forward to reach the goal, on the contrary if is positive it 
should move backwards. Its main feature is that the motor powers are turned to 
full forwards or full backwards depending on the sign of the error. 

This approach has many advantages, such as its simplicity and no need for 
calibration. 
However, since this controller is not always precise, especially when a derivative 
term is needed to avoid oscillations or when we are in steady-state, close to 
desired value, we implemented a variant that moves the the robot at full speed 
unless they are closer than a certain value, avoiding to approach the goal at full 
speed and overshoot it.

\section{Artificial Neural Networks}
\label{sec:nn}
In the field of \ac{ml}, \acp{ann} are mathematical models based on the 
simplification of Biological Neural Networks \cite[][]{zou2008overview}. 

%FIXME missing reference of the image
\begin{figure}[htb]
	\centering
	\includegraphics[width=.7\textwidth]{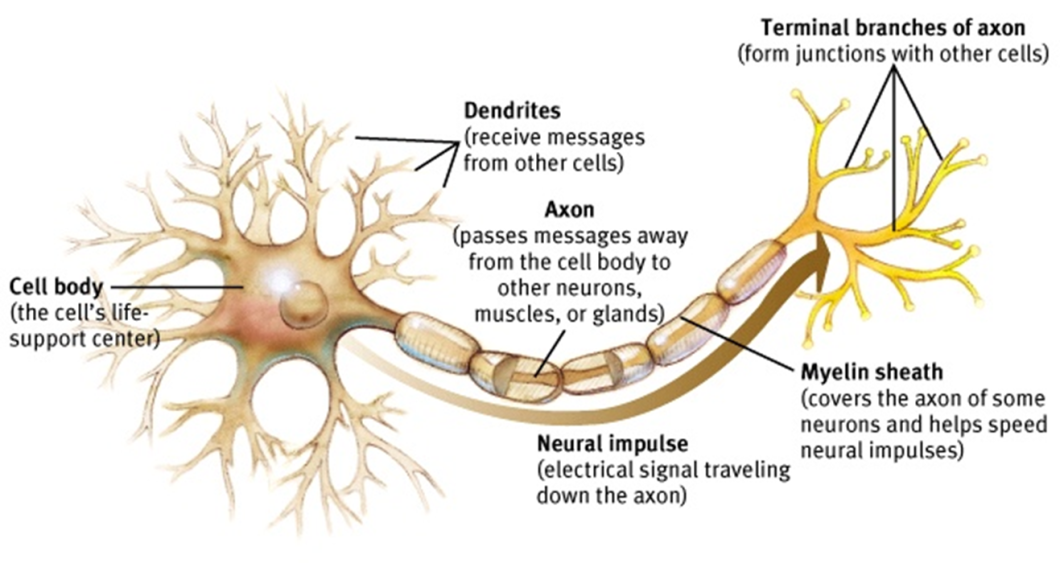}
	\caption{Structure of a Biological Neuron.}
	\label{fig:bioneuron}
\end{figure}

A \ac{nn} can be considered as a dynamic system having the topology of an 
oriented graph, whose nodes model the neurons in a biological brain, while the 
edges represent the synapses — interconnections of information.
Each connection can transmit a signal from one artificial neuron to another,
which are typically aggregated in layers. The stimuli are received by a level of 
input nodes, called processing unit, which processes the signal and transmits it to 
other neurons connected to it.

As we anticipated, \acp{nn} can be seen as mathematical models that define a 
function $f : X \rightarrow Y$. 
The network function of a neuron $f(x)$ is defined as a composition of
other functions $g_i(x)$, which can in turn be decomposed into others.
A widely used representation for the description of traditional \acp{ann} is the 
weighted sum, shown in Equation \ref{eq:mathmodel}.

\begin{Equation}[!h]
	\centering
	\begin{equation}
	f(x)=\phi \bigg( \sum_{i}w_{i}x_i + \theta \bigg)
	\end{equation}
	\caption[Mathematical description of traditional \acp{ann}.]{Function that 
	describes mathematically the traditional \acp{ann} in terms of weighted sum.}
	\label{eq:mathmodel}
\end{Equation}
Each input signal $x_i$ is multiplied by its corresponding weight $w_{i}$, which 
assumes a positive or negative value depending on whether you want to excite or 
inhibit the neuron.
The bias $\theta$ varies according to the propensity of the neuron to activate, 
influencing its output.
Additionally, a predefined function $\phi$ can be applied, also called activation 
function, which is explained in the following paragraph.

\begin{figure}[htb]
	\centering
	\includegraphics[width=.7\textwidth]{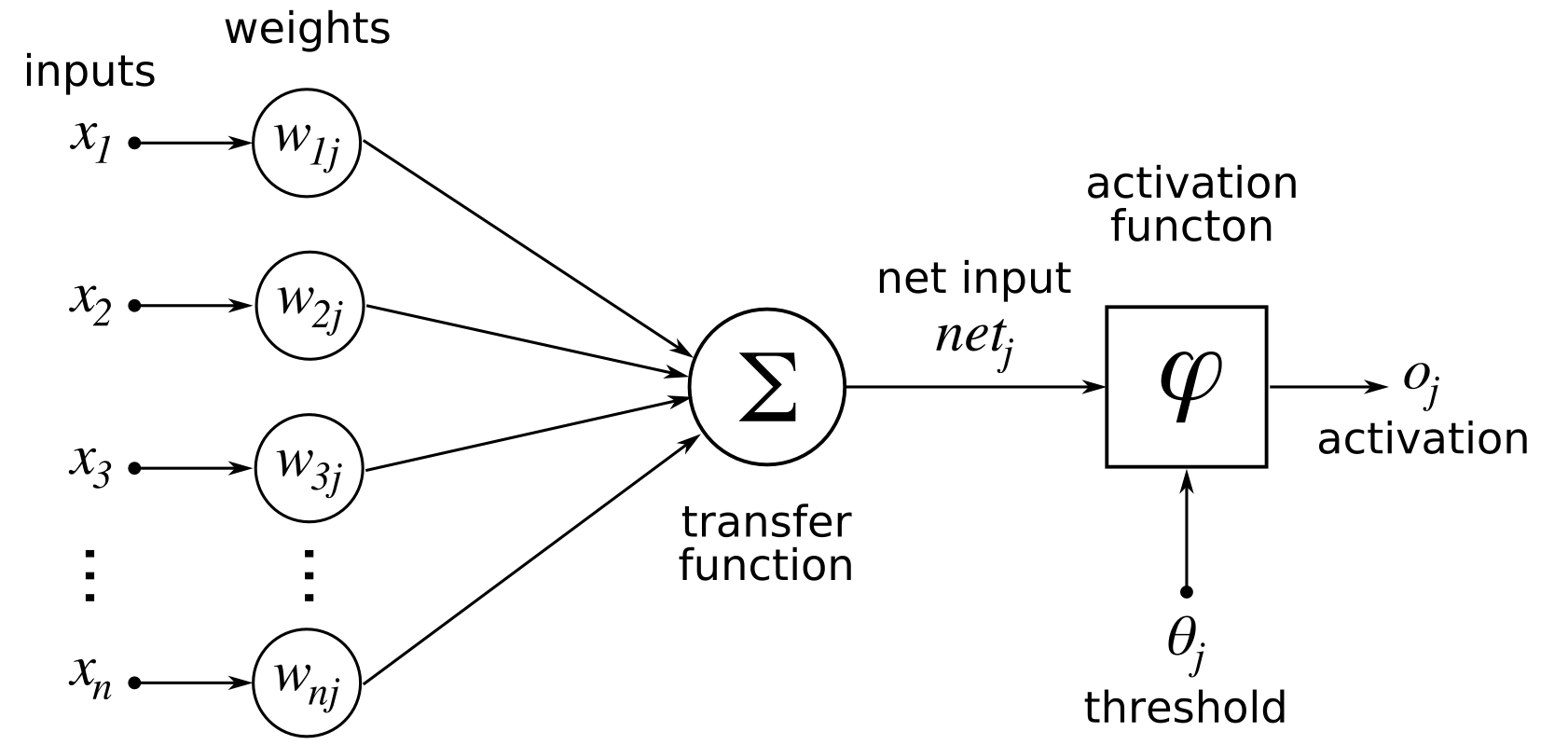}
	\caption{Mathematical model of an Artificial Neuron.}
	\label{fig:neuron}
	\vspace{-0.5cm}
\end{figure}

\subsection{Activation functions}
\label{subsec:activationfun}

An activation function is a fundamental component of the model. It allows the 
network to learn non-linear transformations, in order to be able to compute 
non-trivial problems.
In the course of this study, we used two of the most popular activation functions  
in deep learning, the {hyperbolic tangent} (Tanh) \cite[][]{kalman1992tanh} 
and the {sigmoid} \cite[][]{han1995influence}, visualised in Figure 
\ref{fig:activation}.

\paragraph*{Tanh}
The tanh is a zero-centred function, whose range lies between $(-1, 1)$, and its 
output is given by the following formula:
\begin{Equation}[H]
	\centering
	\begin{equation}
	f(x)= \frac{\sinh (x)}{\cosh (x)} = \bigg( \frac{e^x - e^{-x}}{e^x + 
		e^{-x}}\bigg)
	\end{equation}
	\caption{Hyperbolic Tangent Function (Tanh).}
	\label{eq:tanh}
\end{Equation}

\paragraph*{Sigmoid}
The sigmoid models the frequency of the stimuli emitted by an inactive neuron, 
$\sigma(x)=0$, to one fully saturated with the maximum activation frequency, 
$\sigma(x)=1$. Its  output is given by the following formula:
\begin{Equation}[H]
	\centering
	\begin{equation}
	\sigma(x)= \frac{1}{1 + e^{-x}}
	\end{equation}
	\caption{Sigmoid Function.}
	\label{eq:sigmoid}
\end{Equation}

\begin{figure}[!htb]
	\begin{center}
		\begin{subfigure}[h]{0.495\textwidth}
			\includegraphics[width=.8\textwidth]{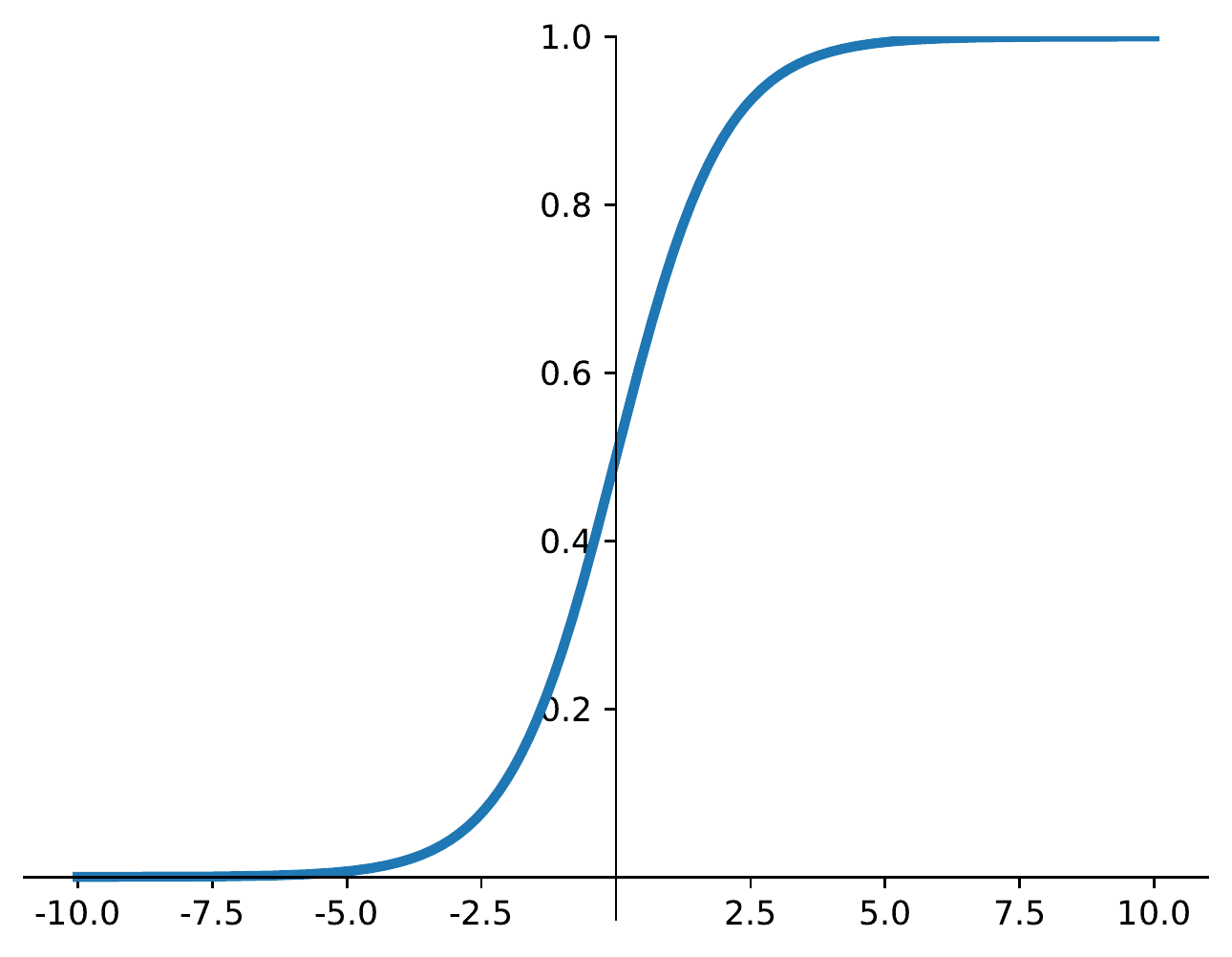}
			\caption{Tanh activation function.}
		\end{subfigure}
		\hfill
		\begin{subfigure}[h]{0.495\textwidth}
			\includegraphics[width=.8\textwidth]{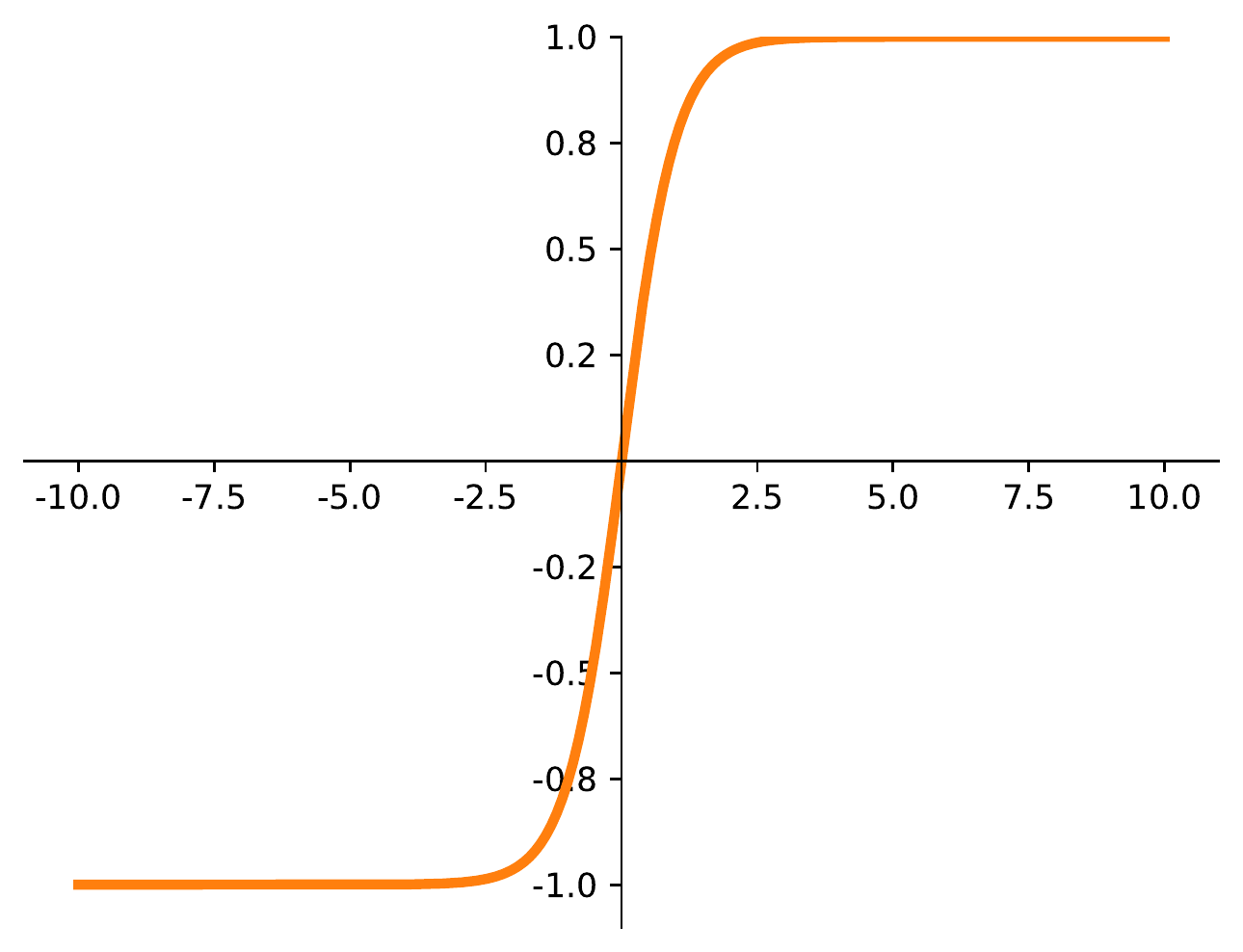}
			\caption{Sigmoid activation function.}
		\end{subfigure}
	\end{center}
	\caption{Trends of two non-linear activation functions.}
	\label{fig:activation}
\end{figure}

%\section{Architetture}
%\label{sec:architetture}
%
%I neuroni vengono organizzati in una struttura detta architettura della rete.
%I dati, partendo da un livello iniziale, chiamato layer di input, attraversano i 
%multipli strati interni della rete, gli hidden layer, raggiungendo l'ultimo livello 
%detto layer di output.
%
%Quando i collegamenti tra i neuroni formano una struttura senza cicli si parla di 
%reti \emph{feed-forward} \cite{svozil1997introduction}.
%
%\subsection{Layer fully-connected}
%\label{subsec:fc}
%
%Un’architettura molto comune nelle reti neurali è una struttura ``densa'', che 
%utilizza \emph{layer fully-connected}, in cui tutti i neuroni del livello precedente 
%sono collegati ad ogni neurone dello strato successivo 
%\cite{sainath2015convolutional}.
%
%Lo scopo di un layer completamente connesso è imparare combinazioni non 
%lineari di feature ad alto livello provenienti dal layer precedente. 
%Una struttura di questo tipo è però caratterizzata da un numero di connessioni 
%che cresce molto velocemente, causando un accrescimento del numero di 
%parametri che la rete deve apprendere.
%Questo comporta un aumento del costo computazionale e un alto rischio di 
%overfitting, approfondito nella sezione \ref{subsec:overfitting}.
%
%Per questo motivo questi vengono spesso sostituiti dai layer convoluzionali.

\subsection{Loss functions}
\label{subsec:lossfunctions}
\acreset{mse}
\acreset{bce}
The learning process is structured as a non-convex optimisation problem in which 
the aim is to minimise a cost function, which measures the distance between a 
particular solution and an optimal one.

In the course of this study we used two different objective functions, depending 
on the strategy to be adopted: to solve the first task, that can be modelled as a 
regression problem, we used the \ac{mse} \cite[][]{wang2009mean}, while for the 
second, that is a binary classification problem, we used the \ac{bce} 
\cite[][]{gomez2018understanding}.

\paragraph*{Mean Squared Error} 
The \ac{mse} computes the deviation between the values observed $\hat y_i$ and 
those predicted by the network $y_i$, over the number of predictions $n$, as 
shown in Equation \ref{eq:mse}.
\begin{Equation}[!htb]
	\centering
	\begin{equation}
	\mathtt{MSE} = \frac{\sum_{i=1}^n (y_i-\hat y_i)^2}{n}
	\end{equation}
	\caption{Mean Squared Error (MSE) loss function.}
	\label{eq:mse}
\end{Equation}
Formally, this criterion measures the average of squared error between 
predictions and targets, and learns to reduce it by penalising big errors in the 
model predictions.

\paragraph*{Binary Cross Entropy} 
The \ac{bce} is a combination of the sigmoid activation and the \ac{ce}. It sets up 
a binary classification problem between two classes, with the following 
formulation:

\begin{Equation}[!htb]
	\centering
	\begin{equation}
	\mathtt{BCE} = -\frac{1}{n} \sum_{i=1}^n y_i \cdot \log(\hat y_i) + (1-y_i) 
	\cdot \log(1 - \hat y_i)
	\end{equation}
	\caption[Binary Cross Entropy (BCE) loss function.\bigskip]{Binary Cross 
	Entropy 
	(\ac{bce}) loss function \cite[][]{sadowski2016notes}.}
	\label{eq:bce}
\end{Equation}

\noindent
where $\hat y_i$ is the $i$-th scalar value in the model output, $y_i$ is the 
corresponding target value, and $n$ is the number of scalar value in the model 
output\footnote{\url{https://peltarion.com/knowledge-center/documentation/modeling-view/build-an-ai-model/loss-functions/binary-crossentropy}}.
 
This loss function should return high values for bad predictions and low values for 
good ones.

\subsection{Optimisation algorithms}
\label{subsec:optimiser}
Optimisation algorithms are needed to minimise the result of a given objective 
function, which depends on the parameters the model has to learn during 
training.
They strongly influence the effectiveness of the learning process as they update 
and calculate the appropriate and optimal values of that model. 
In particular, the extent of the update is determined by the learning rate, which 
guarantees convergence to the global minimum, for convex error surfaces, and to 
a local minimum, for non-convex surfaces.

\paragraph*{Adam}

The optimiser we have chosen for this thesis project is Adam, {an algorithm for 
first-order gradient-based optimisation of stochastic objective functions, based 
on adaptive estimates of lower-order moments} \cite[][]{kingma2014adam, 
loshchilov2017decoupled}. 

\acresetall
\chapter{Tools}
\label{chap:impl}

This chapter introduces the tools used in this work, starting from the description 
of the target platform in Section \ref{sec:thymio}, then describing the simulator 
in Section \ref{sec:enki} and finally mentioning the frameworks used for the 
implementation in Section \ref{sec:learning}.

\section{Thymio II}
\label{sec:thymio}

The target platform is Thymio II, a small differential drive mobile robot developed 
in the context of a collaboration between the MOBOTS group of the \ac{epfl} and 
the \ac{ecal}. 

Thymio runs the Aseba open-source programming environment, an event-based 
modular architecture for distributed control of mobile robots, designed to enable 
beginners to program easily and efficiently \cite[][]{magnenat2010aseba, 
mondada2017bringing}, making it well-suited for robotic education and research.

Another particularity of this tool is the integration with the open-source 
\ac{ros} \cite[][]{quigley2009ros}, through asebaros bridge 
\cite[][]{asebaros}. 

The Thymio II includes sensors that can measure light, sound and distance. 
It can perform actions such as move using two wheels, each powered by its own 
motor, but also turning lights on and off.

\subsection{Motors}
\label{subsection:thymotors}
The robot is equipped with two motors, each connected to one of the two 
wheels, which allow the robot to move forward, backwards but also turn by 
setting the velocity of the wheels at different speeds. The maximum speed 
allowed 
to the agent is $16.6$cm/s.

\subsection{Sensors}
\label{subsec:thysensors}

Thymio II possesses a large number of sensors, but for the purpose of this 
study we focused only on usage of the horizontal proximity ones. 

Around the robot periphery are positioned $7$ distances sensors, five on the 
front and two on the rear. 
These sensors can measure the distances to nearby objects thanks to the use 
of  \ac{ir}, addressed in Section \ref{subsec:sensors}.
\begin{figure}[h!tb]
	\centering
	\includegraphics[width=.6\textwidth]{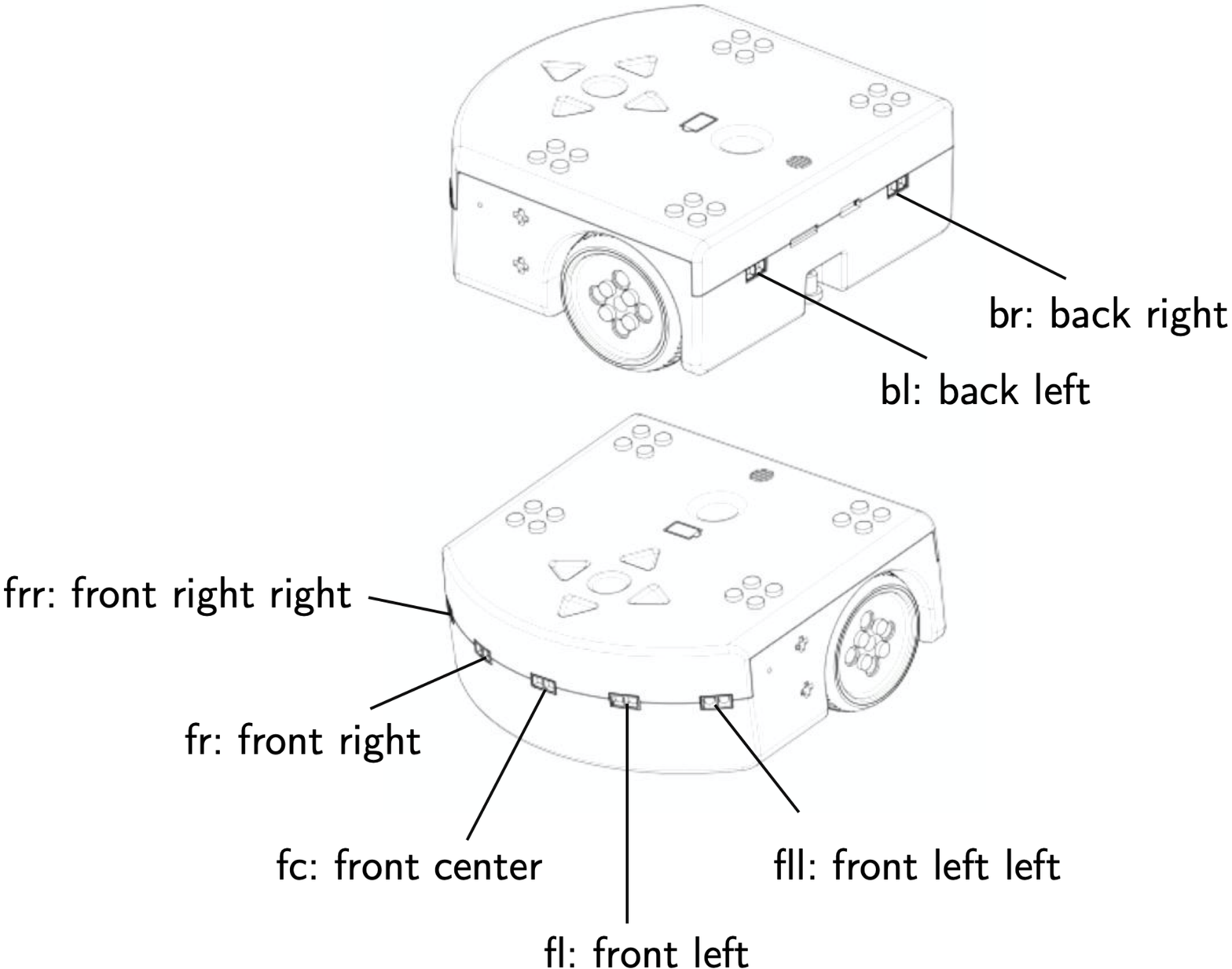}
	\caption{Thymio II sensors}
	\label{fig:thymio sensors}
\end{figure}

\subsection{Communication}
\label{subsec:thymiocomm}

Thymio II can employ its horizontal \ac{ir} sensors to communicate a value 
to robots within a range of about $48$cm. 
In particular, the robot can enable the proximity communication and send an 
integer payload of $11$ bits at $10$Hz, one value transmitted every 
$0.1$s. 
This message is received by other agents if at least one of their proximity sensors 
is visible by one of the emitter’s \acp{led} and if they have enabled the proximity 
communication.
Every time that a message is received by some robot, an event is 
created containing the message, payload data and intensities data.

\subsection{LEDs}
\label{subsec:thymioled}

Thymio II holds many \acp{led} scattered around its body, most of them are 
associated 
with sensors and can highlight their activations.

In this study, specifically for the objectives of Task 2 introduced later in Section 
\ref{sec:task2}, two \ac{rgb} \acp{led} on the top of the robot are used and 
driven together.
It is possible to set the intensities of the top \acp{led}, in particular choosing the 
value for each channel (red, green and blue) in a range from $0$ (off) to $32$ 
(fully lit).

\section{Enki simulator}
\label{sec:enki}

Enki is the open-source robot simulator used in this project. It provides collision 
and limited physics support for robots evolving on a flat surface and can simulate 
groups of robots a hundred times faster than real-time \cite[][]{enki}.

In this project, we exploit the PyEnki package \cite[][]{enki-jguzzi} that provides 
Python bindings to the Enki simulator using Boost::Python 
\cite[][]{boostpython}.
Moreover, it adds some functionalities to the original simulator such as the 
support for the proximity communication between Thymio II.
Another peculiarity is that the simulated world can be run in real time inside a Qt 
application or even without the \ac{gui} as fast as possible.

In the simulator, the environment is represented as a \ac{3d} Cartesian 
coordinate system, defined by three pair-wise perpendicular axes, $x$, $y$ and 
$z$, that go through the origin in $(0, 0, 0)$. 

In the following sections, the functioning of motors and sensors are explored in 
detail, as well as the concept of communication, anticipated in Section 
\ref{subsec:thymiocomm}, giving particular attention to the explanation of their 
use in the simulator.

\subsection{Motors}
\label{subsec:enkimotors}
As for the motors, they are used to establish the speed of the wheels. The target 
wheels' speed can be set by writing the variables 
\texttt{motor\_\{left,right\}\_target}. The velocity is a float value specified in 
centimetres per second (cm/s).
The wheels at maximum maximum allowed speed can provide a velocity of 
$16.6$cm/s.

Another important element used for the construction of the model constraints is 
the distances between the left and right driving wheels, that is fixed to $9.4$ 
cm.

Finally, a little amount of relative noise, about $0.027$, is added to the target 
wheel speed at each control step.

\subsection{Sensors}
\label{subsec:enkisensors}
In Enki, it is possible to access the Thymio sensor readings in two different ways, 
by using \texttt{prox\_values} and \texttt{prox\_comm\_events}.

\paragraph*{\texttt{prox\_values}}
 an array of floats that holds the values of $7$ horizontal distance sensors around 
 its periphery [\texttt{fll} (front left left), \texttt{fll} (front left), \texttt{fc} (front 
 centre), \texttt{fr} (front right right), \texttt{frr} (front right), \texttt{bl} (back left), 
 \texttt{br} (back right)]. 
These values can vary from $0$ — when the robot does not see anything — up to 
$4505$ — when the robot is very close to an obstacle. 
Thymio II updates this array at a frequency of $10$Hz, generating the 
\texttt{prox event} after every update. 
The maximum range of these sensors is $14$cm.

\paragraph*{\texttt{prox\_comm\_events}}

a list of events, one for every received message collected during the last control 
step of the simulation. 
After enabling the proximity communication, using \texttt{prox\_comm\_enable} 
command, the robot can use the horizontal \ac{ir} distance sensors to 
communicate a value to peer robots within a range of about $48$cm. 
The integer payload to be sent is contained in the variable 
\texttt{prox\_comm\_tx}, while the value received is contained in the variable 
\texttt{prox\_comm\_events.rx} of the fired event.
In addition, the IRCommEvent stores, in the variable 
\texttt{prox\_comm\_events.payloads}, a list of $7$ payloads, one for each sensor 
and a list of  $7$ intensities  [\texttt{fll}, \texttt{fll}, \texttt{fc}, \texttt{fr}, 
\texttt{frr}, \texttt{bl}, \texttt{br}], saved in the variable 
\texttt{prox\_comm\_events.intensities}.
The readings of the latter array, together with those contained in 
\texttt{prox\_values}, are those that will be used in the course of this study.

\begin{figure}[!htb]
	\centering
	\includegraphics[width=.6\textwidth]{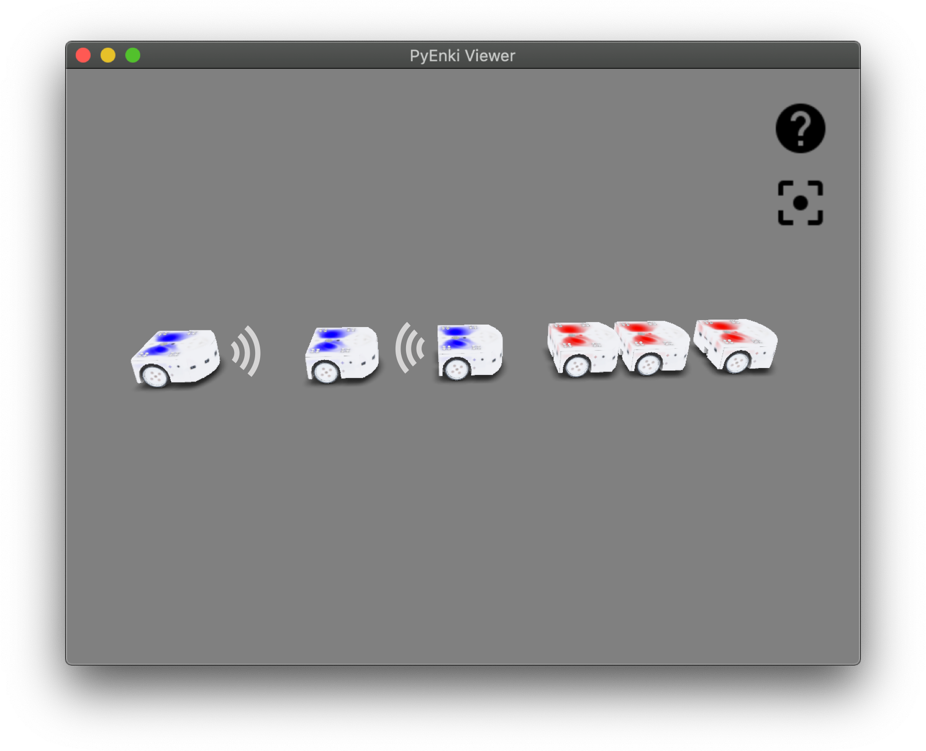}
	\caption[Example of communication with the PyEnki simulator]{The PyEnki 
	simulator viewer showing an example of communication, 
	where the second Thymio is receiving messages from the nearest agents.}
	\label{fig:thymio comm}
\end{figure}

\section{Frameworks}
\label{sec:learning}
The proposed imitation learning approaches have been implemented in 
\texttt{Python} and all the models have been trained on an NVIDIA Tesla 
V100-PCIE-16GB \ac{gpu}.
For this purpose, we used different tools which we briefly mention below.

\paragraph*{\texttt{PyTorch}} is an open source machine learning framework  for 
tensor computation with strong GPU acceleration, uses also to build 
\acp{dnn}\footnote{\url{https://pytorch.org}}.

\paragraph*{\texttt{NumPy}} is a package for scientific computing that provides 
support for a range of utilities for linear algebra and matrix 
manipulation\footnote{\url{https://numpy.org}}.

\paragraph*{\texttt{Pandas}} is an open source library providing 
high-performance, easy-to-use data structures and data analysis 
tools\footnote{\url{https://pandas.pydata.org}}. We used it in particular for 
building 
and working on the datasets generated for training the models. 

\paragraph*{\texttt{Matplotlib}} is a comprehensive package for creating 
static, animated, and interactive visualisations in 
\texttt{Python}\footnote{\url{https://matplotlib.org}}. 

\paragraph*{\texttt{scikit-learn}} is a library that provides simple and efficient 
tools for predictive data analysis, built on \texttt{NumPy}, \texttt{Scipy}, and 
\texttt{Matplotlib}\footnote{\url{https://scikit-learn.org}}.

\acresetall
\chapter{Methodologies}
\label{chap:methods}

This chapter illustrates the methodology used to approach the development 
process of this work. 
We start Section \ref{sec:MAS} by analysing the domain of the problem and then 
defining in Section \ref{sec:imitlrng} the learning method.
We continue with a presentation of the approaches considered in Section 
\ref{sec:approaches}, followed by Section \ref{sec:dataset} that describes the type 
of data used for this purpose and how they are generated.
We conclude Section \ref{sec:controllersmodel} with a detailed explanation of the 
controllers used and of the models implemented.

\section{Multi-agent systems}
\label{sec:MAS}

In this work, we investigate collaborative scenarios in a homogeneous \ac{mas}.

In a given simulated \ac{1d} environment, we consider a team of $N$ interacting 
agents, that are assumed to be interchangeable since they have the same physical 
structure and observation capabilities, which collaborate to solve a given common 
goal \cite[][]{stone2000multiagent, vsovsic2016inverse}.

This system, containing agents also called ``swarming agents’’, in principle is an 
extension of a \ac{decpomdp} \cite[][]{oliehoek2012decentralised}, and can be 
formally defined as a tuple $(S, O, A, R, \pi)$, \cite[][]{schaal1999imitation}, 
where:
\begin{itemize}
	\item $S$ is the state of the system, or the set of local states for each agent, 
	composed of all the possible combinations of positions and observations.
	\item $O$ is the set of possible observations for each agent, obtained through 
	the sensors.
	\item $A$ is the set of possible actions for each agent.
	\item $R$ is the reward function $S \times A \rightarrow \mathbb{R}$.
	\item $\pi$ is the policy $S \rightarrow A$ which determines the action to 
	execute in a given state for every single agent.
\end{itemize}

As introduced in Chapter \hyperref[chap:intro]{Introduction}, through the course 
of this study we tackle two \ac{ma} scenarios. In both of them, the agents have a 
common goal that requires them to cooperate to achieve it. 

Three popular approaches used to solve cooperative \ac{ma} problems are 
Supervised and Unsupervised learning, and \ac{rl}, which are distinguished 
according to the type of feedback provided to the agent: the target output in the 
first case, no feedback is provided in the second, and reward based on the learned 
output in the last one \cite[][]{panait2005cooperative}.
However, in reward-based methods, it is notoriously hard to design a suitable 
reward function, able to lead to the desired behaviour in all possible scenarios, 
even for complex tasks \cite[][]{hadfield2017inverse}.
\ac{il} methods can be used to overcome this problem by learning a policy from 
expert demonstrations without access to a reward signal \cite[][]{song2018multi}.

\section{Imitation learning}
\label{sec:imitlrng}
\acreset{rl}
\acreset{il}
\acreset{nn}

\ac{il} is a class of methods that has been successfully applied to a wide range of 
domains in robotics, for example, autonomous driving.
Unlike reward-based methods, \ac{il} acquires skills by directly observing 
demonstrations of the desired behaviour in order to provide a learning signal to 
the agents \cite[][]{zhang2018deep}.

A typical approach to \ac{il}, also called Behavioural Cloning 
\cite[][]{torabi2018behavioral}, is a supervised technique that consists in 
collecting a certain amount of data, corresponding to a sequence of encountered 
observations and actions performed by a teacher agent which acts according to 
an unknown policy in order to achieve a certain goal.
Then, the demonstrations of the expert’s behaviour are used to learn a controller, 
by training usually a classifier or regressor, that predict behaviour to correctly 
achieve the same goal in a certain environment \cite[][]{ross2011reduction}.
The machine learning model should be able to learn how to extract relevant 
information from the data provided to it, directly learning a mapping from 
observations to actions.

Instead, an unsupervised approach collects only the data that correspond to a 
sequence of encountered observations performed by a teacher agent, and the 
learned controller should be able to infer the correspondence from observations 
to actions and find a way to accomplish the same task \cite[][]{stadie2017third}.
In this case, learning a good model is more challenging since the coordination, 
that is implicit in the demonstrations, has to be inferred as a latent variable 
\cite[][]{le2017coordinated}.

\section{Approaches}
\label{sec:approaches}
Given the two scenarios presented in Chapter \hyperref[chap:intro]{Introduction}, 
distributing the robots in space such that they stand at equal distance from each 
other and colouring the robots in space depending depending on their position 
with respect to the others in the group, through the course of this study we tackle 
two \ac{ma} scenarios. In both of them, the agents have a common goal that 
requires them to cooperate to achieve it. 

While both are excellent examples of distributed tasks, they have an important 
difference: the first problem can be solved without using communication, while 
for the second one it is necessary.
A full explanation of how communication works for Thymio II is covered in 
Section \ref{subsec:thymiocomm}.

\subsection{Distributed approach}
\label{subsec:dist}

As stated before, the first task can be accomplished with a distributed approach 
without communication.
The agents share the same goal: arrange themselves uniformly along the line 
between the two “dead” robots, in such a way they stand at equal distances from 
each other.
This problem represents an example of a cooperative task, for this reason, it will 
be desirable for the agents to cooperate \cite[][]{barrett2017making}. 

On the one hand, when communication is not possible, they can achieve their 
goal without directly interact with each other. 
As stated by Holland in \emph{Multiagent systems: Lessons from social insects and 
collective robotics} \cite[][]{holland1996multiagent}, since the agents exist in the 
same environment, they can affect each other indirectly in several ways.
For instance, they can be sensed from the other robot’s or they can even change 
the state of an agent by applying a force on it, for example, by colliding with it.
The work of Grassè about Stigmergy Theory provides additional details about 
active and passive stigmergy \cite[][]{grasse1959reconstruction}.

On the other hand, the difficulty of the problem we consider is that the agent 
does not have full knowledge of its mates’ behaviours.
Although a communication protocol is not necessary, allowing an explicit 
exchange of messages between the agents may nonetheless increase agent 
performance: they use the exchange of message in order to coordinate more 
effectively and distribute more accurately, finally reaching a more efficient 
solution \cite[][]{panait2005cooperative}.

\subsection{Distributed approach with communication}
\label{subsec:comm}
In the second task also, the agents share a common goal: assuming that they are 
divided into groups, which are unknown to them, their objective is to determine  
their group membership and to colour themselves accordingly. 
For this kind of problems, allow explicit communication is not a plus but a 
necessity.

An important aspect that needs to be considered is the decision of what to 
communicate and when, so that no issues arise.

Regarding the communication content, the agents can, for example, inform the 
others of their current state by sharing the sensor readings, or even information 
about the past \cite[][]{guestrin2002coordinated, panait2005cooperative}.
For this purpose, we used an unsupervised approach in which we do not have to 
specify the communication content, which instead has to be inferred by the 
network as a latent variable.

Of great importance is also the moment in which transmit a message. In fact, if the 
communication is delayed, it can become useless or even cause unwanted 
behaviour \cite[][]{stone2000multiagent}.
As we have already said in Section \ref{subsec:thymiocomm}, each robot transmits 
a message every $0.1$s and likewise receives one for each of the sensors. In 
our case, we expect each agent to receive two communications, one from each of 
its respective neighbours. With real robots, but also in simulation, what we want to 
attain is a synchronous communication update protocol. Formally, each robot 
$n$, given the observations at time $t$, that corresponds to the sensor readings 
$S_n(t)$, and the communications at time $t-1$, in particular $C_{n-1}(t-1)$ 
and $C_{n+1}(t-1)$, calculates the control $V_n(t)$ and the message to 
transmit $C_n(t)$ at time $t$. 
Adopting this technique, it is not important to keep track of the order of the 
robots and it is as if the agents operate simultaneously. 
Ideally, the frequency of updates must be lower than that with which the 
robots exchange messages, however, it can happen that due to delays or 
noise in the sensor readings the communication of some robots is not 
received or transmitted. In this case, the array is not updated and the last 
received message is kept instead, without causing undesired behaviour but simply 
a slowdown.

\section{Data collection}
\label{sec:dataset}

In this work, $N$ robots, all oriented in the same direction, are initially 
randomly placed along the $x$-axis, avoiding collisions and in such a way the 
average gap among them is included in the proximity sensors' ranges. 
All agents act in collaboration to achieve a common goal, except the first and last 
in the row that behave like walls.

Each robot can be considered as a point on the plane, formally described by a 
homogeneous vector with respect to the world coordinate frame $W$, obtained 
multiplying the homogeneous vector of the point w.r.t. the robot coordinate 
frame $A$, by a homogeneous transformation. The relative pose $A$ of each 
agent is identified by a $3 \times 3$ matrix $\mathbf{T}$, with respect to the 
world reference frame $W$. 
\begin{Equation}[!htb]
	\centering
	\begin{equation}
	{^W\!\xi_A} = {^W\!\mathbf{T}_A} 
	=
	\begin{pmatrix}
	^W\!\mathbf{R}_A & ^W\!\mathbf{t}_A\\
	0, 0 & 1
	\end{pmatrix}
	=
	\begin{pmatrix}
	\cos \theta & - \sin \theta & t_x\\
	\sin \theta & \cos \theta & t_y\\
	0 & 0 & 1
	\end{pmatrix}
	\end{equation}
	\caption[Homogeneous transformation matrix.]{The homogeneous 
		transformation matrix, 	$^W\!\mathbf{T}_A$, includes $^W\!\mathbf{R}_A$, 
		a 
		$2 \times 2$ rotation matrix and $^W\!\mathbf{t}_A$, a $2 \times 1$ 
		translation vector.}
	\label{eq:hommatrix}
\end{Equation}
However, since they are arranged on a line, the environment can be considered 
\ac{1d}, hence, the $y$ coordinate is equal to $0$ and also the orientation angle 
$\theta$ must be zero as all the agents are oriented as the world frame. 
Moreover, we can consider the agents as holonomic, since their movements are 
limited to only one dimension. This premise simplifies our system, in which 
consequently we have to keep into account only geometric constraints and not
kinematic.

Of fundamental importance is the approach adopted for the generation of the 
starting positions of the robots.
The initial configurations need to be randomly generated, verifying that there is 
no bias towards those close to the target.
In particular, once established the number of agents to spawn and the average 
gap between them, a vector containing samples, each representing a random 
gap in $\mathbb{R}$, is drawn from a uniform distribution in the interval $[0, 
2*\mathtt{avg\_gap})$. 
The length of the Thymio, that is $10.9$cm, is added to each gap, then the 
final positions are obtained by returning the cumulative sum of the elements in 
the generated vector. 

Another premise regards the sensors of the Thymio: as introduced in Section 
\ref{subsec:enkisensors}, we have available \texttt{prox\_values} and 
\texttt{prox\_comm\_events.intensities}. Before being used, the 
\texttt{prox\_comm\_events.intensities} should be flattened to obtain a single 
array containing the intensities of all recorded events. 
To do so, we decided to create a new array, called \texttt{prox\_comm}, by 
keeping for each element only the value with the maximum intensity among the 
possible values in the corresponding position of the original vectors – for this 
purpose maximum $2$.
In addition, we define another variable, named \texttt{all\_sensors}, which is an 
array of $14$ intensities resulting from the combination of the 
\texttt{prox\_values} and \texttt{prox\_comm} vectors.

The data that we use to train our machine learning models are generated through 
Enki, the simulator introduced in Section \ref{sec:enki}.
In particular, two datasets containing $1000$ simulation runs each are built using 
the omniscient and the manual controllers, that will be explained in Sections 
\ref{subsec:expert} and \ref{subsec:manual}. 
Each run, that differs from the others for the initial positions of the agents, sets up 
a world containing $N$ Thymio. 
In particular, for all the simulations the number of agents $N$ is chosen randomly 
within the range $[5, 10]$, and the \texttt{avg\_gap}, that is the average distance 
among the robot in the final configuration of the run, that can be in the range 
$[5, 24]$.
A simulation run is stopped either immediately after all the robots reach the 
target pose, with a certain tolerance, or after $4$s ($40$ time steps).
At each time step, all the useful information regarding the agents is stored in 
the dataset, such as the sensor readings, the pose of the robot, the motors target, 
the communication transmitted and received and its colour.

All the original datasets are shuffled, based on the single run, in order to improve 
the generalisation on the samples, and then split the resulting collection into 
train, validation and test sets, containing respectively 60-20-20\% of the data.
The dataset generated using the expert controller is the one used to train the 
networks, while the one generated with the manual controller is used as a 
baseline for the comparison with the learned model.

\section{Controllers}
\label{sec:controllersmodel}

\acreset{il}
\acreset{mas}

In a \ac{mas}, each agent can perceive the environment through sensors 
acquiring a total or partial knowledge of it. The observations can be used by a 
controller, the key component of the system introduced in Section 
\ref{subsec:control}, together with the current state of the agents, to determine 
actions, draw inferences and finally solve tasks. 

For the two scenarios that we consider in this study, the state of the agent is the 
combination of four elements: its position on the $x$-axis, the observations, i.e.,
the distances from neighbours recorded in the sensor readings, communication 
messages, one transmitted to the two nearest neighbours, one on the left and the 
other on the right, and two received from peer robots, and finally its colour.
Instead, the set of actions that agents can perform are different depending on 
the task: in the first scenario the agents move forward and backwards along the 
x-axis, therefore the set includes the range of velocity that they can assume; in the 
second one, the agents  can turn on their top RGB LED in red or blu, so the set this 
time is composed by the two possible colours.

In an \ac{il} setting, there are two main controllers involved: an omniscient 
controller, which decide the best action exploiting its perfect knowledge of the 
state of the system, and a learned controller, which imitate the behaviour of the 
expert.
Undoubtedly, one of the main advantages of adopting a \ac{ml} model to solve 
these tasks is that the algorithm must learn how to extract relevant information 
from the data it receives, sidestepping the difficulty of manually implementing the 
perception part.
However, for the tasks that we are going to face, we introduce three controllers: in 
addition to the two mentioned before, we also use a manual controller, which can 
observe only parts of the system. As a consequence, its decisions do not depend 
on the state of the whole swarm \cite[][]{vsovsic2016inverse}.

\subsection{Expert controller}
\label{subsec:expert}

As just described, the first element involved in an imitation learning problem is 
the omniscient controller, also called expert. 
This is a centralised controller that perceives the environment and observes the 
state and the observations of all agents, obtaining a global knowledge of the state 
of the system. In this way, it can use all the information to decide the best action 
to perform for all the agents. 

As the goals to be achieved vary, the controllers should act differently. For this 
reason, in the following paragraphs we define the approaches used for the 
implementation of the controllers in the two scenarios.

\subsubsection{Task 1: Distributing the robots in space}
\label{subsubsec:experttask1}

In the first scenario, the omniscient controller, based on the current poses of the 
robots, moves the agents at a certain speed to reach the target positions. In 
particular, the linear velocity of each agent is computed as a ``signed distance`` 
between the current and the goal position of the robot, along its theta. 

Formally, given the current pose, defined by the triple $(x, y, \theta)$ and the 
target pose $(\overline x, \overline y, \overline \theta)$, the signed distance $d$ 
is computed as follow:
\begin{Equation}[!htb]
	\centering
	\begin{equation}
	d = \left(\overline x * \cos (\theta) + \overline y * \sin (\theta)\right) -
	\left( x * \cos (\theta) + y * \sin (\theta)\right)
	\end{equation}
	\caption[Signed distance function.]{Function used to compute the ``signed 
		distance'' between the current and the goal position of a robot.}
	\label{eq:signeddist}
\end{Equation}

\noindent
To obtain the final velocity of the agent, this quantity is multiplied by a constant, 
we choose $10$ to keep the controller as fast as possible, and then clipped to the 
maximum value supported by Thymio II, $16.6$cm/s.

This controller can be considered as a variant of a Bang Bang controller, 
introduced in Section \ref{subsubsec:bangbang}, since the optimal controller 
moves the robot at maximum speed towards the target unless the target is closer 
than \texttt{control\_step\_duration} $\times$ \texttt{maximum\_speed}. In this 
case, the agent is moved slower than maximum speed so that at the end of the 
time step it is located exactly at the target.

\subsubsection{Task 2: Colouring the robots in space}
%FIXME
In this scenario, the omniscient controller, based on the current poses of each 
robot, is able to determine the order of the agents and turn on their top \ac{led}  
in one time step. For this reason, we decided to use the same dataset obtained for 
the previous task, but adding to it the colour of the robot, in order to provide the 
network with examples consisting of multiple time steps from which it can learn.

\subsection{Manual controller}
\label{subsec:manual}

Unfortunately, centralised solutions for distributed problems are not feasible in 
real situations since agents do not have access to their states but only to their local 
observations. The global state of the system, accessible to a centralised controller, 
in this case, is hidden from each agent who therefore cannot understand its 
absolute position.
Instead, we are interested in situations in which is used a local controller to decide 
the next action, based on the individual agent's observations, or even cases in 
which the programming part of the controller is automated.
In fact, the main purpose of this controller is to draw conclusions about the 
quality of the controller learned.

As before, to different goals to be achieved correspond different controllers 
that should act differently. For this reason, in the following paragraphs we define 
the approaches used for the implementation of the controllers in the two 
scenarios.

\subsubsection{Task 1: Distributing the robots in space}
\label{subsubsec:manualtask1}
\acreset{pid}
In the first scenario, the controller is in charge of moving the robots towards the 
target by minimising the difference between the values recorded by the front and 
rear sensors, trying to maintain the maximum achievable speed.

For each agent, the controller is the same and, if given an identical set of 
observations as input, likewise, the outputs will be equivalent.

The kind of controller we decided to implement for this purpose is a Proportional 
(P) controller, a particular variant of the \ac{pid} controller with only the $K_p$ 
term, described in detail in Section \ref{subsubsec:pid}. 
In particular, the value of the proportional gain has been tuned to yield 
satisfactory performance so that the system is stable, as shown in Figure 
\ref{fig:pid}. 

\begin{figure}[htb]
	\centering
	\includegraphics[width=.5\textwidth]{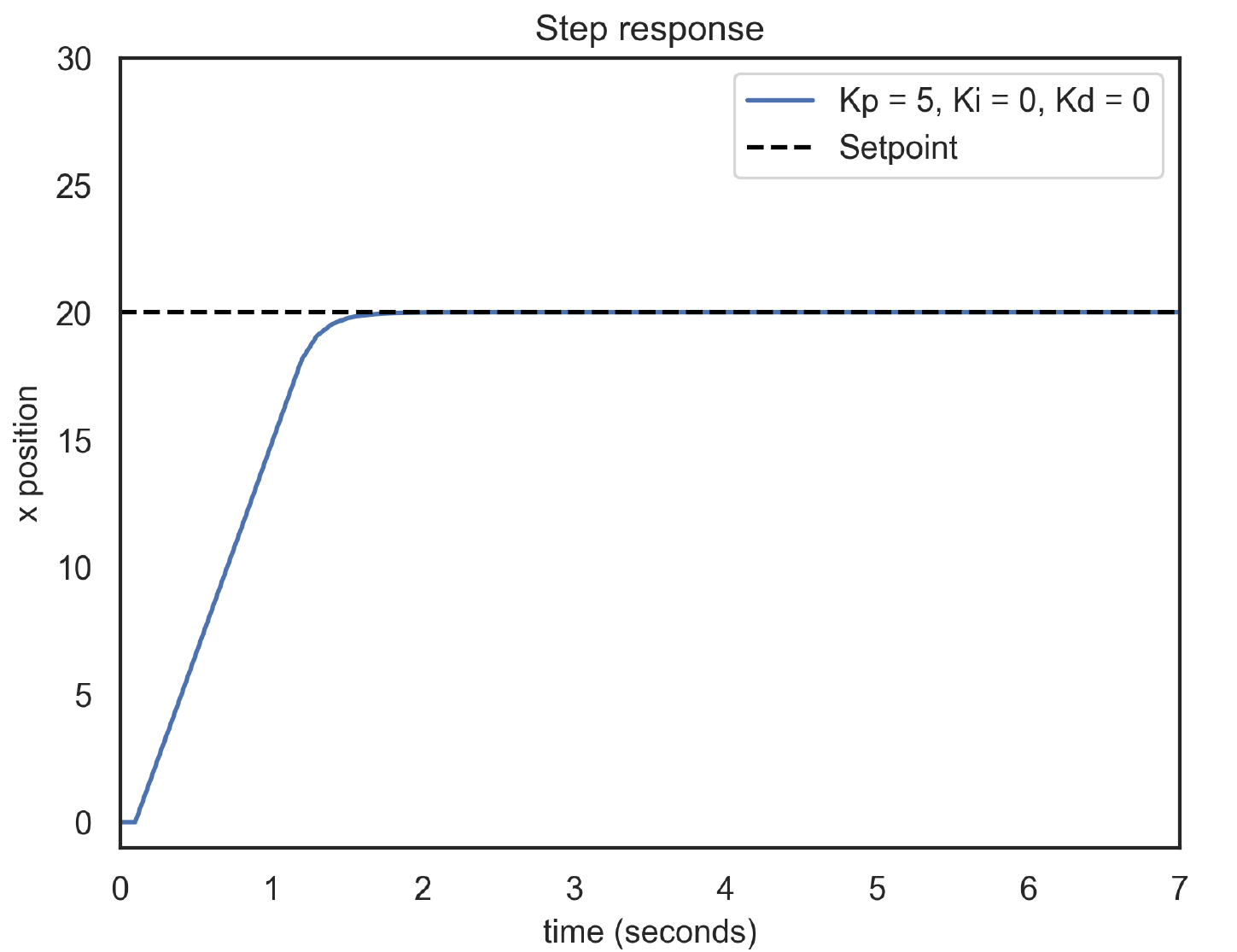}
	\caption[Step response of the proportinal PID controller.]{Visualisation of the 
		step-response of a P controller with proportional 
		gain $5$.}
	\label{fig:pid}
\end{figure}

It is important to notice that the speed returned by the controller is used to set the 
\texttt{motor\_\{left, right\}\_target}, both with the same value, in order to move 
the robots straight ahead. Moreover, the first and the last robots of the line, 
whose sensors never receive a response respectively from the back and from the 
front, never move.

\subsubsection{Task 2: Colouring the robots in space}
In this scenario, the robots observations, in particular, the sensor readings, do not 
provide useful information about the order of the agents, therefore they are not 
considered to accomplish this task.
On the other hand, if an omniscient controller is not employed, it is impossible to 
solve the problem without using communication, since it is the only way for the 
agents to understand their ordering.

Thus, by initially making all robots transmit the same value, i.e.,$0$, we are able 
to establish which are the first and the last robots in the row, or those that do not 
receive any communication respectively from back and front. 
These two agents can at this point start the actual communication by transmitting 
the value they received, increased by $1$, that is $1$. 
The following robots will in turn transmit the value they have received, increased 
by one. Since the messages received by each of them are two, the agents will in a 
sense, learn to count in order to understand which is the correct value to 
transmit.

The protocol used to decide the communication and the colour, which also 
depends on the amount of robots $N$, or if there are even or odd numbers, is 
shown in Listing \ref{lst:manualtask2}. The colour of each agent in the initial 
configuration is randomly chosen between the two possible colours, red and blue.

\medskip
\begin{python}
	c_left, c_right = get_received_communication(state)
	
	if N % 2 == 1:  # if the number of robots is odd
	
	# Case 1: no communication received from left
	if c_left == 0:
	if c_right > N // 2:
	# the agent is in the first half of the row, so its colour is blue
	message = c_right - 1
	colour = 1
	elif c_right == N // 2:
	# the agent is the central one, so its colour is blue
	message = c_right + 1
	colour = 1
	else:
	# the agent is in the second half of the row, so its colour is red
	message = c_right + 1
	colour = 0
	
	# Case 2: no communication received from right
	elif c_right == 0:
	if c_left > N // 2:
	# the agent is in the second half of the row, so its colour is red
	message = c_left - 1
	colour = 0
	elif c_left == N // 2:
	# the agent is the central one, so its colour is blue
	message = c_left + 1
	colour = 1
	else:
	# the agent is in the first half of the row, so its colour is blue
	message = c_left+ 1
	colour = 1
	
	# Case 3: communication received from both sides
	else:
	if c_left > c_right:
	# the agent is in the second half of the row, so its colour is red
	message = c_right + 1
	colour = 0
	else:
	# the agent is in the first half of the row, so its colour is blue
	message = c_left + 1
	colour = 1
	
	
	elif self.N % 2 == 0:  # if the number of robots is even
	
	# Case 1: no communication received from left
	if c_left == 0:
	if c_right > N // 2:
	# the agent is in the first half of the row, so its colour is blue
	# the situation is ambiguous the message to transmit could be c_right or 
	# even c_right - 1
	message = c_right
	colour = 1
	else:
	# the agent is in the second half of the row, so its colour is red
	message = c_right + 1
	colour = 0
	
	# Case 2: no communication received from right
	elif c_right == 0:
	if c_left < N // 2:
	# the agent is in the first half of the row, so its colour is blue
	message = c_left + 1
	colour = 1
	else:
	# the agent is in the second half of the row, so its colour is red
	# the situation is ambiguous the message to transmit could be c_left or 
	# even c_left - 1
	message = c_left
	colour = 0
	
	# Case 3: communication received from both sides
	else:
	if c_left > c_right:
	# the agent is in the second half of the row, so its colour is red
	message = c_right + 1
	colour = 0
	elif c_left < c_right:
	# the agent is in the first half of the row, so its colour is blue
	message = c_left + 1
	colour = 1
	else:
	# the agent is in the second half of the row, so its colour is red
	message = c_left
	colour = 0
\end{python}

\begin{lstlisting}[frame=none,caption={Protocol used by the manual controller 
to decide, for each robot, the message to transmit and the colour.}, 
label=lst:manualtask2]
\end{lstlisting}

\subsection{Learned controller}
\label{subsec:learned}

Usually, to train a network that learns a controller, the states and the actions, 
provided by an expert, need to be observable. For this study, we have decided to 
use the observations of the robots instead of the state, i.e.,the positions. 
The reason behind this decision is that in most environments, agents are never 
actually exposed to the full state of this system. Instead, they receive partial 
observations, often local or incomplete. In addition, it is frequently too expensive 
to provide the agent with the full state of the system, and sometimes it is not even 
clear how to represent it \cite[][]{ml-agents}.

As the goals to be achieved vary, the controllers should act differently. For this 
reason, we define distinct approaches for the implementation of the controllers 
in the two scenarios.
Regarding the first task, we consider two different networks: one distributed that 
act in a supervised way, and one that, in addition to predicting the control output, 
infers a communication protocol between the agents. 
In the second task, we trained one network that predicts the colour output and in 
addition, as before, infers the communication between the robots.

\subsubsection{Task 1: Distributing the robots in space without using 
communication}

Using the data collected through the simulator using the expert controller, it is 
possible to train a very simple ``distributed network'' that takes as input an array 
containing the response values of the sensors – which can be either 
\texttt{prox\_values}, \texttt{prox\_comm} or \texttt{all\_sensors} – and produces 
as output an array containing one float that represents the speed of the wheels, 
which is assumed to be the same both right and left.

The training dataset then contains a fixed number of simulation runs, each 
composed of a variable quantity of time steps. It is important to notice that 
for this approach, unlike the one with communication, it is neither necessary to 
preserve the order of the sequence of time steps, nor to know the exact number 
of agents in the simulation since the network input is the sensing of a single robot.

For this reason, the model is independent of the number of agents and 
consequently it is possible to prove its generalisation capacity, regardless the 
number of robots, by training the networks first on datasets each with a different 
but fixed value of $N$ and then evaluating them on simulations with a variable 
$N$.
It is easy to show that, although the value of $N$ changes, the network structure 
does not, as it is sufficient during the input preprocessing to change the 
dimension of the input in such a way that all the tensors have the same length, 
fixed at the maximum possible value of $N$, padding those tensors with a lower 
number of agents.

The architecture of the network, in Figure \ref{fig:singlenetdistributed1}, is 
straightforward: there are three linear layers of size $\langle\mathtt{input\_size}, 
10\rangle$,  $\langle 10, 10\rangle$ and $\langle 10, 1\rangle$, where 
\texttt{input\_size} is the shape of the sensing that can be either $7$ or $14$.
\begin{figure}[htb]
	\centering
	\begin{subfigure}[h]{0.495\textwidth}
		\centering
		\includegraphics[width=.8\textwidth]{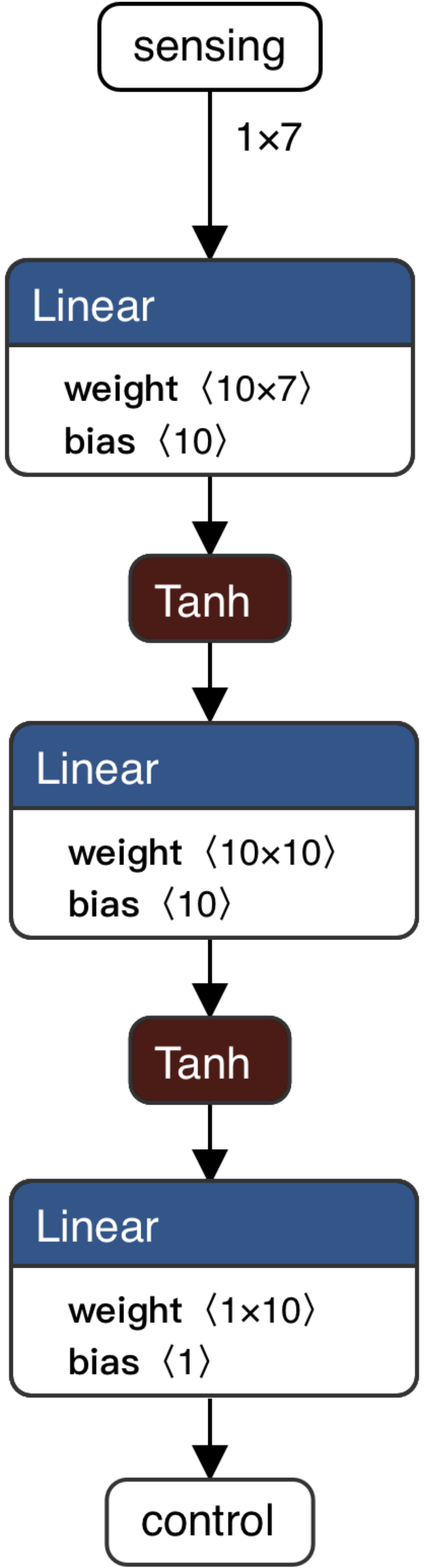}%
		\caption{Network with $7$ input sensing.}
		\label{fig:singlenet-d7distributed1}
	\end{subfigure}
	\hfill
	\begin{subfigure}[h]{0.495\textwidth}
		\centering
		\includegraphics[width=.8\textwidth]{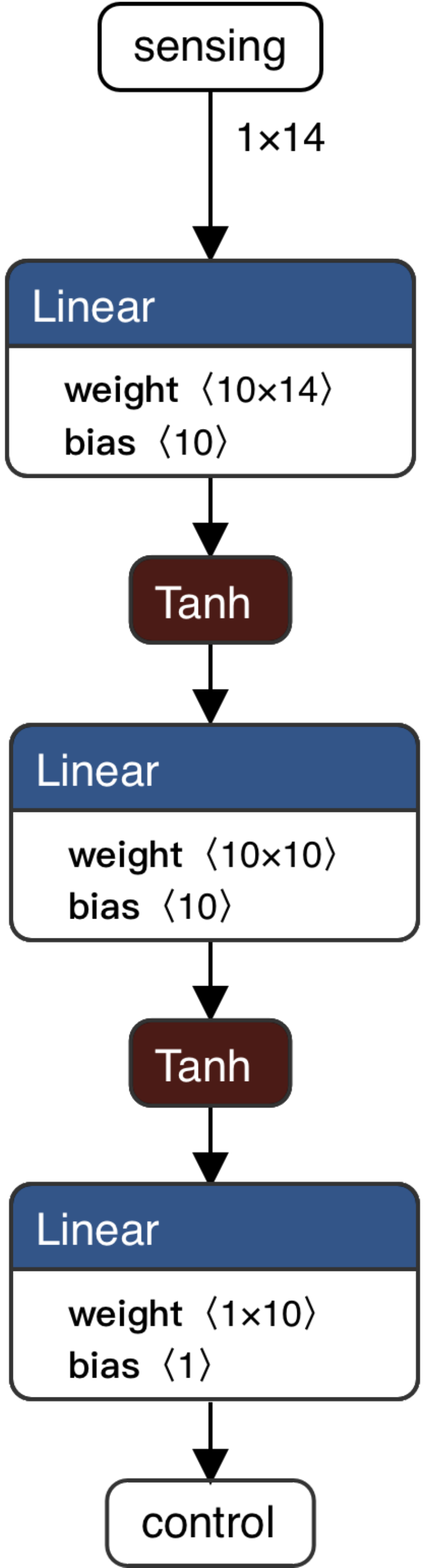}
		\caption{Network with $14$ input sensing.}
		\label{fig:singlenet-d14distributed1}
	\end{subfigure}
	\caption[Network architectures for the distributed approach.]{Visualisation of 
		the network architecture chosen for the distributed approach.}
	\label{fig:singlenetdistributed1}
\end{figure}

The Tanh non-linear activation function introduced in Section 
\ref{subsec:activationfun}, is applied to the first and second layer. 

As optimiser, we chose Adam, introduced in Section \ref{subsec:optimiser}, 
implemented in the \texttt{torch.optim} package, with a learning rate of $0.01$. 

Instead of performing gradient descent on the entire dataset, the training set is 
split in mini-batches of size $100$. In this way, an approximation of the gradient 
is produced, which makes the algorithm faster and at the same time, for 
sufficiently large batches, the result is indistinguishable.
Gradient descent algorithms are susceptible to ``getting stuck'' in local minima.
Mini-batches shuffle facilitate to avoid this problem by enabling the gradient to 
``bounce'' out of eventual local minimum, making it more variable by exploiting 
randomness, thereby helping convergence \cite[][]{meng2019convergence}.

All the models are trained for $50$ epochs and evaluated using the \ac{mse} loss 
function, defined in Section \ref{subsec:lossfunctions}, implemented in the 
\texttt{torch.nn} package.

\subsubsection{Task 1: Distributing the robots in space using communication}
\label{subsubsec:task1comm}

An alternative to the previous approach involves training a distributed network 
that also exploits a communication protocol between agents to decide the output 
control more reliably. 
Thus, using the same data collected before, we build a model that at each time 
step takes as input an array containing the response values of the sensors for each 
robot – \texttt{prox\_values}, \texttt{prox\_comm} or \texttt{all\_sensors} – and 
the messages received in the previous time step, communicated by the nearest 
agents (on the left and on the right), and produces 2 floats as output: the control, 
which is the speed of the wheels as before, and the communication, i.e.,the 
message transmitted by the robot to its two neighbouring agents.

Even for this purpose, the model is independent of the number of agents in 
the simulations. Instead, now it is important to keep track of the time steps order 
since the input of the network requires the communication received which 
corresponds to the messages transmitted in the previous time step. To do so, 
preprocessing is applied to the dataset to combine consecutive 
time steps into a set of sequences. Therefore, we divide each simulation in 
sequences of length $2$, composed of two successive observations for each 
robot, using a stride of $1$.   
Accordingly, the shape of the model input has been transformed from $1 \times 
\mathtt{input\_size}$ to $\mathtt{seq\_length} \times \mathtt{N} \times 
\mathtt{input\_size}$, where \texttt{seq\_length} is fixed at $2$, $N$ is variable 
and \texttt{input\_size} can be $7$ or $14$.

\begin{figure}[!htb]
	\centering
	\includegraphics[width=\textwidth]{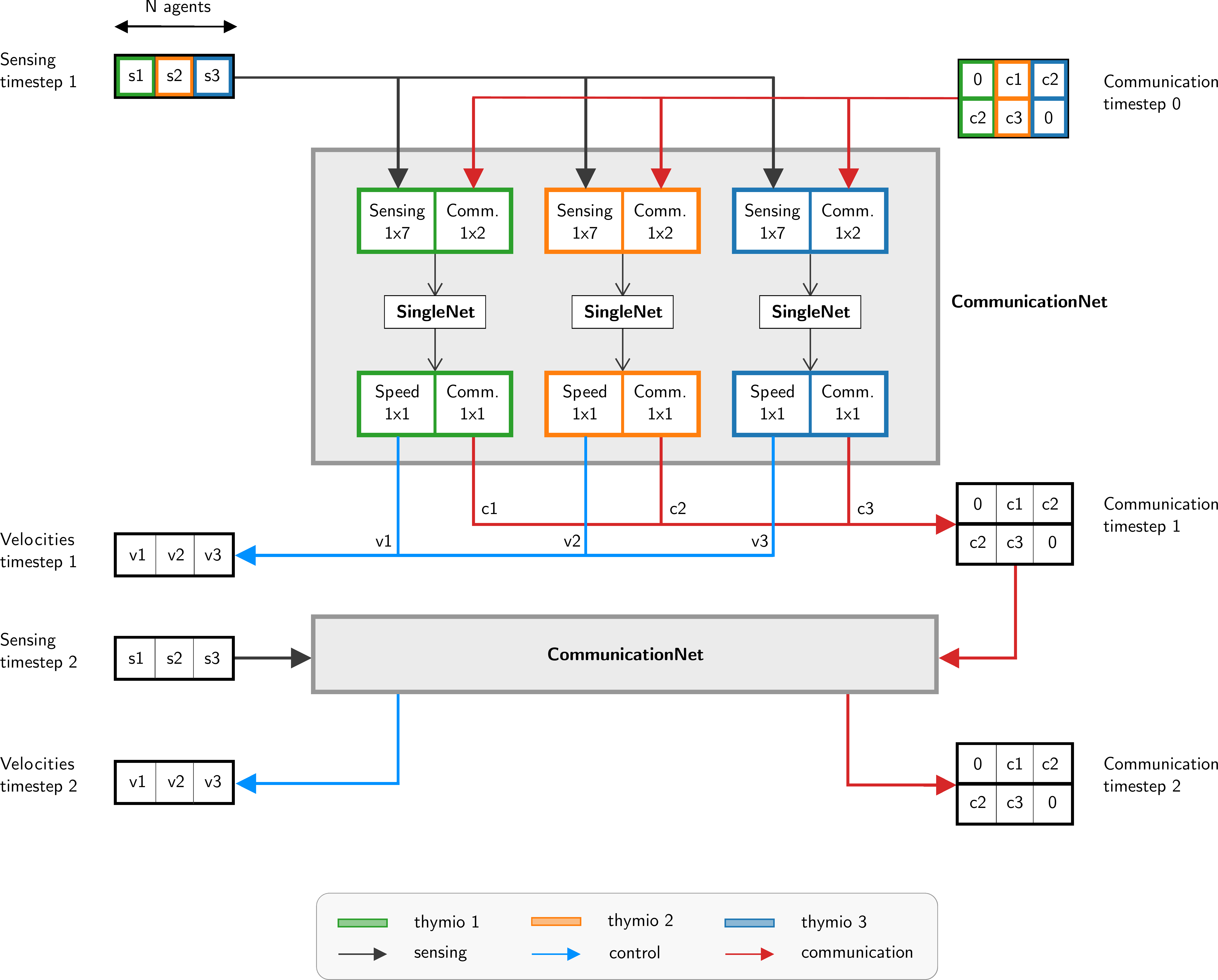}
	\caption[Communication network.]{Visualisation of the forward pass of the 
		communication network with three agents and a sequence composed by two 
		time steps.}
	\label{fig:commnet1}
\end{figure}

It is important to notice that the communication is not in the input since it is not 
contained in the original dataset, instead is treated as a hidden variable to be 
inferred. 
At the beginning of each sequence, there are no previous time steps to consider 
since no messages have been received yet. Therefore, a placeholder is randomly 
initialised, filled with float values in the range $[0, 1]$. 
The size of this array corresponds to the number of agents plus two elements, one 
at the beginning and one at the end of the vector, always set to $0$ since they are 
used to store the fact that the two extreme robots never receive messages 
respectively from the left or from the right. 
The random initialisation of this vector is essential to increase the generalization 
capabilities of the network during its training, showing it different starting 
situations.

As a consequence, we define a recurrent structure of the communication 
network, shown in Figure \ref{fig:commnet1}.
It is composed by two nested modules: in the outer level operates the 
\texttt{CommNet} that handles the sensing of all the agents, while in the inner the 
\texttt{SingleNet} that works on the sensing and the communication received by a 
single agent in a certain time step, producing as output the control and the 
communication to transmit. 
Therefore, this corresponds to a static unroll of a \ac{rnn}.%FIXME \cite[][]{}.

The architecture of the \texttt{SingleNet}, displayed in Figure 
\ref{fig:singlenetcomm1}, is almost the same as the one of the distributed 
model without communication: there are three linear layers each of size 
$\langle\mathtt{input\_size}, 10\rangle$,  $\langle 10, 10\rangle$ and $\langle 
10, 2\rangle$, where \texttt{input\_size} is the sum of the shape of the sensing 
and the two communication values received, one from the left and one from the 
right.

\begin{figure}[!htb]
	\centering
	\begin{subfigure}[h]{0.495\textwidth}
		\centering
		\includegraphics[width=.8\textwidth]{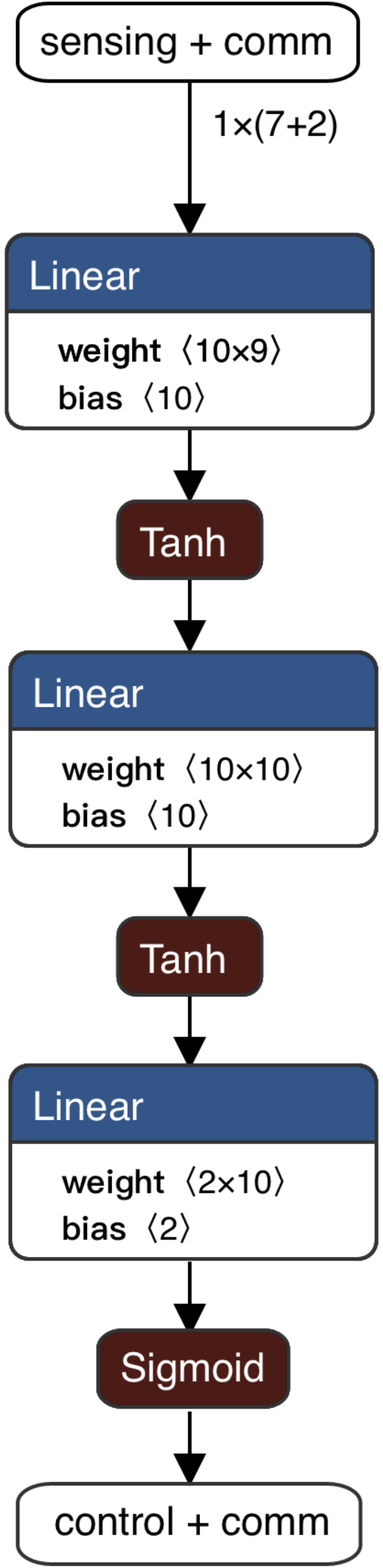}%
		\caption{\texttt{SingleNet} with $7$ input sensing.}
	\end{subfigure}
	\hfill
	\begin{subfigure}[h]{0.495\textwidth}
		\centering
		\includegraphics[width=.8\textwidth]{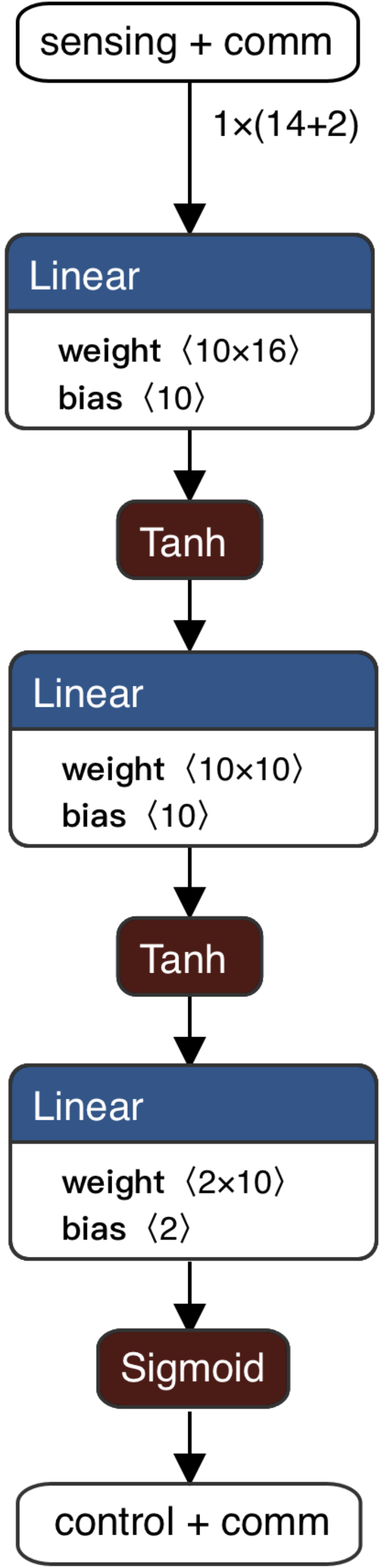}
		\caption{\texttt{SingleNet} with $14$ input sensing.}
	\end{subfigure}
	\caption[Network architectures for the distributed approach with 
	communication.]{Visualisation of the network architecture chosen for the 
		distributed approach with communication in case of 7 or 14 inputs.}
	\label{fig:singlenetcomm1}
\end{figure}

As before, a Tanh non-linear activation function is applied to the first and second 
layer, while a sigmoid, introduced in Section \ref{subsec:activationfun}, is applied 
to the second dimension of the output in order to normalise it in the range $[0, 
1]$.

As before, we use Adam optimiser, addressed in Section \ref{subsec:optimiser}, 
but with a smaller learning rate, $0.001$. 
We split the dataset in mini-batches, this time of size $10$ and then we train 
the models for $500$ epochs. 
Finally, we evaluate the goodness of the predicted control using the \ac{mse} 
loss function, while the communication has to be inferred by the network.
Since the network is fully connected, the communication affects directly the 
output, and consequently, the error minimised. Improving the loss has an impact 
also on the communication latent variable, since the error is propagated through 
%FIXME
the internal network, in order to update the weight during the back-propagation 
step.

\subsubsection{Task 2: Colouring the robots in space}
In this scenario, it is possible to implement a network very similar to the one used 
for the previous task, that is the distributed approach with communication, 
described in Paragraph \ref{subsubsec:task1comm}, but this time ignoring the 
sensors readings.
Thus, using the same data collected before we build a model that at each time 
step takes as input for each robot only the messages received in the previous time 
step, communicated by the nearest agents (on the left and on the right), and 
produces as output an array of 2 floats, the first one is the probability of the agent 
top \ac{led} to be blue and the second is the communication, i.e.,the message to 
be transmitted by the robot.

The communication network, whose structure is shown in Figure 
\ref{fig:commnet2}, is composed by two nested modules: in the outer-level 
operates the \texttt{CommNet} that handle the sensing of all the agents, while in 
the inner-level the \texttt{SingleNet} that works on the communication received 
by a single agent in a certain time step, producing as output the colour and the 
communication to transmit. 
\begin{figure}[H]
	\centering
	\includegraphics[width=.14\textwidth]{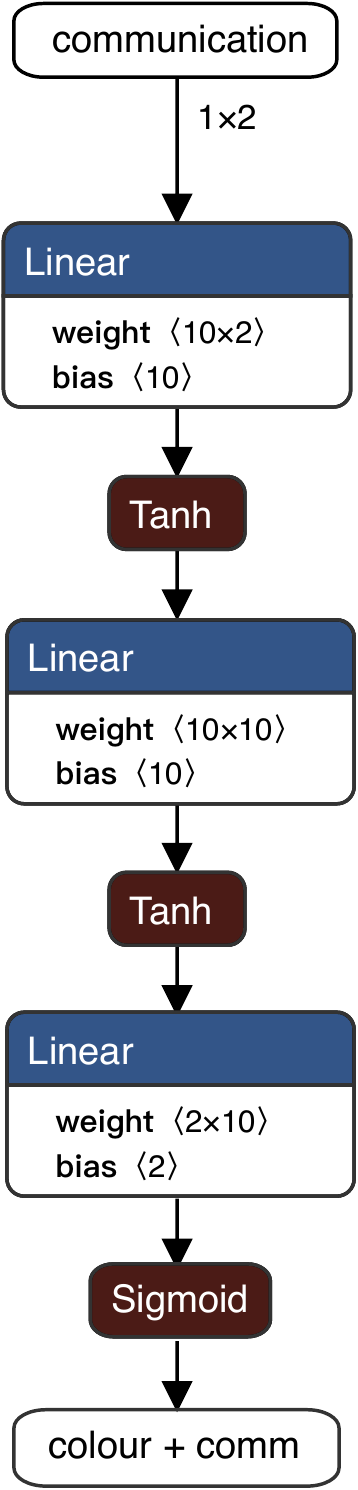}
	\caption[Network architectures for the communication approach.]{Visualisation 
		of the network architecture chosen for the 
		communication approach.}
	\label{fig:singlenetcomm2}
\end{figure}

The \texttt{SingleNet}, displayed in Figure \ref{fig:singlenetcomm2}, is composed 
by three linear layers of size $\langle \mathtt{input\_size}, 10\rangle$,  $\langle 
10, 10\rangle$ and $\langle 10, 2\rangle$, where \texttt{input\_size} 
corresponds to the two communication values received, one from the left and one 
from the right.

\begin{figure}[!htb]
	\centering
	\includegraphics[width=\textwidth]{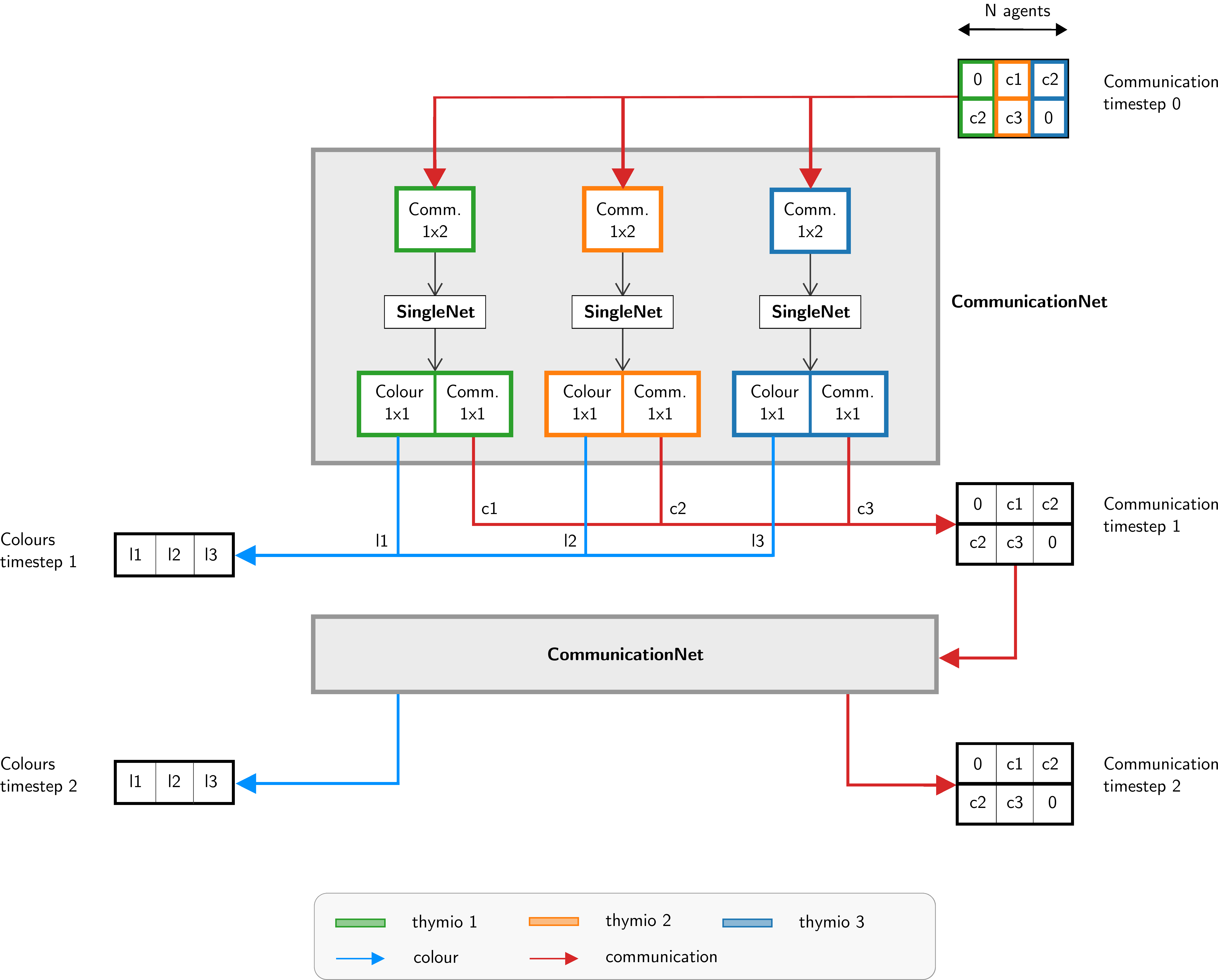}
	\caption[Communication network of the second task.]{Visualisation of the 
		forward pass of the communication network with three agents and a 
		sequence 
		composed by two time steps.}
	\label{fig:commnet2}
\end{figure}

The activation functions used for this purpose are two, and are introduced in 
Section \ref{subsec:activationfun}.
To the first and second layer is applied a Tanh non-linear activation function, 
while a sigmoid to the output, in order to normalise it in the range $[0, 1]$.

As before, we use Adam optimiser but with a smaller learning rate, $0.001$. 
We split the dataset in mini-batches of size $10$ and then we train the models for 
$100$ epochs. 

In order to decide the metric to evaluate the goodness of the prediction, it is 
necessary to analyse the output of the network. As we said, the model returns, in 
addition to the communication, the colour that is actually the probability of the 
agent top \ac{led} to be blue. This means that, when the network produces a 0, 
the probability that the \ac{led} is blue is equal to 0, i.e.,it is red; in the same way, 
1 means that instead, this will be blue. In this way, we define a simple policy 
function that returns the colour blue when the probability is between 0.5 and 1, 
and red otherwise, or when the probability is less than 0.5.
For this reason, instead of using the \ac{mse} loss function, as this is a binary 
classification problem we choose the \ac{bce}, defined in Section 
\ref{subsec:lossfunctions}, implemented in the \texttt{torch.nn} package.
It is important to remember that communication is still inferred as a latent 
variable.

\acresetall
\chapter{Evaluation}
\label{chap:experiments}

In this chapter we present the results of our research. 
For this purpose, we use the tools presented in Chapter \ref{chap:impl} and the 
environment set up provided in Chapter \ref{chap:methods}.
We briefly describe the tasks we face respectively in Sections \ref{sec:task1} and 
\ref{sec:task2}, and then we proceed with the evaluation of the results of the 
experiments.

Throughout the experiments we compare different models, varying the input of 
the network, either \texttt{prox\_values}, \texttt{prox\_comm} or 
\texttt{all\_sensors}, the number of agents and the average gap between them, 
either fixed to a certain value or variable.

\section{Task 1: Distributing the robots in space}
\label{sec:task1}

The first scenario tackles an interesting multi-agent coordination task, 
distributing the robots in space in such a way they stand at equal distance from 
each other. In particular, they have arrange themselves uniformly along the line 
between the two ``dead'' robots.

We focus on two approaches to solve to problem, one distributed that act in a 
supervised way in order to predict the target velocity of the agents, and one that 
in addition to predict the control infers a communication protocol between the 
agents. 

In both cases, we train \acp{dnn} that receive sensor inputs and produce 
commands for the motors, but for the second alternative, the network has an 
addition input – the received communication transmitted by the neighbouring 
agents in the previous time step – and an extra output – the message to be sent.

\subsection{Distributed approach}
\label{subsec:task1-exp-distr}

\subsubsection{Experiment 1: fixed number of agents}
\label{subsubsec:task1-exp-distr-1}

The first group of experiments, summarised in Table \ref{tab:modeln5dist}, 
examines the behaviour of the control learned in the case of the three different 
inputs, \texttt{prox\_values}, \texttt{prox\_comm} or \texttt{all\_sensors}, for a 
number of robots $N$ and an \texttt{avg\_gap} both fixed at $5$ and the second 
chosen between $8$, $13$ and $24$.
\begin{figure}[!htb]
	\centering
	\begin{tabular}{cccc}
		\toprule
		\textbf{Model} \quad & \textbf{\texttt{network\_input}} & 
		\textbf{\texttt{input\_size}} &
		\textbf{\texttt{avg\_gap}} \\
		\midrule
		\texttt{net-d1} 				 & \texttt{prox\_values}	&  $  7$  &  $  8$  \\
		\texttt{net-d2} 				& \texttt{prox\_values}	    &  $  7$  &  $13$ \\
		\texttt{net-d3} 				& \texttt{prox\_values}	    &  $  7$  &  $24$  \\
		\texttt{net-d4} 				 & \texttt{prox\_comm}	  &  $  7$  &  $  8$  \\
		\texttt{net-d5} 				 & \texttt{prox\_comm}	  &  $  7$  &  $13$  \\
		\texttt{net-d6} 				 & \texttt{prox\_comm}	  &  $  7$  &  $24$  \\
		\texttt{net-d7} 				 & \texttt{all\_sensors}	  &  $14$  &  $  8$  \\
		\texttt{net-d8} 				 & \texttt{all\_sensors}	  &  $14$  &  $13$ 	\\
		\texttt{net-d9} 				 & \texttt{all\_sensors}	  &  $14$  &  $24$ 	\\
		\bottomrule
	\end{tabular}
	\captionof{table}[Experiments with $5$ agents (no communication).]{List of the 
	experiments carried out with $5$ agents.}
	\label{tab:modeln5dist}
\end{figure}

First of all we start by showing in Figure \ref{fig:distloss} an overview of the 
models performance in terms of train and validation losses. 
\begin{figure}[!htb]
	\centering
	\includegraphics[width=.85\textwidth]{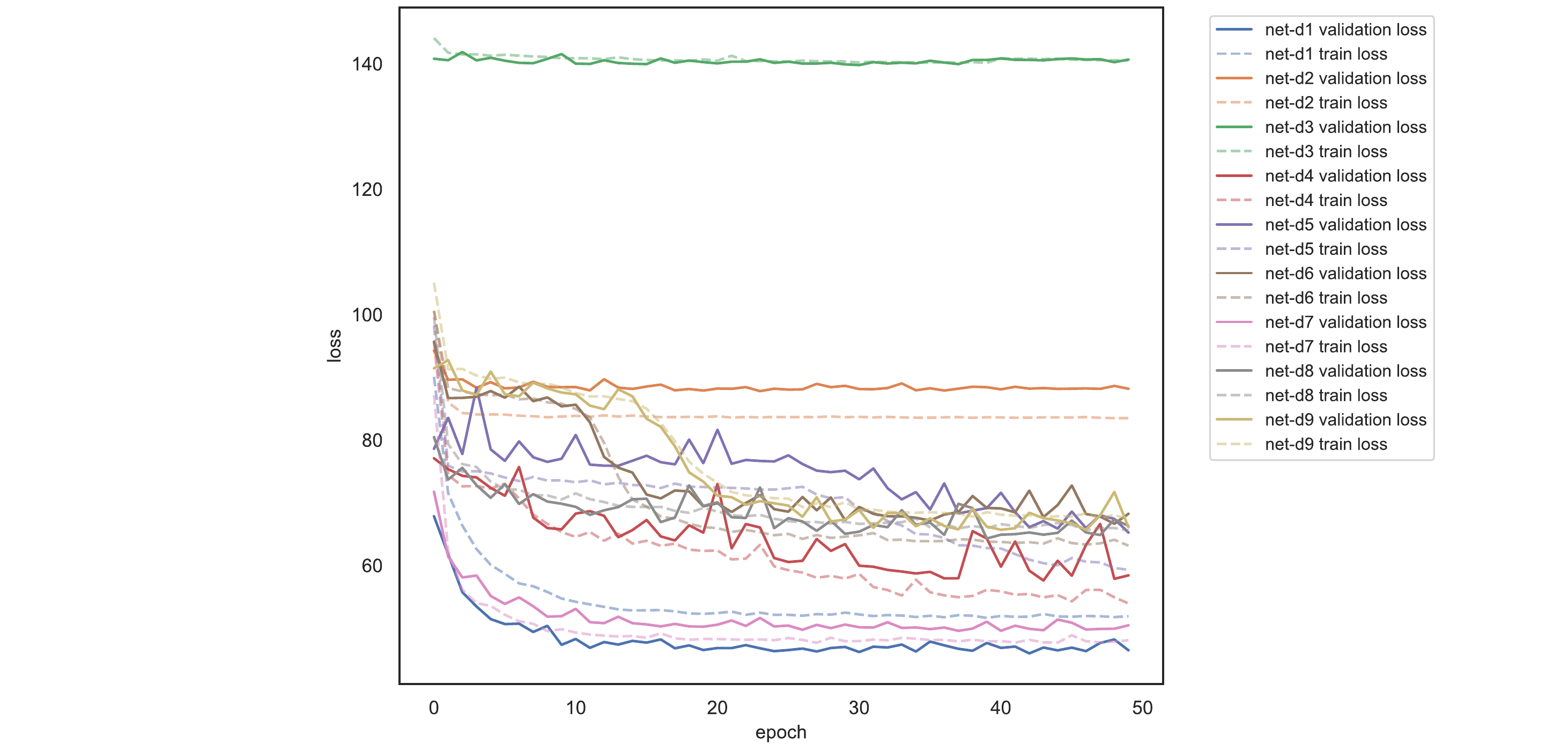}%
	\caption[Comparison of losses of the first set of 
	experiments.]{Comparison of 
	the losses of the models carried out with $5$ agents.}
	\label{fig:distloss}
\end{figure}
It is immediately evident that, in case of \texttt{prox\_values} inputs, the 
experiment performed with an \texttt{avg\_gap} of $24$ is not remarkable since 
the gap exceeds the maximal range of the sensor. In fact, from this analysis we 
generally expect a more stable behaviour using both types of input together, i.e.,
\texttt{all\_sensors}, as they are able to perform with both small and large gaps.

\paragraph*{Results using \texttt{prox\_values} input}
\label{para:1}
We start the analysis by exploring the results of the experiments obtained 
using the \texttt{prox\_values} readings alone as input of the network, continuing 
the with \texttt{prox\_comm} and concluding with \texttt{all\_sensors}.

The performance of \texttt{net-d1} are shown in the following images. In 
particular, in Figure \ref{fig:net-d1r2} is visualised a comparison of the \ac{r2}, 
or coefficient of determination, of the manual and the learned controllers, on the 
validation set.
This score function evidences how well the regression predictions approximate 
the real data points (groundtruth). Since a model which perfectly predicts the data 
has a score of $1$, we assume that a higher score corresponds to a model that 
performs better.
\begin{figure}[!htb]
	\centering
	\begin{subfigure}[h]{0.49\textwidth}
		\centering
		\includegraphics[width=\textwidth]{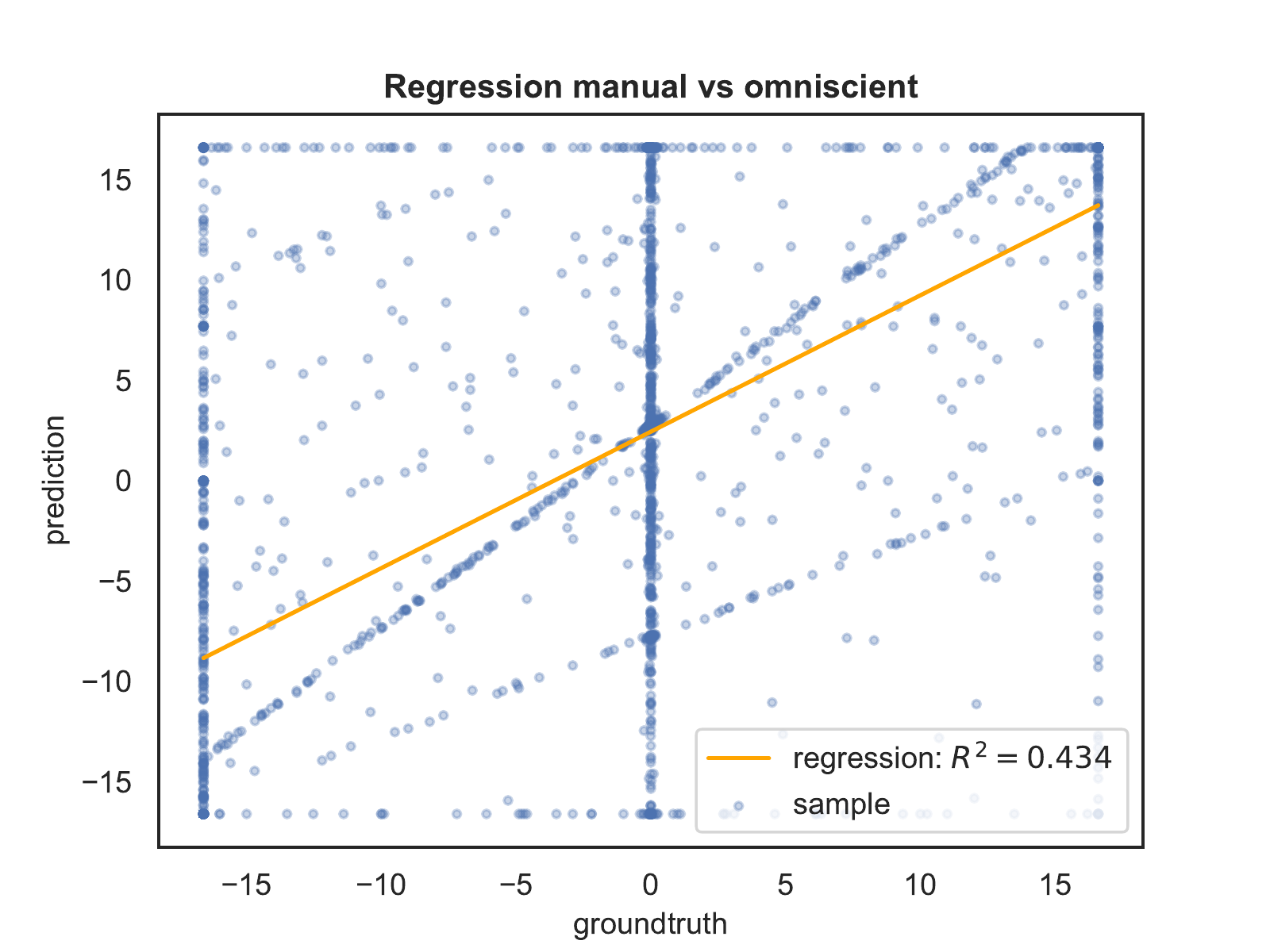}%
	\end{subfigure}
	\hfill
	\begin{subfigure}[h]{0.49\textwidth}
		\centering
		\includegraphics[width=\textwidth]{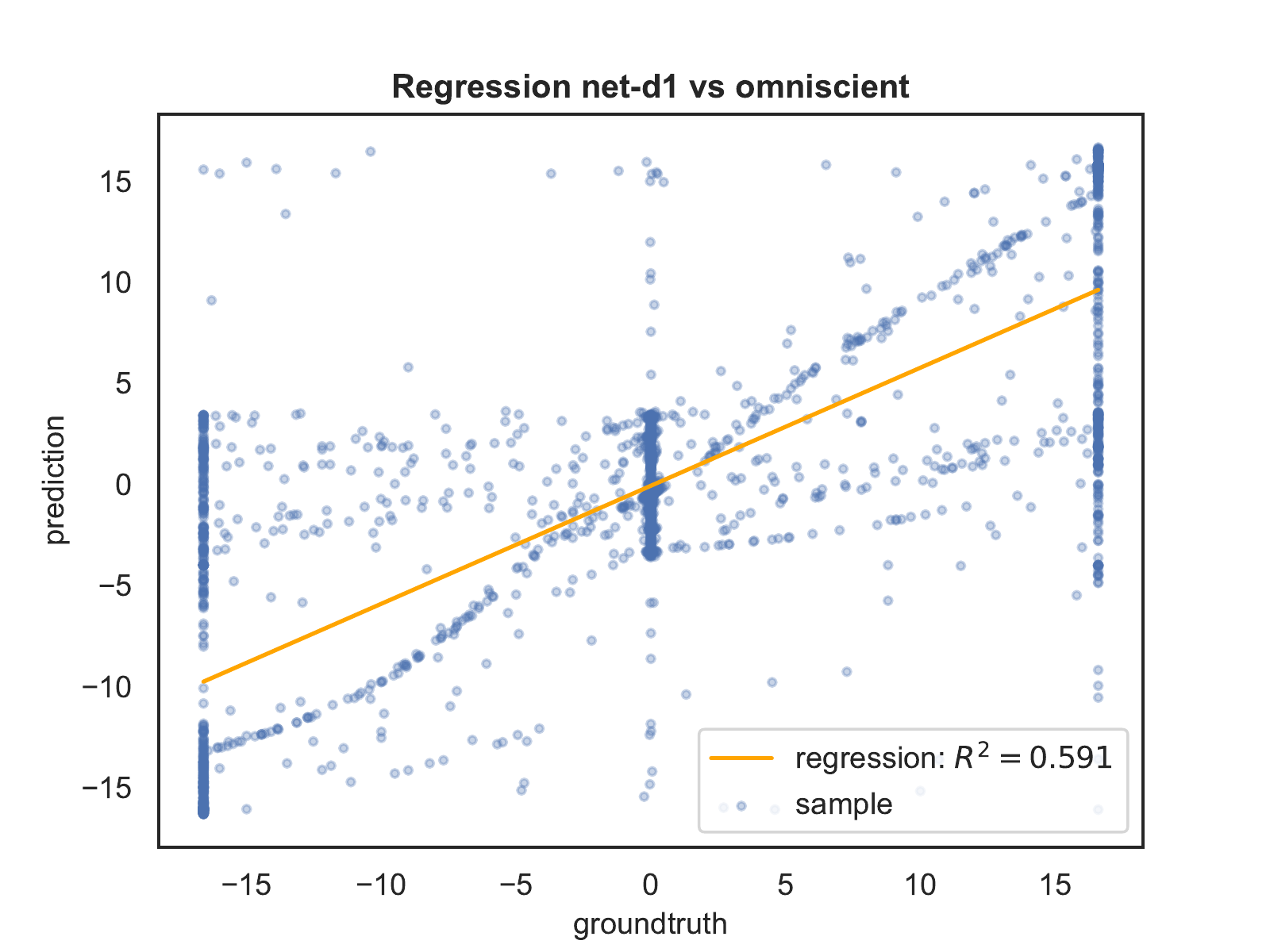}
	\end{subfigure}
	\caption[Evaluation of the \ac{r2} coefficients of \texttt{net-d1} 
	.]{Comparison of 
	the \ac{r2} coefficient of the manual and the controller learned from 
	\texttt{net-d1} with respect to the omniscient one.}
	\label{fig:net-d1r2}
\end{figure}
From these figures we expect that the robots' behaviour using the learned 
controller instead of the manual one is a bit better, even if far from the omniscient 
controller.

\begin{figure}[!htb]
	\centering
	\includegraphics[width=.7\textwidth]{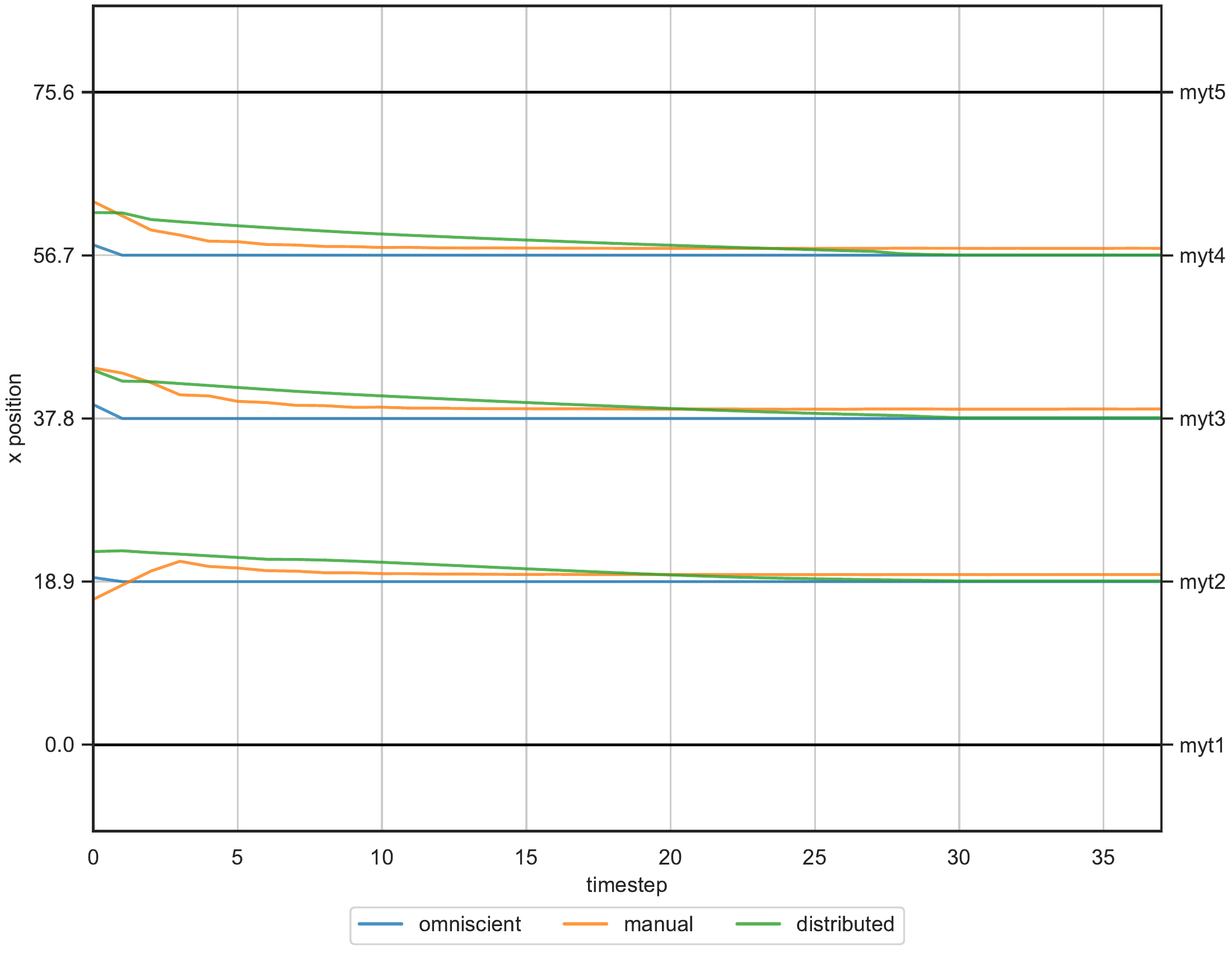}%
	\caption[Evaluation of the trajectories obtained with \texttt{prox\_values} 
	input.]{Comparison of trajectories, of a single simulation, generated using three 
	controllers: the expert, the manual and the one learned from \texttt{net-d1}.}
	\label{fig:net-d1traj1}
\end{figure}
In Figure \ref{fig:net-d1traj1} we first show a sample simulation: on the y-axis is 
represented the position of each agent, while on the x-axis the simulation time 
steps. We compare the trajectories obtained from the three controllers, in 
particular visualising the omniscient one in blue, the manual in orange and the 
learned one in green.
The extreme robots are passive. The agents moved using the omniscient controller 
reach the target very quickly, in a couple of time steps. Those moved using the 
manual controller are slower, they approach the goal position on average in 10 
time steps but never reach it. Instead, the learned controller is even slower than 
the previous one, but in about 25 time steps, the agents manage to arrive in the 
correct final configuration. 

In Figure \ref{fig:net-d1traj} we show a comparison of the expert and the learned 
trajectories, and then between the manual and the learned ones, this time 
summarising the performance over all the validation runs: at each time step, the 
position of each agent is presented 
\begin{figure}[!htb]
	\begin{center}
		\begin{subfigure}[h]{0.49\textwidth}
			\centering
			\includegraphics[width=.9\textwidth]{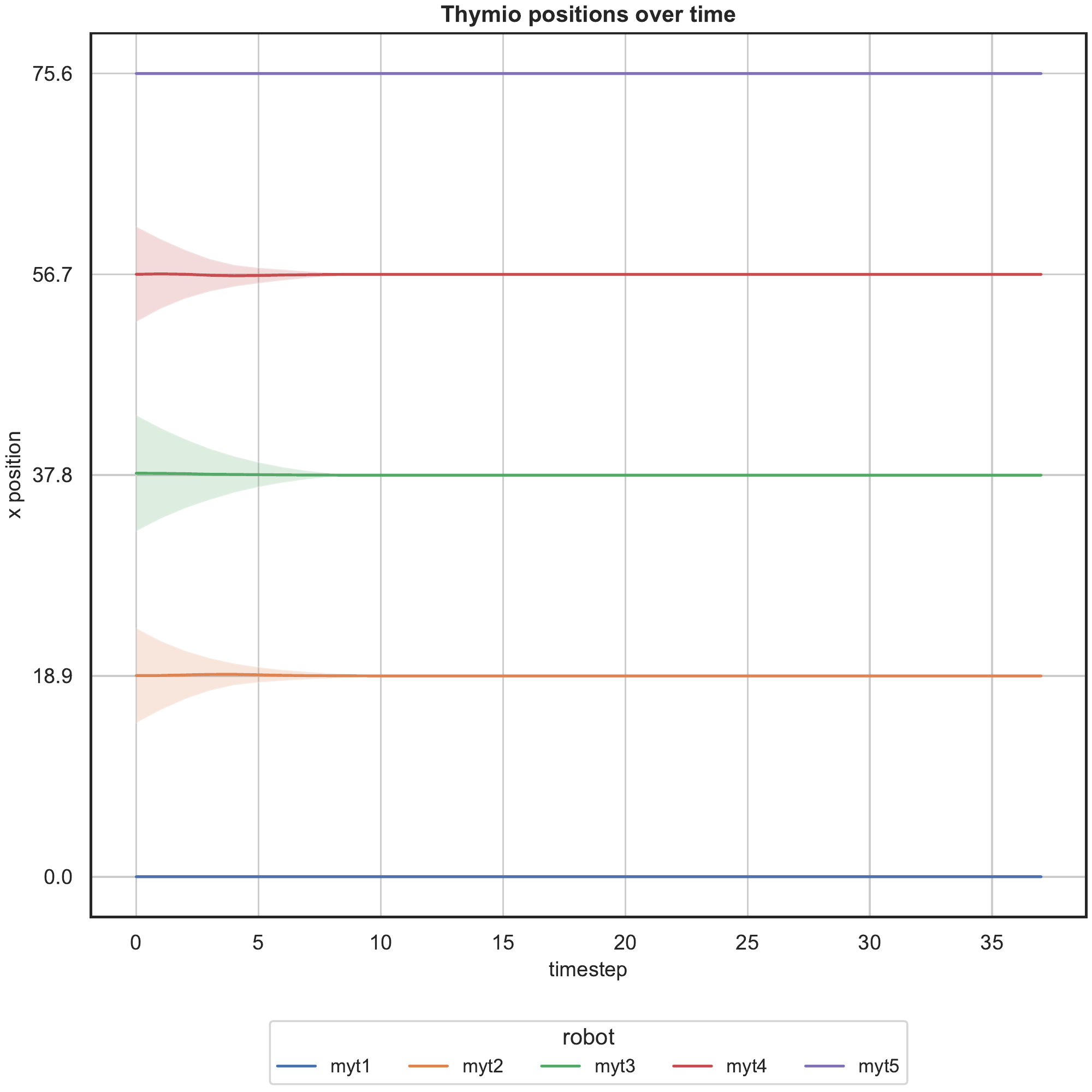}%
			\caption{Expert controller trajectories.}
		\end{subfigure}
		\hfill
		\begin{subfigure}[h]{0.49\textwidth}
			\centering
			\includegraphics[width=.9\textwidth]{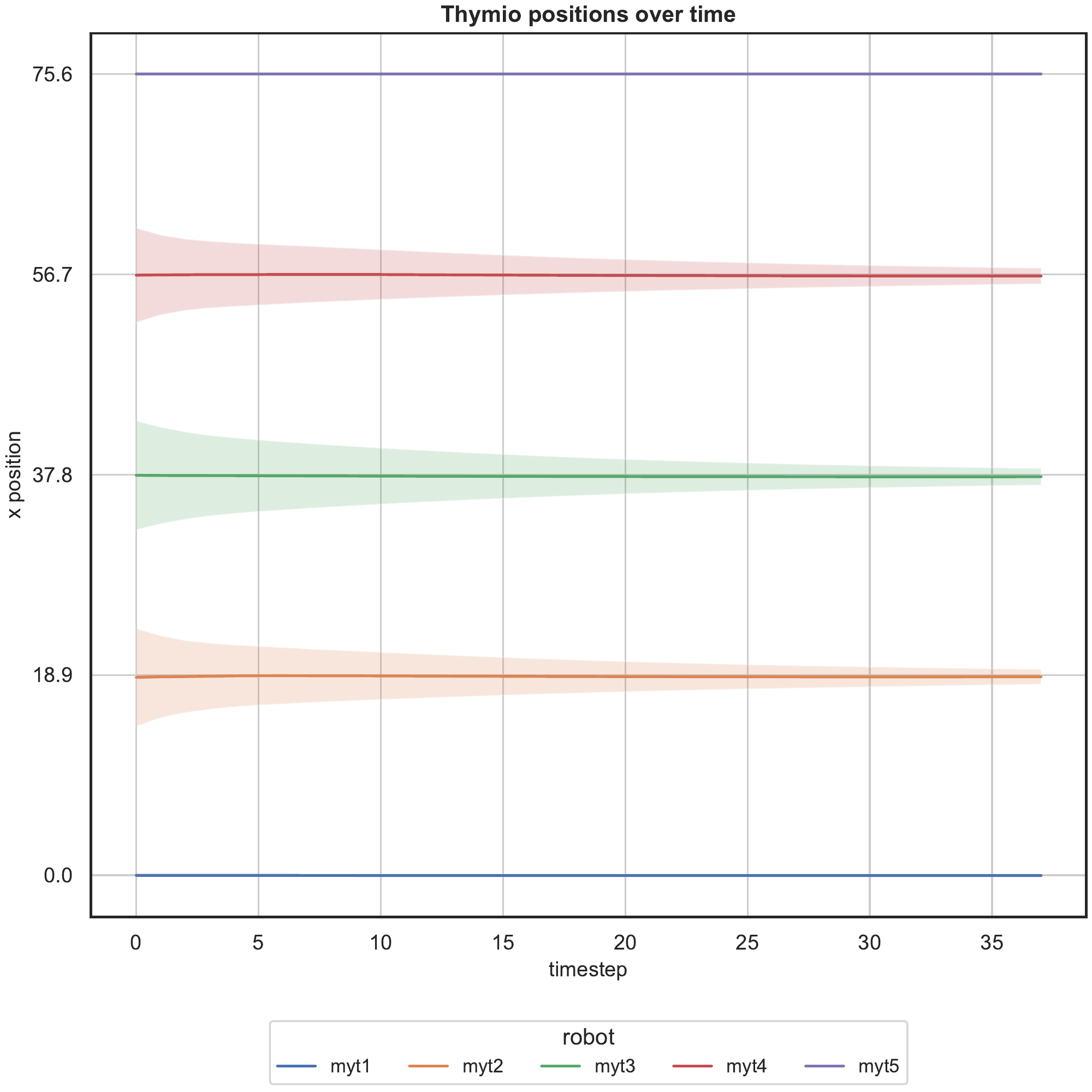}
			\caption{Distributed controller trajectories.}
		\end{subfigure}
	\end{center}
	\begin{center}
		\begin{subfigure}[h]{0.49\textwidth}
			\centering
			\includegraphics[width=.9\textwidth]{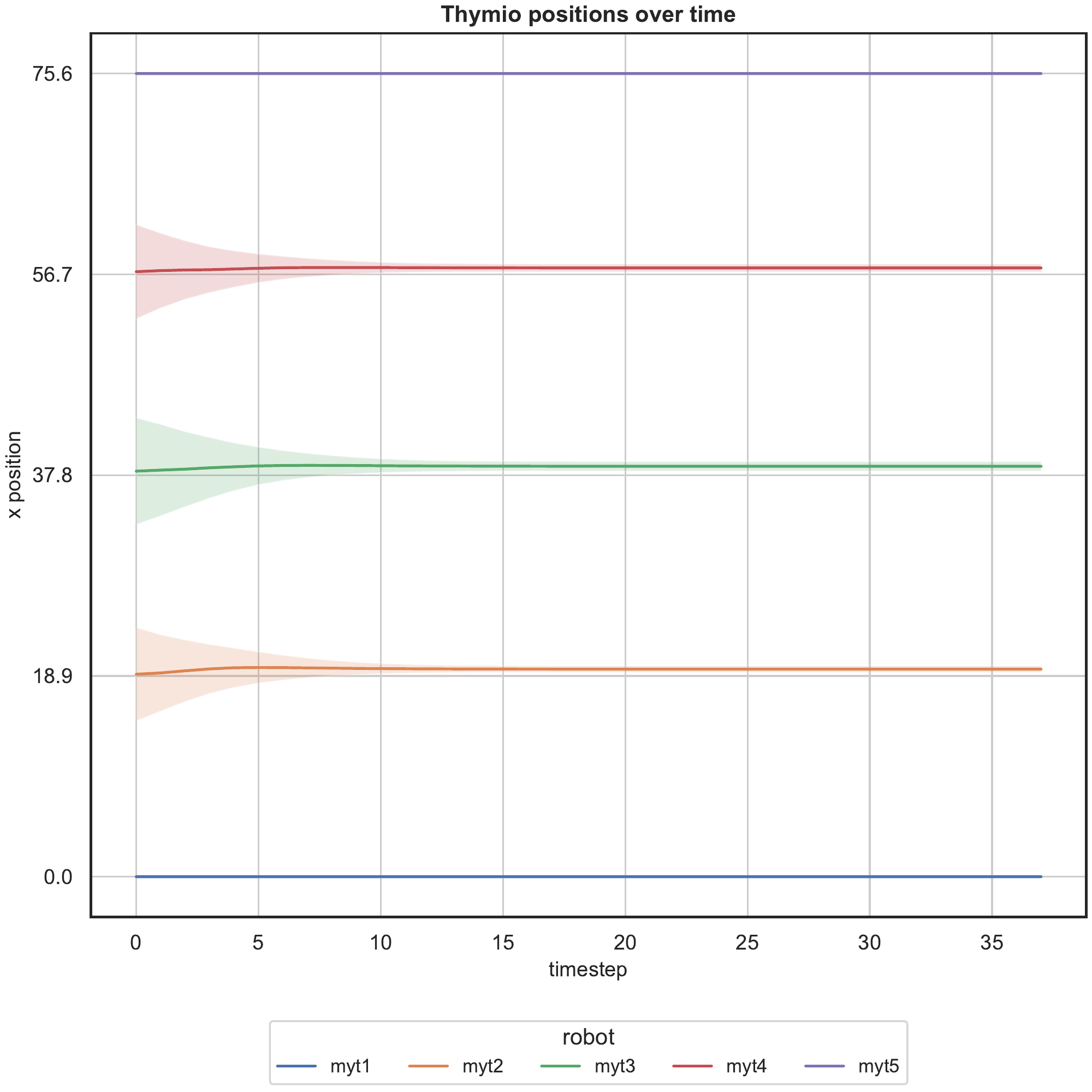}%
			\caption{Manual controller trajectories.}
		\end{subfigure}
		\hfill
		\begin{subfigure}[h]{0.49\textwidth}
			\centering
			\includegraphics[width=.9\textwidth]{contents/images/net-d1/position-overtime-learned_distributed}
			\caption{Distributed controller trajectories.}
		\end{subfigure}
	\end{center}
	\caption[Evaluation of the trajectories learned by \texttt{net-d1}]{Comparison 
	of trajectories, of all the simulation runs, generated 
	using three controllers: the expert, the manual and 
		the one learned from \texttt{net-d1}.}
	\label{fig:net-d1traj}
\end{figure}

\noindent
as an average over all the simulation runs, and 
besides is shown this average minus and plus the standard deviation.
As expected, the convergence of the robots to the target using the omniscient 
controller is much faster than with the manual or the learned one. Generally, the 
learned trajectories require a higher number of time steps to converge to the 
correct configuration, sometimes even $40$ may be necessary, compared to the 
two other controllers that need less than $10$.

Indeed, analysing in Figure \ref{fig:net-d1control} the evolution of the control 
over time, it is possible to notice that the omniscient in the first time steps uses a 
higher speed than that chosen by the manual controller or the one predicted by 
the network. 
After about $10$ time steps the expert reaches the target while the manual need 
about $15$ time steps to arrive to the goal with a certain tolerance, maintaining 
then the speed constant at $0$. Instead, the distributed controller decreases the 
speed of the agents as the time steps pass, reaching zero speed but with a certain 
variance, probably caused by oscillations.
\begin{figure}[!htb]
	\centering
	\begin{subfigure}[h]{0.3\textwidth}
		\centering
		\includegraphics[width=\textwidth]{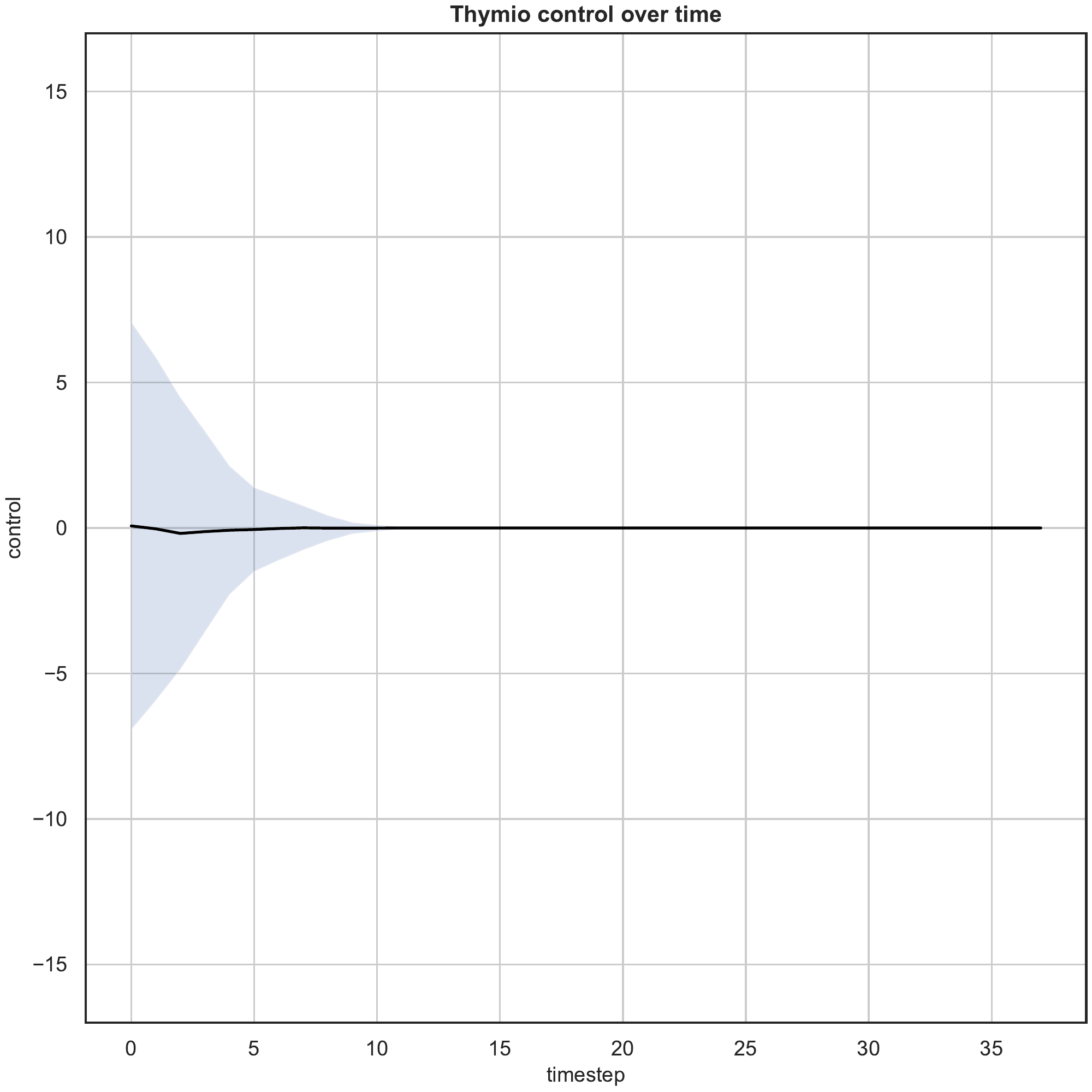}%
		\caption{Expert controller.}
	\end{subfigure}
	\hfill
	\begin{subfigure}[h]{0.3\textwidth}
		\centering
		\includegraphics[width=\textwidth]{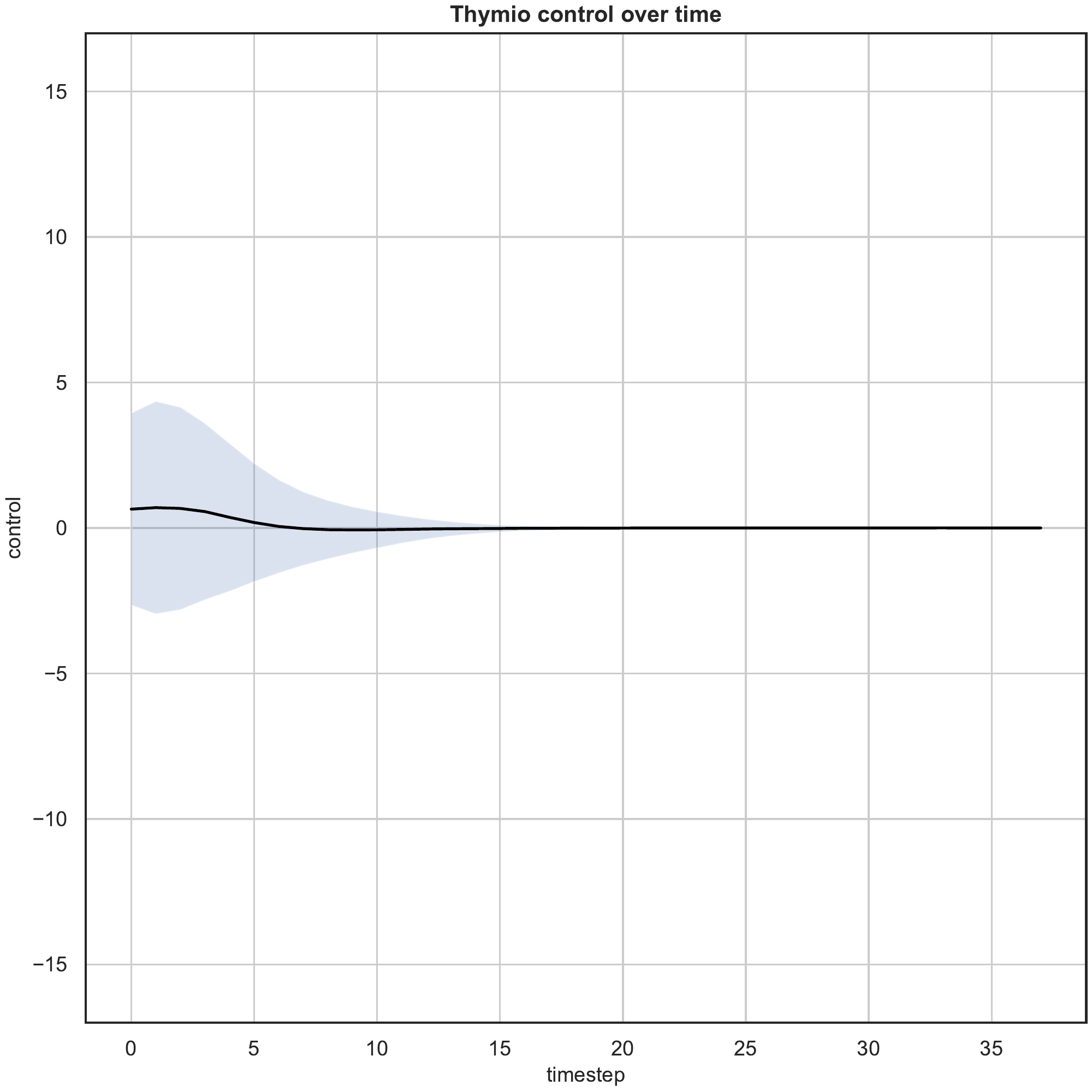}%
		\caption{Manual controller.}
	\end{subfigure}
	\hfill
	\begin{subfigure}[h]{0.3\textwidth}
		\centering
		\includegraphics[width=\textwidth]{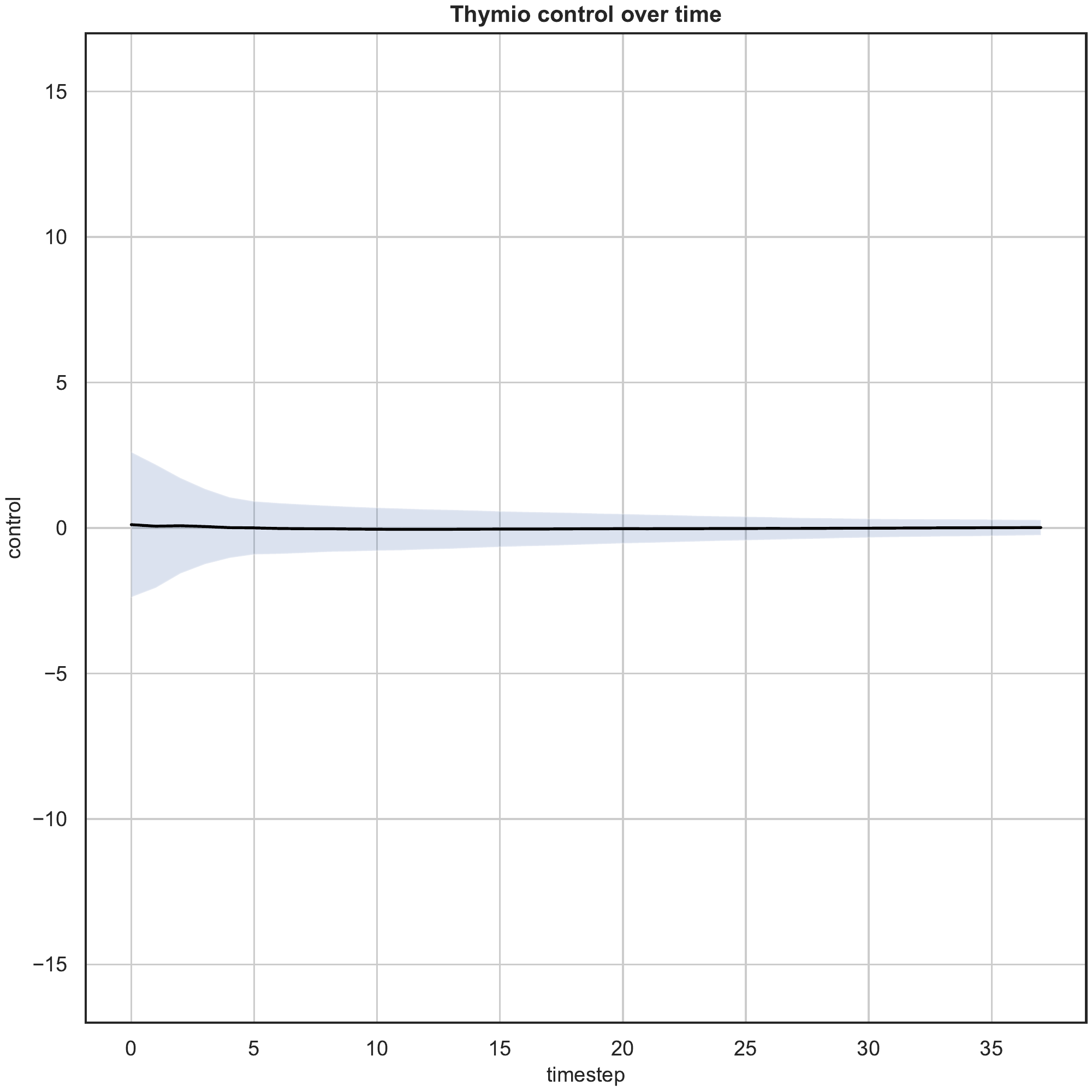}
		\caption{Distributed controller.}
	\end{subfigure}
	\caption[Evaluation of the control learned by \texttt{net-d1}.]{Comparison 
		of output control decided using three controllers: the expert, the manual 
		and the one learned from \texttt{net-d1}.}
	\label{fig:net-d1control}
\end{figure}

In Figure \ref{fig:net-d1responsesensors} is visualised the response of the learned 
controller as the input sensing changes. 
In particular we analyse two cases. The first one shows the control predicted by 
the network when the robot sees only in front and nothing behind, more 
specifically when the given input is  $([0, 0, x, 0, 0, 0, 0])$, with $x$ varying in the 
range $[0, 4500]$.
The second shows the control predicted by the network when the robot instead 
sees nothing in front, more specifically when the given input is  $([0, 0, 0, 0, 0,x , 
x])$, with $x$ varying in the range $[0, 4500]$.
The behaviour is almost as expected. When the robot sees nothing behind but 
something in front, the model returns a negative speed, since the robot has to 
move backwards. 
The absolute value of control increases as the proximity to the obstacle increases.
A complementary behaviour is obtained when the robot sees only behind but 
not in front.
\begin{figure}[!htb]
	\centering
	\begin{subfigure}[h]{0.49\textwidth}
		\centering
		\includegraphics[width=\textwidth]{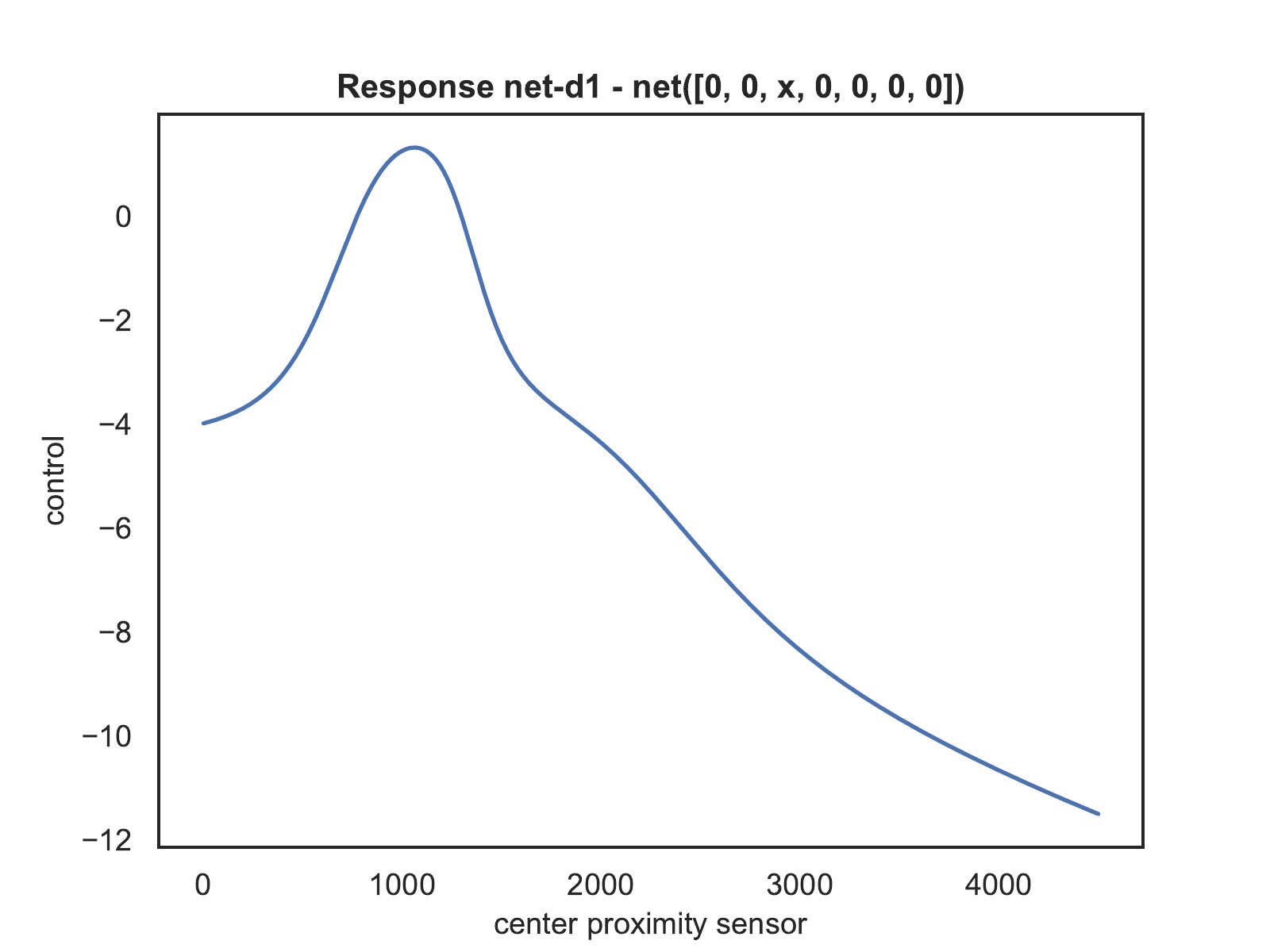}%
		%\caption{response-net-d1-net([0, 0, x, 0, 0, 0, 0]).}
	\end{subfigure}
	\hfill
	\begin{subfigure}[h]{0.49\textwidth}
		\centering
		\includegraphics[width=\textwidth]{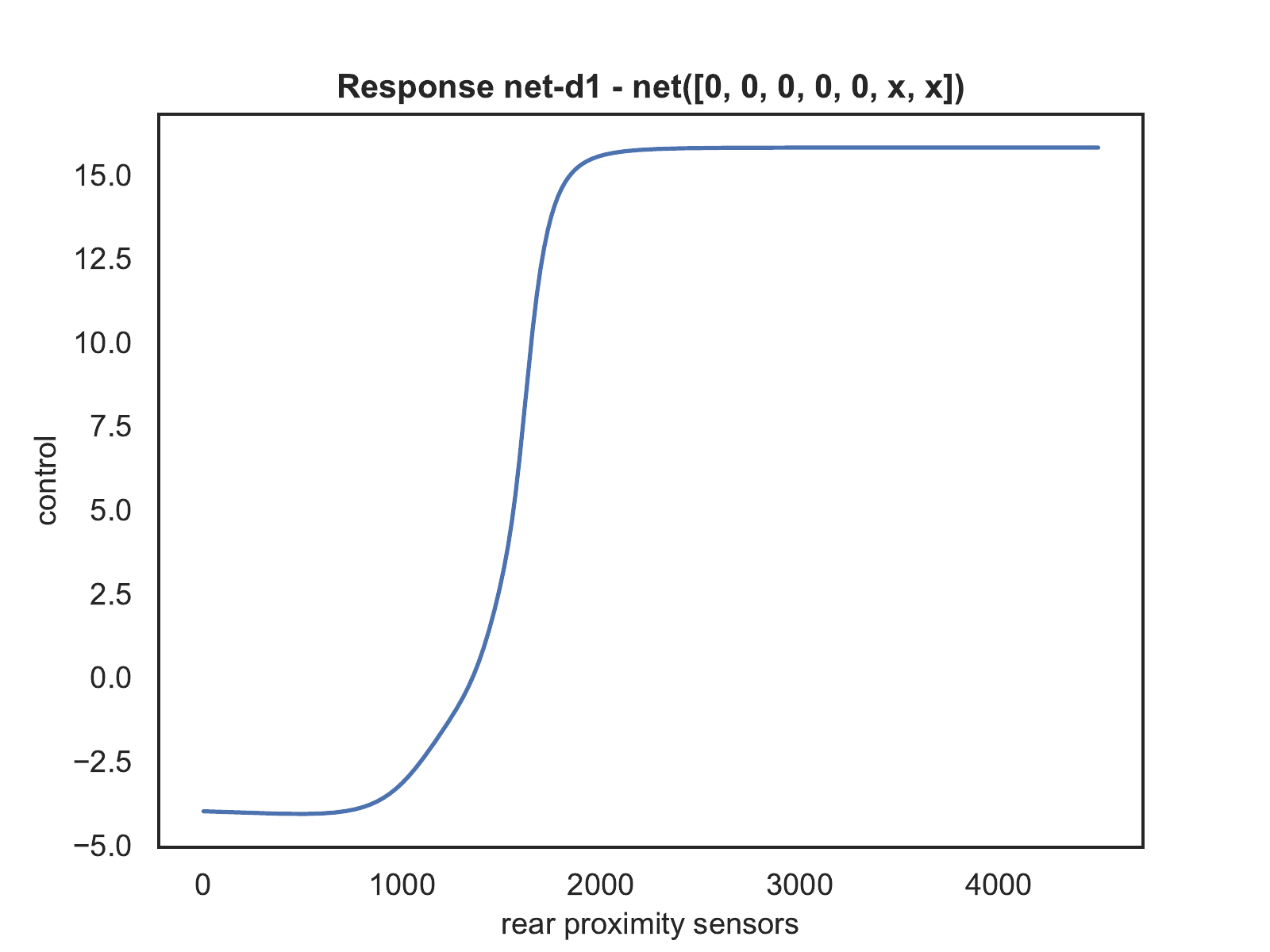}
		%\caption{response-net-d1-net([0, 0, 0, 0, 0, x, x]).}
	\end{subfigure}
	\caption[Response of \texttt{net-d1} by varying the input sensing.]{Response of 
		\texttt{net-d1} by varying the input sensing.}
	\label{fig:net-d1responsesensors}
\end{figure}

In Figure \ref{fig:net-d1responseposition} is displayed the behaviour of a robot 
located between two stationary agents which are already in their place, showing 
the response of the controllers, on the y-axis, by varying the position of the 
moving robot, visualised on the x-axis. The output control is computed as an 
average over $100$ measures in which the pose of the agent $(x, y, \theta)$ 
differs by a certain epsilon uniformly distributed in the range $[-0.5, 0.5]$, thus 
to avoid the effects of noise that would be obtained on a single measurement and 
unrealistic artefacts in which the sensors are not continuous. Besides, are shown 
the bands which represent plus and minus standard deviation. As expected, the 
output is a high value, positive or negative respectively when the robot is close to 
an obstacle on the left or on the right, or it is close to $0$ when the distance from 
right and left is equal.
\begin{figure}[!htb]
	\centering
	\includegraphics[width=.45\textwidth]{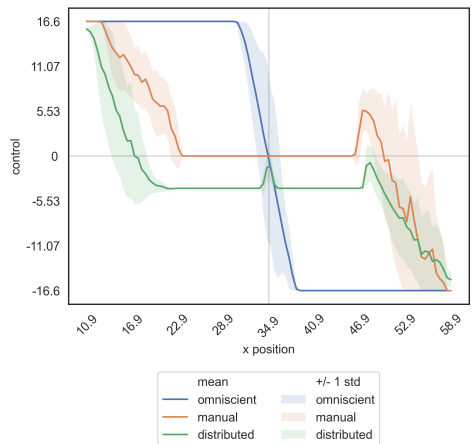}%
	\caption{Response of \texttt{net-d1} by varying the initial position.}
	\label{fig:net-d1responseposition}
\end{figure}

Finally, in Figure \ref{fig:net-d1distance} is presented another useful metric that 
measures the absolute distance of each robot from the target, visualised on the 
y-axis, over time. This value is averaged on all robots among all the simulation 
runs. The median value is shown as well as the interquartile and interdecile ranges.
On average, the distance from goal of the learned controller is lower than the one 
obtained with the manual controller, meaning that in the final configuration 
the robots moved following the learned controller are closer to the target than 
those moved with the manual one, which are on average at a distance of about 
$1$cm from the goal position. 
\begin{figure}[!htb]
	\centering
	\includegraphics[width=.65\textwidth]{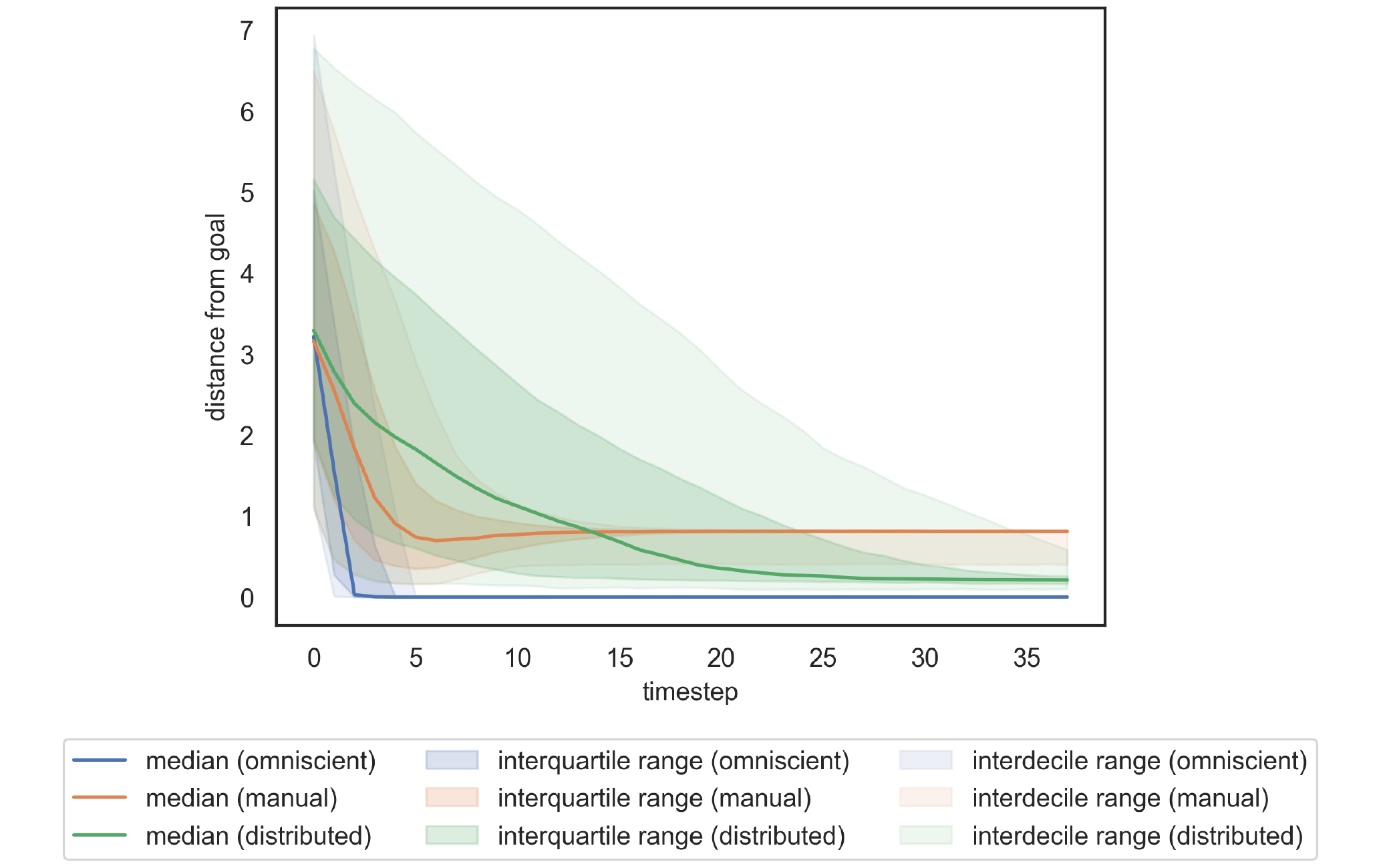}%
	\caption[Evaluation of \texttt{net-d1} distances from goal.]{Comparison of 
		performance in terms of distances from goal obtained using three 
		controllers: 
		the expert, the manual and the one learned from \texttt{net-d1}.}
	\label{fig:net-d1distance}
\end{figure}

As mentioned before, in case of \texttt{prox\_values} inputs the 
experiment performed with an \texttt{avg\_gap} of $24$ is not meaningful since 
this value exceeds the maximal range of the sensor. Similarly, since $14$ is the 
maximum range, it is difficult to use this type of input when the \texttt{avg\_gap}  
is $13$, as shown by the losses in Figure \ref{fig:distlossprox_values}.
\begin{figure}[!htb]
	\centering
	\includegraphics[width=.85\textwidth]{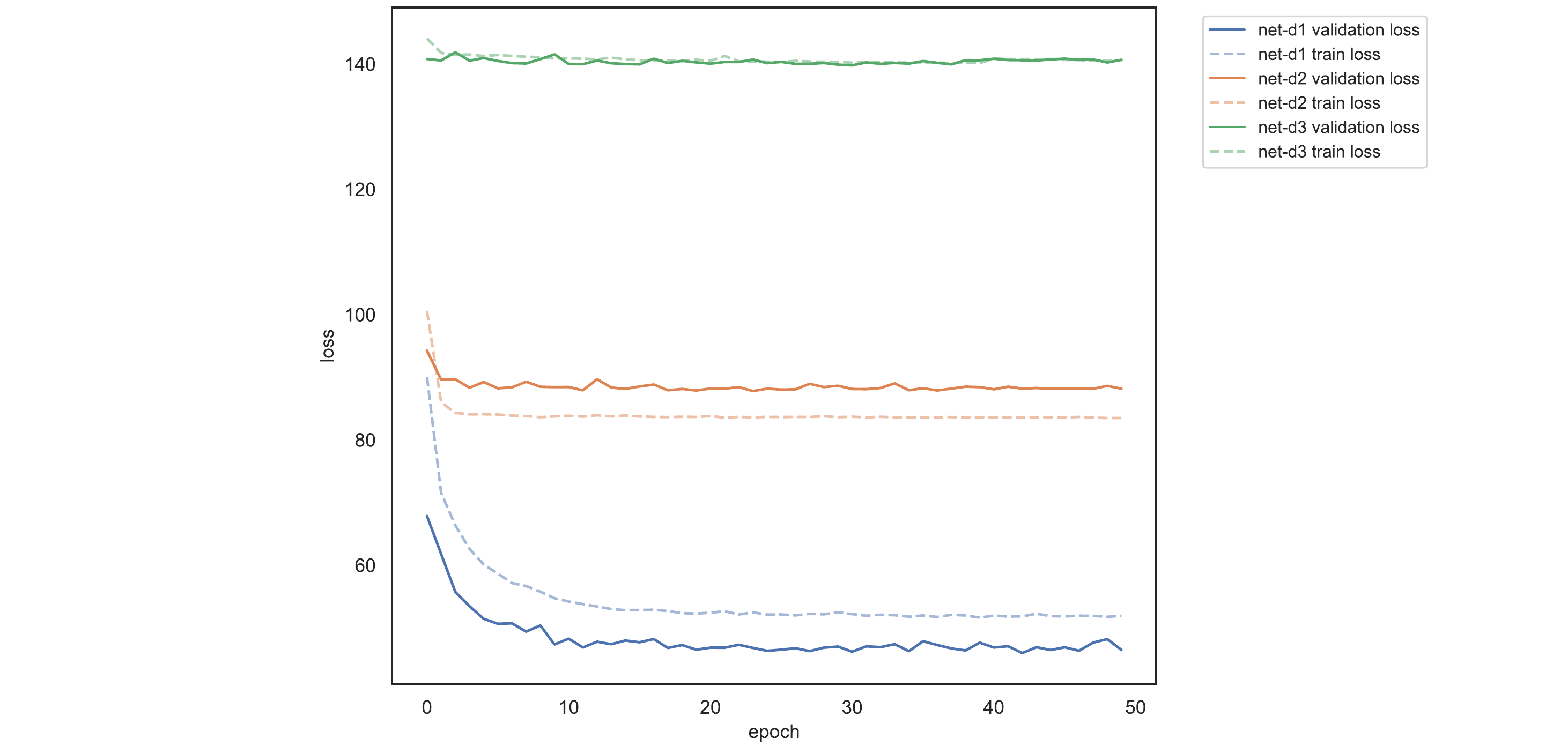}%
	\caption{Comparison of the losses of the models that use \texttt{prox\_values} 
		readings.}
	\label{fig:distlossprox_values}
\end{figure}

\paragraph*{Results using \texttt{prox\_comm} input}
Following are shown the results of the experiments obtained using the 
\texttt{prox\_comm} readings. 
In Figure \ref{fig:distlossprox_comm}, we analyse the losses by varying the 
average gap. From a first observation, the network seems to be able to work with 
all the gaps.
\begin{figure}[!htb]
	\centering
	\includegraphics[width=.85\textwidth]{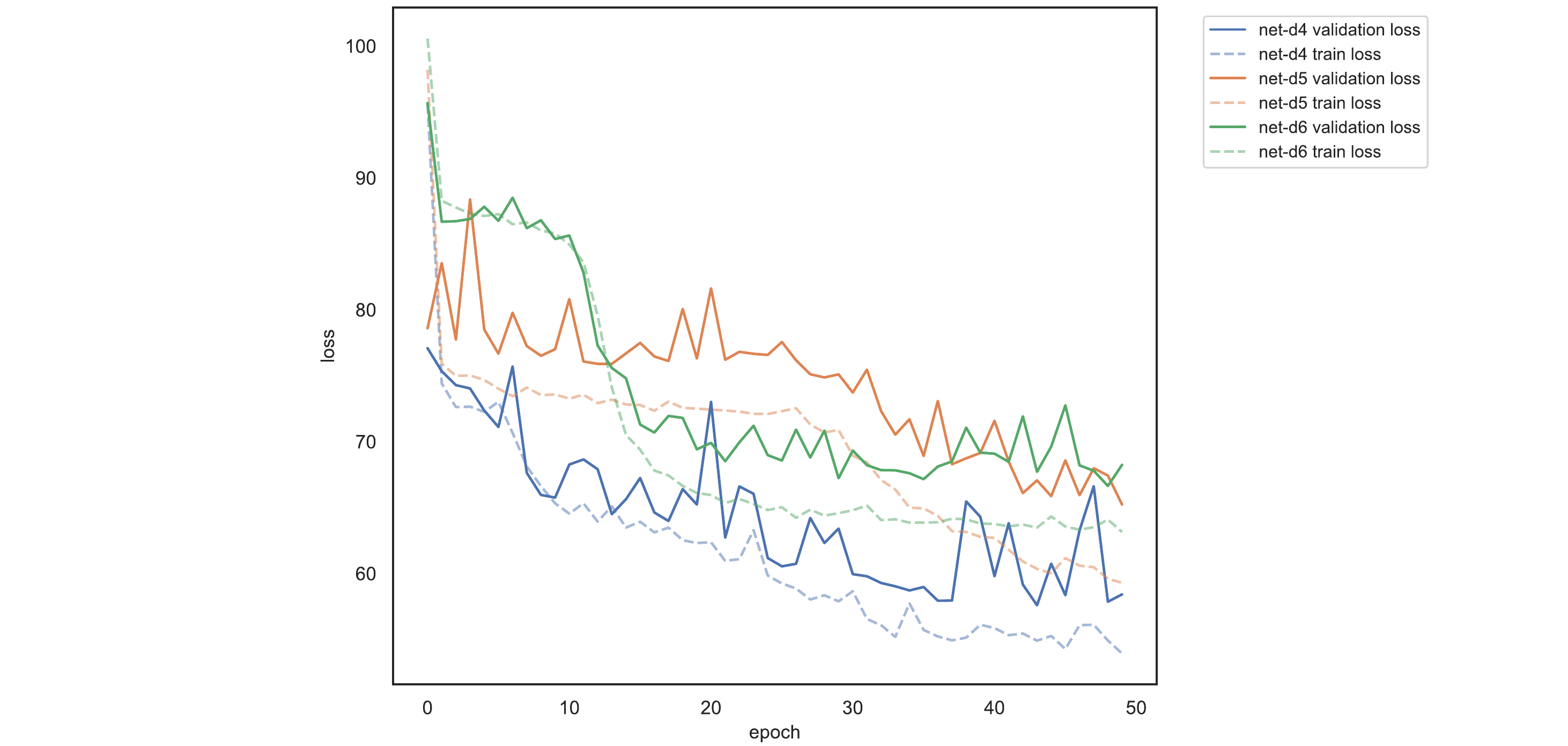}%
	\caption{Comparison of the losses of the models that use \texttt{prox\_comm} 
		readings.}
	\label{fig:distlossprox_comm}
\end{figure}

For the assumptions made before, we believe that the model obtained from  
\texttt{net-d6}, the has a higher average gap, is the more promising. 
Moreover, as shown from the \ac{r2} coefficients in Figure \ref{fig:net-d6r2}, we 
expect that the robots’ behaviour using the learned instead of the manual 
controller is better, even if far from the expert.

%Examining  in Figure \ref{fig:net-d456r2}, the higher 
%value is obtained with. 

%\begin{figure}[!htb]
%	\begin{center}
%		\begin{subfigure}[h]{0.49\textwidth}
%			
%\includegraphics[width=\textwidth]{contents/images/net-d4/regression-net-d4-vs-omniscient}%
%		\end{subfigure}
%		\hfill\vspace{-0.5cm}
%		\begin{subfigure}[h]{0.49\textwidth}
%			
%\includegraphics[width=\textwidth]{contents/images/net-d5/regression-net-d5-vs-omniscient}%
%		\end{subfigure}
%	\end{center}
%	\begin{center}
%		\begin{subfigure}[h]{0.49\textwidth}
%			
%\includegraphics[width=\textwidth]{contents/images/net-d6/regression-net-d6-vs-omniscient}
%		\end{subfigure}
%	\end{center}
%	\caption[Comparison of the \ac{r2} coefficients for \texttt{prox\_comm} 
%	readings.]{Comparison of the \ac{r2} coefficients of the models that use 
%		\texttt{prox\_comm} readings.}
%	\label{fig:net-d456r2}
%\end{figure}
\begin{figure}[!htb]
	\centering
	\begin{subfigure}[h]{0.49\textwidth}
		\centering
		\includegraphics[width=\textwidth]{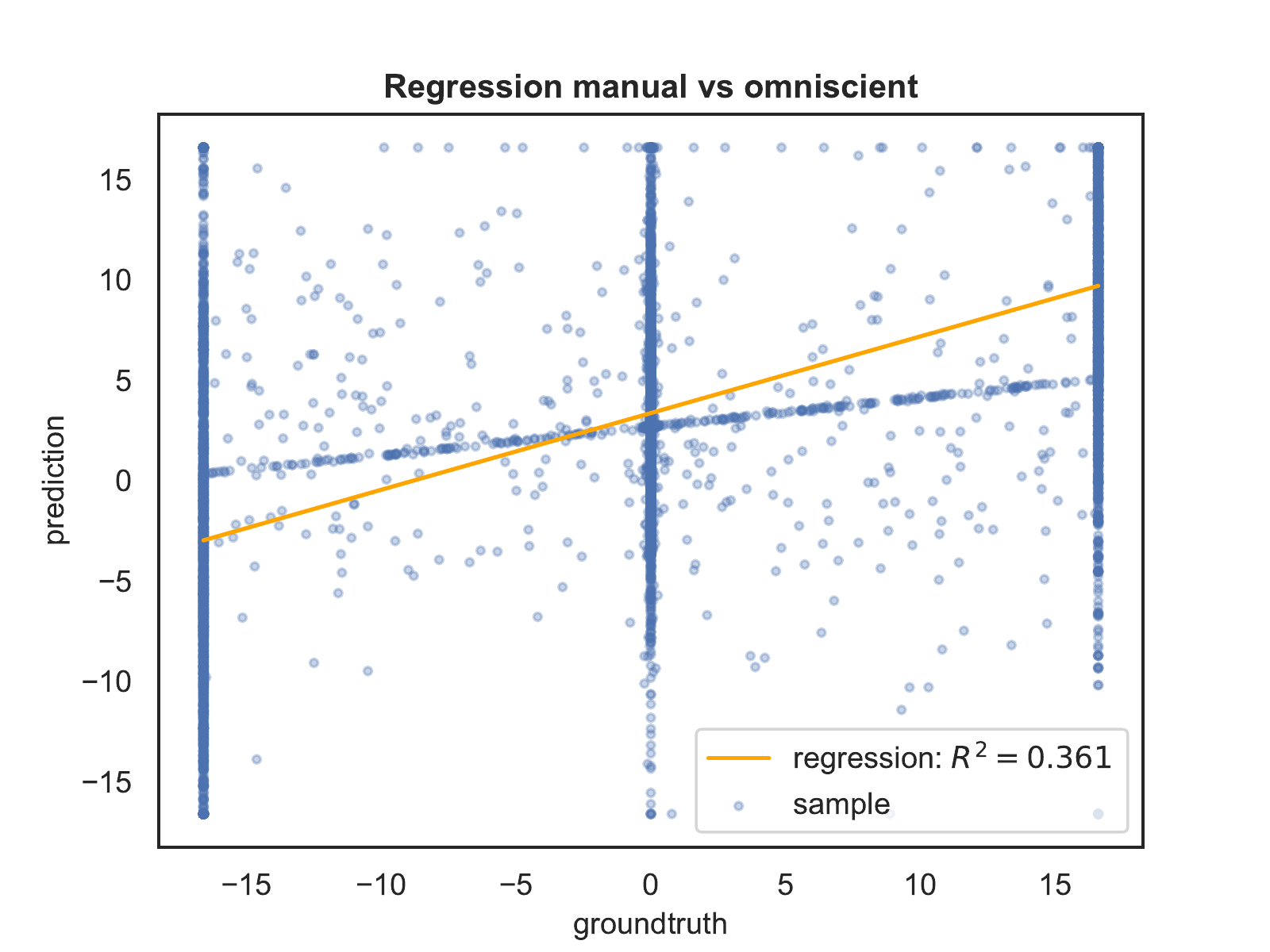}%
	\end{subfigure}
	\hfill
	\begin{subfigure}[h]{0.49\textwidth}
		\centering
		\includegraphics[width=\textwidth]{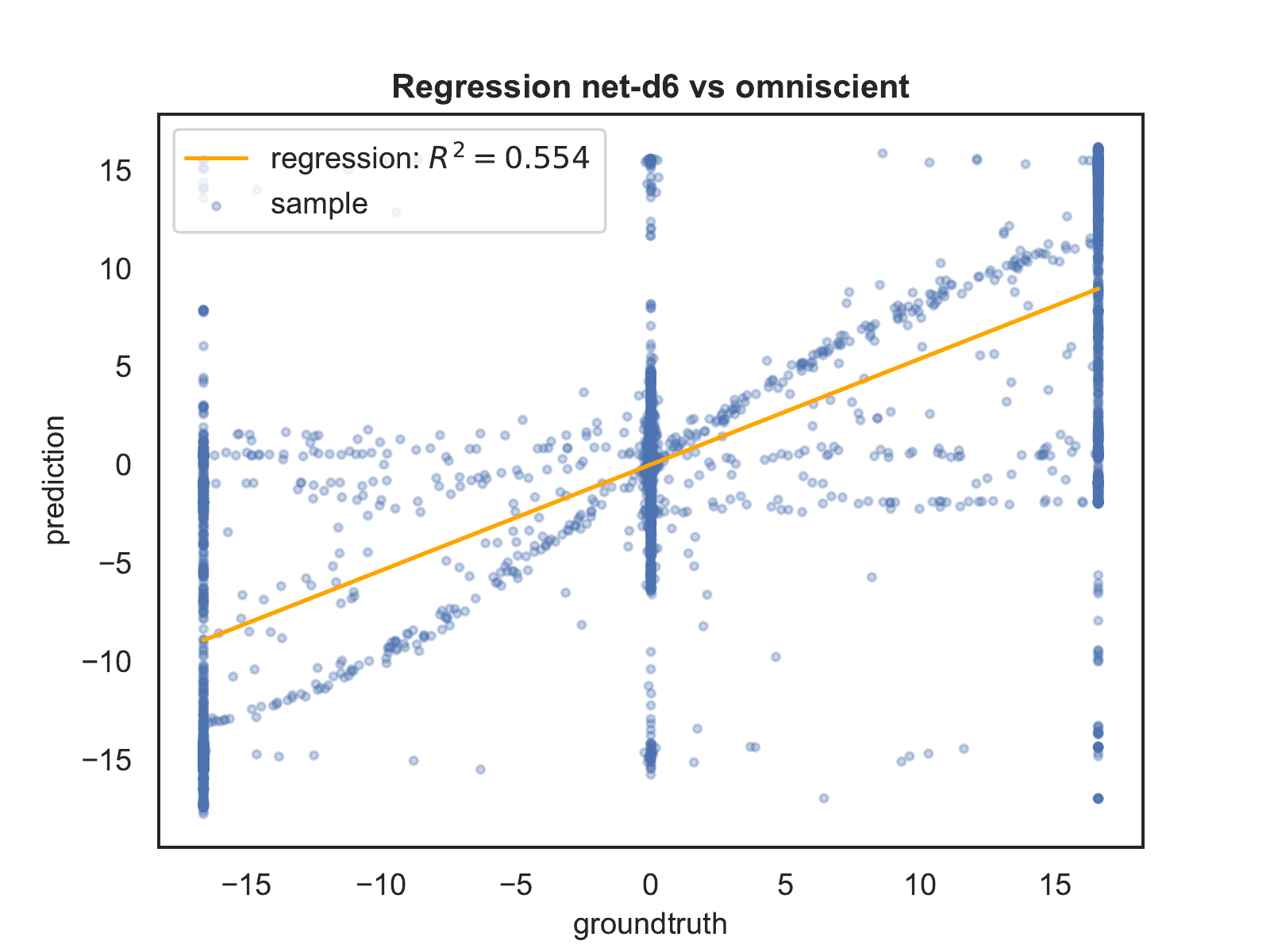}
	\end{subfigure}
	\caption[Evaluation of the \ac{r2} coefficients of \texttt{net-d6} .]{Comparison 
	of the \ac{r2} coefficient of the manual and the controller learned from 
	\texttt{net-d6} with respect to the omniscient one.}
	\label{fig:net-d6r2}
\end{figure}

In Figure \ref{fig:net-d6traj1}, we show a comparison of the trajectories obtained 
for a sample simulation, using the three controllers.
\begin{figure}[!htb]
	\centering
	\includegraphics[width=.7\textwidth]{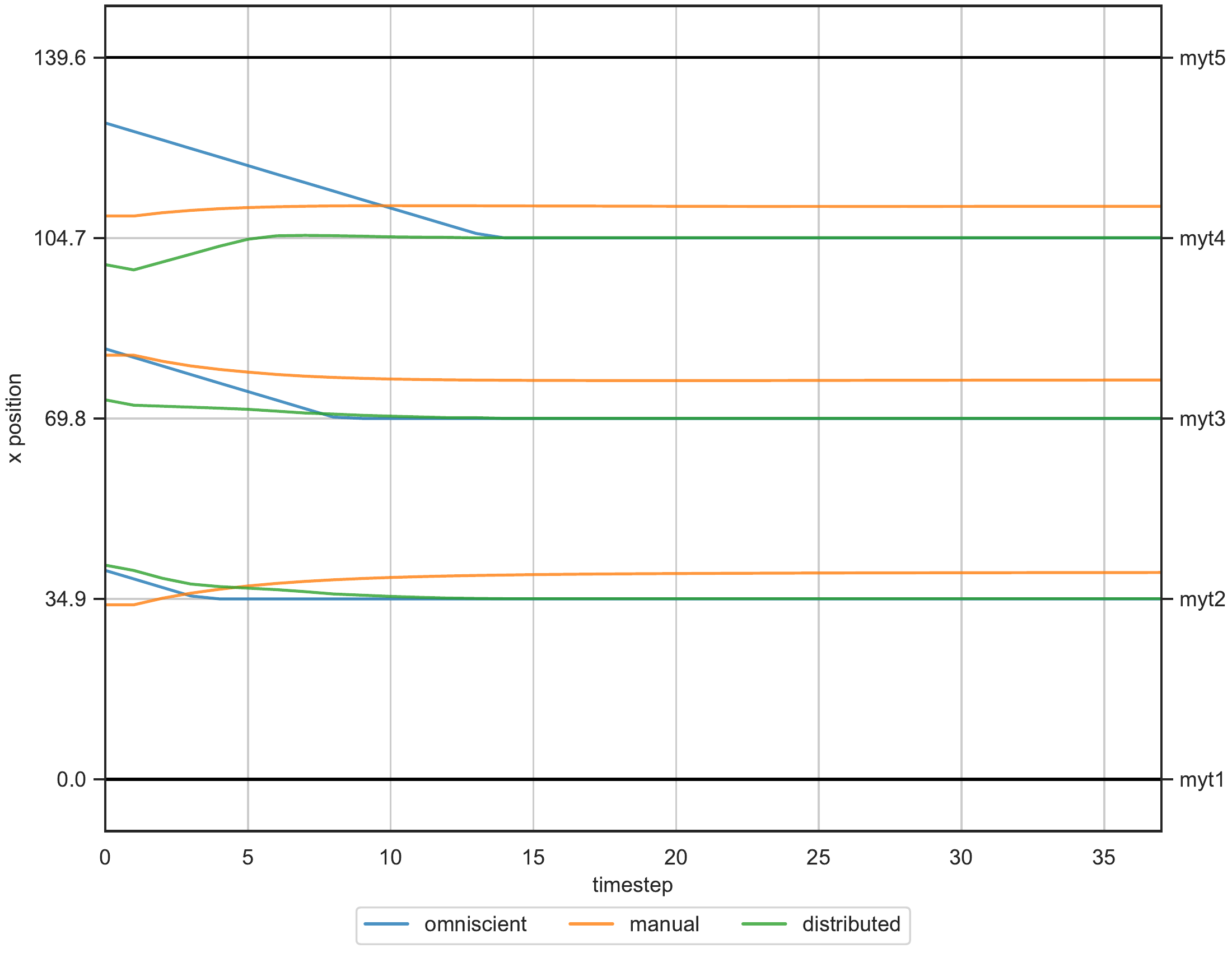}%
	\caption[Evaluation of the trajectories obtained with \texttt{prox\_comm} 
	input.]{Comparison of trajectories, of a single simulation, generated using three 
		controllers: the expert, the manual and the one learned from \texttt{net-d6}.}
	\label{fig:net-d6traj1}
\end{figure}  
We immediately see that the agents moved using the omniscient controller 
reach the target in less then $15$ time steps. Those moved using the manual 
controller did not approach the goal, even if they try to position themselves at 
equal distances. Instead, the learned controller lets the robots arrive in the 
correct final configuration in $10$ time steps, faster than the expert. This 
because, in this case, the initial positions of the agents moved with the omniscient 
controller are farther than in the other case, so it takes longer to reach the goal.

\bigskip
In Figure \ref{fig:net-d6traj} are shown the trajectories obtained employing the 
three controllers, averaged over all the runs.
As expected, the convergence to the target is slower than before, even for the 
expert, since the distance between the robots is greater, but it is still much faster 
than with the other two controllers.
The manual controller has serious
\begin{figure}[!htb]
	\begin{center}
		\begin{subfigure}[h]{0.49\textwidth}
			\centering
			\includegraphics[width=.9\textwidth]{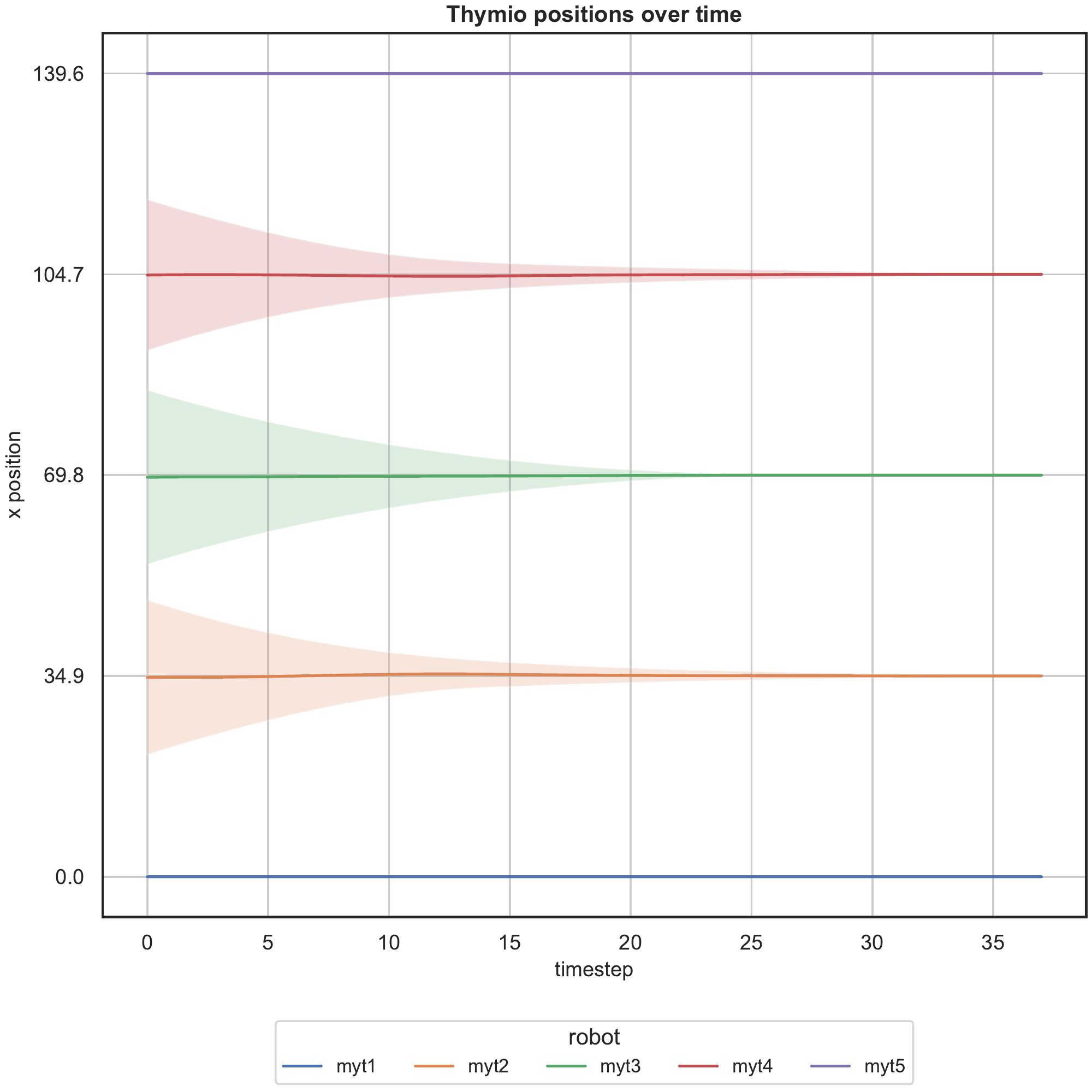}%
			\caption{Expert controller trajectories.}
		\end{subfigure}
		\hfill
		\begin{subfigure}[h]{0.49\textwidth}
			\centering
			\includegraphics[width=.9\textwidth]{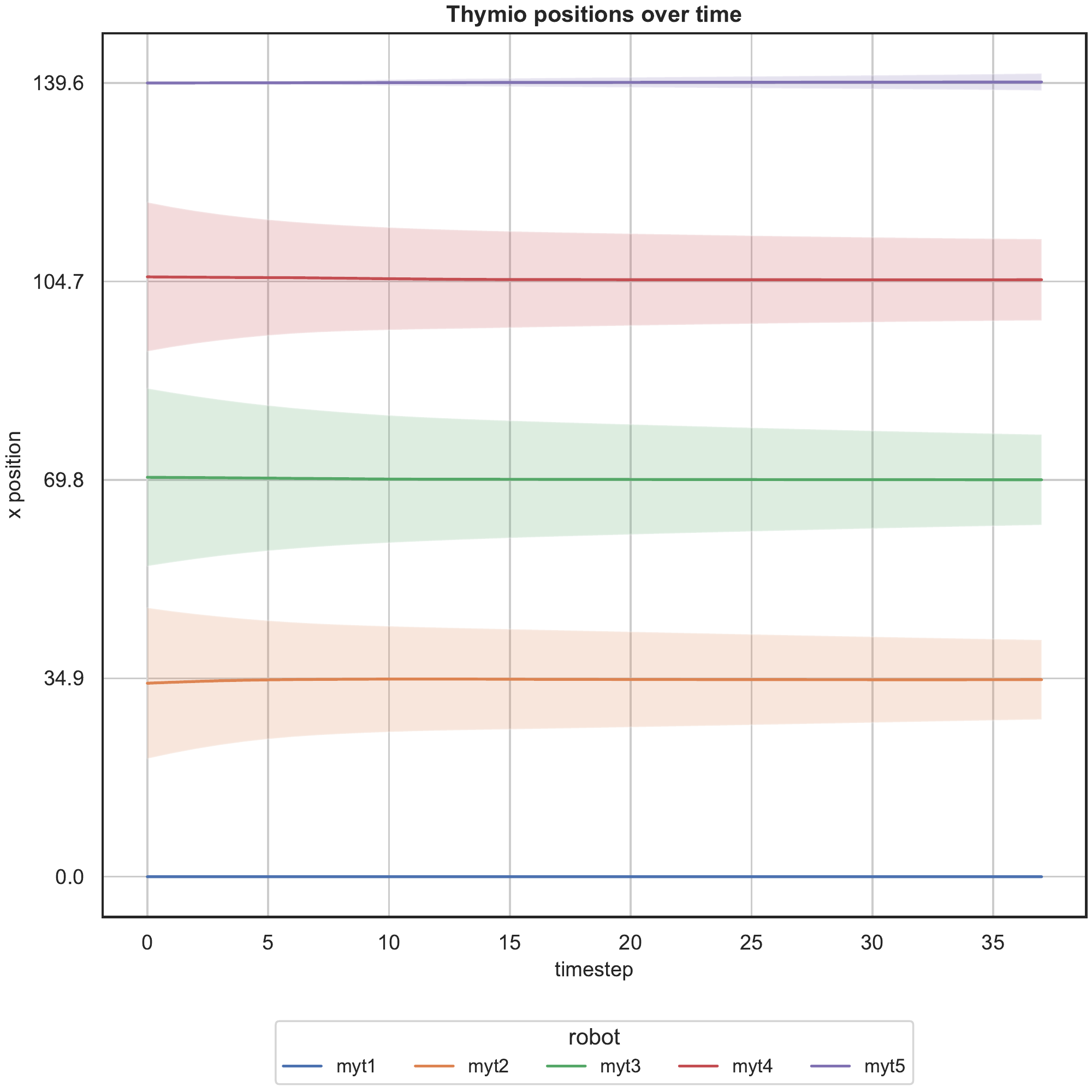}
			\caption{Distributed controller trajectories.}
		\end{subfigure}
	\end{center}
	\caption[Evaluation of the trajectories learned by \texttt{net-d6}.]{Comparison 
	of trajectories, of all the simulation runs, generated using three controllers: the 
	expert, the manual and the one learned from \texttt{net-d6}.}
\end{figure}
\begin{figure}[!htb]\ContinuedFloat
	\begin{center}
	\begin{subfigure}[h]{0.49\textwidth}
		\centering			
		\includegraphics[width=.9\textwidth]{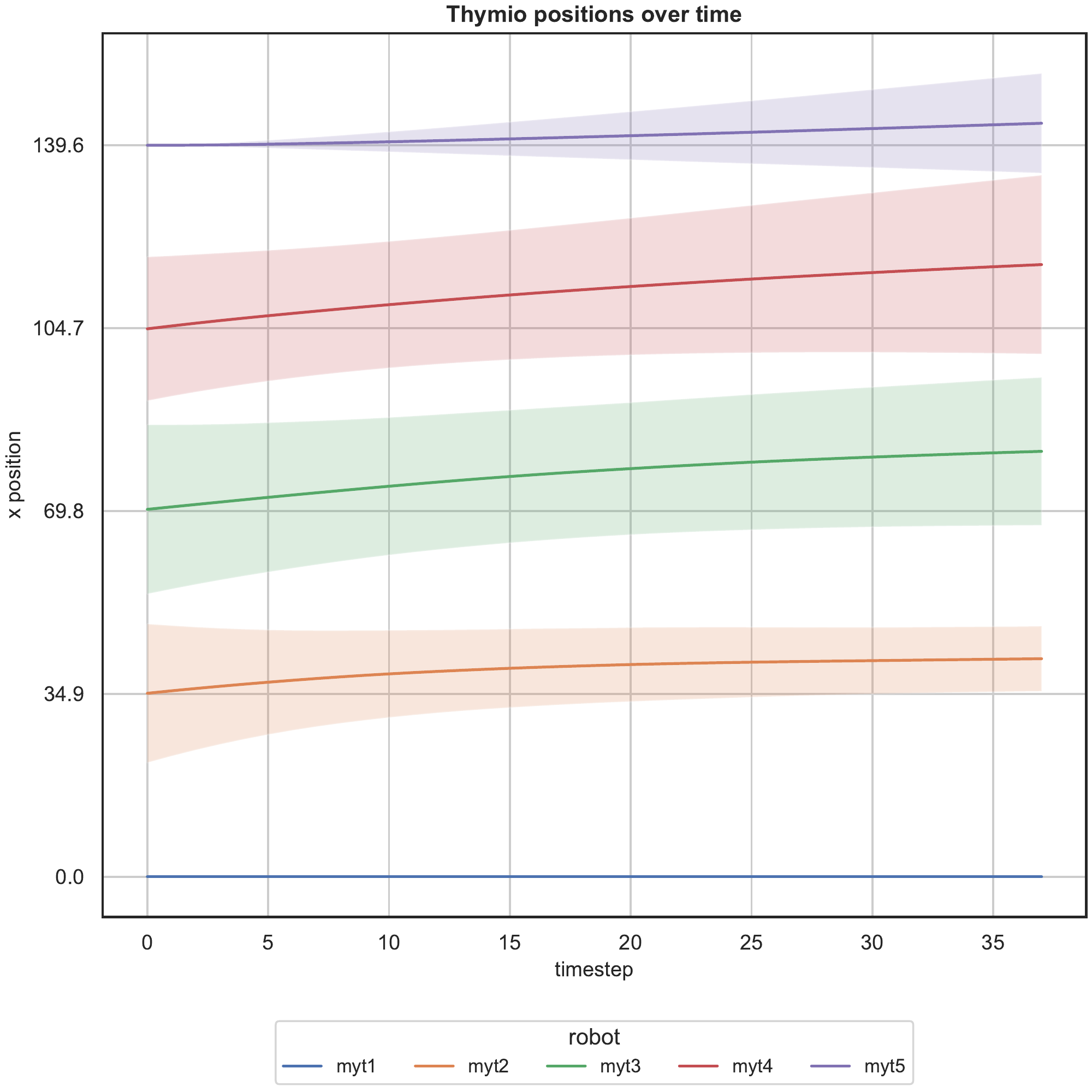}%
		\caption{Manual controller trajectories.}
	\end{subfigure}
	\hfill
	\begin{subfigure}[h]{0.49\textwidth}
		\centering
		\includegraphics[width=.9\textwidth]{contents/images/net-d6/position-overtime-learned_distributed}
		\caption{Distributed controller trajectories.}
	\end{subfigure}
	\end{center}
	\caption[]{Comparison 
	of trajectories, of all the simulation runs, generated using three controllers: the 
	expert, the manual and the one learned from \texttt{net-d6} (cont.).}
	\label{fig:net-d6traj}
\end{figure}

\noindent
problems in reaching the goal: even if the 
agents try to position themselves at equal distances, they tend to increase the 
average gap between them, creating situations in which the last robot in motion 
hits the fixed one. 
Surprisingly, the learned controller allows the agents to converge to the correct 
configuration by taking more time than the expert does.

An immediate examination of the evolution of the control over time, in Figure 
\ref{fig:net-d6control}, highlights the speed of the expert controller, which in all 
the simulation runs, after $25$ time steps, has reached 0. 
In addition, the manual controller always sets a positive speed, which leads to the 
wrong behaviour mentioned earlier, while the slowness of the distributed 
control is explained by the usage of a low speed.
\begin{figure}[!htb]
	\centering
	\begin{subfigure}[h]{0.3\textwidth}
		\centering
		\includegraphics[width=\textwidth]{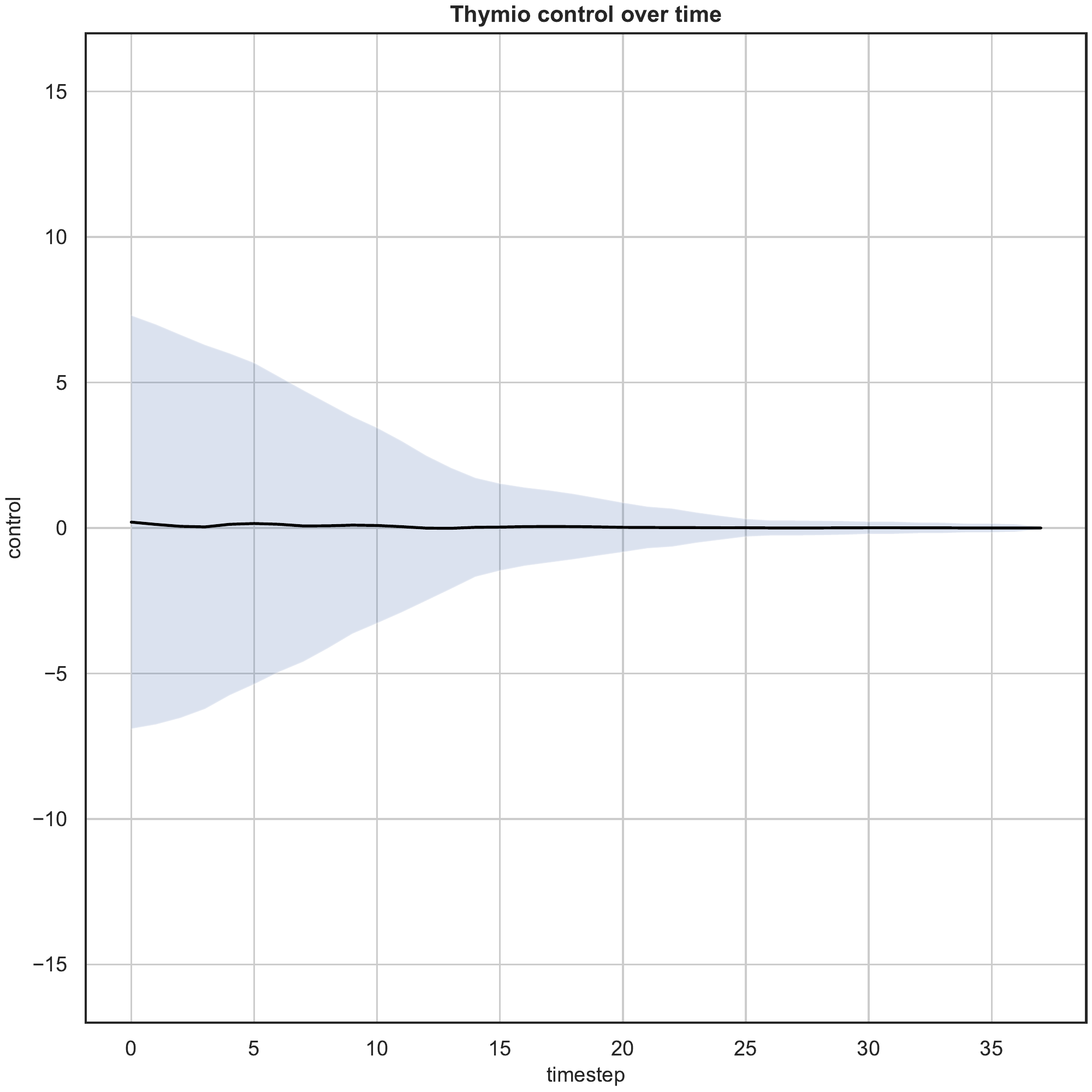}%
		\caption{Expert controller.}
	\end{subfigure}
	\hfill
	\begin{subfigure}[h]{0.3\textwidth}
		\centering
		\includegraphics[width=\textwidth]{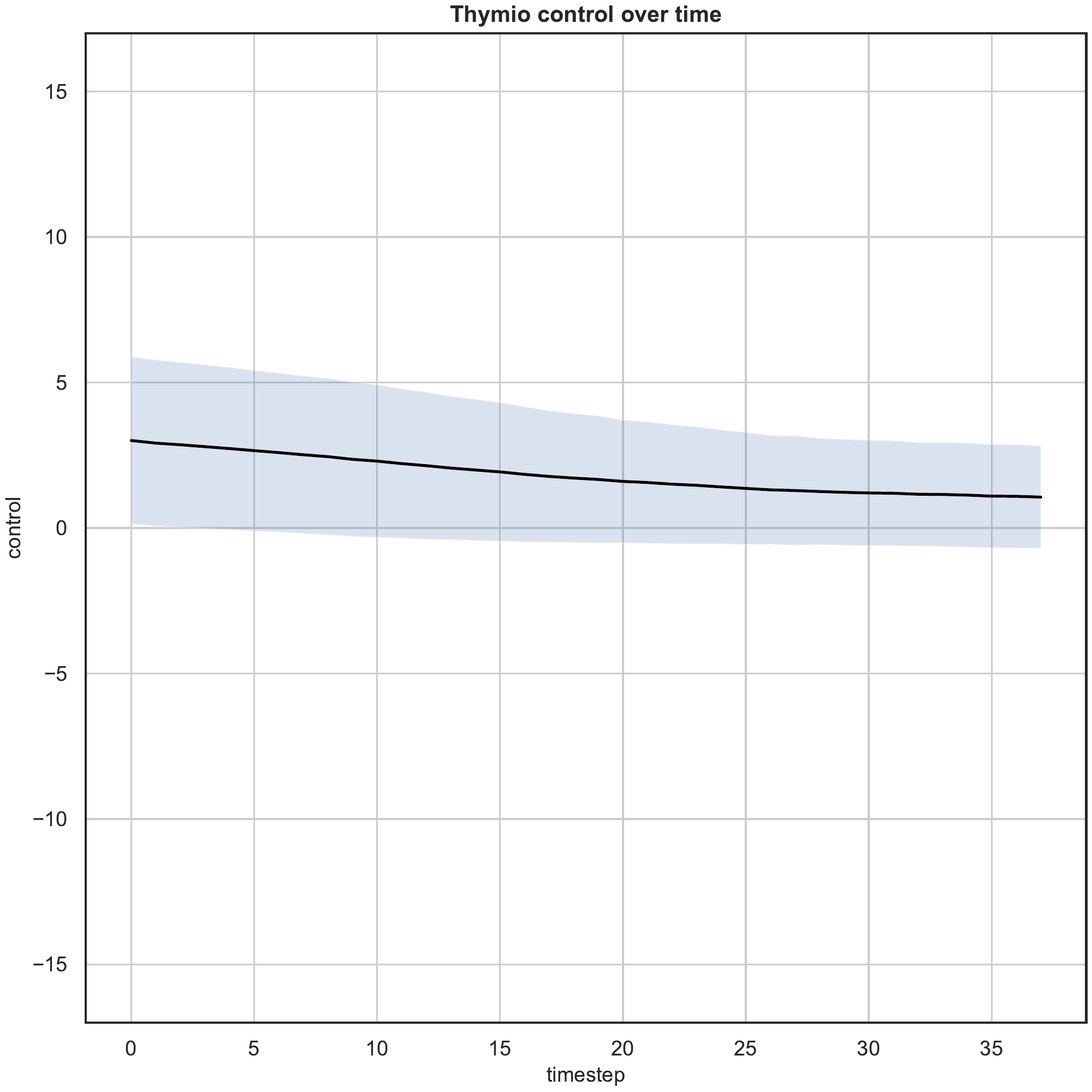}%
		\caption{Manual controller.}
	\end{subfigure}
	\hfill
	\begin{subfigure}[h]{0.3\textwidth}
		\centering
		\includegraphics[width=\textwidth]{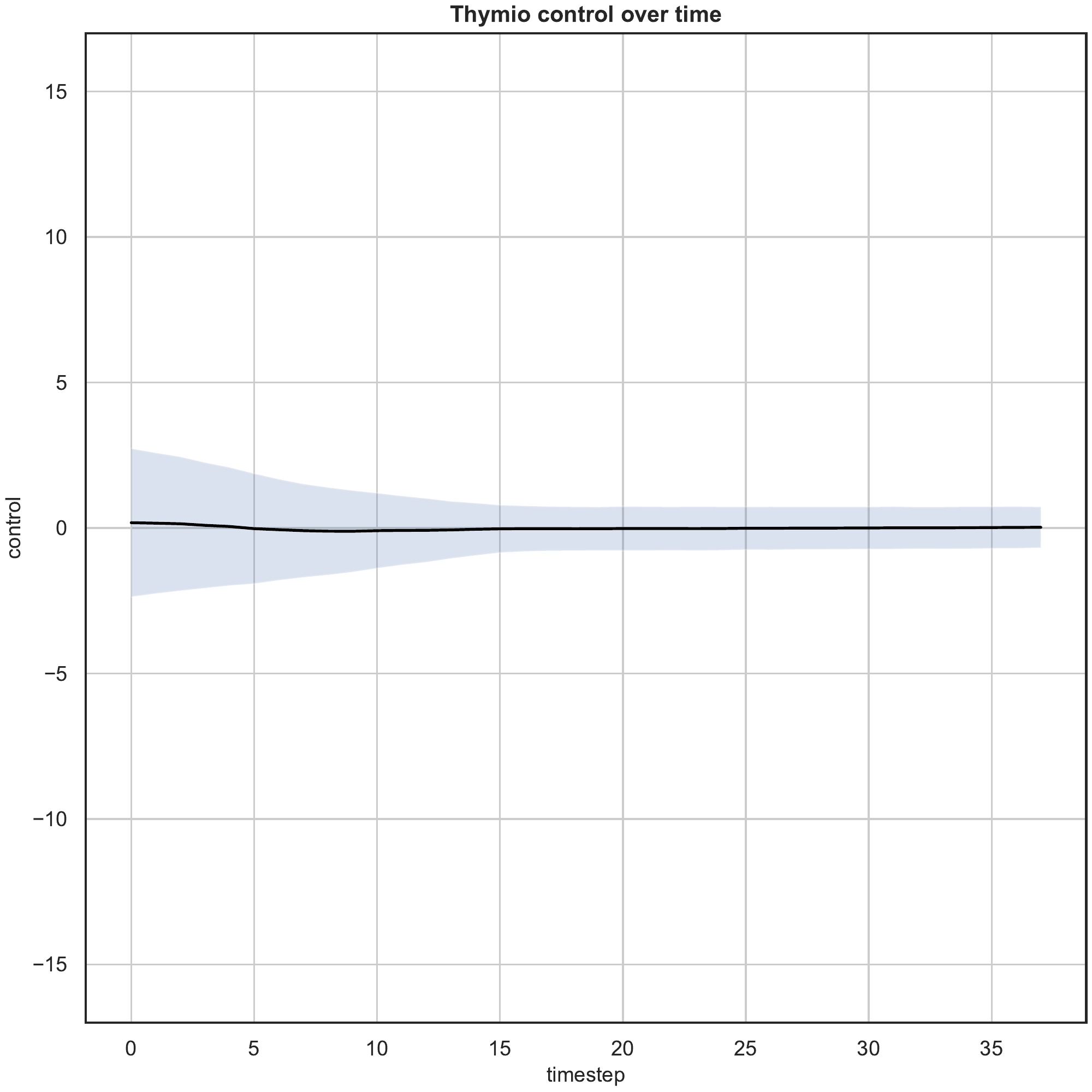}
		\caption{Distributed controller.}
	\end{subfigure}
	\caption[Evaluation of the control decided by \texttt{net-d6}.]{Comparison 
		of output control decided using three controllers: the expert, the manual 
		and the one learned from \texttt{net-d6}.}
	\label{fig:net-d6control}
\end{figure}

Figure \ref{fig:net-d6responsesensors} visualises the response of the learned 
controller as the input sensing changes, analysing the same two cases as before. 
Despite the behaviour is the same obtained using \texttt{prox\_values} when the 
robot sees only behind, this time the trend is different when the robot sees 
nothing behind: since the robot has to move backwards, a negative speed is 
always returned, that is higher when the obstacle is far. 
\begin{figure}[!htb]
	\centering
	\begin{subfigure}[h]{0.49\textwidth}
		\centering
		\includegraphics[width=\textwidth]{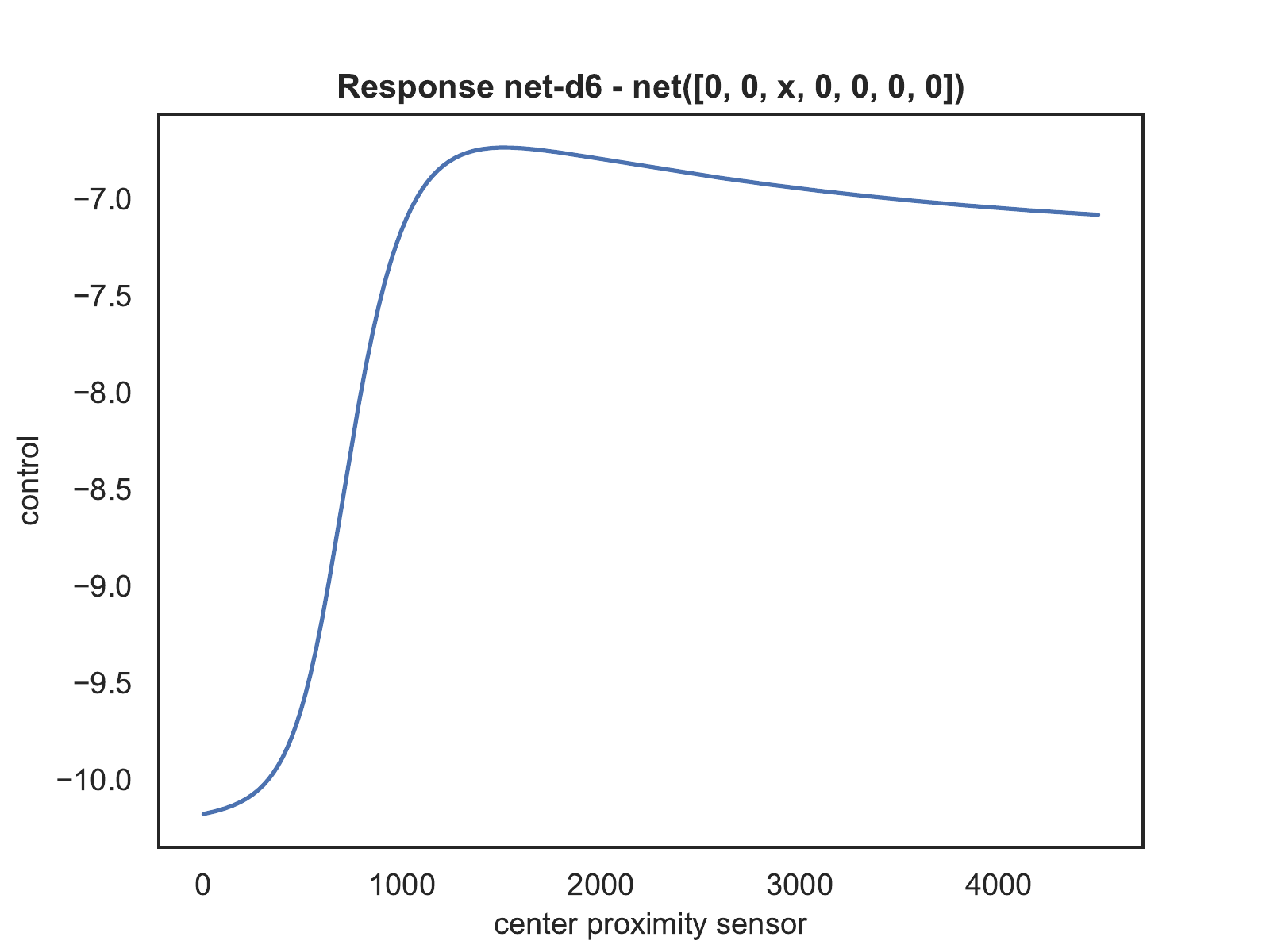}%
	\end{subfigure}
	\hfill
	\begin{subfigure}[h]{0.49\textwidth}
		\centering
		\includegraphics[width=\textwidth]{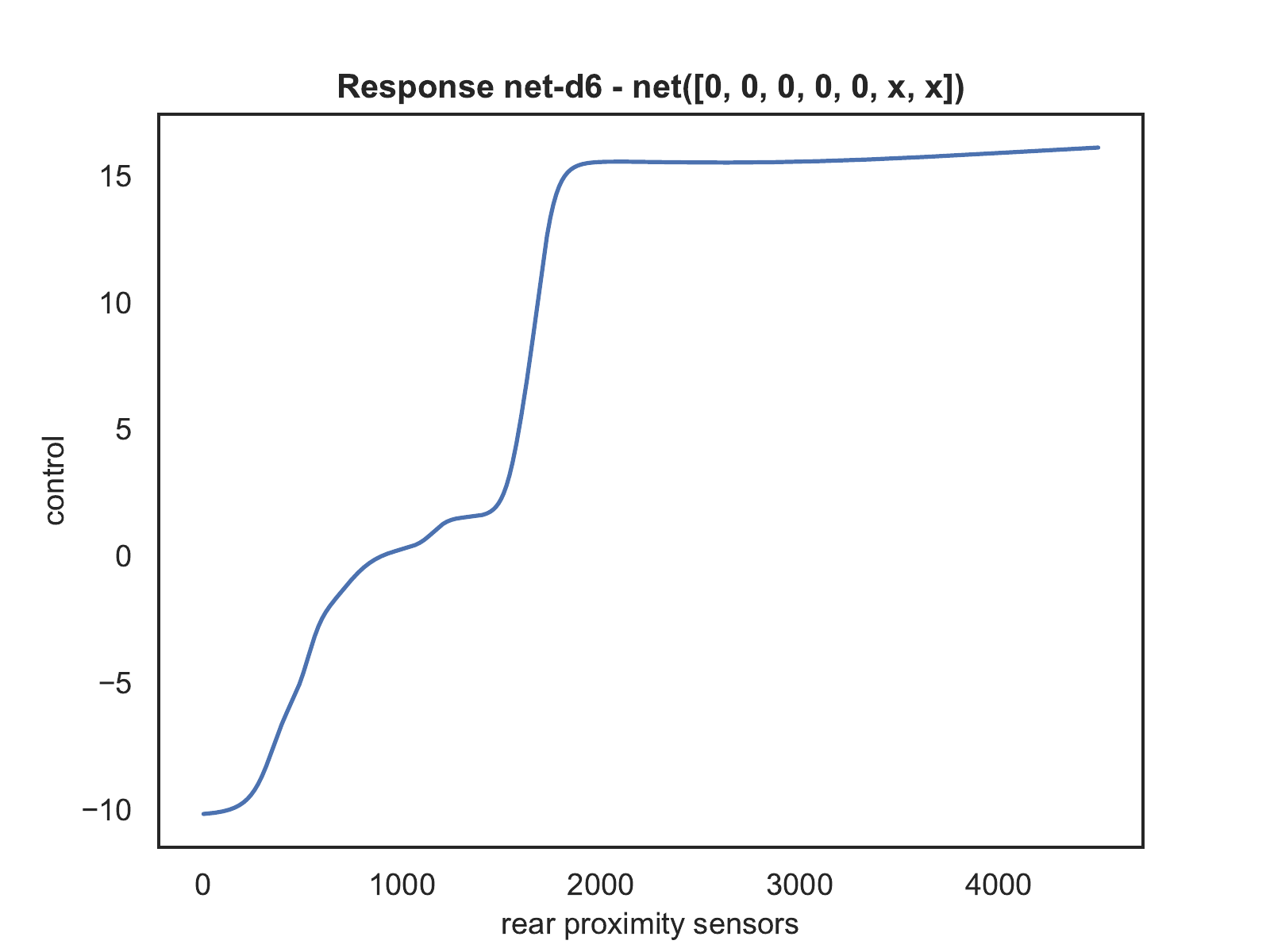}
	\end{subfigure}
	\caption{Response of \texttt{net-d6} by varying the input sensing.}
	\label{fig:net-d6responsesensors}
\end{figure}

In Figure \ref{fig:net-d6responseposition}, the behaviour of a robot 
located between other two that are already in their place is displayed.
\begin{figure}[!htb]
	\centering
	\includegraphics[width=.45\textwidth]{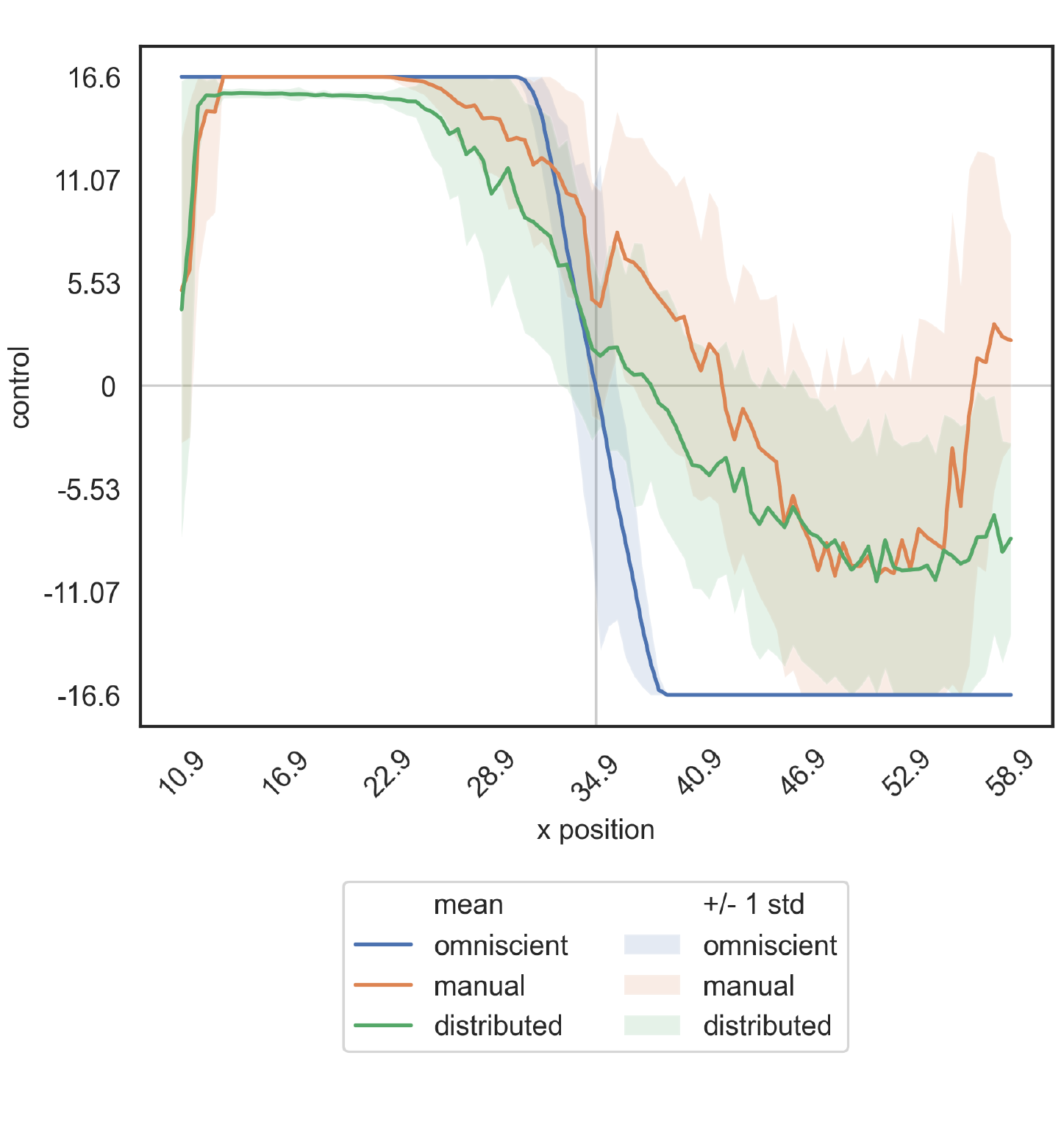}%
	\caption{Response of \texttt{net-d6} by varying the initial position.}
	\label{fig:net-d6responseposition}
\end{figure}
It visualises the response of the learned controller by varying the distance 
between two stationary agents and a robot located among them.
As expected, the output is a high positive value when the robot is close to an 
obstacle on the left, it decreases and reaches $0$ when the distance from 
right and left is equal, and finally becomes negative when there is an obstacle in 
front and not behind. 

\bigskip
Finally, in Figure \ref{fig:net-d6distance}, is presented the average distance of the 
robots from the target among all the simulations. The performance of the learned 
and the manual controllers are similar: in the final configuration they both are at 
about $5$cm from the target. 
\begin{figure}[!htb]
	\centering
	\includegraphics[width=.65\textwidth]{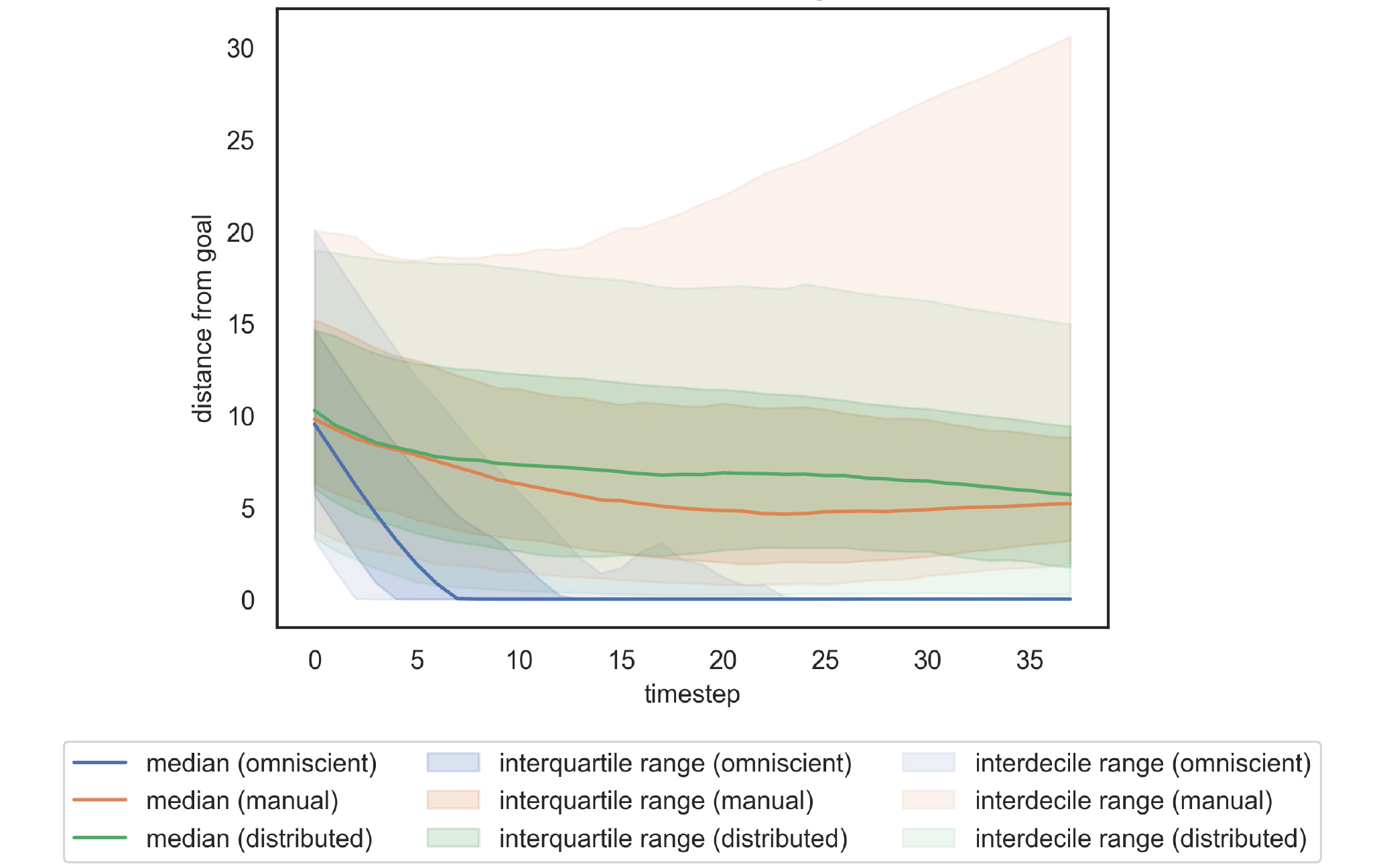}%
		\caption[Evaluation of \texttt{net-d6} distances from goal.]{Comparison of 
		performance in terms of distances from goal obtained using three 
		controllers: the expert, the manual and the one learned from \texttt{net-d6}.}
	\label{fig:net-d6distance}
\end{figure}

We conclude the first group of experiments presenting the results obtained 
using both types of input together from which we expect a more stable and 
robust behaviour. 

\paragraph*{Results using \texttt{all\_sensors} input}
In Figure \ref{fig:distlossall}, an analysis of the losses shows that using 
\texttt{all\_sensors} inputs the network is able to work well with all the 
gaps. 
\begin{figure}[!htb]
	\centering
	\includegraphics[width=.8\textwidth]{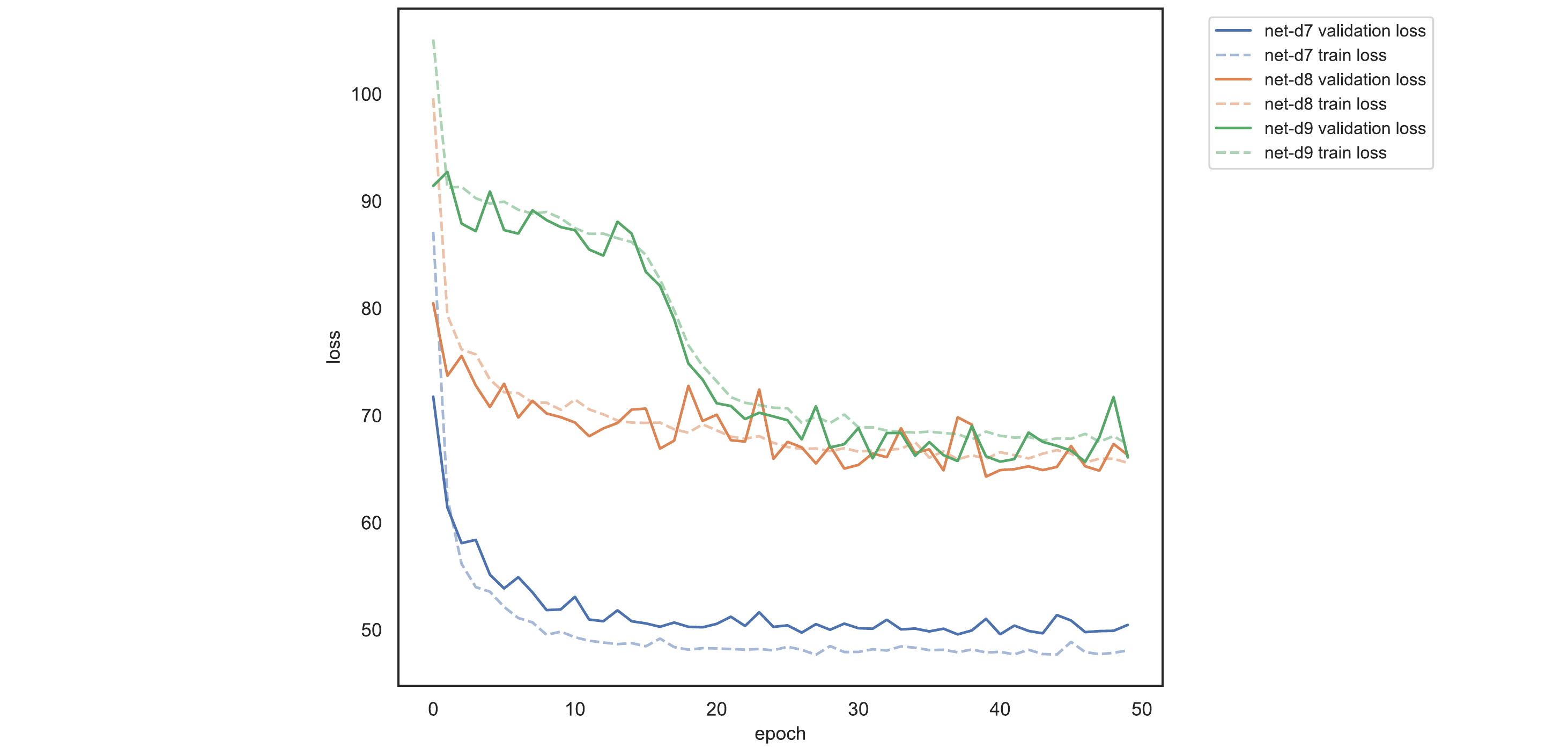}%
	\caption{Comparison of the losses of the models that use \texttt{all\_sensors} 
		readings.}
	\label{fig:distlossall}
\end{figure}

Examining the \ac{r2} coefficients in Figure \ref{fig:net-d789r2}, the behaviour 
obtained with \texttt{net-d7} and \texttt{net-d9} are the more promising. 
\begin{figure}[!htb]
	\begin{center}
		\begin{subfigure}[h]{0.49\textwidth}
 			\includegraphics[width=\textwidth]{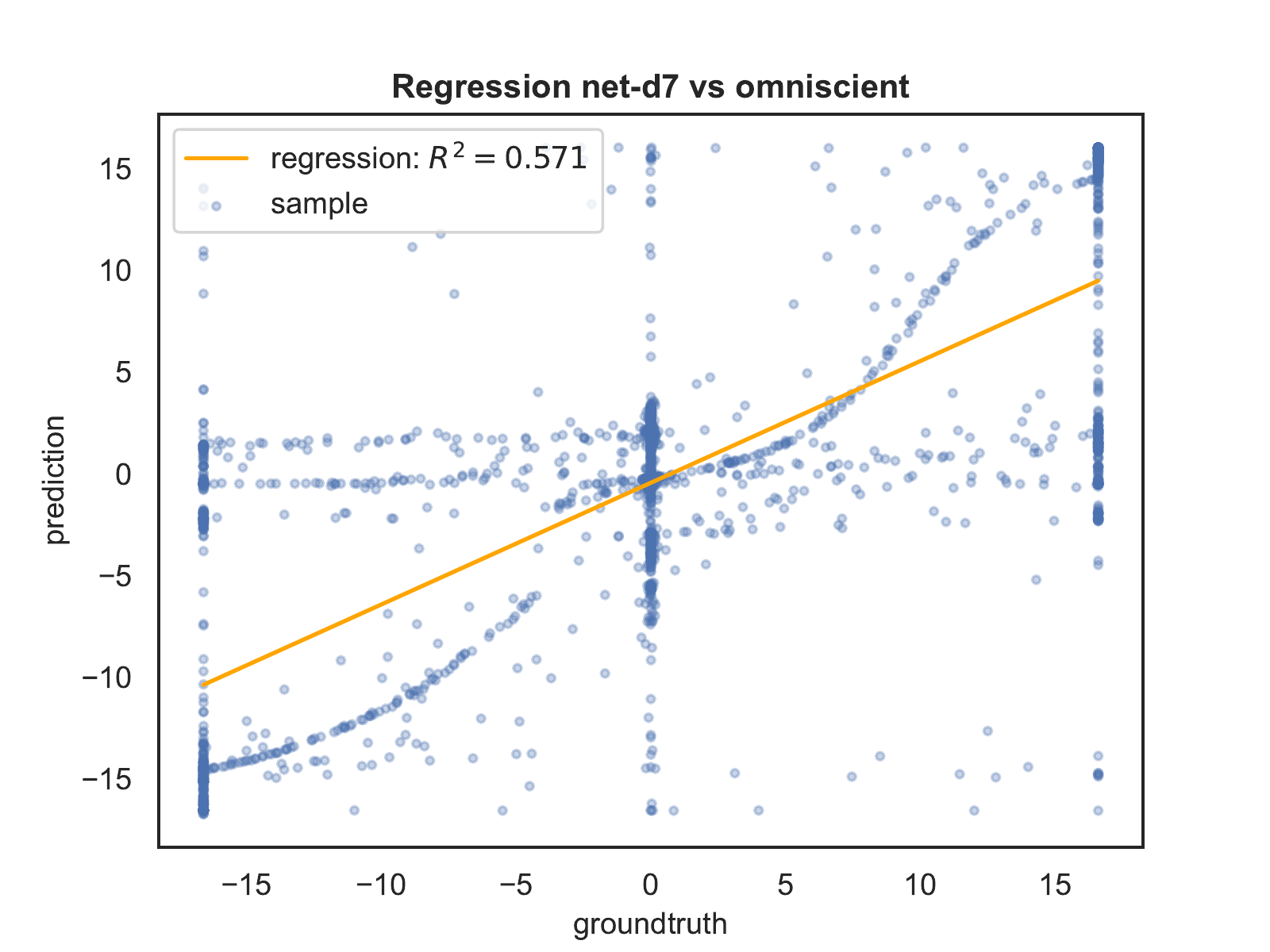}%
		\end{subfigure}
		\hfill
		\begin{subfigure}[h]{0.49\textwidth}
			\includegraphics[width=\textwidth]{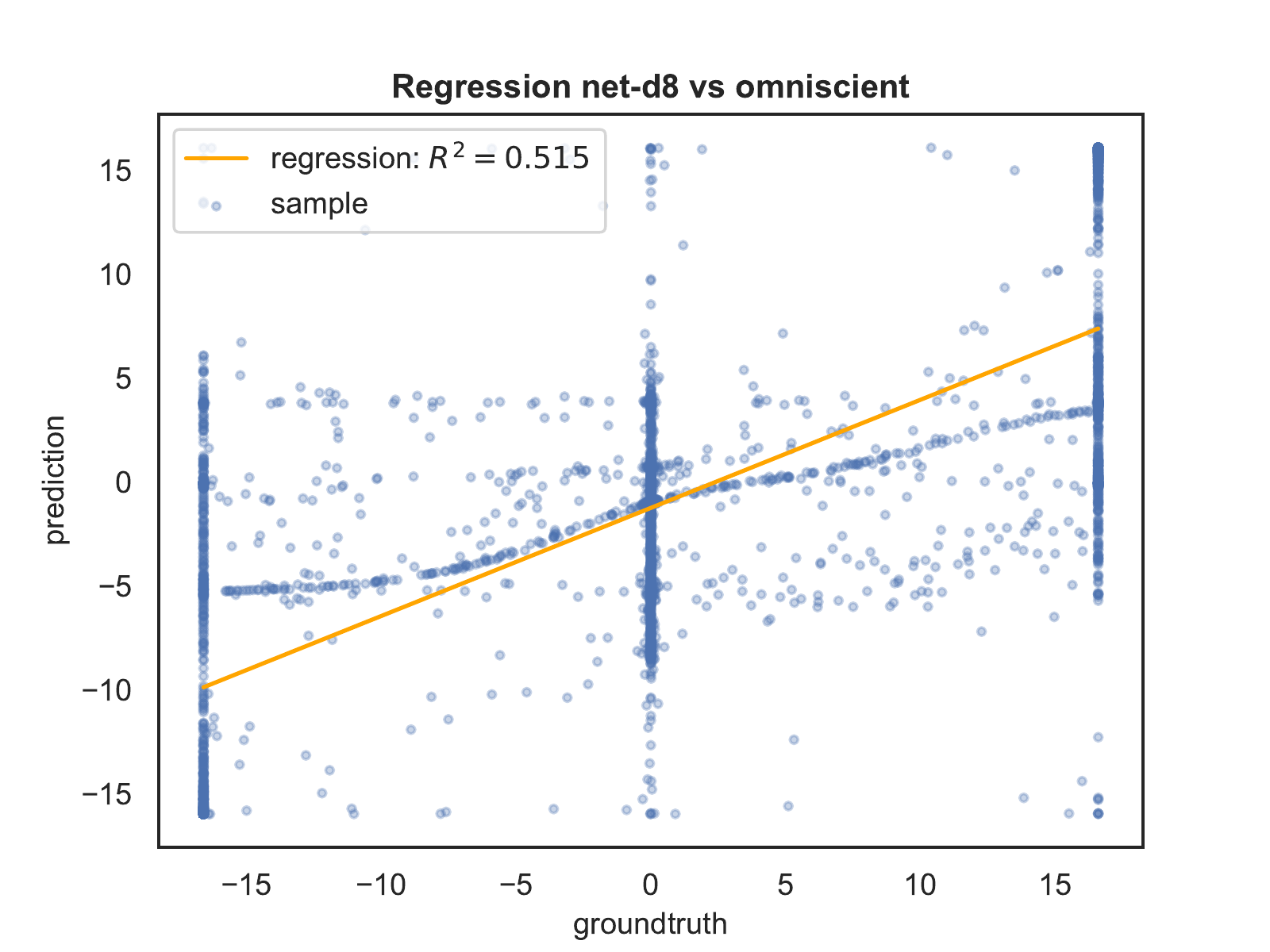}%
		\end{subfigure}
	\end{center}
	\hfil\vspace{-0.8cm}
	\begin{center}
		\begin{subfigure}[h]{0.49\textwidth}
			\includegraphics[width=\textwidth]{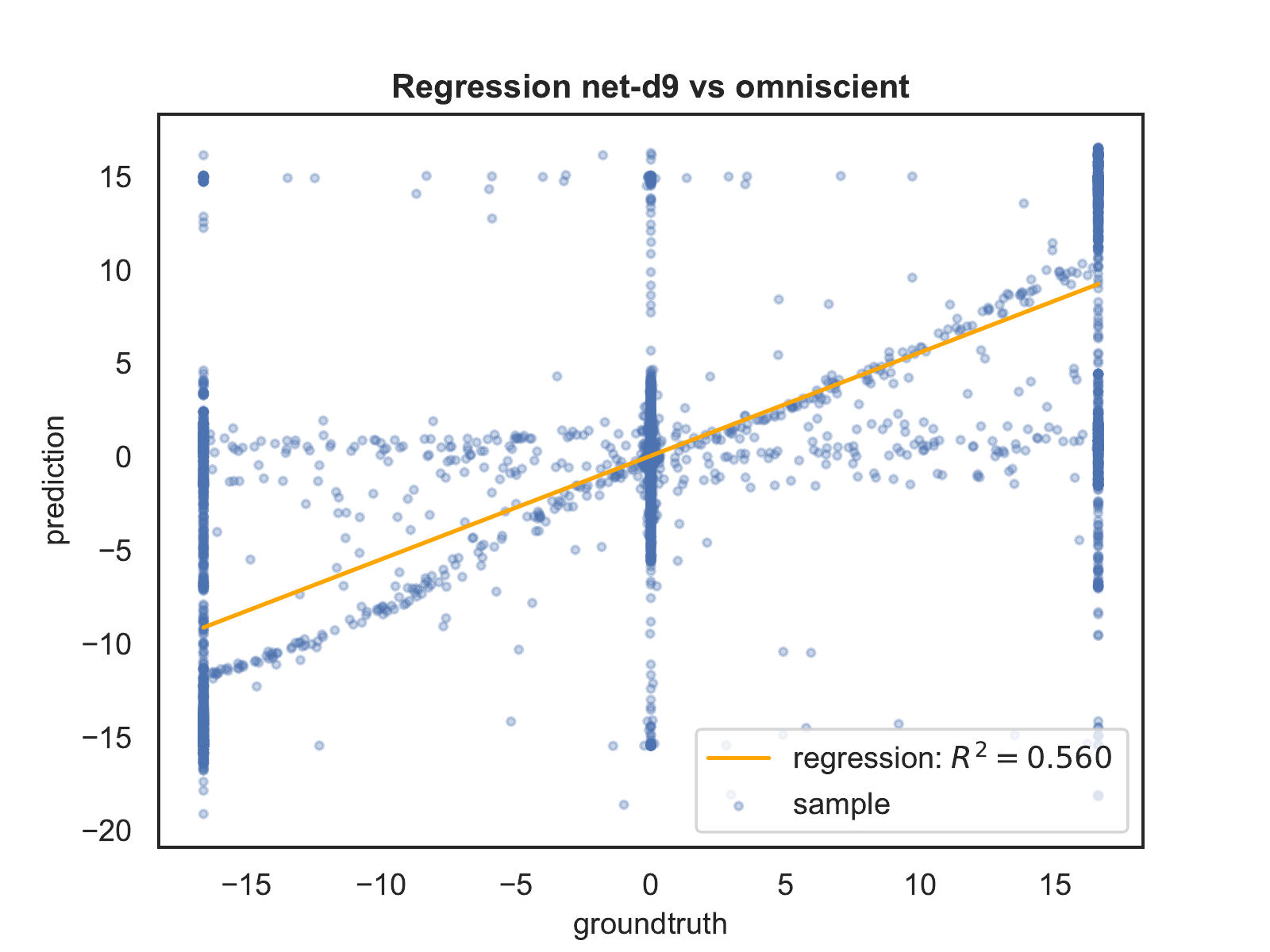}
		\end{subfigure}
	\end{center}
	\caption[Comparison of the \ac{r2} coefficients for \texttt{prox\_comm} 
	readings.]{Comparison of the \ac{r2} coefficients of the models that use 
		\texttt{prox\_comm} readings.}
	\label{fig:net-d789r2}
\end{figure}

\bigskip
Considering the more complex case, that is the one with the greatest average gap,
the superiority of this controller is further supported by the comparisons in Figure
\ref{fig:net-d9r2}. 
\begin{figure}[!htb]
	\centering
	\begin{subfigure}[h]{0.49\textwidth}
		\centering
		\includegraphics[width=\textwidth]{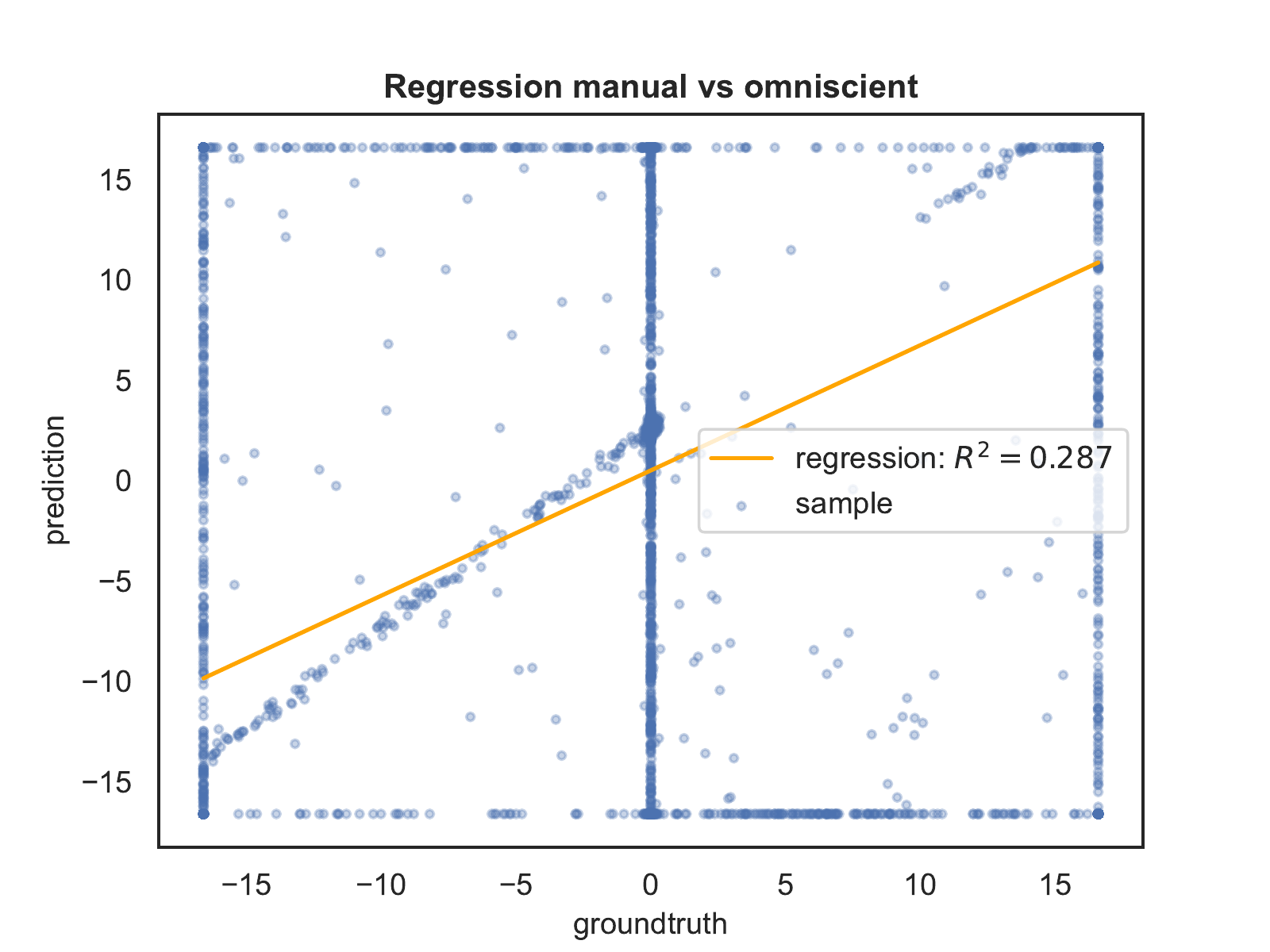}%
	\end{subfigure}
	\hfill
	\begin{subfigure}[h]{0.49\textwidth}
		\centering
		\includegraphics[width=\textwidth]{contents/images/net-d9/regression-net-d9-vs-omniscient}
	\end{subfigure}
	\caption[Evaluation of the \ac{r2} coefficients of \texttt{net-d9}.]{Comparison 
		of the \ac{r2} coefficient of the manual and the controller learned from 
		\texttt{net-d9} with respect to the omniscient one.}
	\label{fig:net-d9r2}
\end{figure}

In Figure \ref{fig:net-d9traj1} is shown a comparison of the trajectories for a 
sample simulation.
As before, the performance obtained using the omniscient and the learned 
controllers are comparable: both are very fast and in less than 10 time steps they 
reach the target. 
Instead, when the agents moved using the manual controller have almost 
approached the target, they start to oscillate. This problem is not surprising, since 
the parameters of the manual controller have been tuned on some specific cases 
and are always the same for all datasets, sometimes it does not work correctly.
\begin{figure}[!htb]
	\centering
	\includegraphics[width=.65\textwidth]{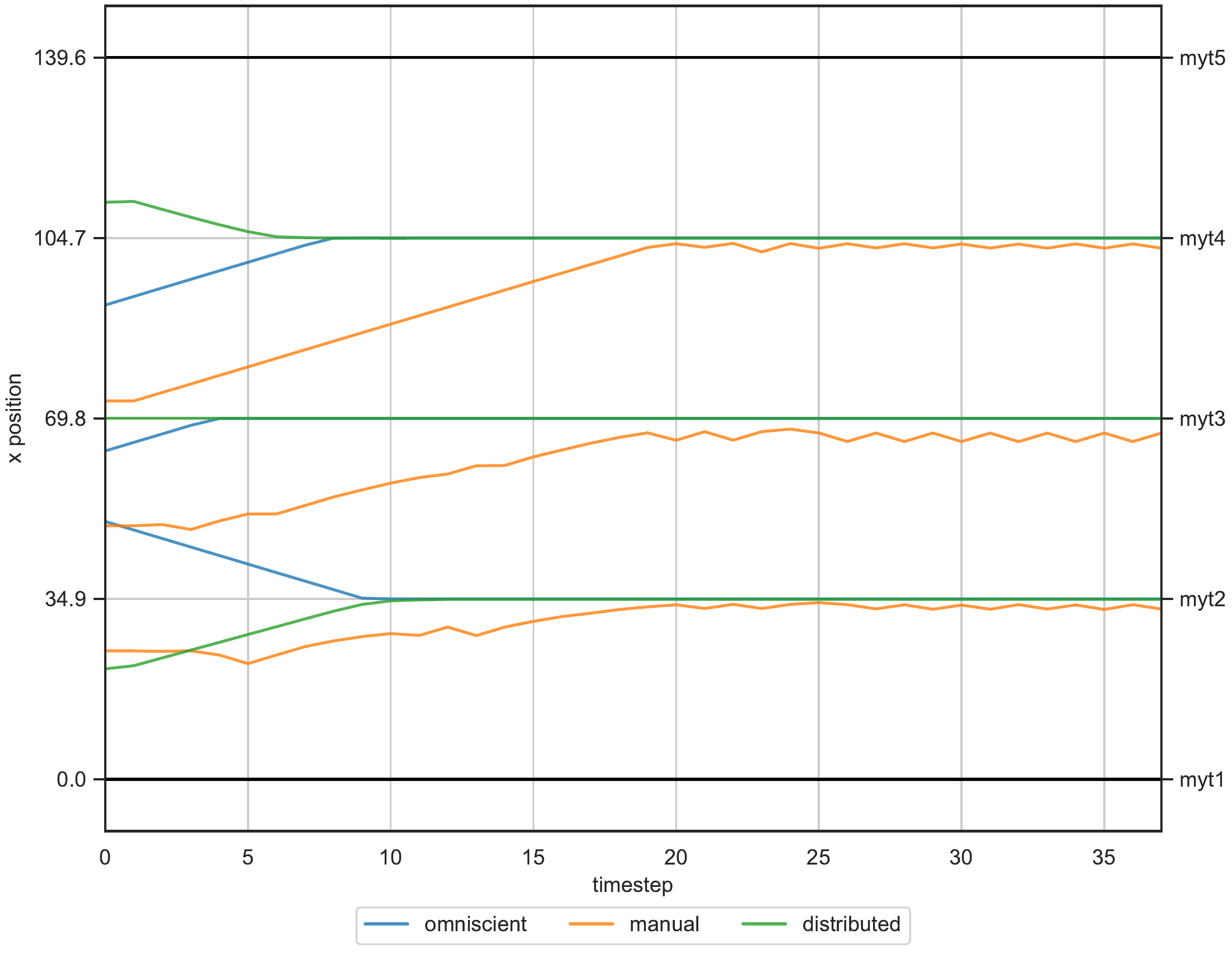}%
	\caption[Evaluation of the trajectories obtained with \texttt{all\_sensor} 
	input.]{Comparison of trajectories, of a single simulation, generated using three 
		controllers: the expert, the manual and the one learned from \texttt{net-d9}.}
	\label{fig:net-d9traj1}
\end{figure}  

\begin{figure}[!htb]
	\begin{center}
		\begin{subfigure}[h]{0.49\textwidth}
			\centering
			\includegraphics[width=.9\textwidth]{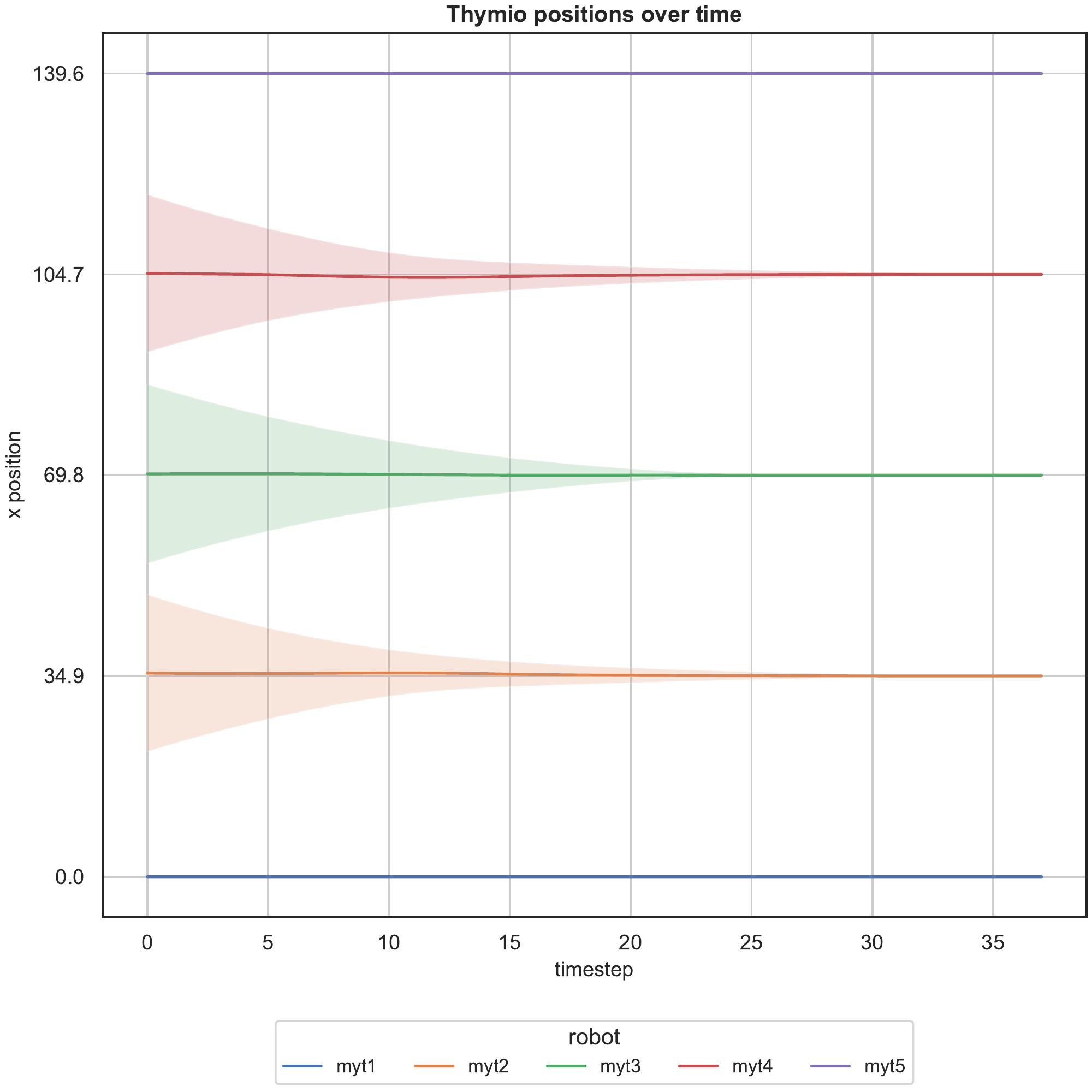}%
			\caption{Expert controller trajectories.}
		\end{subfigure}
		\hfill
		\begin{subfigure}[h]{0.49\textwidth}
			\centering
			\includegraphics[width=.9\textwidth]{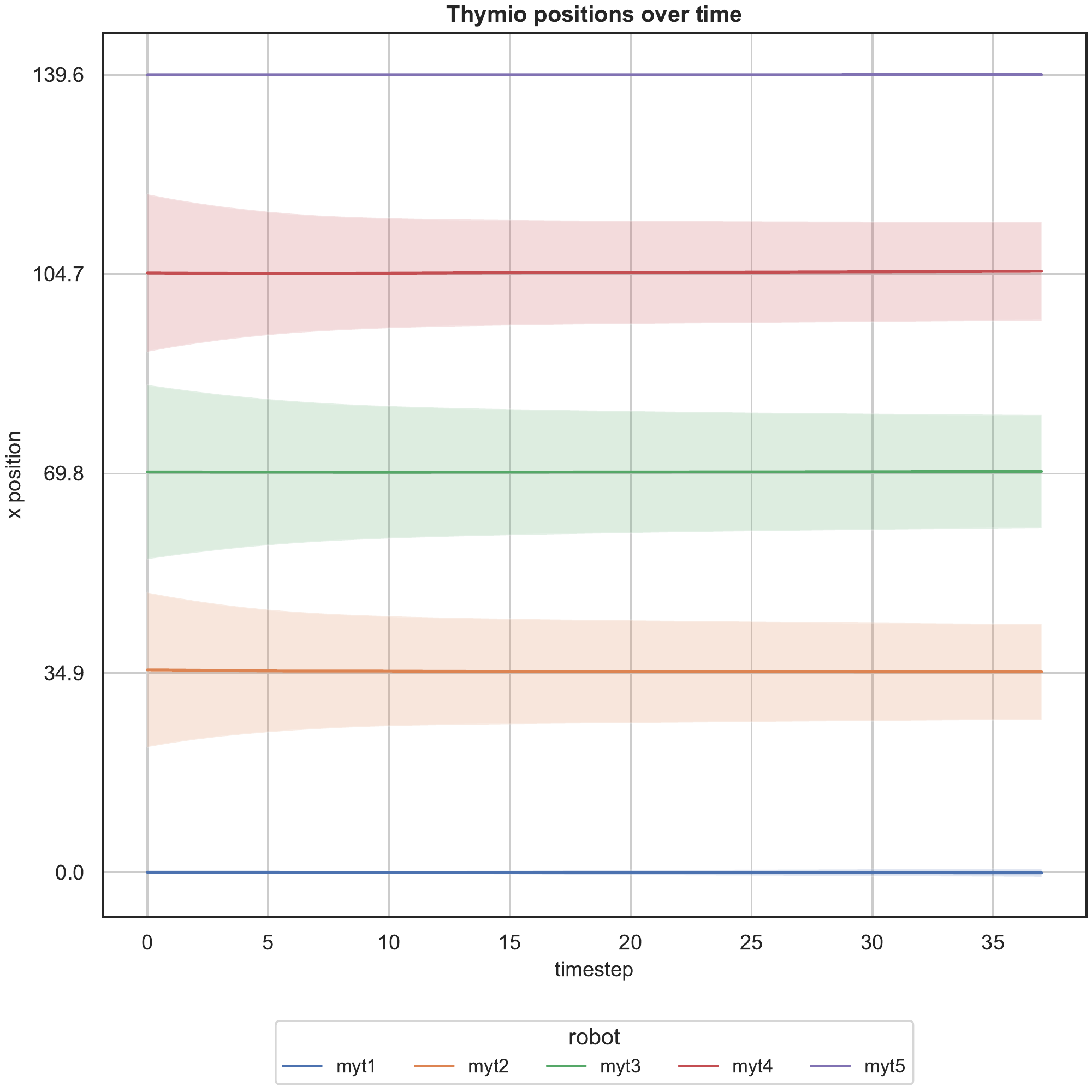}
			\caption{Distributed controller trajectories.}
		\end{subfigure}
	\end{center}
\vspace{-0.5cm}
	\caption[Evaluation of the trajectories learned by 
\texttt{net-d9}.]{Comparison of trajectories, of all the simulation runs, 
	generated using three controllers: the expert, the manual and the one learned 
	from \texttt{net-d9}.}
\end{figure}
\begin{figure}[!htb]\ContinuedFloat
	\begin{center}
		\begin{subfigure}[h]{0.49\textwidth}
			\centering
			\includegraphics[width=.9\textwidth]{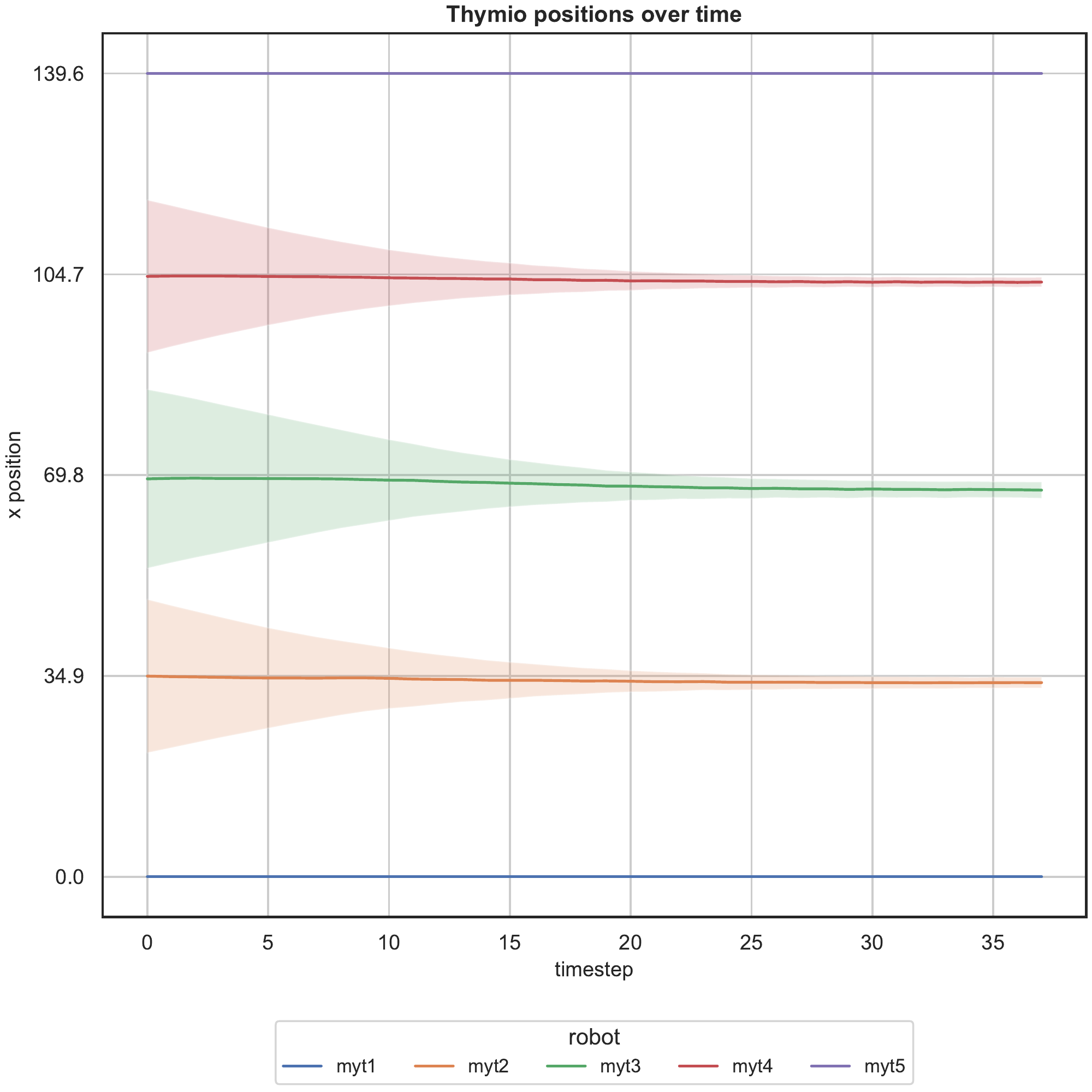}%
			\caption{Manual controller trajectories.}
		\end{subfigure}
		\hfill
		\begin{subfigure}[h]{0.49\textwidth}
			\centering
			\includegraphics[width=.9\textwidth]{contents/images/net-d9/position-overtime-learned_distributed}
			\caption{Distributed controller trajectories.}
		\end{subfigure}
	\end{center}
	\caption[]{Comparison of trajectories, of all the simulation runs, 
	generated using three controllers: the expert, the manual and the one learned 
	from \texttt{net-d9} (cont.).}
	\label{fig:net-d9traj}
\end{figure}

In Figure \ref{fig:net-d9traj} are shown trajectories obtained employing the three 
controllers. The convergence to the target is still slow, even if this time the expert 
needs less time steps than before. 
The manual controller does not show the same problem as before, while the 
learned controller is still the slowest to end up in the correct configuration.

Examining the evolution of the output control, in Figure \ref{fig:net-d9control}, 
the plots of the expert and the learned controllers are similar, although the speed 
in the second is much lower.
\begin{figure}[!htb]
	\centering
	\begin{subfigure}[h]{0.3\textwidth}
		\centering
		\includegraphics[width=\textwidth]{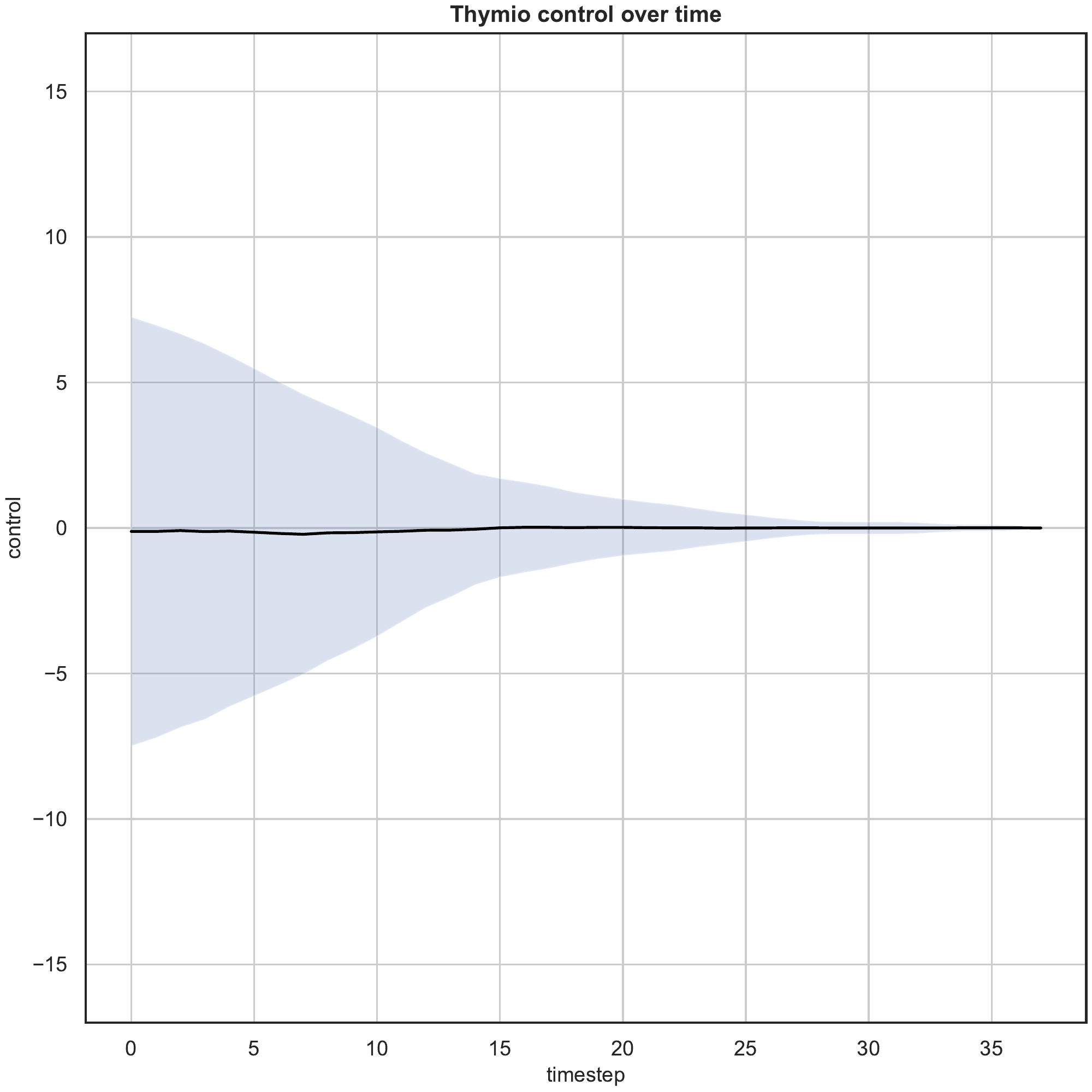}%
		\caption{Expert controller.}
	\end{subfigure}
	\hfill
	\begin{subfigure}[h]{0.3\textwidth}
		\centering
		\includegraphics[width=\textwidth]{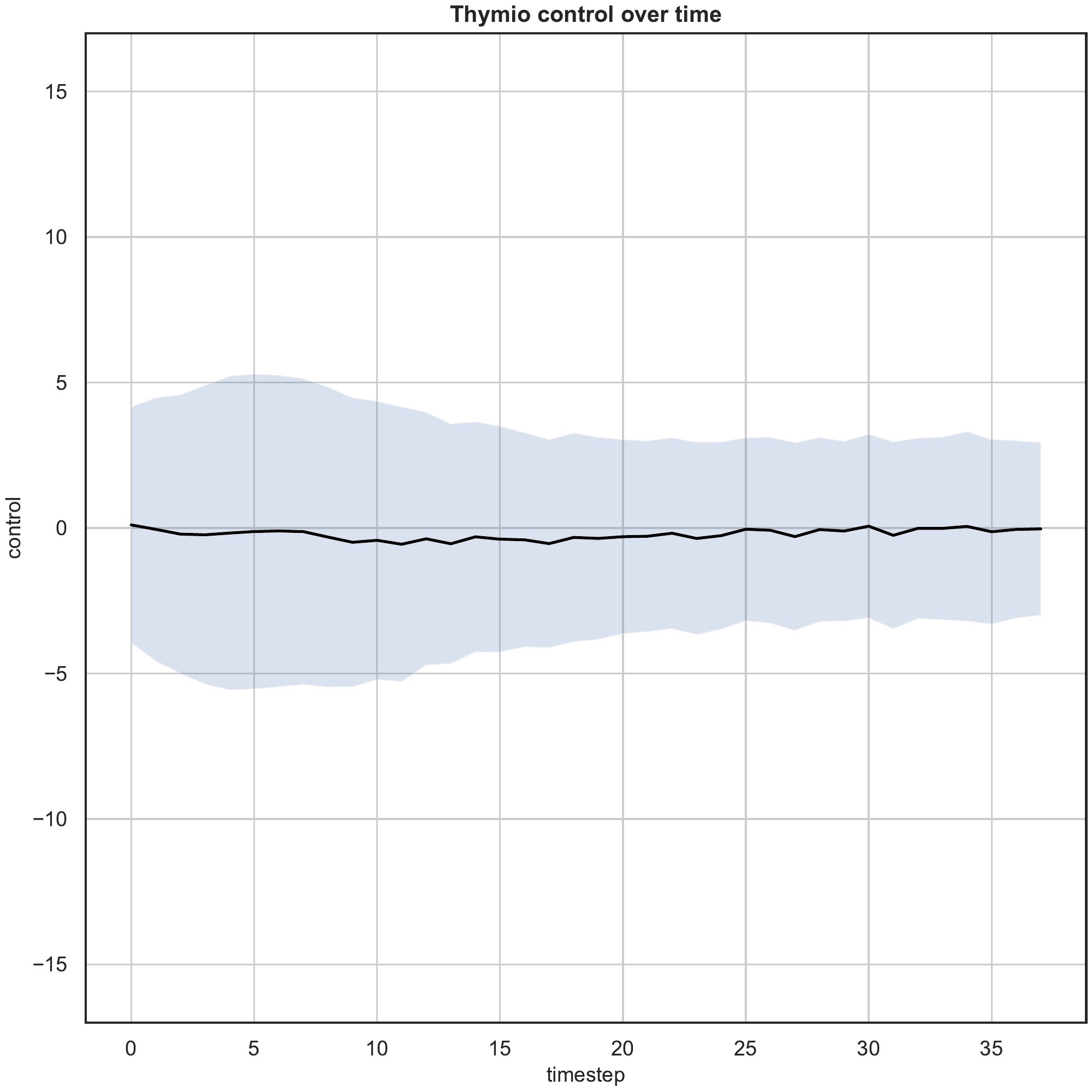}%
		\caption{Manual controller.}
	\end{subfigure}
	\hfill
	\begin{subfigure}[h]{0.3\textwidth}
		\centering
		\includegraphics[width=\textwidth]{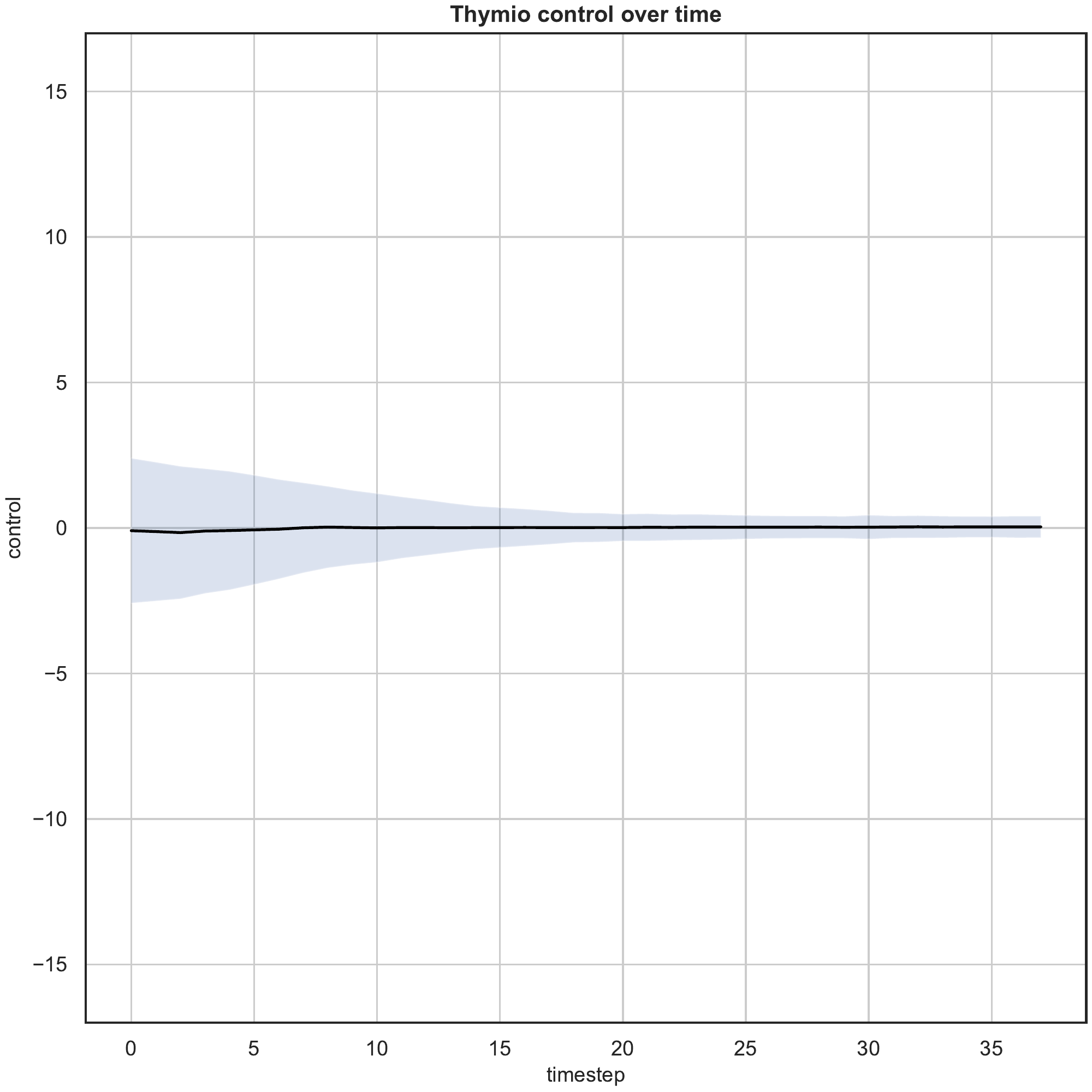}
		\caption{Distributed controller.}
	\end{subfigure}
	\caption[Evaluation of the control decided by \texttt{net-d9}.]{Comparison 
		of output control decided using three controllers: the expert, the manual 
		and the one learned from \texttt{net-d9}.}
	\label{fig:net-d9control}
\end{figure}

In Figure \ref{fig:net-d9responseposition} is displayed the behaviour of a robot 
located between other two that are already in their place.
This time the trend of the three curves shows how the behaviour of the model 
learned and of the manual controller are similar to that of the expert.
\begin{figure}[!htb]
	\centering
	\includegraphics[width=.45\textwidth]{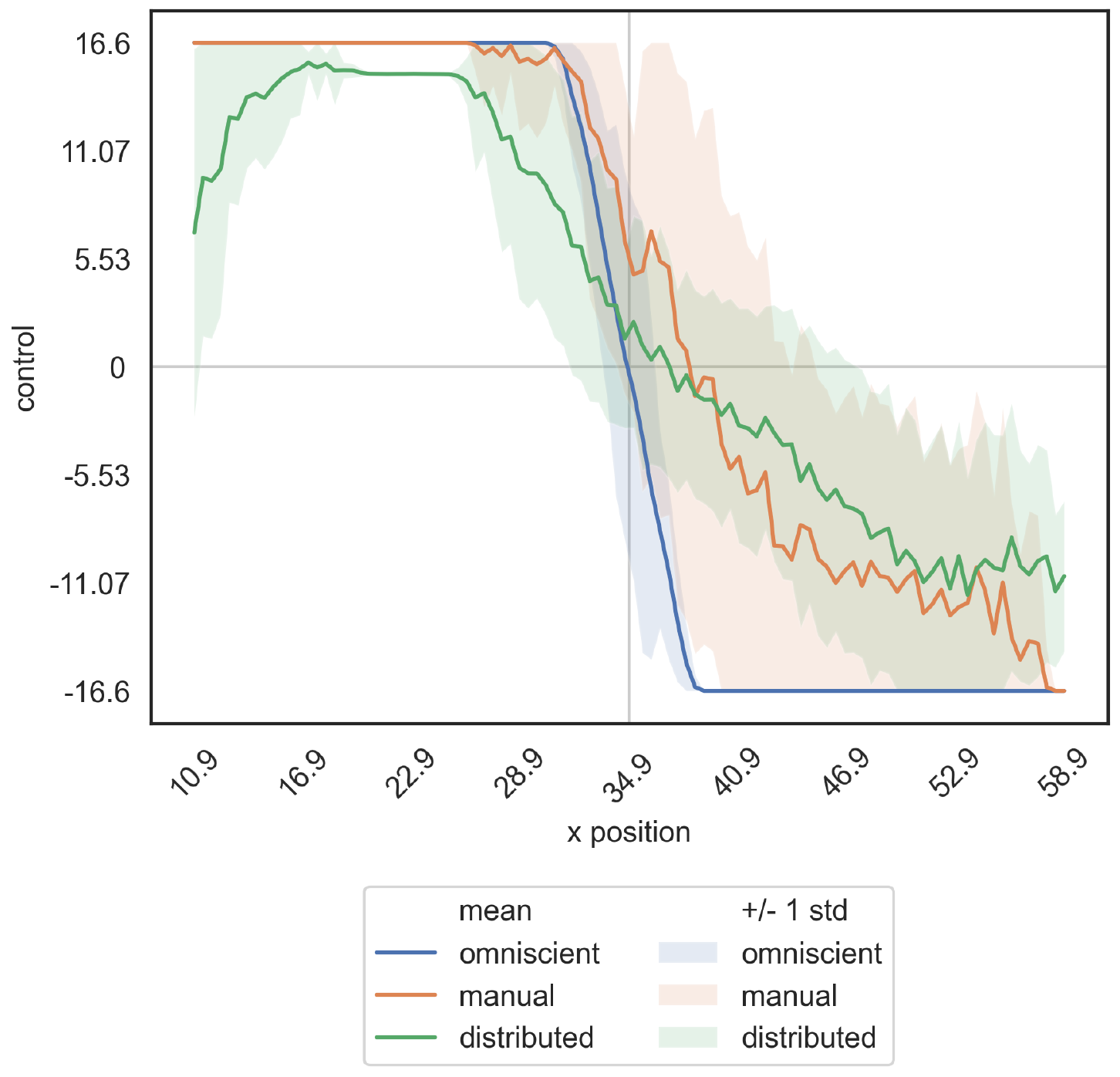}%
	\caption{Response of \texttt{net-d9} by varying the initial position.}
	\label{fig:net-d9responseposition}
\end{figure}

Finally, in Figure \ref{fig:net-d9distance}, is presented the absolute distance of 
each robot from the target, averaged on all robots among all the simulation runs. 
The median value is shown as well as the interquartile and interdecile ranges.
\begin{figure}[!htb]
	\centering
	\includegraphics[width=.63\textwidth]{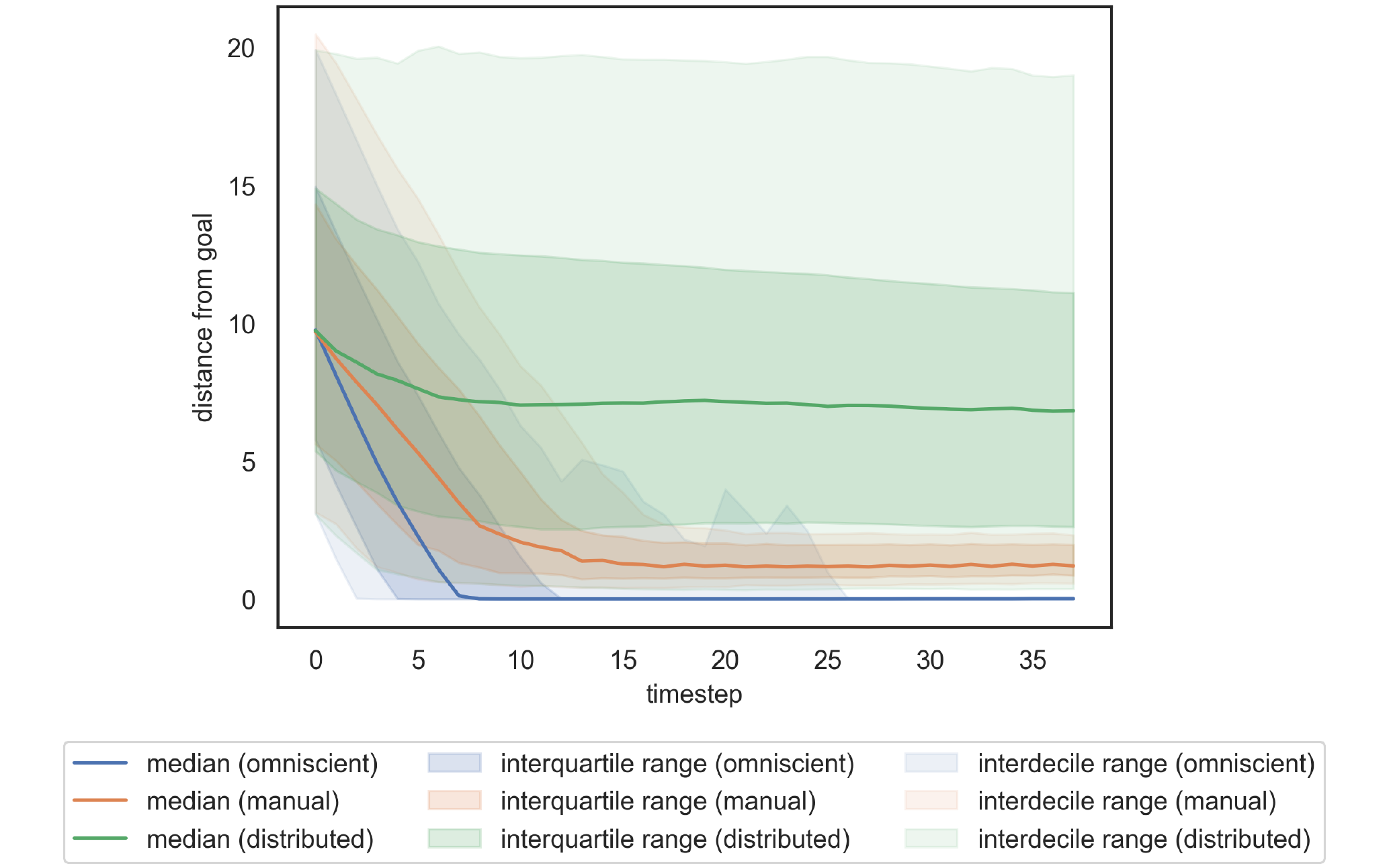}%
	\caption[Evaluation of \texttt{net-d9} distances from goal.]{Comparison of 
		performance in terms of distances from goal obtained using three 
		controllers: the expert, the manual and the one learned from \texttt{net-d9}.}
	\label{fig:net-d9distance}
\end{figure}
As anticipated by the trajectories in Figure \ref{fig:net-d9traj}, the controller 
learned from \texttt{net-d9} is slower to converge than the manual one. In fact, 
this plot confirms that the agents moved following a manual controller in the final 
configuration are closer to the target than those moved by the distributed 
controller: they are on average $2$ or $7$cm away from the goal position 
respectively.

\paragraph*{Summary}
We show in the figures below the losses of the trained models as the different 
inputs of the network vary, in particular, we represent with the blue, the orange 
and the green lines \texttt{prox\_values}, \texttt{prox\_comm} and
\texttt{all\_sensors} inputs respectively.
\begin{figure}[!htb]
	\centering
	\begin{subfigure}[h]{0.3\textwidth}
		\centering
		\includegraphics[width=\textwidth]{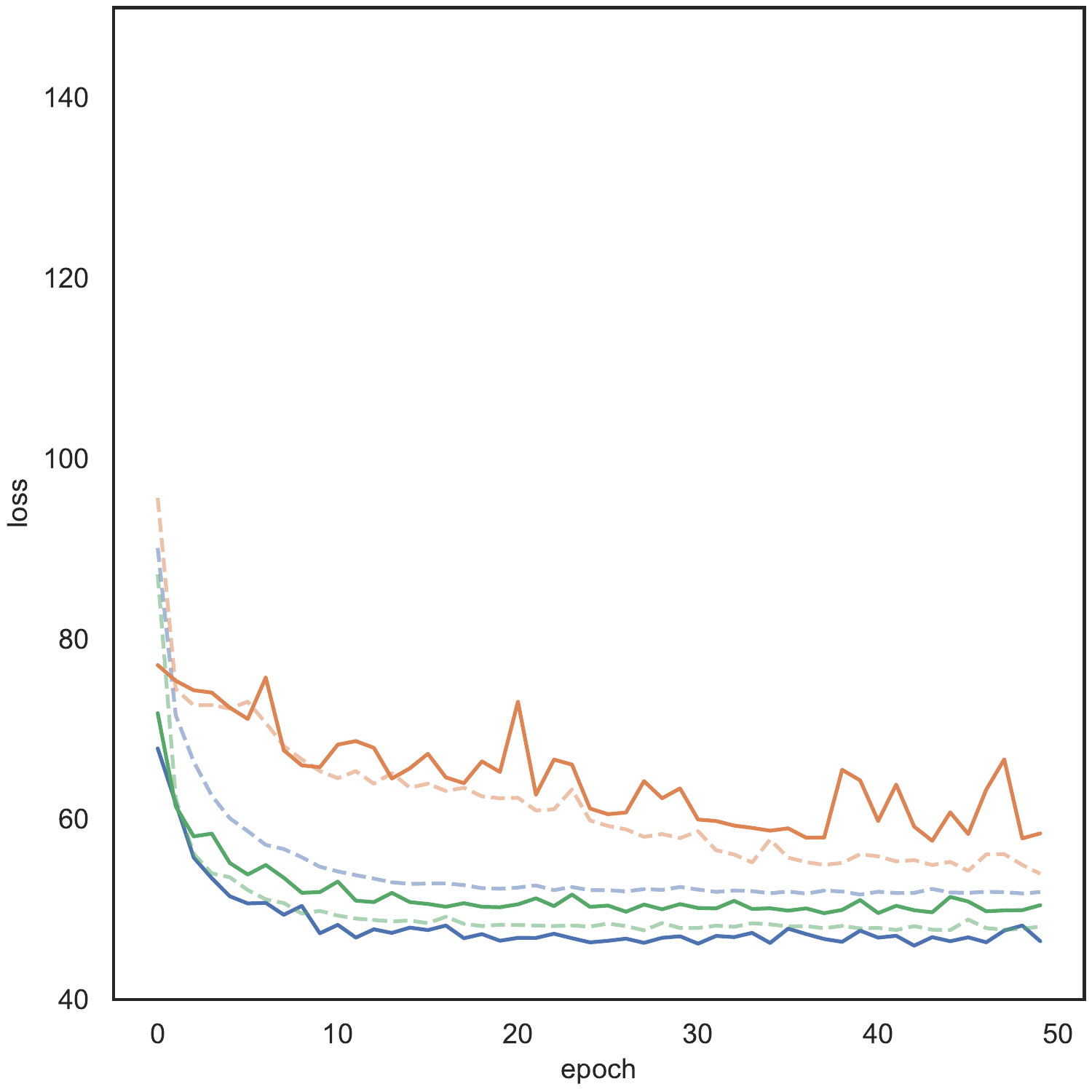}%
		\caption{\texttt{avg\_gap} of $8$cm.}
	\end{subfigure}
	\hfill
	\begin{subfigure}[h]{0.3\textwidth}
		\centering
		\includegraphics[width=\textwidth]{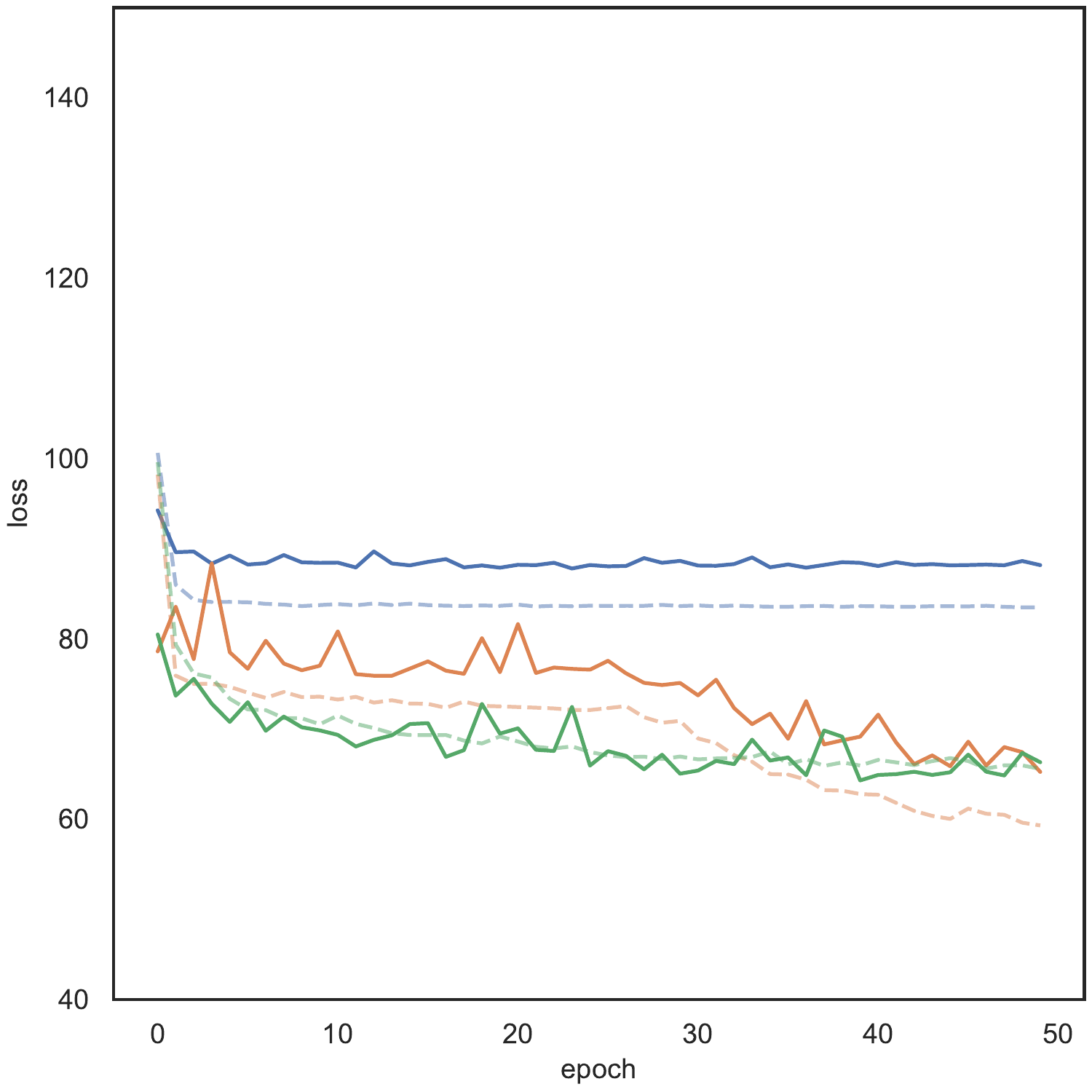}%
		\caption{\texttt{avg\_gap} of $13$cm.}
	\end{subfigure}
	\hfill
	\begin{subfigure}[h]{0.3\textwidth}
		\centering
		\includegraphics[width=\textwidth]{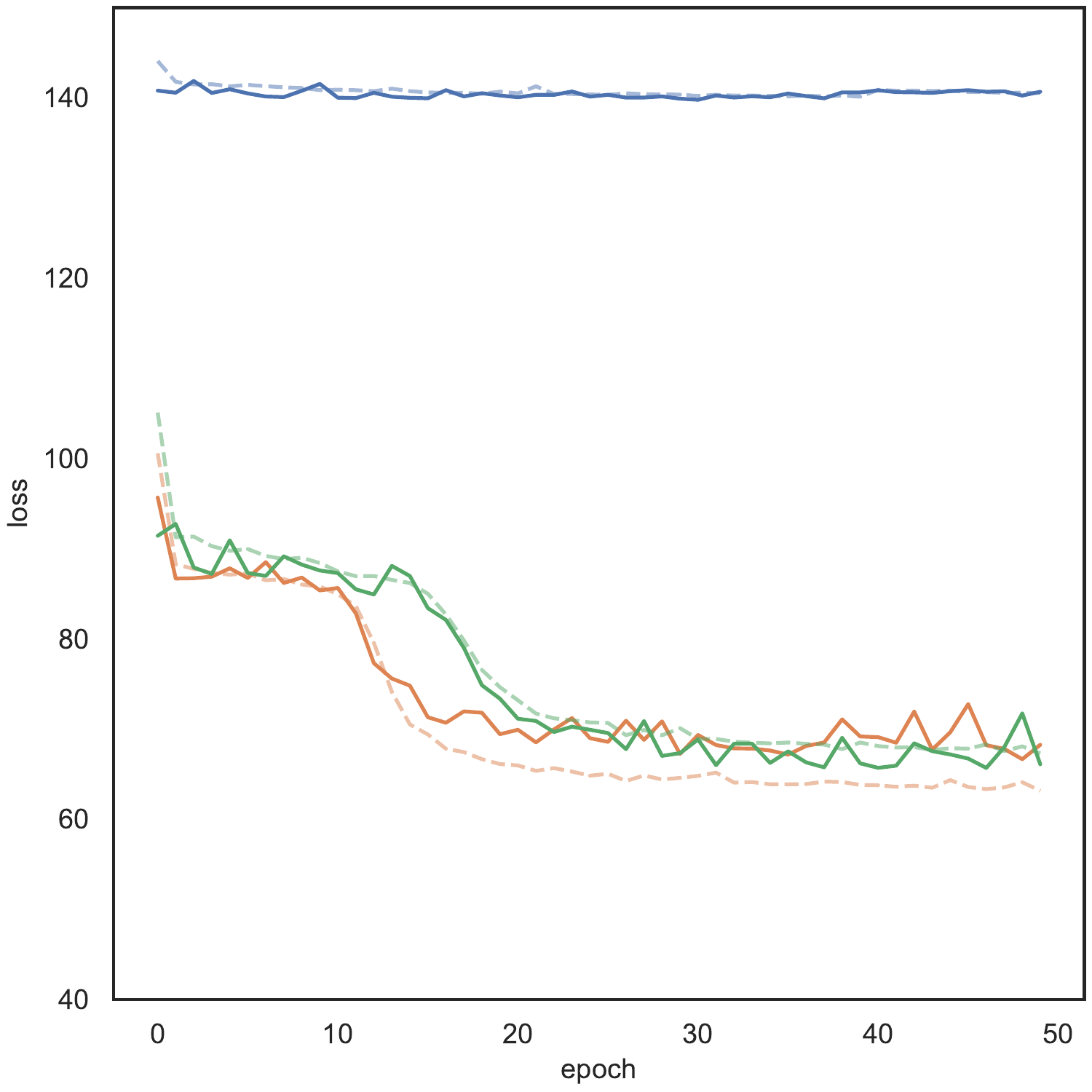}
		\caption{\texttt{avg\_gap} of $24$cm.}
	\end{subfigure}
	\caption[Losses summary of the first set of experiments.]{Comparison 
		of the losses by varying the input of the networks for the three gaps.}
	\label{fig:distloss81324}
\end{figure}

In case of an \texttt{avg\_gap} of $8$cm, the model trained using 
\texttt{prox\_values} has a lower loss, following is the network that employ 
\texttt{all\_sensors}, with similar results, and finally the model that works with 
\texttt{prox\_comm}.
This performance is expected, as well as the fact that \texttt{all\_sensors} cannot 
perform better than \texttt{prox\_values} with small gaps, since in this case the 
data coming from \texttt{prox\_comm} contains only zeros in the the second half 
of the array, making it unusable.
In a complementary way, by increasing the gap to $13$cm,  
\texttt{prox\_values} alone is not able to achieve satisfactory results, while used 
together with \texttt{prox\_comm}, \texttt{all\_sensors} reaches good 
performances that on the validation set are comparable to those obtained using 
\texttt{prox\_comm} alone.
Finally, by increasing the gap even more, up to $24$cm, 
\texttt{prox\_values} becomes completely unusable, while \texttt{prox\_comm} 
and \texttt{all\_sensors} still have excellent performances similar to each other.

\subsubsection{Experiment 2: variable number of agents}
\label{subsubsec:task1-exp-distr-2}
The second group of experiments we carried out using a distributed approach, 
examines the behaviour of the control learned using \texttt{all\_sensors} inputs. 
\begin{figure}[H]
	\centering
	\begin{tabular}{ccccc}
		\toprule
		\textbf{Model} \quad & \textbf{\texttt{network\_input}} & 
		\textbf{\texttt{input\_size}} & \textbf{\texttt{avg\_gap}} & \textbf{\texttt{N}}\\
		\midrule
		\texttt{net-d10} 	& \texttt{all\_sensors}		&  $14$  &  $8$		 	 &	$5$ \\
		\texttt{net-d11} 	& \texttt{all\_sensors}		&  $14$  &  $20$		&	$5$ \\
		\texttt{net-d12} 	& \texttt{all\_sensors}		&  $14$  &  variable   &	$5$ \\
		\texttt{net-d13} 	& \texttt{all\_sensors}	  	&  $14$  &  $8$			 &	  $8$ \\
		\texttt{net-d14} 	& \texttt{all\_sensors}	  	&  $14$  &  $20$   		&	 $8$ \\
		\texttt{net-d15} 	& \texttt{all\_sensors}	  	&  $14$  &  variable	&	 $8$ \\
		\texttt{net-d16} 	& \texttt{all\_sensors}	  	&  $14$  &  $ 8$		  &	 variable\\
		\texttt{net-d17} 	& \texttt{all\_sensors}	  	&  $14$  &  $20$		 &	variable\\
		\texttt{net-d18} 	& \texttt{all\_sensors}	  	&  $14$  &  variable	 &	
		variable\\
		\bottomrule
	\end{tabular}
	\captionof{table}[Experiments with variable agents and gaps (no 
	communication).]{List of the experiments carried out using a variable 
		number of agents and of gaps.}
	\label{tab:modeldist}
\end{figure}
In this situation, the simulation runs use a different number of robots $N$, that 
can be fixed at $5$ or $8$ for the entire simulation, or even vary in the range $[5, 
10]$. The same reasoning is applied to the choice of the \texttt{avg\_gap}, that 
can be a fixed value in all the runs, chosen between $8$ or $20$, but also vary in 
the range $[5, 24]$. 
The objective of this set of experiments, summarised in Table \ref{tab:modeldist}, 
is to verify the robustness of the models, proving that it is possible to train 
networks that handle a variable number of agents. 

First of all, we start by showing in Figure \ref{fig:distlossext} an overview of the 
models performance in terms of train and validation losses. 
\begin{figure}[!htb]
	\centering
	\includegraphics[width=.8\textwidth]{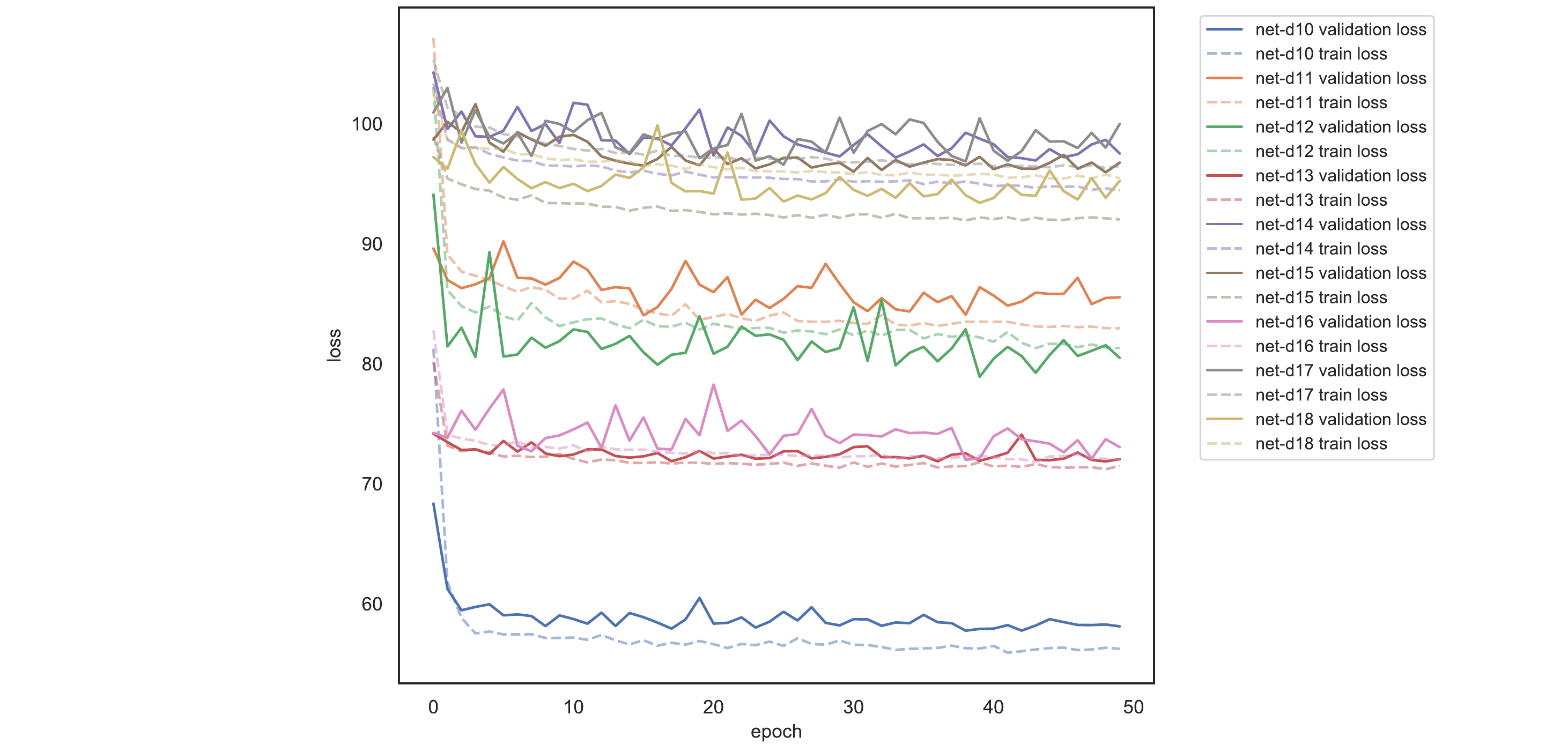}%
	\caption[Comparison of losses of the second set of 
	experiments.]{Comparison of the losses of the models carried out using a 
		variable number of agents and of average gap.}
	\label{fig:distlossext}
\end{figure}

\paragraph*{Results using 5 agents}
In Figure \ref{fig:distlossn5} are analysed the experiments performed using a 
fixed number of agents, the same used for the group of experiments presented 
above, i.e., $5$, in order to show the difference of performance using a gap that 
is first small, then large and finally variable.

\begin{figure}[!htb]
	\centering
	\includegraphics[width=.8\textwidth]{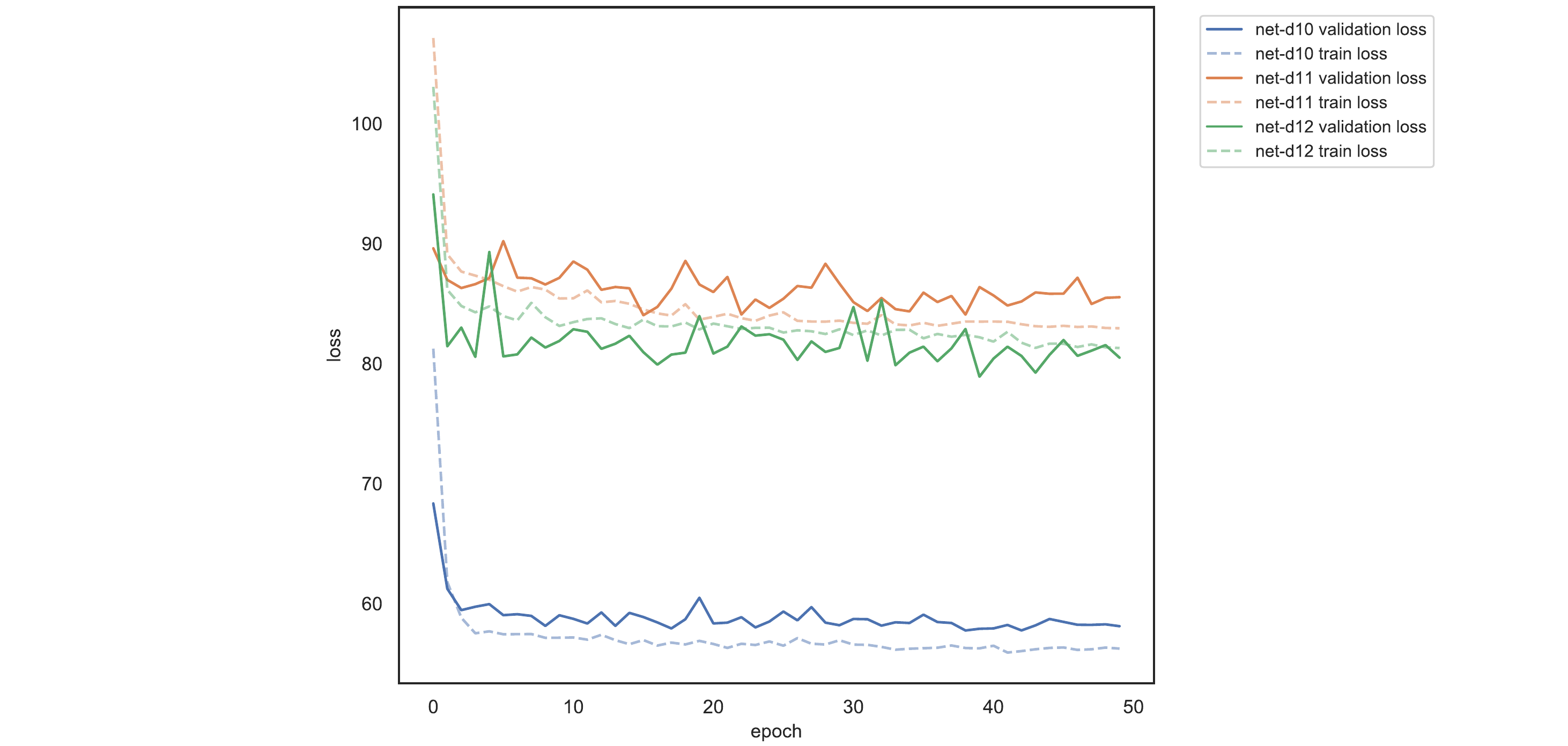}%
	\caption[Comparison of the losses of the models that use $5$ 
	agents.]{Comparison of the losses of the models that use $5$ agents as 
	the gap varies.}
	\label{fig:distlossn5}
\end{figure}

Examining more in detail the case in which the model is trained using a 
variable average gap, in Figure \ref{fig:net-d12r2} is visualised a comparison of 
the \ac{r2} of the manual and the learned controllers, on the validation set. 
The robots' behaviour using the learned instead of the manual controller is a 
bit better, even if far from the expert.
\begin{figure}[!htb]
	\centering
	\begin{subfigure}[h]{0.49\textwidth}
		\centering
		\includegraphics[width=\textwidth]{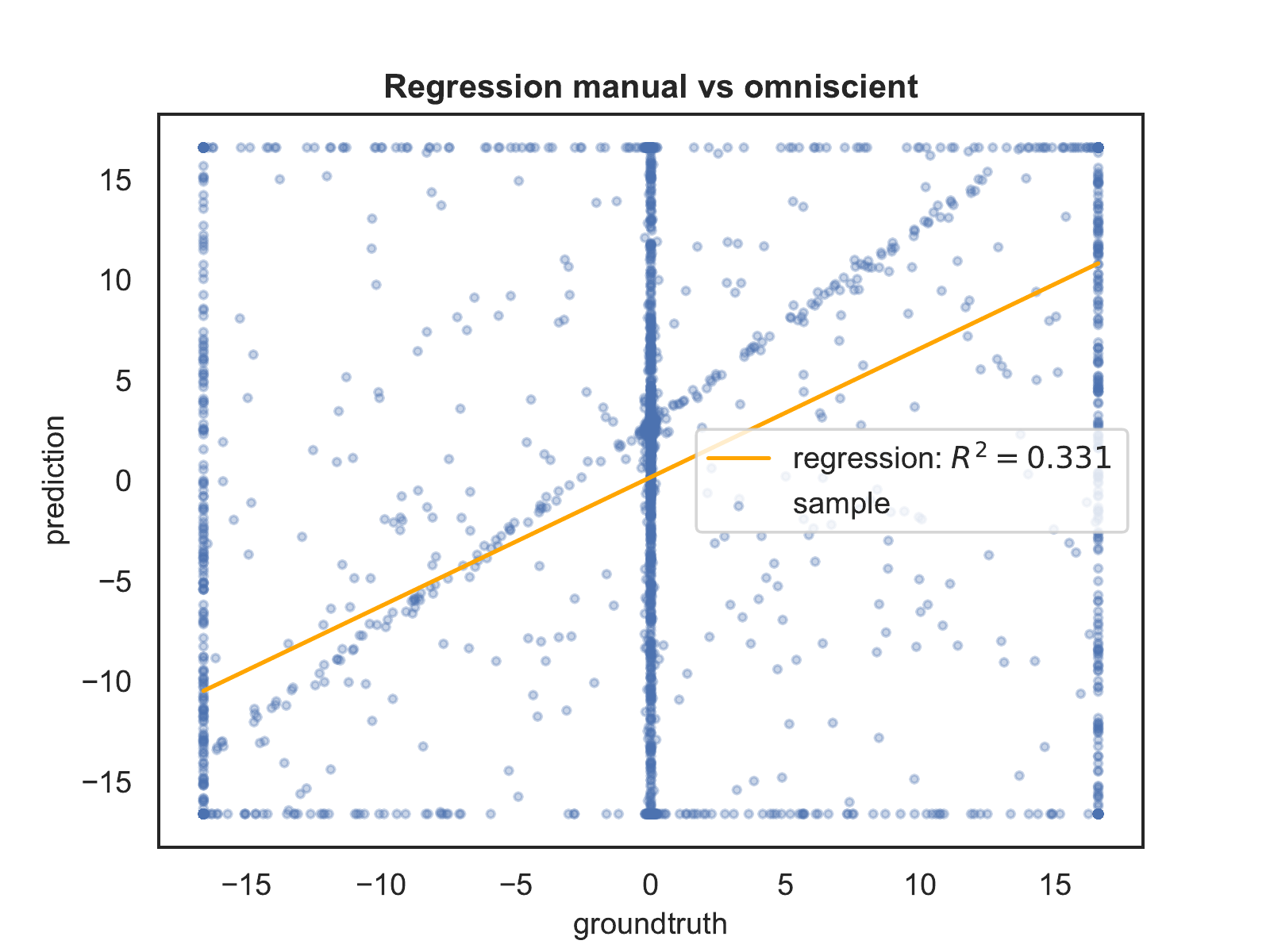}%
	\end{subfigure}
	\hfill
	\begin{subfigure}[h]{0.49\textwidth}
		\centering
		\includegraphics[width=\textwidth]{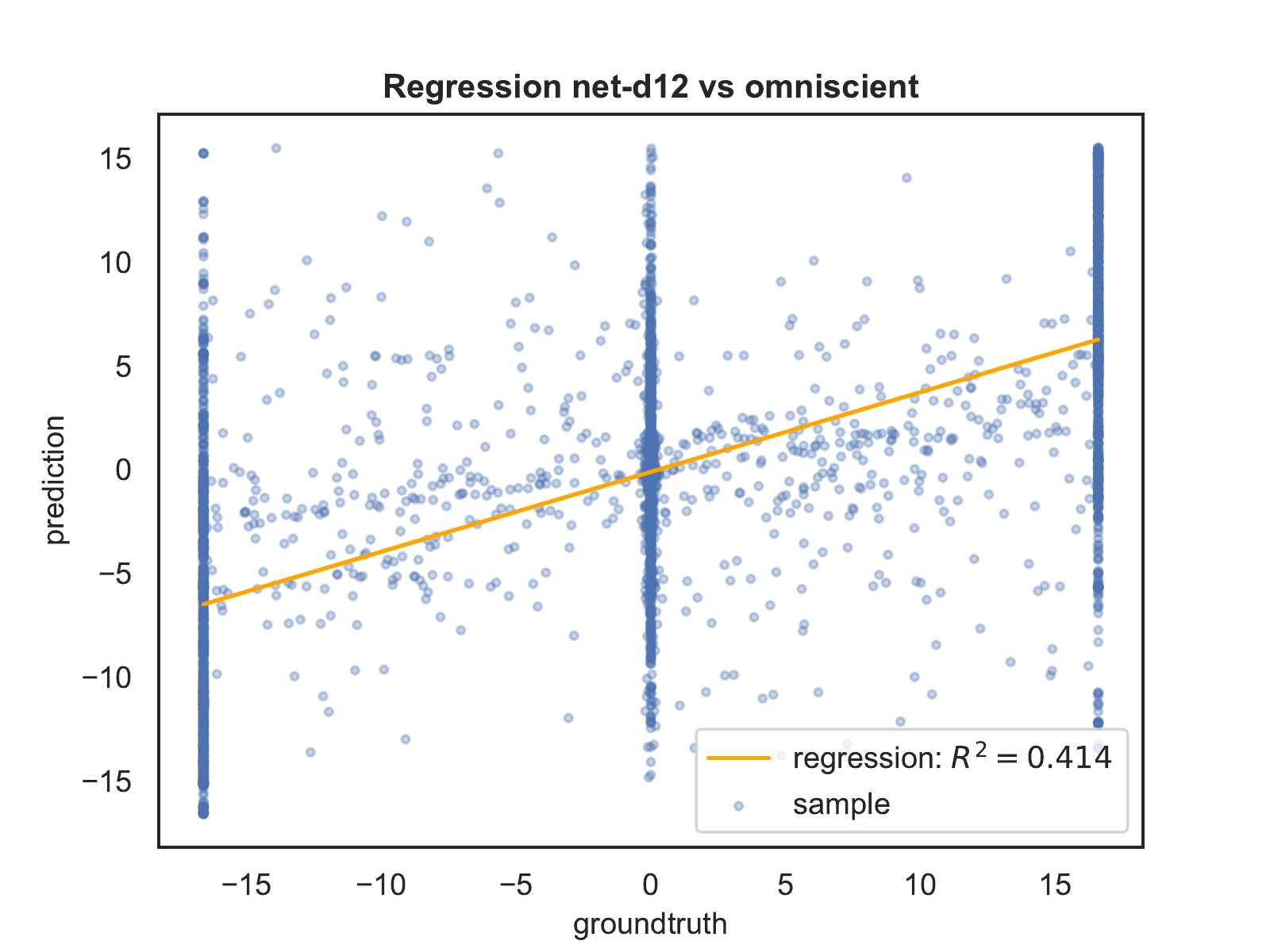}
	\end{subfigure}
	\caption[Evaluation of the \ac{r2} coefficients of \texttt{net-d12} 
	.]{Comparison 
		of the \ac{r2} coefficients of the manual and the controller learned from 
		\texttt{net-d12} with respect to the omniscient one.}
	\label{fig:net-d12r2}
\end{figure}

In Figure \ref{fig:net-d12traj1}, we first show a sample simulation: on the y-axis 
are visualised the positions of the agents, while on the x-axis the simulation time 
steps. 
\begin{figure}[!htb]
	\centering
	\includegraphics[width=.7\textwidth]{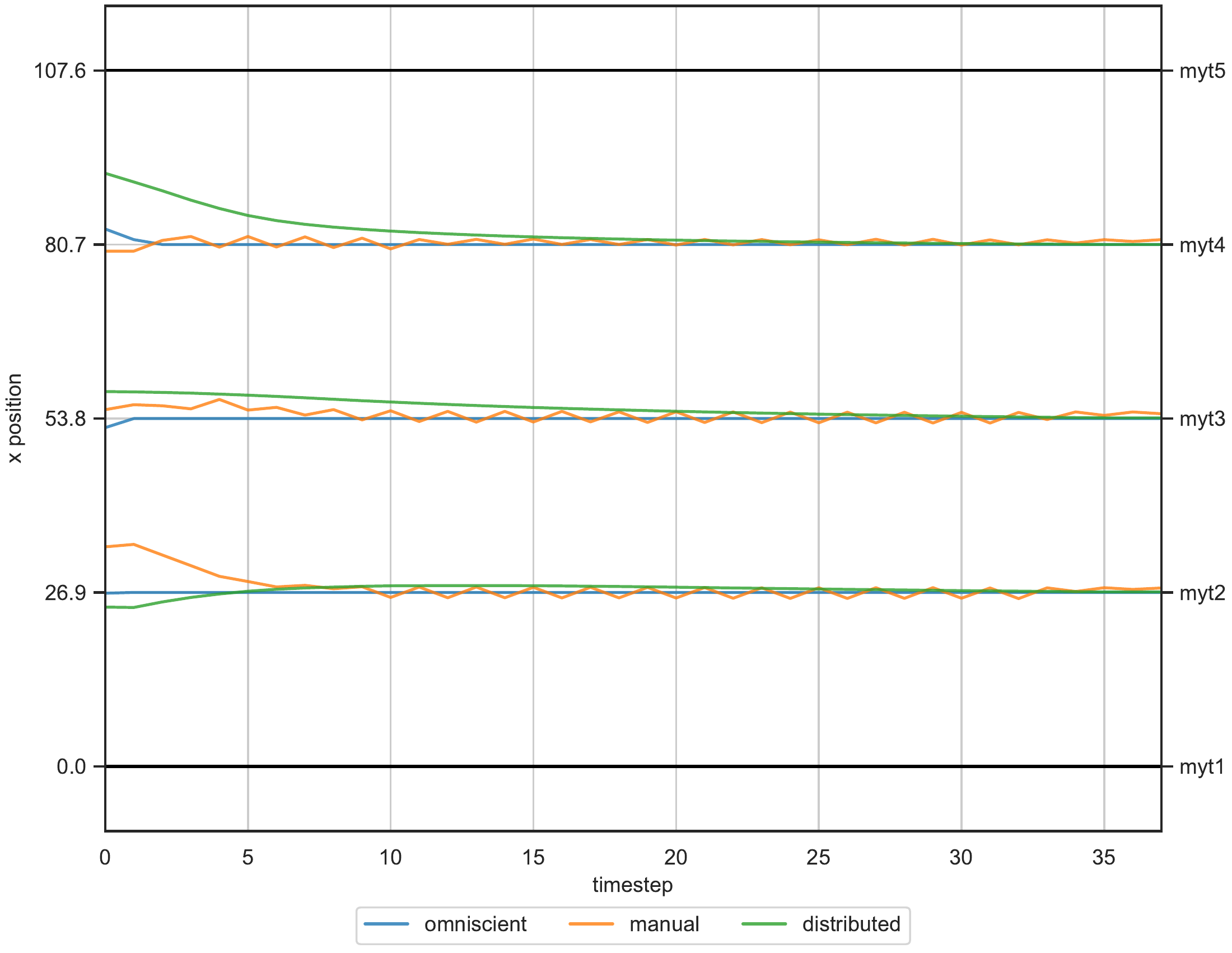}%
	\caption[Evaluation of the trajectories obtained with 5 agents.]{Comparison of 
	trajectories, of a single simulation, generated using three 
		controllers: the expert, the manual and the one learned from 
		\texttt{net-d12}.}
	\label{fig:net-d12traj1}
\end{figure}
The agents moved using the omniscient controller are, as expected, those that 
reach the target faster. Those moved using the manual controller are slower and 
moreover, as for the case shown in Figure \ref{fig:net-d9traj1}, they start to 
oscillate when they approach the target. Instead, the learned controller, even if is 
slower than the other two, is able to reach the correct configuration. 

In Figure \ref{fig:net-d12traj}, we show first a comparison of the expert and the 
learned trajectories, and then between the manual and the learned ones. 
In particular, on the y-axis is visualised the position of each agent over time, 
averaged over all the simulation runs, while on the x-axis the simulation 
time steps. It is important to note that there is a difference in these graphs 
compared to those of the previous group of experiments. 
\begin{figure}[!htb]
	\begin{center}
		\begin{subfigure}[h]{0.49\textwidth}
			\centering
			\includegraphics[width=.95\textwidth]{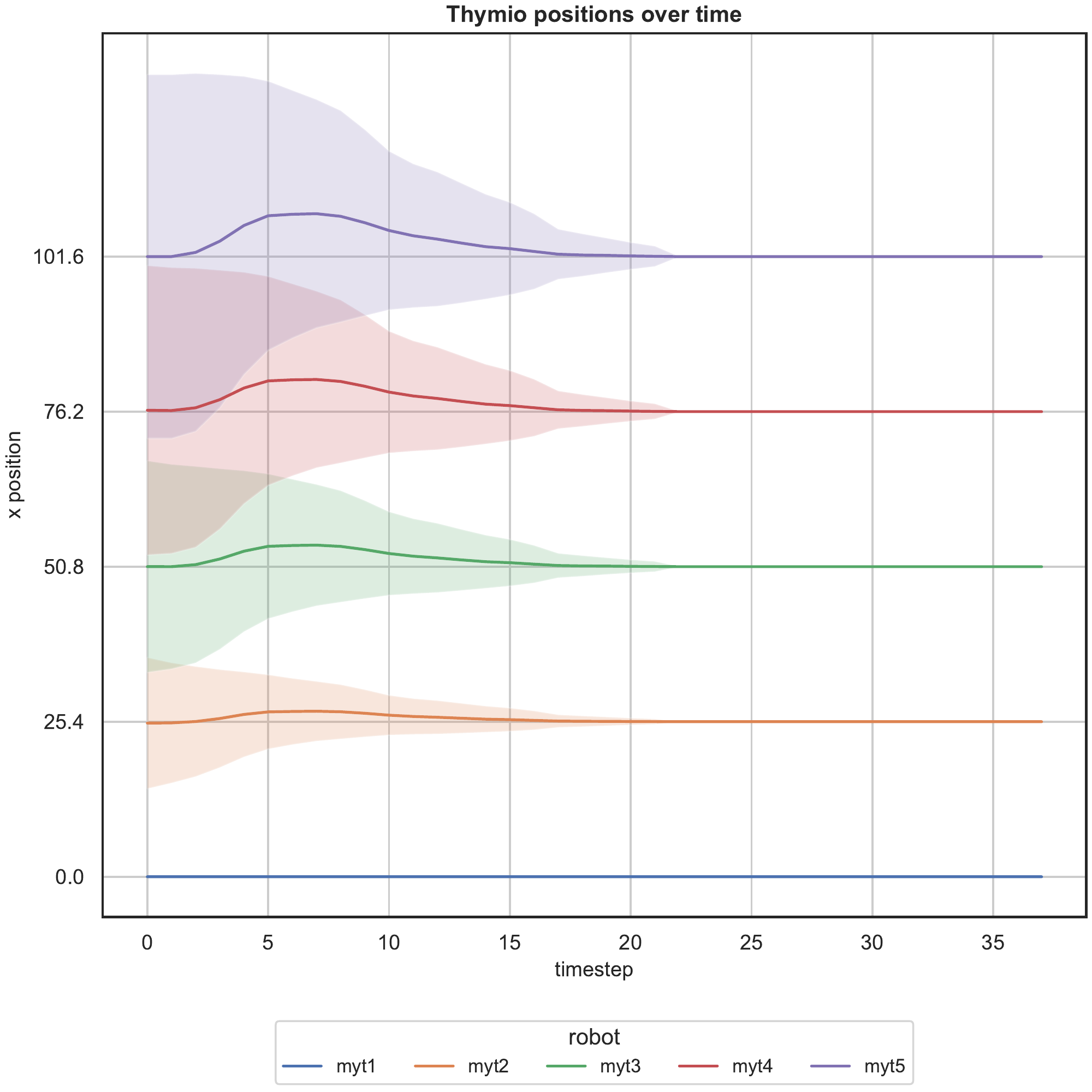}%
			\caption{Expert controller trajectories.}
		\end{subfigure}
		\hfill
		\begin{subfigure}[h]{0.49\textwidth}
			\centering
			\includegraphics[width=\textwidth]{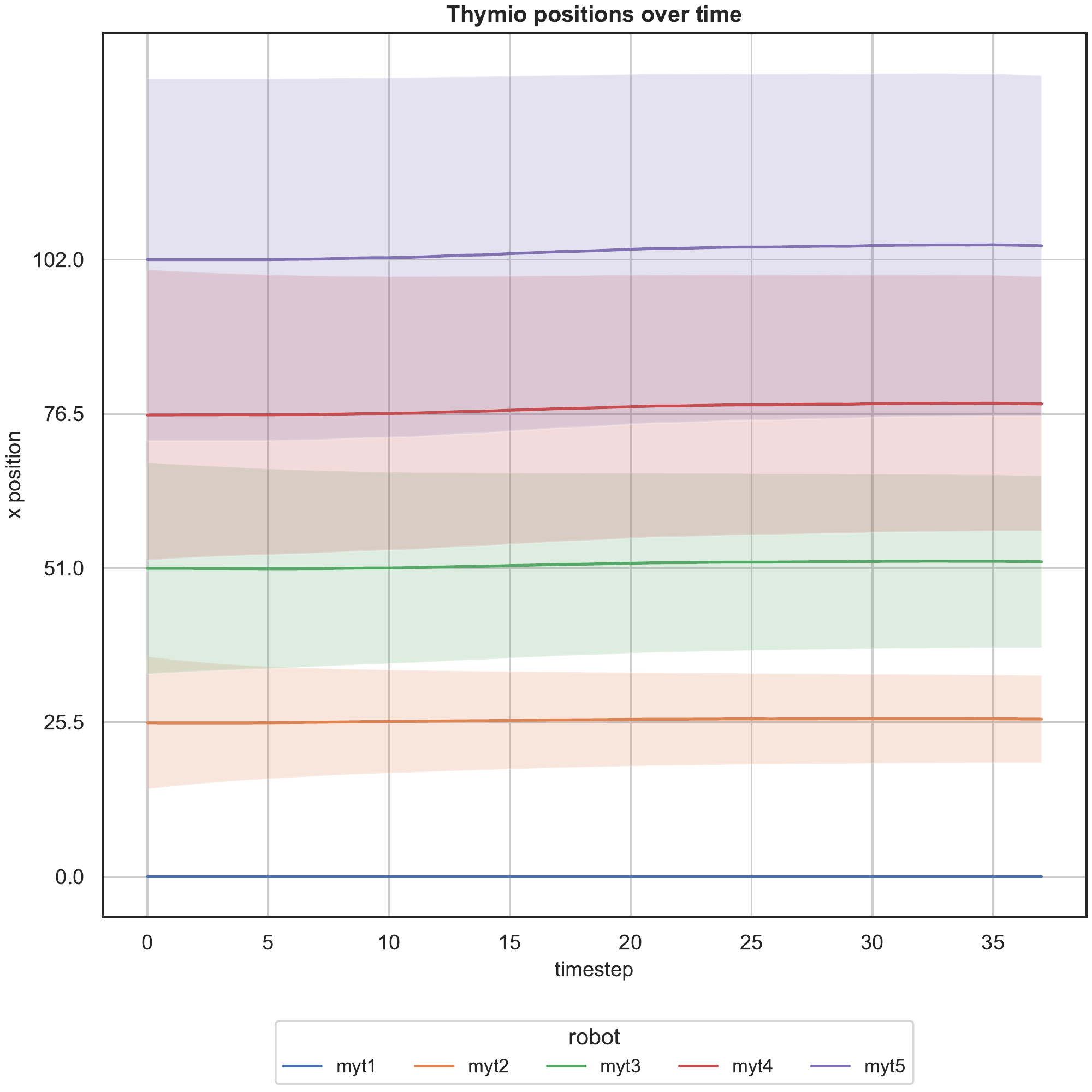}
			\caption{Distributed controller trajectories.}
		\end{subfigure}
	\end{center}
	\begin{center}
		\begin{subfigure}[h]{0.49\textwidth}
			\centering			
			\includegraphics[width=.95\textwidth]{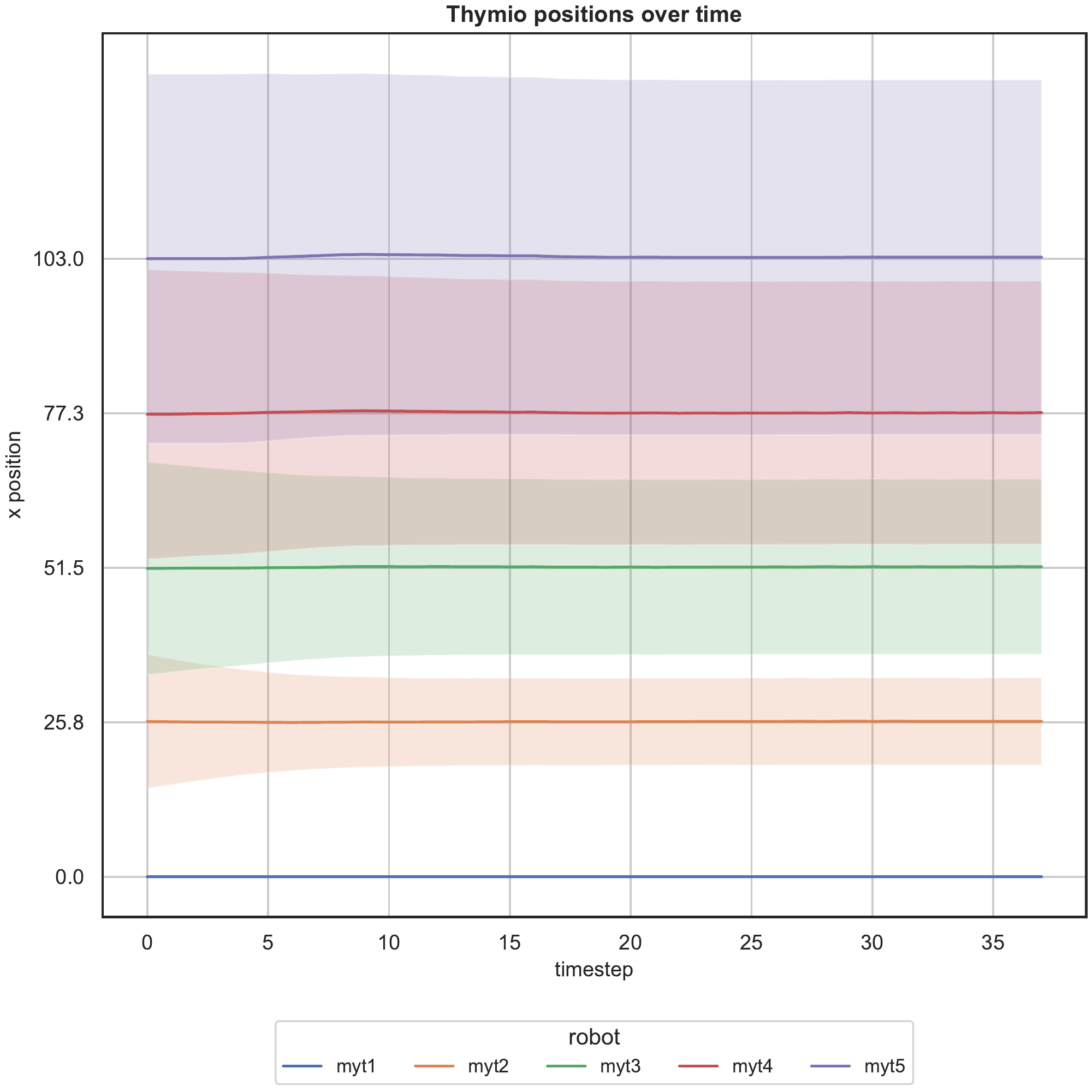}%
			\caption{Manual controller trajectories.}
		\end{subfigure}
		\hfill
		\begin{subfigure}[h]{0.49\textwidth}
			\centering
			\includegraphics[width=\textwidth]{contents/images/net-d12/position-overtime-learned_distributed}
			\caption{Distributed controller trajectories.}
		\end{subfigure}
	\end{center}
	\vspace{-0.5cm}
	\caption[Evaluation of the trajectories learned by 
	\texttt{net-d12}.]{Comparison of trajectories, of all the simulation runs, 
	generated using three controllers: the expert, the manual and the one learned 
	from \texttt{net-d12}.}
	\label{fig:net-d12traj}
\end{figure}

\noindent
Observing the deviation of the position of the robots with respect to the average, 
the last agent of the row did not maintain the same initial and goal positions 
throughout the simulations since the average gap set is different for every run.
The convergence of the robots to the target is guaranteed in 20 time steps using 
the expert, while the manual and learned controller still manage to reach the 
correct configuration with more time.

Analysing the evolution of the control over time, in Figure 
\ref{fig:net-d12control}, we observe that the speeds set by the manual controller 
and the one learned from the network are significantly lower and therefore do 
not allow to reach the target in a satisfactory time.
\begin{figure}[!htb]
	\centering
	\begin{subfigure}[h]{0.3\textwidth}
		\centering
		\includegraphics[width=\textwidth]{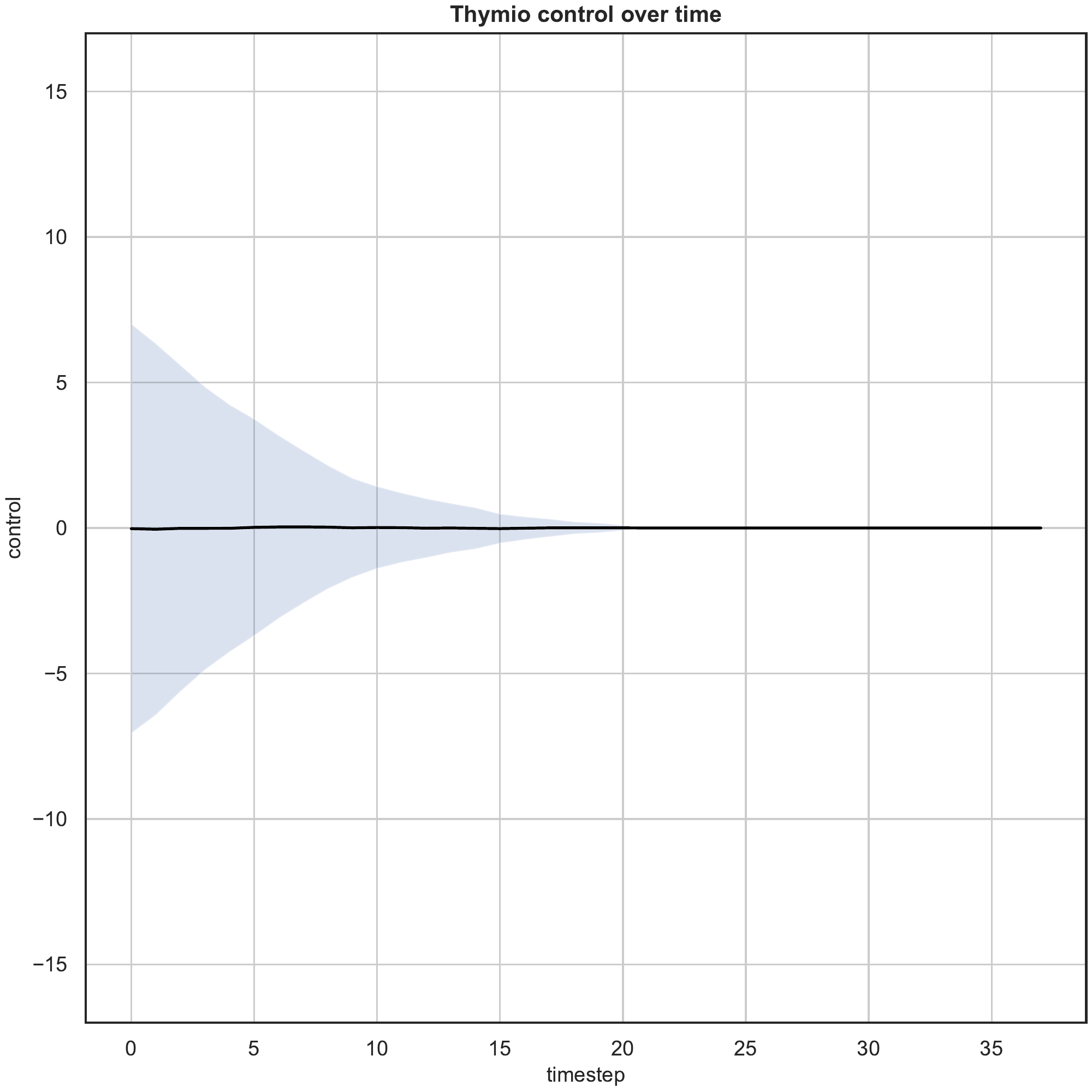}%
		\caption{Expert controller.}
	\end{subfigure}
	\hfill
	\begin{subfigure}[h]{0.3\textwidth}
		\centering
		\includegraphics[width=\textwidth]{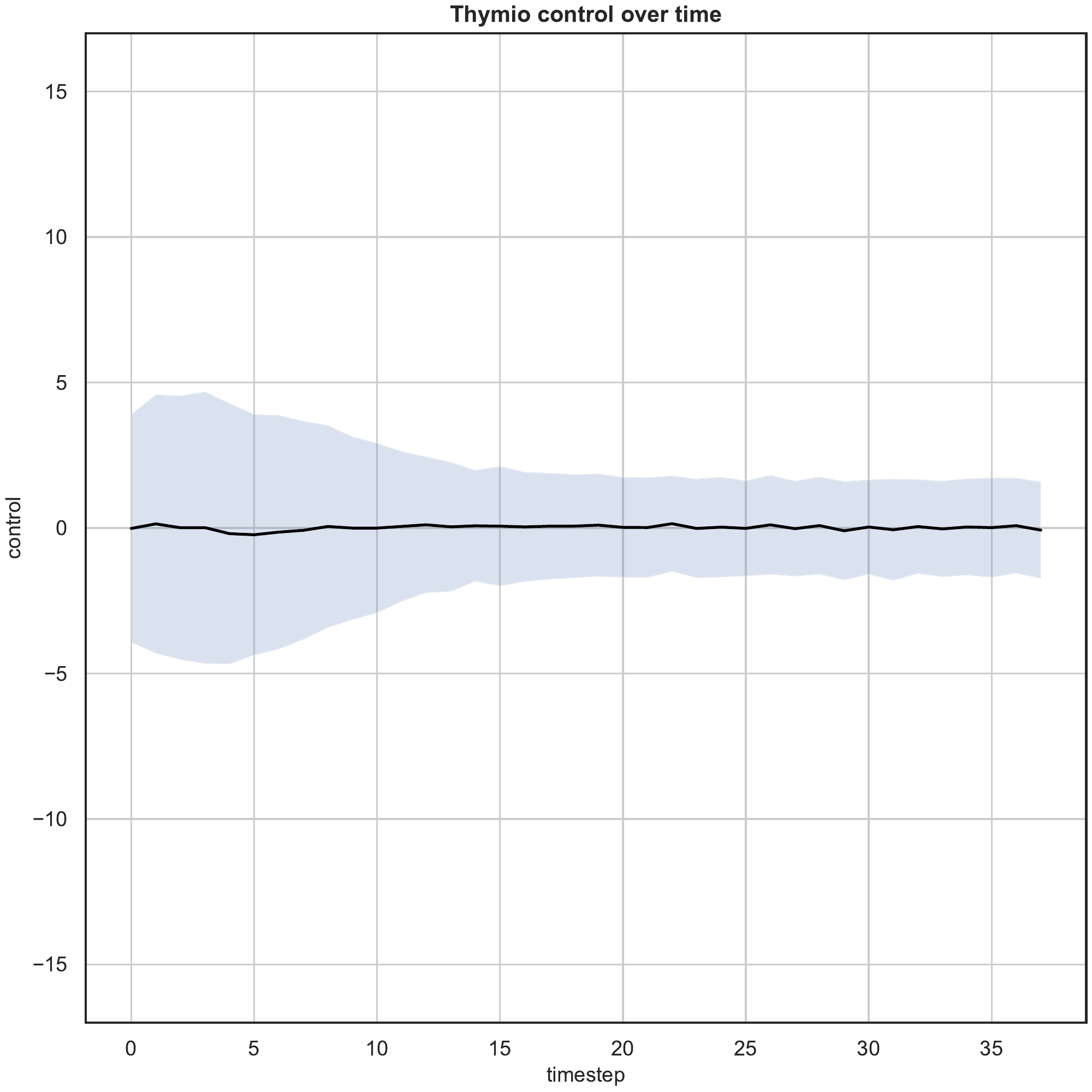}%
		\caption{Manual controller.}
	\end{subfigure}
	\hfill
	\begin{subfigure}[h]{0.3\textwidth}
		\centering
		\includegraphics[width=\textwidth]{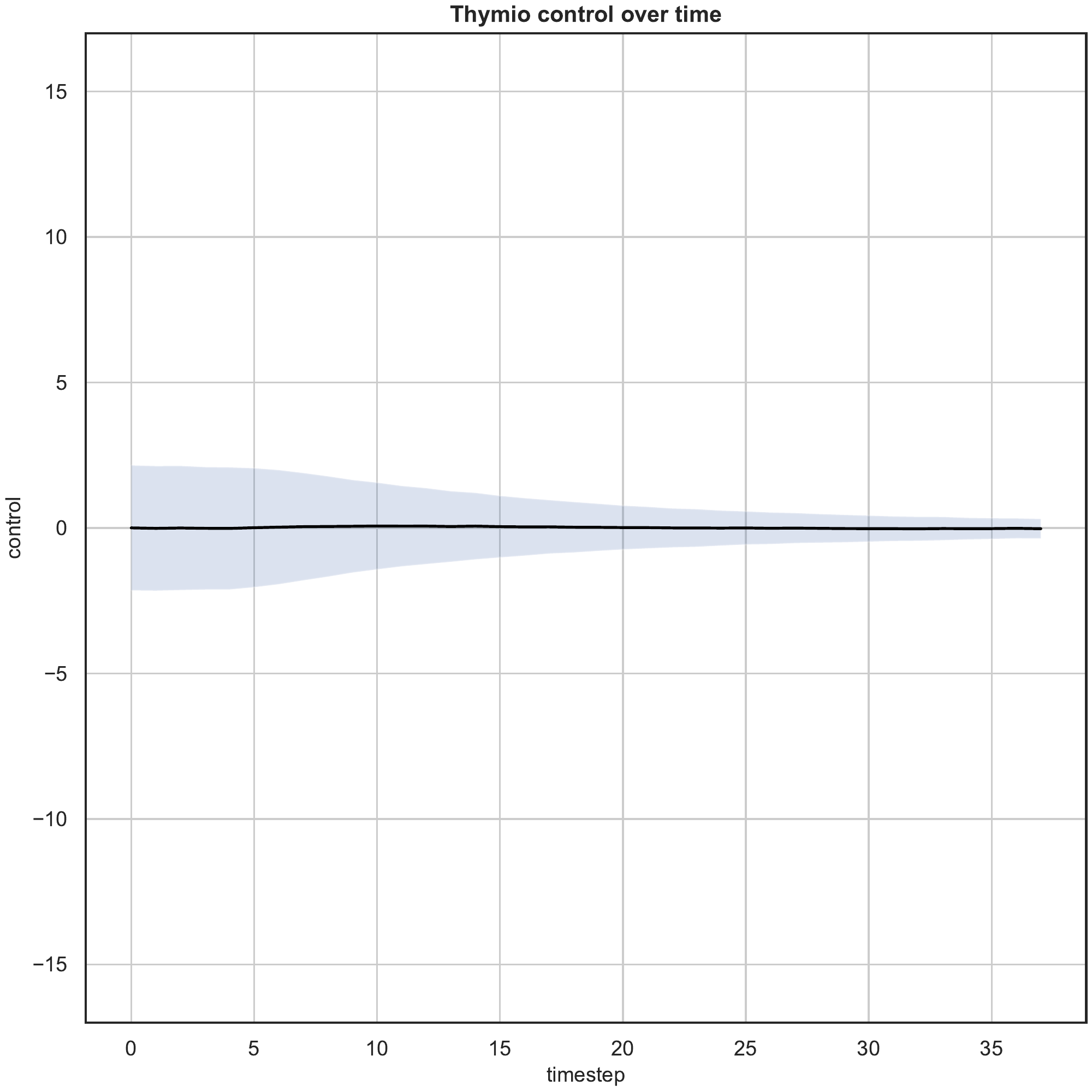}
		\caption{Distributed controller.}
	\end{subfigure}
	\caption[Evaluation of the control decided by \texttt{net-d12}.]{Comparison of 
	output control decided using three controllers: the expert, the manual and the 
	one learned from \texttt{net-d12}.}
	\label{fig:net-d12control}
\end{figure}

In Figure \ref{fig:net-d12responseposition}, another informative plot displays 
the behaviour of a robot located between other two that are already in their place.
As expected, the trend of the three curves shows how the behaviour of the 
model learned and of the manual controller are similar.
\begin{figure}[!htb]
	\centering
	\includegraphics[width=.45\textwidth]{contents/images/net-d12/response-varying_init_position-distributed}%
	\caption{Response of \texttt{net-d12} by varying the initial position.}
	\label{fig:net-d12responseposition}
\end{figure}
However, the performance of the manual controller, when the robot is close to 
the one that preceded it, is practically comparable to that obtained by the 
expert.
In contrast, when the agent is close to the following one, it worsens and 
presents a lot of variability.

Finally, in Figure \ref{fig:net-d12distance}, is shown the absolute distance of 
each robot from the target, averaged on all robots among all the simulation runs, 
over time. 
Despite we expected performances similar to those presented in the Figure 
\ref{fig:net-d9distance}, in this circumstance the agents moved following the 
distributed controller are closer to the target, in particular in the final 
configuration they are on average $2$cm away from the goal position. 
Furthermore, it is shown that there are far fewer cases away from the average 
behaviour.
\begin{figure}[!htb]
	\centering
	\includegraphics[width=.65\textwidth]{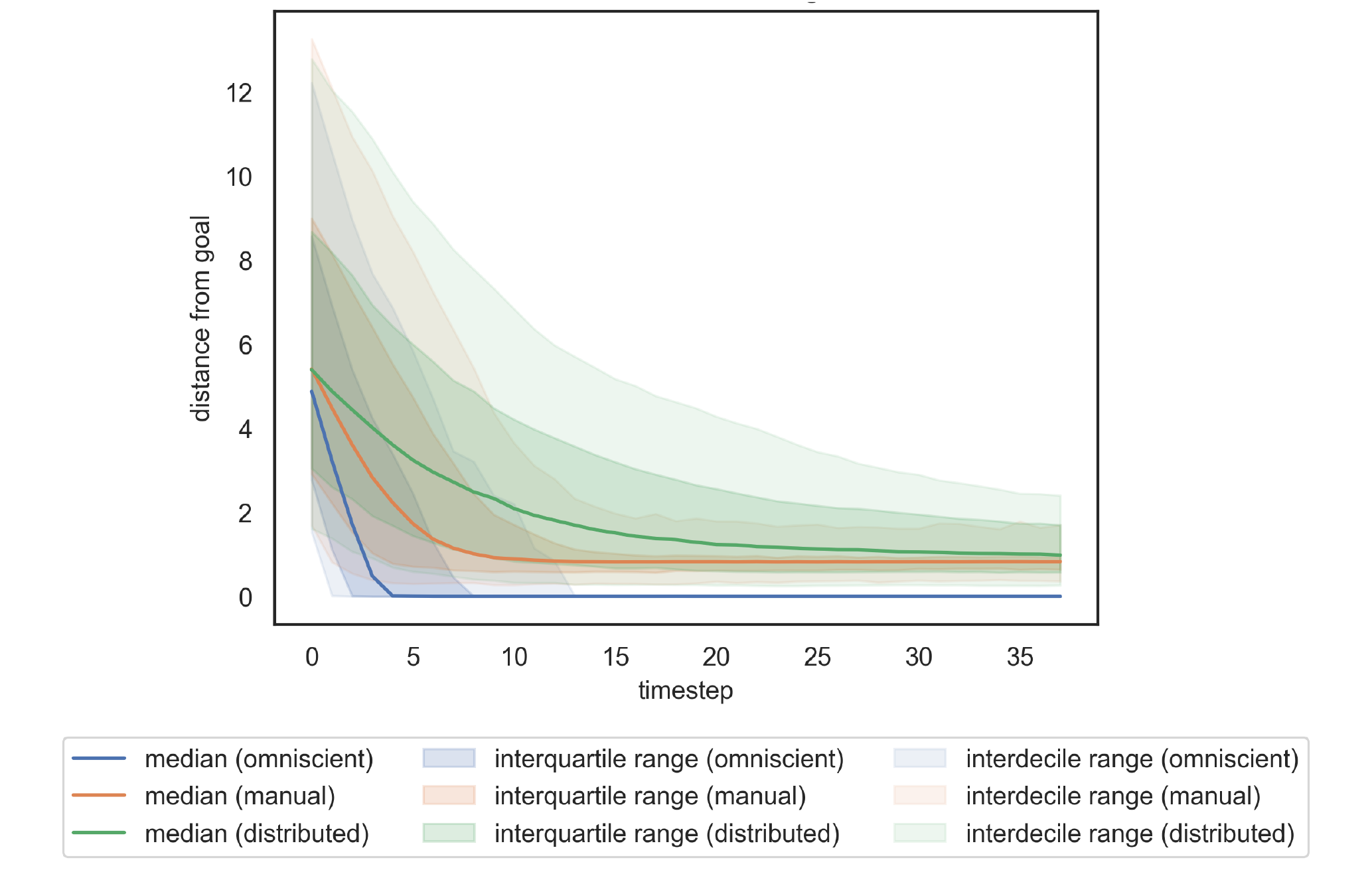}%
	\caption[Evaluation of \texttt{net-d1} distances from goal.]{Comparison of 
		performance in terms of distances from goal obtained using three 
		controllers: 
		the expert, the manual and the one learned from \texttt{net-d12}.}
	\label{fig:net-d12distance}
\end{figure}

\paragraph*{Results using 8 agents}
Following are shown the losses of the models trained using an higher number of 
agents, i.e.,8, by varying the average gap. From a first observation, 
\begin{figure}[H]
	\centering
	\includegraphics[width=.75\textwidth]{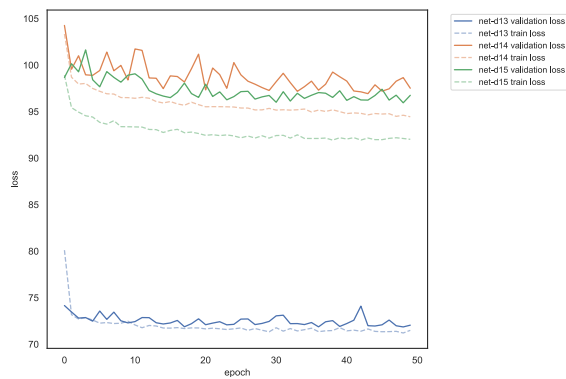}%
	\caption[Comparison of the losses of the models that use $8$ 
	agents.]{Comparison of the losses of the models that use $8$ agents as the gap 
	varies.}
	\label{fig:distlossn8}
\end{figure}

\noindent
the network seems to be able to work with all the gaps.
As before, for the network it is easier to perform a task using a smaller gap.
For this reason, it is more interesting to analyse the case in which the model 
is trained using a variable gap.

In Figure \ref{fig:net-d15r2} is visualised a comparison of the \ac{r2} 
coefficients of the manual and the learned controller. In both cases, the 
coefficients are very low, since in most of the cases in which the controllers have 
to decide a zero or maximum speed a wrong value is predicted, however, the 
coefficient obtained from the network is slightly better.
\begin{figure}[!htb]
	\centering
	\begin{subfigure}[h]{0.49\textwidth}
		\centering
		\includegraphics[width=\textwidth]{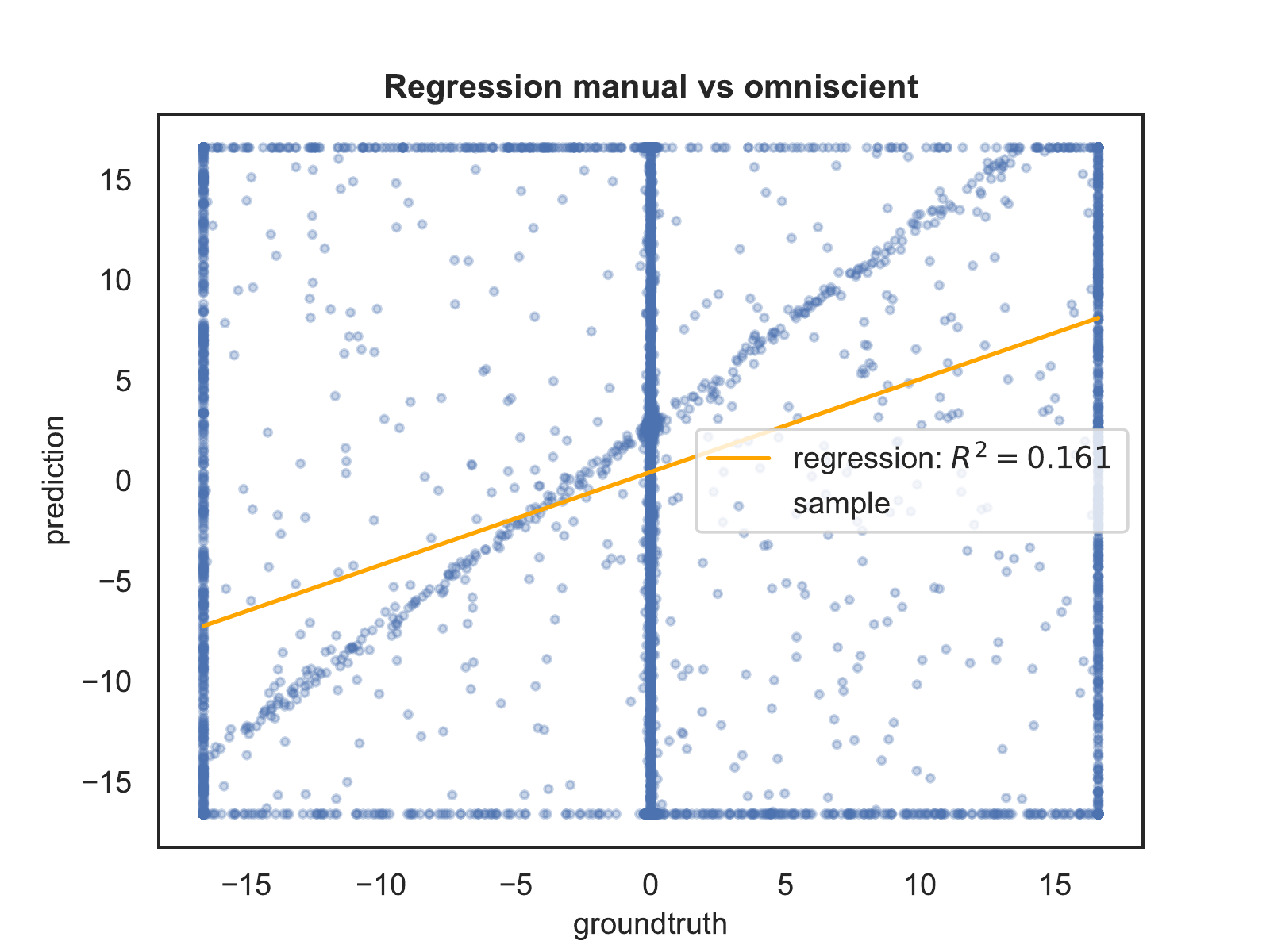}%
	\end{subfigure}
	\hfill
	\begin{subfigure}[h]{0.49\textwidth}
		\centering
		\includegraphics[width=\textwidth]{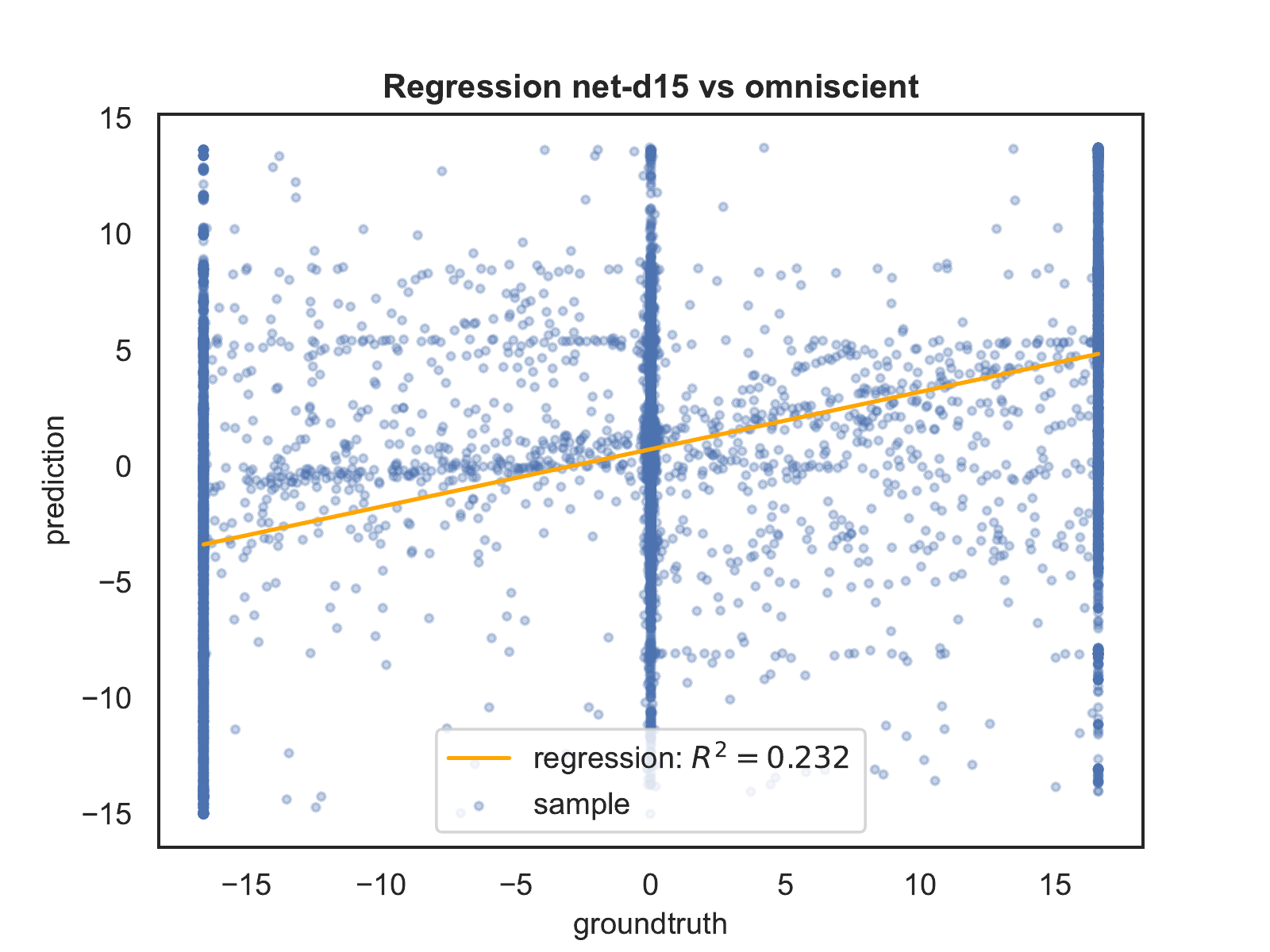}
	\end{subfigure}
	\caption[Evaluation of the \ac{r2} coefficients of \texttt{net-d15} 
	.]{Comparison of the \ac{r2} coefficients of the manual and the controller 
		learned from \texttt{net-d15} with respect to the omniscient one.}
	\label{fig:net-d15r2}
\end{figure}

In Figure \ref{fig:net-d15traj1} is displayed a sample simulation that shows on 
the y-axis position of  
\begin{figure}[H]
	\centering
	\includegraphics[width=.7\textwidth]{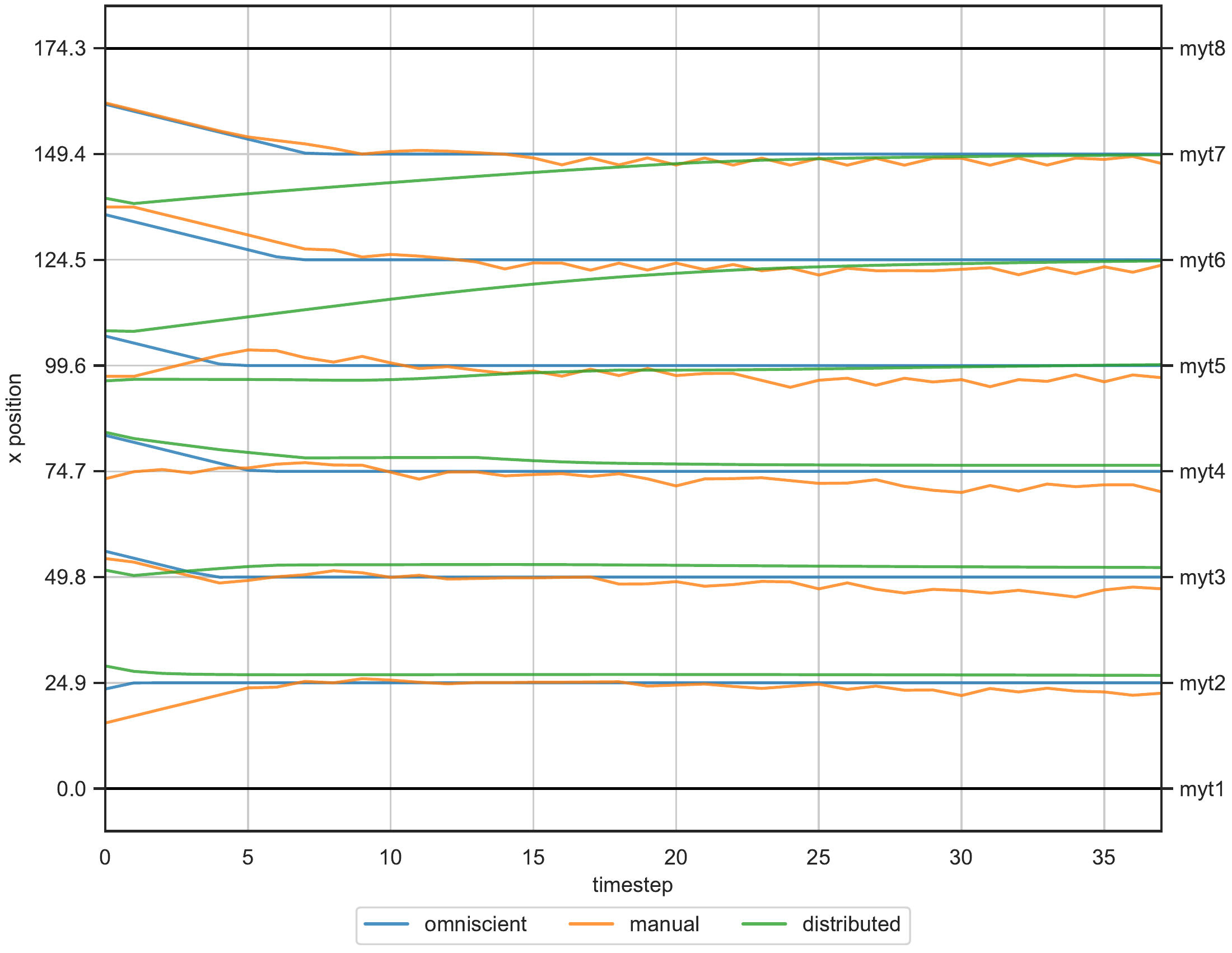}%
	\caption[Evaluation of the trajectories obtained with 8 agents.]{Comparison of 
	trajectories, of a single simulation, generated using three controllers: the 
	expert, the manual and the one learned from \texttt{net-d15}.}
	\label{fig:net-d15traj1}
\end{figure}

\noindent
each agent, while on the x-axis the simulation time steps.
The agents moved using the learned controller, even if in the initial configuration 
are far from the target, they are able to reach the goal. 
Instead, the agents moved using the manual controller, when have almost 
approached the target, they start to oscillate. 

In Figure \ref{fig:net-d15traj} are shown the trajectories obtained employing 
the three controllers, averaged over all the simulation runs.
Compared to the previous case, a greater number of robots implies a 
slowdown in reaching the correct position, even when using an expert 
controller.
As before, the convergence of the robots using the manual and learned 
controllers needs more time.
\begin{figure}[!htb]
	\begin{center}
		\begin{subfigure}[h]{0.49\textwidth}
			\centering
			\includegraphics[width=.9\textwidth]{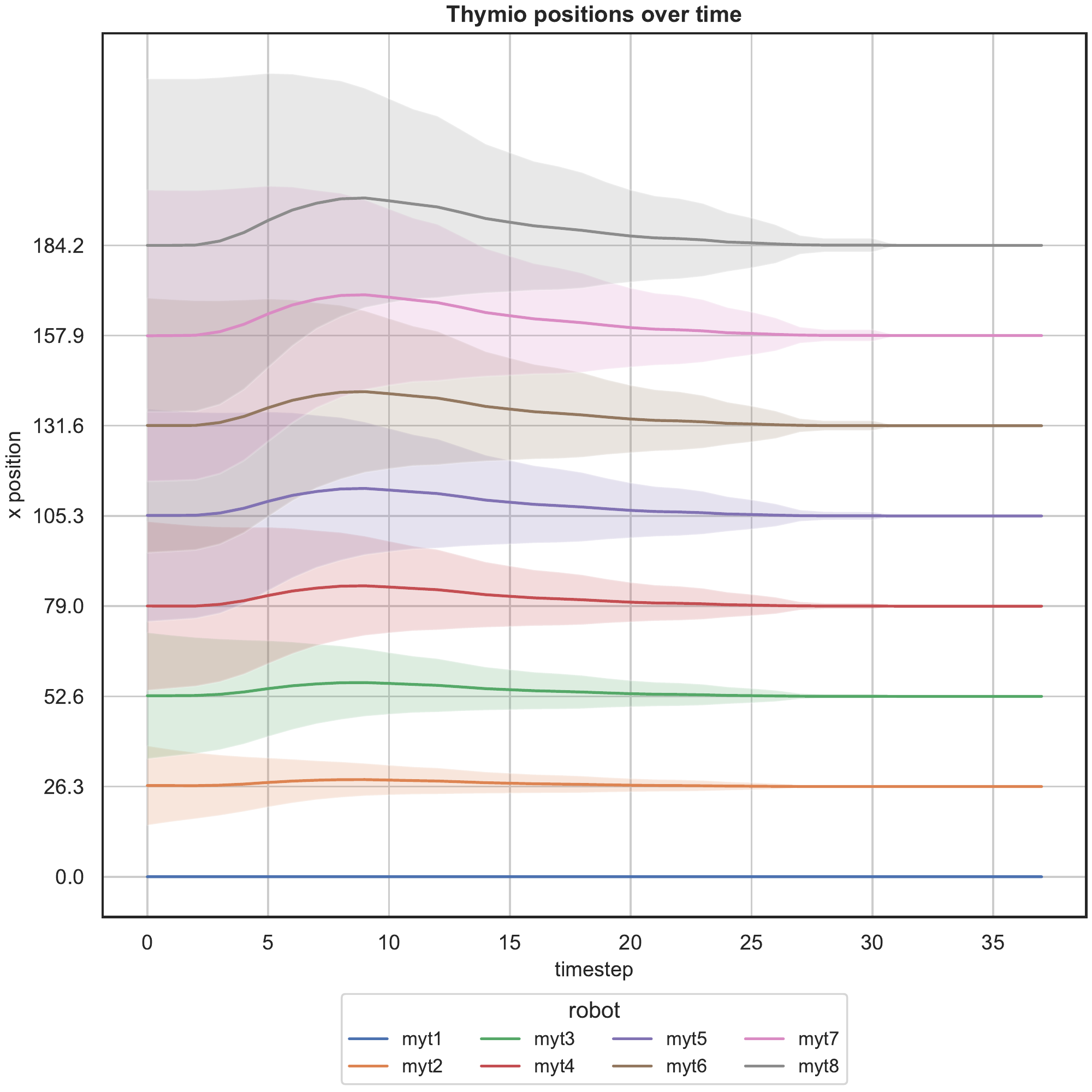}%
			\caption{Expert controller trajectories.}
		\end{subfigure}
		\hfill
		\begin{subfigure}[h]{0.49\textwidth}
			\centering
			\includegraphics[width=.9\textwidth]{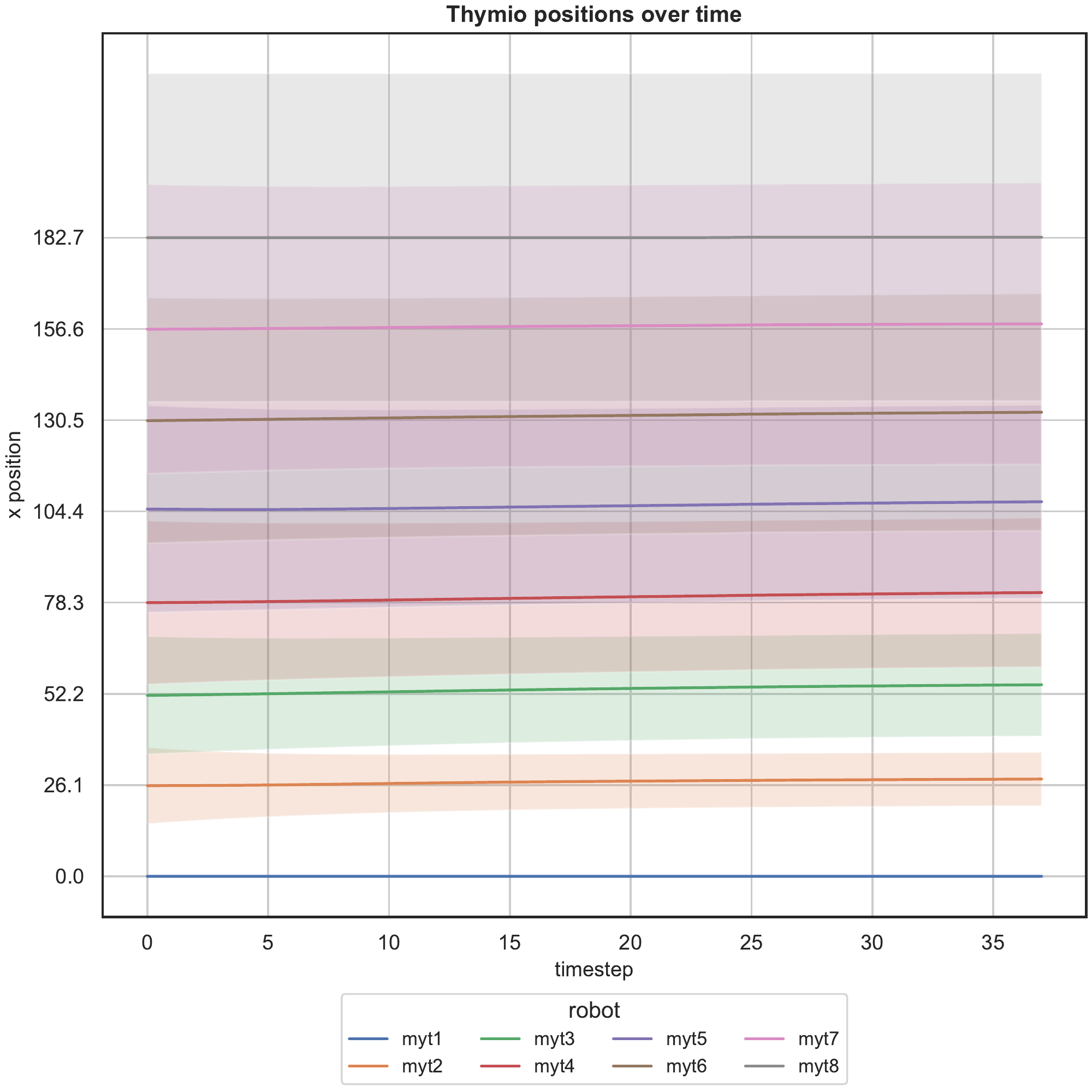}
			\caption{Distributed controller trajectories.}
		\end{subfigure}
	\end{center}
	\begin{center}
		\begin{subfigure}[h]{0.49\textwidth}
			\centering			
			\includegraphics[width=.9\textwidth]{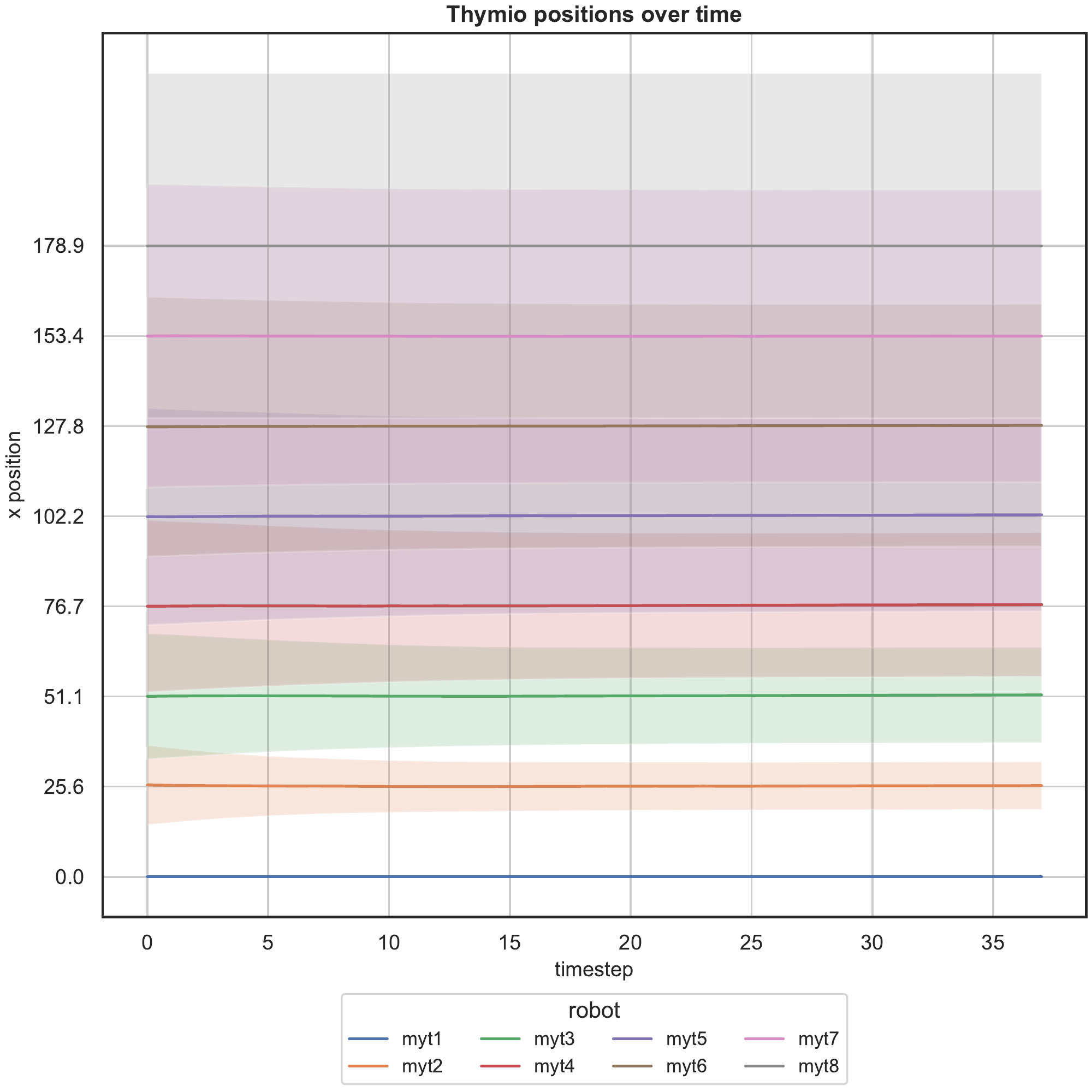}%
			\caption{Manual controller trajectories.}
		\end{subfigure}
		\hfill
		\begin{subfigure}[h]{0.49\textwidth}
			\centering
			\includegraphics[width=.9\textwidth]{contents/images/net-d15/position-overtime-learned_distributed}
			\caption{Distributed controller trajectories.}
		\end{subfigure}
	\end{center}
	\caption[Evaluation of the trajectories learned by 
	\texttt{net-d15}.]{Comparison of trajectories, of all the simulation runs, 
	generated using three controllers: the expert, the manual and the one learned 
	from \texttt{net-d15}.}
	\label{fig:net-d15traj}
\end{figure}

Examining in Figure \ref{fig:net-d15control} the evolution of the control over 
time, the graph of the distributed controller highlights how the speed decided by 
the model has further decreased due to the increase in the amount of agents in 
the simulation.
\begin{figure}[!htb]
	\centering
	\begin{subfigure}[h]{0.3\textwidth}
		\centering
		\includegraphics[width=\textwidth]{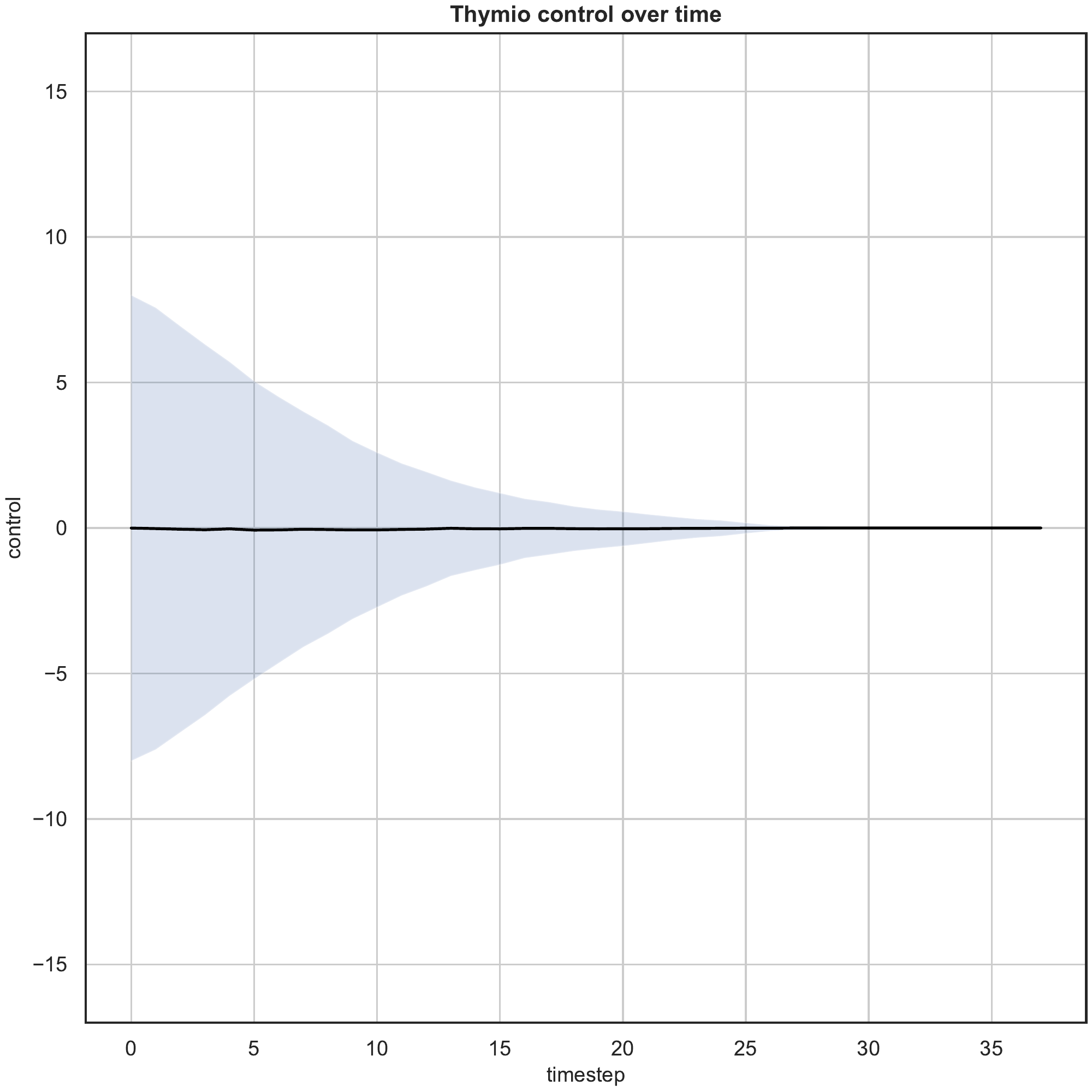}%
		\caption{Expert controller.}
	\end{subfigure}
	\hfill
	\begin{subfigure}[h]{0.3\textwidth}
		\centering
		\includegraphics[width=\textwidth]{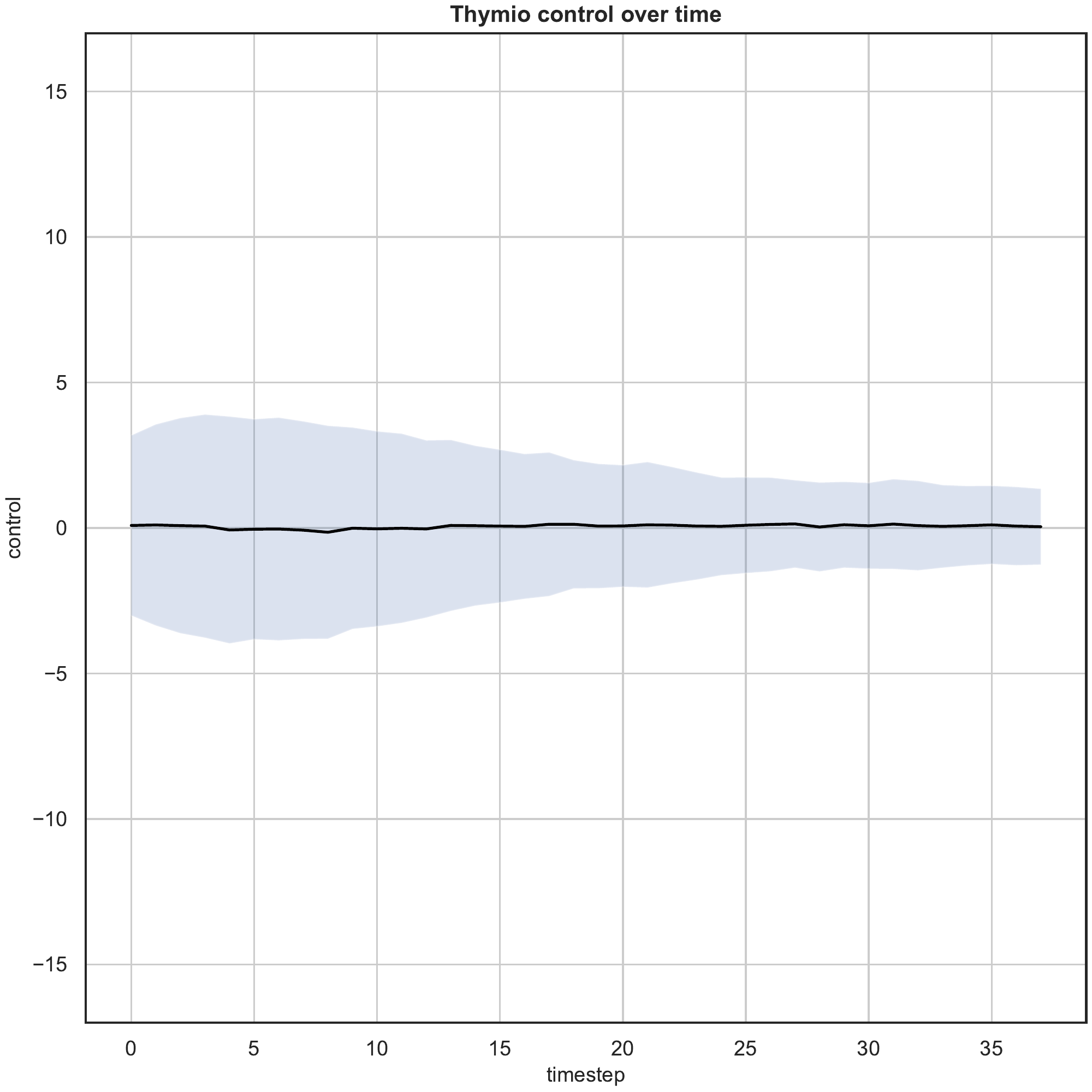}%
		\caption{Manual controller.}
	\end{subfigure}
	\hfill
	\begin{subfigure}[h]{0.3\textwidth}
		\centering
		\includegraphics[width=\textwidth]{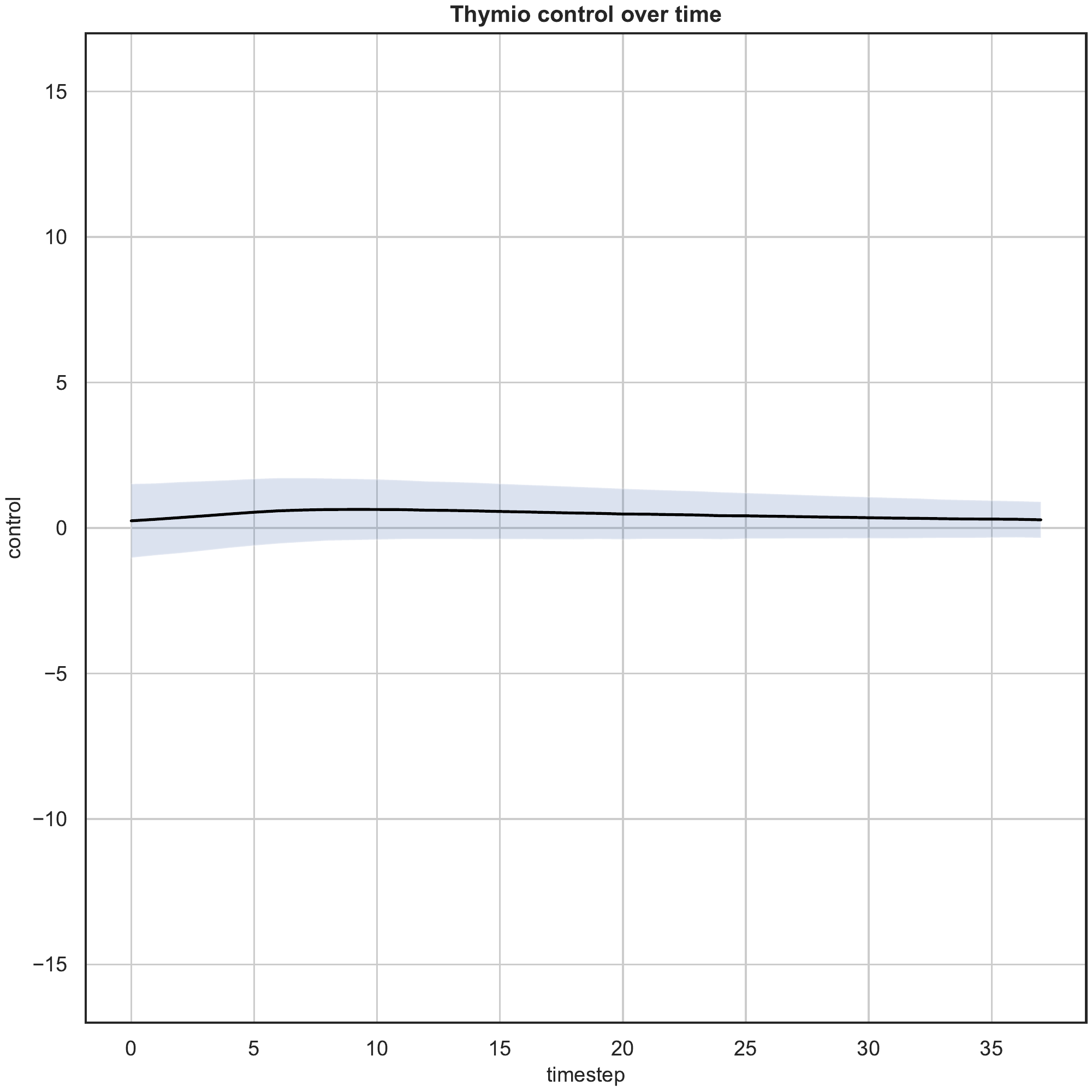}
		\caption{Distributed controller.}
	\end{subfigure}
	\caption[Evaluation of the control decided by \texttt{net-d15}.]{Comparison 
		of output control decided using three controllers: the expert, the manual 
		and the one learned from \texttt{net-d15}.}
	\label{fig:net-d15control}
\end{figure}

Figure \ref{fig:net-d15responseposition} displays the behaviour of a robot 
located between other two that are already in their place.
Analysing the way of acting of the three controllers for this experiment, from 
the plot arises an important difference in the decisions taken by the distributed 
and the manual controllers.
The learned controller, whether a robot is closer to the one that precedes it 
or to the one following it, sets a proportional speed, lower than the optimal one, 
that leads it to move respectively back and forth to reach the desired position.
Instead, the manual controller when an agent is closer to the one in front 
sets a very high speed to move quickly to the desired position, just like the expert 
does, unlike when the robot is closer to the one following it, where it sets a 
negative speed but not high enough, a bit like the distributed controller does.
\begin{figure}[!htb]
	\centering
	\includegraphics[width=.45\textwidth]{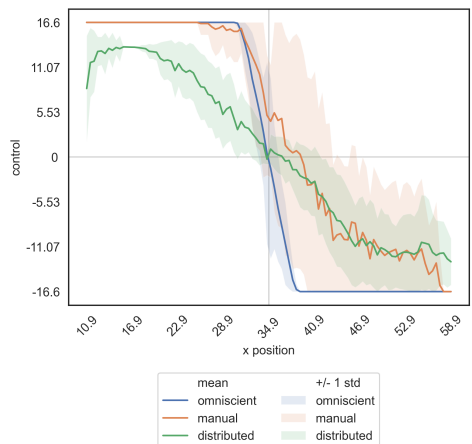}%
	\caption{Response of \texttt{net-d15} by varying the initial position.}
	\label{fig:net-d15responseposition}
\end{figure}

\bigskip
Finally, in Figure \ref{fig:net-d15distance} is presented the average distance of 
the robots from the target among all the simulations. The performance of the 
learned and the manual controllers are different from before: \texttt{net-d15} is 
slower to converge. In fact, this plot 
confirms that the agents moved following a manual controller in the final 
configuration are closer to the target than those moved by the distributed 
controller, respectively, they are on average $2$ or $3.5$cm away from the 
goal position. Moreover, observing the coloured bands we see that there is a lot of 
variance in the distributed controller final positions, in fact there are runs in which 
some agents can be even $10$cm far from the target.
\begin{figure}[!htb]
	\centering
	\includegraphics[width=.65\textwidth]{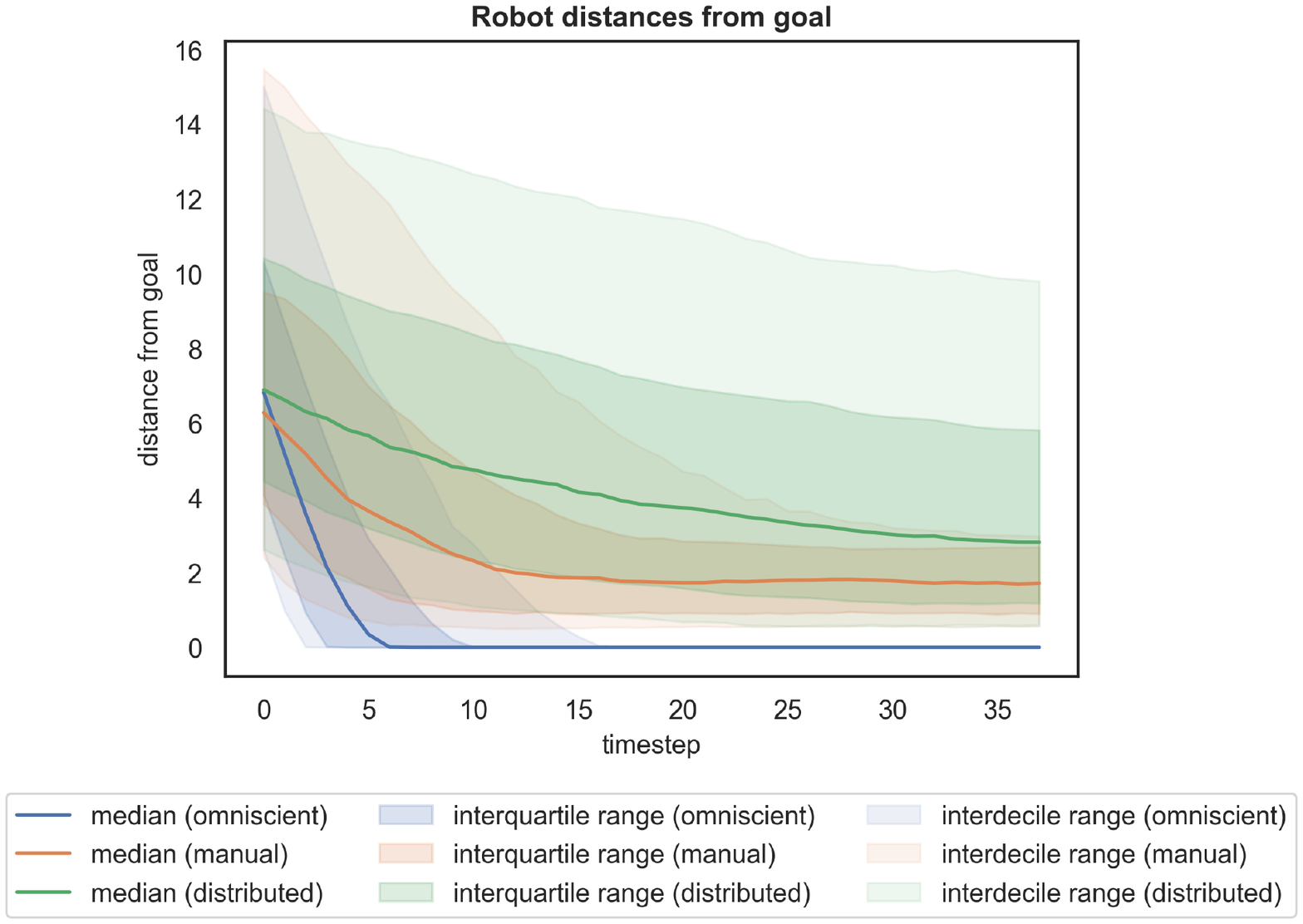}%
	\caption[Evaluation of \texttt{net-d15} distances from goal.]{Comparison 
	of performance in terms of distances from goal obtained using three 
	controllers: the expert, the manual and the one learned from \texttt{net-d15}.}
	\label{fig:net-d15distance}
\end{figure}

\paragraph*{Results using variable agents}
We conclude the experiments performed using a distributed approach by 
presenting the results obtained with a variable number of agents. 
\begin{figure}[H]
	\centering
	\includegraphics[width=.8\textwidth]{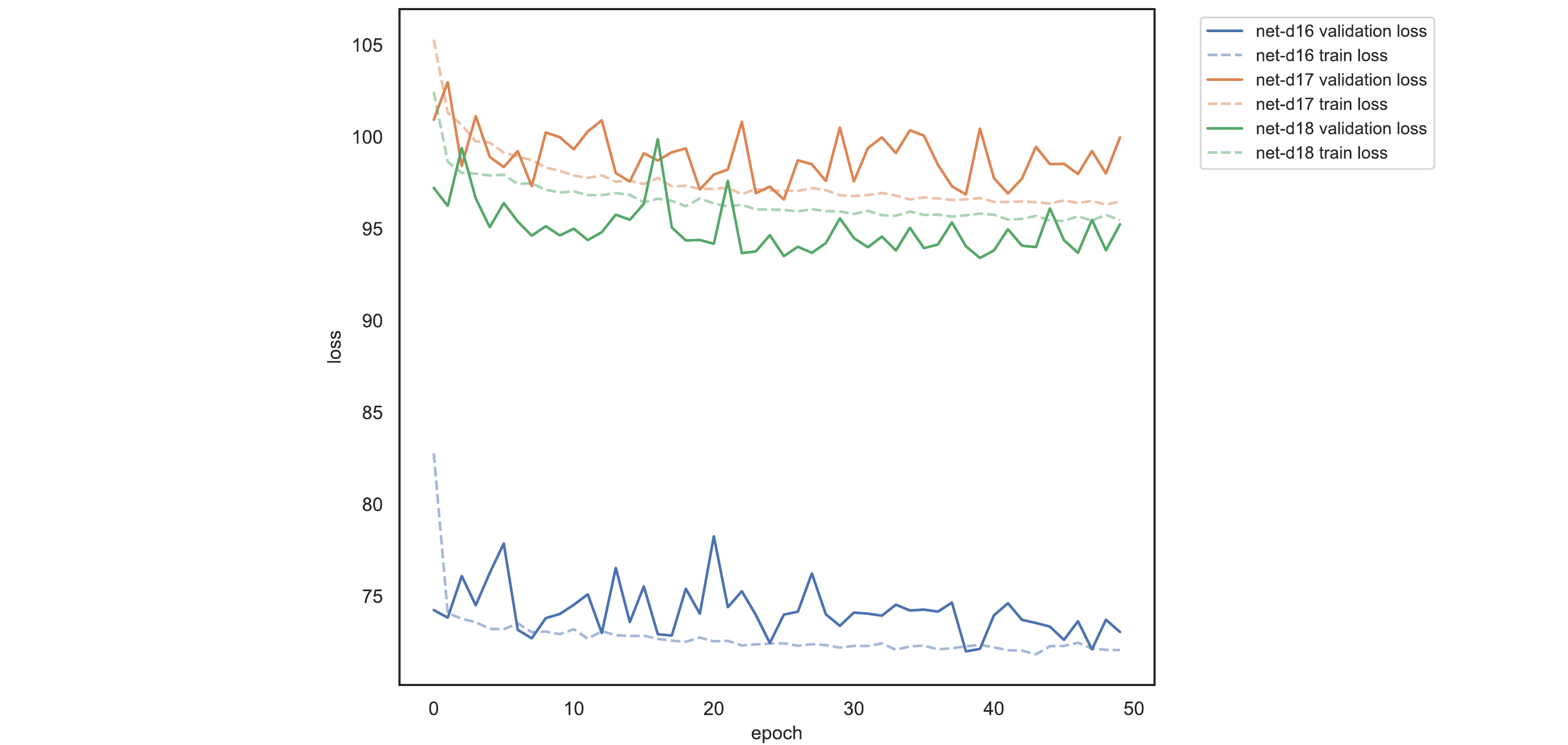}%
	\caption[Comparison of the losses of the models that use variable 
	agents.]{Comparison of the losses of the models that use variable agents 
	and gaps.}
	\label{fig:distlossnvar}
\end{figure}

\noindent
In Figure \ref{fig:distlossnvar}, are analysed the losses by varying the average 
gap. As before, for the network it is easier to perform a task using a smaller gap 
and in general training the model on a variable gap performs better than on a 
fixed but big gap.

Dwelling on the most interesting case, the one in with both average gap and 
number of agents variable, the \ac{r2} coefficients shown in Figure 
\ref{fig:net-d18r2} are still very low.
\begin{figure}[!htb]
	\centering
	\begin{subfigure}[h]{0.49\textwidth}
		\centering
		\includegraphics[width=\textwidth]{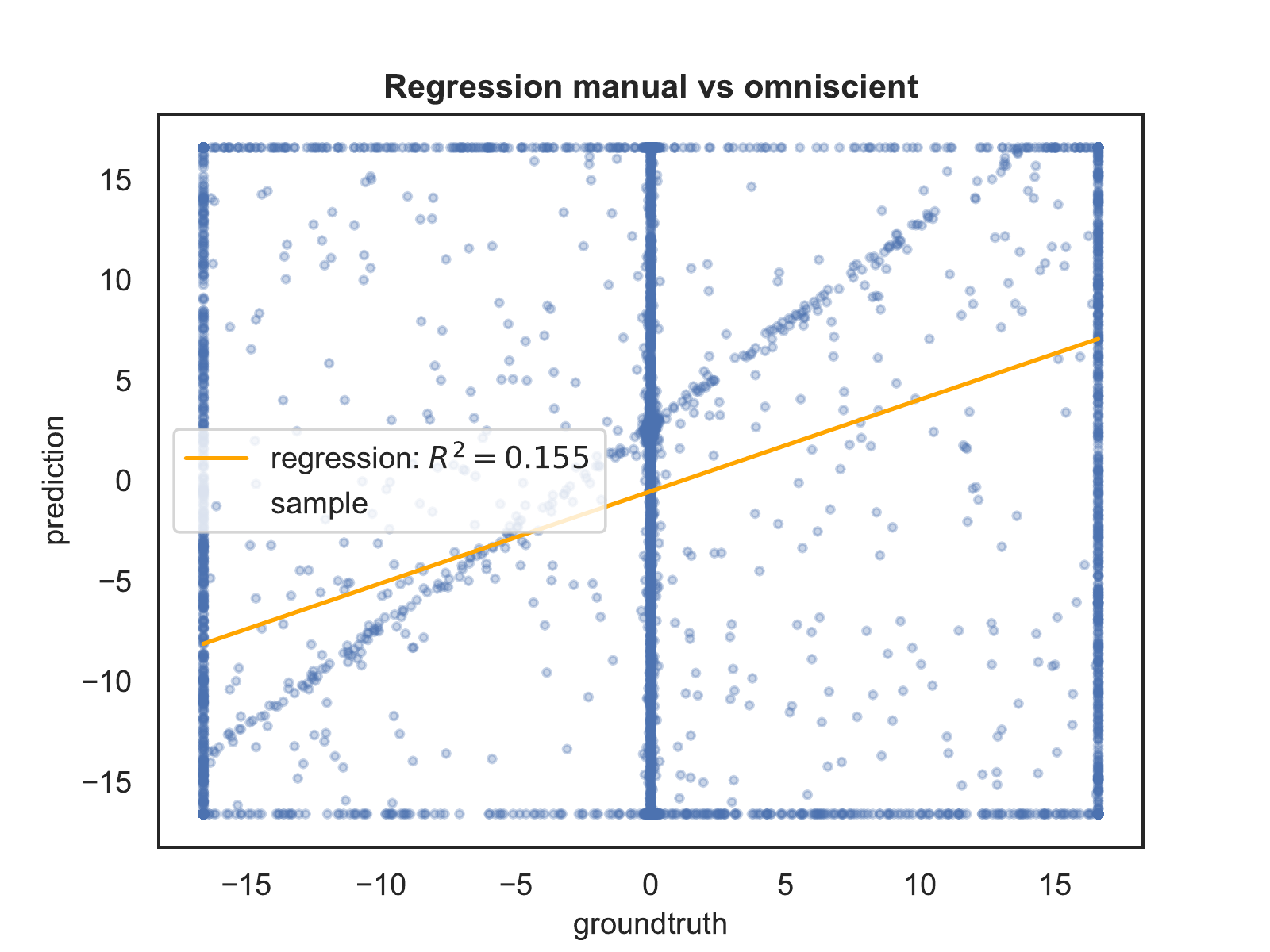}%
	\end{subfigure}
	\hfill
	\begin{subfigure}[h]{0.49\textwidth}
		\centering
		\includegraphics[width=\textwidth]{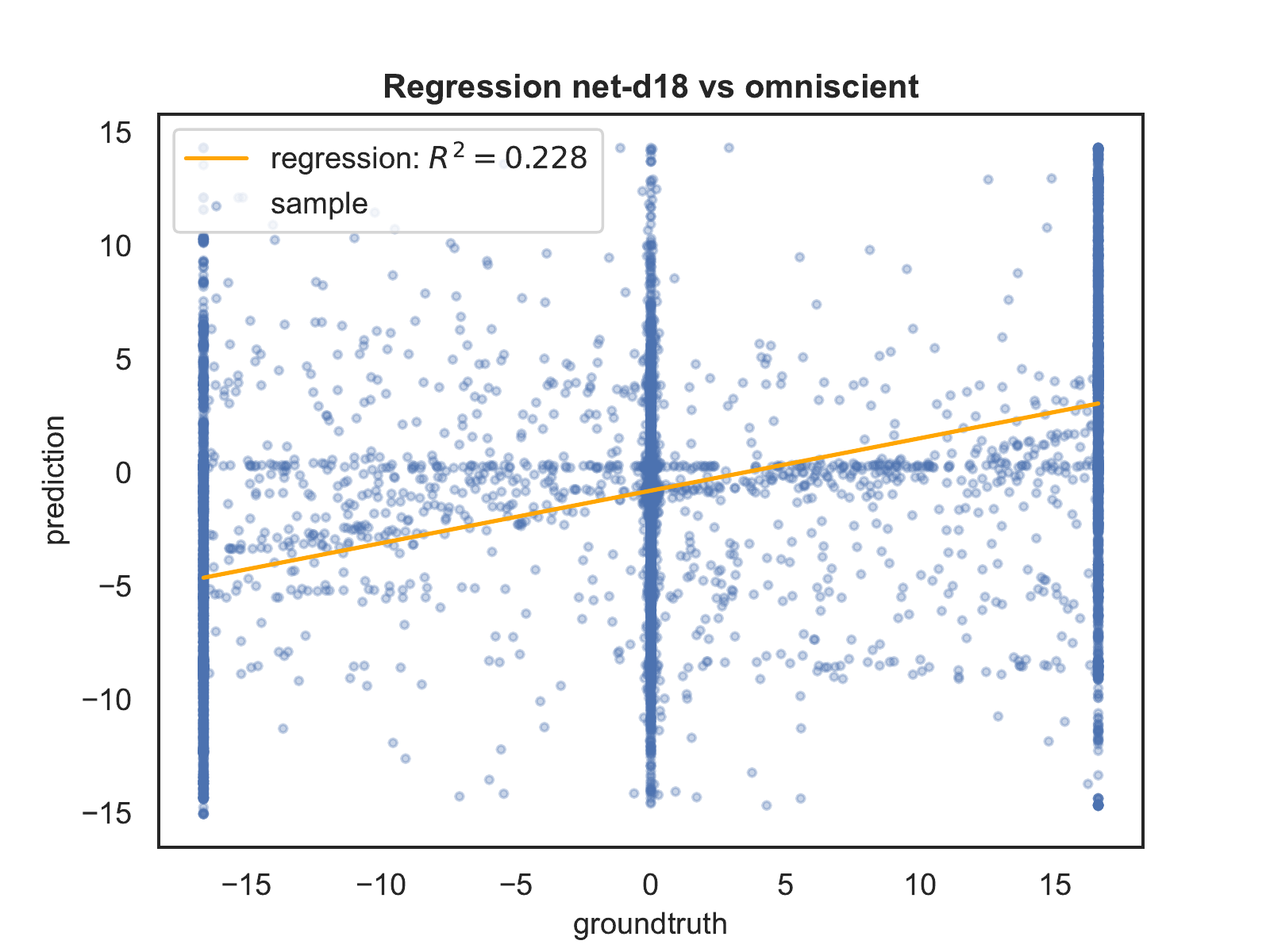}
	\end{subfigure}
	\caption[Evaluation of the \ac{r2} coefficients of \texttt{net-d18} 
	.]{Comparison of the \ac{r2} coefficient of the manual and the controller 
	learned from \texttt{net-d18} with respect to the omniscient one.}
	\label{fig:net-d18r2}
\end{figure}

In Figure \ref{fig:net-d9traj1} is shown a comparison of the trajectories obtained 
for a sample simulation.
\begin{figure}[!htb]
	\centering
	\includegraphics[width=.65\textwidth]{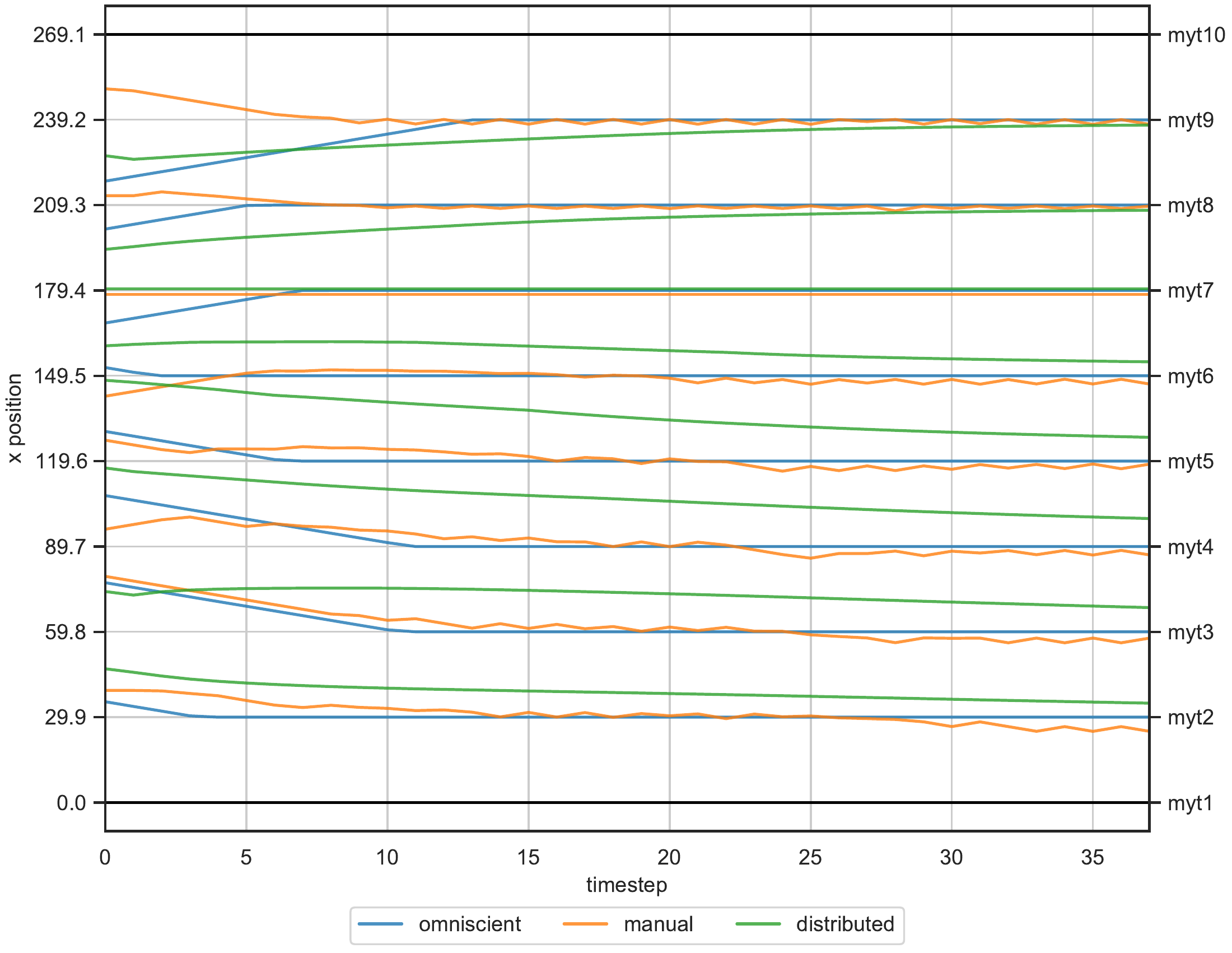}%
	\caption[Evaluation of the trajectories obtained with variable 
	agents.]{Comparison of trajectories, of a single simulation, generated using 
		three controllers: the expert, the manual and the one learned from 
		\texttt{net-d18}.}
	\label{fig:net-d18traj1}
\end{figure}  
In this example, in the simulation there are 10 agents that are always able to 
reach the target when moved using an omniscient controller.
When they use the manual controller, the same oscillation issue occurs in 
proximity  to the goal. Instead, the learned controller is certainly the slowest, and 
after 38 time steps not all robots are in the correct position.

Since the number of agents is variable, we show different plots, depending on this 
quantity, for the trajectories obtained employing the three controllers: we analyse, 
in the following figures, cases with 5, 8 and 10 agents. 
\begin{figure}[!htb]
	\begin{center}
		\begin{subfigure}[h]{0.325\textwidth}
			\centering
			\includegraphics[width=\textwidth]{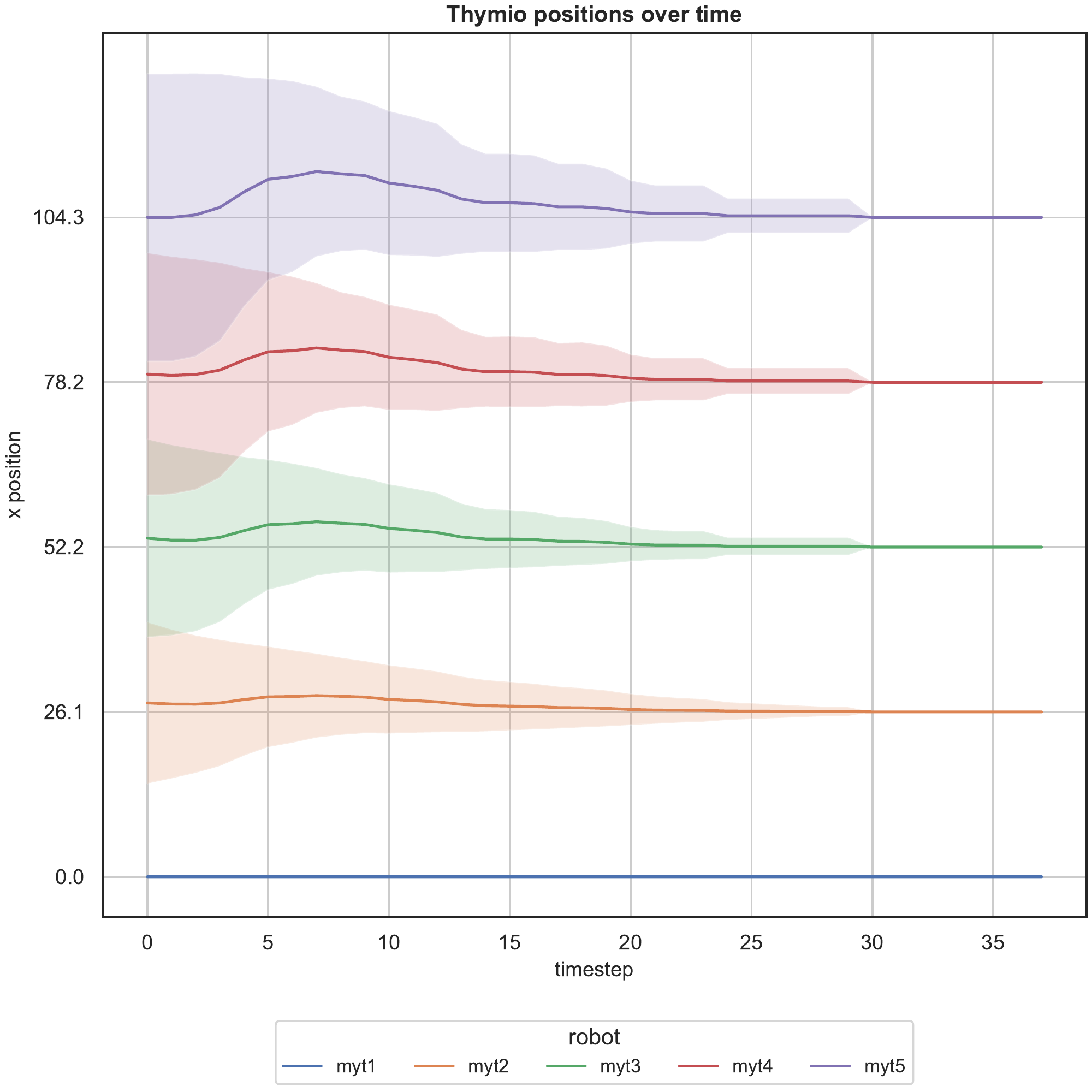}%
			\caption{Expert controller.}
		\end{subfigure}
		\hfill
		\begin{subfigure}[h]{0.325\textwidth}
			\centering
			\includegraphics[width=\textwidth]{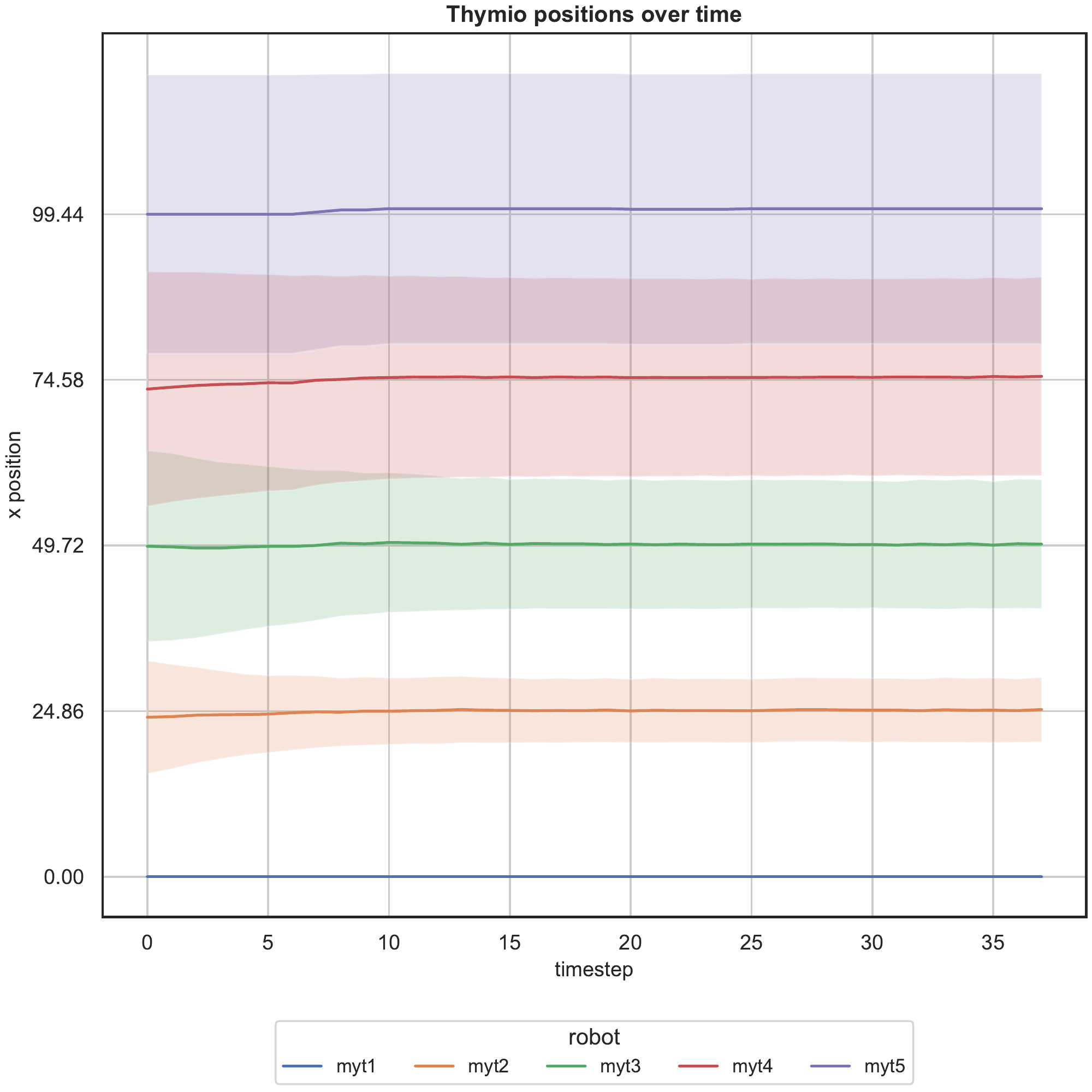}%
			\caption{Manual controller.}
		\end{subfigure}
		\hfill
		\begin{subfigure}[h]{0.325\textwidth}
			\centering
			\includegraphics[width=\textwidth]{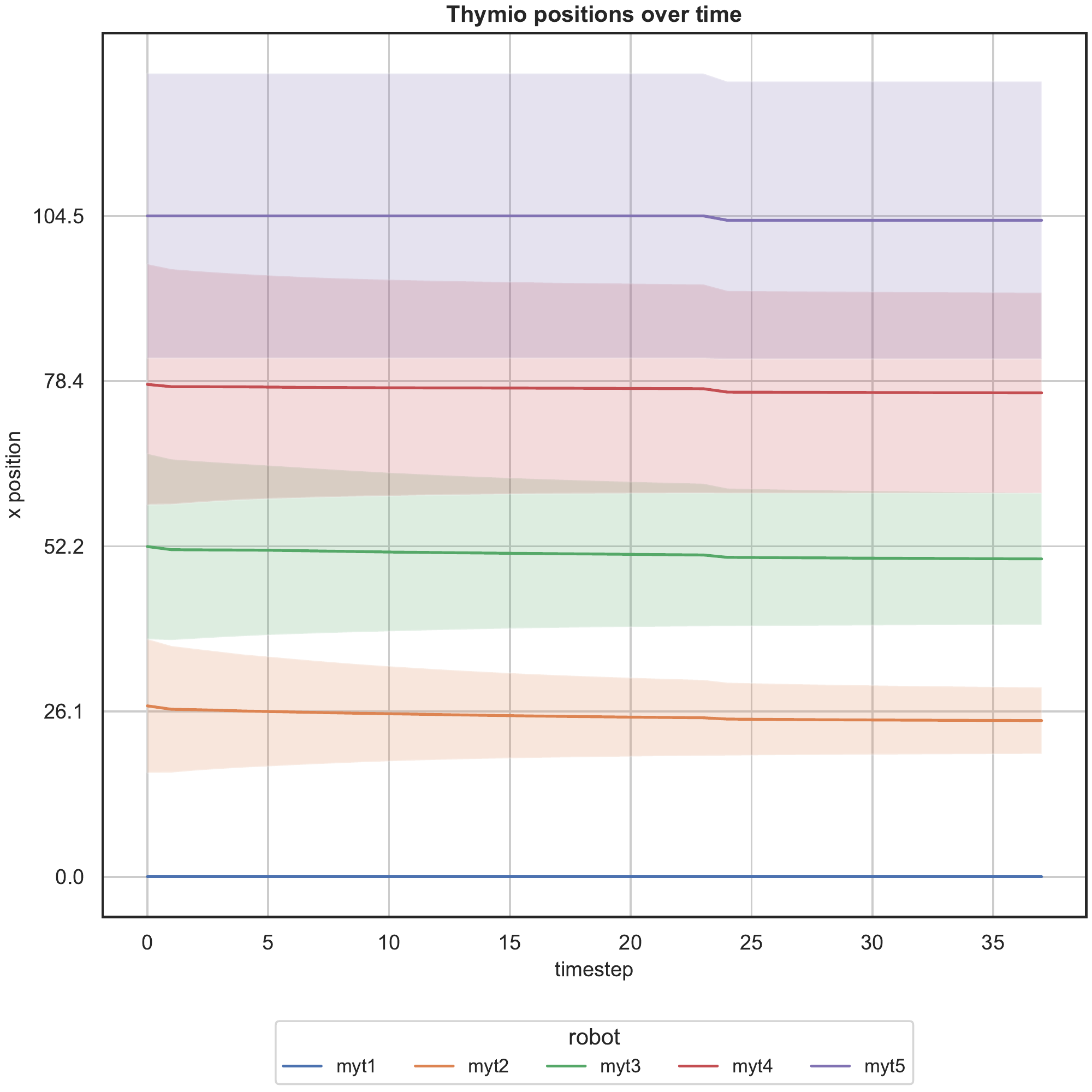}
			\caption{Distributed controller.}
		\end{subfigure}
	\end{center}
	\begin{center}
		\begin{subfigure}[h]{0.325\textwidth}
			\centering
			\includegraphics[width=\textwidth]{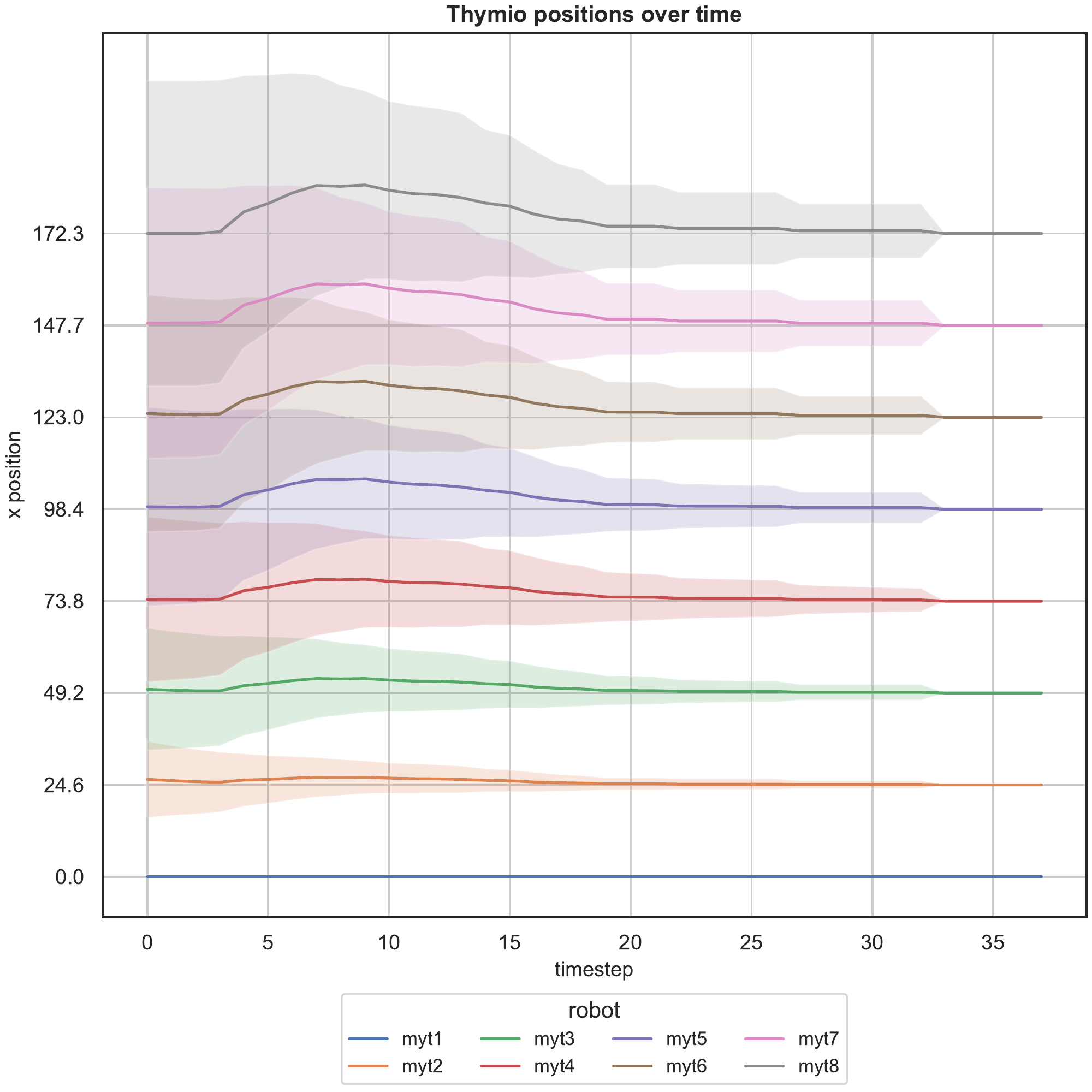}%
			\caption{Expert controller.}
		\end{subfigure}
		\hfill
		\begin{subfigure}[h]{0.325\textwidth}
			\centering
			\includegraphics[width=\textwidth]{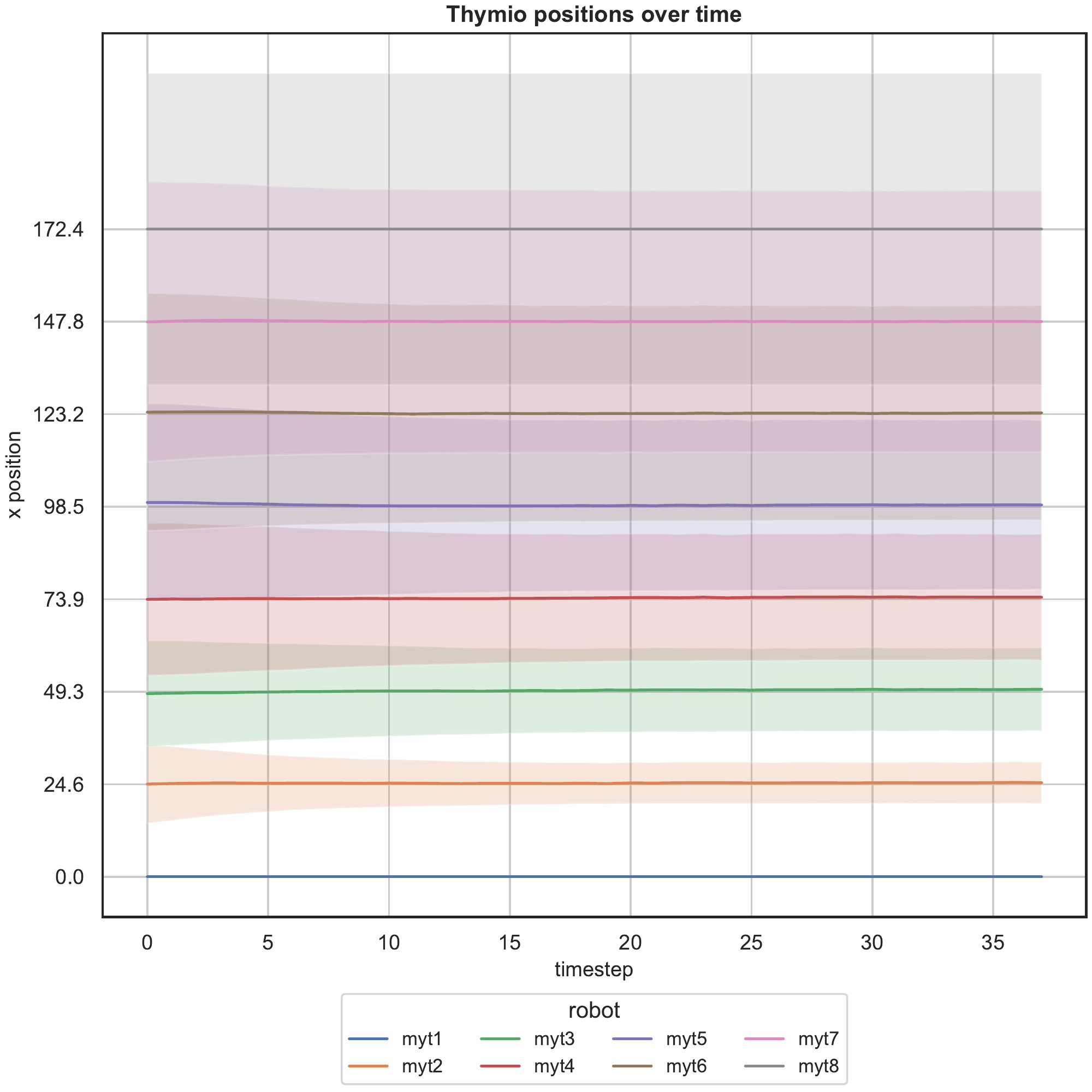}%
			\caption{Manual controller.}
		\end{subfigure}
		\hfill
		\begin{subfigure}[h]{0.325\textwidth}
			\centering
			\includegraphics[width=\textwidth]{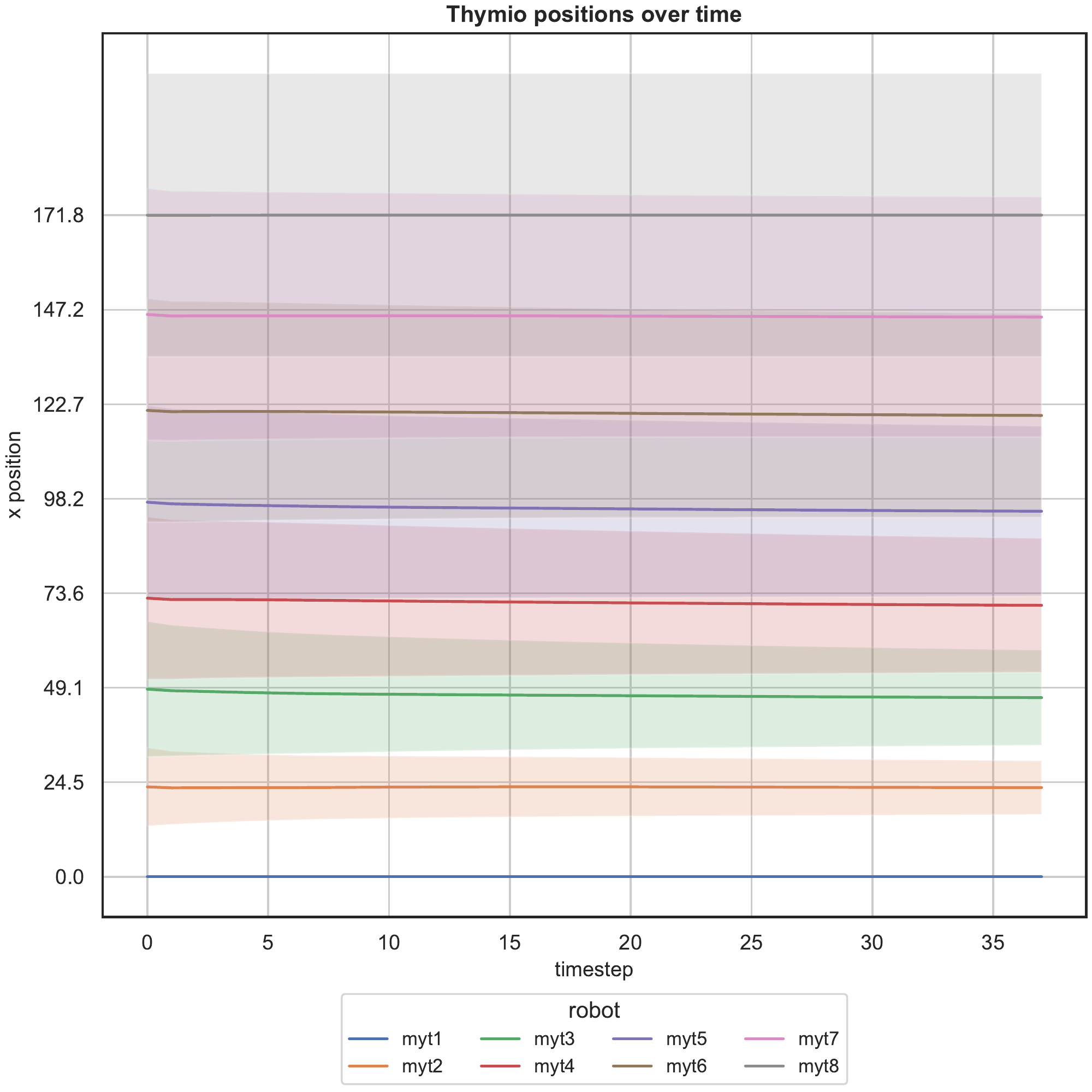}
			\caption{Distributed controller.}
		\end{subfigure}
	\end{center}
	\begin{center}
		\begin{subfigure}[h]{0.325\textwidth}
			\centering
			\includegraphics[width=\textwidth]{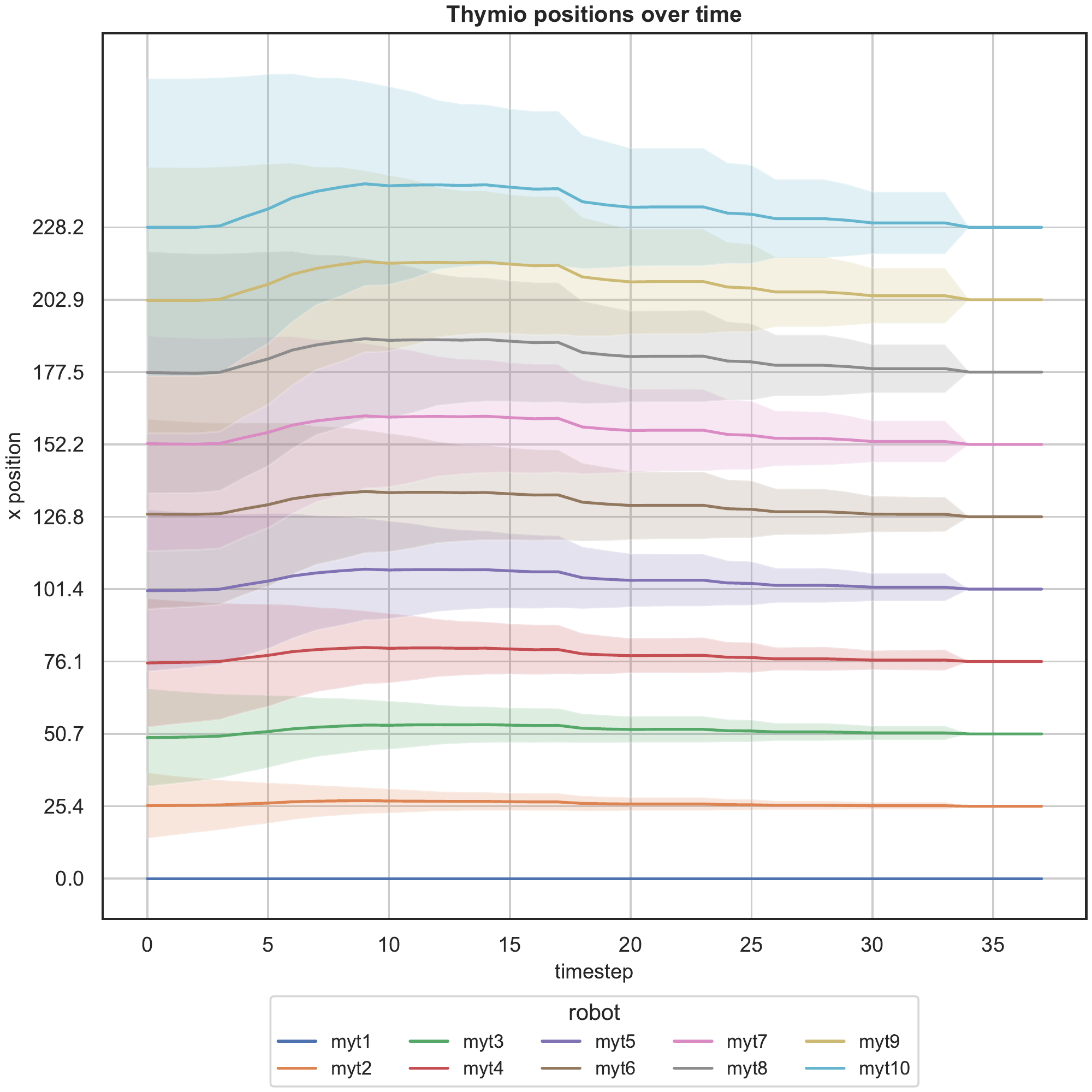}%
			\caption{Expert controller.}
		\end{subfigure}
		\hfill
	\begin{subfigure}[h]{0.325\textwidth}
		\centering
		\includegraphics[width=\textwidth]{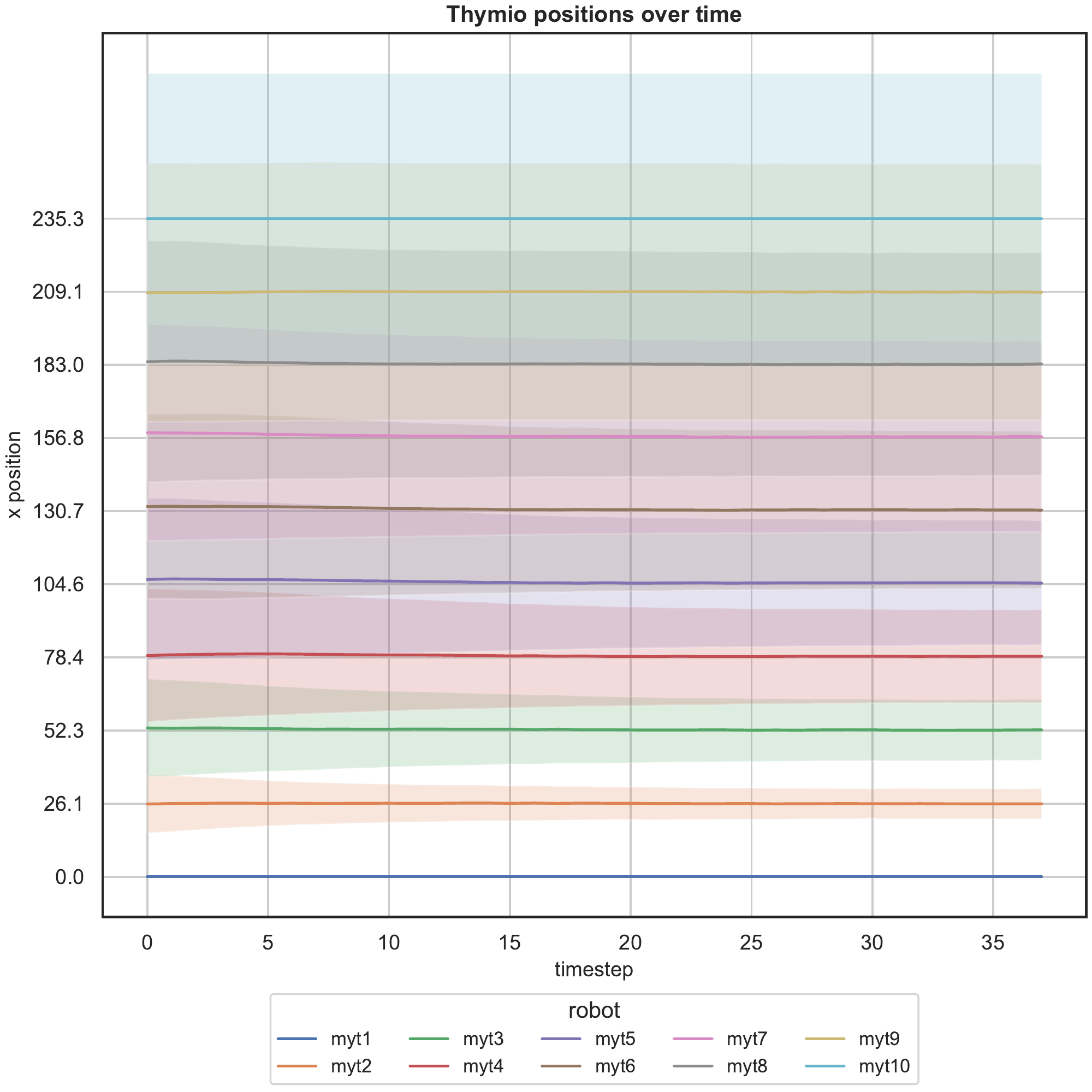}%
		\caption{Manual controller.}
	\end{subfigure}
	\hfill
	\begin{subfigure}[h]{0.325\textwidth}
		\centering
		\includegraphics[width=\textwidth]{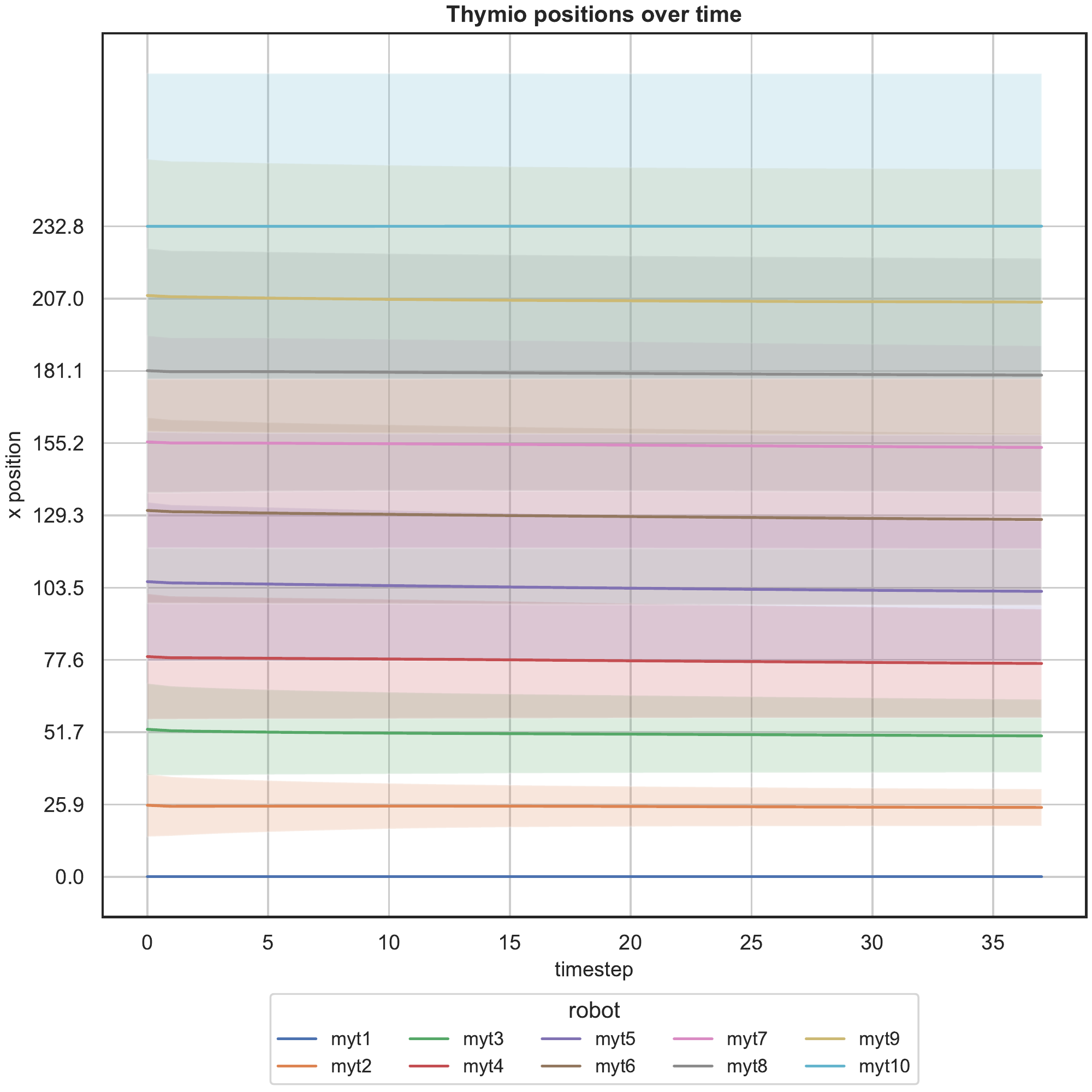}
		\caption{Distributed controller.}
	\end{subfigure}
\end{center}
\vspace{-0.5cm}
	\caption[Trajectories learned by \texttt{net-d18} using 5, 8 and 10 
	agents.]{Comparison of trajectories, of all the simulation runs, of 5, 8 and 10 
	agents generated using three controllers.}
	\label{fig:net-d18traj10}
\end{figure}
From a first observation it is confirmed that increasing the number of robots in 
the simulation implies a greater number of time steps to reach the final 
configuration.
Furthermore, with a large number of agents it is common for biases to add up 
and for the error to become more significant, in particular the one of the central 
robot of the group.
The convergence is still slow, even if this time the expert need less time steps than 
before. 
The learned controller is still the slowest to end up in the correct configuration.

Examining the evolution of the output control, in Figure \ref{fig:net-d18control}, 
the graph of the distributed controller highlights how the speed decided by the 
model has decreased due to the increase in the amount of agents in the 
simulation.
\begin{figure}[!htb]
	\centering
	\begin{subfigure}[h]{0.3\textwidth}
		\centering
		\includegraphics[width=\textwidth]{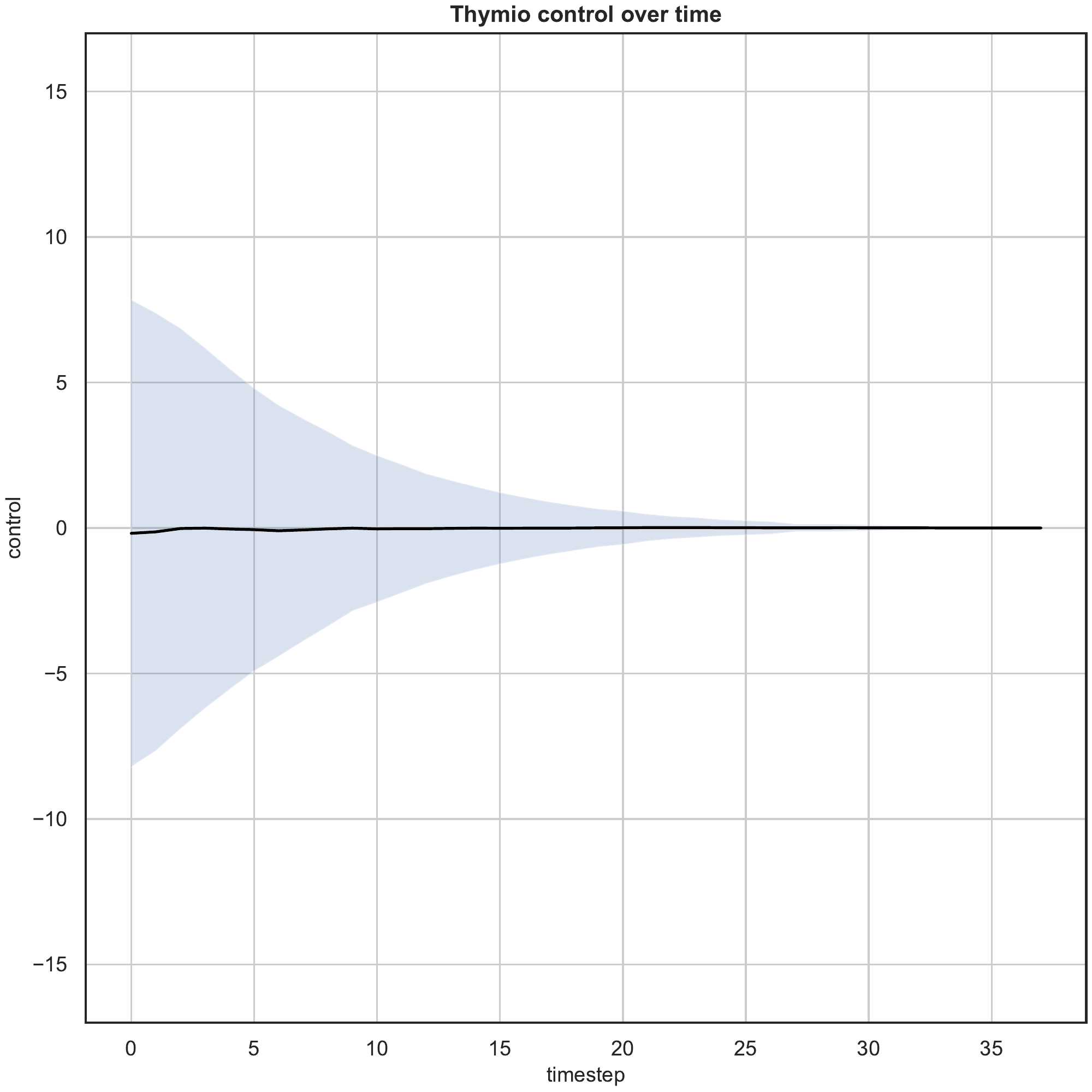}%
		\caption{Expert controller.}
	\end{subfigure}
	\hfill
	\begin{subfigure}[h]{0.3\textwidth}
		\centering
		\includegraphics[width=\textwidth]{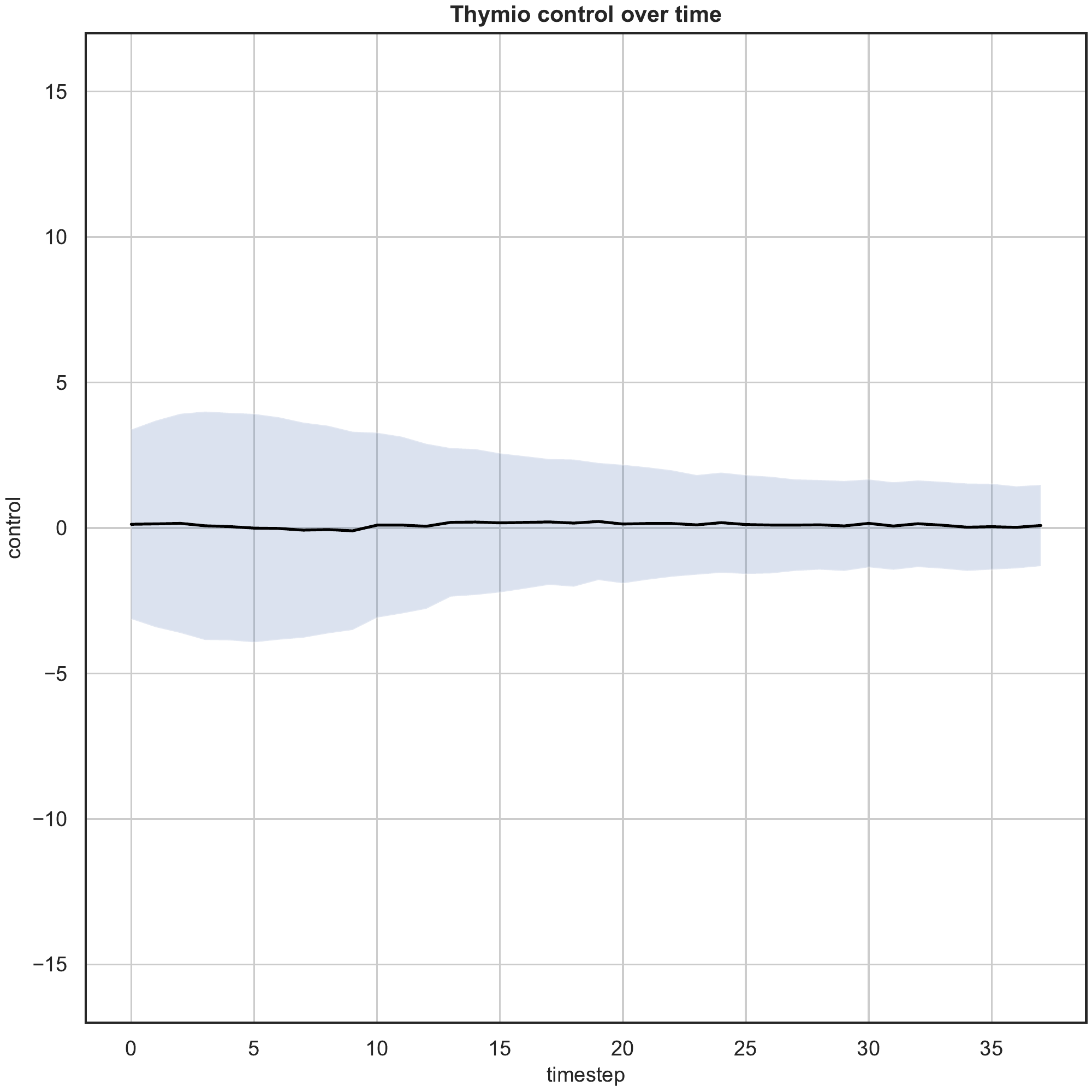}%
		\caption{Manual controller.}
	\end{subfigure}
	\hfill
	\begin{subfigure}[h]{0.3\textwidth}
		\centering
		\includegraphics[width=\textwidth]{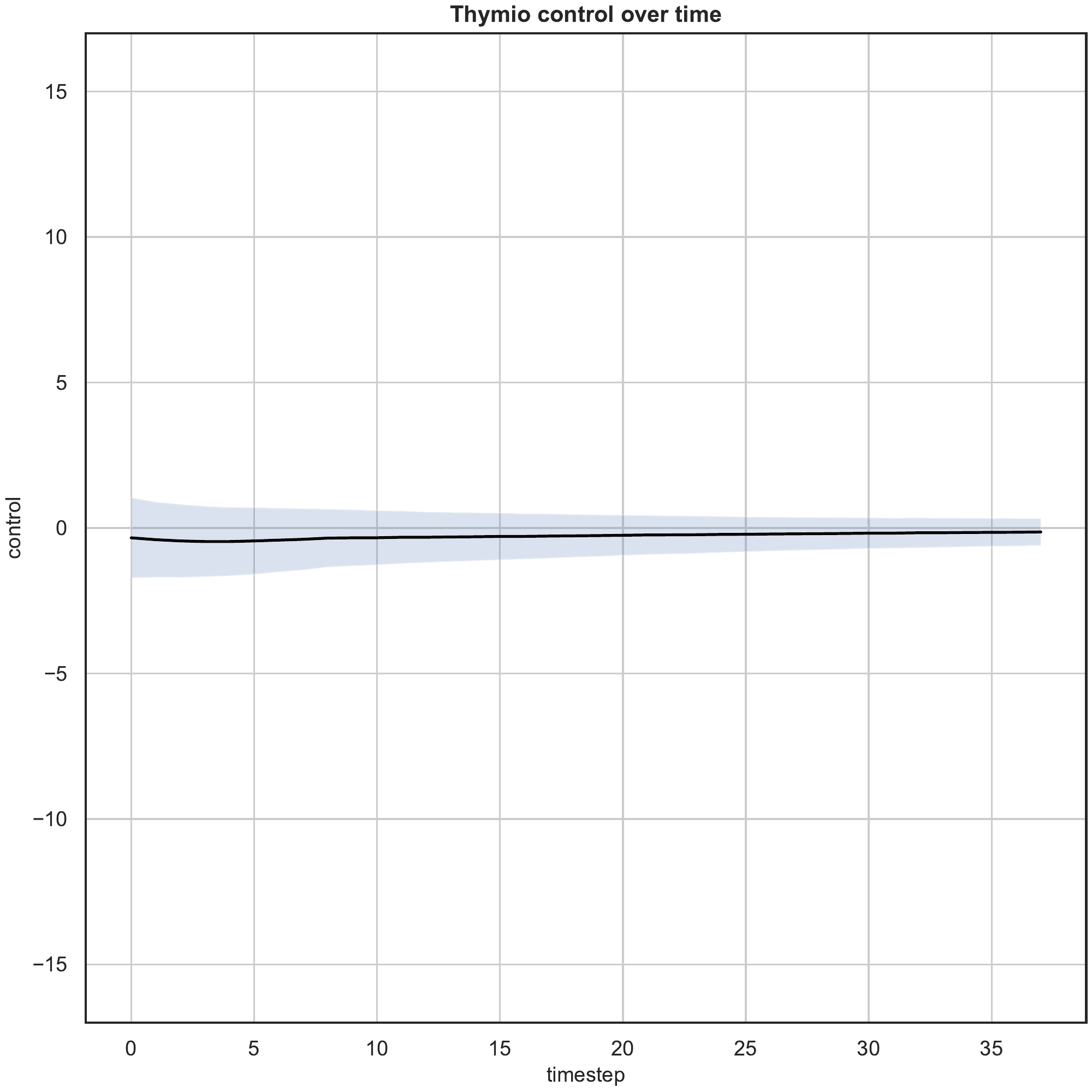}
		\caption{Distributed controller.}
	\end{subfigure}
	\caption[Evaluation of the control decided by \texttt{net-d18}.]{Comparison 
		of output control decided using three controllers: the expert, the manual and 
		the one learned from \texttt{net-d18}.}
	\label{fig:net-d18control}
\end{figure}

In Figure \ref{fig:net-d18responseposition} is displayed the behaviour of a robot 
located between other two that are already in their place.
\begin{figure}[!htb]
	\centering
	\includegraphics[width=.45\textwidth]{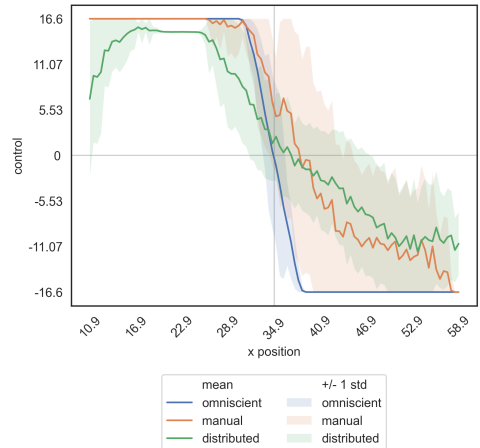}%
	\caption{Response of \texttt{net-d18} by varying the initial position.}
	\label{fig:net-d18responseposition}
\end{figure}
In this case the same reasoning made for Figure 
\ref{fig:net-d15responseposition} applies.

Focusing on the absolute distance of each robot from the target, presented in 
Figure \ref{fig:net-d18distance}, we observe once again that the agents moved 
following a manual controller in the final configuration are closer to the target 
than those moved with the learned one.
\begin{figure}[!htb]
	\centering
	\includegraphics[width=.65\textwidth]{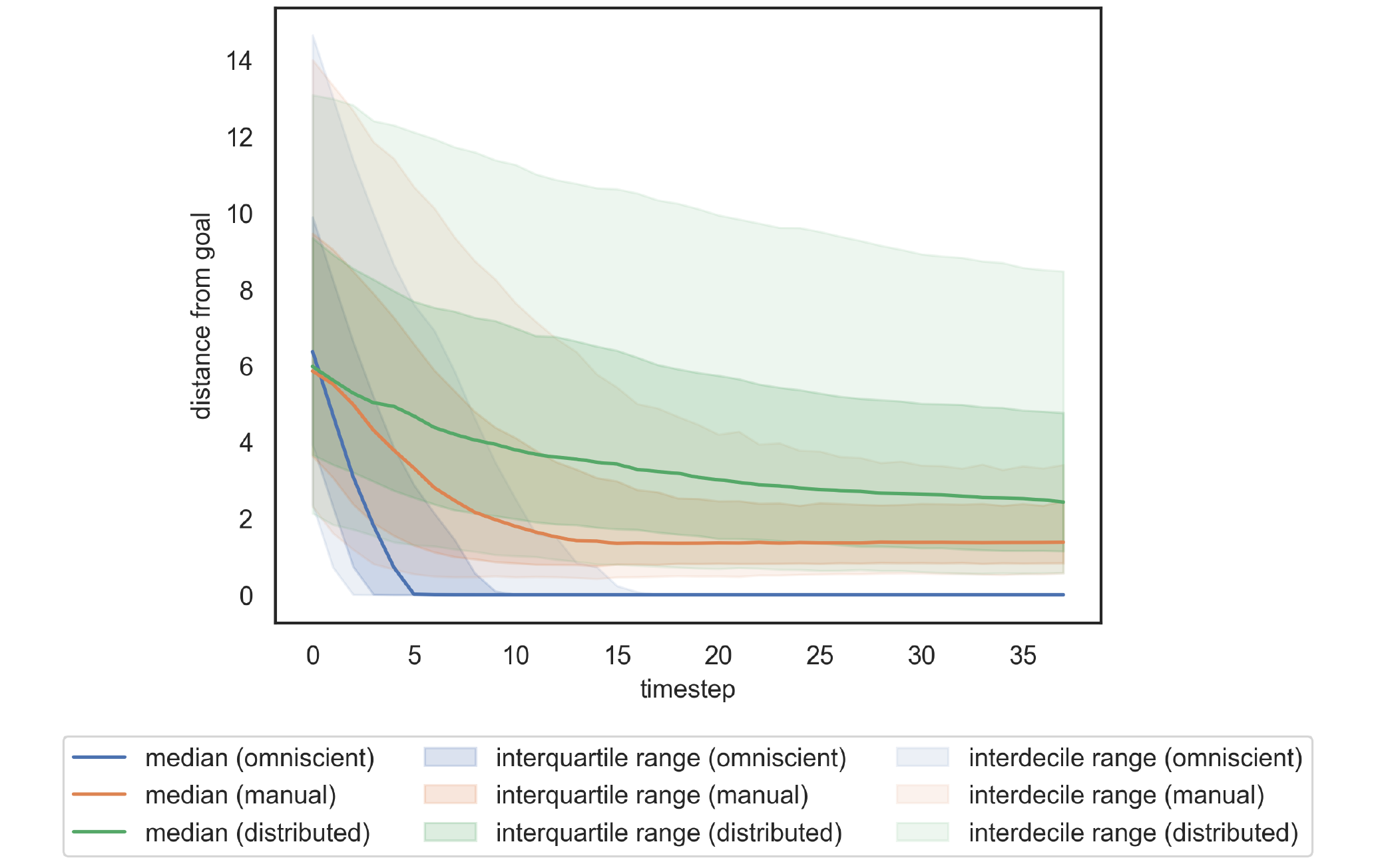}%
	\caption[Evaluation of \texttt{net-d18} distances from goal.]{Comparison of 
		performance in terms of distances from goal obtained using three 
		controllers: the expert, the manual and the one learned from 
		\texttt{net-d18}.}
	\label{fig:net-d18distance}
\end{figure}

\paragraph*{Summary}
To summarise the performance, as the number of agents vary for each gap, we 
show once again in the figures below the losses of the trained models.
In case of an \texttt{avg\_gap} of $8$cm, the model trained using 
a minor number of agents, as expected has a lower loss, following, with very 
similar values the model that employs 8 robots and that with a variable number of 
agents.
Finally, by choosing a variable gap and number of agents the performance are 
better than those generated with fixed but high number of robots. While again, 
the results obtained using fewer agents are the best.
\begin{figure}[!htb]
	\centering
	\begin{subfigure}[h]{0.3\textwidth}
		\centering
		\includegraphics[width=\textwidth]{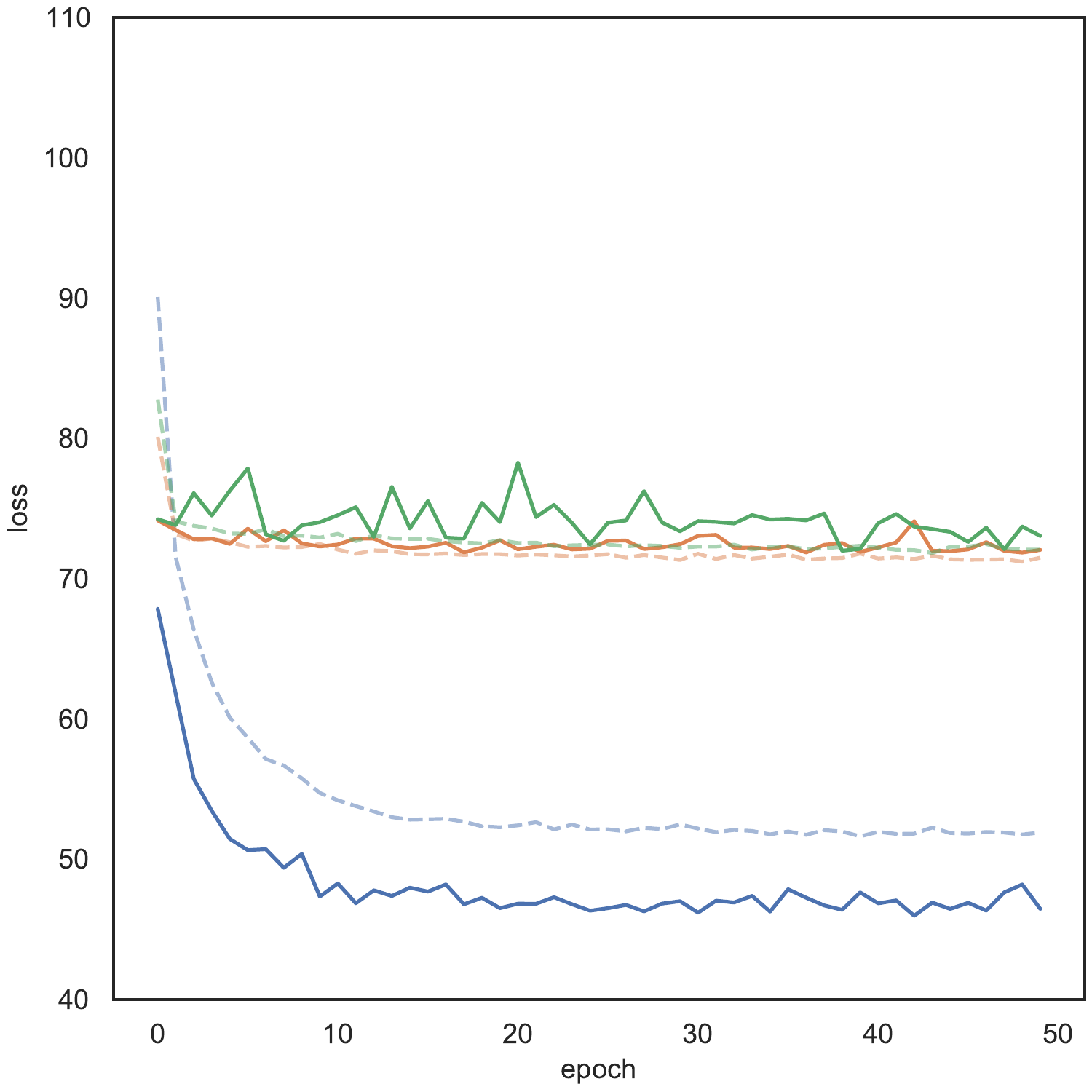}%
		\caption{\texttt{avg\_gap} of $8$cm.}
	\end{subfigure}
	\hfill
	\begin{subfigure}[h]{0.3\textwidth}
		\centering
		\includegraphics[width=\textwidth]{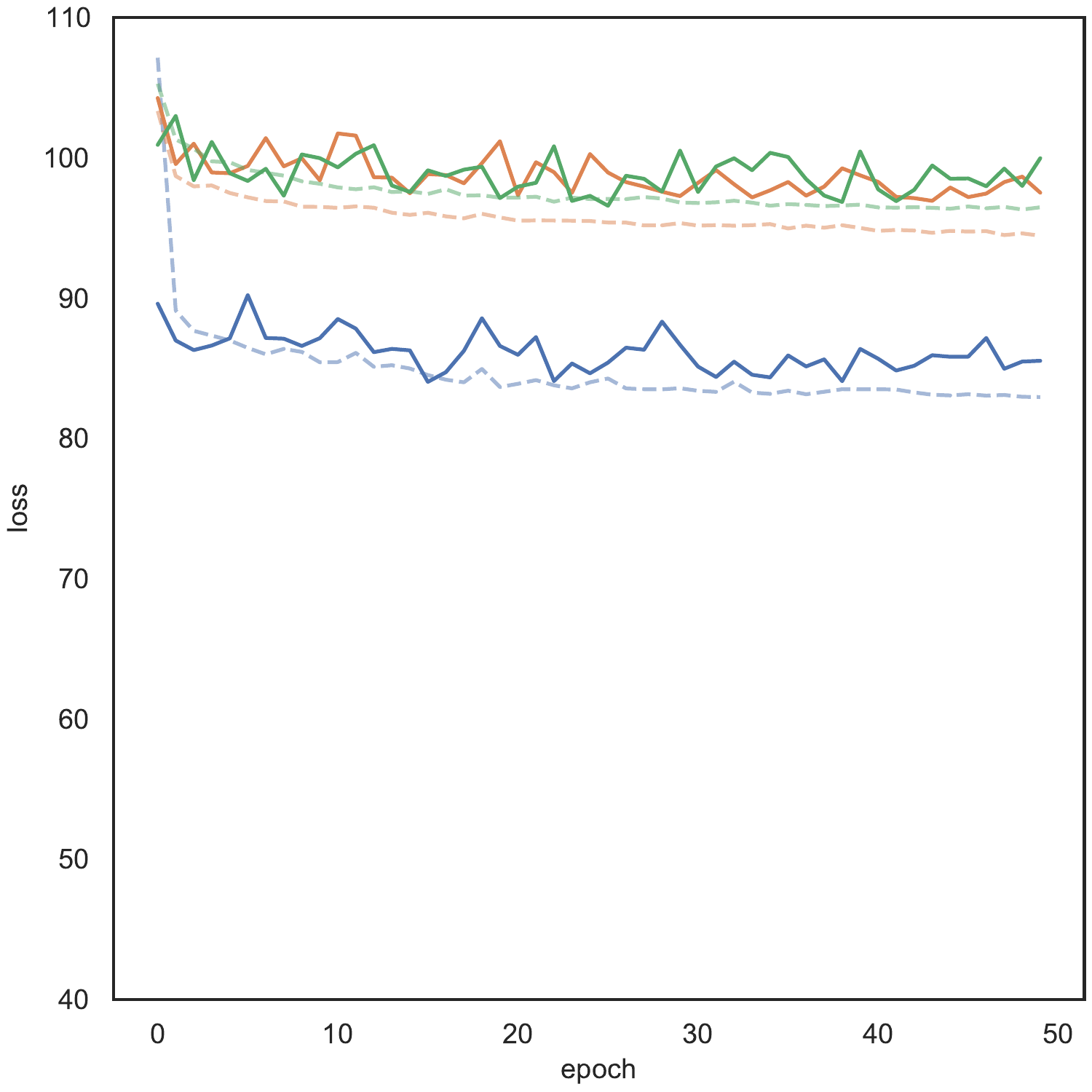}%
		\caption{\texttt{avg\_gap} of $20$cm.}
	\end{subfigure}
	\hfill
	\begin{subfigure}[h]{0.3\textwidth}
		\centering
		\includegraphics[width=\textwidth]{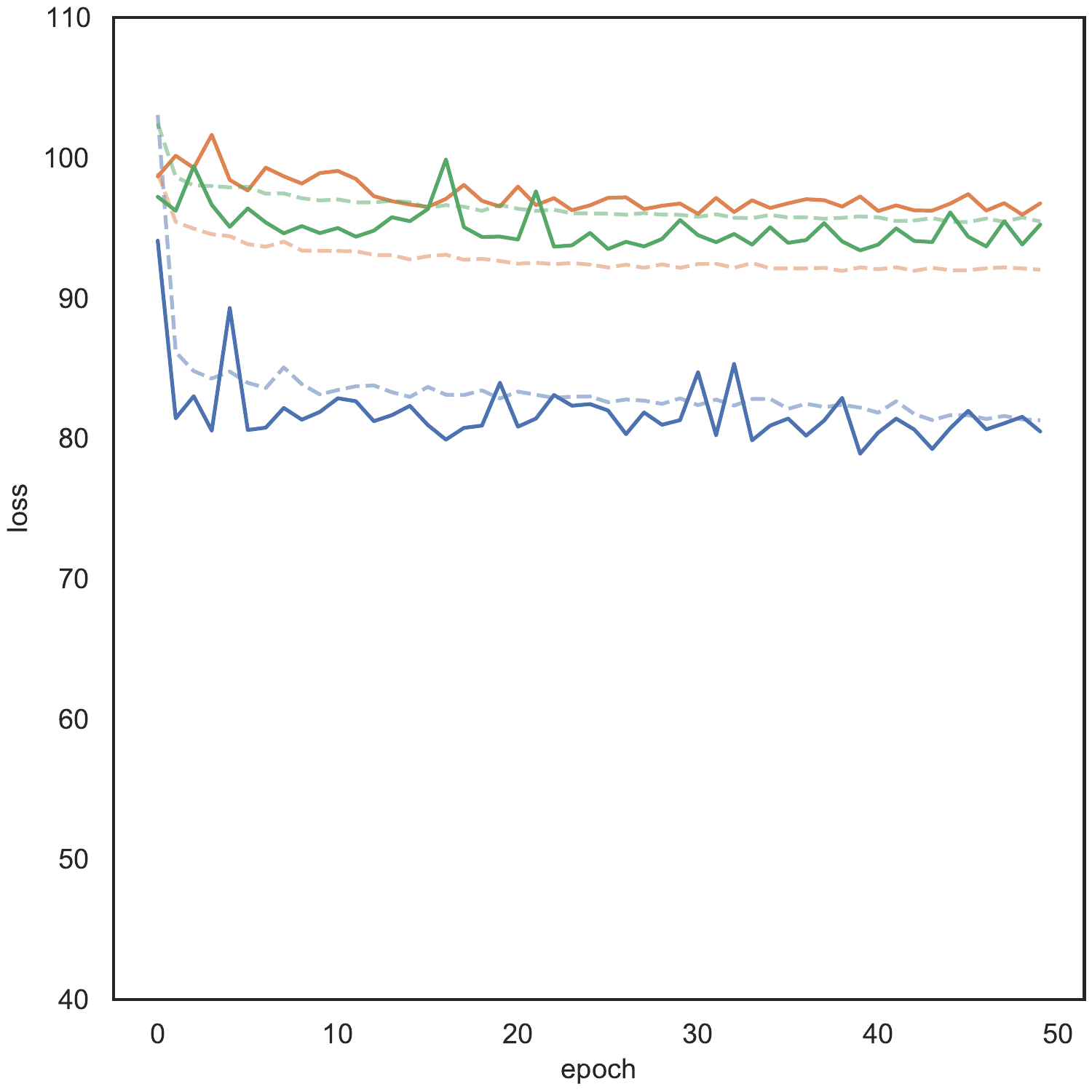}
		\caption{\texttt{avg\_gap} variable.}
	\end{subfigure}
		\caption[Losses summary of the second set of 
		experiments.]{Comparison of the losses by varying the input of the networks 
		for different gaps.}
	\label{fig:distloss820var}
	\vspace{-0.5cm}
\end{figure}

\subsubsection{Experiment 3: increasing number of agents}
\label{subsubsec:task1-exp-distr-3}
Multi-agent systems present interesting scalability challenges. For this reason, the 
objective of the last group of experiments is to show the behaviour of a network 
trained using \texttt{all\_sensors} input, variable gaps and number of agents, 
applied on simulations with a higher number of robots, from 5 up to 50.
\begin{figure}[!htb]
	\centering
	\includegraphics[width=.5\textwidth]{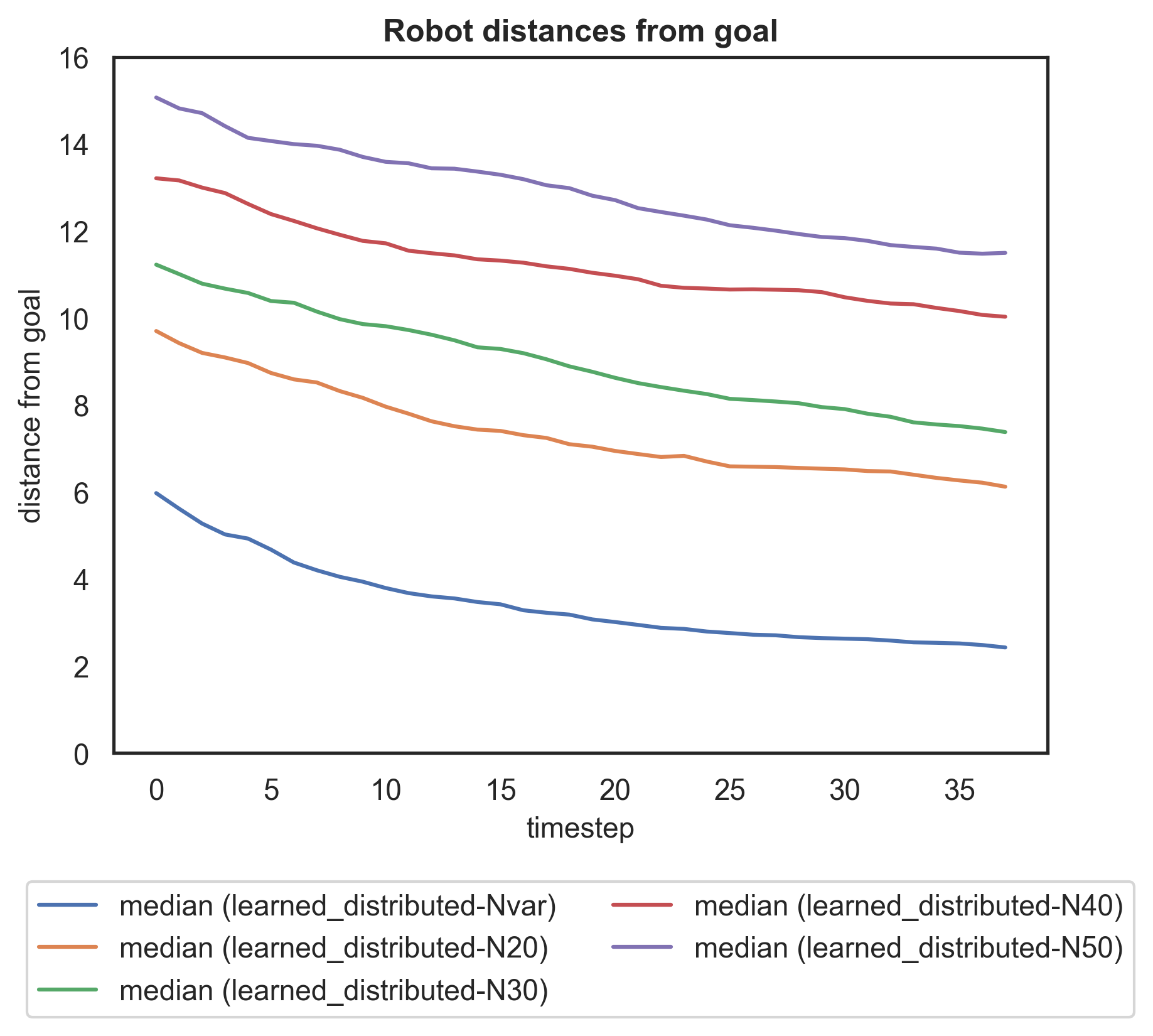}%
	\caption[Evaluation of distances from goal for a high number of 
	robots.]{Comparison of performance in terms of distances from goal obtained 
	on simulations with an increasing number of robots.}
	\label{fig:distdistr}
\end{figure}
In Figure \ref{fig:distdistr} is visualised, for 5 different experiments, the absolute 
distance of each robot from the target over time. This value is averaged on all 
robots among all the simulation runs. 
In general, as expected, the complexity grows rapidly as the number of agents 
increases and it is common for biases to add up and for the error to become more 
significant.
In the experiment performed with a variable number of agents, i.e.,in the range 
$[5, 10]$, in the final configuration the robots are on average at about $3$cm 
from the goal position. Increasing the number of agents, first to 20, then 30, 40 
and finally 50, the robots are more and more distant, in the worst case $13$cm 
from the target.

\subsubsection{Remarks}
\label{subsubsec:remarks-task1-dist}
In this section, we have shown that using a distributed controller learned by 
imitating an expert it is possible to obtain results more or less comparable to 
those reached employing a manual controller.
The problem presents interesting scalability challenges, and in general, a greater 
number of robots implies a slowdown in reaching the correct positions.
However, this approach is not enough to achieve satisfactory performance. In the 
following section we are going to describe a second approach that solve the 
problem by exploiting a communication protocol between agents.
\bigskip
\subsection{Distributed approach with communication}
\label{subsec:task1-exp-distr-comm}

\subsubsection{Experiment 1: fixed number of agents}
\label{subsubsec:task1-exp-comm-1}

In this section, we explore the same experiments carried out for the distributed approach without communication addressed in Section \ref{subsubsec:task1-exp-distr-1}, paying more attention to the cases with 
variable gaps and robots. 

\begin{figure}[!htb]
	\centering
	\begin{tabular}{cccc}
		\toprule
		\textbf{Model} \quad & \textbf{\texttt{network\_input}} & 
		\textbf{\texttt{input\_size}} &
		\textbf{\texttt{avg\_gap}} \\
		\midrule
		\texttt{net-c1} 				 & \texttt{prox\_values}	&  $  7$  &  $  8$  \\
		\texttt{net-c2} 			 	 & \texttt{prox\_values}	&  $  7$  &  $13$ \\
		\texttt{net-c3} 				 & \texttt{prox\_values}	&  $  7$  &  $24$  \\
		\texttt{net-c4} 				 & \texttt{prox\_comm}	  &  $  7$  &  $  8$  \\
		\texttt{net-c5} 				 & \texttt{prox\_comm}	  &  $  7$  &  $13$  \\
		\texttt{net-c6} 				 & \texttt{prox\_comm}	  &  $  7$  &  $24$  \\
		\texttt{net-c7} 				 & \texttt{all\_sensors}	  &  $14$  &  $  8$  \\
		\texttt{net-c8} 				 & \texttt{all\_sensors}	  &  $14$  &  $13$ 	\\
		\texttt{net-c9} 				 & \texttt{all\_sensors}	  &  $14$  &  $24$ 	\\
		\bottomrule
	\end{tabular}
	\captionof{table}[Experiments with $5$ agents (communication).]{List of the 
		experiments carried out with $5$ agents using communication.}
	\label{tab:modeln5comm}
\end{figure}

The first analysis, which we summarise in Table \ref{tab:modeln5comm}, concerns 
the behaviour of the learned controllers in case of the three different inputs, 
\texttt{prox\_values}, \texttt{prox\_comm} or \texttt{all\_sensors}, for a number 
of robots $N$ and an \texttt{avg\_gap} both fixed respectively at $5$ and the 
second chosen between $8$, $13$ and $24$.
The performance of these model in terms of train and validation losses are shown 
in Figure \ref{fig:commlossallt}. It is immediately evident that the trend of the 
curves are very similar to that obtained with the distributed approach. 
\begin{figure}[!htb]
	\centering
	\includegraphics[width=.8\textwidth]{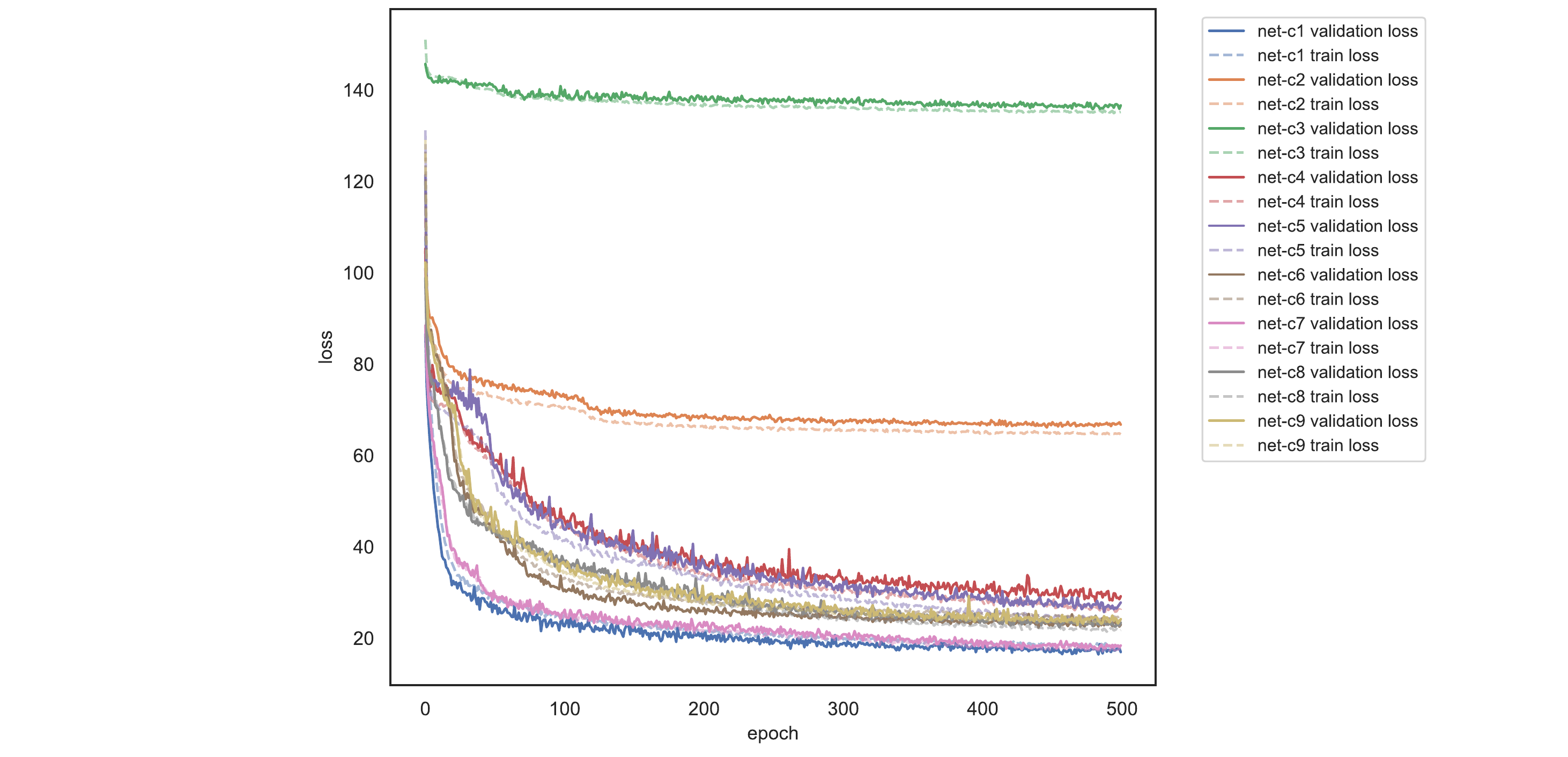}%
	\caption[Comparison of losses of the second set of experiments.]{Comparison 
	of the losses of the models carried out using a variable number of agents and of 
	average gap.}
	\label{fig:commlossallt}
	\vspace{-0.5cm}
\end{figure}

\paragraph*{Results using \texttt{prox\_values} input}
We continue by analysing and comparing the performances obtained from these 
experiments and those in Section \ref{para:1}.

\begin{figure}[!htb]
	\begin{center}
		\begin{subfigure}[h]{0.49\textwidth}
			\centering
			\includegraphics[width=.65\textwidth]{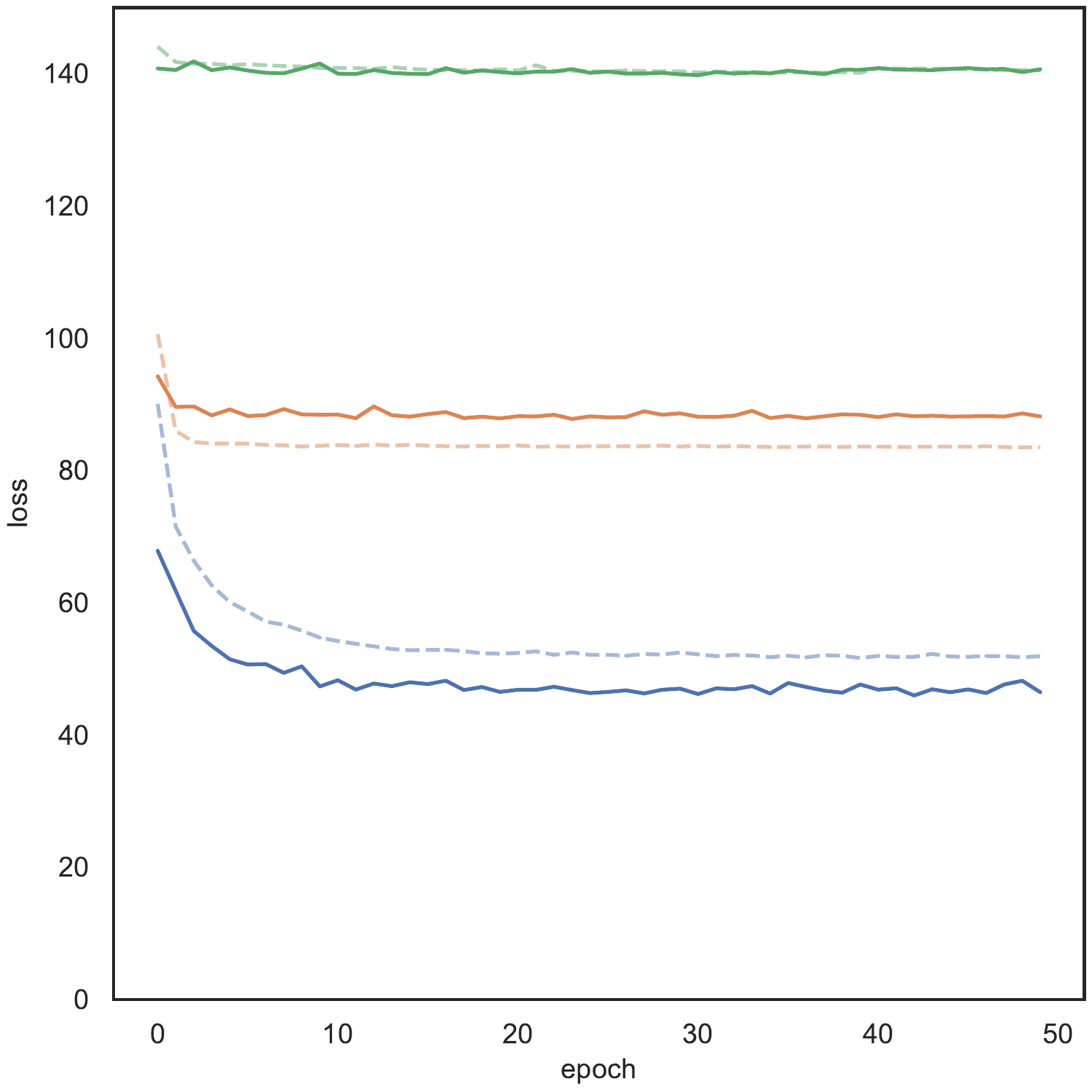}
			\caption{Distributed approach.}
		\end{subfigure}
		\hfill
		\begin{subfigure}[h]{0.49\textwidth}
			\centering
			\includegraphics[width=.7\textwidth]{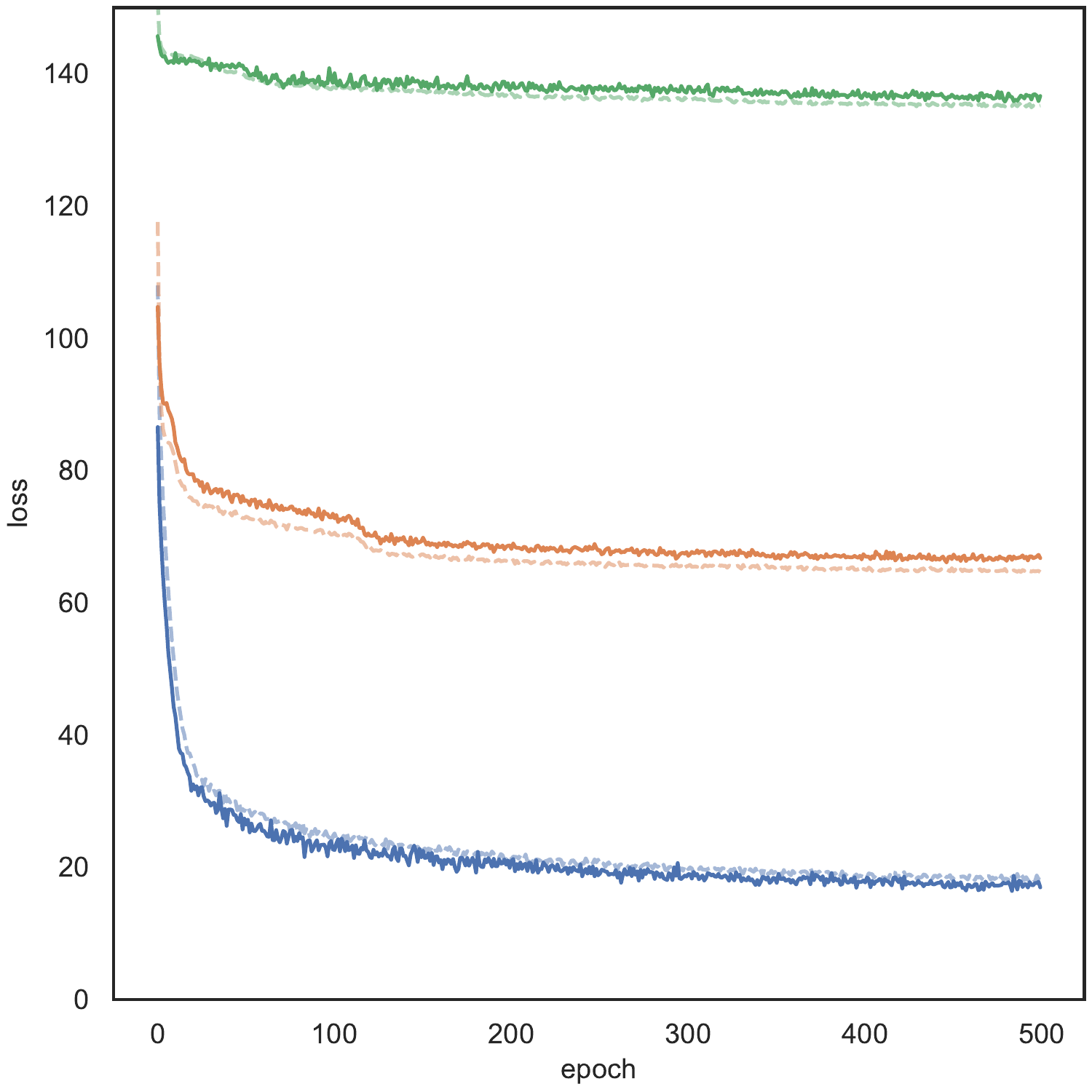}
			\caption{Distributed approach with communication.}
		\end{subfigure}	
	\end{center}
	\vspace{-0.5cm}
	\caption{Comparison of the losses of the models that use \texttt{prox\_values} 
		readings.}
	\label{fig:commlossprox_values}
\end{figure}

In Figure \ref{fig:commlossprox_values} is shown an overview of the performance, 
in terms of loss, of the models trained using \texttt{prox\_values} and varying the 
average gap: the blue, orange and green lines represent respectively gaps of $8$, 
$13$ and $24$cm.
The loss in case of the smaller gap is decreased from $40$ to $20$, meaning an 
improvement over the previous approach.

Focusing on the models that use the \texttt{prox\_values} readings as input and 
the dataset generated using $8$cm as average gap, in Figure 
\ref{fig:net-c1r2} we observe the \ac{r2} of the manual and the learned 
controllers, in both cases, i.e.,with and without communication, on the validation 
sets.
From these figures, as previously mentioned, we expect that the behaviour of the 
robots using the distributed controller is better than the manual one, even if 
far from the expert. On the other hand, adding the communication to the model 
produces an increase in the coefficient \ac{r2} from $0.59$ to $0.85$, thus 
promising superior performance and an attitude more similar to that of the 
omniscient controller.
\begin{figure}[!htb]
	\begin{center}
		\begin{subfigure}[h]{0.49\textwidth}
			\includegraphics[width=\textwidth]{contents/images/net-d1/regression-net-d1-vs-omniscient}%
		\end{subfigure}
		\hfill\vspace{-0.5cm}
		\begin{subfigure}[h]{0.49\textwidth}
			\includegraphics[width=\textwidth]{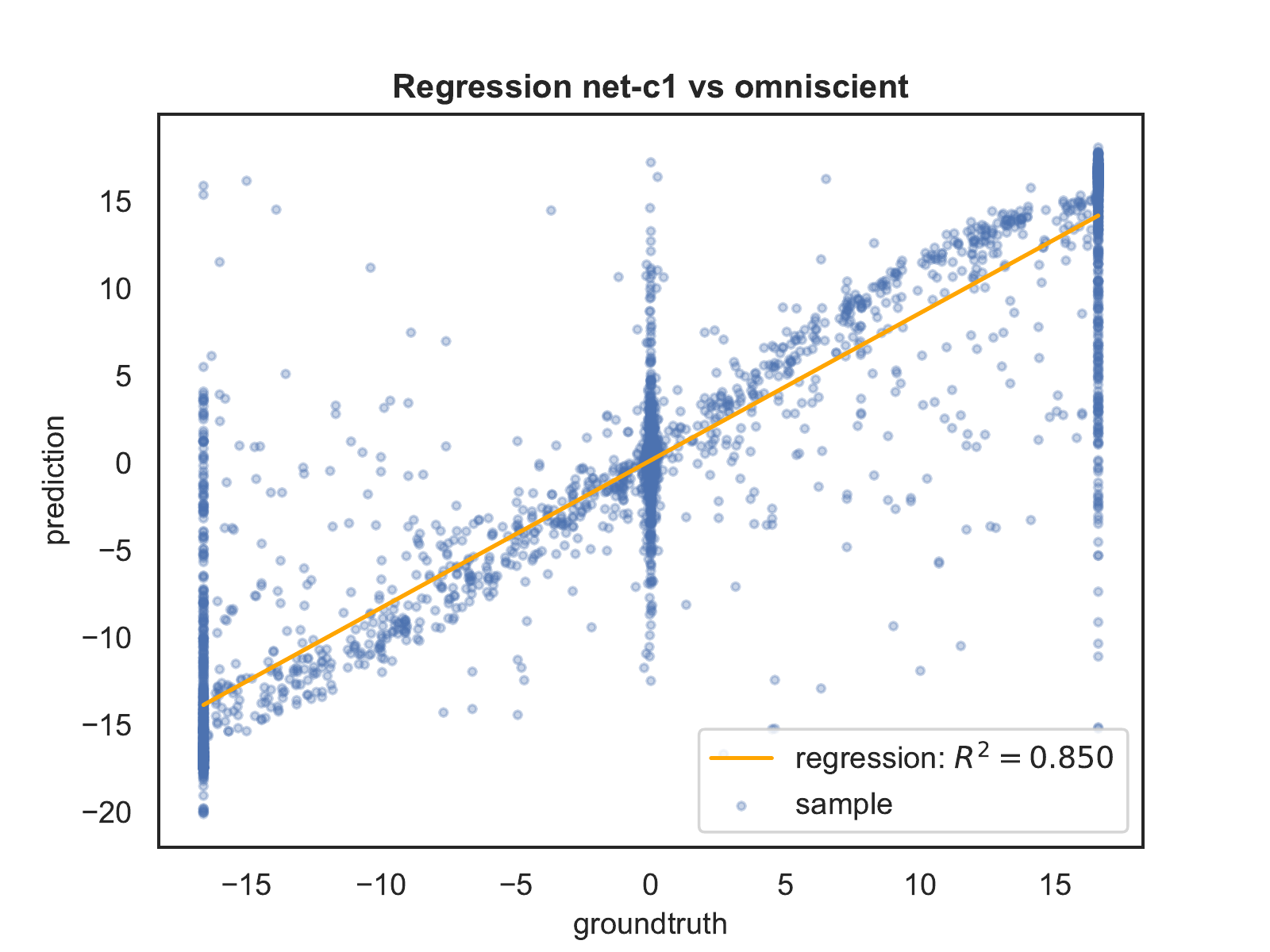}%
		\end{subfigure}
	\end{center}
	\caption[Evaluation of the \ac{r2} coefficients of \texttt{net-c1}.]{Comparison 
		of the \ac{r2} coefficients of the controllers learned from \texttt{net-d1} and 
		\texttt{net-c1}, with respect to the omniscient one.}
	\label{fig:net-c1r2}
\end{figure}

In Figure \ref{fig:net-c1traj1} are visualised the trajectories of the agents, in a 
sample simulation, over time.
The agents moved using the omniscient controller are those that reach the target  
faster. Those moved using the manual controller approach the goal position, on 
average in 10 time steps, but never reach it. The controller learned from 
\texttt{net-d1} in about 25 time steps is able to let the agents arrive in the correct 
final configuration. Instead, the new controller, learned from \texttt{net-c1}, is 
faster than both the previous and the manual one: in less than 5 time steps the 
robots reach their goal.
\begin{figure}[!htb]
	\centering
	\includegraphics[width=.65\textwidth]{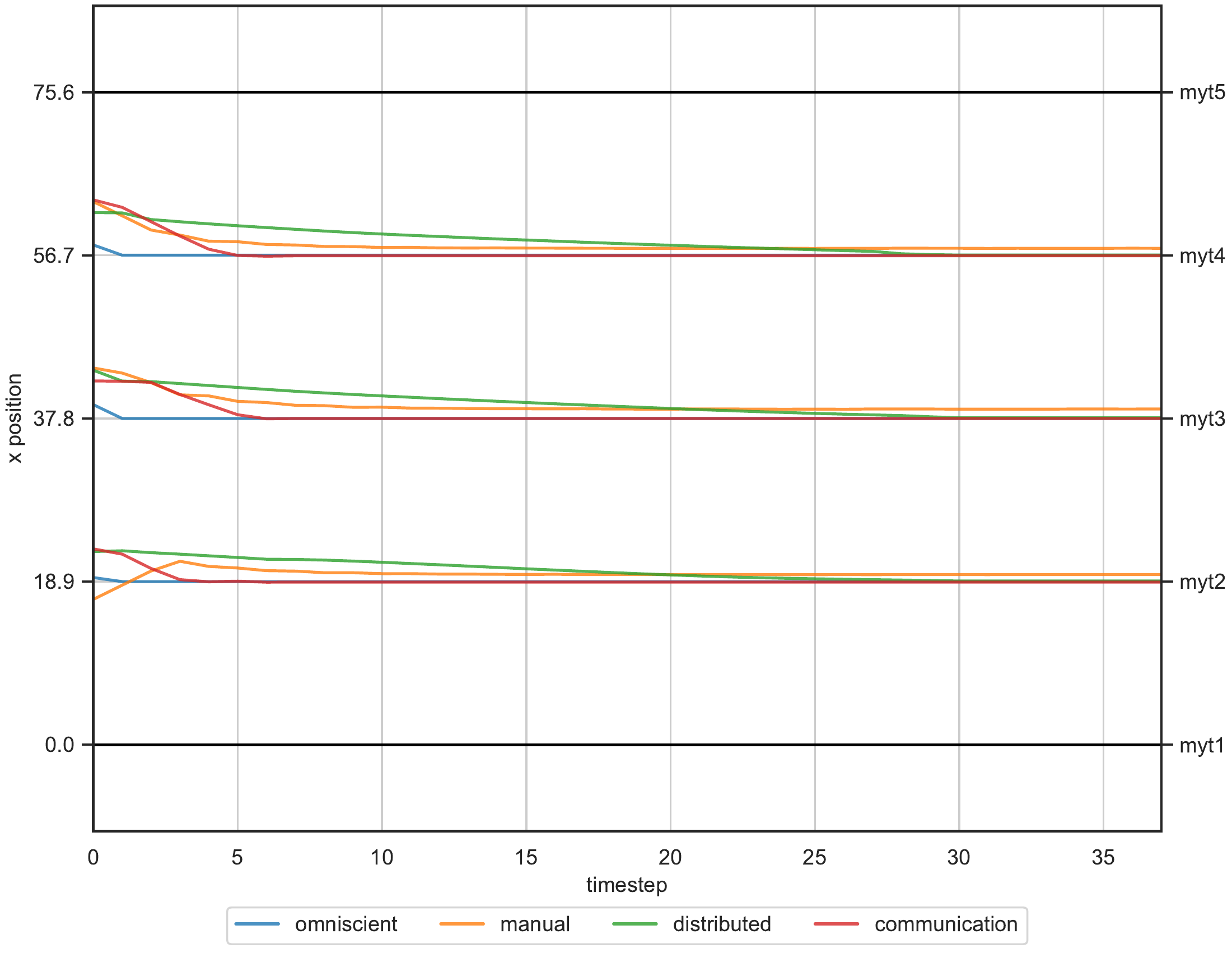}%
	\caption[Evaluation of the trajectories obtained with \texttt{prox\_values} 
	input.]{Comparison of trajectories, of a single simulation, generated using four 
	controllers: the expert, the manual and the two learned from \texttt{net-d1} 
	and \texttt{net-c1}.}
	\label{fig:net-c1traj1}
\end{figure}

\begin{figure}[H]
	\begin{center}
		\begin{subfigure}[h]{0.49\textwidth}
			\centering
			\includegraphics[width=.9\textwidth]{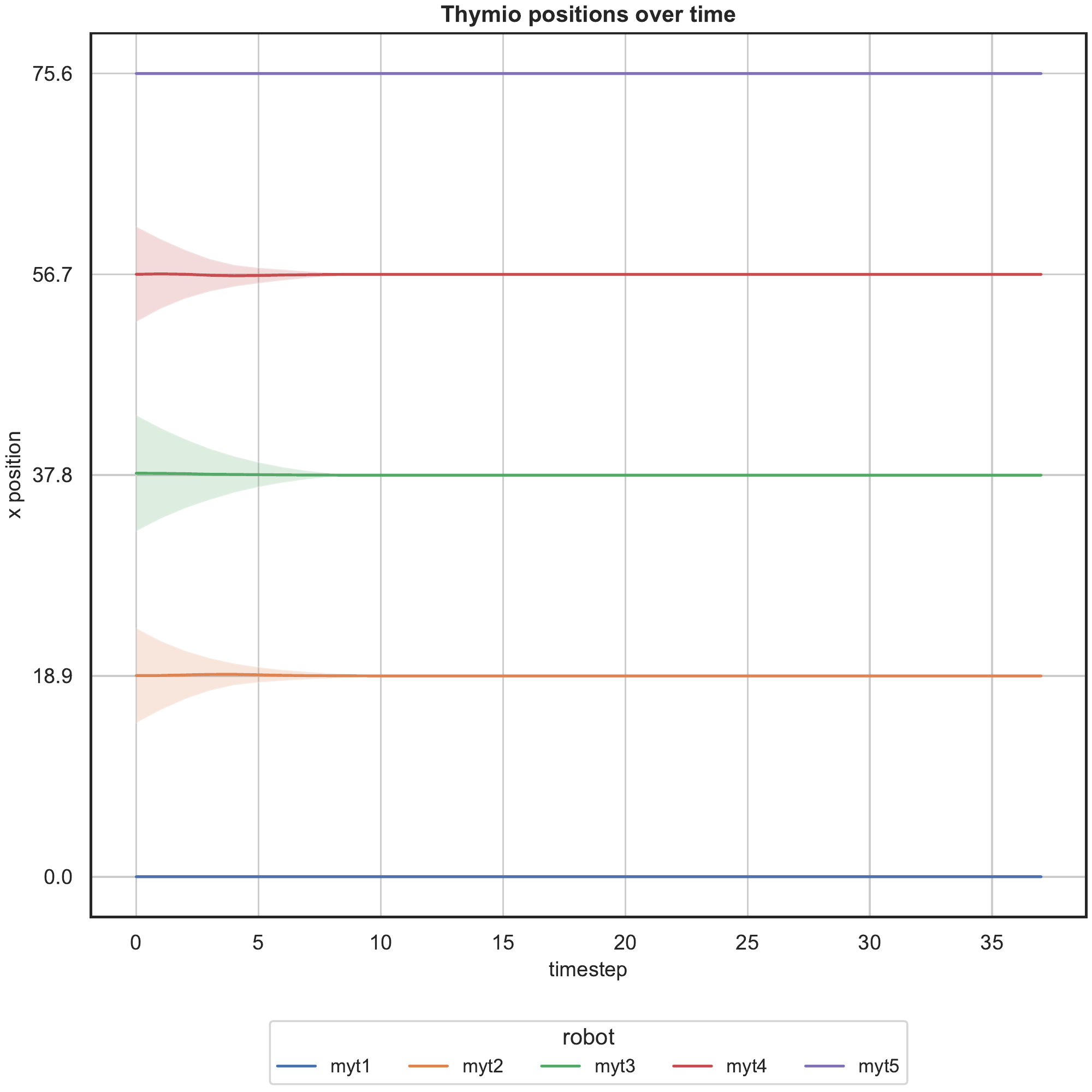}%
			\caption{Expert controller trajectories.}
		\end{subfigure}
		\hfill
		\begin{subfigure}[h]{0.49\textwidth}
			\centering
			\includegraphics[width=.9\textwidth]{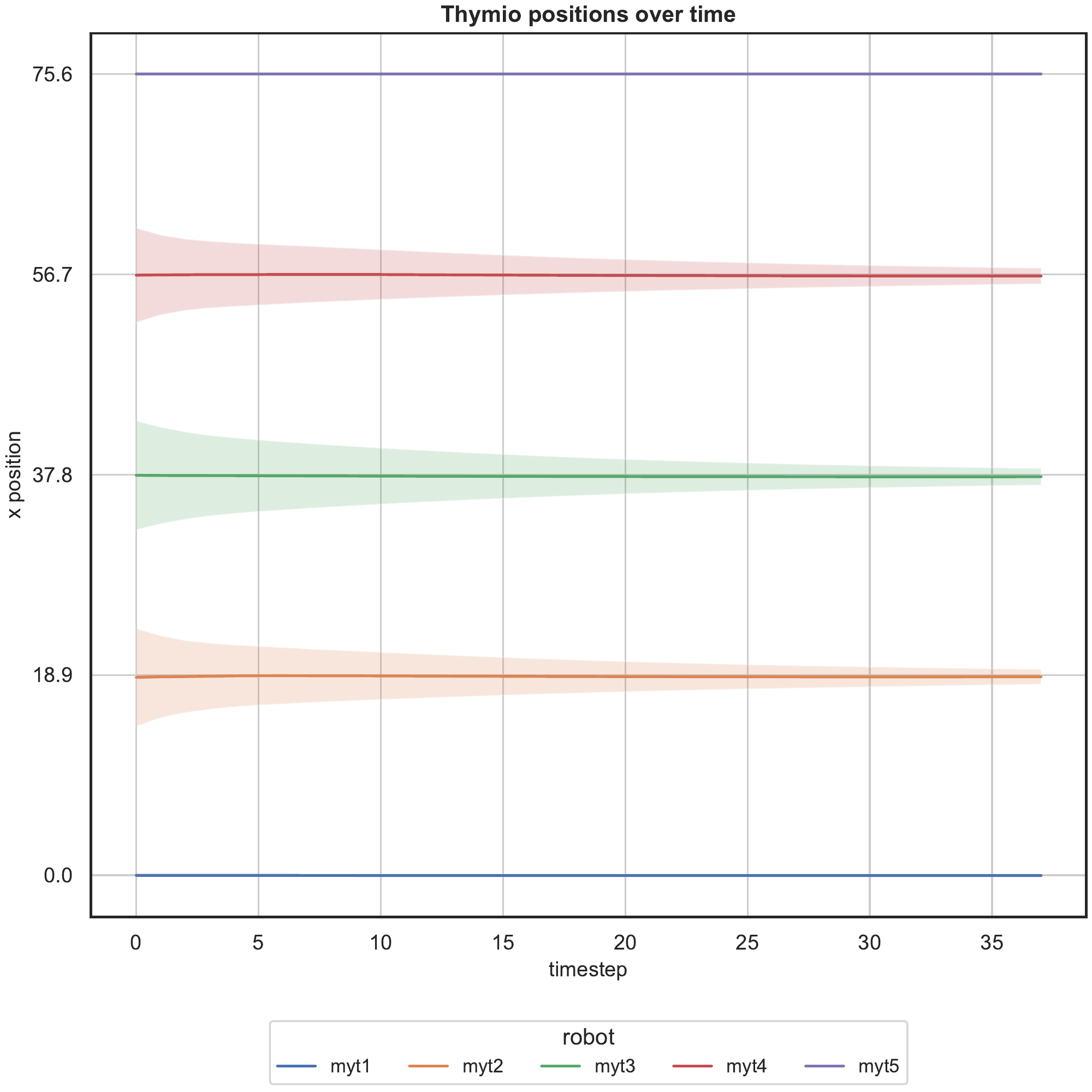}
			\caption{Distributed controller trajectories.}
		\end{subfigure}
	\end{center}
	\vspace{-0.5cm}
	\caption[Evaluation of the trajectories learned by \texttt{net-c1}.]{Comparison 
	of trajectories, of all the simulation runs, generated using four controllers: the 
	expert, the manual and the two learned from \texttt{net-d1} and 
	\texttt{net-c1}.}
\end{figure}

\begin{figure}[!htb]\ContinuedFloat
	\begin{center}
		\begin{subfigure}[h]{0.49\textwidth}
			\centering			
			\includegraphics[width=.9\textwidth]{contents/images/net-d1/position-overtime-manual}%
			\caption{Manual controller trajectories.}
		\end{subfigure}
		\hfill
		\begin{subfigure}[h]{0.49\textwidth}
			\centering
			\includegraphics[width=.9\textwidth]{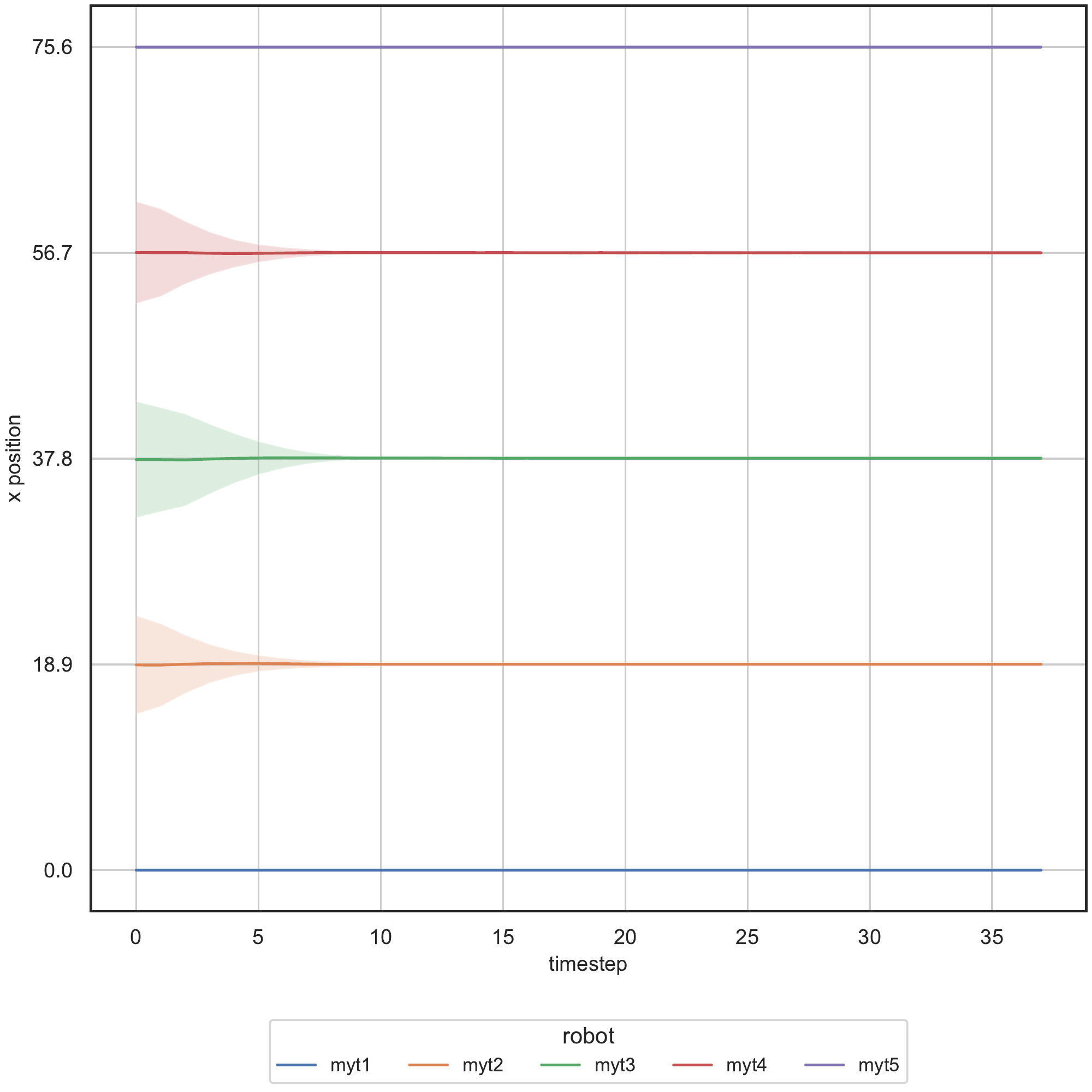}
			\caption{Communication controller trajectories.}
		\end{subfigure}
	\end{center}
	\vspace{-0.5cm}
	\caption[]{Comparison of trajectories, of all the simulation runs, generated 
	using four controllers: the expert, the manual and the two learned from 
	\texttt{net-d1} and \texttt{net-c1} (cont.).}
	\label{fig:net-c1traj}
\end{figure}

To confirm this improvement, we show in Figure \ref{fig:net-c1traj} the 
trajectories obtained employing the four controllers. 
Clearly, the convergence of the robots to the target using the communication is 
much faster that with the distributed controller alone, but it is even better than 
the manual, to such an extent that it can be compared to the expert.

Moreover, analysing the evolution of the control over time, in Figure 
\ref{fig:net-c1control}, we observe that the control learned from the network with 
communication is much more similar to that decided by the expert.
After about $10$ time steps both reach the target and set the speed at $0$, while 
the manual and the distributed need more time.

In Figure \ref{fig:net-c1responseposition} is shown the behaviour, in terms of 
response of the controllers, of a robot located between other two which are 
already in the correct position.
\begin{figure}[H]
	\begin{center}
		\begin{subfigure}[h]{0.35\textwidth}
			\includegraphics[width=\textwidth]{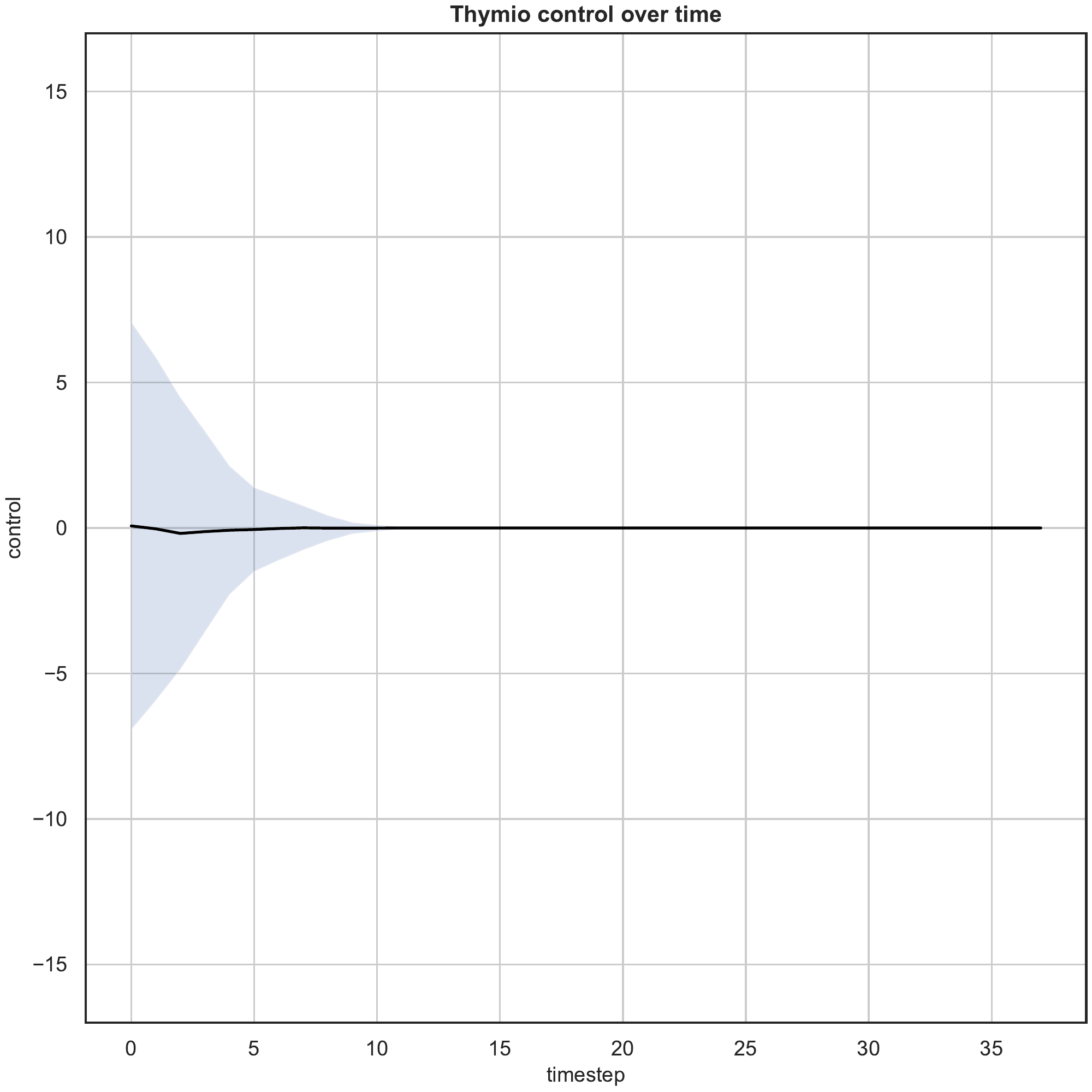}%
			\caption{Expert control.}
		\end{subfigure}
		\hspace{1cm}
		\begin{subfigure}[h]{0.35\textwidth}
			\includegraphics[width=\textwidth]{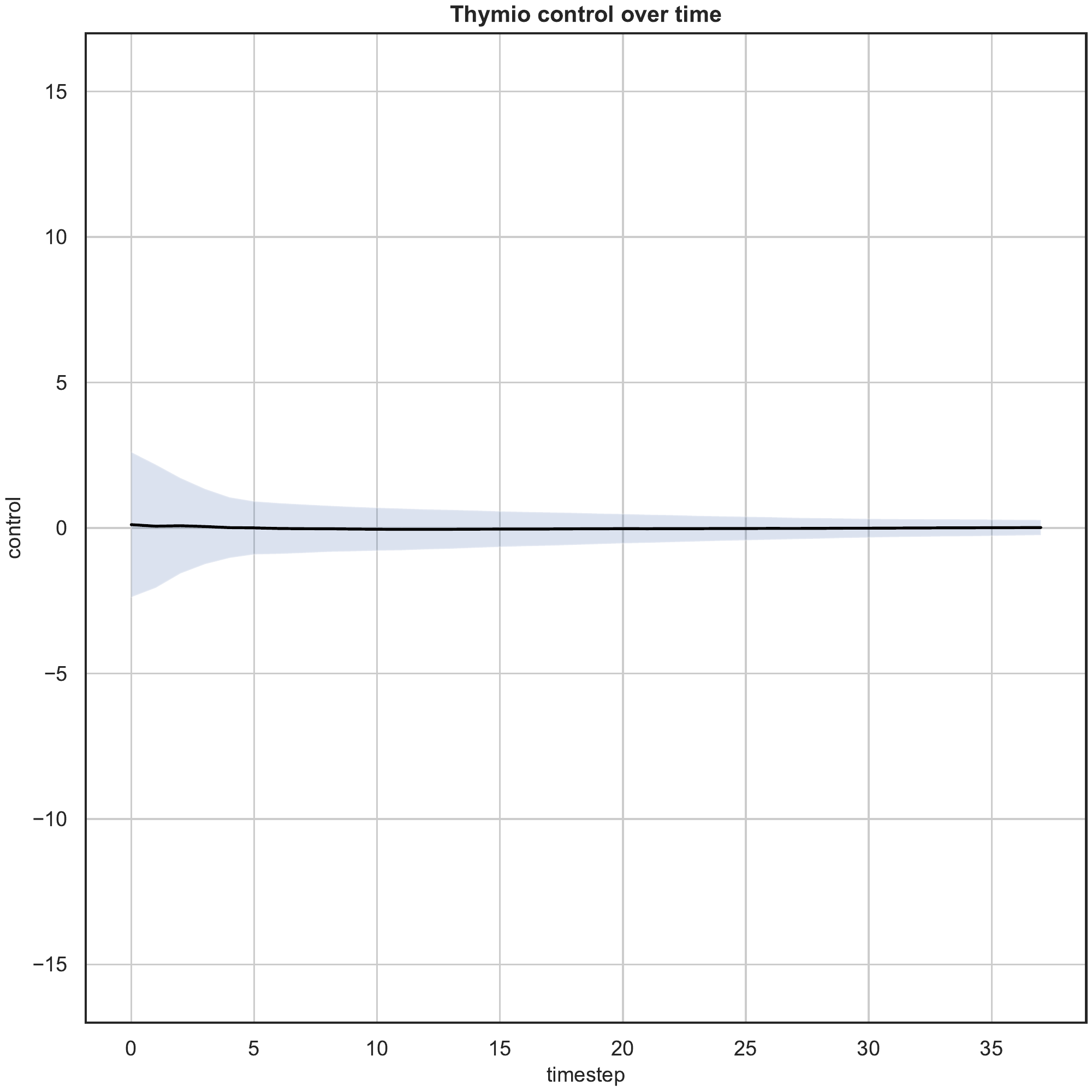}
			\caption{Distributed control.}
		\end{subfigure}
	\end{center}
	\vspace{-0.5cm}
	\caption[Evaluation of the control decided by \texttt{net-c1}.]{Comparison of 
	output control decided using four controllers: the expert, the manual and the 
	two learned from \texttt{net-d1} and \texttt{net-c1}.}
\end{figure}
\begin{figure}[!htb]\ContinuedFloat
	\begin{center}
		\begin{subfigure}[h]{0.35\textwidth}			
			\includegraphics[width=\textwidth]{contents/images/net-d1/control-overtime-manual}%
			\caption{Manual control.}
		\end{subfigure}
		\hspace{1cm}
		\begin{subfigure}[h]{0.35\textwidth}
			\includegraphics[width=\textwidth]{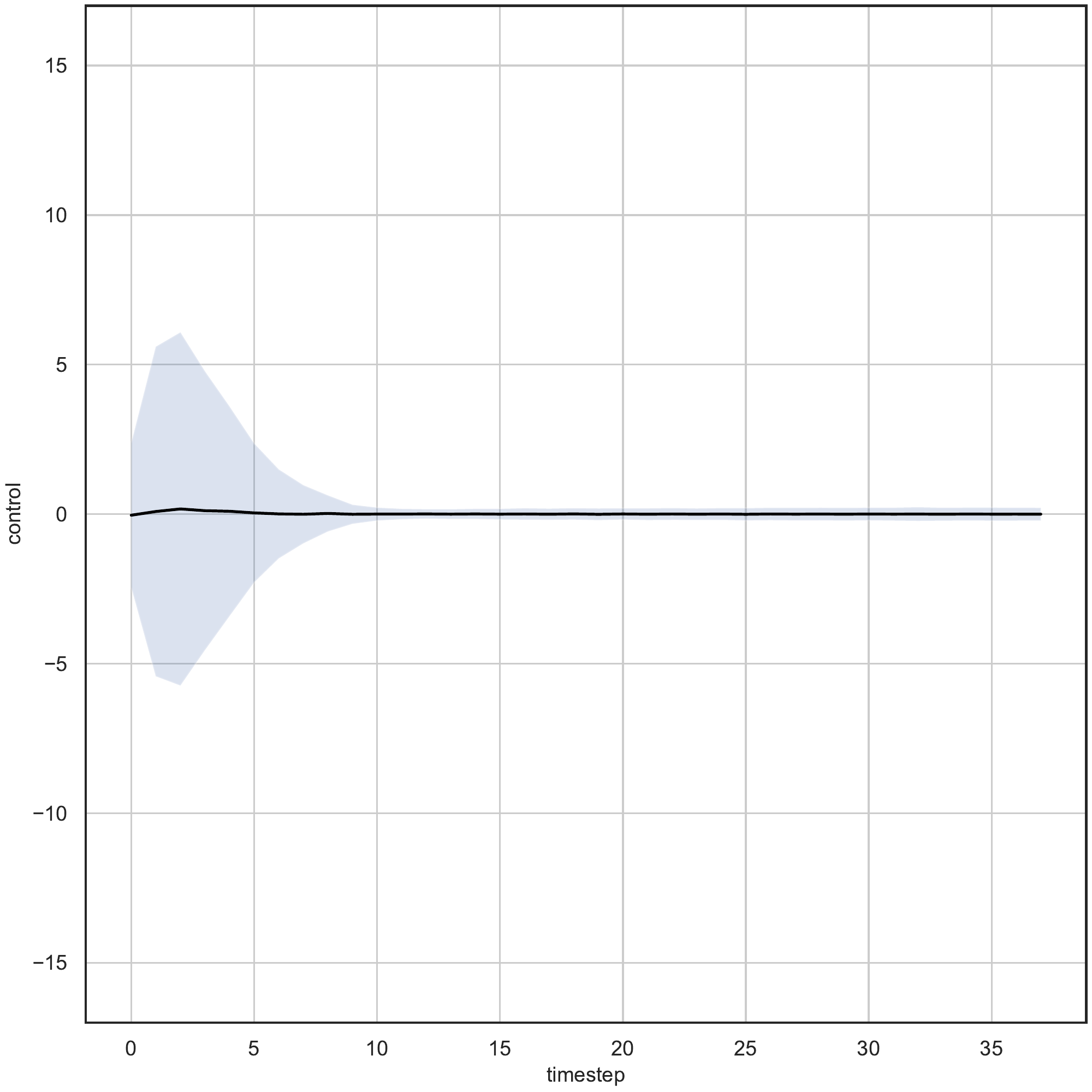}
			\caption{Communication control.}
		\end{subfigure}
	\end{center}
	\vspace{-0.5cm}
	\caption[]{Comparison of 
		output control decided using four controllers: the expert, the manual and the 
		two learned from \texttt{net-d1} and \texttt{net-c1} (cont.).}
	\label{fig:net-c1control}
\end{figure}

\noindent
on the y-axis, by varying the position of the moving robot, on the x-axis.  
As expected, the output is a high value, positive or negative when the robot is 
near to an obstacle on the left or on the right, or close to $0$ when the distance is 
equal on both side.
When the moving robot is located halfway between the two stationary agents, the 
behaviour of the controller with communication is more similar to the one desired.
\begin{figure}[!htb]
	\centering
	\includegraphics[width=.45\textwidth]{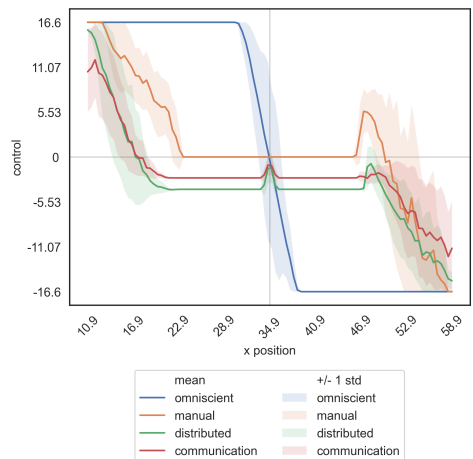}%
	\caption{Response of \texttt{net-c1} by varying the initial position.}
	\label{fig:net-c1responseposition}
\end{figure}

\begin{figure}[!htb]
	\centering
	\includegraphics[width=.65\textwidth]{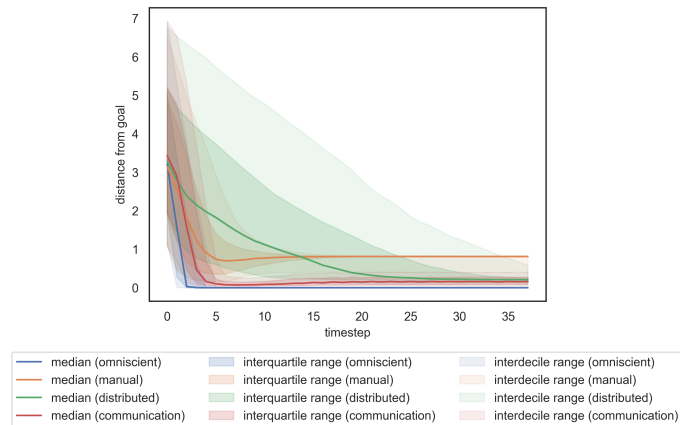}%
	\caption[Evaluation of \texttt{net-c1} distances from goal.]{Comparison of 
		performance in terms of distances from goal obtained using four controllers: 
		the expert, the manual and the two learned from \texttt{net-d1} and 
		\texttt{net-c1}.}
	\label{fig:net-c1distance}
\end{figure}

Finally, in Figure \ref{fig:net-c1distance} are presented the absolute distances of 
each robot from the target, visualised on the y-axis, over time.
On average, the distance from goal of the communication controller is far better  
than that obtained with the distributed and the manual. After about $5$ time 
steps, the robots are in the final configuration, while using the distributed 
alone, $25$ time steps are necessary to get closer to the target. Instead, the 
manual controller does not reach the goal, remaining $1$cm away. 

For this experiment, it seems that the addition of the communication allowed the 
distributed controller to assume an efficient behaviour, for this reason it is 
interesting to investigate how this happens since the communication protocol is 
inferred and learned from the network.
\begin{figure}[!htb]
	\centering
	\includegraphics[width=.6\textwidth]{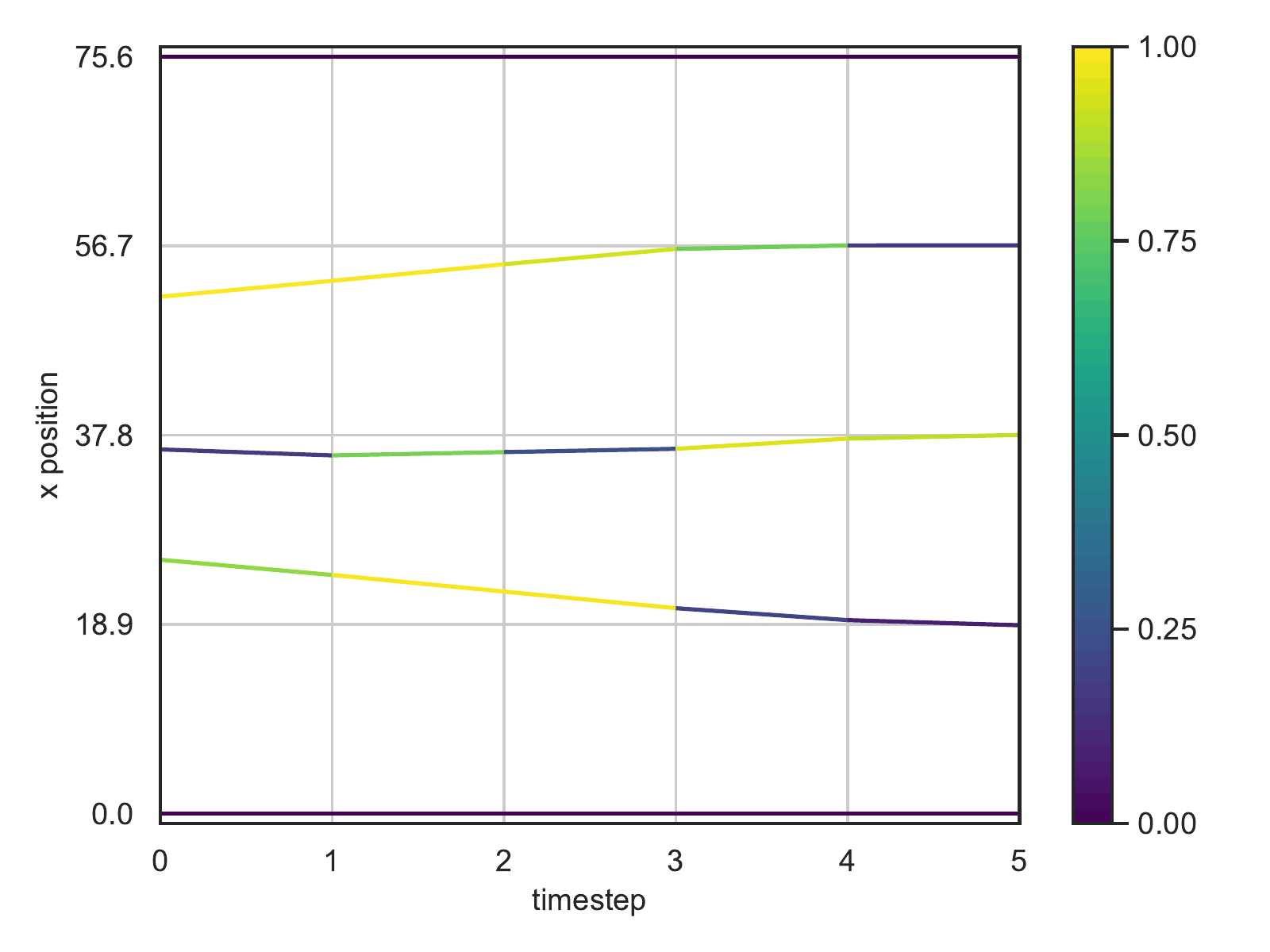}
	\vspace{-0.5cm}
	\caption[Evaluation of the communication learned by 
	\texttt{net-c1}.]{Visualisation of the communication values transmitted by each 
		robot over time using the controller learned from \texttt{net-c1}.}	
	\label{fig:net-c1comm}
\end{figure}
Analysing the graph in Figure \ref{fig:net-c1comm}, which shows the trajectories 
of the agents over time coloured in such a way that the messages transmitted are 
shown through a colour bar whose spectrum is included in the range $[0, 1]$, i.e.,
the maximum and minimum value of communication transmitted. 
We observe that the second and fourth robots in the first time steps transmit 
values close to one, while the central one transmits a value very close to 0. 
The agents begin to move towards the desired position and, as they approach the 
target, the extreme ones transmit decreasing values while the central one 
transmits a higher value. In this case, 5 time steps are sufficient to reach the final 
configuration.
It is difficult, however, only from this image to understand the criteria which the 
network uses to decide the communication.
From a further analysis in fact it would seem that the value predicted by the 
network is not linearly correlated neither to the distance from the goal nor to the 
speed assumed by the agents.

\paragraph*{Results using \texttt{prox\_comm} input}
Following are presented the results of the experiments performed using 
\texttt{prox\_comm} readings. 

In Figure \ref{fig:commlossprox_comm}, are shown the losses by varying the 
average gap, as before the blue, orange and green lines represent respectively 
gaps of $8$, $13$ and $24$cm. From a first observation, the network seems 
to be able to work with all the gaps and the approach with communication 
demonstrate lower loss values than the previous one.
\begin{figure}[!htb]
	\begin{center}
		\begin{subfigure}[h]{0.49\textwidth}
			\centering
			\includegraphics[width=.7\textwidth]{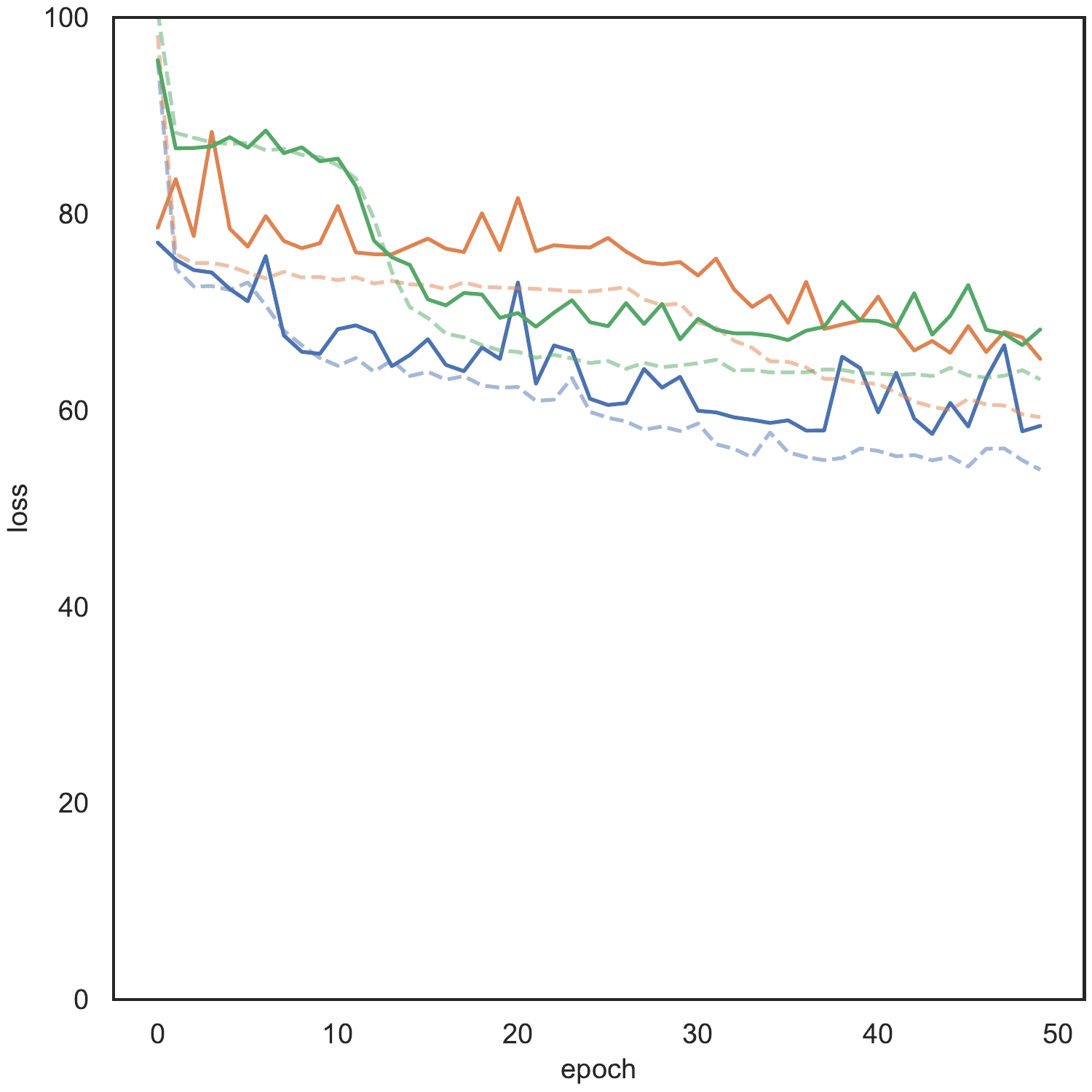}
			\caption{Distributed approach.}
		\end{subfigure}
		\hfill
		\begin{subfigure}[h]{0.49\textwidth}
			\centering
			\includegraphics[width=.7\textwidth]{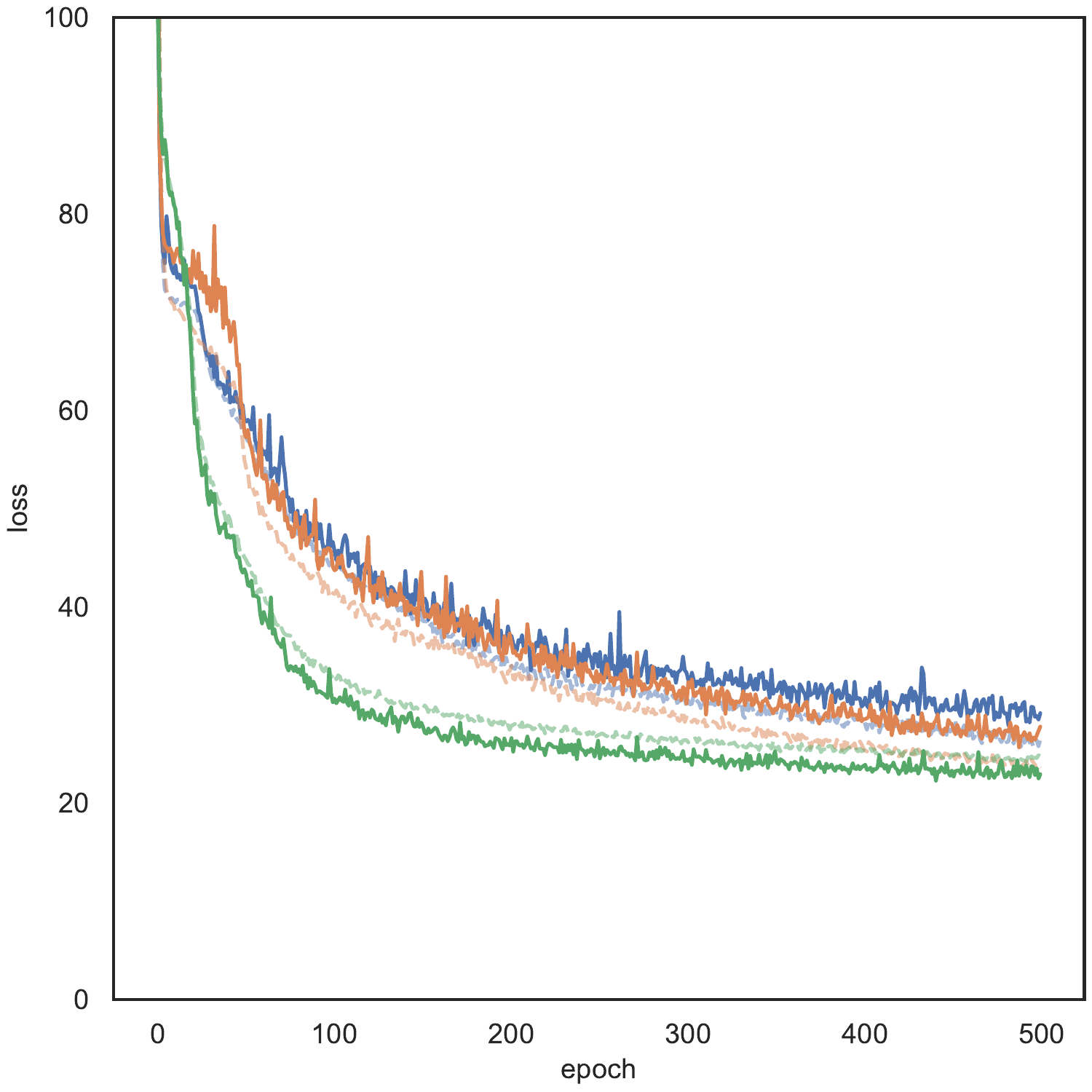}
			\caption{Distributed approach with communication.}
		\end{subfigure}	
	\end{center}
	\vspace{-0.5cm}
	\caption{Comparison of the losses of the models that use \texttt{prox\_comm} 
		readings.}
	\label{fig:commlossprox_comm}
\end{figure}

Focusing on the models trained on the dataset with average gap $24$cm, in 
Figure \ref{fig:net-c6r2} we observe the \ac{r2} of the manual and the learned 
controllers, both the one with
\begin{figure}[H]
	\begin{center}
		\begin{subfigure}[h]{0.49\textwidth}
			\includegraphics[width=\textwidth]{contents/images/net-d6/regression-net-d6-vs-omniscient}%
		\end{subfigure}
		\hfill\vspace{-0.5cm}
		\begin{subfigure}[h]{0.49\textwidth}
			\includegraphics[width=\textwidth]{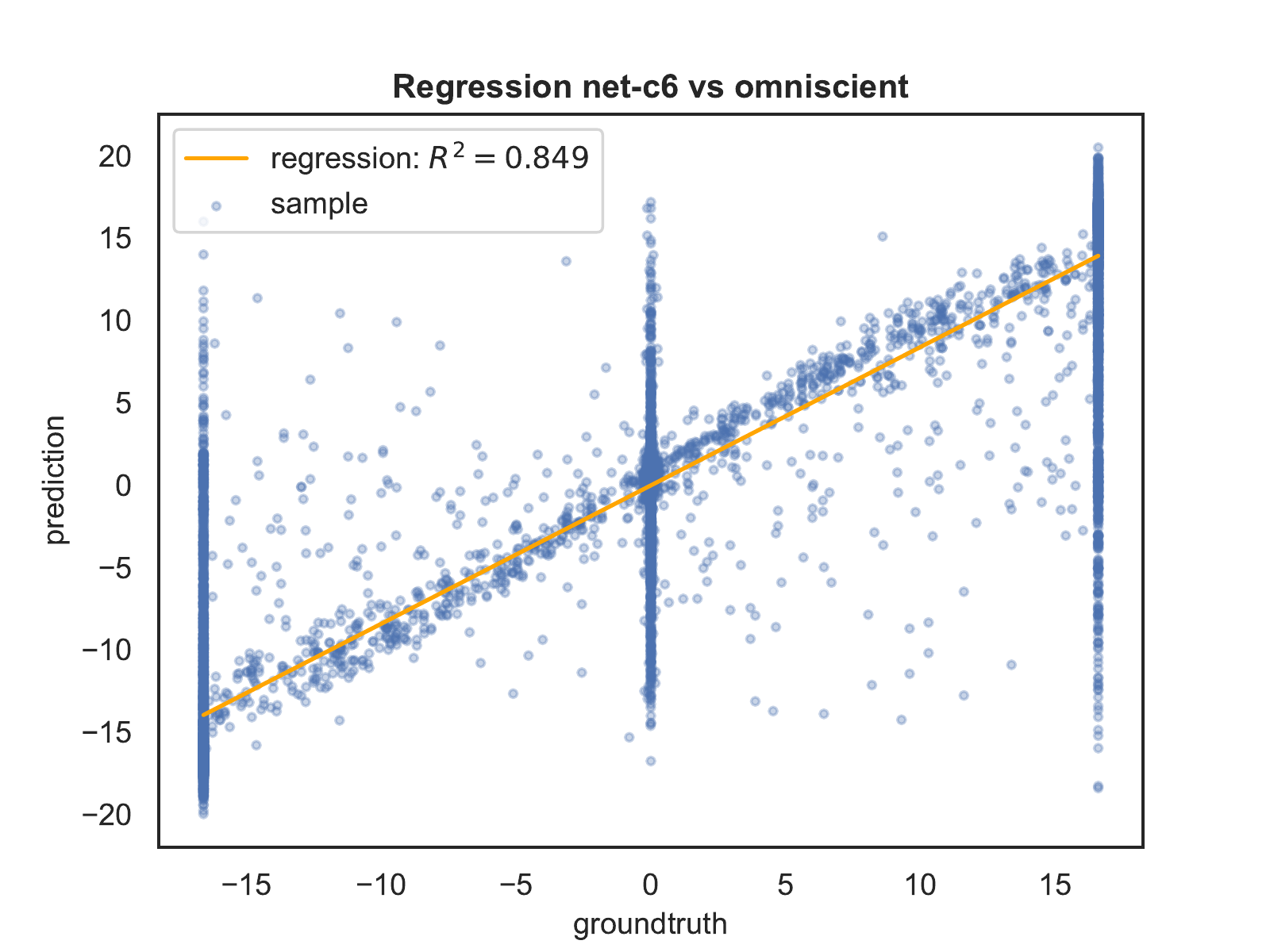}%
		\end{subfigure}
	\end{center}
	\caption[Evaluation of the \ac{r2} coefficients of \texttt{net-c6}.]{Comparison 
		of the \ac{r2} coefficients of the controllers learned from \texttt{net-d6} and 
		\texttt{net-c6}, with respect to the omniscient one.}
	\label{fig:net-c6r2}
\end{figure}

\noindent
and the one without communication, on the validation sets.
Once again, we expect better performance using the new approach than the 
previous, given the fact that the coefficient is increased from $0.55$ up to $0.85$.

In Figure \ref{fig:net-c6traj1} we show a comparison of the trajectories obtained 
for a sample simulation, using the four controllers. We immediately see that the 
omniscient controller and the two \texttt{net-d6} and \texttt{net-c6} are the 
fastest and allow agents to always reach the target. 
Instead, those moved using the manual controller did not approach the goal, even 
if they try to position themselves at equal distances.
\begin{figure}[!htb]
	\centering
	\includegraphics[width=.6\textwidth]{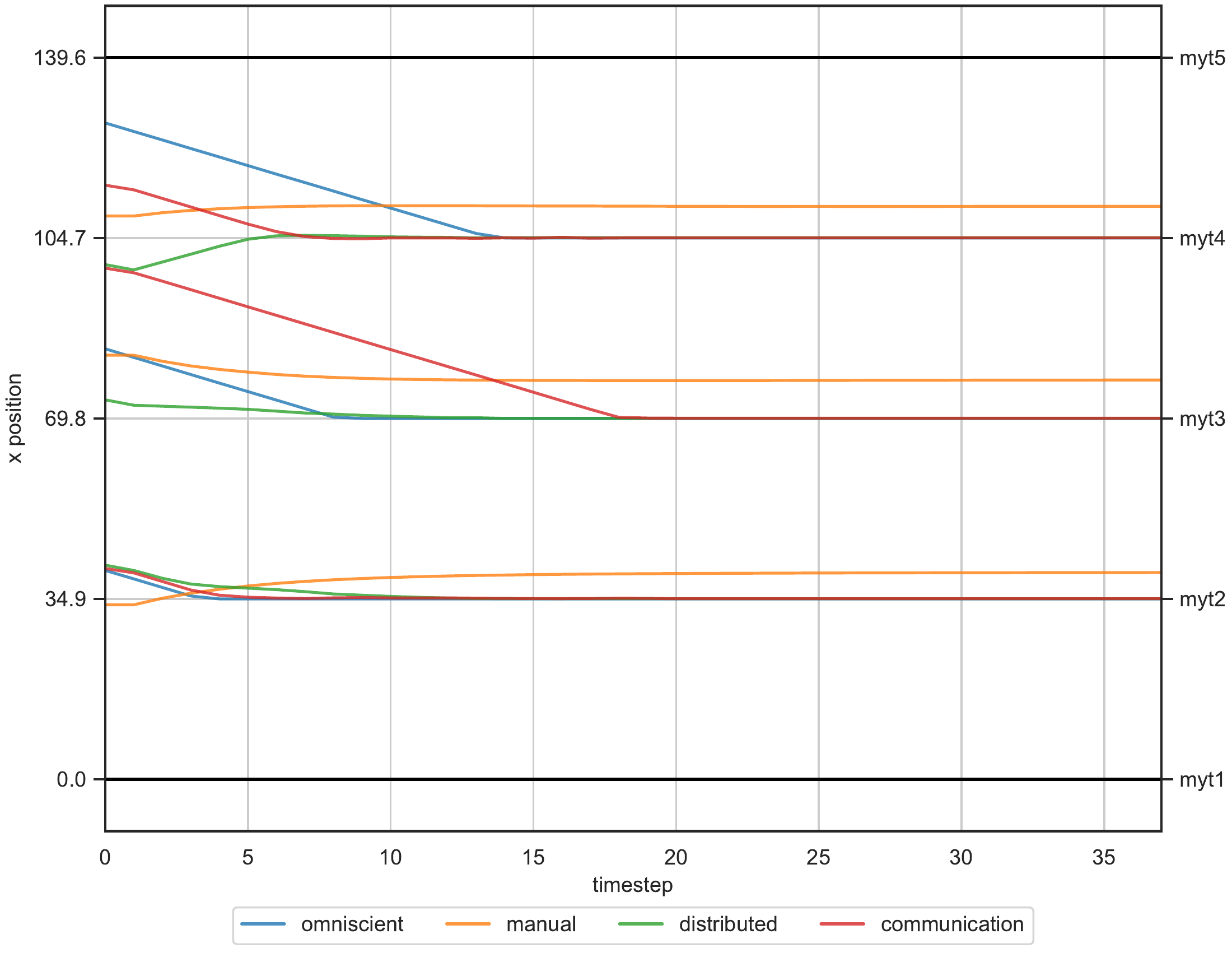}%
	\caption[Evaluation of the trajectories obtained with \texttt{prox\_comm} 
	input.]{Comparison of trajectories, of a single simulation, generated using four 
	controllers: the expert, the manual and the two learned from \texttt{net-d6} 
	and \texttt{net-c6}.}
	\label{fig:net-c6traj1}
\end{figure}  

\begin{figure}[H]
	\begin{center}
		\begin{subfigure}[h]{0.49\textwidth}
			\centering
			\includegraphics[width=.9\textwidth]{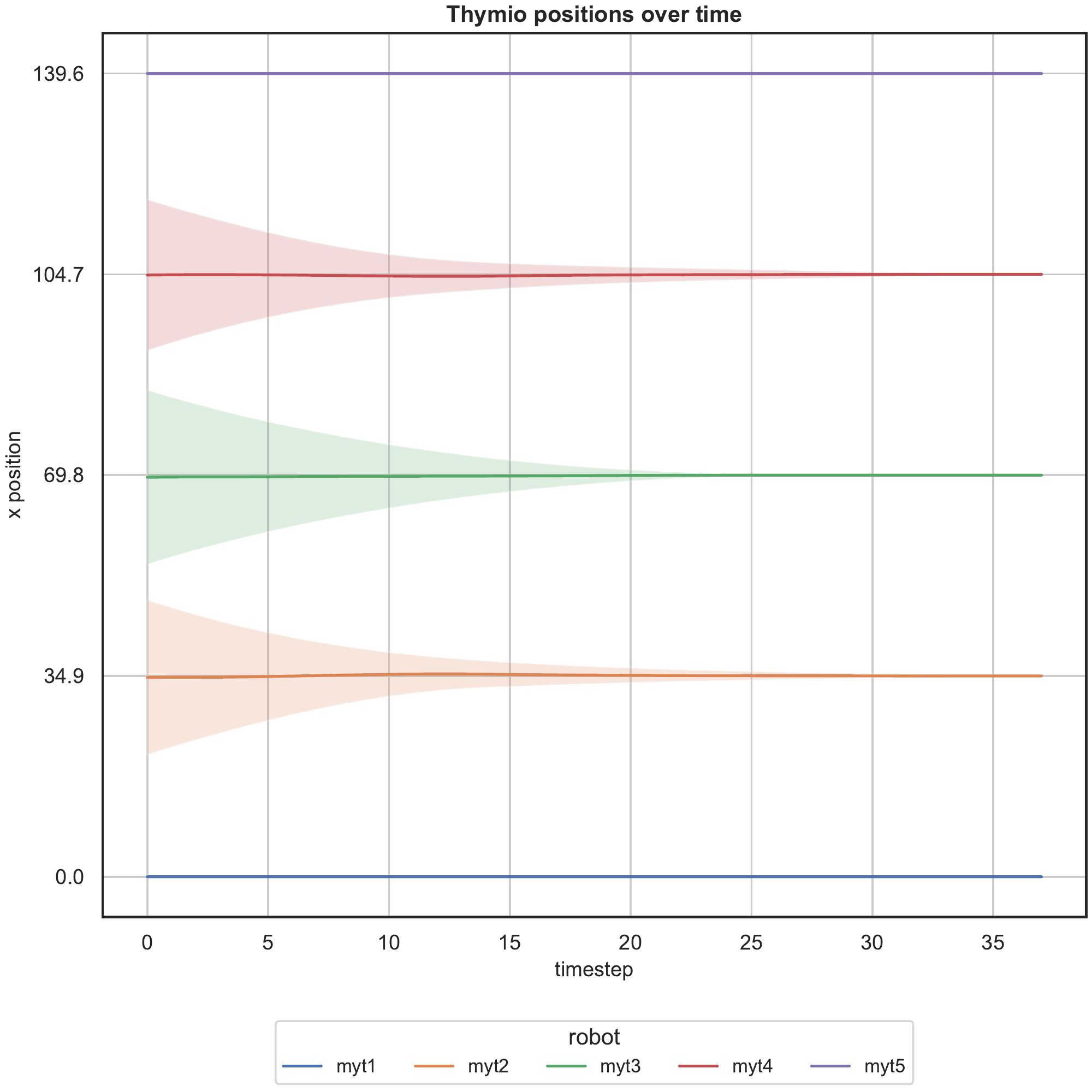}%
			\caption{Expert controller trajectories.}
		\end{subfigure}
		\hfill
		\begin{subfigure}[h]{0.49\textwidth}
			\centering
			\includegraphics[width=.9\textwidth]{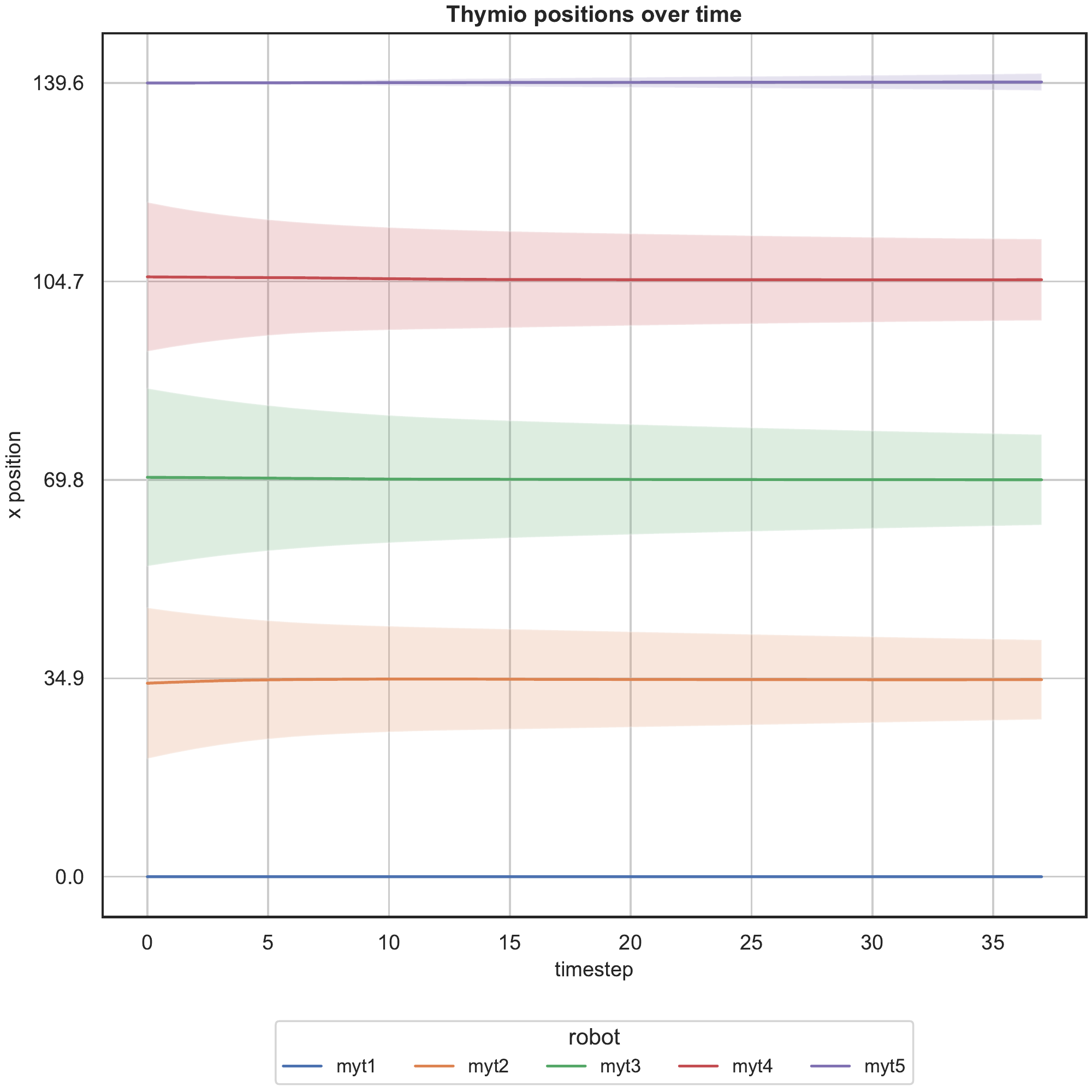}
			\caption{Distributed controller trajectories.}
		\end{subfigure}
	\end{center}
	\vspace{-0.5cm}
	\caption[Evaluation of the trajectories learned by \texttt{net-c6}.]{Comparison 
	of trajectories, of all the simulation runs, generated using four controllers: the 
	expert, the manual and the two learned from \texttt{net-d6} and 
	\texttt{net-c6}.}
\end{figure}

\begin{figure}[!htb]\ContinuedFloat
	\begin{center}
		\begin{subfigure}[h]{0.49\textwidth}
			\centering			
			\includegraphics[width=.9\textwidth]{contents/images/net-d6/position-overtime-manual}%
			\caption{Manual controller trajectories.}
		\end{subfigure}
		\hfill
		\begin{subfigure}[h]{0.49\textwidth}
			\centering
			\includegraphics[width=.9\textwidth]{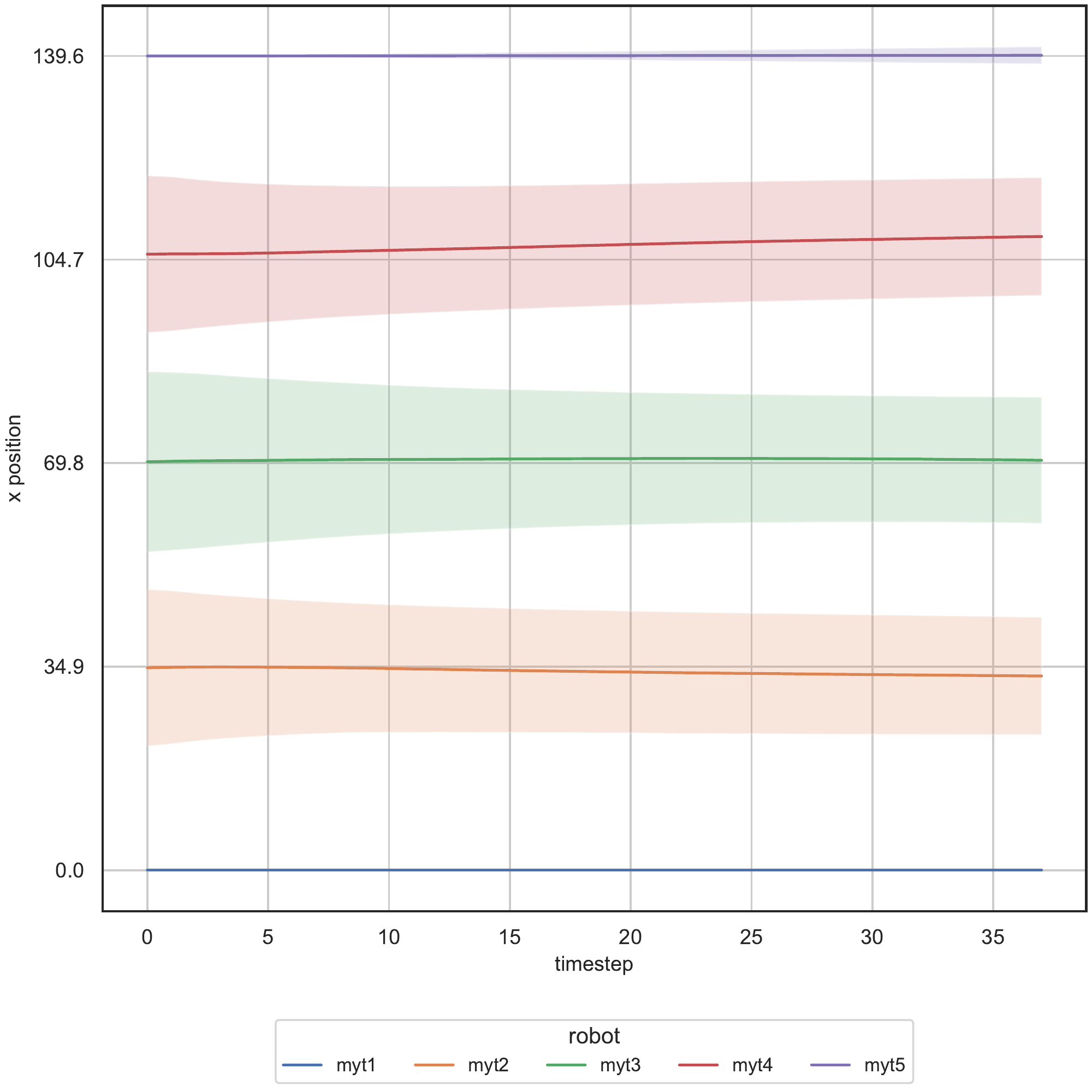}
			\caption{Communication controller trajectories.}
		\end{subfigure}
	\end{center}
	\vspace{-0.5cm}
	\caption[]{Comparison of trajectories, of all the simulation runs, generated 
	using four controllers: the expert, the manual and the two learned from 
	\texttt{net-d6} and \texttt{net-c6} (cont.).}
	\label{fig:net-c6traj}
\end{figure}

In Figure \ref{fig:net-c6traj} are shown the trajectories obtained employing the 
four controllers. Even for the expert, the convergence to the target is slower than 
before, since the distance between the robots is greater, but it is still much faster 
than with the other two controllers. As we said for Figure \ref{fig:net-d6traj}, the 
manual controller has serious problems in reaching the goal. The two learned 
controllers can still try to approach the desired position employing more time 
steps.

In addition, the analysis of the evolution of the control over time in Figure 
\ref{fig:net-c6control} suggests that the reason for this behaviour is the fact the 
both the learned controllers use a lower speed than expected.
\begin{figure}[!htb]
	\begin{center}
		\begin{subfigure}[h]{0.35\textwidth}
			\includegraphics[width=\textwidth]{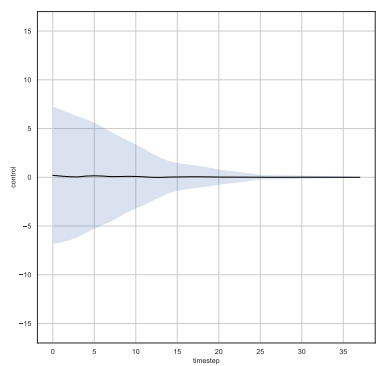}%
			\caption{Expert control.}
		\end{subfigure}
		\hspace{1cm}
		\begin{subfigure}[h]{0.35\textwidth}
			\includegraphics[width=\textwidth]{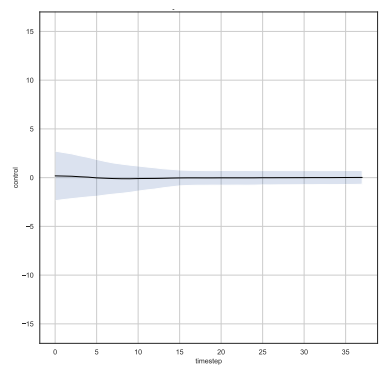}
			\caption{Distributed control.}
		\end{subfigure}
	\end{center}
	\caption[Evaluation of the control decided by \texttt{net-c6}.]{Comparison of 
	output control decided using four controllers: the expert, the manual and the 
	two learned from \texttt{net-d6} and \texttt{net-c6}.}
\end{figure}

\begin{figure}[!htb]\ContinuedFloat
	\begin{center}
		\begin{subfigure}[h]{0.35\textwidth}			
			\includegraphics[width=\textwidth]{contents/images/net-d6/control-overtime-manual}%
			\caption{Manual control.}
		\end{subfigure}
		\hspace{1cm}
		\begin{subfigure}[h]{0.35\textwidth}
			\includegraphics[width=\textwidth]{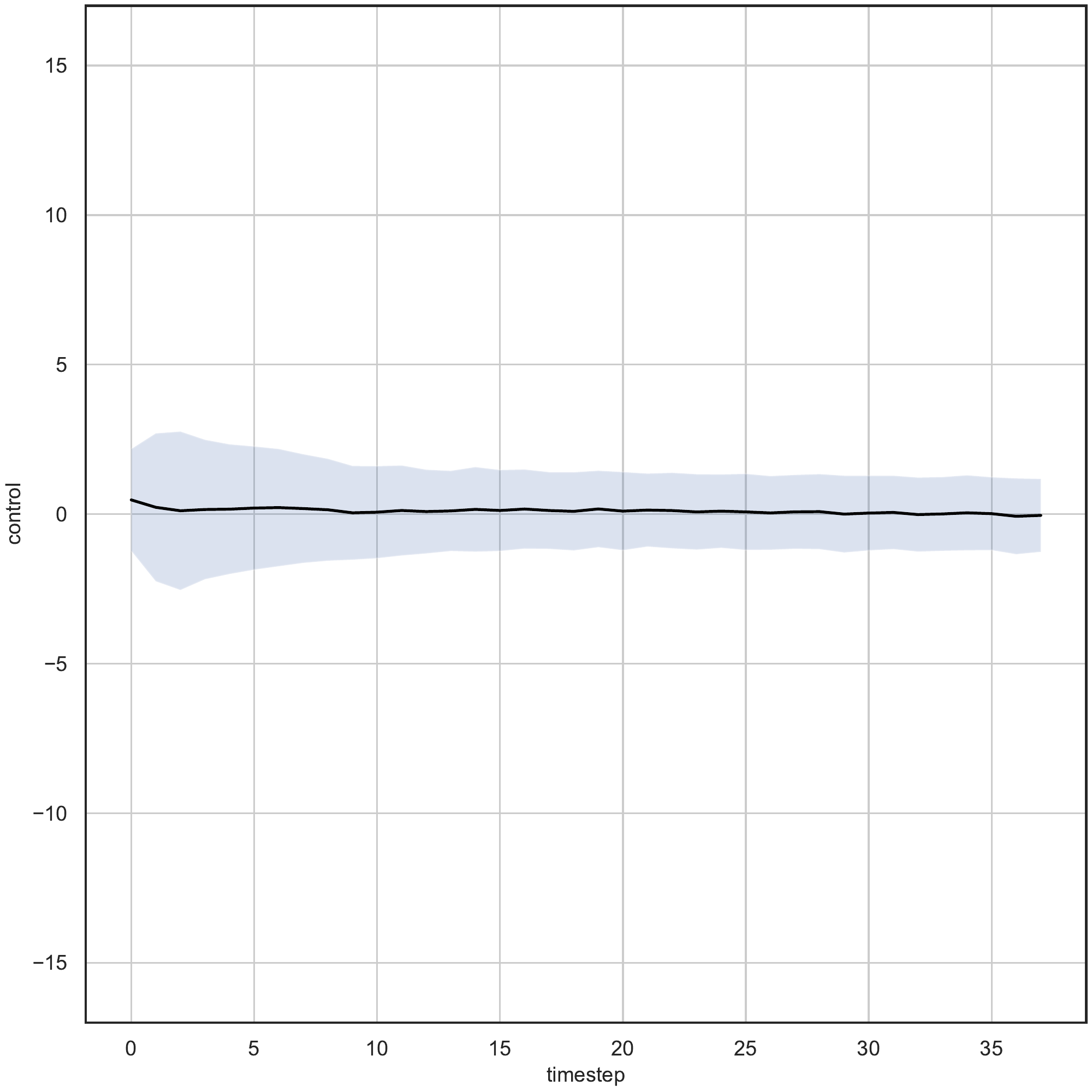}
			\caption{Communication control.}
		\end{subfigure}
	\end{center}
	\caption[]{Comparison of output control decided using four controllers: the 
	expert, the manual and the two learned from \texttt{net-d6} and 
	\texttt{net-c6} (cont).}
	\label{fig:net-c6control}
\end{figure}

In Figure \ref{fig:net-c6responseposition} is displayed the behaviour of a robot 
located between other two stationary agents, showing the response of the 
controllers, on the y-axis, by varying the position of the moving robot, on the 
x-axis.  
As expected, the output is a high positive value when the robot is close to an 
obstacle on the left, negative when there is an obstacle in front and not behind, 
and $0$ when the distance from right and left is equal.
\begin{figure}[!htb]
	\centering
	\includegraphics[width=.45\textwidth]{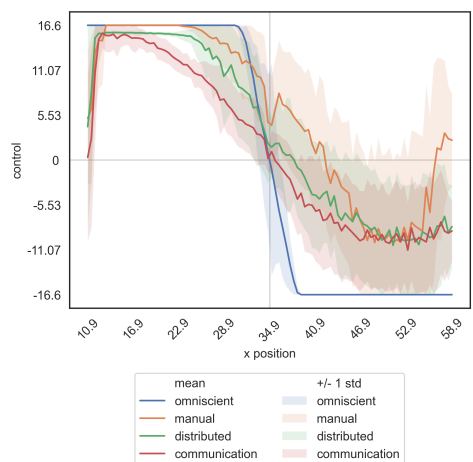}%
	\caption{Response of \texttt{net-c6} by varying the initial position.}
	\label{fig:net-c6responseposition}
\end{figure}

\noindent
The behaviour of the controller that uses the communication is most accurate 
when the moving robot is halfway between the two stationary.

Finally, in Figure \ref{fig:net-c6distance} is presented a metric that measures the 
absolute distance of each robot from the target over time.
Unlike the non-optimal performances obtained with the manual and distributed 
controllers, in which in both cases the robots in the final configuration are 
located on average at about $5$cm from the target, the distance from goal 
of the communication controller is far better. In just $11$ time steps, $4$ more 
than the expert, the agents reach the goal position.
\begin{figure}[!htb]
	\centering
	\includegraphics[width=.65\textwidth]{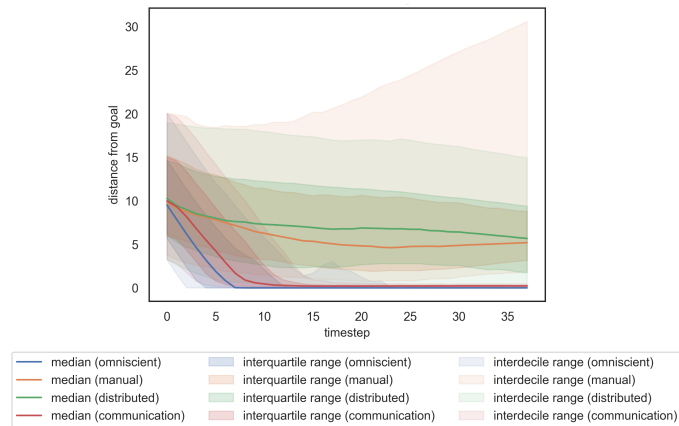}%
	\caption[Evaluation of \texttt{net-c6} distances from goal.]{Comparison of 
		performance in terms of distances from goal obtained using four controllers: 
		the expert, the manual and the two learned from \texttt{net-d6} and 
		\texttt{net-c6}.}
	\label{fig:net-c6distance}
\end{figure}

As before, it is very difficult to analyse the communication transmitted over time 
through the visualisation in Figure \ref{fig:net-c6comm}. In this case, the agents 
seem to communicate initially a high value, except the central one that is sending 
values in the range $[0.6, 1]$.
\begin{figure}[H]
	\centering
	\includegraphics[width=.6\textwidth]{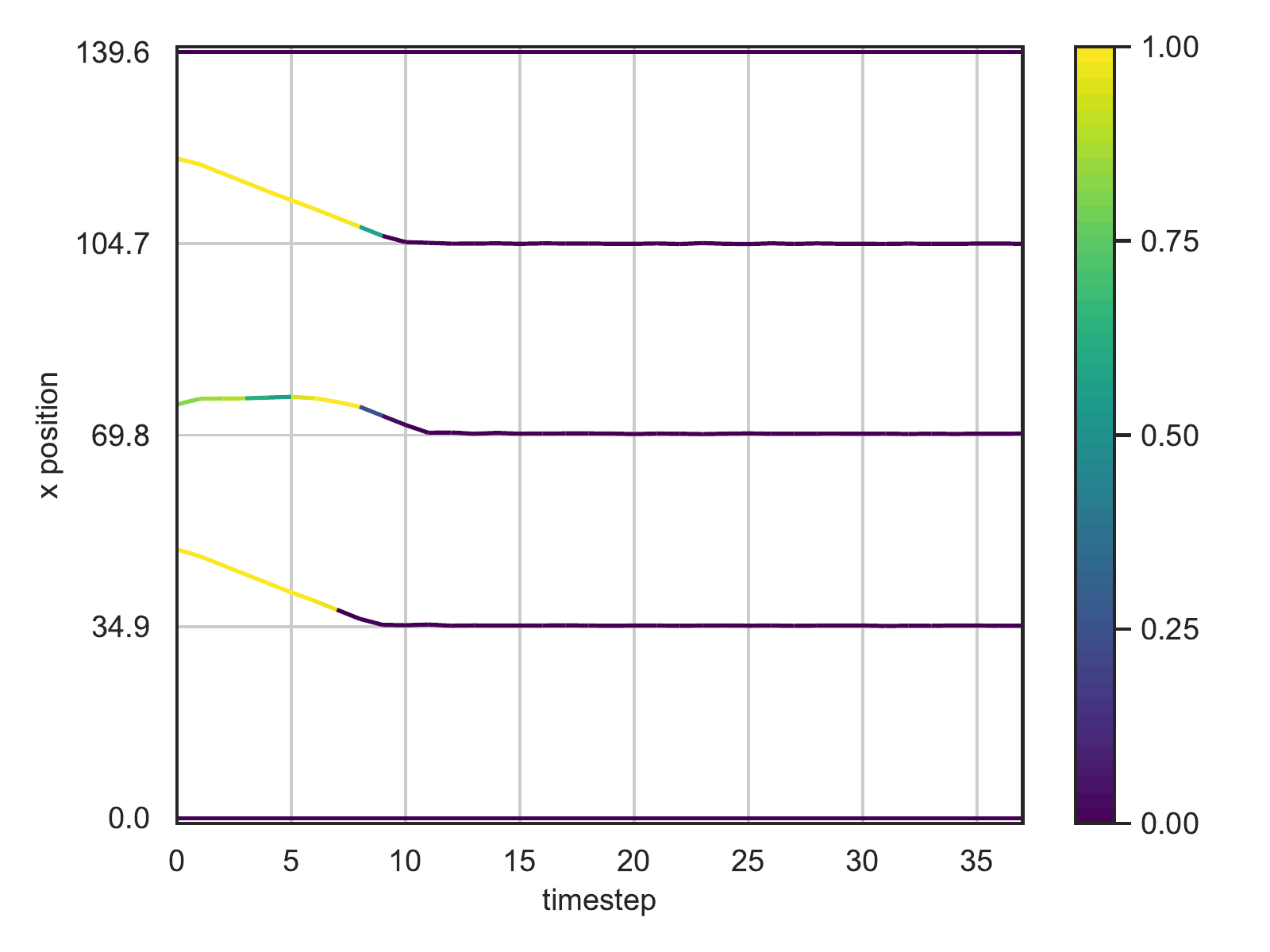}
	\vspace{-0.5cm}
	\caption[Evaluation of the communication learned by 
	\texttt{net-c6}.]{Visualisation of the communication values transmitted by each 
		robot over time using the controller learned from \texttt{net-c6}.}	
	\label{fig:net-c6comm}
\end{figure}

\noindent
 In all the cases, these values decrease as the robots approach the target position.

\paragraph*{Results using \texttt{all\_sensors} input}
We conclude the first group of experiments presenting the results obtained using 
both types of input together from which we expect a more stable and robust 
behaviour.

In Figure \ref{fig:commlossall_sensors} are summarised the performance, 
in terms of loss, of the models trained using \texttt{all\_sensors} input and 
different gaps: the blue, orange and green lines represent respectively average 
gaps of $8$, $13$ and $24$cm.
It is immediately evident that using the new approach the trend of the curves are 
very similar to each other, and in general the loss is considerably decreased down 
to $20$.
\begin{figure}[!htb]
	\begin{center}
			\begin{subfigure}[h]{0.49\textwidth}
			\centering
			\includegraphics[width=.7\textwidth]{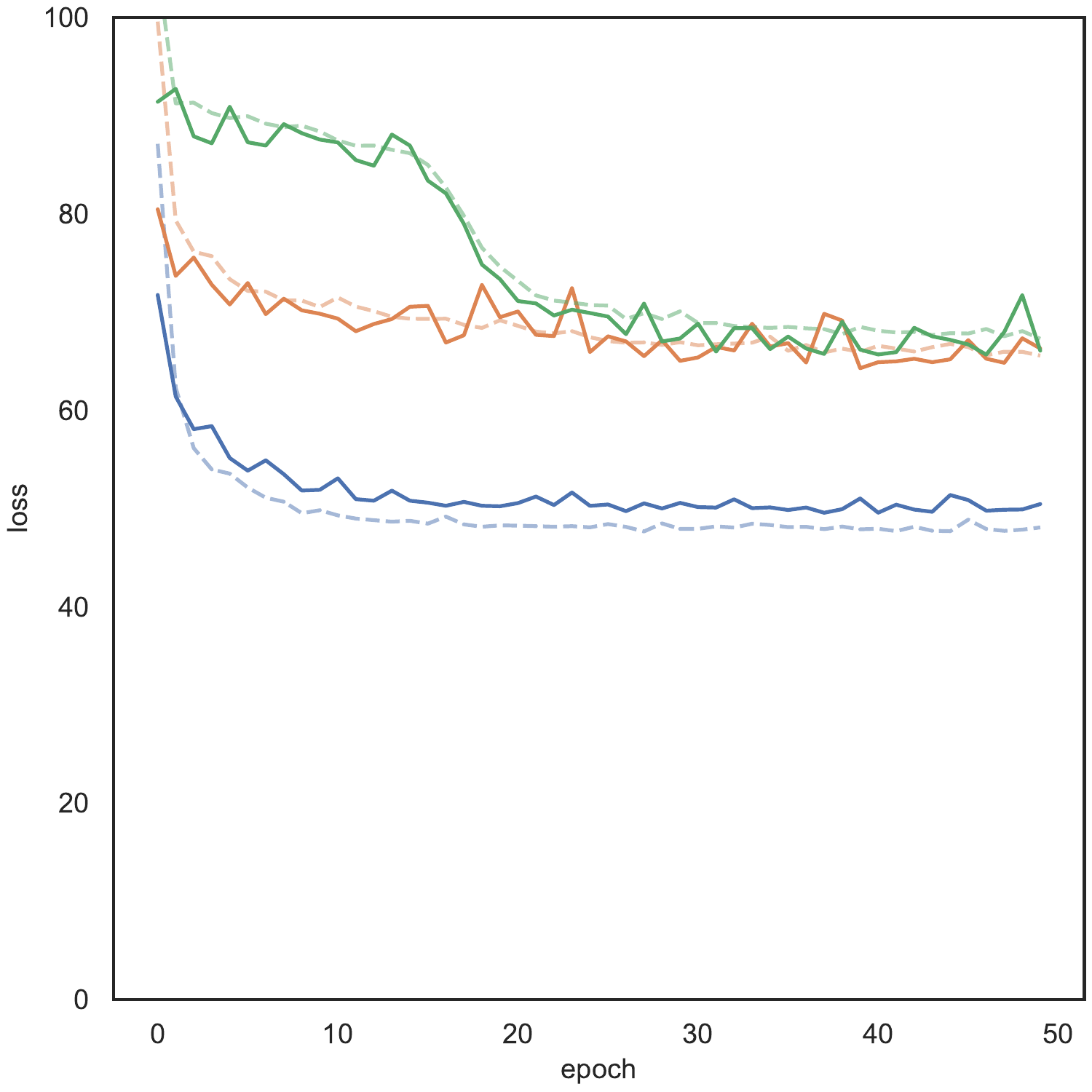}
			\caption{Distributed approach.}
		\end{subfigure}
		\hfill
		\begin{subfigure}[h]{0.49\textwidth}
			\centering
			\includegraphics[width=.7\textwidth]{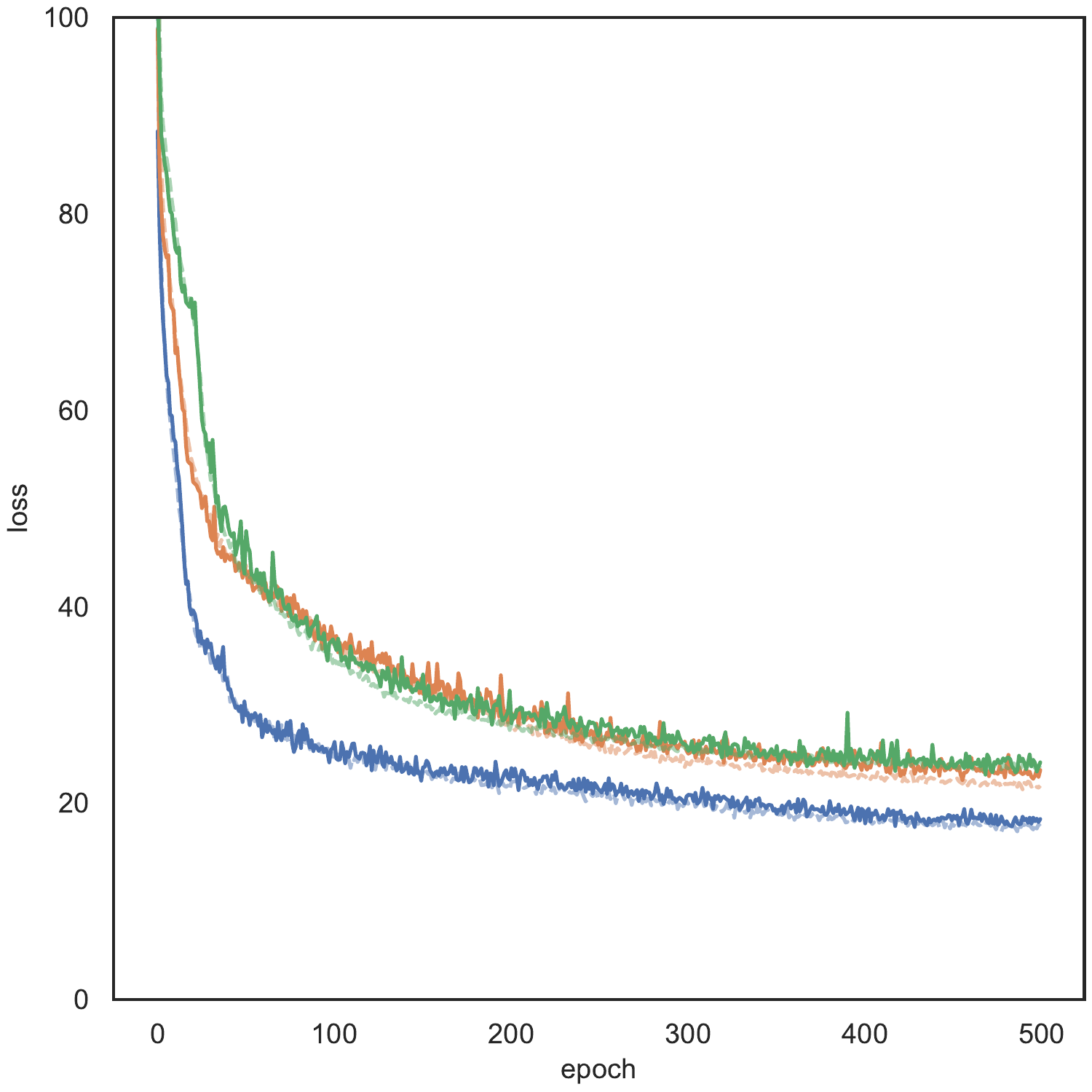}
			\caption{Distributed approach with communication.}
		\end{subfigure}	
	\end{center}
	\vspace{-0.5cm}
	\caption{Comparison of the losses of the models that use \texttt{all\_sensors} 
		readings.}
	\label{fig:commlossall_sensors}
\end{figure}

Considering, as before, the more complex case, for instance the one with the 
greatest average gap, in Figure \ref{fig:net-c9r2} are visualised the \ac{r2} of the 
manual and the learned controllers, with and without communication.
The superiority of the new approach is once again confirmed by the increase in 
the coefficient \ac{r2} from $0.56$ to $0.84$. 
\begin{figure}[!htb]
	\begin{center}
		\begin{subfigure}[h]{0.49\textwidth}
			\includegraphics[width=\textwidth]{contents/images/net-d9/regression-net-d9-vs-omniscient}%
		\end{subfigure}
		\hfill\vspace{-0.5cm}
		\begin{subfigure}[h]{0.49\textwidth}
			\includegraphics[width=\textwidth]{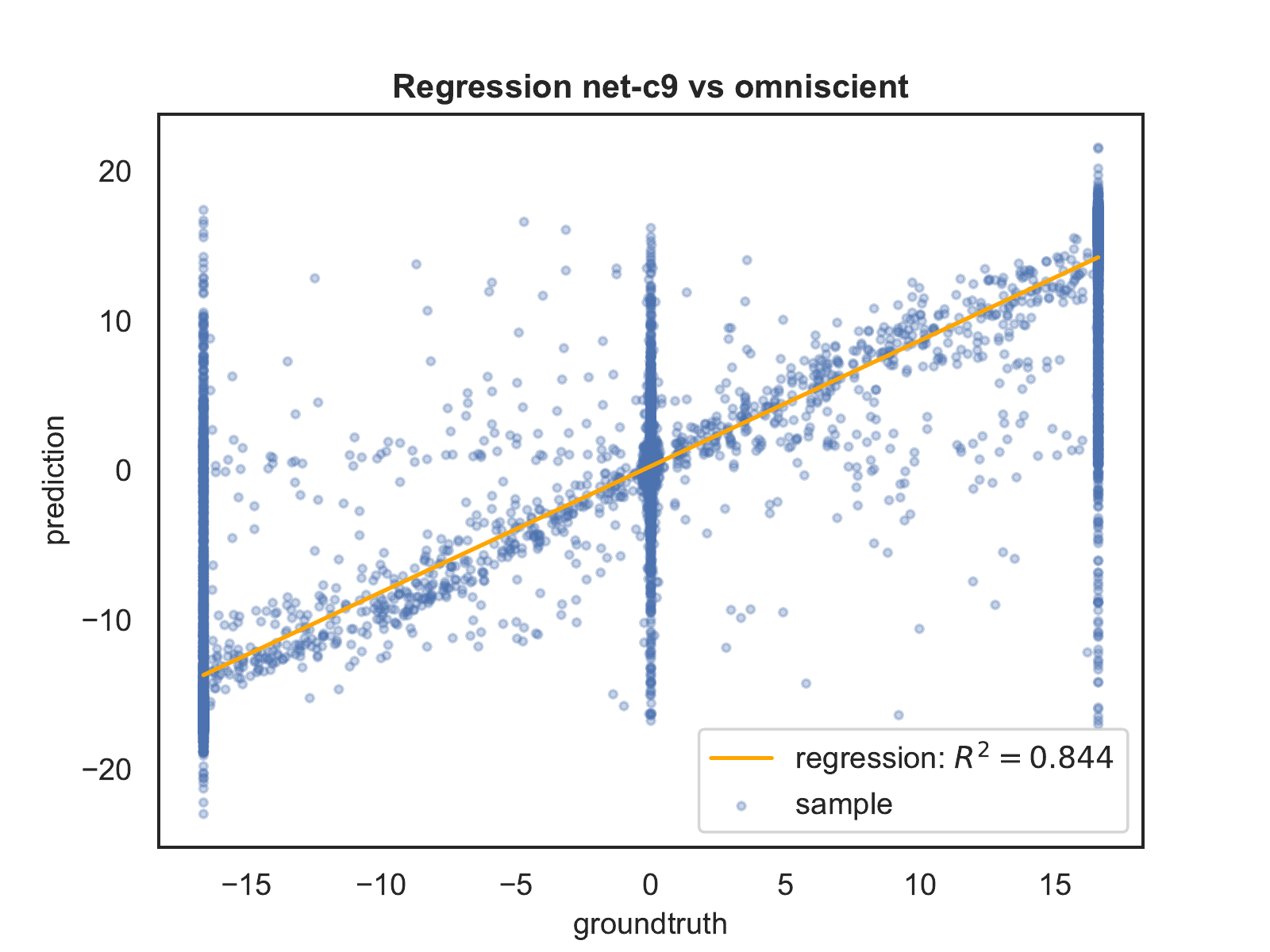}%
		\end{subfigure}
	\end{center}
	\caption[Evaluation of the \ac{r2} coefficients of \texttt{net-c9}.]{Comparison 
		of the \ac{r2} coefficients of the controllers learned from 
		\texttt{net-d9} and \texttt{net-c9}, with respect to the omniscient one.}
	\label{fig:net-c9r2}
\end{figure}

In Figure \ref{fig:net-c9traj1} is shown a comparison of the trajectories obtained 
for a sample simulation.
As before, the performance obtained using the omniscient and the learned 
controllers are comparable: they all are very fast and reach the target in less than 
10 time steps. 
\begin{figure}[!htb]
	\centering
	\includegraphics[width=.65\textwidth]{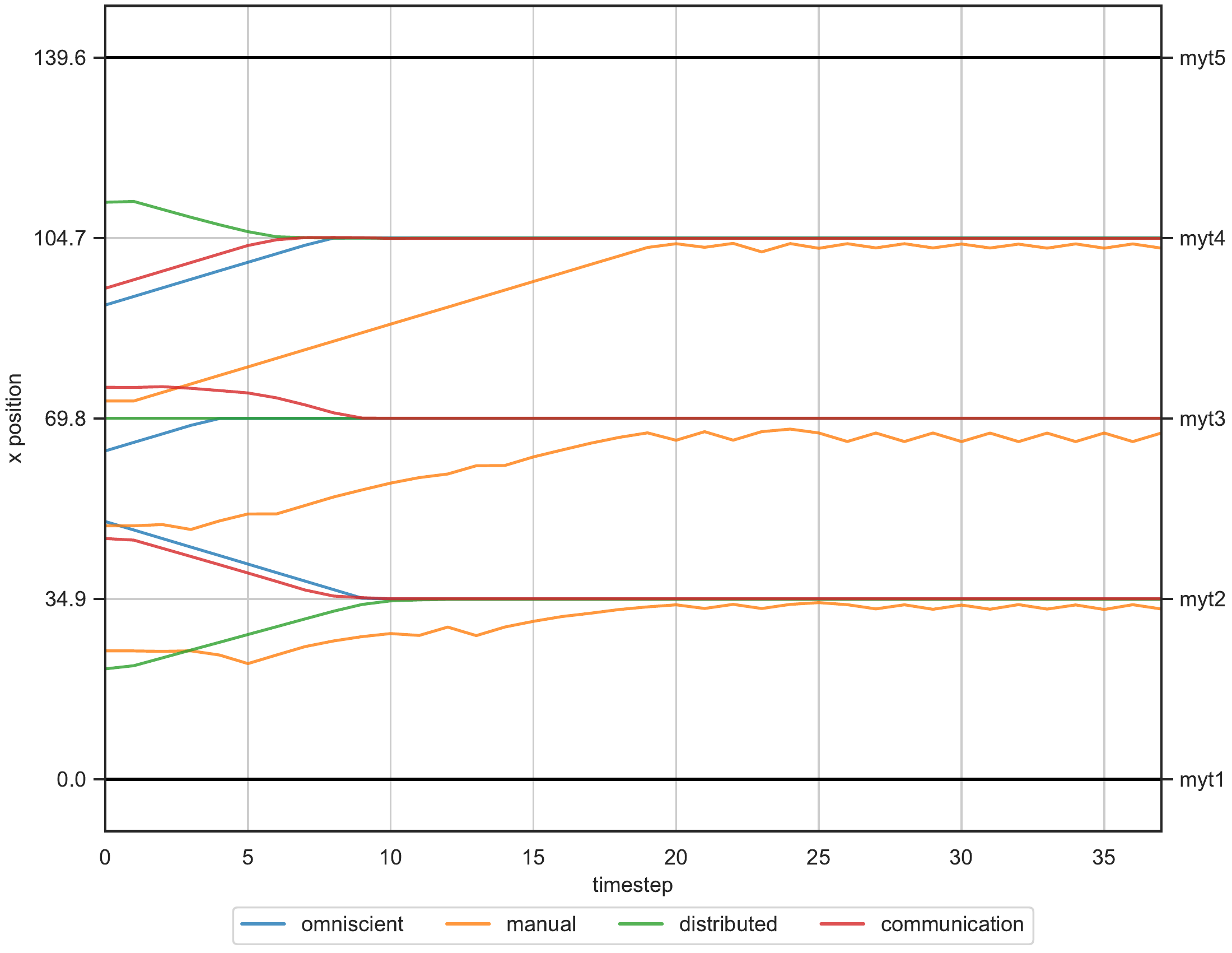}%
	\caption[Evaluation of the trajectories obtained with \texttt{all\_sensors} 
	input.]{Comparison of trajectories, of a single simulation, generated using four 
		controllers: the expert, the manual and the two learned from \texttt{net-d9} 
		and \texttt{net-c9}.}
	\label{fig:net-c9traj1}
\end{figure}

This improvement is further supported by the trajectories shown in Figure 
\ref{fig:net-c9traj}. 
The convergence to the target is still slow due to the high distance between the 
robots, but this time the controller with communication is much faster than the 
distributed controller but also than the manual. The bands of the deviation show 
that this approach can reach the same results of the expert.

Moreover, from the evolution of the control over time in Figure 
\ref{fig:net-c9control}, we observe that the control learned from the 
communication network is very similar to the expert.
\begin{figure}[H]
	\begin{center}
		\begin{subfigure}[h]{0.49\textwidth}
			\centering
			\includegraphics[width=.9\textwidth]{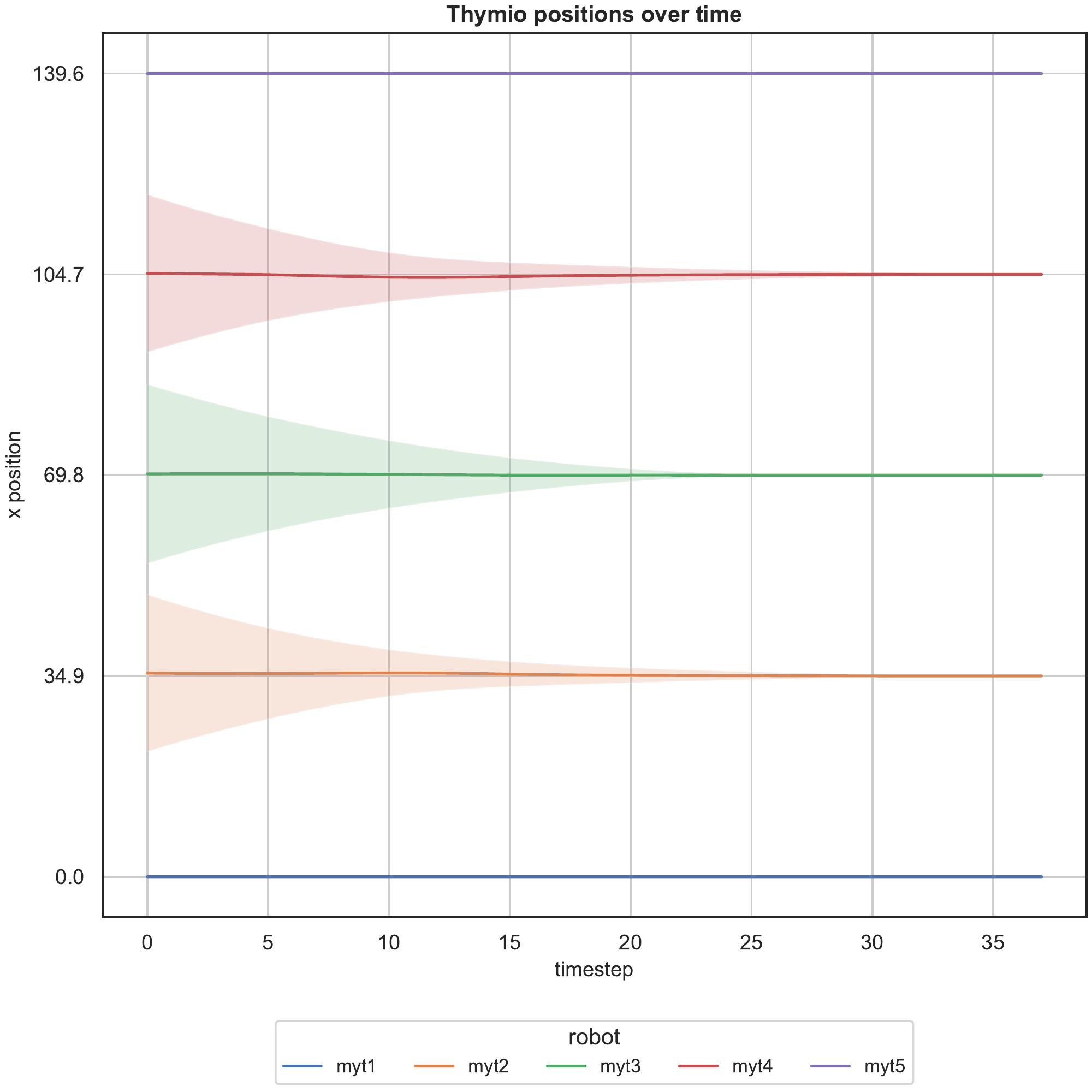}%
			\caption{Expert controller trajectories.}
		\end{subfigure}
		\hfill
		\begin{subfigure}[h]{0.49\textwidth}
			\centering
			\includegraphics[width=.9\textwidth]{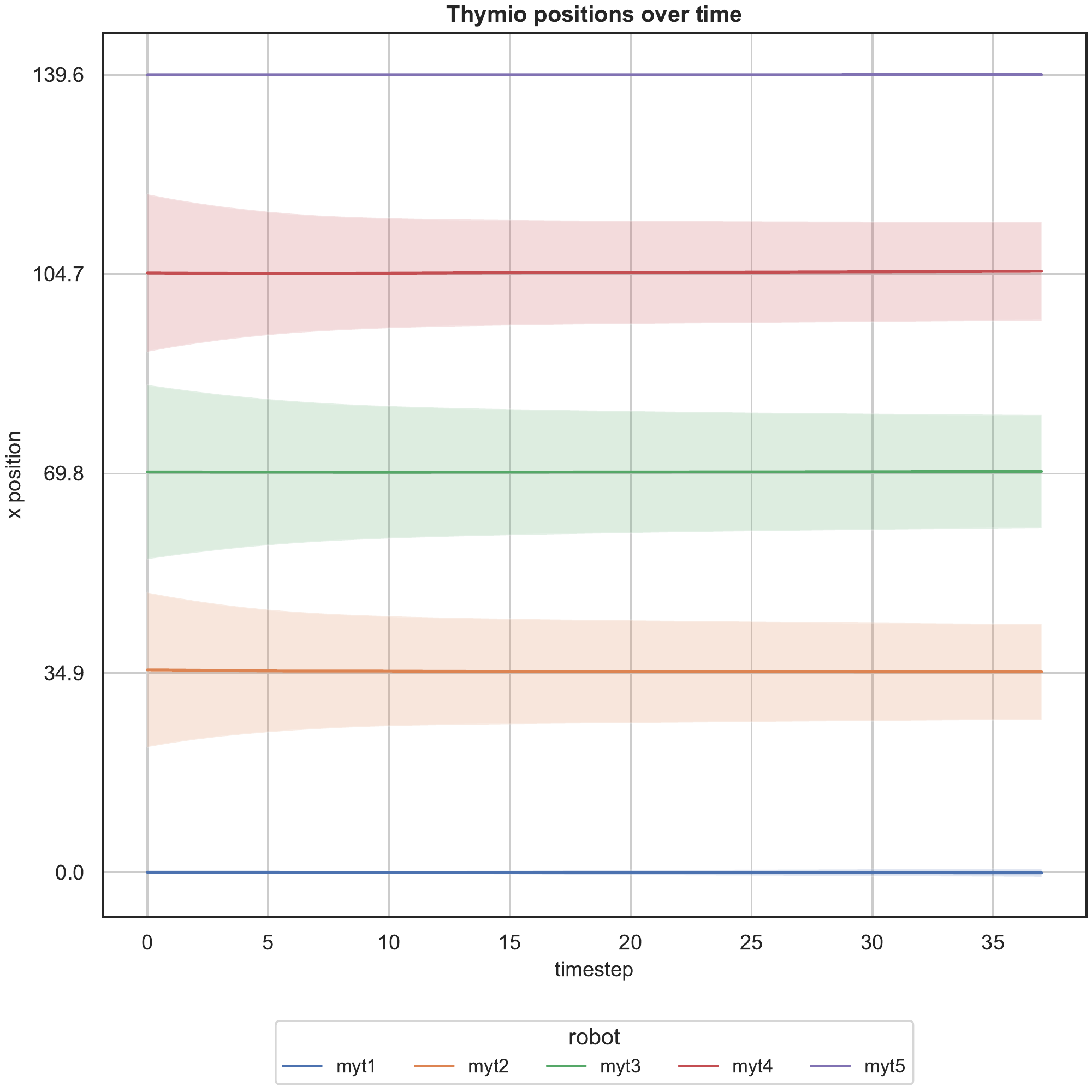}
			\caption{Distributed controller trajectories.}
		\end{subfigure}
	\end{center}
	\vspace{-0.5cm}
	\caption[Evaluation of the trajectories learned by \texttt{net-c9}.]{Comparison 
	of trajectories, of all the simulation runs, generated using four controllers: the 
	expert, the manual and the two learned from \texttt{net-d9} and 
	\texttt{net-c9}.}
\end{figure}

\begin{figure}[!htb]\ContinuedFloat
	\begin{center}
		\begin{subfigure}[h]{0.49\textwidth}
			\centering			
			\includegraphics[width=.9\textwidth]{contents/images/net-d9/position-overtime-manual}%
			\caption{Manual controller trajectories.}
		\end{subfigure}
		\hfill
		\begin{subfigure}[h]{0.49\textwidth}
			\centering
			\includegraphics[width=.9\textwidth]{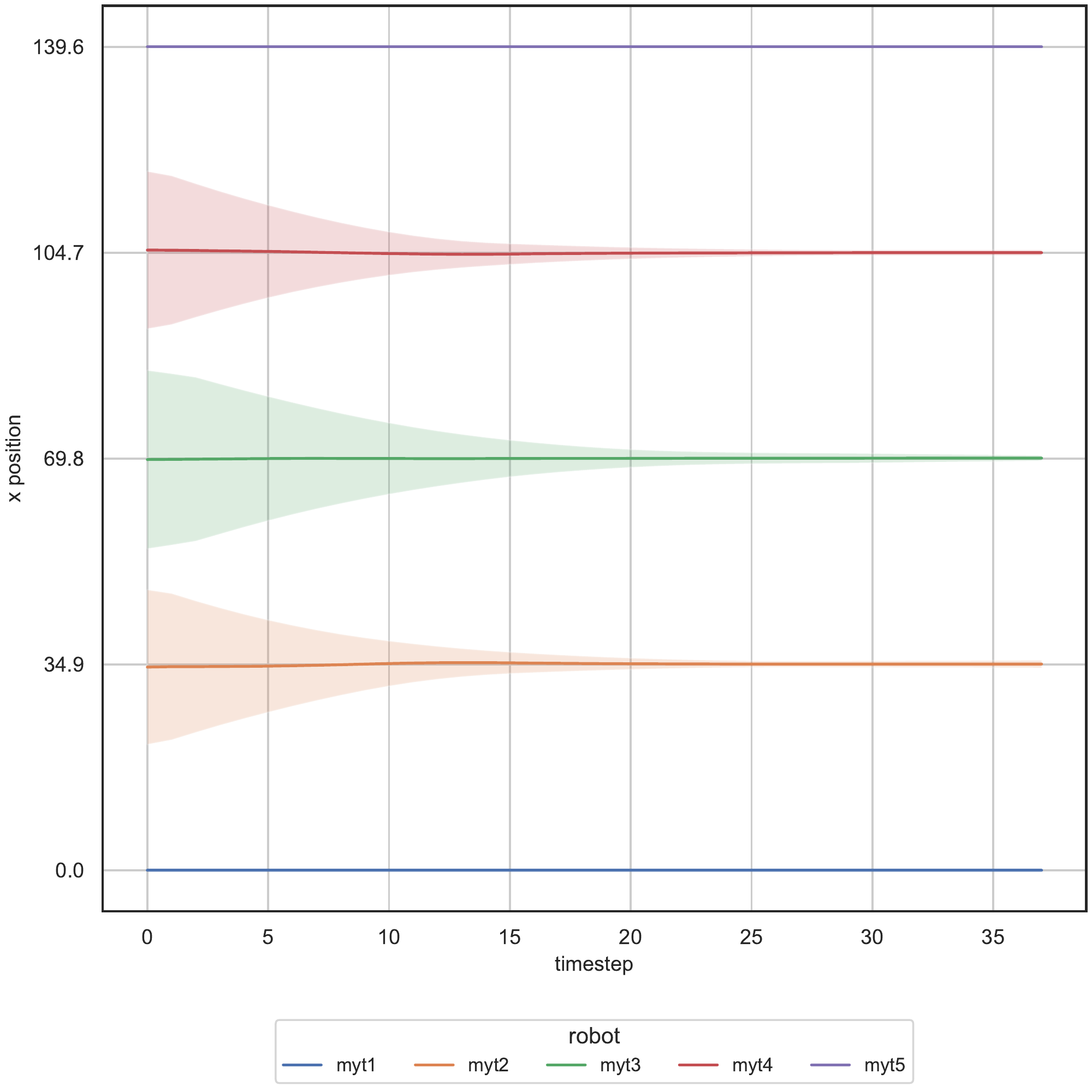}
			\caption{Communication controller trajectories.}
		\end{subfigure}
	\end{center}
\vspace{-0.5cm}
	\caption[]{Comparison of trajectories, of all the simulation runs, generated 
	using four controllers: the expert, the manual and the two learned from 
	\texttt{net-d9} and \texttt{net-c9} (cont.).}
	\label{fig:net-c9traj}
\end{figure}

\begin{figure}[H]
	\begin{center}
		\begin{subfigure}[h]{0.33\textwidth}
			\includegraphics[width=\textwidth]{contents/images/net-d9/control-overtime-omniscient}%
			\caption{Expert control.}
		\end{subfigure}
		\hspace{1cm}
		\begin{subfigure}[h]{0.33\textwidth}
			\includegraphics[width=\textwidth]{contents/images/net-d9/control-overtime-learned_distributed}
			\caption{Distributed control.}
		\end{subfigure}
	\end{center}
	\begin{center}
		\begin{subfigure}[h]{0.33\textwidth}			
			\includegraphics[width=\textwidth]{contents/images/net-d9/control-overtime-manual}%
			\caption{Manual control.}
		\end{subfigure}
		\hspace{1cm}
		\begin{subfigure}[h]{0.33\textwidth}
			\includegraphics[width=\textwidth]{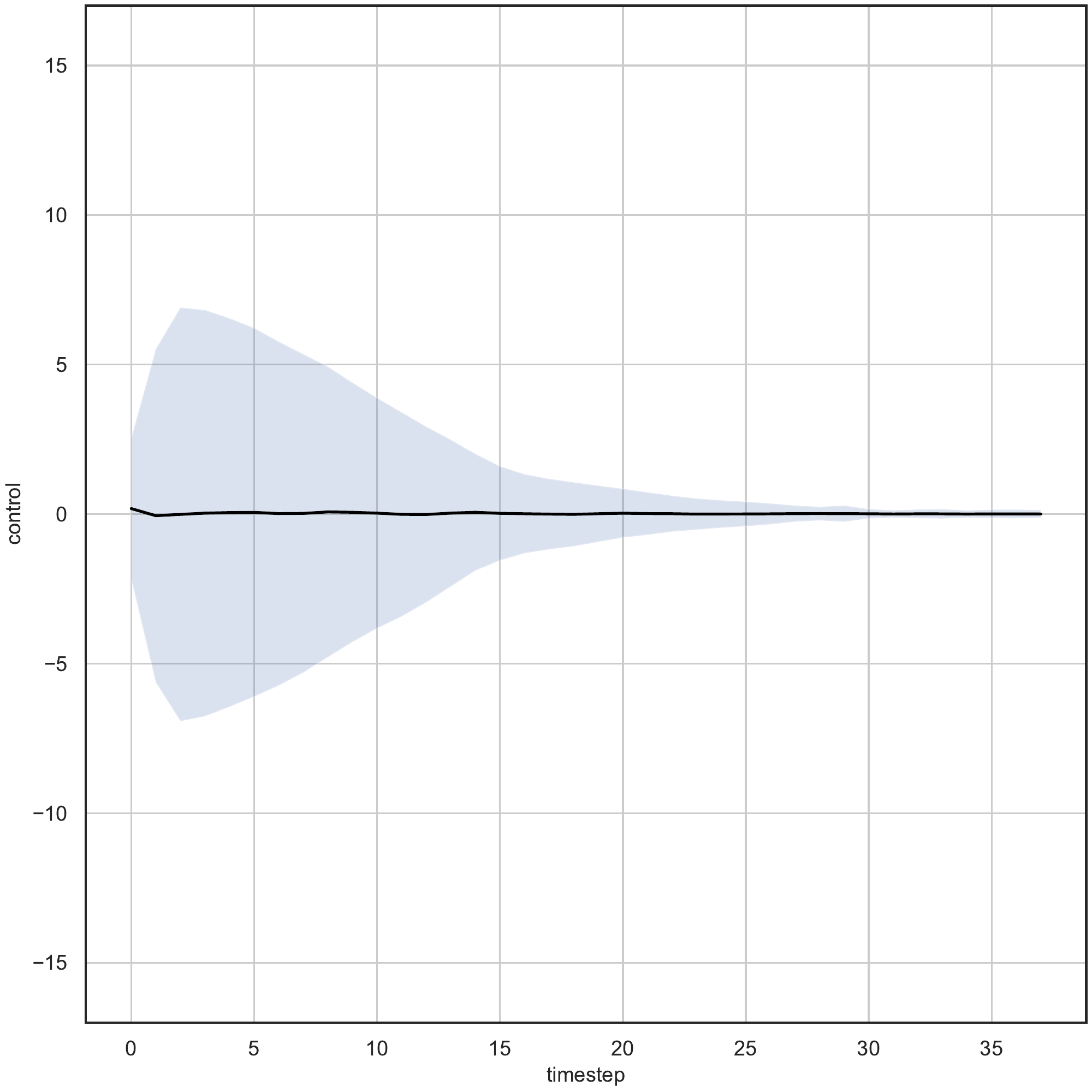}
			\caption{Communication control.}
		\end{subfigure}
	\end{center}
	\vspace{-0.5cm}
	\caption[Evaluation of the control decided by \texttt{net-c9}.]{Comparison of 
	output control decided using four controllers: the expert, the manual and the 
	two learned from \texttt{net-d9} and \texttt{net-c9}.}
	\label{fig:net-c9control}
\end{figure}

In Figure \ref{fig:net-c9responseposition} is displayed the behaviour of a robot 
located between other two stationary agents which are already in the correct 
position, 
showing the response of the controllers, on the y-axis, by varying the position of 
the moving robot, visualised on the x-axis. As expected, the output is a high 
value, positive or negative respectively when the robot is close to an obstacle on 
the left or on the right, or it is close to $0$ when the distance from right and left 
is equal.
\begin{figure}[!htb]
	\centering
	\includegraphics[width=.5\textwidth]{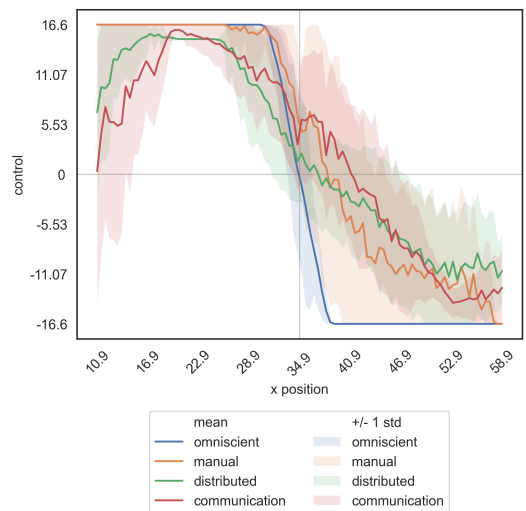}%
	\caption{Response of \texttt{net-c9} by varying the initial position.}
	\label{fig:net-c9responseposition}
\end{figure}

Finally, in terms of absolute distance of each robot from the target, in Figure 
\ref{fig:net-c9distance} we show that exploiting the communication, the robots 
are able to end up in the correct position by using a couple more time steps than 
expert does. 
\begin{figure}[!htb]
	\centering
	\includegraphics[width=.7\textwidth]{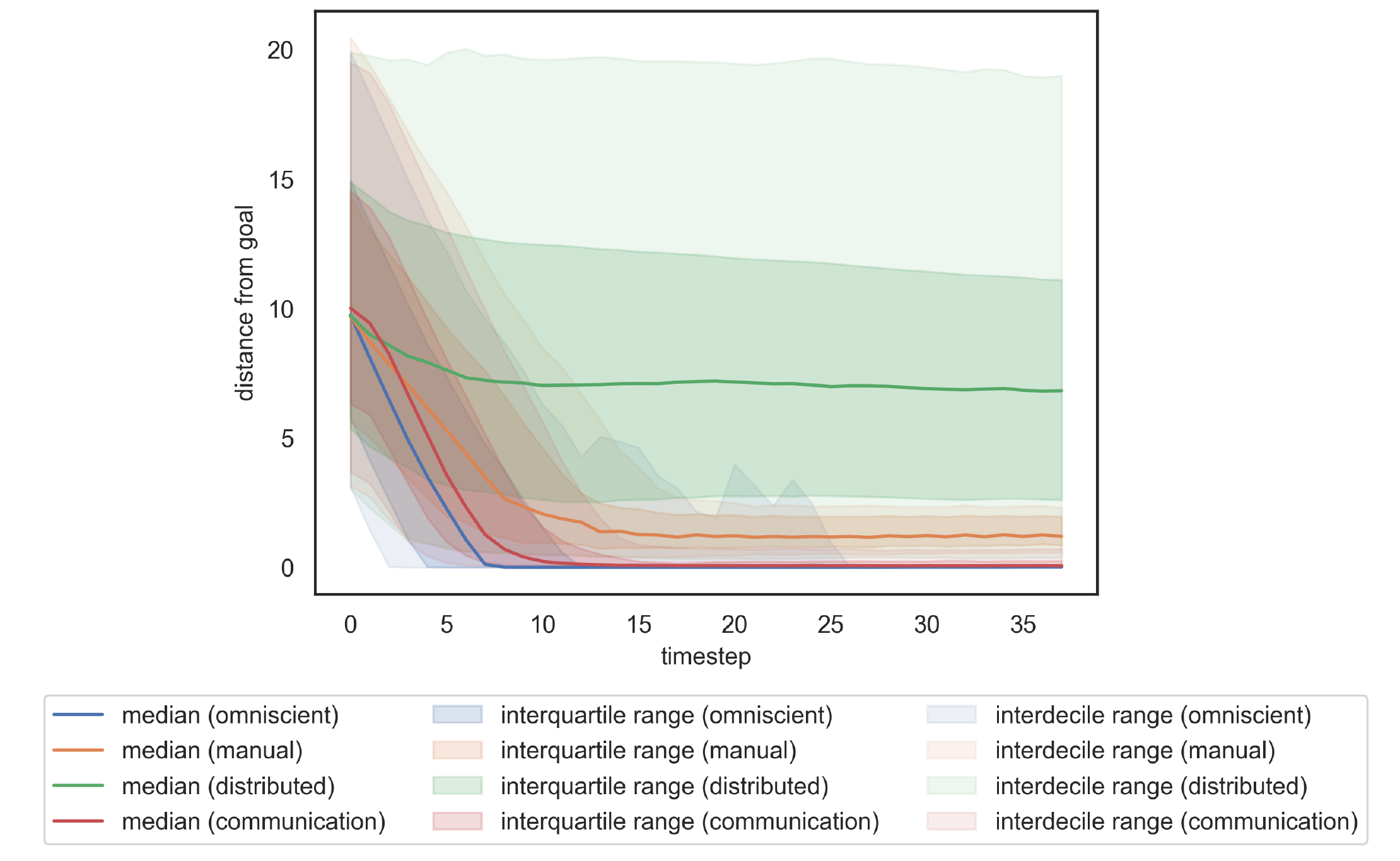}%
	\caption[Evaluation of \texttt{net-c9} distances from goal.]{Comparison of 
		performance in terms of distances from goal obtained using four controllers: 
		the expert, the manual and the two learned from \texttt{net-d9} and 
		\texttt{net-c9}.}
	\label{fig:net-c9distance}
\end{figure}

In this case, the analysis of the communication transmitted over time, in Figure 
\ref{fig:net-c9comm}, contradicts the hypotheses made for the previous 
experiments. The robots do not decrease or increase the value transmitted by 
approaching the target, demonstrating, as anticipated that this value is 
uncorrelated to the distance from goal, reaffirming the difficulty of understanding 
the protocol learned from the network.
\begin{figure}[!htb]
	\centering
	\includegraphics[width=.6\textwidth]{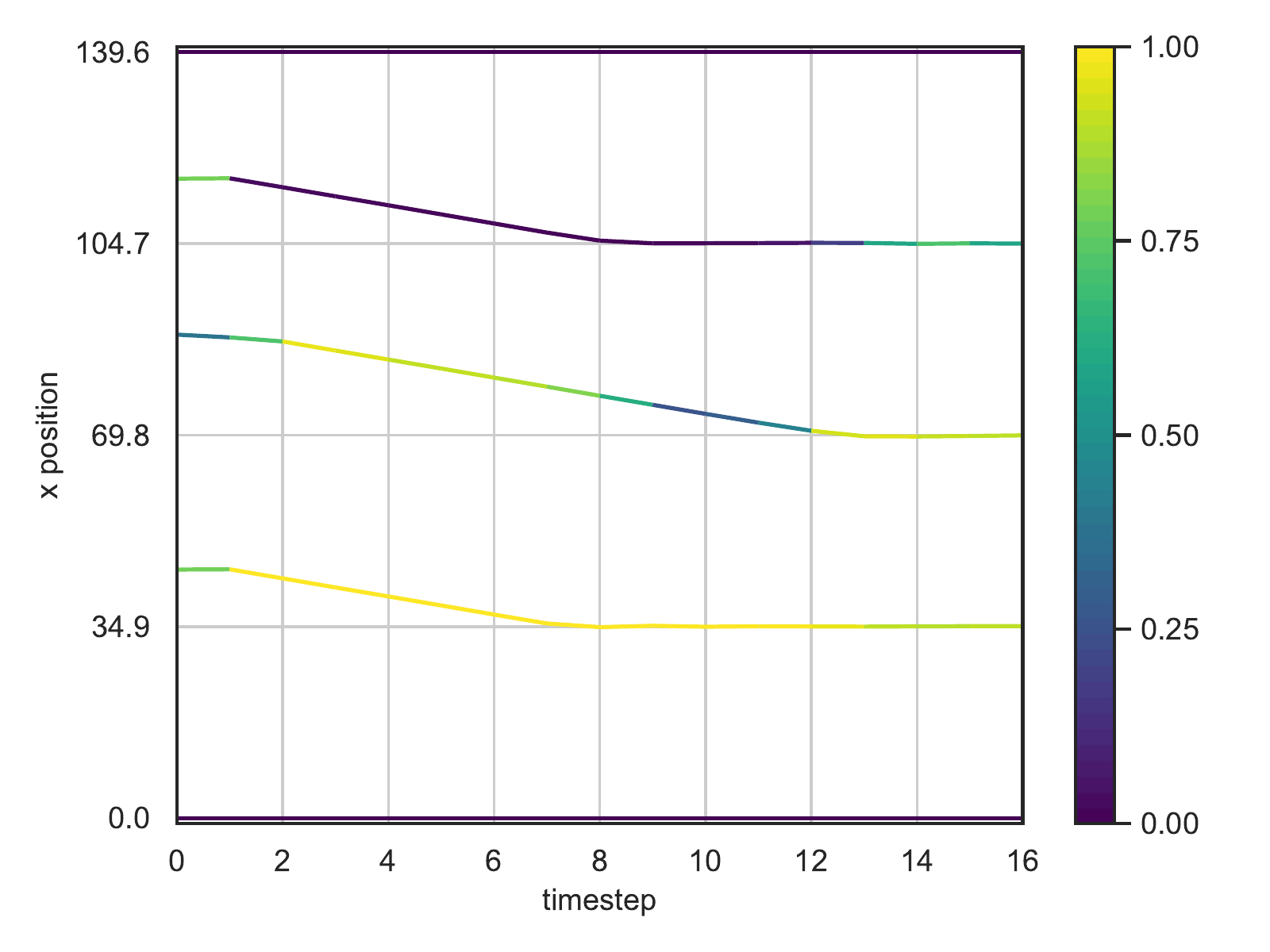}
	\vspace{-0.5cm}
	\caption[Evaluation of the communication learned by 
	\texttt{net-c9}.]{Visualisation of the communication values transmitted by each 
		robot over time using the controller learned from \texttt{net-c9}.}	
	\label{fig:net-c9comm}
\end{figure}

\paragraph*{Summary}

To sum up, we show in the figures below the losses of the trained models as the 
different inputs vary for each gap. In all the figures the blue line represents the 
loss using \texttt{prox\_values} as input, in orange \texttt{prox\_comm} and 
finally in green \texttt{all\_sensors}.
In both approaches, with \texttt{avg\_gap} of $8$cm, the model trained 
using \texttt{prox\_values} has a lower loss, followed, with a very close value, by 
the network that employ \texttt{all\_sensors}.
\begin{figure}[!htb]
	\begin{center}
		\begin{subfigure}[h]{0.32\textwidth}
			\includegraphics[width=\textwidth]{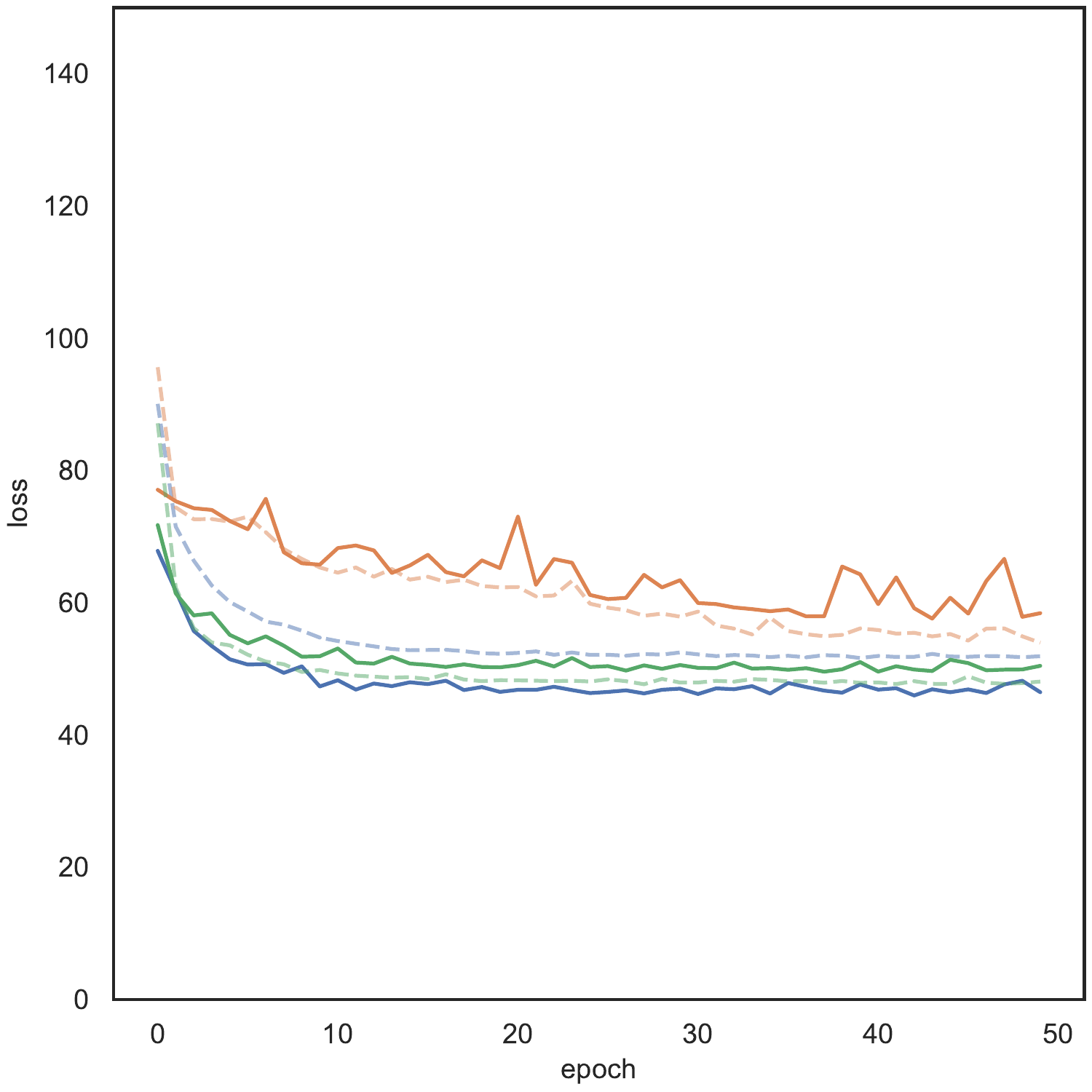}
			\caption{\texttt{avg\_gap} of $8$cm.}
		\end{subfigure}
		\hfill
		\begin{subfigure}[h]{0.32\textwidth}
			\includegraphics[width=\textwidth]{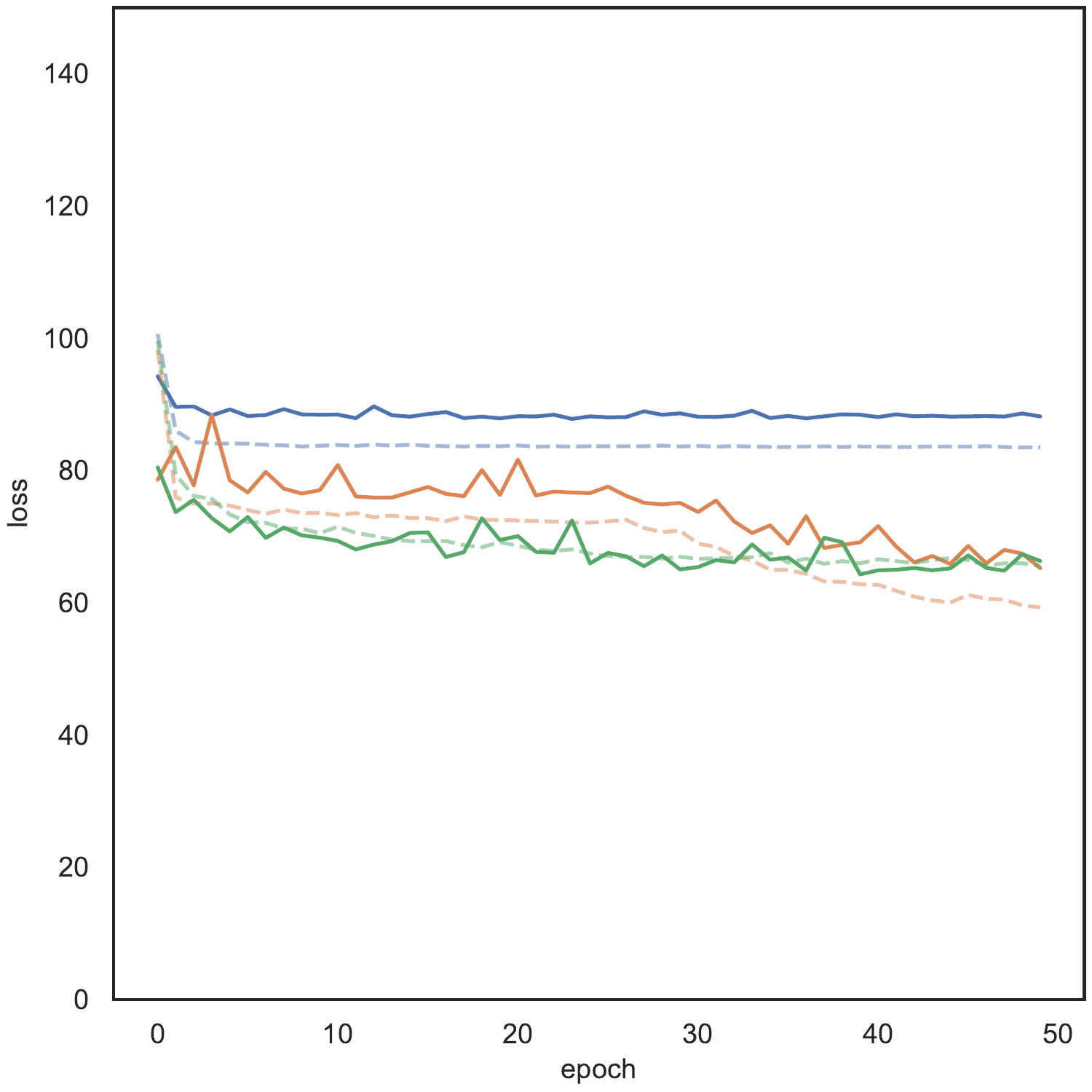}%
			\caption{\texttt{avg\_gap} of $13$cm.}
		\end{subfigure}
		\hfill
		\begin{subfigure}[h]{0.32\textwidth}
			\includegraphics[width=\textwidth]{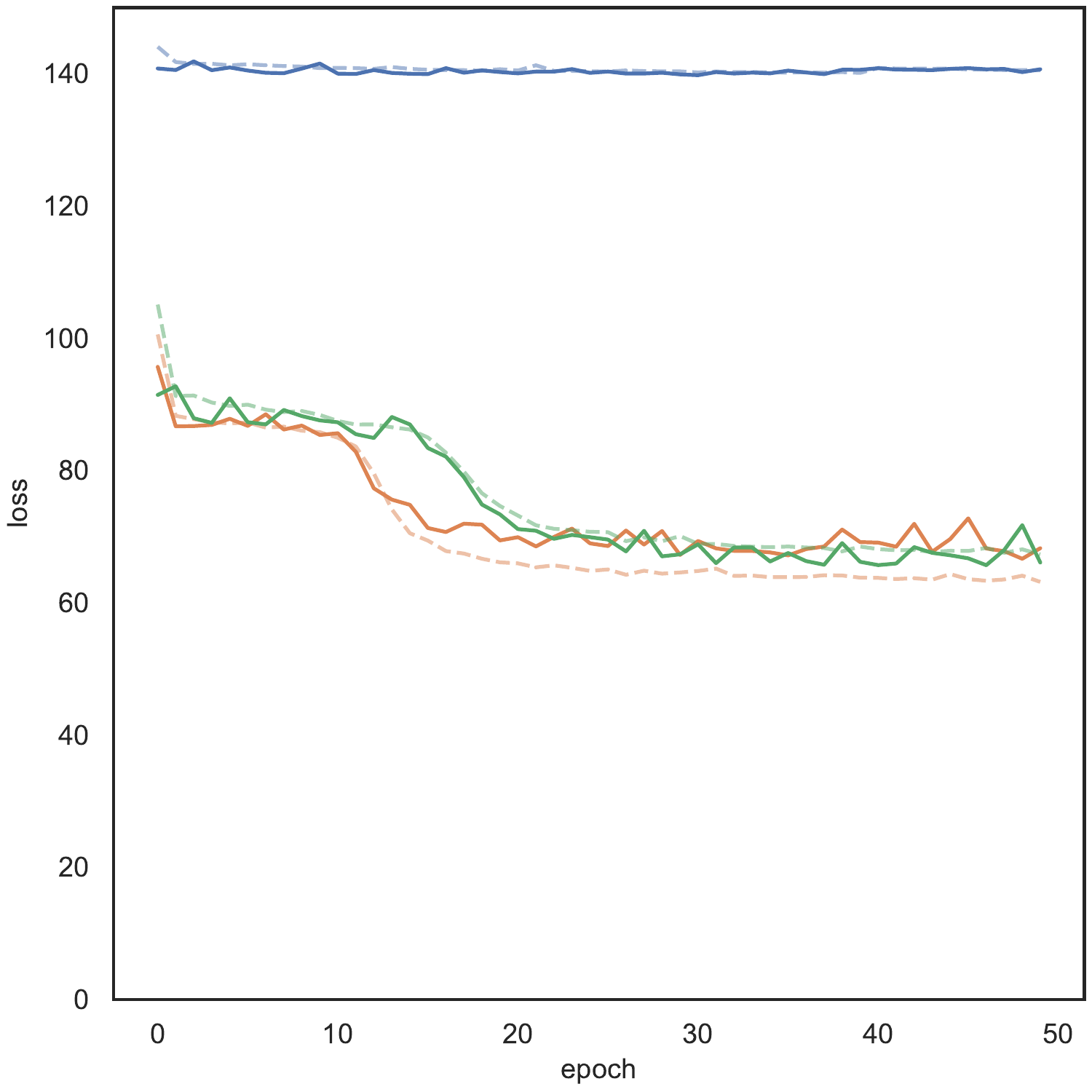}
			\caption{\texttt{avg\_gap} of $24$cm.}
		\end{subfigure}
	\end{center}
	\vspace{-0.5cm}
	\caption[Losses summary of the first set of experiments (no 
	communication).]{Comparison of the losses of the model trained without 
		communication, by varying the input of the networks for the three gaps.}
	\label{fig:distlossgaps}
\end{figure}

\begin{figure}[!htb]
	\begin{center}
		\begin{subfigure}[h]{0.32\textwidth}
			\includegraphics[width=\textwidth]{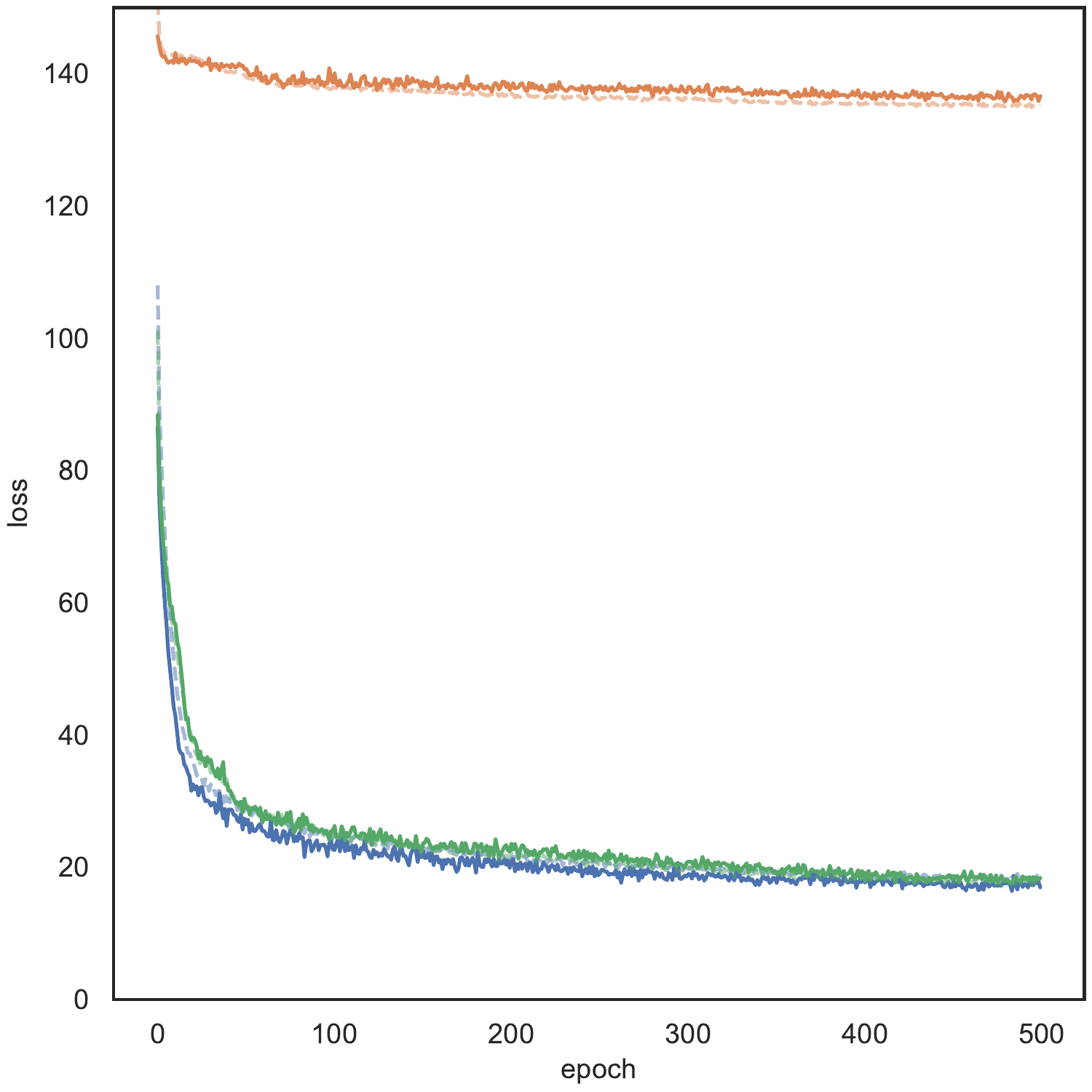}%
			\caption{\texttt{avg\_gap} of $8$cm.}
		\end{subfigure}
		\hfill
		\begin{subfigure}[h]{0.32\textwidth}
			\includegraphics[width=\textwidth]{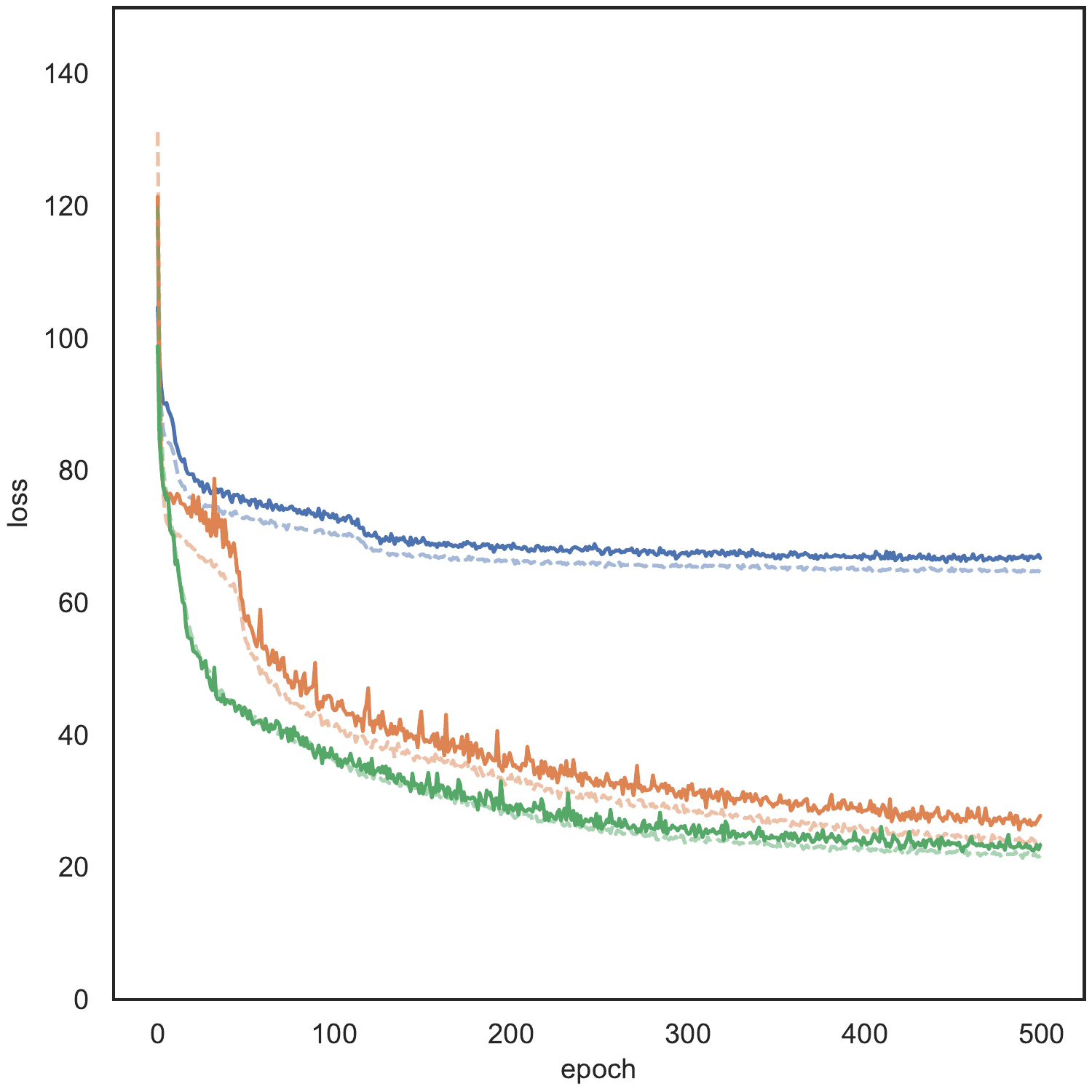}%
			\caption{\texttt{avg\_gap} of $13$cm.}
		\end{subfigure}
		\hfill
		\begin{subfigure}[h]{0.32\textwidth}
			\includegraphics[width=\textwidth]{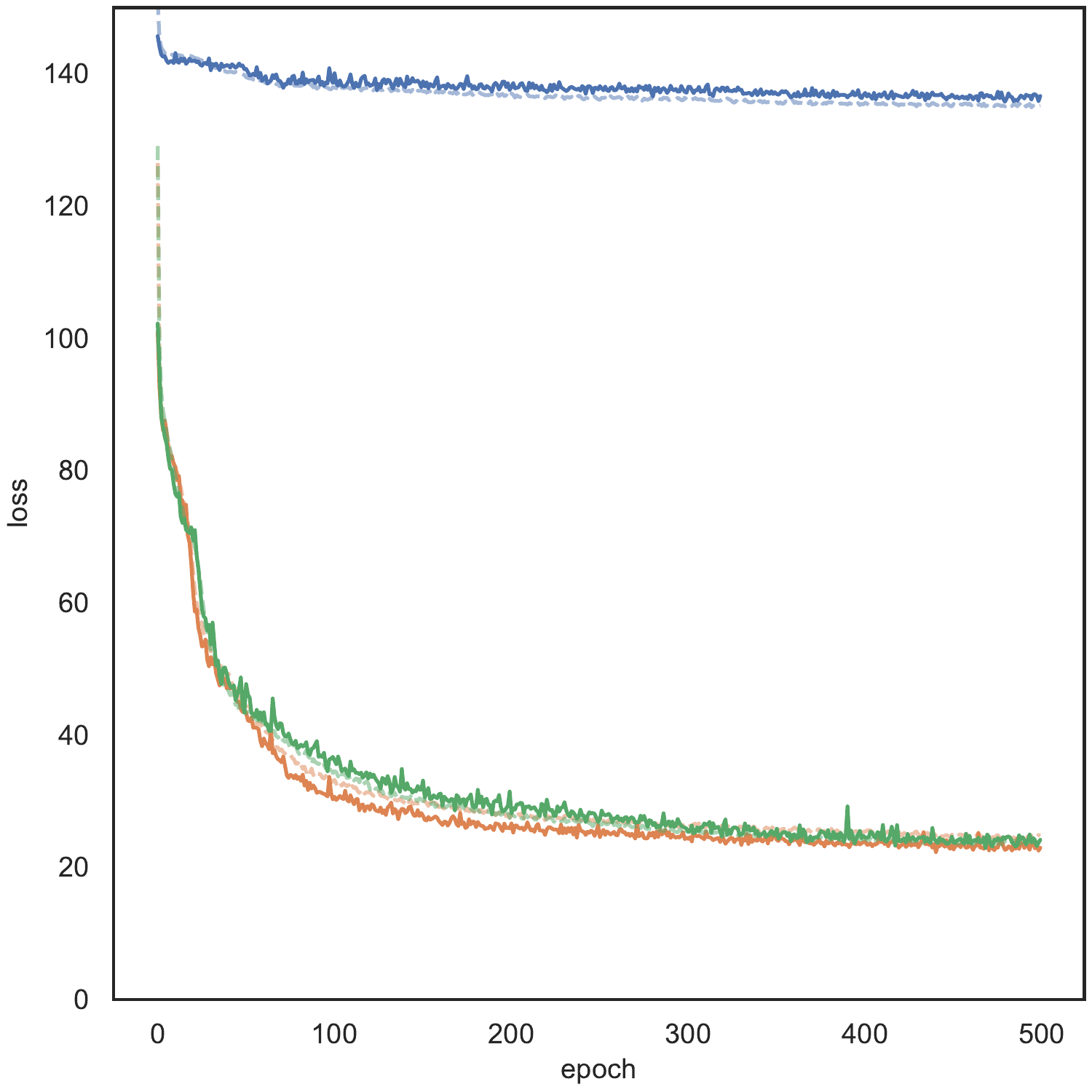}
			\caption{\texttt{avg\_gap} of $24$cm.}
		\end{subfigure}
	\end{center}
	\vspace{-0.5cm}
	\caption[Losses summary of the first set of experiments 
	(communication).]{Comparison of the losses of the model trained with 
		communication, by varying the input of the networks for the three gaps.}
	\label{fig:commloss81324}
\end{figure}

\noindent
Next is the model that works with \texttt{prox\_comm}, that is not able to work 
well with small gaps, even when using communication. 
Similarly, by increasing the gap to $13$cm, \texttt{prox\_values} is not able 
to achieve satisfactory results in both approaches, while used together with 
\texttt{prox\_comm}, \texttt{all\_sensors} reaches good performances, which can 
even be improved by using communication.
Finally, by increasing the gap even more, up to $24$cm, 
\texttt{prox\_values} becomes completely unusable, while \texttt{prox\_comm} 
and \texttt{all\_sensors} have excellent responses, in particular with the new 
approach.

\subsubsection{Experiment 2: variable number of agents}
\label{subsubsec:task1-exp-comm-2}

The second group of experiments we carried out using a distributed approach 
with communication, summarised in Table \ref{tab:modelcomm}, examines the 
behaviour of the control learned using \texttt{all\_sensors} inputs. 
\begin{figure}[!htb]
	\centering
	\begin{tabular}{ccccc}
		\toprule
		\textbf{Model} \quad & \textbf{\texttt{network\_input}} & 
		\textbf{\texttt{input\_size}} & \textbf{\texttt{avg\_gap}} & \textbf{\texttt{N}}\\
		\midrule
		\texttt{net-c10} 	& \texttt{all\_sensors}		&  $14$  &  $8$		 	 &	 $5$ \\
		\texttt{net-c11} 	& \texttt{all\_sensors}		&  $14$  &  $20$		&	$5$ \\
		\texttt{net-c12} 	& \texttt{all\_sensors}		&  $14$  &  variable   &    $5$\\
		\texttt{net-c13} 	& \texttt{all\_sensors}	  	&  $14$  &  $8$			 &	  $8$ \\
		\texttt{net-c14} 	& \texttt{all\_sensors}	  	&  $14$  &  $20$   		&	 $8$ \\
		\texttt{net-c15} 	& \texttt{all\_sensors}	  	&  $14$  &  variable	&	 $8$ \\
		\texttt{net-c16} 	& \texttt{all\_sensors}	  	&  $14$  &  $ 8$		  &	 variable\\
		\texttt{net-c17} 	& \texttt{all\_sensors}	  	&  $14$  &  $20$		 &	variable\\
		\texttt{net-c18} 	& \texttt{all\_sensors}	  	&  $14$  &  variable	 &	
		variable\\
		\bottomrule
	\end{tabular}
	\captionof{table}[Experiments with variable agents and gaps 
	(communication).]{List of the experiments carried out using a variable number 
		of agents and of gaps.}
	\label{tab:modelcomm}
\end{figure}

The objective of this set of experiments is to verify the robustness of the models, 
proving that it is possible to train networks that use a variable number of agents.
\begin{figure}[!htb]
	\centering
	\includegraphics[width=.8\textwidth]{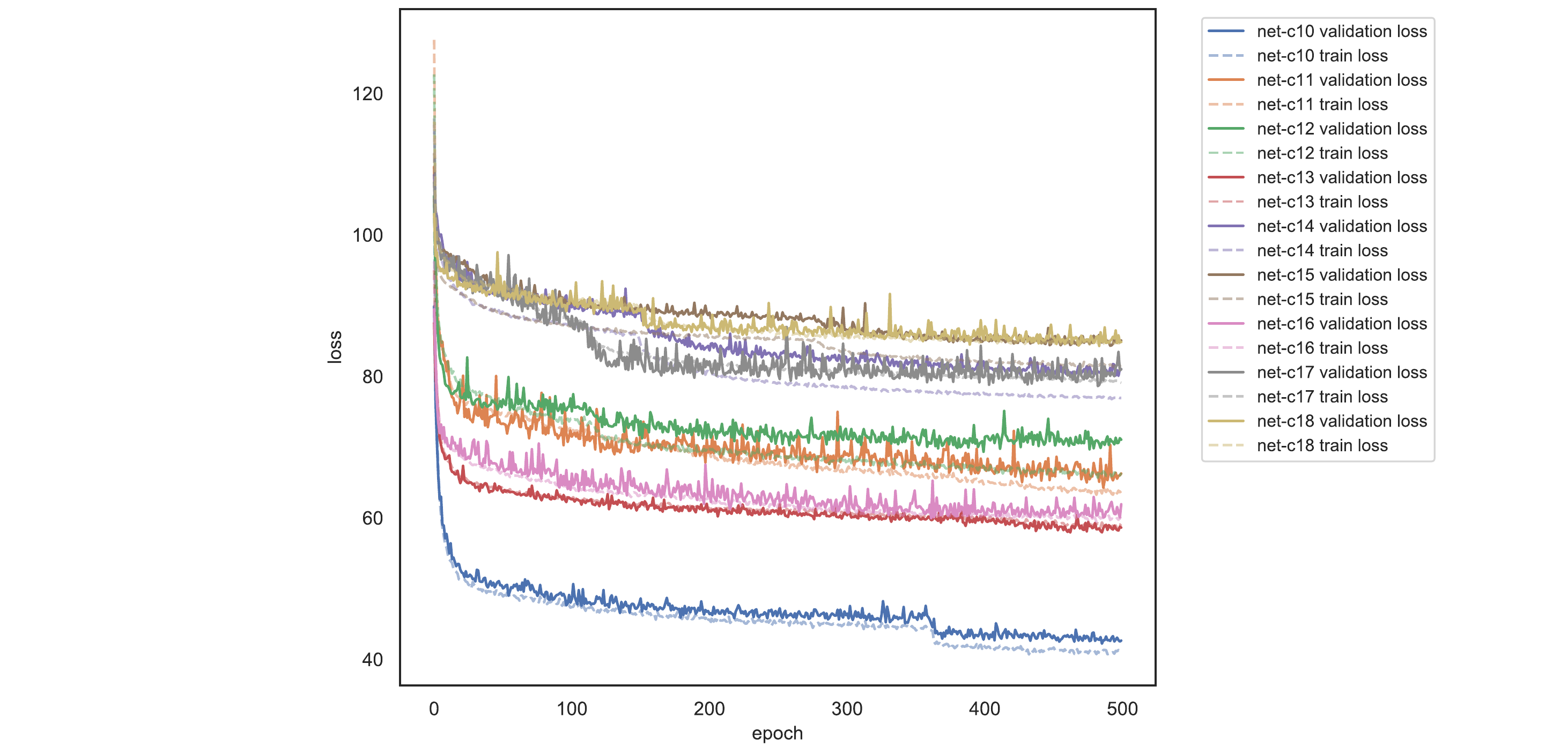}%
	\caption[Comparison of losses of the second set of experiments.]{Comparison 
		of the losses of the models carried out using a variable number of agents and 
		of average gap.}
	\label{fig:commexlossallt}
\end{figure}

We analyse the same experiments of the previous approach, presented in Section 
\ref{subsec:task1-exp-distr-comm}.
We focus our examination by inspecting the behaviour of the network trained on 
simulations with different number of robots $N$ and variable average gap, i.e 
\texttt{net-c12}, \texttt{net-c15} and \texttt{net-c18}. Then we compare the 
performances obtained for these models to the corresponding distributed 
networks, i.e., \texttt{net-d12}, \texttt{net-d15} and \texttt{net-d18}.
In Figure \ref{fig:commexlossallt} os shown an overview of the train and 
validation losses obtained for these models.

\paragraph*{Results using 5 agents}
Proceeding step by step, we summarise in Figure \ref{fig:commlossn5} the losses 
of the experiments carried out using a $5$ agents, the same of the first group of 
experiments presented above, in order to highlight the difference of performance 
using a gap that is first small, then large and finally variable, respectively 
represented 
\begin{figure}[!htb]
	\begin{center}
		\begin{subfigure}[h]{0.49\textwidth}
			\centering
			\includegraphics[width=.7\textwidth]{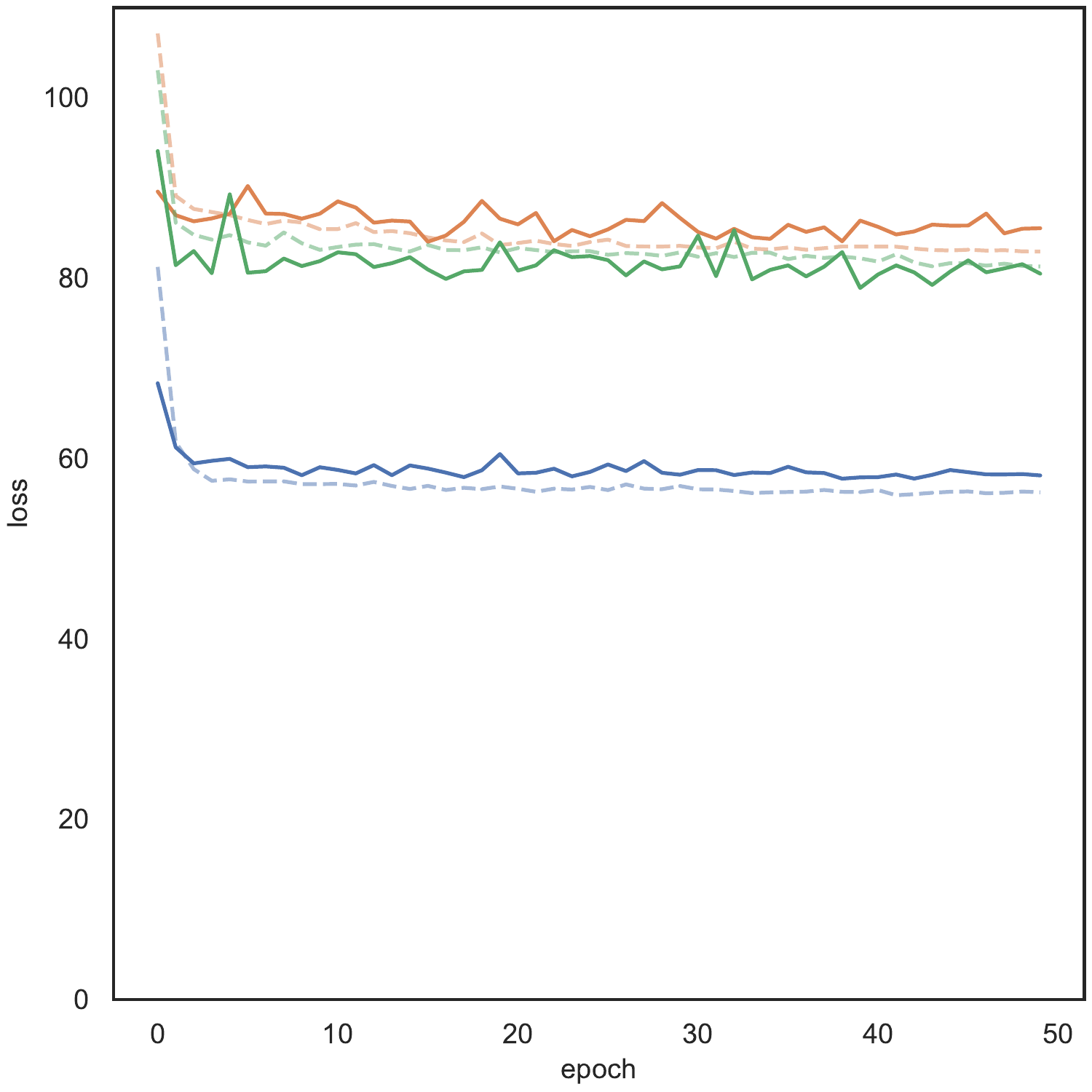}
			\caption{Distributed approach.}
		\end{subfigure}
		\hfill
		\begin{subfigure}[h]{0.49\textwidth}
			\centering
			\includegraphics[width=.7\textwidth]{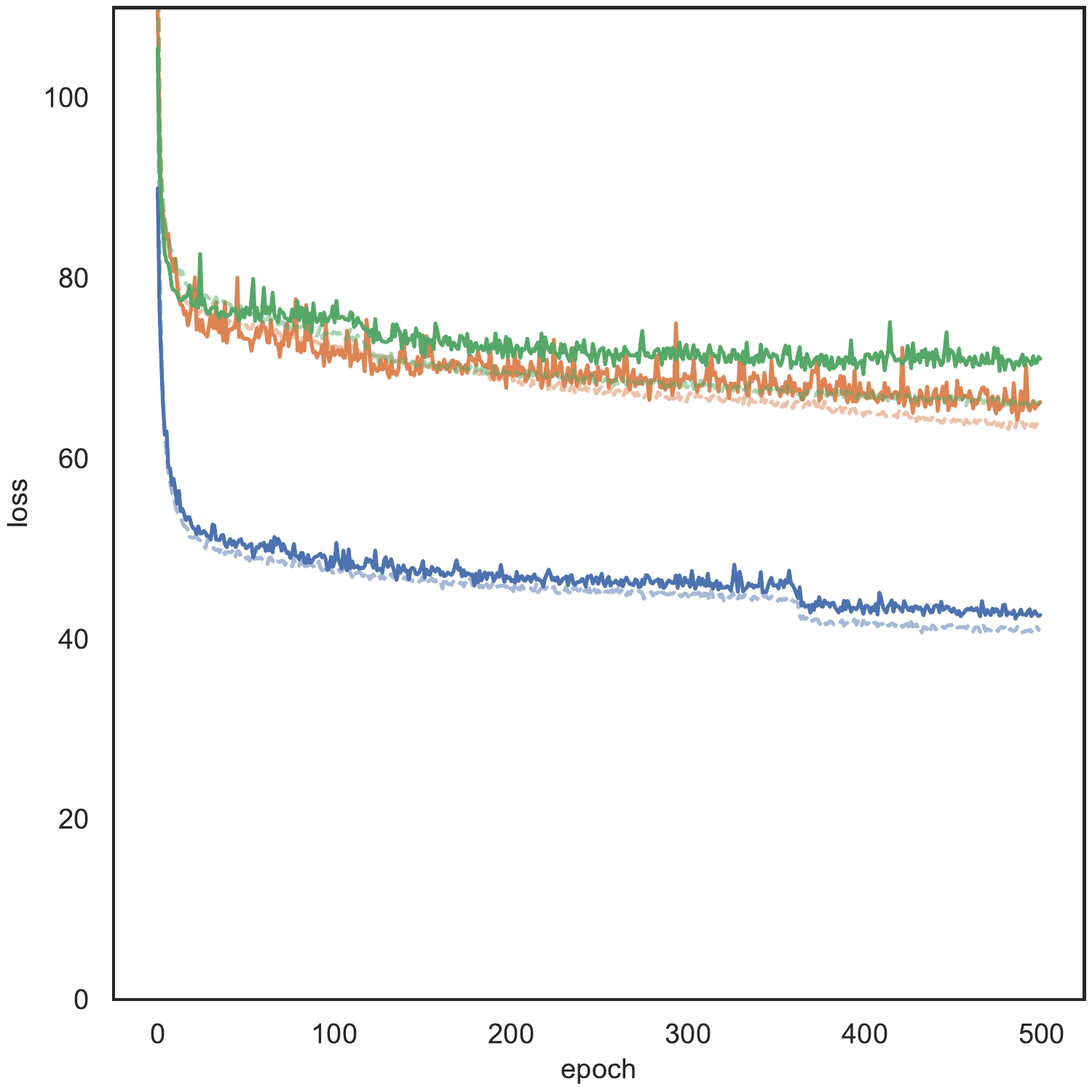}
			\caption{Distributed approach with communication.}
		\end{subfigure}	
	\end{center}
	\vspace{-0.5cm}
	\caption[Comparison of the losses of the models that use $5$ 
	agents.]{Comparison of the losses of the models that use $5$ agents as 
		the gap varies.}
	\label{fig:commlossn5}
\end{figure}
by the blue, the orange and the green lines.
These results are also compared to those obtained without employing the 
communication.
Clearly, in case of small gaps the network performs better. We can also note that 
in general by enabling the communication, the losses decrease, meaning an 
improvement over the previous approach.

Considering the model trained using a variable gap, in Figure \ref{fig:net-c12r2} 
are shown the \ac{r2} coefficients of the manual and the learned controllers, for 
both approaches.
From these we expect that the behaviour of the robots using the learned 
controller with communication instead of the manual or the distributed alone is 
better, even if far from the expert. The new model produces an increase in the 
coefficient from $0.41$ to $0.49$.
\begin{figure}[!htb]
	\begin{center}
		\begin{subfigure}[h]{0.49\textwidth}
			\includegraphics[width=\textwidth]{contents/images/net-d12/regression-net-d12-vs-omniscient}%
		\end{subfigure}
		\hfill\vspace{-0.5cm}
		\begin{subfigure}[h]{0.49\textwidth}
			\includegraphics[width=\textwidth]{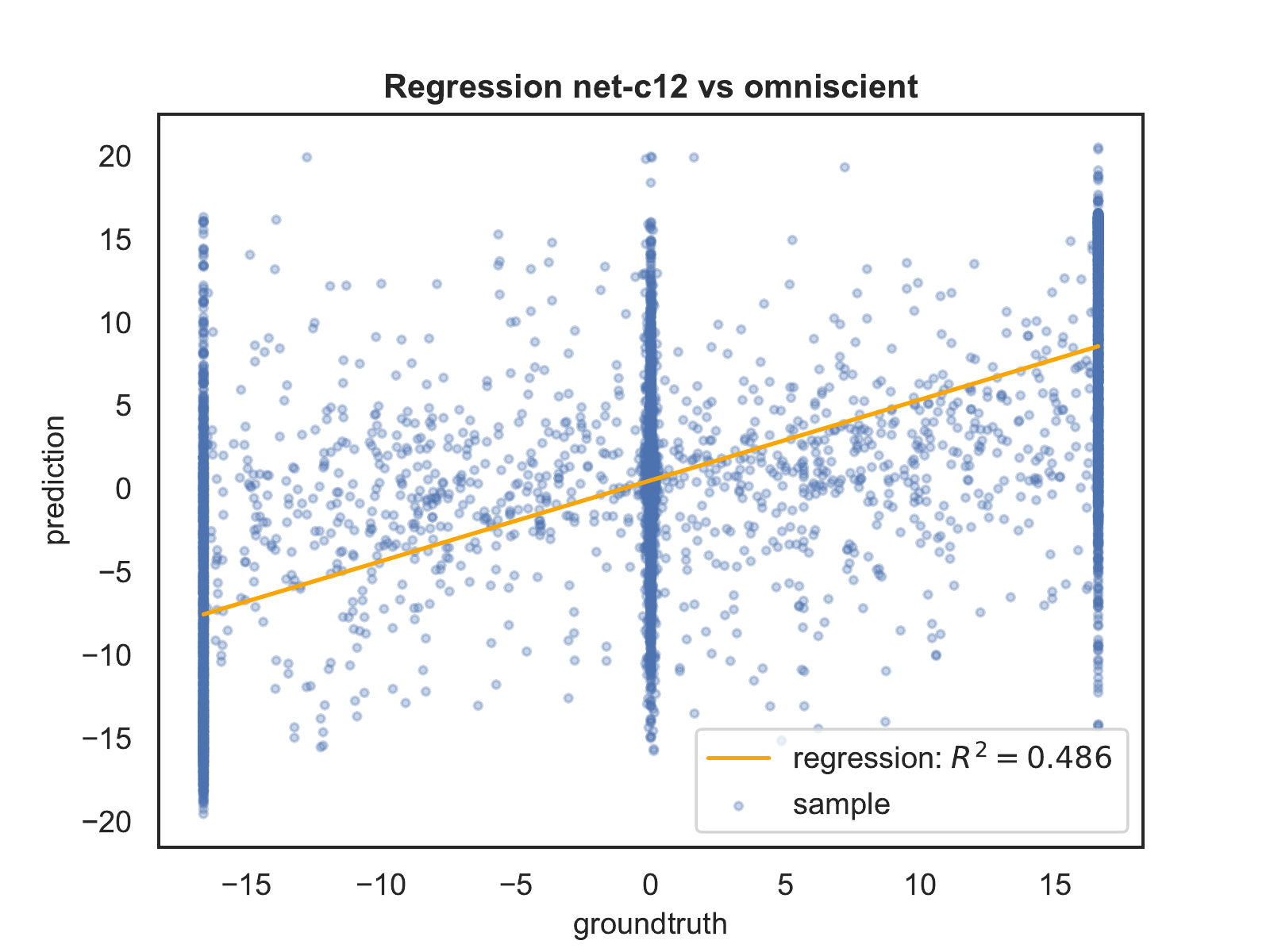}%
		\end{subfigure}
	\end{center}
	\caption[Evaluation of the \ac{r2} coefficients of \texttt{net-c12}.]{Comparison 
		of the \ac{r2} coefficients of the controllers learned from 
		\texttt{net-d12} and \texttt{net-c12}, with respect to the omniscient one.}
	\label{fig:net-c12r2}
\end{figure}

\begin{figure}[!htb]
	\centering
	\includegraphics[width=.7\textwidth]{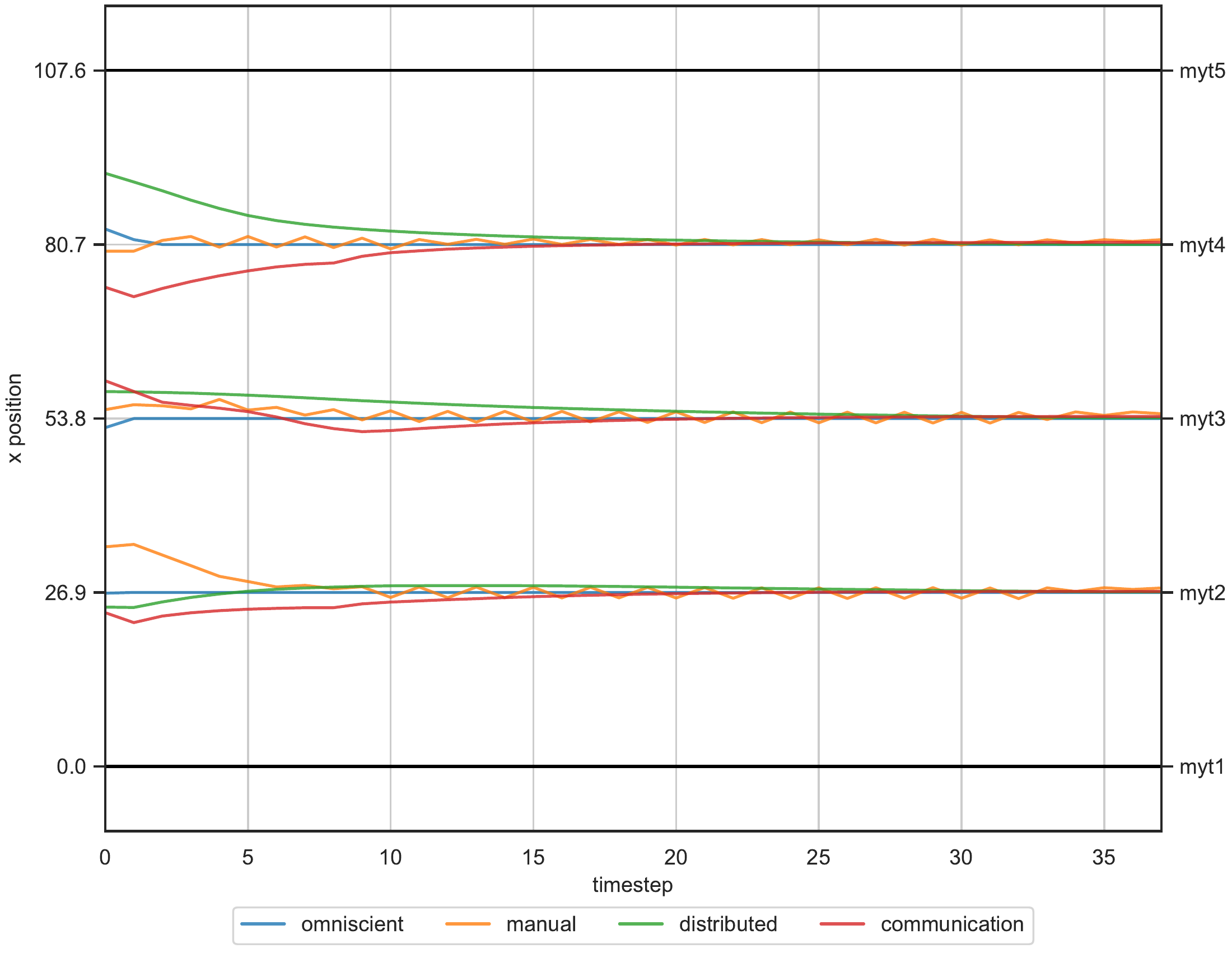}%
	\caption[Evaluation of the trajectories obtained with 5 agents.]{Comparison of 
		trajectories, of a single simulation, generated using four 
		controllers: the expert, the manual and the two learned from \texttt{net-d12}
		and \texttt{net-c12}.}
	\label{fig:net-c12traj1}
\end{figure}
In Figure \ref{fig:net-c12traj1} we first show a sample simulation: on the y-axis 
are visualised the positions of the agents, while on the x-axis the simulation time 
steps. 
The behaviour of the agents moved using the new learned controller seems very 
similar to that obtained using the distributed one.

Analysing in Figure \ref{fig:net-c12traj} the trajectories obtained employing the 
four controllers, it is difficult to demonstrate the improvement over the previous 
approach as having a variable gap we observe a deviation of the position of the 
robots with respect to the average that in this case it does not indicate a limitation 
of the model but just that the target positions are different among the simulation 
runs. 
\begin{figure}[!htb]
	\begin{center}
		\begin{subfigure}[h]{0.49\textwidth}
			\centering
			\includegraphics[width=.9\textwidth]{contents/images/net-d12/position-overtime-omniscient}%
			\caption{Expert controller trajectories.}
		\end{subfigure}
		\hfill
		\begin{subfigure}[h]{0.49\textwidth}
			\centering
			\includegraphics[width=.9\textwidth]{contents/images/net-d12/position-overtime-learned_distributed}
			\caption{Distributed controller trajectories.}
		\end{subfigure}
	\end{center}
	\begin{center}
		\begin{subfigure}[h]{0.49\textwidth}
			\centering			
			\includegraphics[width=.9\textwidth]{contents/images/net-d12/position-overtime-manual}%
			\caption{Manual controller trajectories.}
		\end{subfigure}
		\hfill
		\begin{subfigure}[h]{0.49\textwidth}
			\centering
			\includegraphics[width=.9\textwidth]{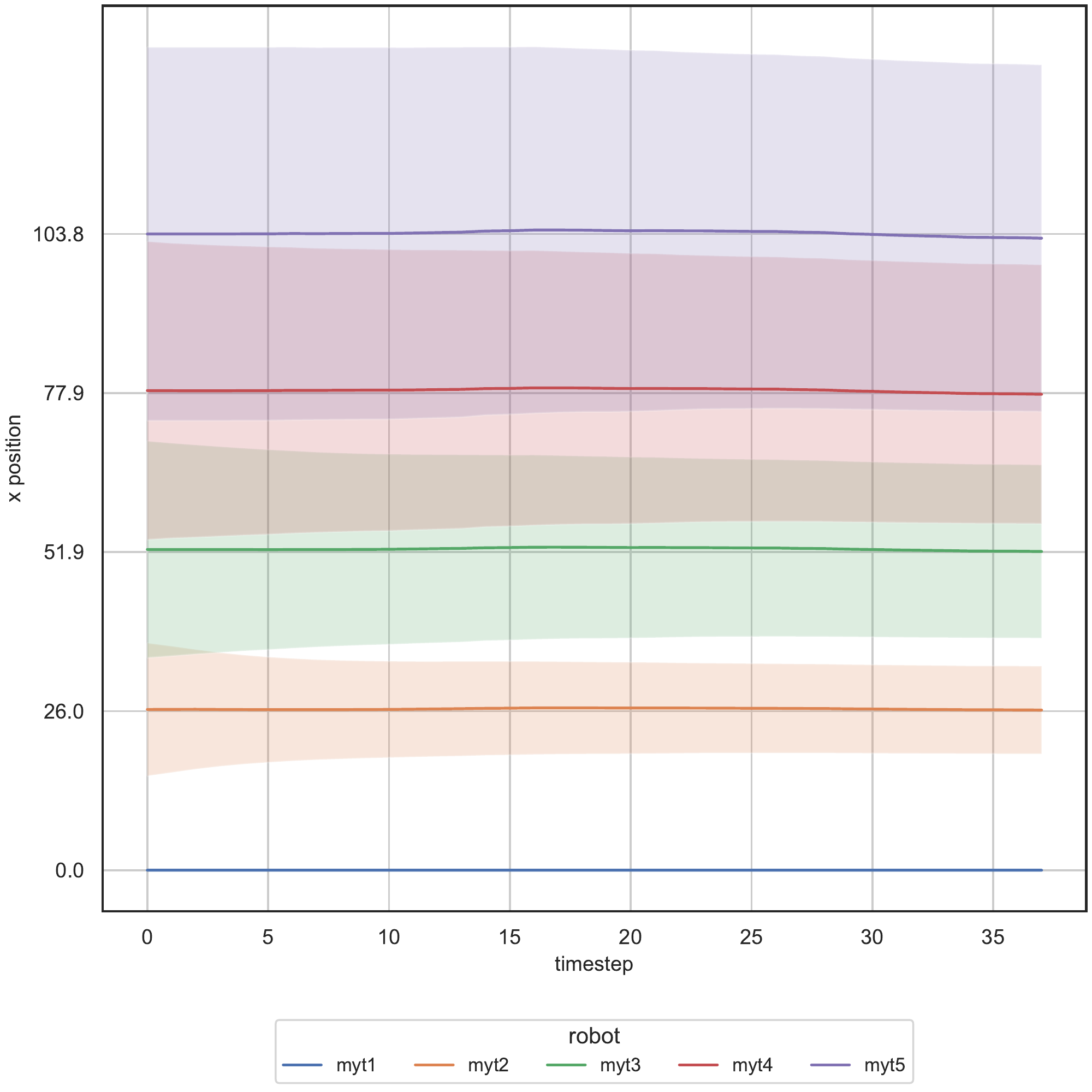}
			\caption{Communication controller trajectories.}
		\end{subfigure}
	\end{center}
	\vspace{-0.5cm}
	\caption[Evaluation of the trajectories learned by \texttt{net-c12}.]{Comparison 
	of trajectories, of all the simulation runs, generated using four controllers: the 
	expert, the manual and the two learned from \texttt{net-d12} and 
	\texttt{net-c12}.}
	\label{fig:net-c12traj}
\end{figure}

\bigskip
An analysis of the evolution of the control over time in Figure 
\ref{fig:net-c12control} evidence that the control decided by the communication 
network is similar to that set by the expert.
\begin{figure}[!htb]
	\begin{center}
		\begin{subfigure}[h]{0.35\textwidth}
			\includegraphics[width=\textwidth]{contents/images/net-d12/control-overtime-omniscient}%
			\caption{Expert control.}
		\end{subfigure}
		\hspace{1cm}
		\begin{subfigure}[h]{0.35\textwidth}
			\includegraphics[width=\textwidth]{contents/images/net-d12/control-overtime-learned_distributed}
			\caption{Distributed control.}
		\end{subfigure}
	\end{center}
	\begin{center}
		\begin{subfigure}[h]{0.35\textwidth}			
			\includegraphics[width=\textwidth]{contents/images/net-d12/control-overtime-manual}%
			\caption{Manual control.}
		\end{subfigure}
		\hspace{1cm}
		\begin{subfigure}[h]{0.35\textwidth}
			\includegraphics[width=\textwidth]{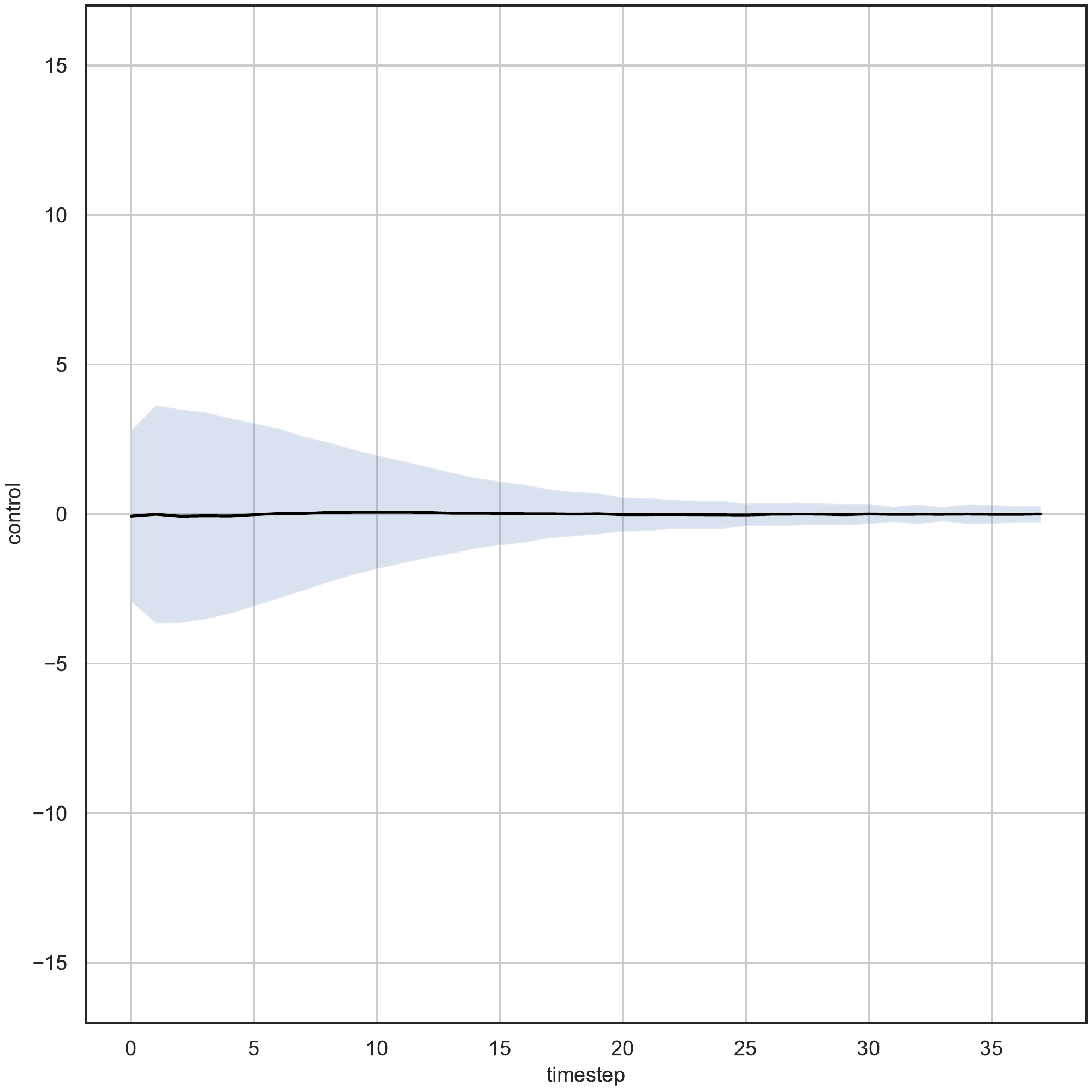}
			\caption{Communication control.}
		\end{subfigure}
	\end{center}
	\vspace{-0.5cm}
	\caption[Evaluation of the control decided by \texttt{net-c12}.]{Comparison of 
	output control decided using four controllers: the expert, the manual and the 
	two learned from \texttt{net-d12} and \texttt{net-c12}.}
	\label{fig:net-c12control}
\end{figure}

In Figure \ref{fig:net-c12responseposition} is displayed the behaviour of a robot 
located between other two stationary agents which are already in the correct 
position, 
showing the response of the controllers, on the y-axis, by varying the position of 
the moving robot, on the x-axis.  
As expected, the output is a high positive value when the robot is close to an 
obstacle on the left, negative when there is an obstacle in front and not behind, 
and $0$ when the distance from right and left is equal.
The behaviour of the controller that uses the communication is the most accurate 
when the moving robot is halfway between the two stationary.

Finally, in Figure \ref{fig:net-c12distance} are presented the absolute distances of 
each robot from the target, visualised on the y-axis, over time.
On average, the distance from goal of the communication controller is a bit better 
than that obtained with the distributed while similar to the manual, even if slower. 
In the final configuration in all three cases the agents are $1$cm away from 
the target. 

\begin{figure}[!htb]
	\centering
	\includegraphics[width=.46\textwidth]{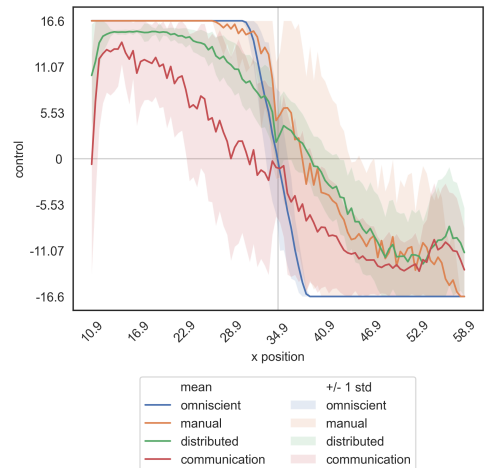}%
	\caption{Response of \texttt{net-c12} by varying the initial position.}
	\label{fig:net-c12responseposition}
	\vspace{0.5cm}
\end{figure}

\begin{figure}[!htb]
	\centering
	\includegraphics[width=.68\textwidth]{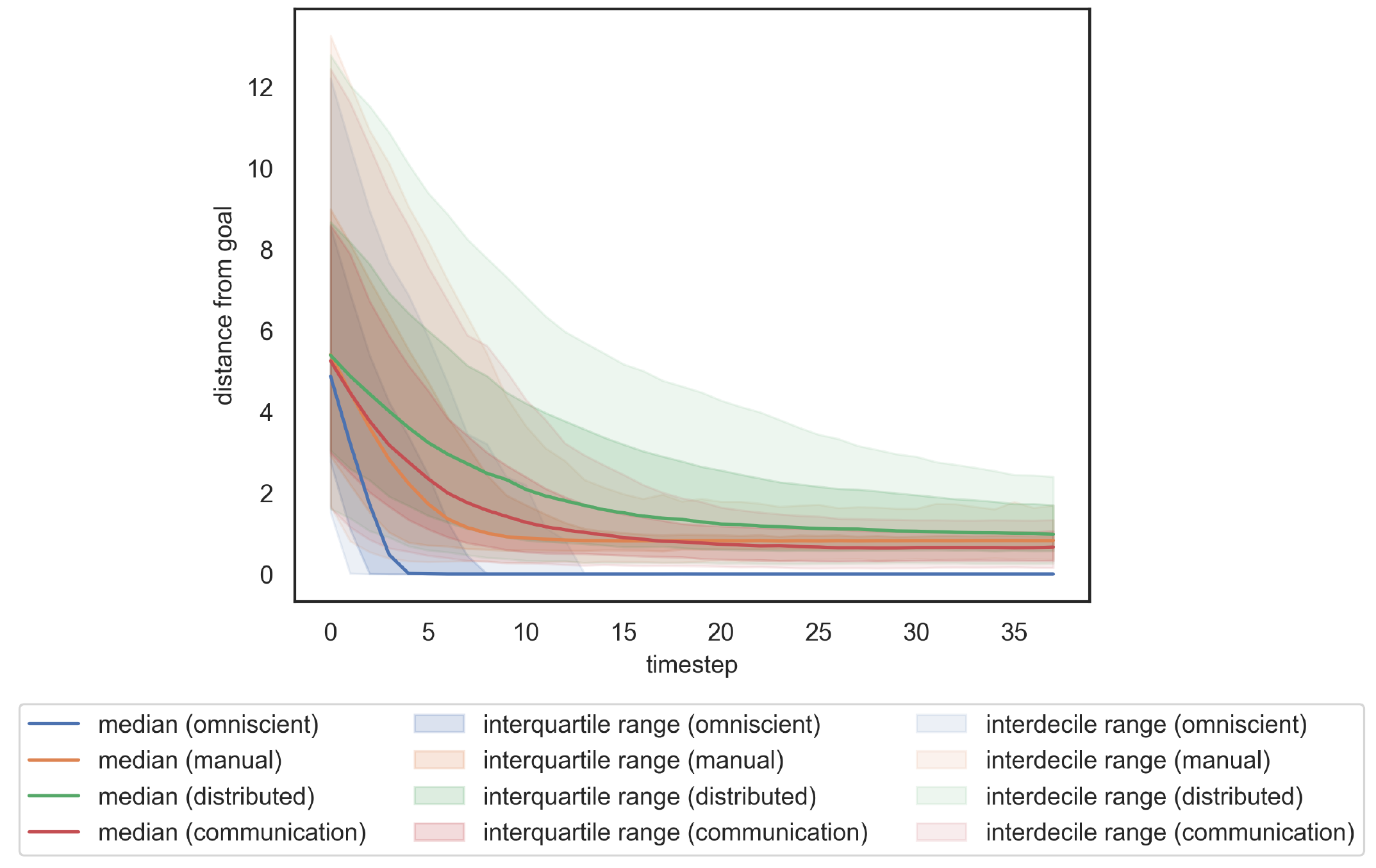}%
	\caption[Evaluation of \texttt{net-c12} distances from goal.]{Comparison of 
		performance in terms of distances from goal obtained using four controllers: 
		the expert, the manual and the two learned from \texttt{net-d12} and 
		\texttt{net-c12}.}
	\label{fig:net-c12distance}
\end{figure}

\paragraph*{Results using 8 agents}
Following are presented the results of the experiments performed using 
$8$ agents. In Figure \ref{fig:commlossN8}, are shown the losses by varying the 
average gap, as before the blue, orange and green lines represent respectively 
gaps of $8$cm, $20$cm and variable. From a first observation we see 
that the networks perform better in case of small gaps and when using 
communication.

Focusing on the models that use the dataset generate using a variable average 
gap, in Figure \ref{fig:net-c15r2} we observe the \ac{r2} of the manual and the 
learned controllers, both the one with and the one without communication, on 
the validation sets.
Given the fact that the coefficient is increased from $0.23$ up to $0.32$, we 
expect superior performance.  

\begin{figure}[!htb]
	\begin{center}
		\begin{subfigure}[h]{0.49\textwidth}
			\centering
			\includegraphics[width=.72\textwidth]{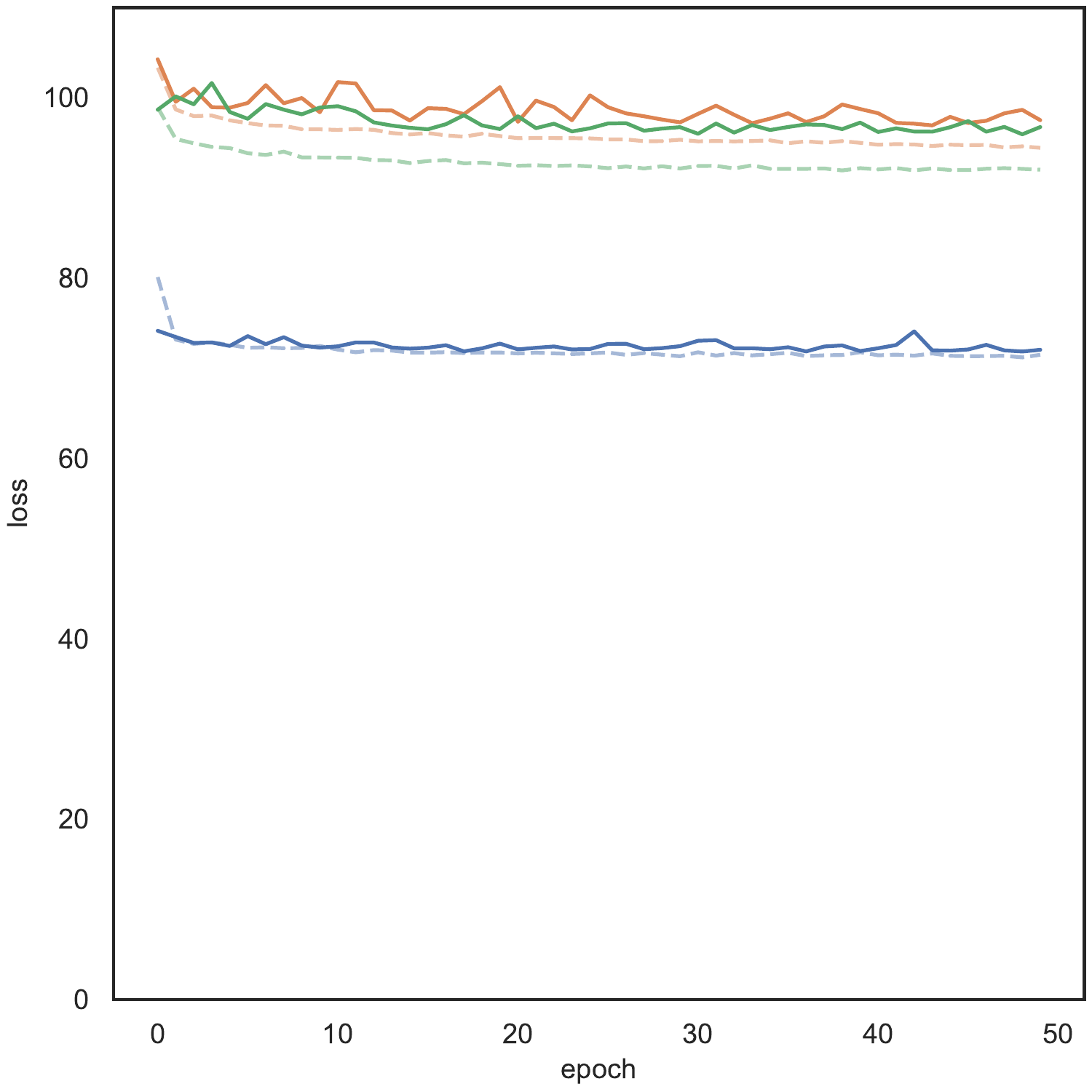}
			\caption{Distributed approach.}
		\end{subfigure}
		\hfill
		\begin{subfigure}[h]{0.49\textwidth}
			\centering
			\includegraphics[width=.72\textwidth]{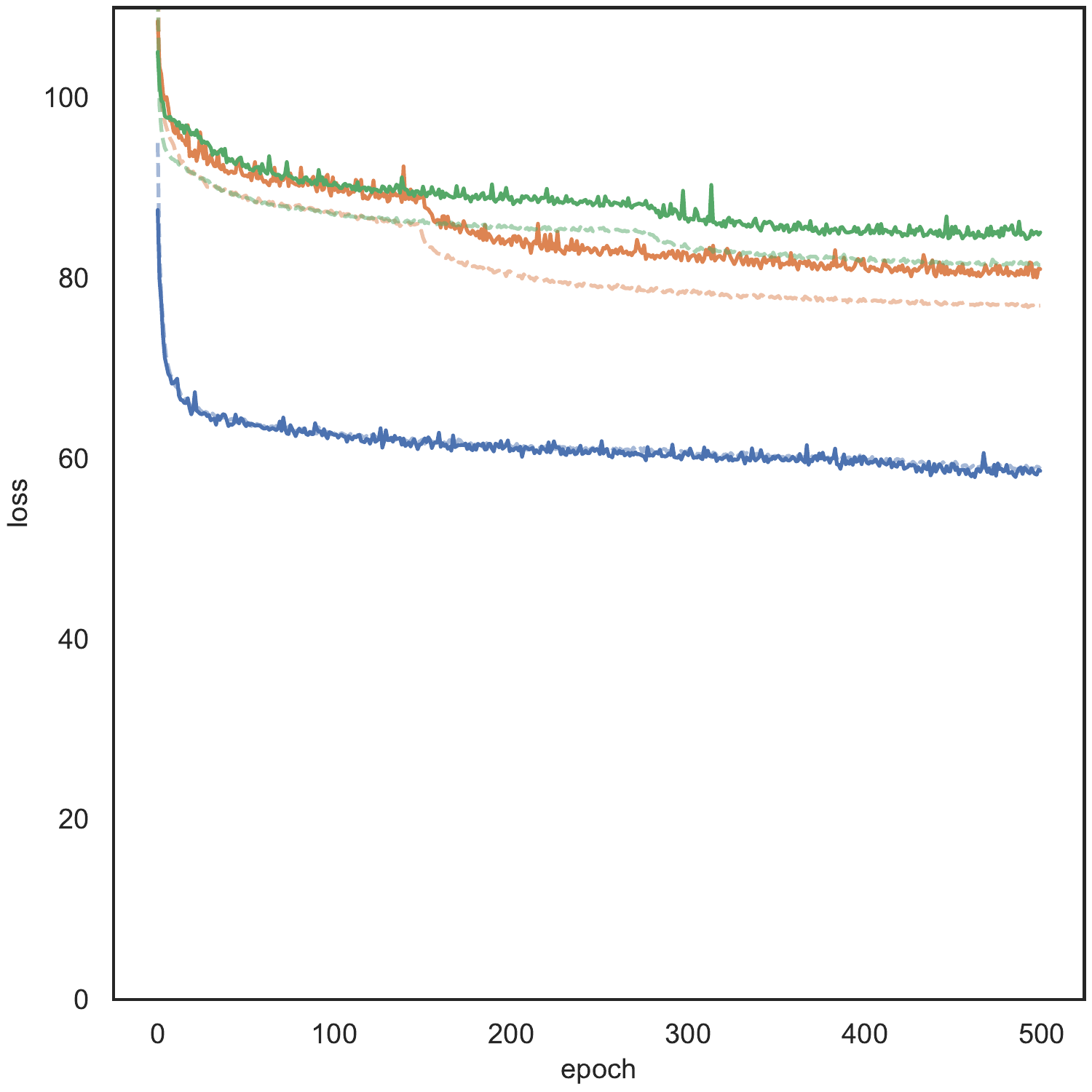}
			\caption{Distributed approach with communication.}
		\end{subfigure}	
	\end{center}
	\vspace{-0.5cm}
	\caption[Comparison of the losses of the models that use $8$ 
	agents.]{Comparison of the losses of the models that use $8$ agents as the gap 
	varies.}
	\label{fig:commlossN8}
\end{figure}

\begin{figure}[!htb]
	\begin{center}
		\begin{subfigure}[h]{0.49\textwidth}
			\includegraphics[width=\textwidth]{contents/images/net-d15/regression-net-d15-vs-omniscient}%
		\end{subfigure}
		\hfill\vspace{-0.5cm}
		\begin{subfigure}[h]{0.49\textwidth}
			\includegraphics[width=\textwidth]{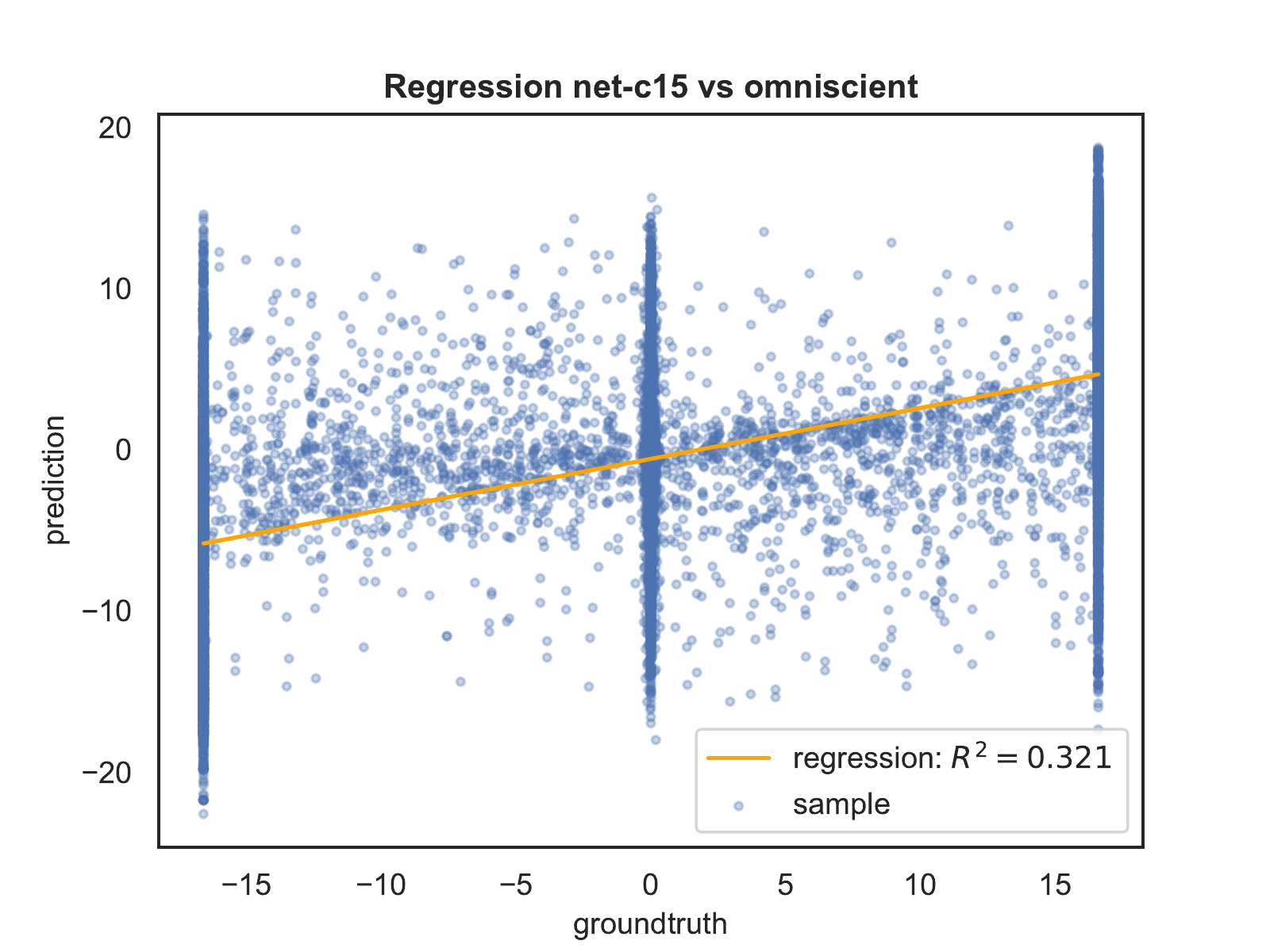}%
		\end{subfigure}
	\end{center}
	\caption[Evaluation of the \ac{r2} coefficients of \texttt{net-c15}.]{Comparison 
		of the \ac{r2} coefficients of the controllers learned from 
		\texttt{net-d15} and \texttt{net-c15}, with respect to the omniscient one.}
	\label{fig:net-c15r2}
\end{figure}

\begin{figure}[H]
	\centering
	\includegraphics[width=.65\textwidth]{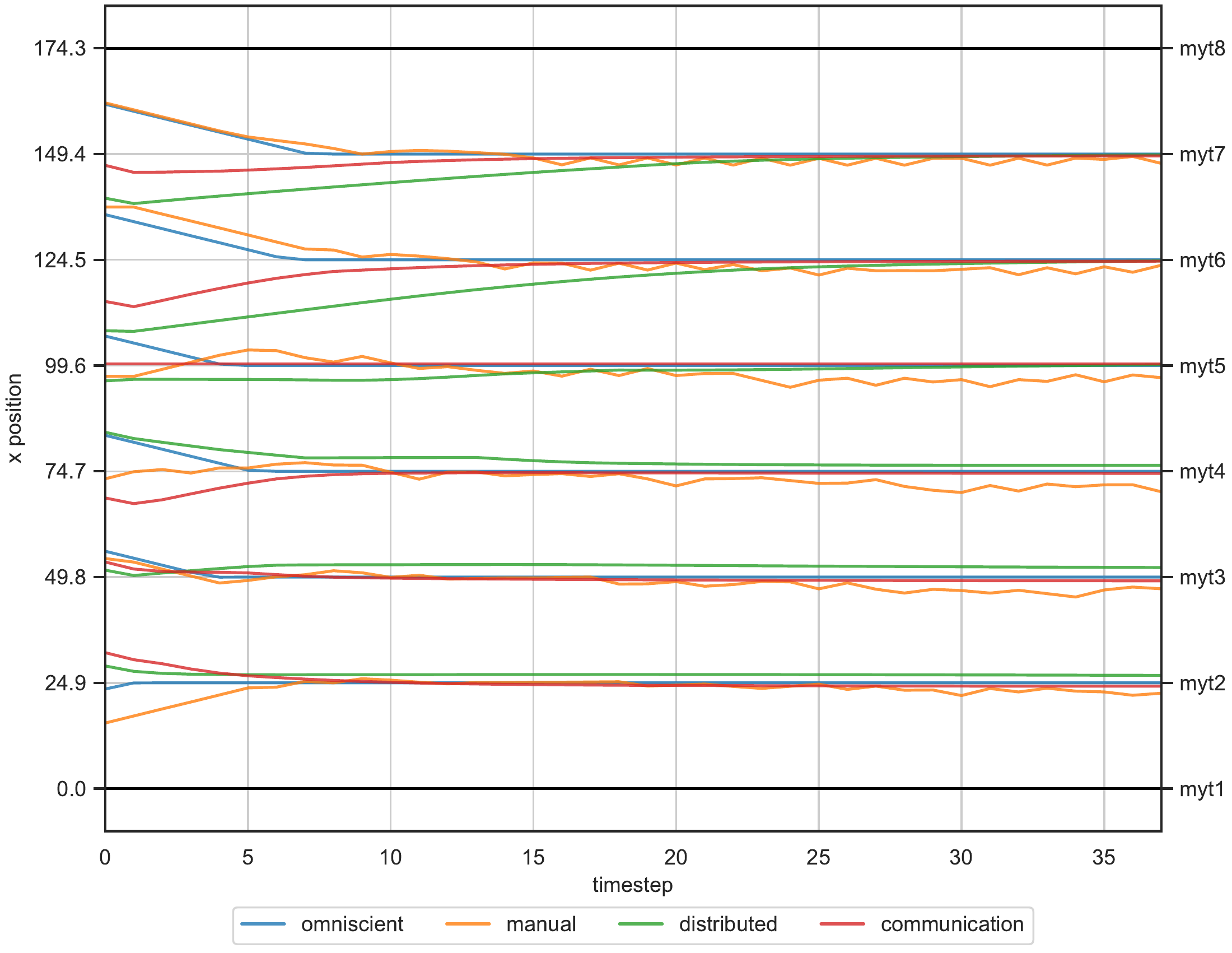}%
	\caption[Evaluation of the trajectories obtained with 8 agents.]{Comparison of 
		trajectories, of a single simulation, generated using four controllers: the 
		expert, the manual and the two learned from \texttt{net-d15} and 
		\texttt{net-c15} .}
	\label{fig:net-c15traj1}
\end{figure}
In Figure \ref{fig:net-c15traj1} is visualised a sample simulation that shows on 
the y-axis position of each agent, while on the x-axis the simulation time steps. 
The agents moved using the manual controller, when have almost approached the 
target, they start to oscillate.
those that use the distributed controller, even if in the initial configuration are far 
from the target, they are able to reach the goal. Instead, the communication 
controller has a more promising behaviour: the convergence is faster than before.

\begin{figure}[!htb]
	\begin{center}
		\begin{subfigure}[h]{0.49\textwidth}
			\centering
			\includegraphics[width=.9\textwidth]{contents/images/net-d15/position-overtime-omniscient}%
			\caption{Expert controller trajectories.}
		\end{subfigure}
		\hfill
		\begin{subfigure}[h]{0.49\textwidth}
			\centering
			\includegraphics[width=.9\textwidth]{contents/images/net-d15/position-overtime-learned_distributed}
			\caption{Distributed controller trajectories.}
		\end{subfigure}
	\end{center}
	\begin{center}
		\begin{subfigure}[h]{0.49\textwidth}
			\centering			
			\includegraphics[width=.9\textwidth]{contents/images/net-d15/position-overtime-manual}%
			\caption{Manual controller trajectories.}
		\end{subfigure}
		\hfill
		\begin{subfigure}[h]{0.49\textwidth}
			\centering
			\includegraphics[width=.9\textwidth]{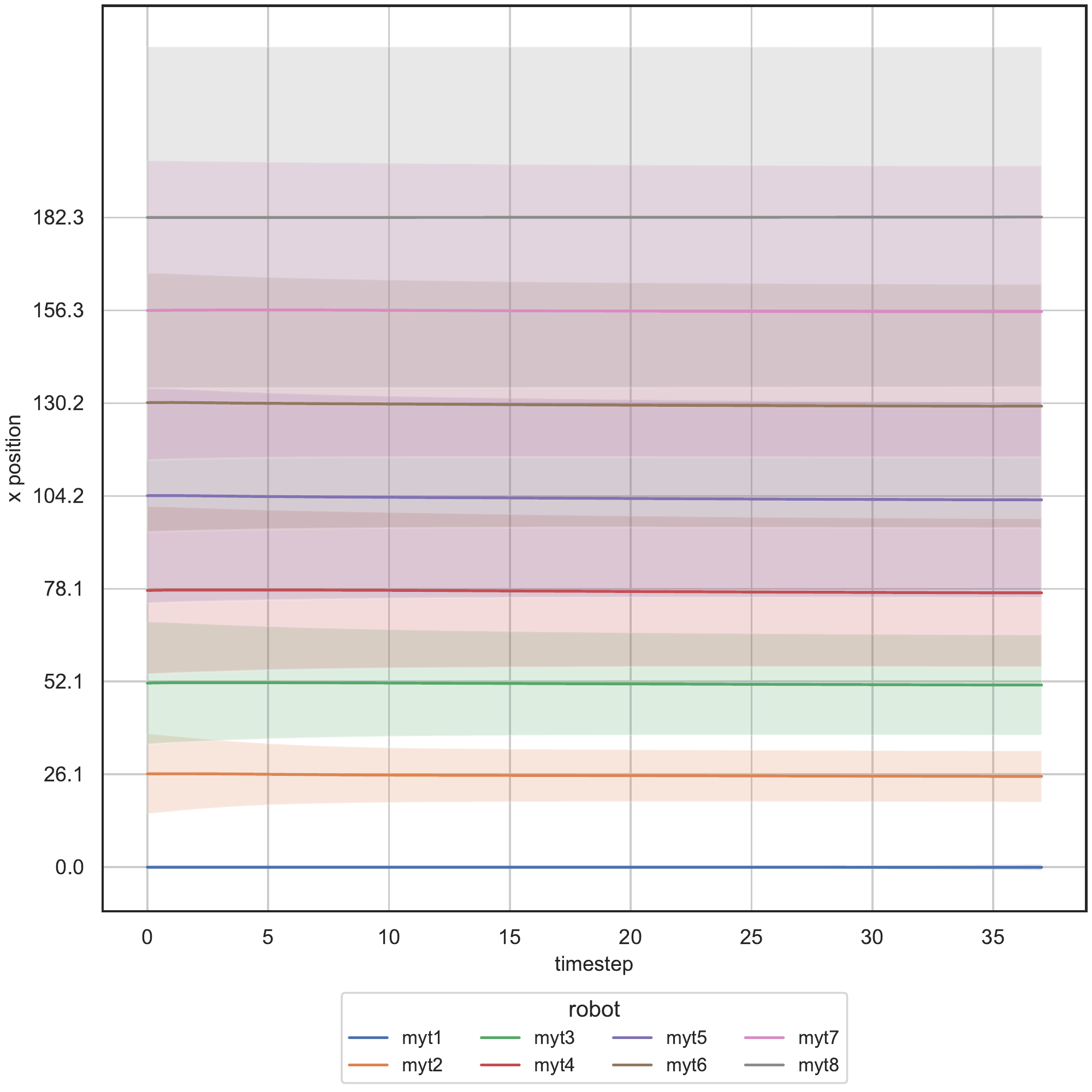}
			\caption{Communication controller trajectories.}
		\end{subfigure}
	\end{center}
	\vspace{-0.5cm}
	\caption[Evaluation of the trajectories learned by \texttt{net-c15}]{Comparison 
	of trajectories, of all the simulation runs, generated using four controllers: the 
	expert, the manual and the two learned from 
	\texttt{net-d15} and \texttt{net-c15}.}
	\label{fig:net-c15traj}
\end{figure}

In Figure \ref{fig:net-c15traj} are shown the trajectories obtained employing the 
four controllers.  On average, all the robots seems to approach the desired 
position, some with less and some with more time steps, but is still difficult 
to demonstrate improvement over the previous approach due to the deviation in 
the graph caused by the different target positions.

Even the analysis of the evolution of the control over time in Figure 
\ref{fig:net-c15control} is not able to provide further considerations.
\begin{figure}[!htb]
	\begin{center}
		\begin{subfigure}[h]{0.35\textwidth}
			\includegraphics[width=\textwidth]{contents/images/net-d15/control-overtime-omniscient}%
			\caption{Expert control.}
		\end{subfigure}
		\hspace{1cm}
		\begin{subfigure}[h]{0.35\textwidth}
			\includegraphics[width=\textwidth]{contents/images/net-d15/control-overtime-learned_distributed}
			\caption{Distributed control.}
		\end{subfigure}
	\end{center}
	\begin{center}
		\begin{subfigure}[h]{0.35\textwidth}			
			\includegraphics[width=\textwidth]{contents/images/net-d15/control-overtime-manual}%
			\caption{Manual control.}
		\end{subfigure}
		\hspace{1cm}
		\begin{subfigure}[h]{0.35\textwidth}
			\includegraphics[width=\textwidth]{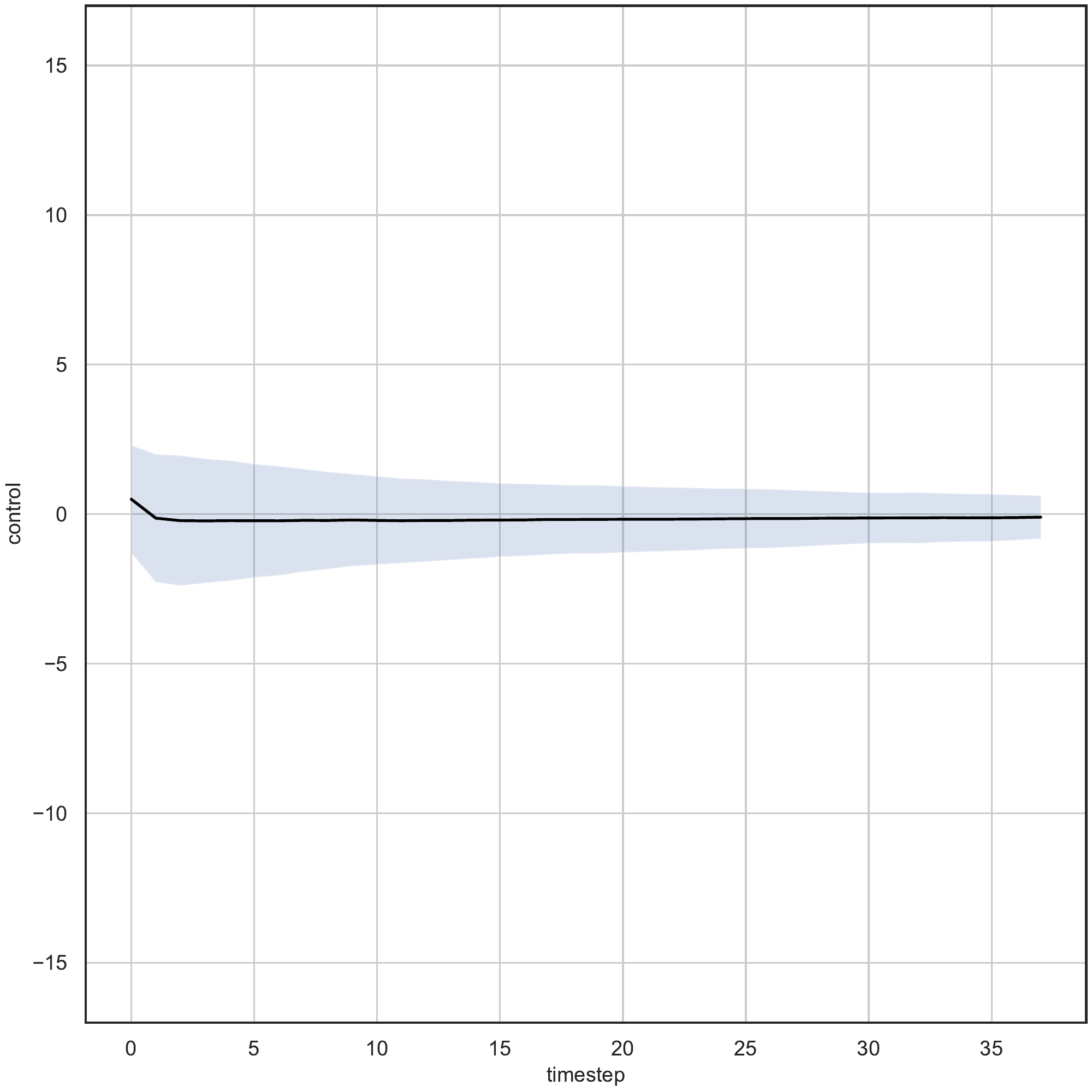}
			\caption{Communication control.}
		\end{subfigure}
	\end{center}
\vspace{-0.5cm}
	\caption[Evaluation of the control decided by \texttt{net-c15}.]{Comparison of 
	the output control decided using four controllers: the 
	expert, the manual and the two learned from \texttt{net-d15} and 
	\texttt{net-c15}.}
	\label{fig:net-c15control}
\end{figure}

\bigskip
In Figure \ref{fig:net-c15responseposition} is displayed the behaviour of a robot 
located between other two stationary agents, showing the response of the 
controllers, 
on the y-axis, by varying the position of the moving robot, visualised on the 
x-axis.  This time the trend of the curve obtained from the communication 
controller is 
different than the desired one.
\begin{figure}[!htb]
	\centering
	\includegraphics[width=.45\textwidth]{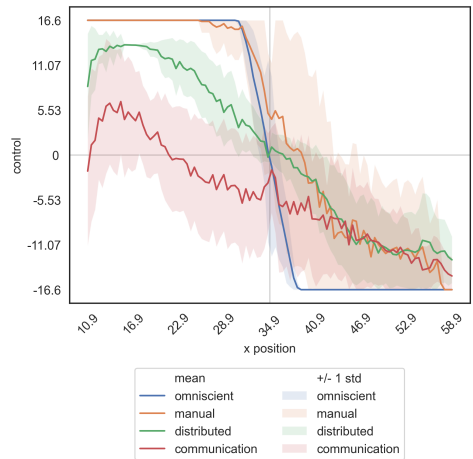}%
	\caption{Response of \texttt{net-c15} by varying the initial position.}
	\label{fig:net-c15responseposition}
\end{figure}

Finally, in Figure \ref{fig:net-c15distance} is presented a metric that measures the 
absolute distance of each robot from the target over time.
Unlike the non-optimal performance obtained with the distributed controller, in 
which the robots in the final configuration are located on average at about 
$5$cm from the target, the distance from goal of the communication 
controller is far better but similar to that obtained with the manual. Even if the 
new approach need more time to converge, at the end the robots are $2$ away 
from the goal.
\begin{figure}[H]
	\centering
	\includegraphics[width=.65\textwidth]{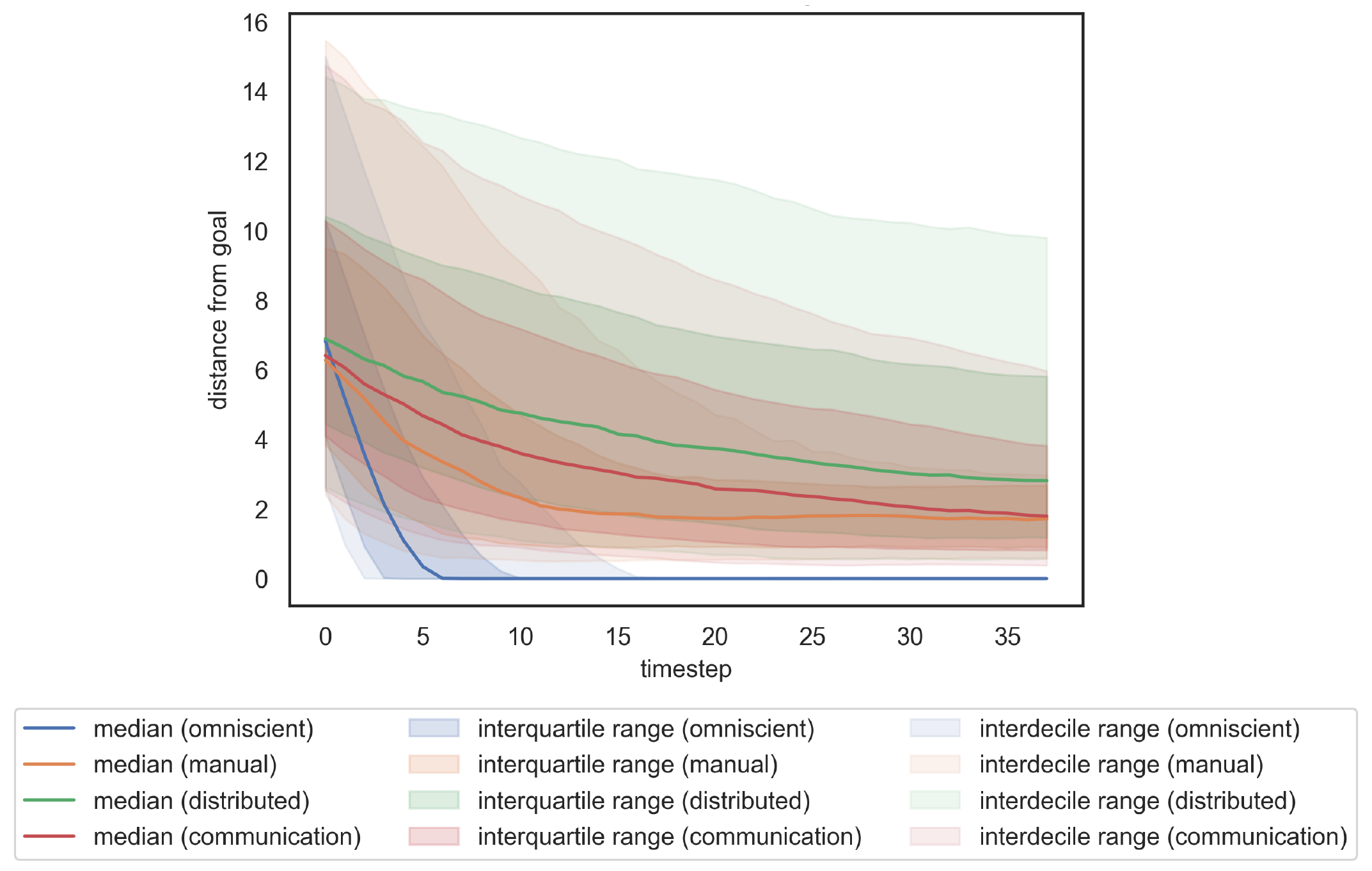}%
	\caption[Evaluation of \texttt{net-c15} distances from goal.]{Comparison of 
		performance in terms of distances from goal obtained using four controllers: 
		the expert, the manual and the two learned from \texttt{net-d15} and 
		\texttt{net-c15}.}
	\label{fig:net-c15distance}
\end{figure}

\paragraph*{Results using variable agents}
We conclude the experiments on task 1 presenting the results obtained using 
variable number of agents, in particular, in Figure \ref{fig:commlossNvar} are 
summarised the performance in terms of loss, as before we used blue, orange and 
green lines to represent respectively average gaps of $8$cm, $13$cm 
and variable.
Using the new approach we observe from the trend of the curves that in general 
the losses are decreased and for the network it is easier to perform the task by 
using a smaller gap.
\begin{figure}[!htb]
	\begin{center}
		\begin{subfigure}[h]{0.49\textwidth}
			\centering
			\includegraphics[width=.7\textwidth]{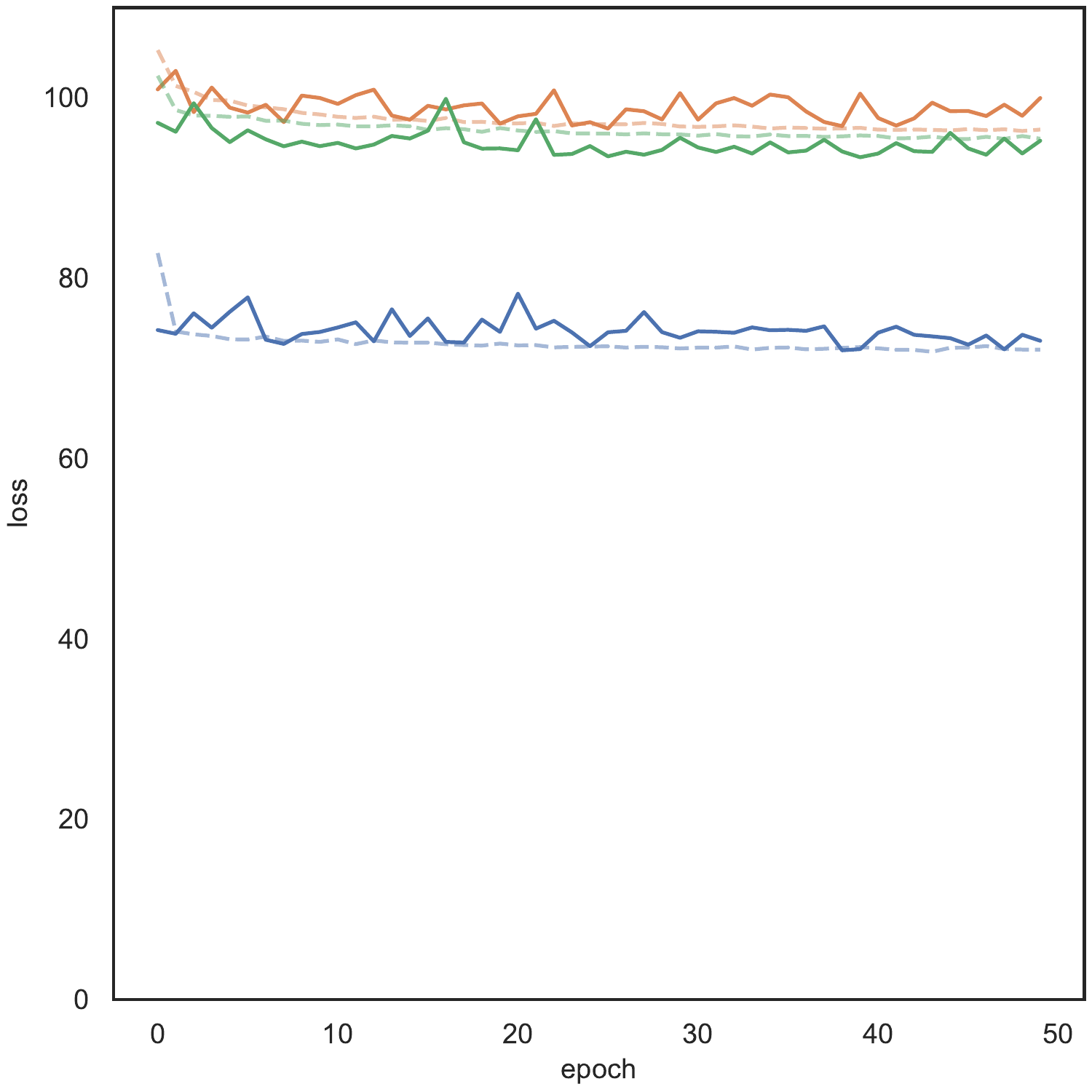}
			\caption{Distributed approach.}
		\end{subfigure}
		\hfill
		\begin{subfigure}[h]{0.49\textwidth}
			\centering
			\includegraphics[width=.7\textwidth]{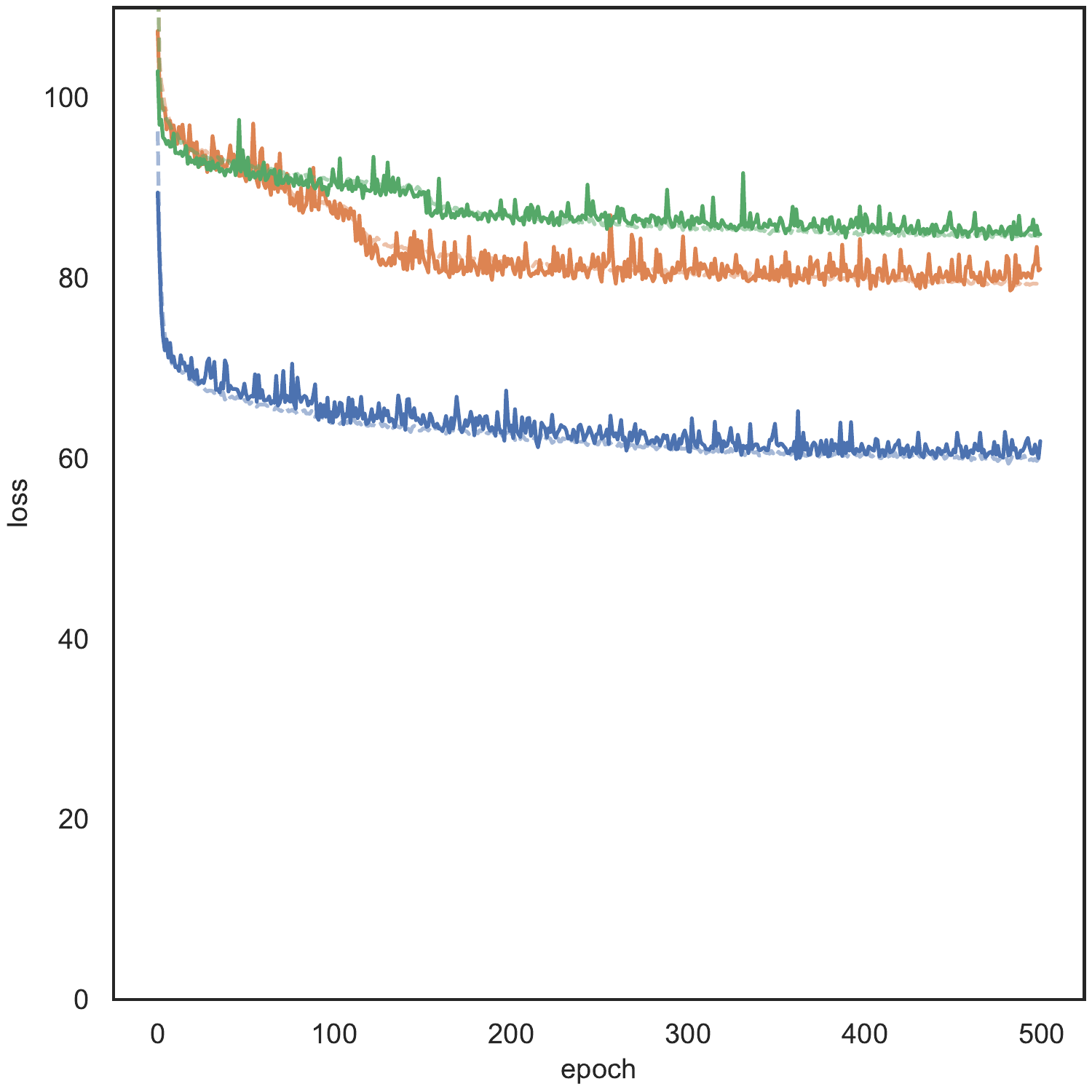}
			\caption{Distributed approach with communication.}
		\end{subfigure}	
	\end{center}
	\vspace{-0.5cm}
	\caption[Comparison of the losses of the models that use viariable 
	agents.]{Comparison of the losses of the models that use variable agents 
	and gaps.}
	\label{fig:commlossNvar}
\end{figure}

Considering, as before, the more complex case, for instance the one with variable 
average gap, in Figure \ref{fig:net-c18r2} are visualised the \ac{r2} of 
the manual and the learned controllers, with and without communication.
A small improvement in the new approach is confirmed by the increase in the 
coefficient \ac{r2} from $0.23$ to $0.30$. 
\begin{figure}[!htb]
	\begin{center}
		\begin{subfigure}[h]{0.49\textwidth}
			\includegraphics[width=\textwidth]{contents/images/net-d18/regression-net-d18-vs-omniscient}%
		\end{subfigure}
		\hfill\vspace{-0.5cm}
		\begin{subfigure}[h]{0.49\textwidth}
			\includegraphics[width=\textwidth]{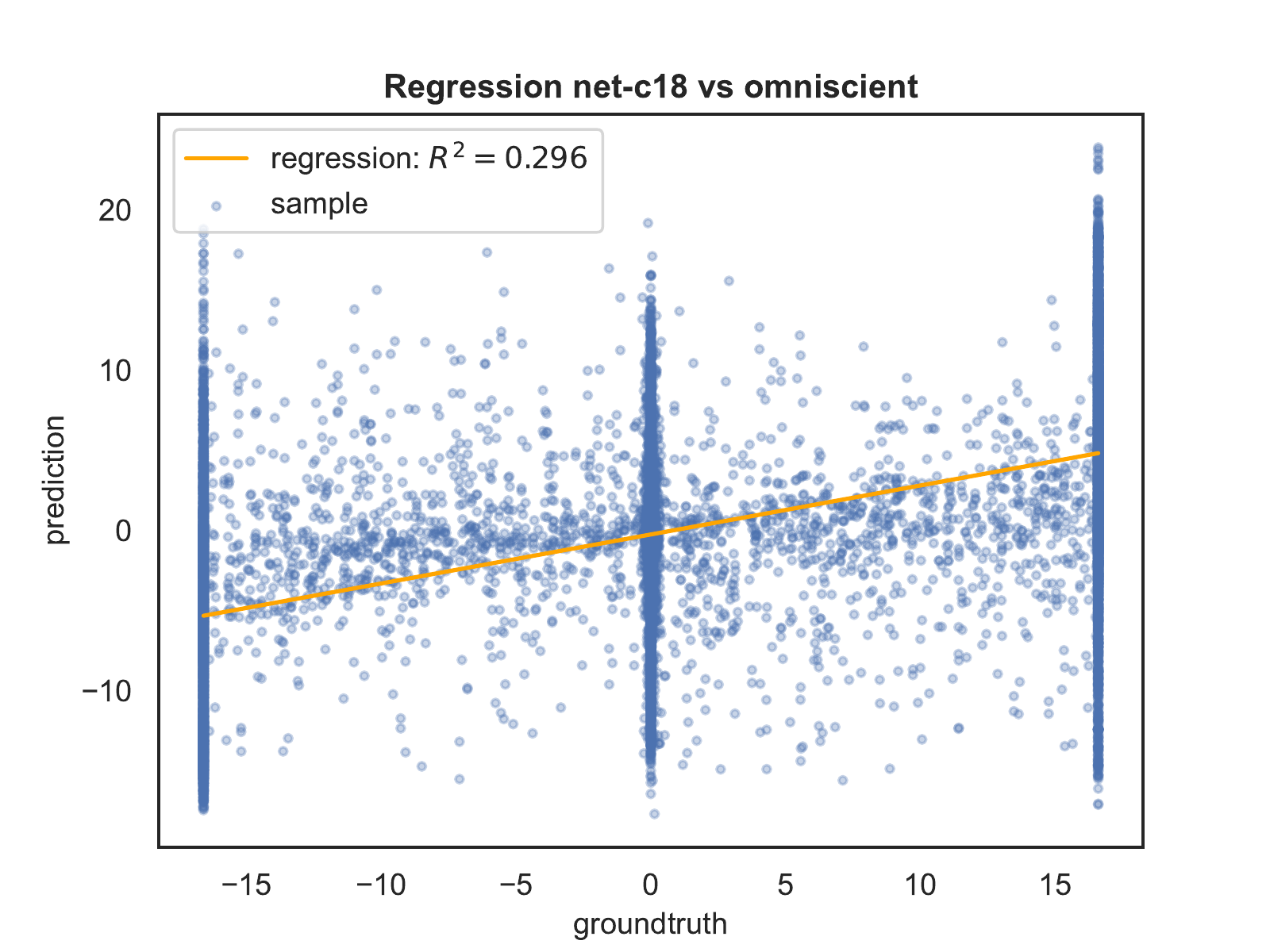}%
		\end{subfigure}
	\end{center}
	\caption[Evaluation of the \ac{r2} coefficients of 
	\texttt{net-c18}.]{Comparison 
		of the \ac{r2} coefficients of the manual and the controllers learned from 
		\texttt{net-d18} and \texttt{net-c18}, with respect to the omniscient 
		one.}
	\label{fig:net-c18r2}
\end{figure}

In this experiment is difficult to demonstrate improvement over the previous 
approach through the trajectories plots, as well as by the evolution of the control, 
due to the variable average gap. For this reason we show, in Figure 
\ref{fig:net-c18traj1} is shown a comparison of the trajectories obtained 
for a sample simulation.
\begin{figure}[!htb]
	\centering
	\includegraphics[width=.75\textwidth]{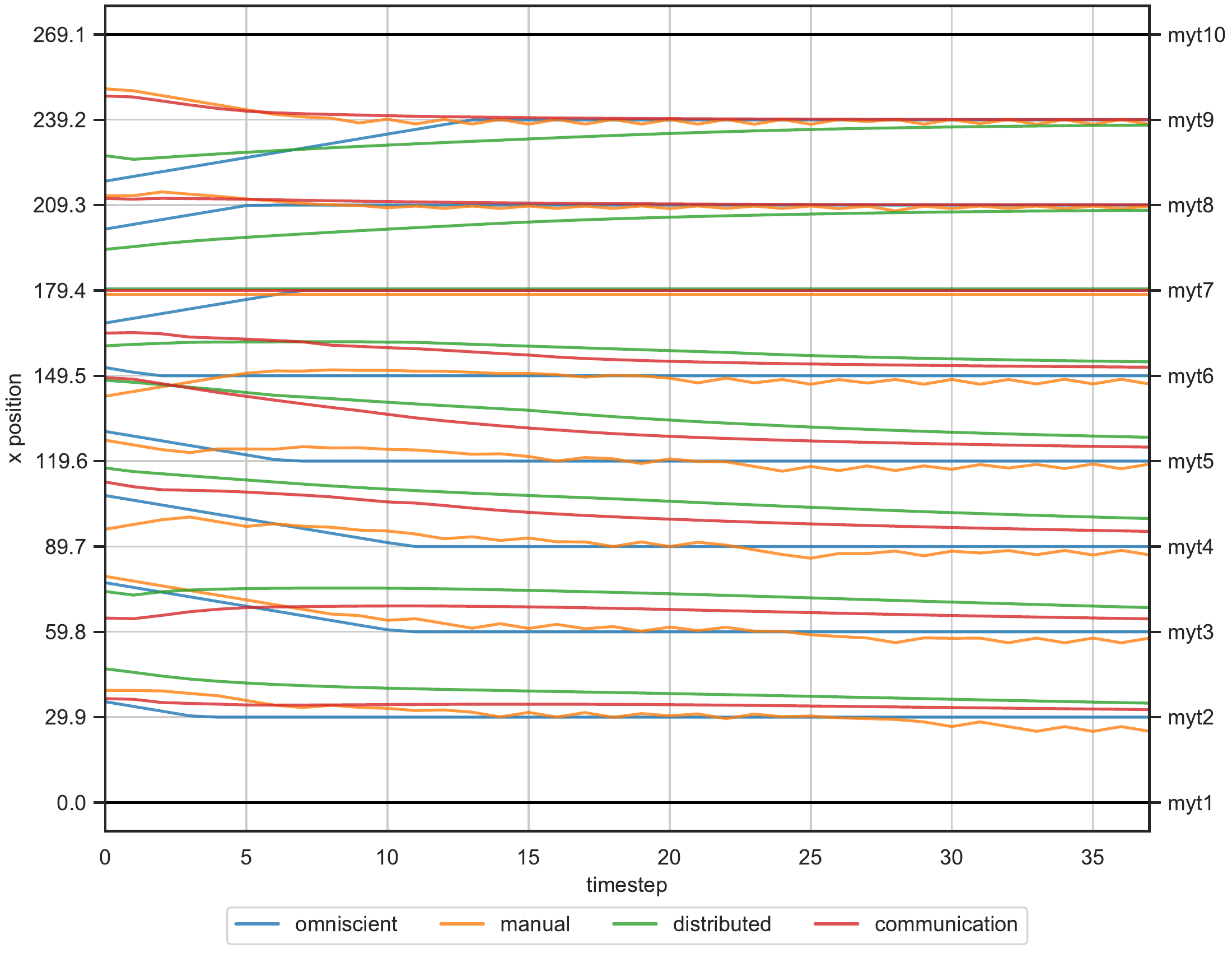}%
	\caption[Evaluation of the trajectories obtained with variable 
	agents.]{Comparison of trajectories, of a single simulation, generated using 
	four controllers: the expert, the manual and the two learned from 
	\texttt{net-d18} and \texttt{net-c18}.}
	\label{fig:net-c18traj1}
\end{figure}  
In this example, the simulation presents 10 agents. They are always able to reach 
the target when moved using the omniscient controller, this is not always true for 
others.
When they use the manual controller, the same oscillation issue occurs in 
proximity  to the goal. The distributed controller is certainly the slowest, and after 
38 time steps not all robots are in the correct position.
Instead, adding the communication speeds up the achievement of the goal. In 
fact, this controller seems to behave in most cases better than the previous two, 
although with results that are still lower than those obtained by the expert.

We move on analysing in Figure \ref{fig:net-c18responseposition} the behaviour 
of a robot located between other two stationary agents which are already in the 
correct position, showing the response of the controllers by varying the position 
of the moving robot. 
\begin{figure}[!htb]
	\centering
	\includegraphics[width=.45\textwidth]{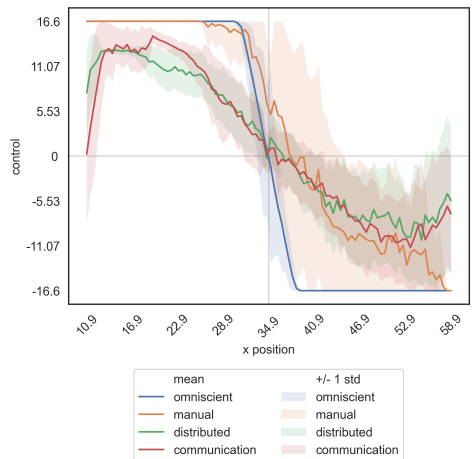}%
	\caption{Response of \texttt{net-c18} by varying the initial position.}
	\label{fig:net-c18responseposition}
\end{figure}
As expected, the output is a high value, positive or negative respectively when the 
robot is close to an obstacle on the left or on the right, or it is close to $0$ when 
the distance from right and left is equal.

Finally, in terms of absolute distance of each robot from the target, in Figure 
\ref{fig:net-c18distance}  
\begin{figure}[H]
	\centering
	\includegraphics[width=.65\textwidth]{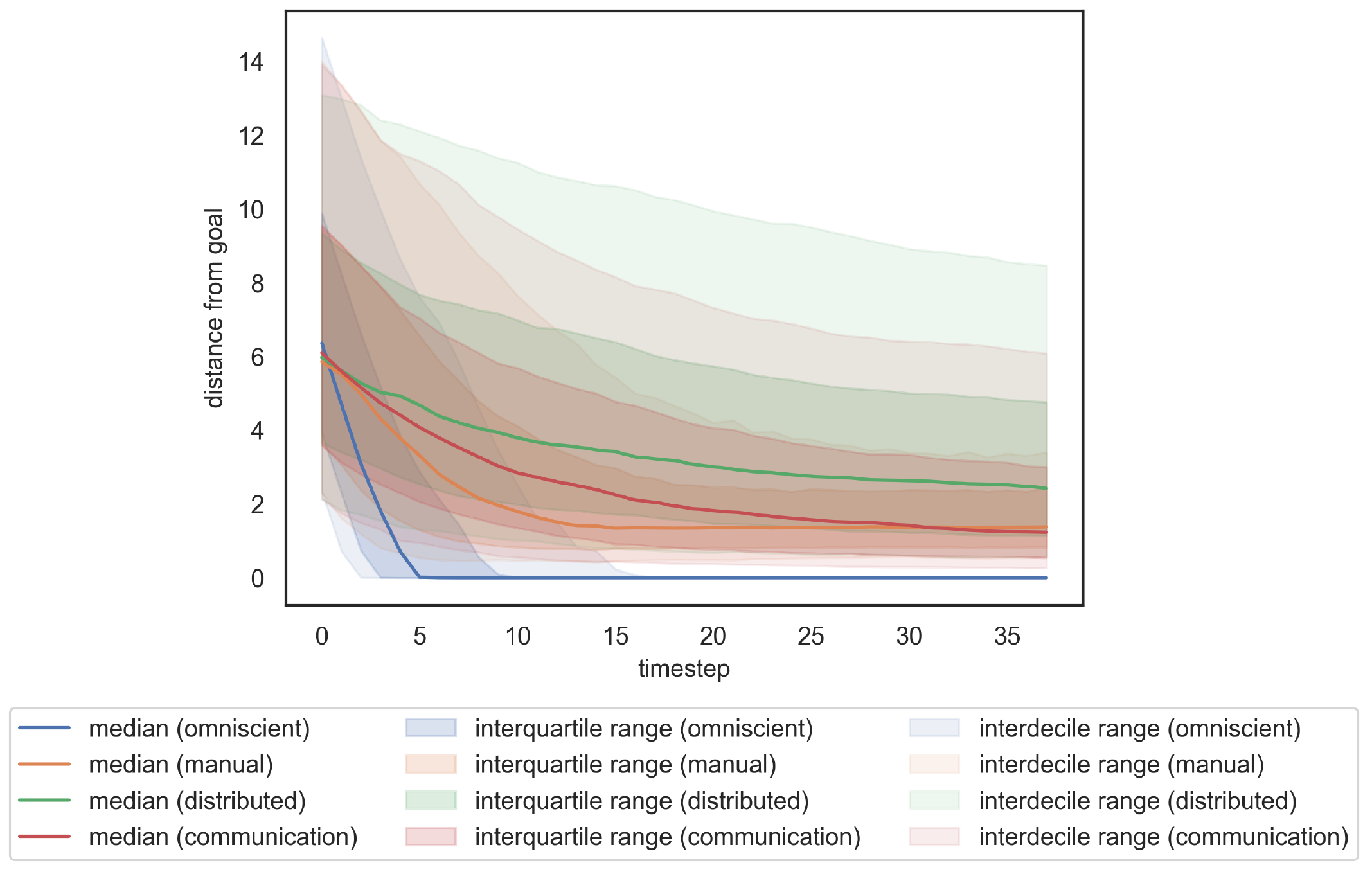}%
	\caption[Evaluation of \texttt{net-c18} distances from goal.]{Comparison of 
		performance in terms of distances from goal obtained using four controllers: 
		the expert, the manual and the two learned from \texttt{net-d18} and 
		\texttt{net-c18}.}
	\label{fig:net-c18distance}
\end{figure}

\noindent
is shown that, exploiting the communication, the robots in the final configuration 
are closer to the target than those moved using the distributed or even the 
manual controller.

\paragraph*{Summary}
To sum up, we finally show in the figures below the losses of the trained 
models as the number of agents vary for each gap. In all the figures are 
represented the losses of the models that use $5$, $8$ and variable agents, 
respectively in blue, orange and green.
In case of an \texttt{avg\_gap} of $8$cm, with or without communication, 
the model trained using a minor number of agents, as expected, has a lower loss.
Instead, very similar are the losses obtained in case of $8$ or variable agents, in 
which the model with less robots is still the better.
In general, using the communication has improved the performance.
\begin{figure}[!htb]
	\begin{center}
		\begin{subfigure}[h]{0.32\textwidth}
			\includegraphics[width=\textwidth]{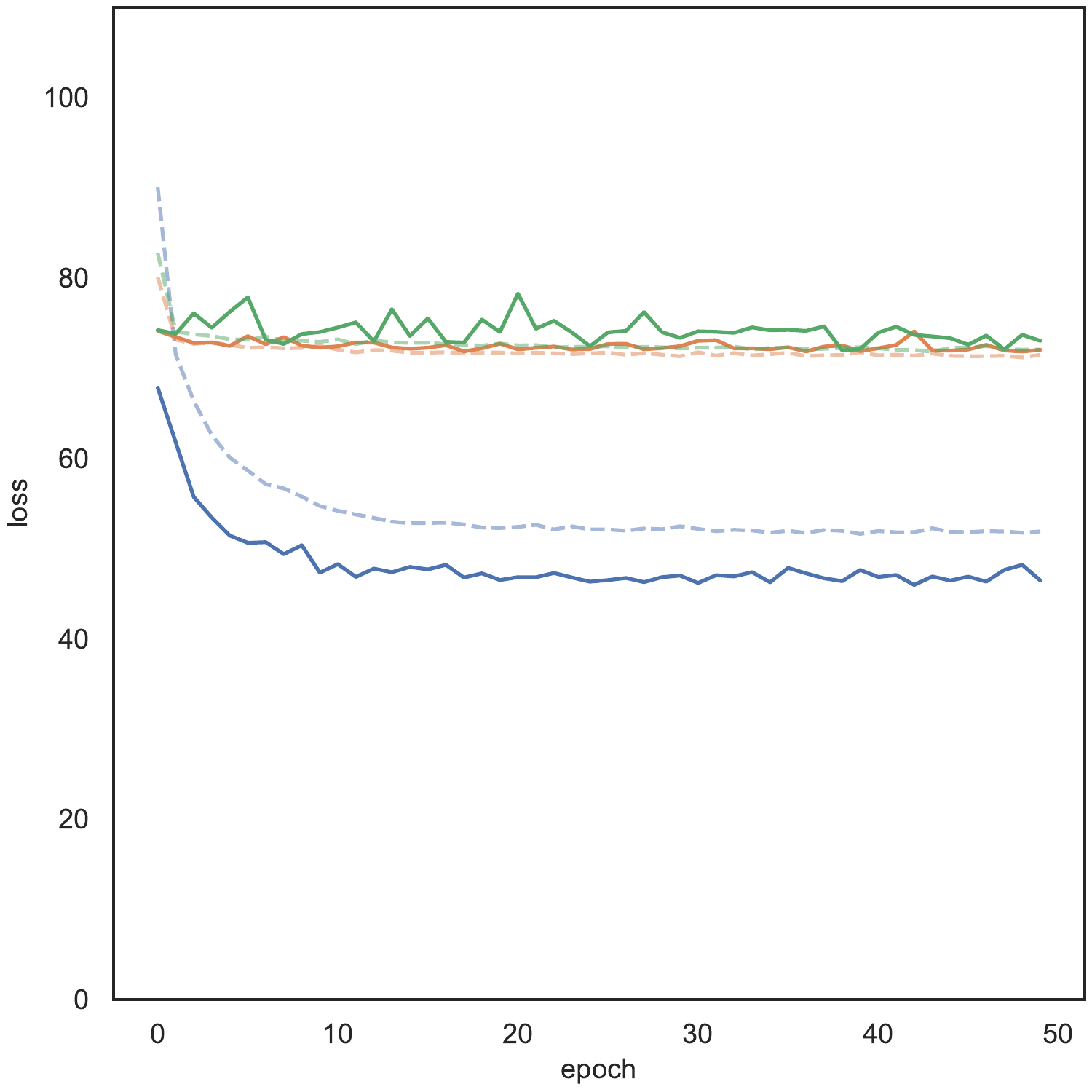}
			\caption{\texttt{avg\_gap} of $8$cm.}
		\end{subfigure}
		\hfill
		\begin{subfigure}[h]{0.32\textwidth}
			\includegraphics[width=\textwidth]{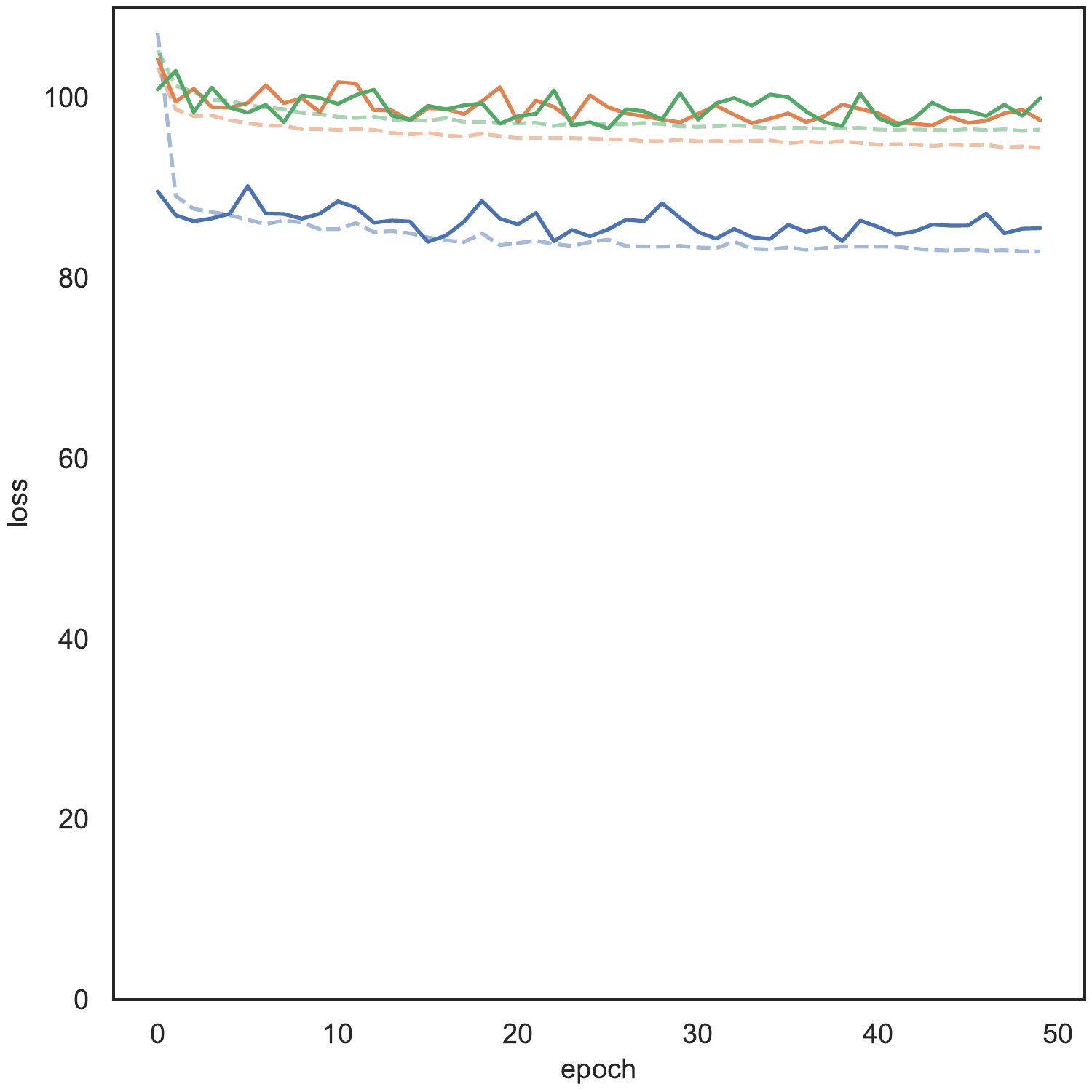}%
			\caption{\texttt{avg\_gap} of $20$cm.}
		\end{subfigure}
		\hfill
		\begin{subfigure}[h]{0.32\textwidth}
			\includegraphics[width=\textwidth]{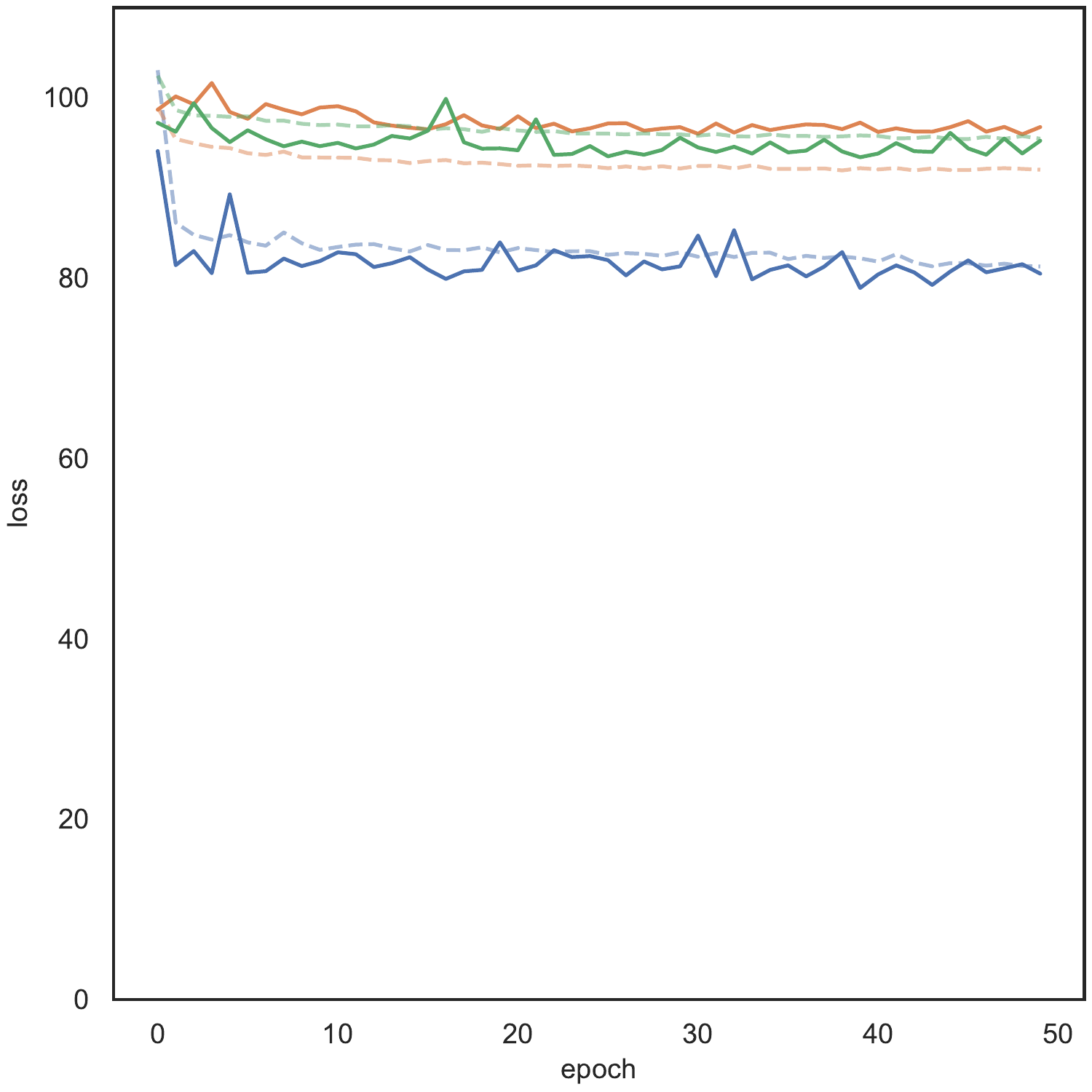}
			\caption{\texttt{avg\_gap} variable.}
		\end{subfigure}
	\end{center}
	\vspace{-0.5cm}
	\caption[Losses summary of the second set of experiments (no 
	communication).]{Comparison of the losses of the model trained without 
		communication, by varying the number of agents for the three gaps.}
	\label{fig:distlossgapsexte}
\end{figure}
\begin{figure}[!htb]
	\begin{center}
		\begin{subfigure}[h]{0.32\textwidth}
			\includegraphics[width=\textwidth]{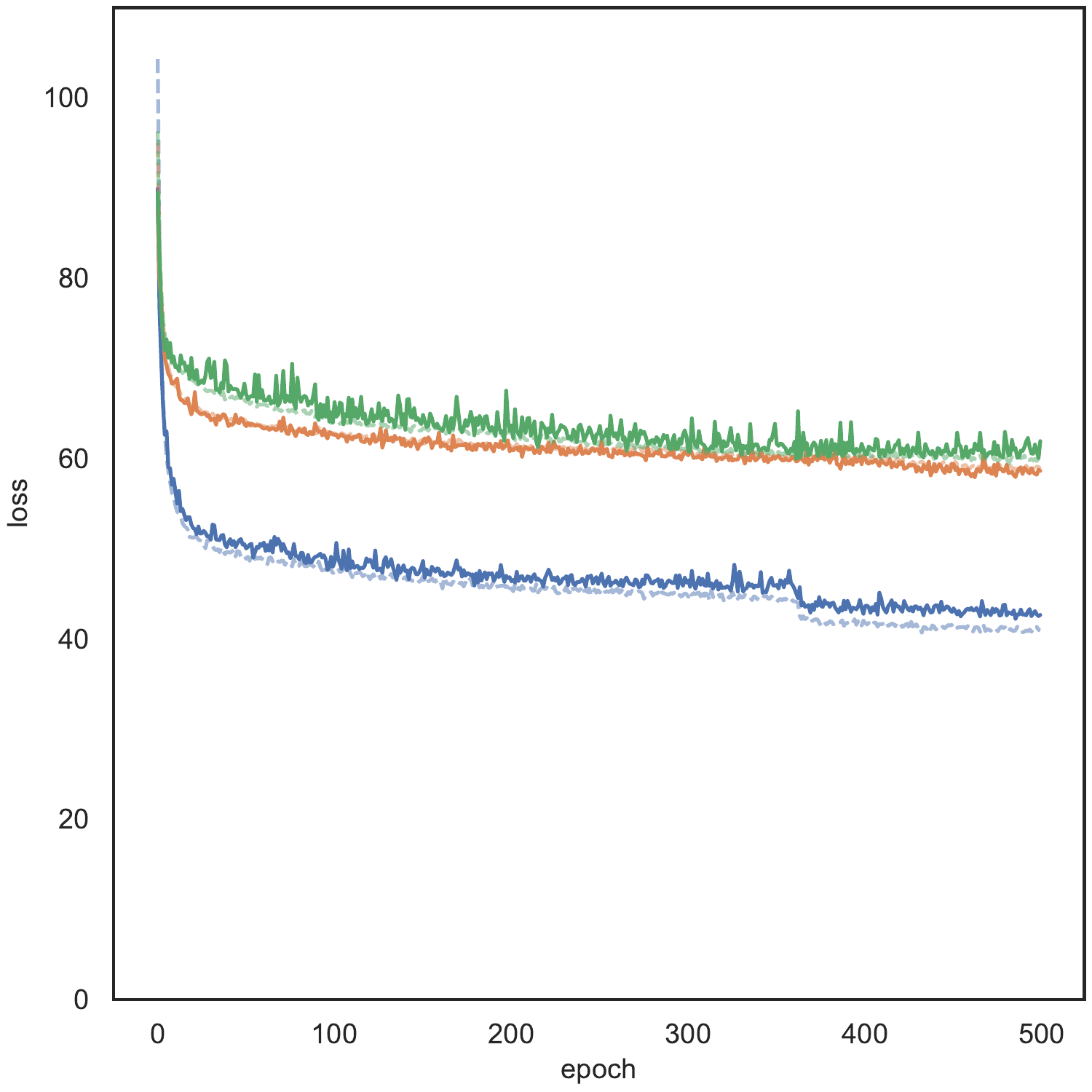}%
			\caption{\texttt{avg\_gap} of $8$cm.}
		\end{subfigure}
		\hfill
		\begin{subfigure}[h]{0.32\textwidth}
			\includegraphics[width=\textwidth]{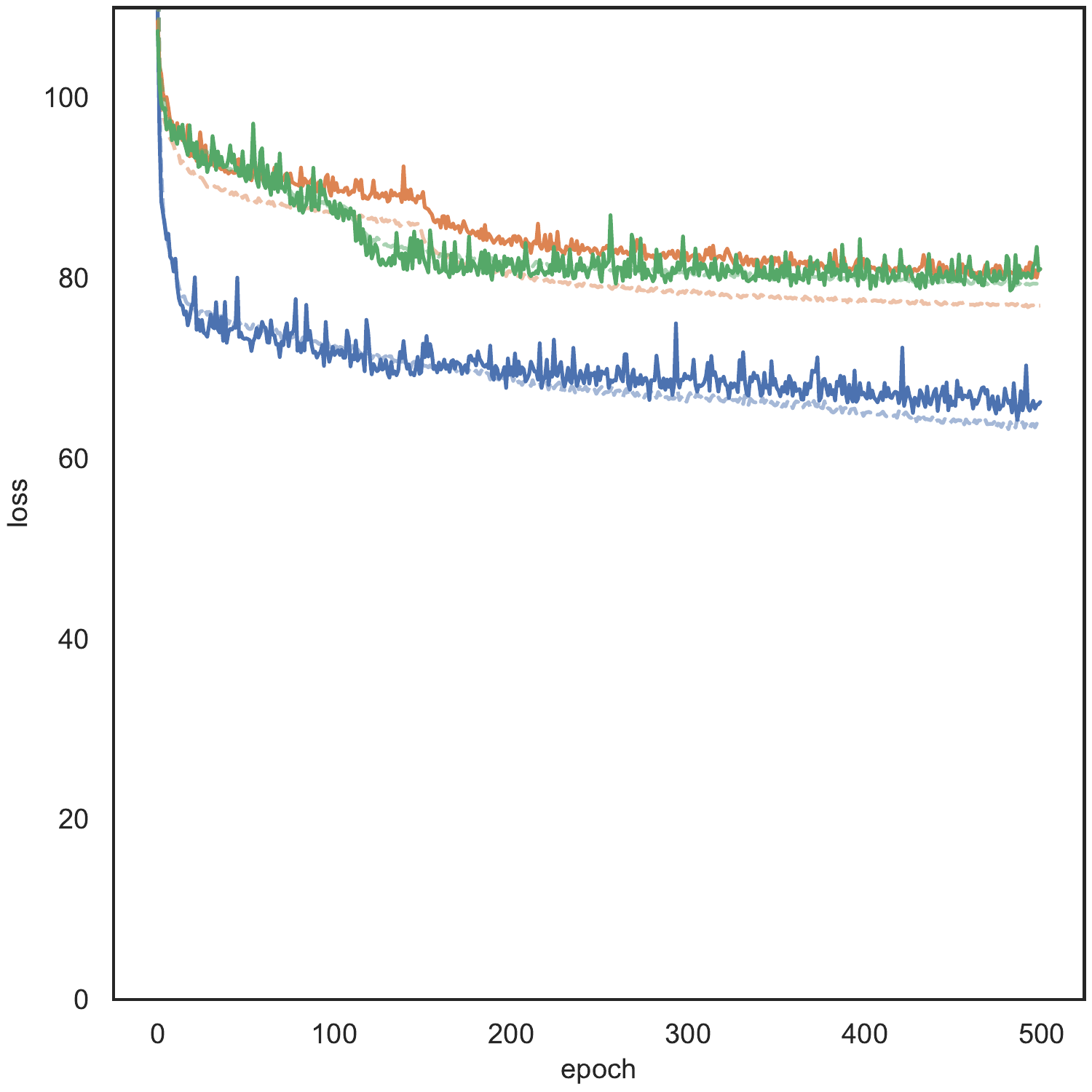}%
			\caption{\texttt{avg\_gap} of $20$cm.}
		\end{subfigure}
		\hfill
		\begin{subfigure}[h]{0.32\textwidth}
			\includegraphics[width=\textwidth]{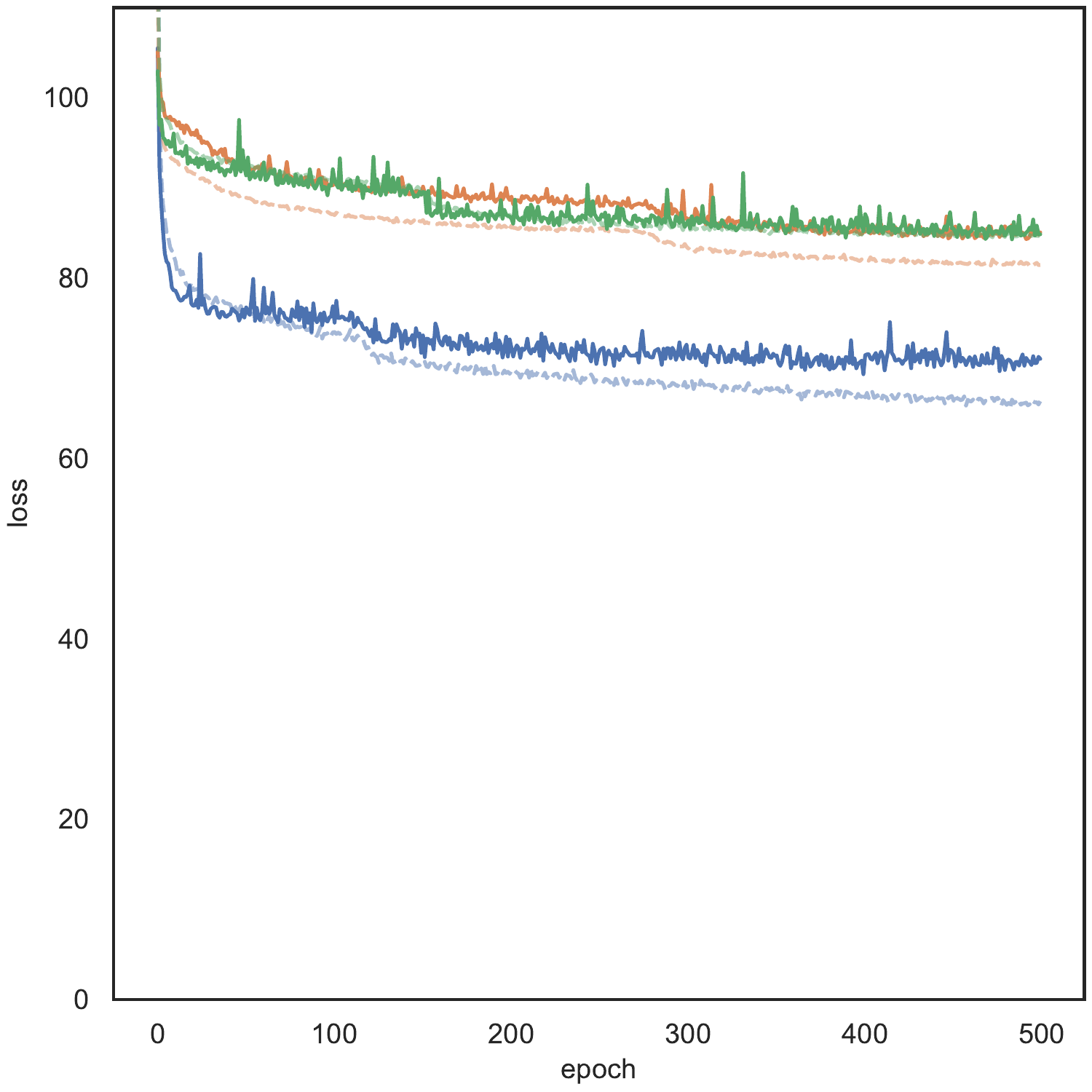}
			\caption{\texttt{avg\_gap} variable.}
		\end{subfigure}
	\end{center}
	\vspace{-0.5cm}
	\caption[Losses summary of the second set of experiments 
	(communication).]{Comparison of the losses of the model trained with 
		communication, by varying the number of agents for the three gaps.}
	\label{fig:commlossexte81324}
\end{figure}

\subsubsection{Experiment 3: increasing number of agents}
\label{subsubsec:task1-exp-comm-3}
The last group of experiments focuses on the scalability properties of a 
multi-agent system, showing the behaviour of the network trained using 
\texttt{all\_sensors} input, variable gaps and number of agents, applied on 
simulations with a higher number of robots, from 5 up to 50.
\begin{figure}[!htb]
	\centering
	\includegraphics[width=.55\textwidth]{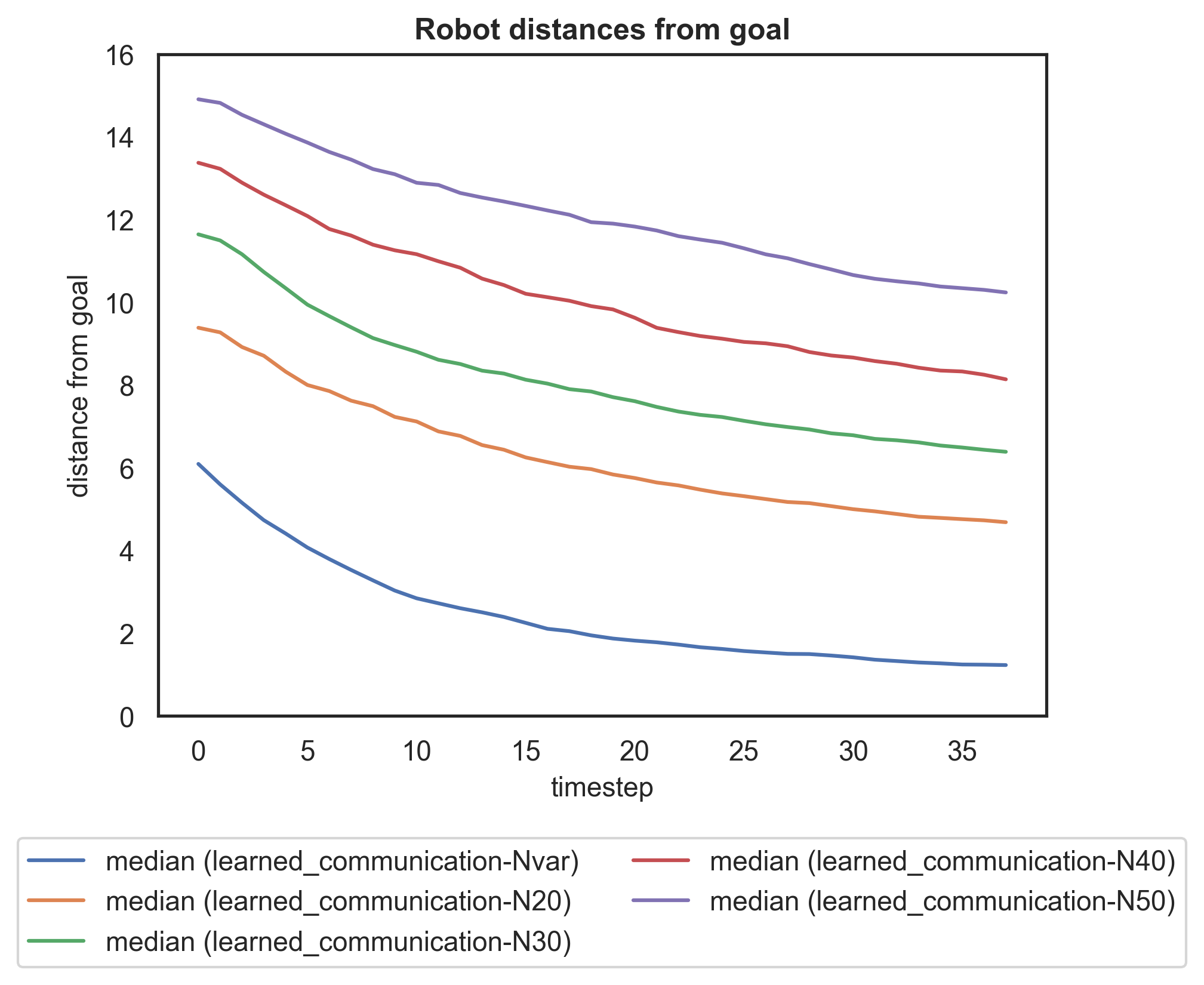}%
	\caption[Evaluation of distances from goal for a high number of 
robots.]{Comparison of performance in terms of distances from goal obtained 
	on simulations with an increasing number of robots.}
	\label{fig:distdistrcomm}
\end{figure}
In Figure \ref{fig:distdistrcomm} is visualised, for 5 different experiments, the 
absolute distance of each robot from the target over time. This value is averaged 
on all robots among all the simulation runs. 
The results obtained using the communication approach are a bit better then the 
previous, even if increasing the number of robots still produces a slowdown in 
reaching the correct positions.
In the final configuration, the robots are on average at about $1$cm from the 
from the goal position, instead of $3$ as before. Increasing the number of agents, 
first to 20, then 30, 40 and finally 50, the robots are in the worst case $10$cm 
from the target, $3$cm closer than without using communication.

\subsubsection{Remarks}
\label{subsubsec:remarks-task1-comm}

In this section, we have shown that using a distributed controller learned by 
imitating an expert and exploiting a communication protocol among the 
agents, it is possible to obtain results more or less comparable to those reached 
employing an expert controller, albeit a bit worse when using variable number of 
agents and gaps.

\section{Task 2: Colouring the robots in space}
\label{sec:task2}

The second scenario tackles another multi-agent coordination task, assuming that 
the agents are divided into groups, their purpose is to colour themselves, by 
turning on their top \ac{rgb} \ac{led}, depending on their group membership. 
As for the previous task, the problem can be solved performing imitation 
learning, but the role of communication is fundamental. In fact, what makes the 
difference are not the distances perceived by the robot sensors but the messages 
exchanged between the agents, which are they only mean to determine their 
order. 
In this scenario, the two ``dead'' robots play an important role: they always 
communicate a message that indicates that they are the only two agents that 
receive communication just from one side.

\subsection{Distributed approach with communication}
\label{subsec:task2-exp-comm}

\subsubsection{Experiment 1: variable number of agents}
\label{subsubsec:task2-exp-comm-1}
In this section, we explore the experiments carried out using the communication 
approach, in particular, examining the behaviour of the control learned from 9 
networks 
\begin{figure}[!htb]
	\centering
	\begin{tabular}{ccc}
		\toprule
		\textbf{Model} \quad & \textbf{\texttt{avg\_gap}} & \textbf{\texttt{N}}\\
		\midrule
		\texttt{net-v1}   &  $8$		 &	 $5$ \\
		\texttt{net-v2}   &  $20$		&	$5$ \\
		\texttt{net-v3}   &  variable   &    $5$\\
		\texttt{net-v4}   &  $8$		 &	  $8$ \\
		\texttt{net-v5}   & $20$   		&	 $8$ \\
		\texttt{net-v6}   &  variable	&	 $8$ \\
		\texttt{net-v7}   &  $ 8$		  &	 variable\\
		\texttt{net-v8}   &  $20$		 &	variable\\
		\texttt{net-v9}   &  variable	 &	variable\\
		\bottomrule
	\end{tabular}
	\captionof{table}[Experiments with variable agents and gaps 
	(communication).]{List of the experiments carried out using a variable number 
		of agents and of gaps.}
	\label{tab:modelcommt2}
\end{figure}

\noindent
based on different simulation runs that use a number of robots $N$ that 
can be fixed at $5$ or $8$ for the entire simulation, or even vary in the range $[5, 
10]$, and an \texttt{avg\_gap} that can be a fixed value in all the runs, chosen 
between $8$ or $20$, but also vary in the range $[5, 24]$. 
The objective of this set of experiments, summarised in Table 
\ref{tab:modelcommt2}, is to verify the robustness of the communication 
protocol and prove also the scalability of the network on the number of agents.

First of all, we show in Figure \ref{fig:t2lossallt} an overview of the train and 
validation losses obtained for these models.
\begin{figure}[!htb]
	\centering
	\includegraphics[width=.8\textwidth]{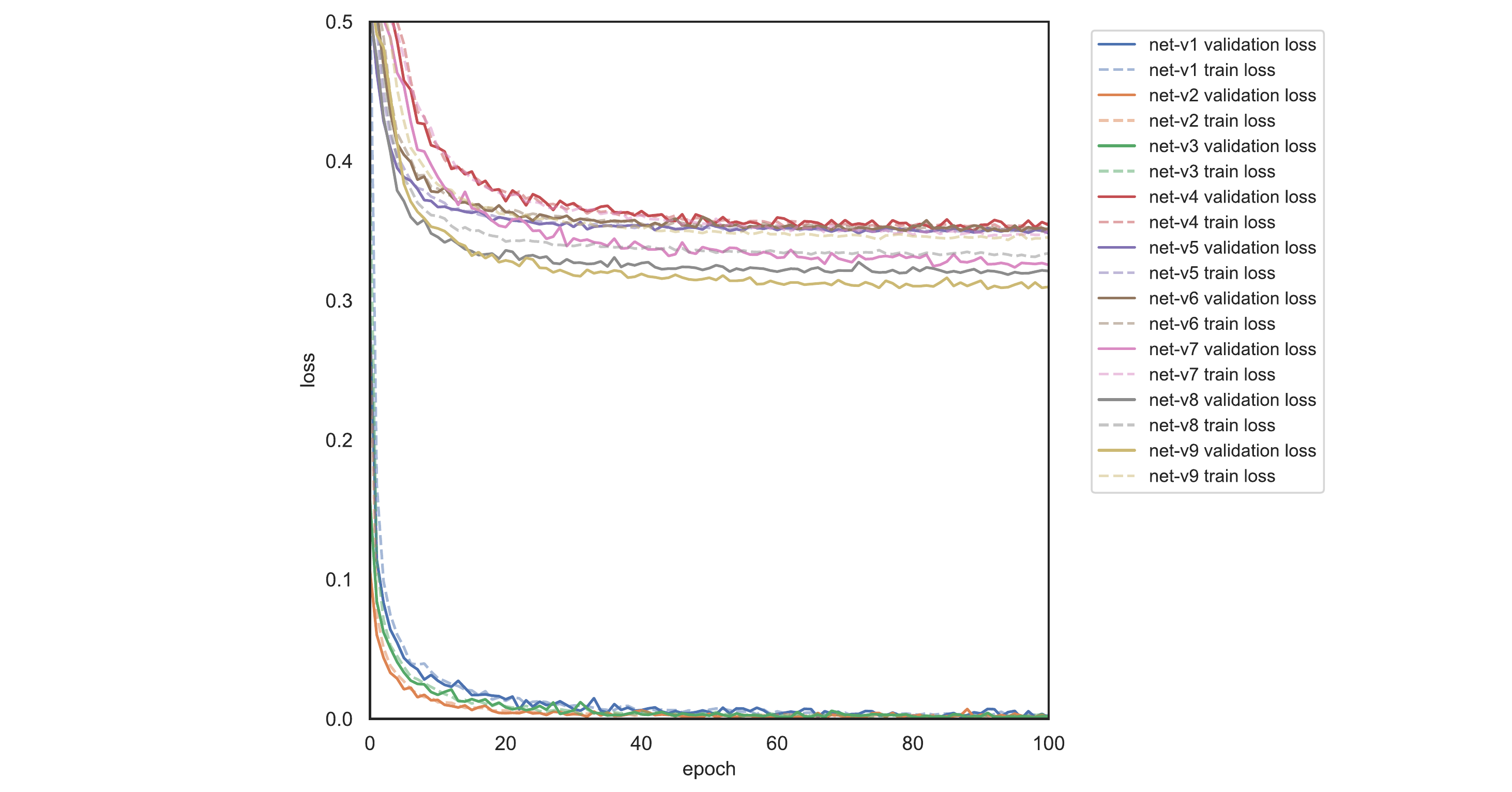}%
	\caption[Comparison of losses of the second set of experiments.]{Comparison 
		of the losses of the models carried out using a variable number of agents and 
		of average gap.}
	\label{fig:t2lossallt}
\end{figure}

\paragraph*{Results using 5 agents}

We start our examination by inspecting the behaviour of the network trained on 
simulations with variable average gap, i.e., \texttt{net-v3}, \texttt{net-v6} and 
\begin{figure}[!htb]
	\centering
	\includegraphics[width=.45\textwidth]{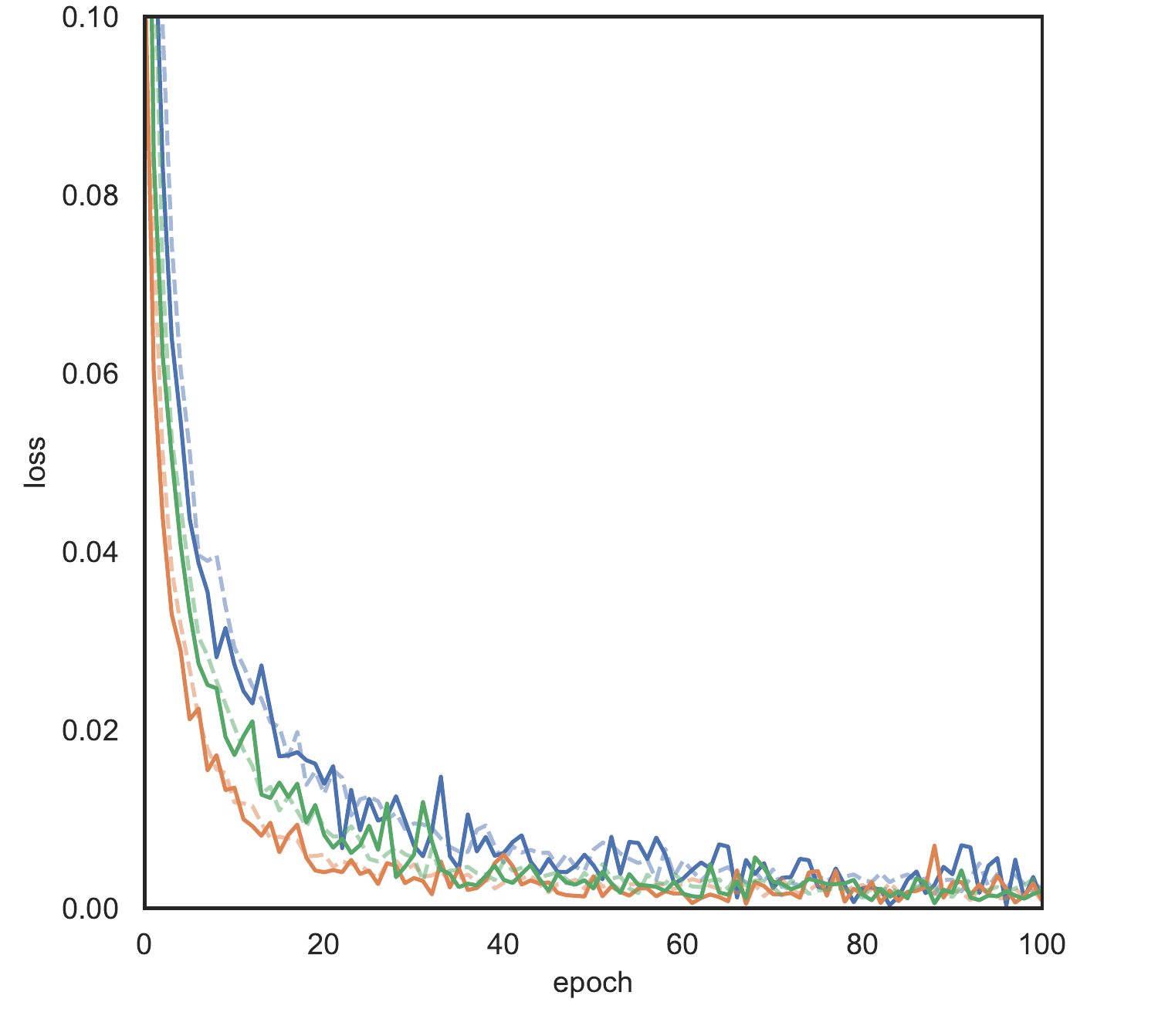}
	\caption[Comparison of the losses of the models that use $5$ 
	agents.]{Comparison of the losses of the models that use $5$ agents as 
		the gap varies.}
	\label{fig:commlossn5t2}
\end{figure}

\noindent
\texttt{net-v9}, summarising, in Figure \ref{fig:commlossn5t2}, the losses of 
these experiments in order to highlight the difference of performance using a gap 
that is first small, then large and finally variable, respectively represented by the 
blue, the orange and the green lines.
Clearly, in case of small gaps the network performs better, albeit slightly, as the 
agents are already close to the target.

Then, we move to explore the results of the experiments by showing in Figure 
\ref{fig:net-v3auc} the \ac{roc} curve of the model 
\cite[][]{fawcett2006introduction}, a visualisation of the performance of our 
classification model, in terms of \ac{tpr} versus \ac{fpr}, at all classification 
thresholds.
In particular we use the \ac{auc} to evaluate the classifier: by measuring the 
\ac{2d} area under the ROC curve, from $[0, 0]$ to $[1, 1]$, the \ac{auc} is able 
to provide an aggregate measure of performance as the discrimination threshold 
varies.
We assume that a model whose predictions are 100\% correct has an \ac{auc} of 
1, as in this case.
\begin{figure}[!htb]
	\centering
	\includegraphics[width=.5\textwidth]{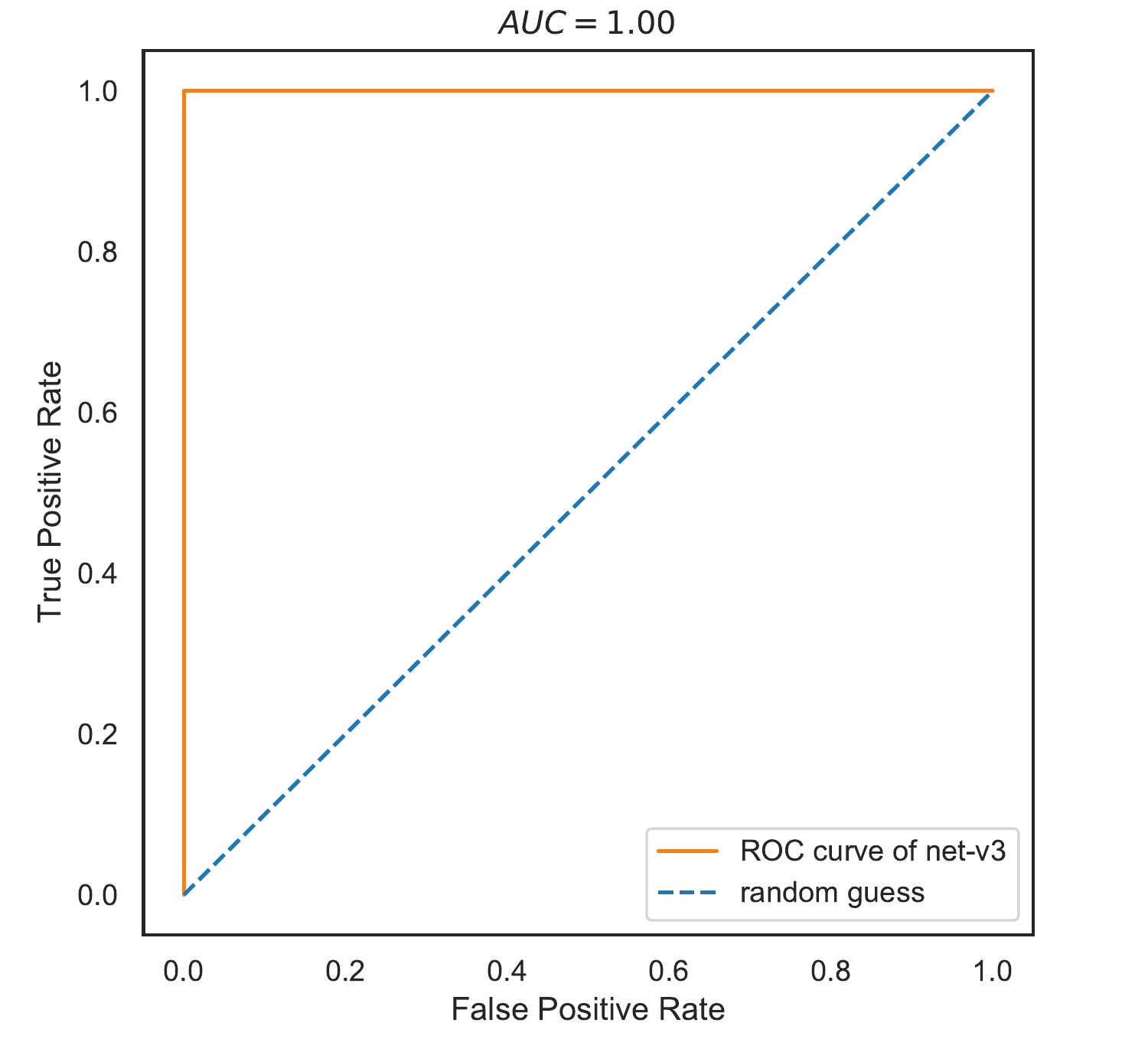}%
	\caption[Evaluation of the ROC of \texttt{net-v3}.]{Visualisation of the 
		\ac{roc} curve of \texttt{net-v3}.}
	\label{fig:net-v3auc}
\end{figure}

Also for this task, it is interesting to analyse the type of communication protocol 
inferred by the network, also comparing it with the one implemented by the 
manual controller. 
In Figure \ref{fig:net-v3commcolour} are shown, for a simulation run, first the 
messages transmitted by the agents over time, through a colour bar whose 
spectrum is included in the range [0, 1], i.e.,the maximum and minimum value of 
communication transmitted, and then the colour assumed by the robot in a 
certain time step, for both the manual and the learned controllers.
The extreme robots always transmit the same message using both controllers, 
while using the learned one, the central robots seems to transmit the same value, 
i.e.,1, but despite this they are able to achieve their goal in only two time steps, 
one less than with the manual. This behaviour cannot scale to a number of robot 
higher then $5$. For instance, in case of $5$ agents, in the first time step, 
\texttt{myt2} and \texttt{myt4} receive respectively the messages $(0, 1)$ and $(1, 
0)$, so they immediately know their position with respect to the central robot, 
which in turn knows its position since it receives $(1, 1)$. Then they communicate 
their message and in the following time step all the agents have coloured 
themselves in the right way, achieving the goal.
Consequently if the number of robots is greater, the central robots are not able to 
localise themselves. 
\begin{figure}[!htb]
	\begin{subfigure}[h]{\textwidth}
		\centering
		\includegraphics[width=.6\textwidth]{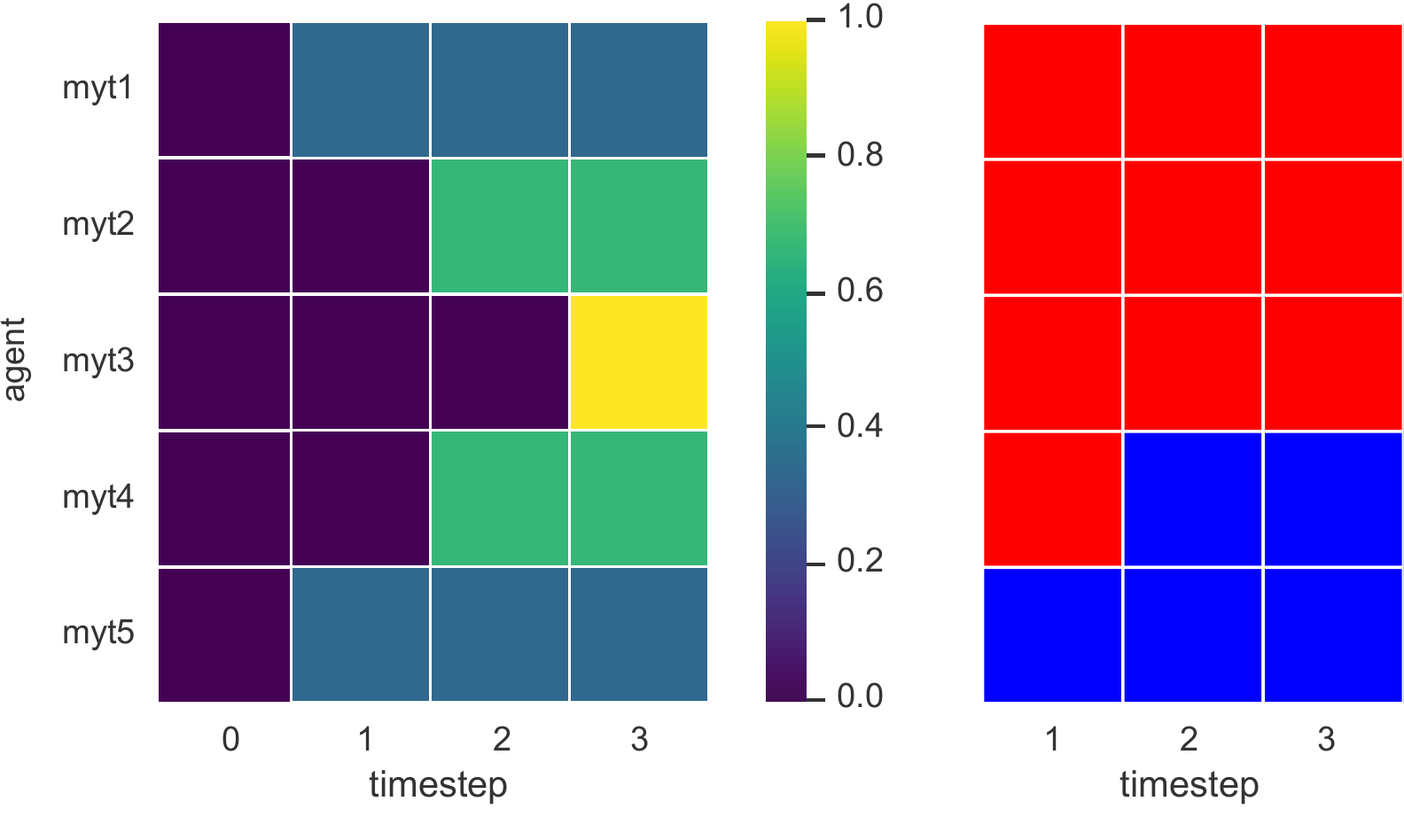}
		\caption{Communication and colour decided using the manual controller.}
	\end{subfigure}
	\hspace*{\fill}%          % empty line absolutely necessary!
	\vspace*{8pt}%  
	\hspace*{\fill}%  
	\begin{subfigure}[h]{\textwidth}
		\centering			
		\includegraphics[width=.6\textwidth]{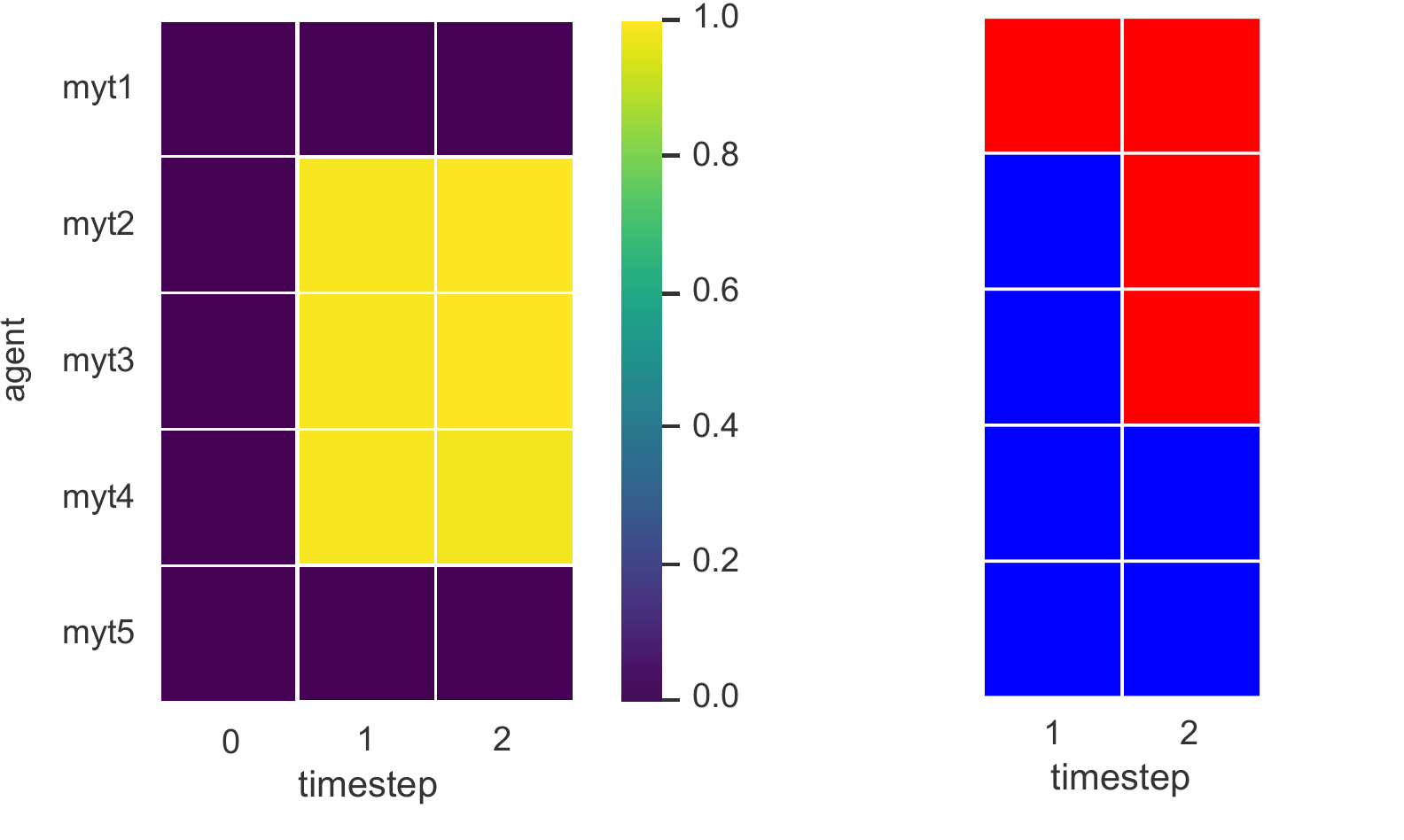}
		\caption{Communication and colour decided using the learned controller.}
	\end{subfigure}
	\caption[Evaluation of the communication learned by 
	\texttt{net-v3}.]{Visualisation of the communication transmitted by each 
		robot over time and the colour decided by the controller learned from 
		\texttt{net-v3}.}	
	\label{fig:net-v3commcolour}
\end{figure}
\vspace{0.5cm}

In Figure \ref{fig:net-v3error} is presented a useful metric that measures the 
amount of wrong expected colours, on the y-axis, over time, averaged for all the 
robots among the simulation runs. In particular, at each time step we count the 
number of agents that have the wrong colour and divide it by the number of 
simulations.
The mean value is shown as well as the bands representing minus and plus the 
standard deviation.
On average, the amount of correct colours is higher for the manual controller 
than the learned one. 
\begin{figure}[!htb]
	\centering
	\includegraphics[width=.5\textwidth]{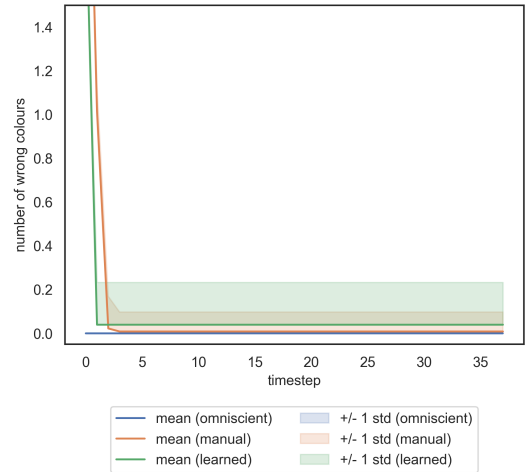}%
	\caption[Evaluation of \texttt{net-v3} amount of wrong expected 
	colours.]{Comparison of performance in terms of amount of wrong expected 
		colours obtained using the controller learned from \texttt{net-v3}.}
	\label{fig:net-v3error}
	\vspace{-0.5cm}
\end{figure}

\paragraph*{Results using 8 agents}
Following are presented the results of the experiments performed using $8$ 
agents. 
\begin{figure}[H]
	\centering
	\includegraphics[width=.5\textwidth]{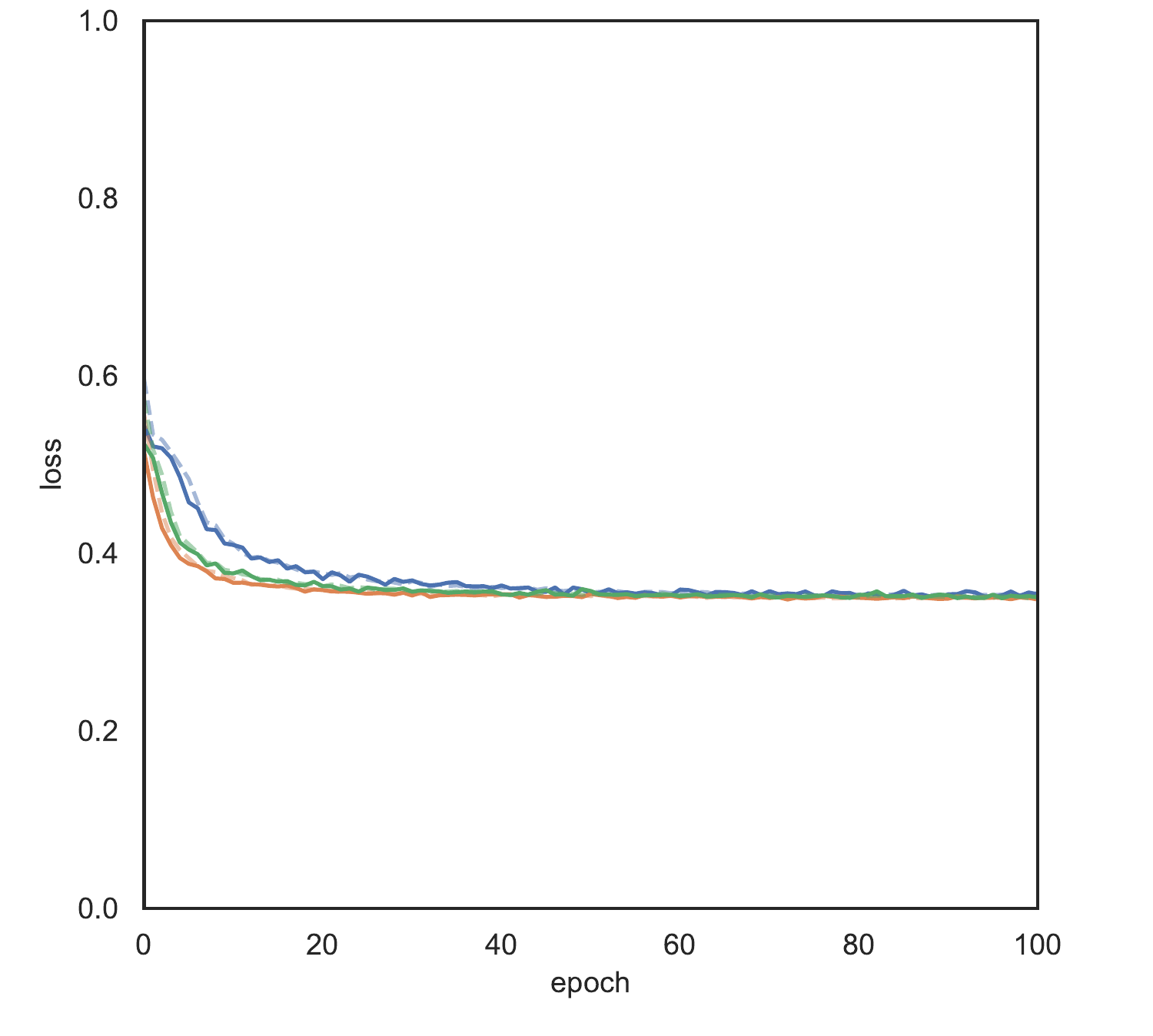}
	\caption[Comparison of the losses of the models that use $8$ 
	agents.]{Comparison of the losses of the models that use $8$ agents as 
	the gap varies.}
	\label{fig:commlossn8t2}
\end{figure}

\noindent
In Figure \ref{fig:commlossn8t2} are summarised the performance in terms of 
train and validation losses, by varying the average gap, as before the blue, orange 
and green lines represent respectively gaps of $8$cm, $20$cm and 
variable. 
From a first observation we see that the losses are higher than before, this is 
because a great number of agents reduce the performance, since more time steps 
are necessary to achieve the goal.

From the \ac{roc} curve of the model in Figure \ref{fig:net-v6auc} we observe 
that this time the \ac{auc} is decreased from 1 to 0.87 with respect to the 
previous model examined. 
%, while the accuracy is increased from $66\%$ up to $73\%$, 
\begin{figure}[!htb]
	\centering
	\includegraphics[width=.5\textwidth]{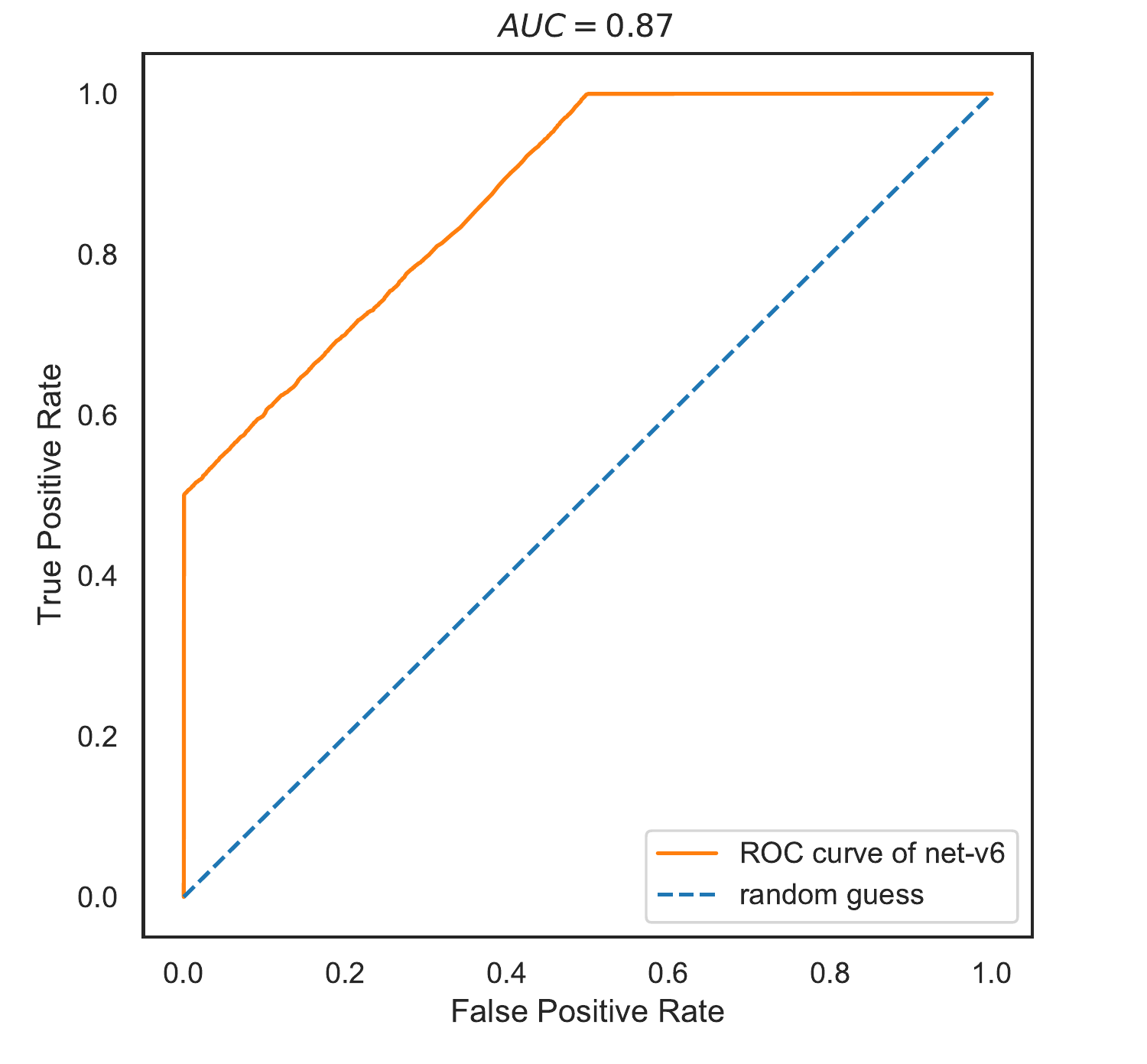}%
	\caption[Evaluation of the \ac{roc} of \texttt{net-v6}.]{Visualisation of the 
		\ac{roc} curve of \texttt{net-v6} based on \ac{bce} Loss.}
	\label{fig:net-v6auc}
\end{figure}

\bigskip
In Figure \ref{fig:net-v6error} is presented the measure of the amount of wrong 
expected colours, on the y-axis, over time, averaged on all robots among all the 
simulation runs. 
\begin{figure}[!htb]
	\centering
	\includegraphics[width=.5\textwidth]{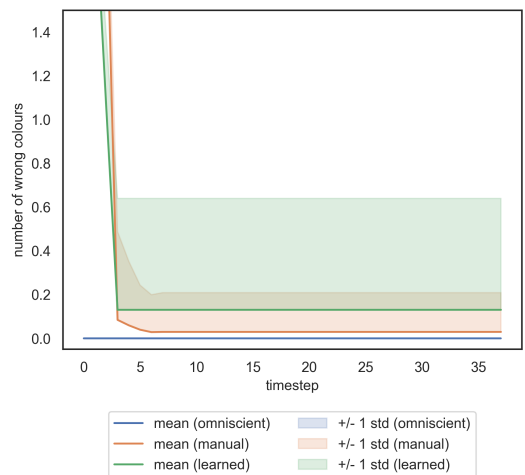}%
	\caption[Evaluation of \texttt{net-v6} amount of wrong expected 
	colours.]{Comparison of performance in terms of amount of wrong expected 
		colours obtained using the controller learned from \texttt{net-v6}.}
	\label{fig:net-v6error}
\end{figure}

\noindent
As expected, the number of colours correctly predicted by the learned controller 
is lower than before, while the manual controller still has the same performance.  

Finally, we visualise, in Figure \ref{fig:net-v6commcolour}, the communication 
protocol inferred by the network and the one chosen by the manual controller, as 
well as the colour assumed by each robot, we immediately see a difference in both 
the figures.
This time it is possible to hypothesize the policy adopted by the network to send 
messages: as before, the extreme robots always send the same message but this 
time the central ones communicate a value interpreted as a reward. In detail, 
starting from the edges, the value 0 is transmitted, then, once the robot that 
follows or precedes receives this value in turn communicates 0, until all the agents 
have received the message and therefore have clear their positional order.
This type of reward acts in such a way as to colour as desired the following robot, 
for those in the first half, or the one that precedes, for those in the second, 
communicating 0 respectively when there is a red agent behind it or when in front 
there is a blue one. In this way the control is able to stabilise and achieve its goal.
\begin{figure}[!htb]
	\begin{subfigure}[h]{\textwidth}
		\centering
		\includegraphics[width=.55\textwidth]{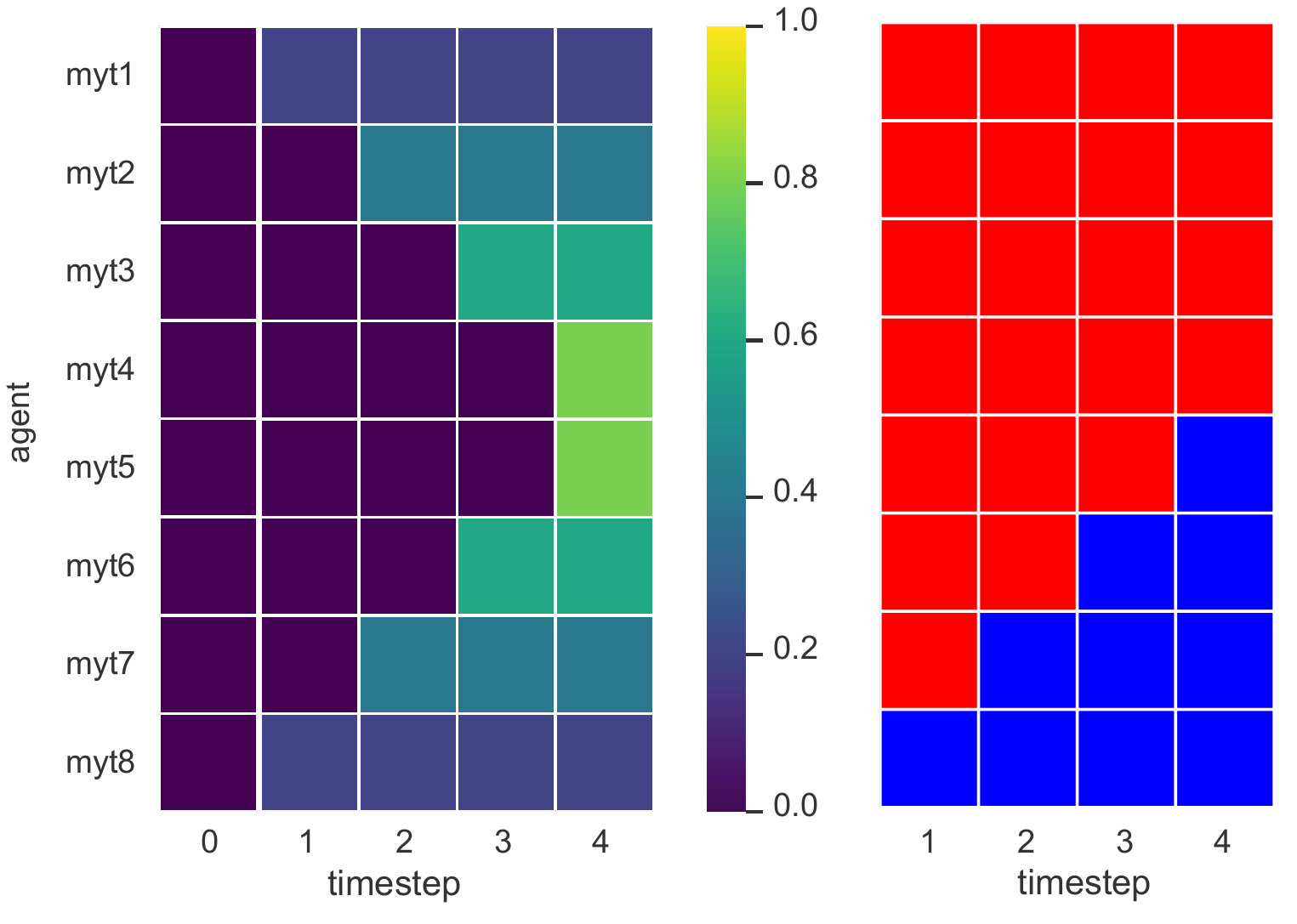}
		\caption{Communication and colour decided using the manual controller.}
	\end{subfigure}
	\hspace*{\fill}%          % empty line absolutely necessary!
	\vspace*{8pt}%  
	\hspace*{\fill}%  
	\begin{subfigure}[h]{\textwidth}
		\centering			
		\includegraphics[width=.55\textwidth]{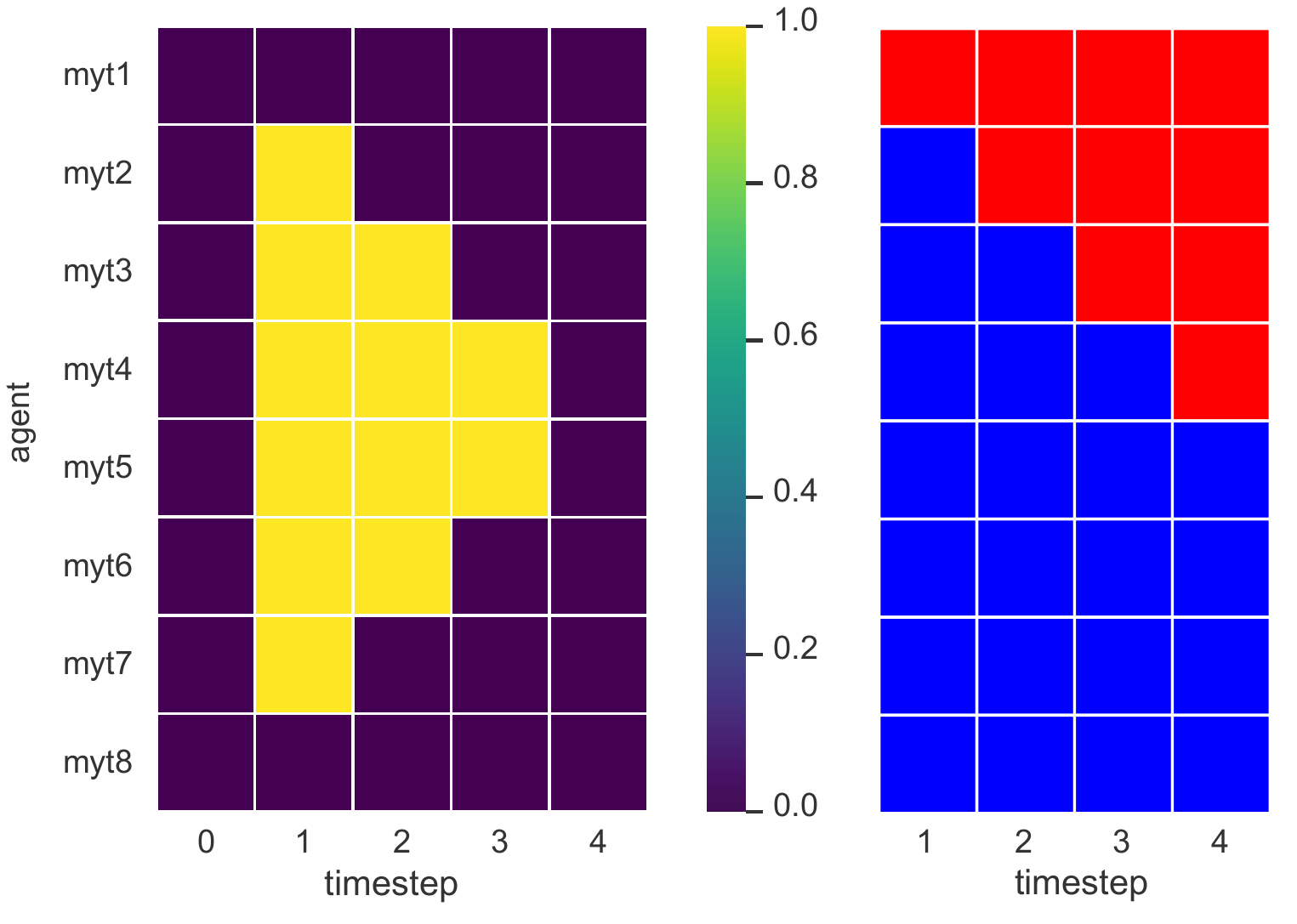}
		\caption{Communication and colour decided using the learned controller.}
	\end{subfigure}
	\caption[Evaluation of the communication learned by 
	\texttt{net-v6}.]{Visualisation of the communication transmitted by each 
		robot over time and the colour decided by the controller learned from 
		\texttt{net-v6}.}	
	\label{fig:net-v6commcolour}	
	\vspace{-0.5cm}
\end{figure}

\paragraph*{Results using variable agents}
We conclude the experiment on this task by presenting the results obtained using 
variable number of agents. In Figure \ref{fig:commlossnvart2} are summarised the 
performance in terms of loss, as before we used blue, orange and green lines to 
represent respectively average gaps of $8$cm, $13$cm and variable. 
As before we observe that in general the losses are increased respect the first 
experiment that use a smaller number of agents, since a higher amount of robots 
reduce the performance of the models, instead it is decreased respect the last 
experiment examined.
\begin{figure}[!htb]
	\centering
	\includegraphics[width=.47\textwidth]{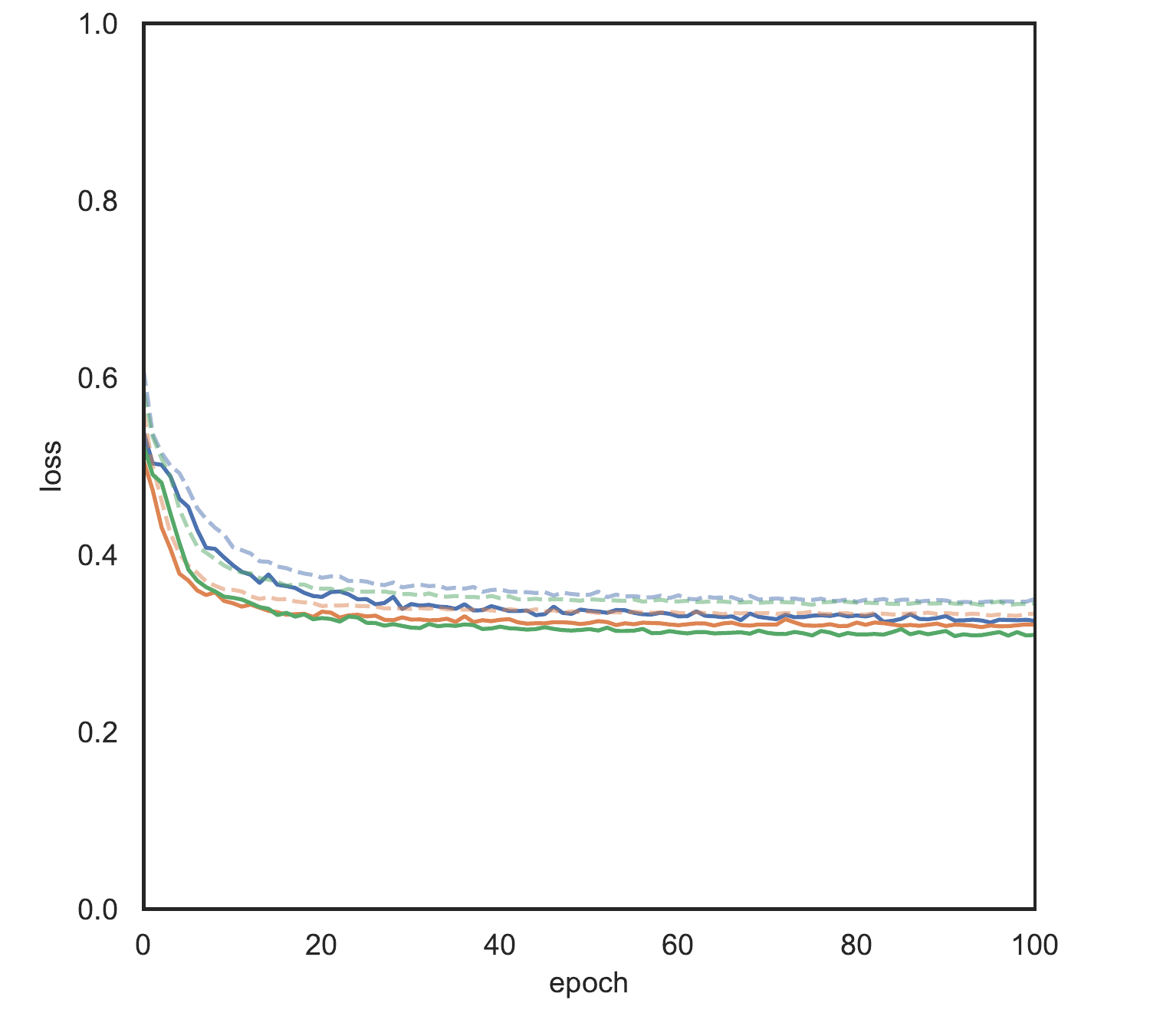}
	\caption[Comparison of the losses of the models that use variable 
	agents.]{Comparison of the losses of the models that use variable agents 
	as the gap varies.}
	\label{fig:commlossnvart2}
\end{figure}

From the \ac{roc} curve of the model in Figure \ref{fig:net-v9auc}  we observe 
that this time the \ac{auc} is a bit increased respect the previous experiment, 
going from $0.87$ up to $0.89$, but still worse than the first one.
\begin{figure}[!htb]
	\centering
	\includegraphics[width=.47\textwidth]{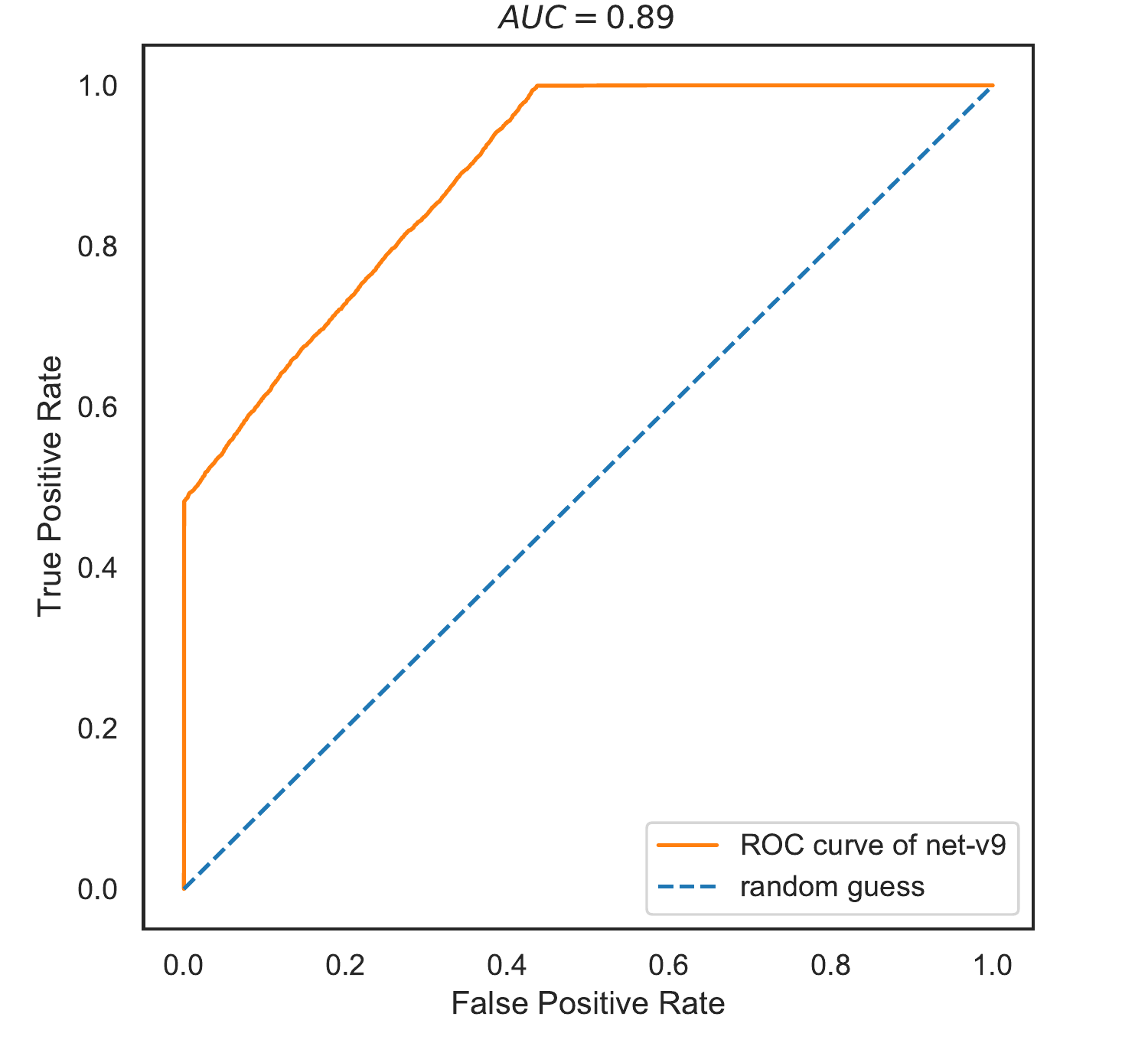}%
	\caption[Evaluation of the \ac{roc} of \texttt{net-v9}.]{Visualisation of the 
		\ac{roc} curve of \texttt{net-v9}.}
	\label{fig:net-v9auc}
\end{figure}

This time we visualise in Figures \ref{fig:net-v9commcolour} and 
\ref{fig:net-v9commcolour2} two examples of communication protocol inferred 
by the network and the one chosen by the manual controller, as well as the colour 
assumed by each robot.
The first visualisation is obtained from a simulation with 10 agents. 
As before, the policy adopted by the network to send messages is very similar to 
the previous one. 
\begin{figure}[!htb]
	\begin{subfigure}[h]{\textwidth}
		\centering
		\includegraphics[width=.6\textwidth]{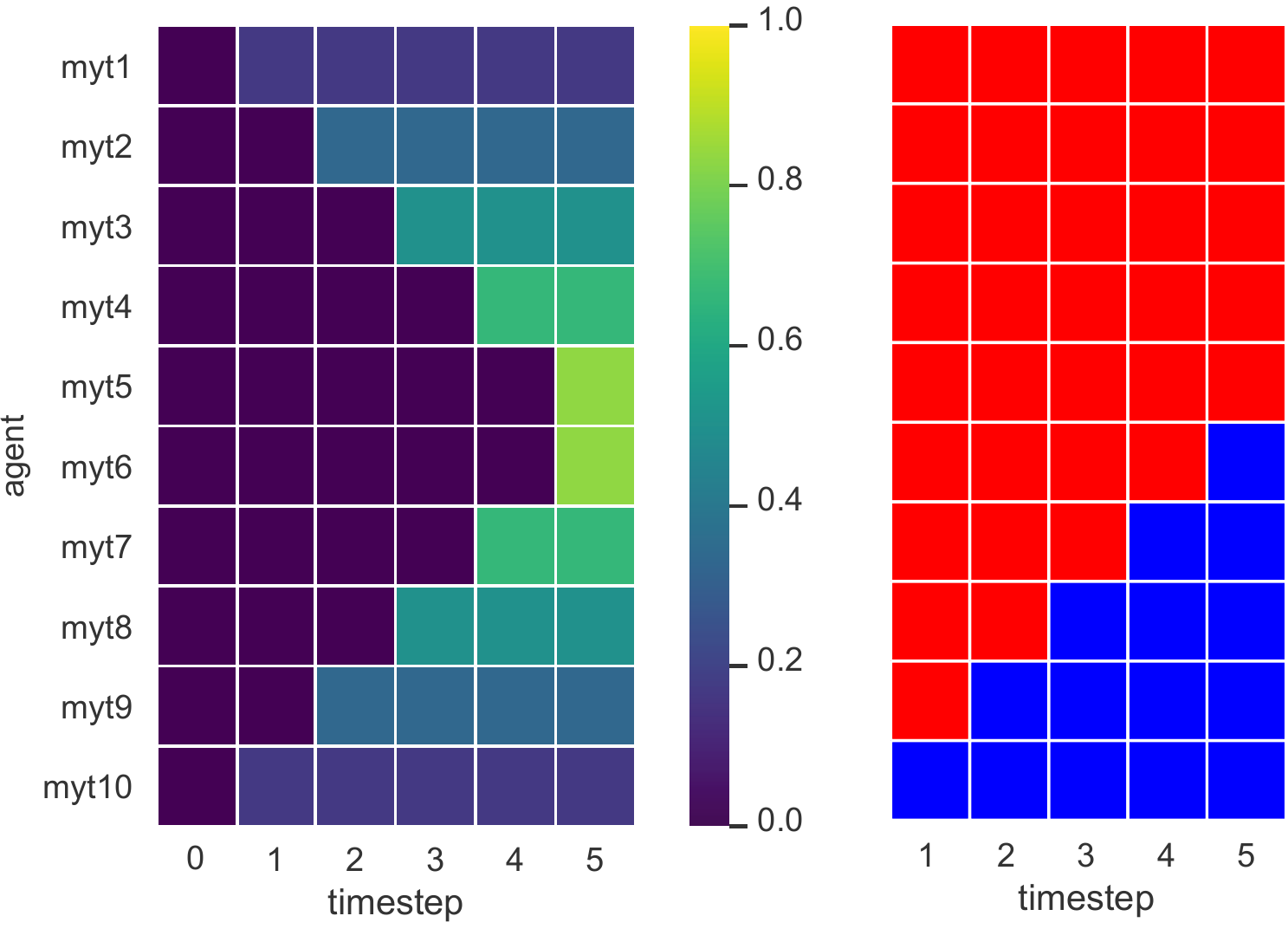}
		\caption{Communication and colour decided using the manual controller.}
	\end{subfigure}
	\hspace*{\fill}%          % empty line absolutely necessary!
	\vspace*{8pt}%  
	\hspace*{\fill}%  
	\begin{subfigure}[h]{\textwidth}
		\centering			
		\includegraphics[width=.6\textwidth]{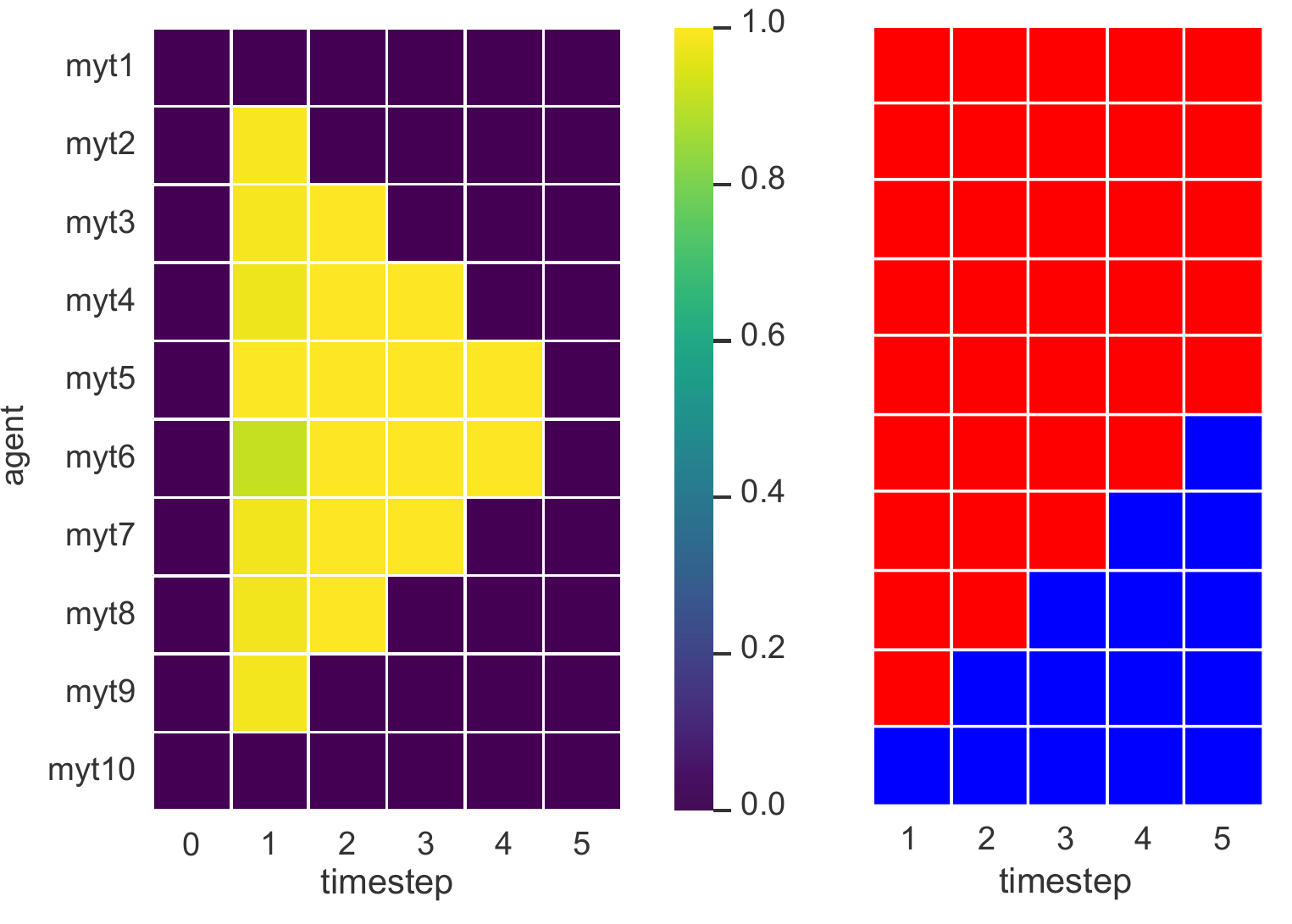}
		\caption{Communication and colour decided using the learned controller.}
	\end{subfigure}
	\caption[Evaluation of the communication learned by 
	\texttt{net-v9}.]{Visualisation of the communication transmitted by each 
		robot over time and the colour decided by the controller learned from 
		\texttt{net-v9}.}	
	\label{fig:net-v9commcolour}
\end{figure}
The robots at the edges start to transmit the value 0. Then, once 
the next robots receive the message in turn they communicates 0 or a value very 
close to it, until all the agents have received and sent the the message and have 
finally clear their positional order.
This time the manual and learned controllers achieve the goal in the same 
number of time steps.
The second visualisation is obtained from a simulation with 6 agents. 
The policy adopted by the network is the same as before, this time is even more 
efficient than the protocol used from the manual controller. In fact, in 2 time 
steps, one less then the other controller, the model is able to achieve the goal.

Finally, in Figure \ref{fig:net-v9error} is presented the measure of the amount of 
wrong expected colours, on the y-axis, over time, averaged on all robots among 
all the simulation runs. 

\begin{figure}[!htb]
	\begin{subfigure}[h]{\textwidth}
		\centering
		\includegraphics[width=.5\textwidth]{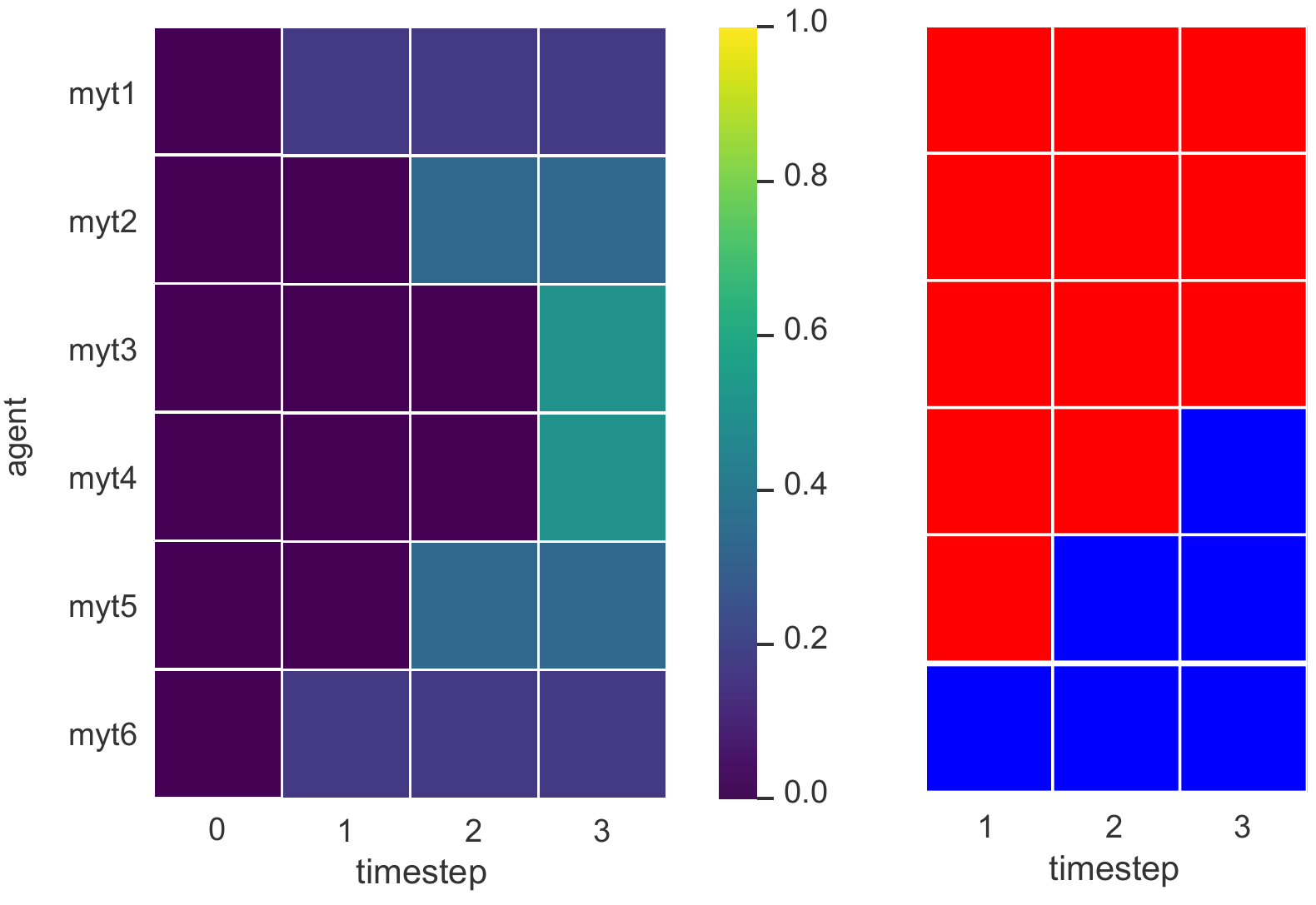}
		\caption{Communication and colour decided using the manual controller.}
	\end{subfigure}
	\hspace*{\fill}%          % empty line absolutely necessary!
	\vspace*{8pt}%  
	\hspace*{\fill}%  
	\begin{subfigure}[h]{\textwidth}
		\centering			
		\includegraphics[width=.5\textwidth]{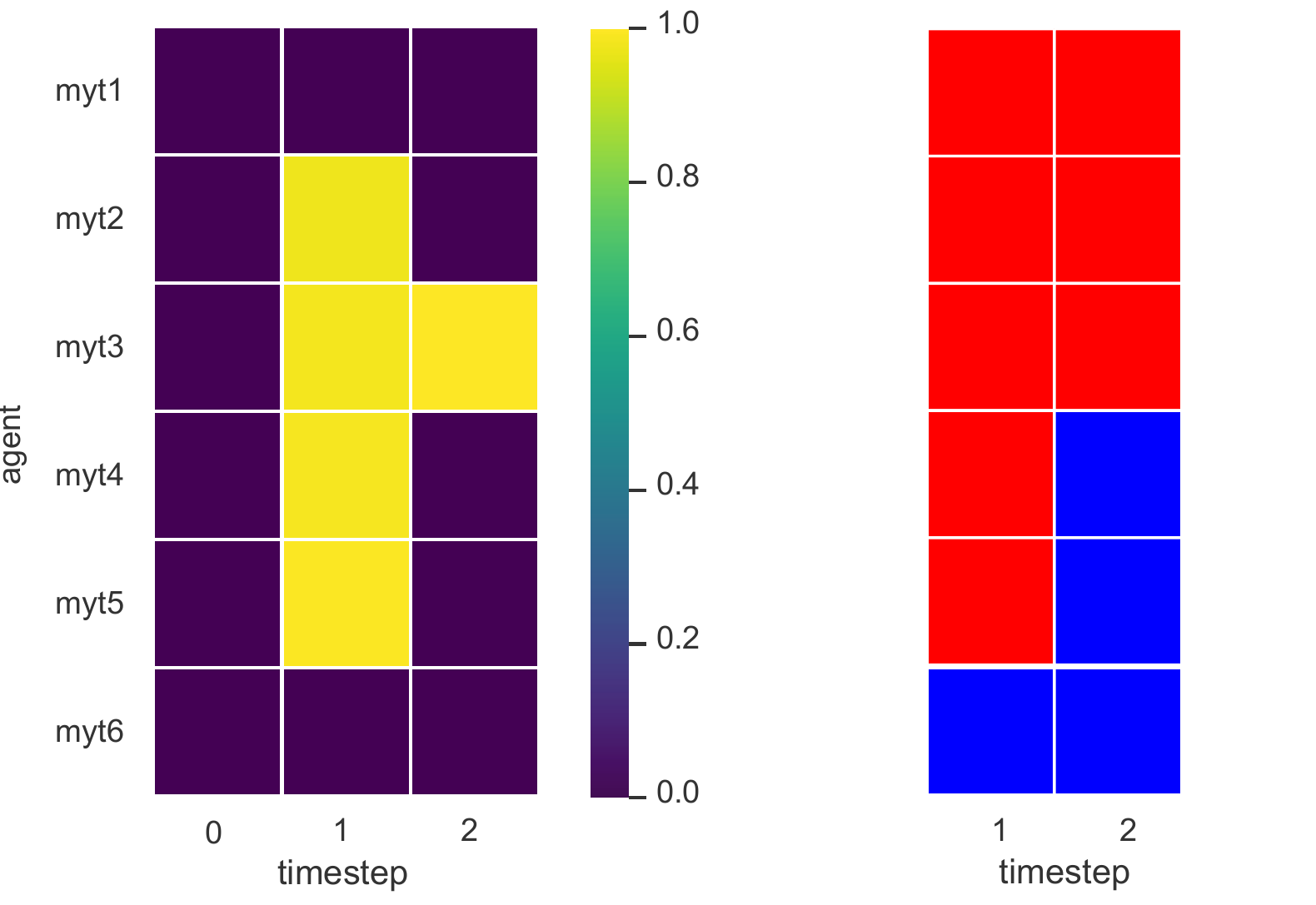}
		\caption{Communication and colour decided using the learned controller.}
	\end{subfigure}
	\caption[Evaluation of the communication learned by 
	\texttt{net-v9}.]{Visualisation of the communication transmitted by each 
		robot over time and the colour decided by the controller learned from 
		\texttt{net-v9}.}	
	\label{fig:net-v9commcolour2}
\end{figure}

\begin{figure}[H]
\centering
\includegraphics[width=.5\textwidth]{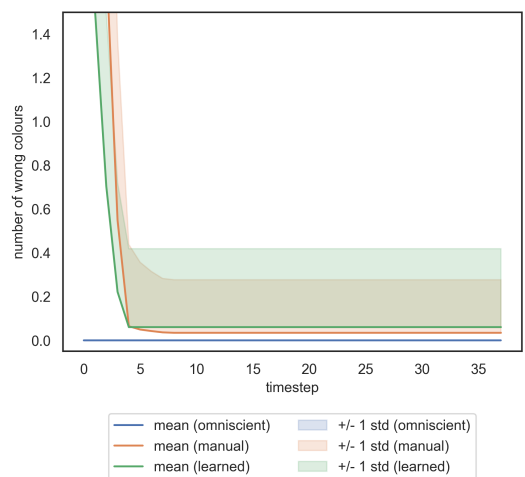}%
\caption[Evaluation of \texttt{net-v9} amount of wrong expected 
colours.]{Comparison of performance in terms of amount of wrong expected 
	colours obtained using the controller learned from \texttt{net-v9}.}
\label{fig:net-v9error}
\end{figure}

\noindent
for this experiment the number of colours correctly predicted by the learned 
controller is a bit less than that obtained by the manual controller, and even if the 
variance for the model is higher the performance are acceptable.

\paragraph*{Summary}
To sum up, we show the losses of the trained models as the number of robots vary 
for each gap, in particular, in blue, orange and green we refer to the simulation 
with $5$, $8$ and variable agents. 
Unlike the previous task, here no clear differences are highlighted varying the 
gap. 
The performance obtained with a smaller number of agents are clearly superior, 
while the other two, obtained by increasing the amount of robots, are very similar 
and tend to move away from each other as the gap grows.
\begin{figure}[!htb]
	\begin{center}
		\begin{subfigure}[h]{0.32\textwidth}
			\includegraphics[width=\textwidth]{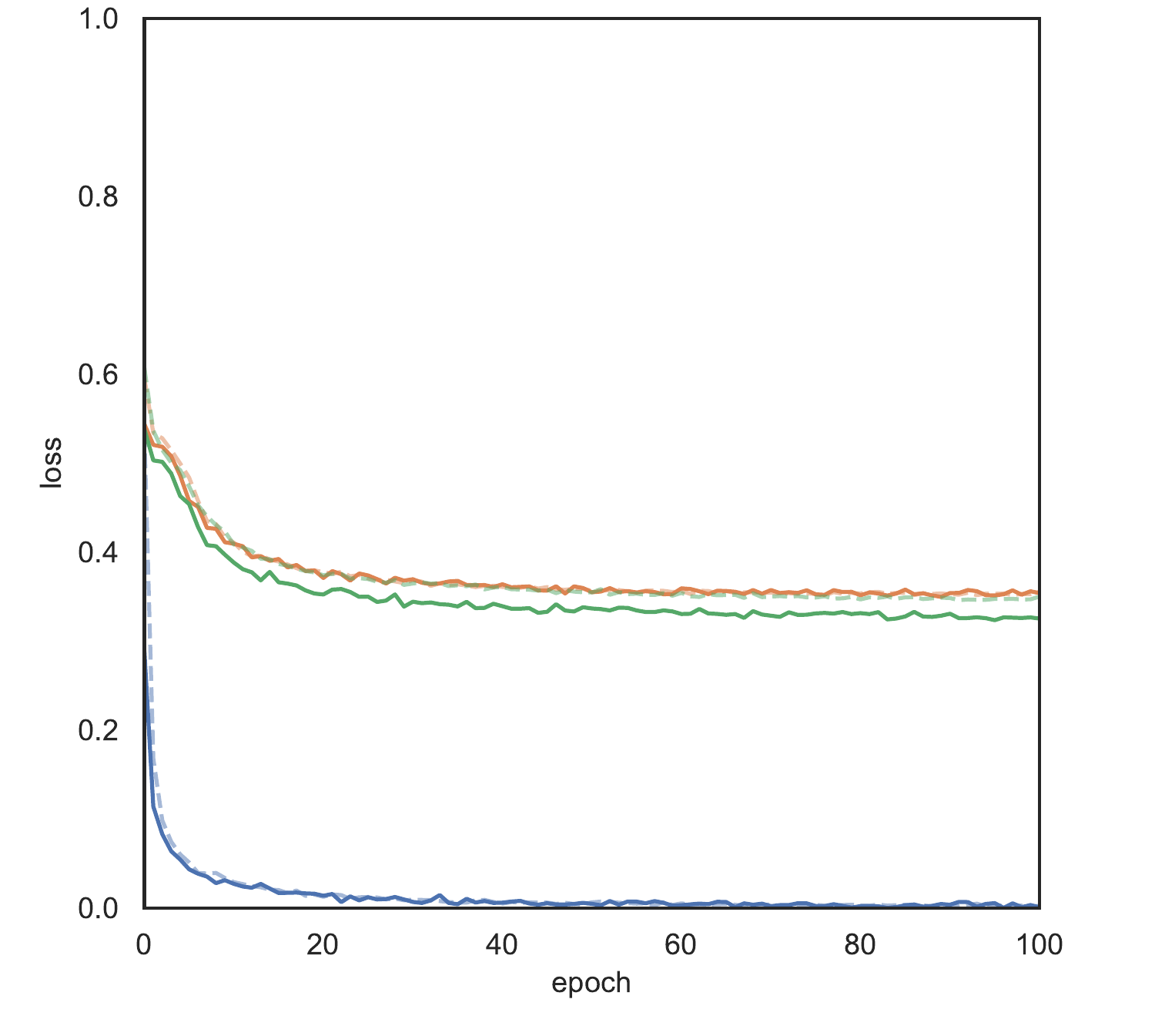}%
			\caption{\texttt{avg\_gap} of $8$cm.}
		\end{subfigure}
		\hfill
		\begin{subfigure}[h]{0.32\textwidth}
			\includegraphics[width=\textwidth]{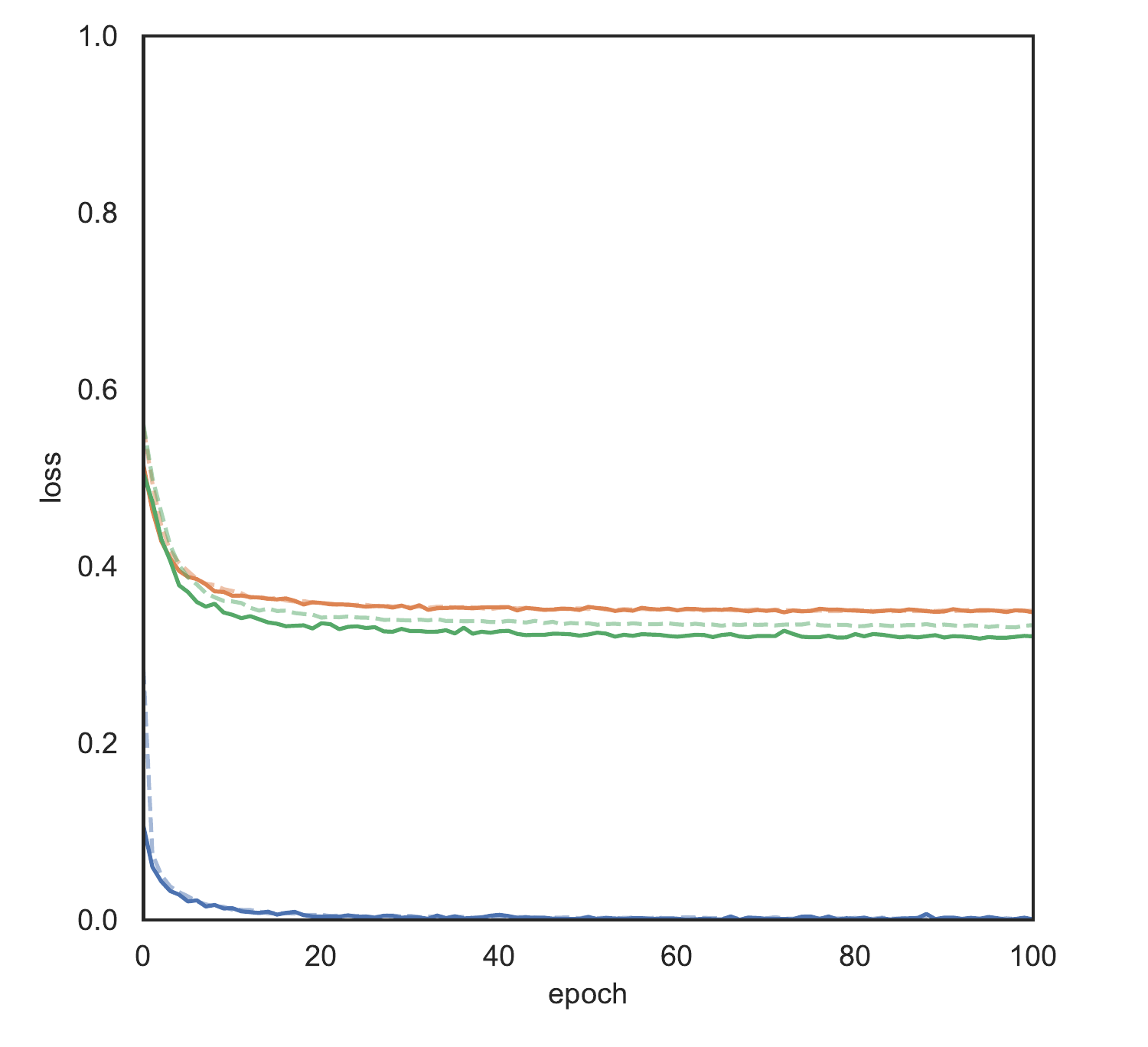}%
			\caption{\texttt{avg\_gap} of $20$cm.}
		\end{subfigure}
		\hfill
		\begin{subfigure}[h]{0.32\textwidth}
			\includegraphics[width=\textwidth]{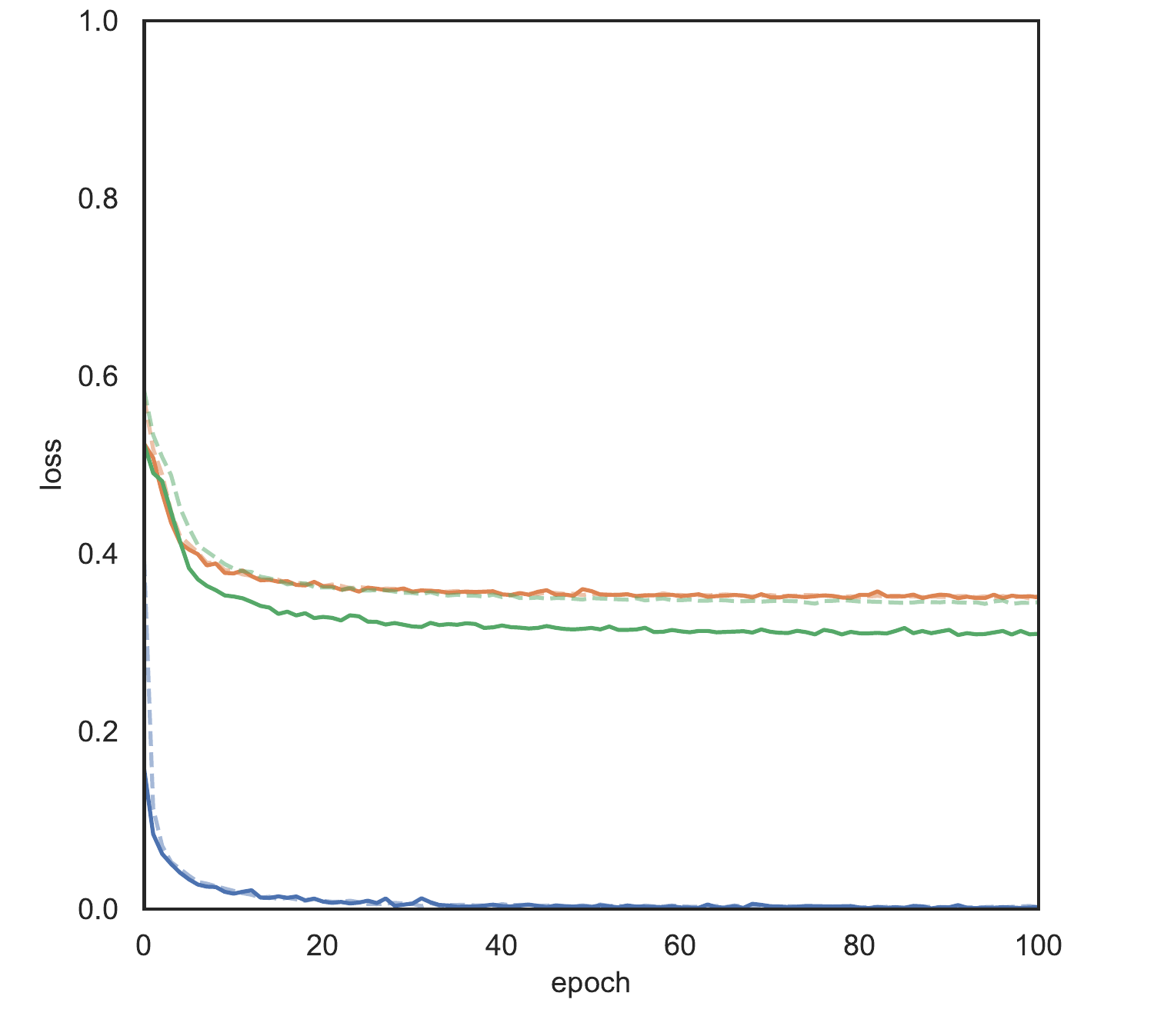}
			\caption{\texttt{avg\_gap} variable.}
		\end{subfigure}
	\end{center}
	\vspace{-0.5cm}
	\caption[Losses summary of the second task 
	(communication).]{Comparison of the losses of the model trained with 
	communication, by varying the number of agents for the three gaps.}
	\label{fig:commlosst2}
	\vspace{-0.5cm}
\end{figure}

\subsubsection{Experiment 2: increasing number of agents}
\label{subsubsec:task2-exp-comm-2}

The last group of experiments focuses on the scalability properties of a 
multi-agent system, showing the behaviour of the network trained using variable 
gaps and number of agents, applied on simulations with a higher number of 
robots, from 5 up to 50.

In Figure \ref{fig:errorcomm} is visualised, for 5 different experiments, the 
expected percentage of wrong colours over time, averaged for all the robots 
among the simulation runs. 
In all experiments, the number of errors in the simulation corresponds to 50\%, 
i.e.,half of the colours are wrong. This results are expected since the colours at the 
first time step are chosen randomly. As the time steps pass, the number of errors 
decreases at a constant speed, about two robots per time step. In general, 
$\frac{N}{2}$ time steps are required to achieve convergence, sometimes 
$\frac{N}{2} - 1$ when the number of agents is odd. 
Despite this, when using a number of robots between 5 and 10, on average 1\% of 
the colours are wrong in the final configuration. Increasing the amount of agents 
this value increases, reaching 10\% in the case of 50 robots.
Despite this, the performances are very promising and the network is able to scale 
well by increasing the number of robots.
\begin{figure}[!htb]
	\centering
	\includegraphics[width=.5\textwidth]{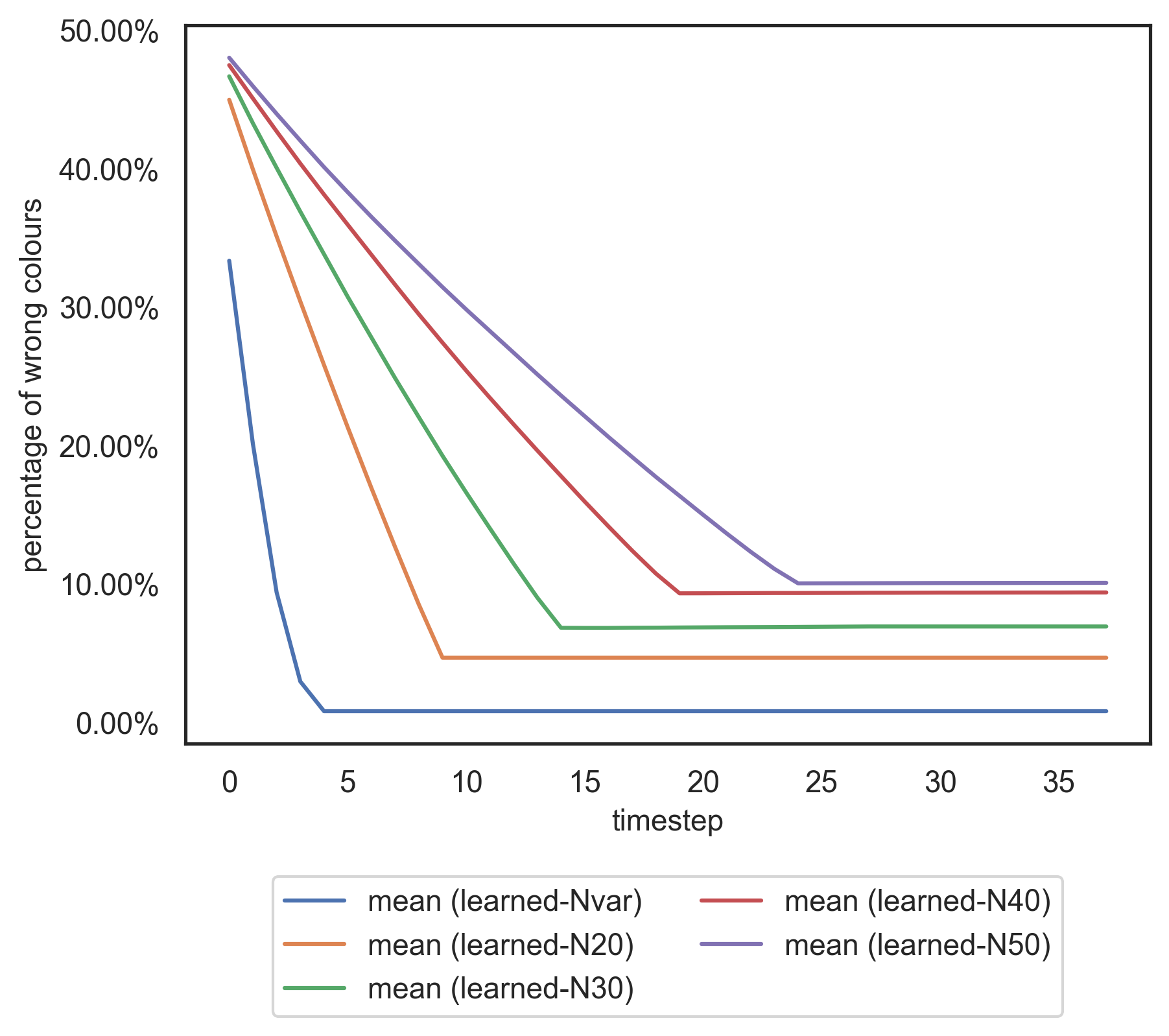}%
	\caption[Evaluation of distances from goal for a high number of 
	robots.]{Comparison of performance in terms of distances from goal obtained 
		on simulations with an increasing number of robots.}
	\label{fig:errorcomm}
\end{figure}

\subsubsection{Remarks}
\label{subsubsec:remarks-task2-comm}

In this section we have shown that, for some problems, communication is a 
necessity. 
Therefore, using a controller learned through imitation learning, which 
autonomously infers a communication protocol between the agents, it is possible 
to obtain excellent results and solve the task in many cases more effectively than 
the baseline.
Moreover, the network is able to scale with the increase of the number of robots, 
without worsening performance.

\acresetall
\chapter{Conclusion and perspectives}
\addcontentsline{toc}{chapter}{Conclusion and perspectives}
\label{chap:concl}
 
This chapter presents first, in Section \ref{sec:concl}, our concluding thoughts and 
finally suggestions for possible future research lines in Section \ref{sec:future}.

\section{Concluding thoughts}
\label{sec:concl}
Robotics research has dedicated extensive attention to cooperative multi-agent 
problems, proposing different approaches that allow the collaboration and 
communication of swarm of robots to achieve a common goal.

Our work makes a further step in direction of \ac{il} approaches, and aims to 
find feasible solutions to different multi-agent scenarios.

We explore two alternative tasks: distributing the robots in space such that they 
stand at equal distance from each other, and, assuming that the agents are 
divided into two sets, colouring the robots in space depending on their group 
membership.
To solve these examples of cooperative tasks, we propose two main models: both 
learn decentralised controllers, via observation of the demonstrations of a 
centralised controller, by training end-to-end \acp{nn} in which is possible to
introduce a communication protocol, is inferred by the network, consisting in an 
explicit exchange of messages between the robots.

For the first task, we build two models: one that at each time step takes as input 
an array containing the response values of the sensors for each robot and 
produces as output the speed of the agent, the other one that as input takes also 
the messages received in the previous time step, communicated by the nearest 
agents (on the left and on the right), and produces as output, in addition to the 
control, the communication, i.e.,the message transmitted by the robot to the two 
neighbours. 

For the second task, we implement a single model, similar to the previous, but this 
time ignoring the sensors readings. Thus, the network, at each time step, takes as 
input for each robot only the message received in the previous time step, 
communicated by the nearest agents, and returns as output an array of 2 floats, 
the first one is the probability of the agent top LED to be blue and the second is 
the communication, i.e.,the message to be transmitted by the robot.

Throughout the experiments, in addition to comparing the approach with or 
without communication, we also analyse the effects of varying the inputs of the 
networks, the average gap and the number of agents chosen. 

First of all, we examine the behaviour of the learned controllers by varying the 
input of the network — either \texttt{prox\_values}, \texttt{prox\_comm} or 
\texttt{all\_sensors}. Since at each input corresponds a different range of the 
proximity sensors, the performance mainly depends on the average gap chosen: 
for \texttt{prox\_values} the best results are obtained by using small gaps, less 
than 12cm, while with \texttt{prox\_comm} using larger ones. In general, 
\texttt{all\_sensors} input is able to work with arbitrary gaps, obtaining more 
stable behaviours and achieving satisfactory results in both approaches.

We continue analysing the performance by varying the average gap between the 
agents — either fixed to a certain value or variable in the range $[5, 24]$. 
Similarly to the previously mentioned experiment, the results are heavily 
influenced by the input used this time too. Considering \texttt{all\_sensors} input, 
the networks have excellent performance with any gap, from a smaller to a larger 
one, even with a variable one. Unlike in the first task, in the second one, no clear 
differences emerge varying the gap. 

Then, we proceed to compare the behaviour of the models varying the number of 
agents — either fixed among the simulation runs or variable in the range $[5, 
10]$. The main reason of these experiments is to verify the robustness of the 
models and proving that it is possible to train networks that handle a variable 
number of agents.
The results show, as expected, that in both approaches it is easier to obtain a 
correct behaviour of the controller using a small number of agents.

Finally, since multi-agent systems present interesting scalability challenges, we 
focus on the behaviour of a network that takes as input \texttt{all\_sensors}, 
variable gaps and number of agents applied on simulations with a higher number 
of robots, from 5 up to 50.
In the first task, a greater number of robots implies a slowdown in reaching the 
correct positions in both approaches, even when using an expert controller. 
In fact, the complexity grows rapidly as the number of robots increases and it is 
common for biases to add up and for the error to become more significant.
In the second scenario, regardless the number of agents, the goal is reached at 
constant speed: in general, $\frac{N}{2}$ time steps, sometimes $\frac{N}{2} - 1$, 
are necessary to reach the correct final configuration. Therefore the network is 
able to scale with the increase of the number of robots, without worsening 
performance.

In conclusion, using a distributed controller learned by imitating an expert, the 
first task obtains performance more or less comparable to those reached with 
a manual controller. Instead, applying a communication strategy improves the 
performance of the distributed model, letting it decide which actions to take 
almost as precisely and quickly as the expert controller.
The second task shows that it is possible to let the network autonomously infer a 
communication protocol, obtaining excellent results and solving the task more 
effectively, in many cases, than the baseline.

\section{Future works}
\label{sec:future}

We demonstrated the effectiveness of our method in a couple of simulated 
scenarios, so a following experiment could be implemented to prove the power of 
our model in the real world.
Another possible expansion of our work could be the application of our models to 
new problems and scenarios, such as colouring the robots using different criteria 
and supporting their repositioning in the row.
In this study we focus on relatively simple environments, indeed an interesting 
extension of this approach would be consider more realistic situations, moving to 
multi-dimensional environments, first \ac{2d} and then \ac{3d}, working with 
drones rather than differential drive robots.

\endgroup

\appendix %optional, use only if you have an appendix

%\begingroup
%\chapter{List of abbreviations}
%\label{chp:abbreviations}
%\let\clearpage\relax
%\vspace*{-90pt}
%\printglossary[title= ]
%\endgroup

\begingroup
\chapter{List of Acronyms}
\label{chp:acronyms}
\let\clearpage\relax
\acuseall
\printacronyms[heading=none]
%\glsaddall  % comment to hide unused acronyms
% \vspace*{-120pt}
% \printglossary[type=acronym]
\endgroup

\backmatter

\nocite{*}
\bibliography{biblio}
%\cleardoublepage
%\theindex %optional, use only if you have an index, must use
%\makeindex in the preamble

\end{document}